\documentclass[10pt,journal,cspaper,compsoc]{IEEEtran}
\usepackage[pdfborder=false,colorlinks,bookmarksnumbered,bookmarksopen,linkcolor=black,citecolor=black,urlcolor=magenta]{hyperref}
\usepackage[tight,footnotesize]{subfigure}
\usepackage{rotating}
\usepackage{graphicx}
\usepackage{amsmath,amssymb} % define this before the line numbering.
\usepackage{epsfig}
\usepackage{array}
\usepackage{multirow}
\usepackage{colortbl}
\usepackage{amsfonts}
\usepackage{pifont}
\usepackage{xspace}
\usepackage{footnote}
\usepackage{etoolbox}
\usepackage{cite}
\usepackage{overpic}
\usepackage{epstopdf}

\def\ie{{\em i.e.}}
\def\eg{{\em e.g.}}
\def\etal{{\em et al.}}

\newcommand{\Figref}[1]{Fig.~\ref{#1}}
\newcommand{\tabref}[1]{Fig.~\ref{#1}}

\newcommand{\myPara}[1]{\vspace{.05in}\noindent\textbf{#1.}}
\newcommand{\secref}[1]{Sec.~\ref{#1}}

\definecolor{gray1}{rgb}{.8,.8,.8}
\newcommand{\layer}{level}
\newcommand{\layers}{layers}
\newcommand{\hzhide}[1]{}
\newcommand{\rforest}{Random Forest}
\graphicspath{ {fig/}{fig/qual_comp/}{fig/failure/}{fig/Authors/}}

\begin{document}

\title{Salient Object Detection: A Discriminative Regional Feature Integration Approach}

\author{Huaizu Jiang,~Zejian Yuan,~Ming-Ming Cheng,
    Yihong Gong, \\
    Nanning Zheng, and Jingdong Wang %  \IEEEmembership{Member,~IEEE}% <-this % stops a space
    \IEEEcompsocitemizethanks{
        \IEEEcompsocthanksitem H. Jiang, Z. Yuan, Y. Gong and N. Zheng, are with Xi'an Jiaotong University.
        % \IEEEcompsocthanksitem H. Jiang, Z. Yuan, and N. Zheng, are with Xi'an Jiaotong University.
            M.M. Cheng is with Oxford University.
            J. Wang is with Microsoft Research Asia.
        \IEEEcompsocthanksitem A preliminary version of this work appeared at CVPR \cite{jiang2013salient}.
        \IEEEcompsocthanksitem Project website \href{http://jianghz.com/drfi}{jianghz.com/drfi}.
    }% <-this % stops a space
    \thanks{}
}

% \markboth{IEEE TRANSACTIONS ON PATTERN ANALYSIS AND MACHINE INTELLIGENCE,~Vol.XXX, No.XXX, XXXXX~2010}%
% {Shell \MakeLowercase{\textit{{\em et al.}}}: xx IEEE TRANSACTIONS ON PATTERN ANALYSIS AND MACHINE INTELLIGENCE}

\IEEEcompsoctitleabstractindextext{%
\begin{abstract}
Salient object detection has been attracting a lot of interest,
and recently various heuristic computational models have been designed.
In this paper,
we formulate saliency map computation as a regression problem.
Our method, which is based on multi-\layer~image segmentation,
utilizes the supervised learning approach
to map the regional feature vector to a saliency score.
Saliency scores across multiple \layers~are finally fused to produce the saliency map.
The contributions lie in two-fold.
One is that we propose a discriminate regional feature integration approach for salient object detection. Compared with existing heuristic models, our proposed method is able to automatically integrate high-dimensional regional saliency features and choose discriminative ones. The other is that by investigating standard generic region properties as well as two widely studied concepts for salient object detection, \ie, regional contrast and backgroundness, our approach significantly outperforms state-of-the-art methods on six benchmark datasets. Meanwhile, we demonstrate that our method runs as fast as most existing algorithms.
%, taking around 0.4 seconds to produce a saliency map.
% \hzhide{we show our approach,
% which integrates the regional contrast,
% regional property and regional backgroundness descriptors
% together to form the master saliency map,
% is able to produce superior saliency maps to existing algorithms most of
% which combine saliency maps heuristically computed from different types of features.
% The other is that we introduce a new regional feature vector, backgroundness,
% to characterize the background,
% which can be regarded as a counterpart of the objectness descriptor~\cite{alexe2012measuring}.
% The performance evaluation on several popular benchmark data sets
% validates that our approach outperforms existing state-of-the-arts.}
\end{abstract}

% \begin{keywords}
% Salient object detection, visual attention, saliency map, discriminative learning
% \end{keywords}
}

\maketitle

\IEEEdisplaynotcompsoctitleabstractindextext

\IEEEpeerreviewmaketitle

\section{Introduction}

%\IEEEPARstart{V}{isual}
Visual saliency has been a fundamental problem
in neuroscience, psychology, neural systems,
and computer vision for a long time.
It is originally defined as a task of
predicting the eye-fixations on images~\cite{Itti98}. 
% Driven by applications, \eg, content-aware image resizing and photos visualization, it is recently extended to
% identifying a region containing the salient object.
Recently it is extended to
identifying a region~\cite{ma2003contrast,Tie07} containing the salient object, known as \emph{salient object detection} or \emph{salient region detection}.
\hzhide{ that is also the focus of this paper.}
Applications of salient object detection 
include object detection and recognition~\cite{KananC10, WaltherK06},
image compression~\cite{Itti04},
image cropping~\cite{MarchesottiCC09},
photo collage~\cite{GofermanTZ10, WangQSTS06}, dominant color
detection~\cite{WangZWWHL12, WangZZW12} and so on.

The study on human visual systems
suggests that
the saliency is related to
uniqueness, rarity and surprise
of a scene,
characterized by primitive features
like color, texture, shape, etc.
Recently a lot of efforts
have been made to
design various heuristic algorithms to
compute the saliency~\cite{AchantaHES09, BorjiI12, ChengZMHH11, GaoMV07, GofermanZT10, KleinF11, LiuYSWZTS11, LuZLX11, PerazziKPH12}.
Built upon the feature integration theory~\cite{Itti98, TreismanG80},
almost all the approaches compute conspicuity (feature) maps
from different saliency cues
and then combine them together
to form the final saliency map. Hand-crafted integration rules, however, are fragile and poor to generalize. For instance, in a recent survey~\cite{BorjiSI12b}, none of the algorithms can consistently outperforms others on the benchmark data sets. Though some learning-based salient object detection algorithms are proposed~\cite{LiuYSWZTS11, KhuwuthyakornRZ10, MehraniS10}, the potential of supervised learning is not deeply investigated.

In this paper,
we formulate salient object detection as a regression problem,
 learning a regressor
that directly maps the regional feature vector to a saliency score.
Our approach consists of three main steps.
The first one is multi-\layer~segmentation,
which decomposes the image to multiple segmentations.
% from a fine \layer to a coarse one.
Second,
we conduct a region saliency computation step
with a \rforest~regressor
that maps the regional features to a saliency score.
Last, a saliency map is computed
by fusing the saliency maps across multiple \layers~of segmentations.

The key contributions lie in the second step,
region saliency computation. Firstly,
unlike most existing algorithms
that compute saliency maps heuristically from various features
and combine them to get the saliency map,
which we call saliency integration,
we learn a \rforest~regressor
that directly maps the feature vector of each region to a saliency score,
which we call discriminative regional feature integration (DRFI).
This is a principle way in image
classification~\cite{DBLP:journals/ijcv/HoiemEH07},
but rarely studied in salient object detection.
%except a few light touches~\cite{KhuwuthyakornRZ10, MehraniS10}.
Secondly, by investigating standard generic region properties and two widely studied concepts in salient object detection, \ie, regional contrast and backgroundness, our proposed approach consistently outperforms state-of-the-art algorithms on all six benchmark data sets with large margins. Rather than heuristically hand-crafting special features, it turns out that
the learned regressor is able to
automatically integrate features and pick up discriminative ones for saliency. Even though the regressor is trained on a small set of images, it demonstrates good generalization ability to other data sets.

The rest of this paper is organized as follows.~\secref{sec:relatedWork} introduces related work and discusses their differences with our proposed method. The saliency computation framework is presented in~\secref{sec:imgSaliencyComputation}.~\secref{sec:RegionalFeatures} describes the regional saliency features adopted in this paper.~\secref{sec:Learning} presents the learning framework of our approach. Empirical analysis of our proposed method and comparisons with other algorithms are demonstrated in~\secref{sec:Experiment}. Finally,~\secref{sec:Conclusion} discusses and concludes this paper.

\section{Related work}
\label{sec:relatedWork}
Salient object detection, stemming from eye fixation prediction, aims to separate the entire salient object from the background. Since the pioneer work of Itti~\etal~\cite{Itti98}, it attracts more and more research interests in computer vision, driven by applications such as content-aware image resizing~\cite{MarchesottiCC09}, picture collage~\cite{WangQSTS06}, etc. In the following, we focus on salient object detection (segmentation) and briefly review existing algorithms. A comprehensive survey can be found from a recent work~\cite{BorjiSI12b}. A literature review of eye fixation prediction can be seen in~\cite{BorjiI12b}, which also includes some analysis on salient object detection. We simply divide existing algorithms into two categories: unsupervised and supervised, according to if the groundtruth annotations of salient objects are adopted.

\myPara{Unsupervised approaches} Most salient object detection algorithms characterize the uniqueness of a scene as salient regions following the center-surround contrast framework~\cite{Itti98}, where different kinds of features are combined according to the feature integration theory~\cite{TreismanG80}. The multi-scale pixel contrast is studied in~\cite{LiuG06, LiuYSWZTS11}. The discriminant center-surround hypothesis is analyzed in~\cite{GaoMV07, GaoV07}. Color histograms, computed to represent the center and the surround,
are used to evaluate the center-surround dissimilarity~\cite{LiuYSWZTS11}.
An information theory perspective is introduced
to yield a sound mathematical formulation,
computing the center-surround divergence based on feature
statistics~\cite{KleinF11}. A cost-sensitive SVM is trained to measure the separability of a center region w.r.t. its surroundings~\cite{LiLSDH13Contextual}. The uniqueness can also be captured in a global scope by comparing a patch to its $k$ nearest neighbors~\cite{GofermanZT10} or as its distance to the average patch over the image along the principal component axis coordinates~\cite{MargolinTZ2013}.

% Goferman~\etal~\cite{GofermanZT10} propose to compare a patch to its most similar patches to compute its saliency, where their spatial distances are taken into consideration. In~\cite{MargolinTZ2013}, the distinctness of each patch is measured by the its distance to the average patch
% over the image along the principle component axis coordinates.

The center-surround difference framework is also
investigated to compute the saliency
from region-based image representation.
The multi-\layer~image segmentation is adopted for salient object detection based on local regional contrast~\cite{JiangWYLZL11}. The global regional contrast is studied in~\cite{ChengZMHH11,PerazziKPH12} as well. To further enhance the performance, saliency maps on hierarchical segmentations
are computed and finally combined through a tree model via dynamic
programming~\cite{YanXSJ2013}. Both the color and textural global uniqueness are investigated in~\cite{scharfenberger2013statistical, shi2013pisa}.

% \hzhide{In~\cite{JiangWYLZL11},
% the difference between the color histogram
% of a region and its immediately neighboring regions
% are used to evaluate the saliency score.
% The saliency maps,
% computed from multi-scale image segmentation
% to capture non-local contrast,
% are then combined together
% to form a master saliency map.
% The global contrast based approach~\cite{ChengZMHH11},
% computing the saliency map by comparing each region with others,
% aims to directly compute the global uniqueness. The global regional contrast, defined in the form of their Euclidean distance, can be efficiently computed using the high-dimensional Gaussian filter~\cite{PerazziKPH12}. The saliency map
% is generated by propagating the saliency scores of regions to the pixels.
% Similar to~\cite{JiangWYLZL11}, saliency maps on hierarchical segmentations
% are computed and finally combined through a tree model via dynamic
% programming~\cite{YanXSJ2013}. Both the color and textural global uniqueness are investigated in~\cite{scharfenberger2013statistical, shi2013pisa} for salient object detection.}

Recently, Cheng~\etal~\cite{ChengWLZVC13Efficient} propose a soft image abstraction using a Gaussian Mixture Model (GMM), where each pixel maintains a probability belonging to all the regions instead of a single hard region label, to better compute the saliency. The global uniqueness can also be captured with the low-rank matrix recovery framework~\cite{DBLP:conf/cvpr/ShenW12, zou2013segmentation, peng2013salient}. 
%The basic assumption is that the entire image can be approximated well by a low-rank matrix plus sparse noises in a feautre space. 
The low-rank matrix corresponds to the background regions while sparse noises are indications of salient regions. A submodular salient object detection algorithm is presented in~\cite{JiangD13submodular}, where superpixels are gradually grouped to form potential salient regions by iteratively optimizing a submodular facility location problem. The Bayesian fraemwork is introduced for salient object detection in~\cite{DBLP:conf/eccv/RahtuKSH10,XieEtAlTIP2013}. A partial differential equation (PDE) is also introduced for salient object detection in a recent work~\cite{liu2014adaptive}.

In addition to capturing the uniqueness, many other priors are also proposed for saliency computation.
Centeral prior,
\ie,
the salient object usually lies in the center of an image,
is investigated in~\cite{JiangWYLZL11, WangWZFZL12}.
Object prior,
such as
connectivity prior~\cite{VicenteKR08},
concavity context~\cite{LuZLX11}, and
auto-context cue~\cite{WangXZH11}, backgroundness prior~\cite{WeiWZ012, YangZLRY2013, li2013saliency, Jiang2013Saliency}, generic objectness prior~\cite{ChangLCL11, JiangLYP13UFO, jia2013category}, and background connectivity prior~
\cite{zou2013segmentation, zhang2013boolean, zhu2014saliency}
are also studied for saliency computation. Example-based approaches,
searching for similar images of the input,
are developed for salient object detection~\cite{MarchesottiCC09,WangKIJR11}.
% A top-down approach via joint conditional random fields
% and dictionary learning
% is introduced~\cite{YangY12}.
The depth cue is leveraged
for saliency analysis derived from stereopsic image pairs in~\cite{NiuGLL12} and a depth camera (\eg, Kinect) in~\cite{desingh2013depth}. Li~\etal~\cite{li2014saliency}~adopt the light field camera for salient object detection.
Besides,
spectral analysis in the frequency domain
is used to detect salient regions~\cite{AchantaHES09}.

% Additionally, there are several works
% directly checking if an image window contains an object.
% The generic objectness measure is defined
% by combining several image cues
% to quantify the possibility
% that an image window contains an object~\cite{AlexeDF10}.
% A category independent object detection cascade,
% which uses superpixel boundary integral, edge distribution and window symmetry
% to describe objectness,
% is learnt to rank a number of object window candidates~\cite{RahtuKB11}.
% Salient object detection by composition~\cite{FengWTZS11}
% checks if the content within an window
% can be composed by neighbor regions.
% A random forest regression approach
% is adopted
% to directly regress the object rectangle
% from the saliency map~\cite{WangWZFZL12}.

\myPara{Supervised approaches} Inspired by the feature integration theory, some approaches focus on learning the linear fusion weight of saliency features. Liu~\etal~\cite{LiuYSWZTS11} propose to learn the linear fusion weight of saliency features in a Conditional Random Field (CRF) framework. Recently, the large-margin framework was adopted to learn the weights in~\cite{lu2014learning}. Due to the highly non-linear essence of the saliency mechanism, the linear mapping might not perfectly capture the characteristics of saliency. In~\cite{KhuwuthyakornRZ10}, a mixture of linear Support Vector Machines (SVM) is adopted to partition the feature space into a set of sub-regions that were linearly separable using a divide-and-conquer strategy. Alternatively, a Boosted Decision Tree (BDT) is learned to get an initial saliency map, which will be further refined using a high dimensional color transform~\cite{kim2014salient}. In~\cite{MehraniS10}, generic regional properties are investigated for salient object detection. Li~\etal~\cite{li2014secrets} propose to generate a saliency map by adaptively averaging the object proposals~\cite{DBLP:conf/cvpr/CarreiraS10} with their foreground probabilities that are learned based on eye fixations features using the \rforest~regressor. Additionally, Wang~\etal~\cite{wang2012salient} learn a \rforest~to directly localize the salient object on thumbnail images. In~\cite{Moosmann06learningsaliency}, a saliency map is used to guide the sampling of sliding windows for object category recognition, which is online learned during the classification process.

\vspace{.05in}Our proposed discriminative regional feature integration (DRFI) approach is a supervised salient object detection algorithm. Compared with unsupervised methods, our approach extends the contrast value used in existing algorithms to the contrast vector to represent a region. More importantly, instead of designing heuristic integration rules, our approach is able to \emph{automatically} combine the high-dimensional saliency features in a data-driven fashion and pick up the discriminative ones. 
%The trained Random Forest region saliency regressor can automatically discover the fusion rules from training data and pick up the most discriminative ones. 
Compared with existing supervised methods, our method learns a highly non-linear combination of saliency features and does not require any assumption of the feature space. The most similar approaches to ours might be~\cite{MehraniS10, kim2014salient}. \cite{MehraniS10} is a light touch on the discriminative feature integration without presenting a deep investigation, which only considers the regional property descriptor. In~\cite{kim2014salient}, the learned saliency map is only used as a pre-processing step to provide a coarse estimation of salient and background regions while our approach directly output the saliency map.

It is noted that some supervised learning approaches exist to predict eye fixation~\cite{DBLP:conf/iccv/JuddEDT09, LuZJX12}. The features, \eg, the local energy of the steerable pyramid
ﬁlters~\cite{DBLP:conf/icip/SimoncelliF95} in~\cite{DBLP:conf/iccv/JuddEDT09}
and the perceptual Gestalt grouping cues in~\cite{LuZJX12}, seem to be more
suitable for eye fixation prediction, while our approach is specifically designed for salient object detection. We also note the discriminative feature fusion has also been studied in image classification~\cite{FernandoFMS12}, which learns the adaptive
weights of features according to the classification task to better distinguish
one class from others. Instead, our approach integrates three types of regional features
in a discriminative strategy for the saliency regression on multiple
segmentations.

\section{Image saliency computation}
\label{sec:imgSaliencyComputation}
\begin{figure}[t]
    \centering
    \includegraphics[width=0.5\textwidth,keepaspectratio]{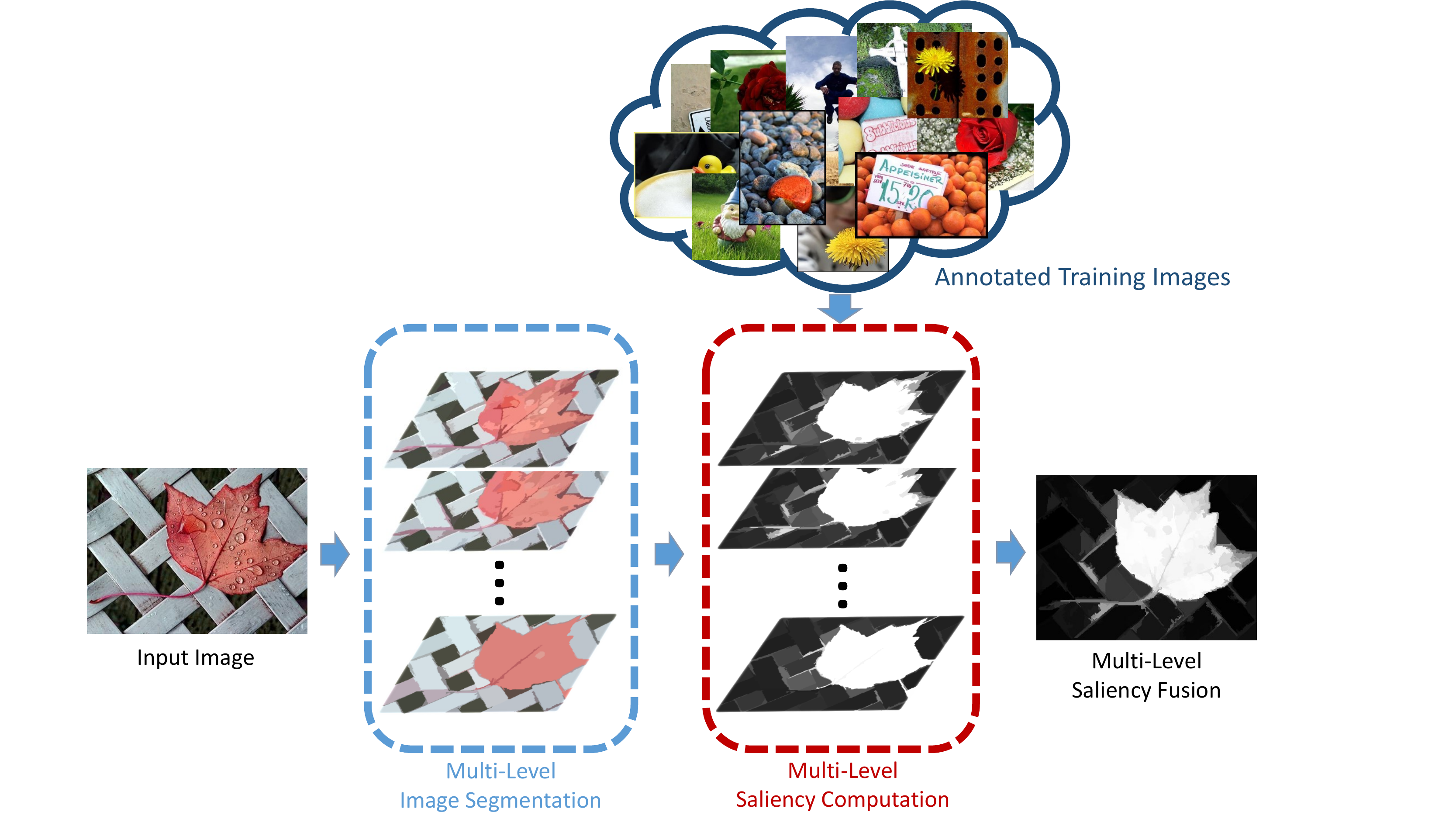}
    \caption{The framework of our proposed discriminative regional feature integration 
        (DRFI) approach.
    }\label{fig:pipeline}
\end{figure}

The pipeline of our approach
consists of three main steps:
multi-\layer~segmentation
that decomposes an image into regions,
regional saliency computation
that maps the features extracted from each region
to a saliency score,
and multi-\layer~saliency fusion
that combines the saliency maps
over all the \layers~of segmentations
to get the final saliency map.
The whole process is illustrated in~\Figref{fig:pipeline}.

\begin{figure*}
    \centering
    \small
    \renewcommand{\arraystretch}{0.8}
    \renewcommand{\tabcolsep}{1.44mm}
    {
    \begin{tabular}{|l|l|c||c|c||c|c|}
    \hline
    \multicolumn{3}{|c||}{Color and texture features} &
    \multicolumn{2}{|c||}{~Differences of features~} &
    \multirow{2}{*}{Contrast} & \multirow{2}{*}{Backgroundness} \\
    \cline{1-5}
    \cline{1-5}
     & ~~~~~~~~~~~~~~~~~~~features & ~dim~   &  ~~~definition~~~ & ~~~dim~~~ &  &
    \\
    \hline\hline
    $\mathbf{a}_1$ &  average RGB values & $3$ & $d(\mathbf{a}_1^{R_i},
    \mathbf{a}_1^{S})$ & $3$ & $c_1\sim c_3$ & $b_1\sim b_3$\\
    $\mathbf{h}_1$ &  RGB histogram & 256 & $\chi^2(\mathbf{h}_1^{R_i},
    \mathbf{h}_1^{S})$ & $1$ & $c_4$ & $b_4$\\
    $\mathbf{a}_2$ &  average HSV values & $3$ & $d(\mathbf{a}_2^{R_i},
    \mathbf{a}_2^{S})$ & $3$ & $c_5\sim c_7$ & $b_5\sim b_7$\\
    $\mathbf{h}_2$ &  HSV histogram & 256 & $\chi^2(\mathbf{h}_2^{R_i},
    \mathbf{h}_2^{S})$ & $1$ & $c_8$ & $b_8$\\
    $\mathbf{a}_3$ &  average L*a*b* values & $3$   & $d(\mathbf{a}_3^{R_i},
    \mathbf{a}_3^{S})$ & $3$ & $c_9\sim c_{11}$ & $b_9\sim b_{11}$\\
    $\mathbf{h}_3$ &  L*a*b* histogram & 256 &
    $\chi^2(\mathbf{h}_3^{R_i}, \mathbf{h}_3^{S})$ & $1$& $c_{12}$ & $b_{12}$\\
    $\mathbf{r}$ &  absolute response of LM filters & $15$  &
    $d(\mathbf{r}^{R_i}, \mathbf{r}^{S})$ & $15$ & $c_{13}\sim c_{27}$& $b_{13}\sim
    b_{27}$\\
    $\mathbf{h}_4$ &  max response histogram of the LM filters &  $15$
    & $\chi^2(\mathbf{h}_4^{R_i}, \mathbf{h}_4^{S})$ & $1$ & $c_{28}$ & $b_{28}$\\
    $\mathbf{h}_5$ &  histogram of the LBP feature &  $256$
    & $\chi^2(\mathbf{h}_4^{R_i}, \mathbf{h}_5^{S})$ & $1$ & $c_{29}$ & $b_{29}$\\
    \hline
    \end{tabular}
    }\\
    \caption{\small Color and texture features describing
    the visual characteristics of a region which are used to compute the regional
    feature vector.
    $d(\mathbf{x}_1, \mathbf{x}_2) = \left(|x_{11}-x_{21}|,\cdots,
    |x_{1d}-x_{2d}|\right)$ where $d$ is the number of elements in the vectors
    $\mathbf{x}_1$ and $\mathbf{x}_2$. And $\chi^2(\mathbf{h}_1, \mathbf{h}_2)
    =\sum_{i=1}^b \frac{2(h_{1i} - h_{2i})^2}{h_{1i} + h_{2i}}$
    with $b$ being the number of histogram bins.
    The last two columns denote the symbols for
    regional contrast and backgroundness descriptors. (In the definition of a
    feature, $S$ corresponds to $R_j$ for the regional contrast descriptor and $B$ for
    the regional backgroundness descriptor, respectively.)}
    \label{tab:RegionContrastBackgroundness}
\end{figure*}

\myPara{Multi-\layer~segmentation}
Given an image $I$,
we represent it by a set of $M$-\layer~segmentations
$\mathcal{S} = \{\mathcal{S}_1, \mathcal{S}_2,\cdots,\mathcal{S}_M\}$,
where each segmentation $\mathcal{S}_m$ is a decomposition of the image $I$. We apply the graph-based image segmentation approach~\cite{FelzenszwalbH04} to
generate multiple segmentations using $M$ groups of different parameters.
% and consists of $K_m$ regions.
% $\mathcal{S}_1$ is the finest segmentation
% consisting of the largest number of regions,
% and $\mathcal{S}_M$ is the coarsest segmentation
% consisting of the smallest number of regions.

Due to the limitation of low-level cues, none of the current
segmentation algorithms can reliably segment the salient object. Therefore, we resort to the multi-\layer~segmentation for robustness purpose.
%choose to compute saliency on multi-\layer~segmentation for robustness purpose.
In~\secref{sec:TrainingSamples}, we will further demonstrate how to utilize multi-\layer~segmentation to
generate a large amount of training samples.

\myPara{Regional saliency computation}
% Our algorithm computes the saliency score for each region.
% It seems
% that the separate computation ignores the relation
% of neighboring regions.
% However, our algorithm essentially takes into consideration
% of such relations
% because we conduct the region saliency computation on multi-\layer segmentations.
% The spatial consistency of saliency scores
% for neighboring regions
% is imposed
% as the neighboring regions in the finer-\layer segmentation may
% form a single region
% in the coarser \layer.
In our approach, we predict saliency scores for each region that is jointly represented by three types of features:
regional contrast, regional property, and regional backgroundness,
which will be described in~\secref{sec:RegionalFeatures}.
At present, we denote the feature as a vector $\mathbf{x}$.
Then the feature $\mathbf{x}$ is passed
into a random forest regressor $f$,
yielding a saliency score.
The Random Forest regressor is learnt
from the regions of the training images
and integrates the features
together in a discriminative strategy.
The learning procedure will be given in~\secref{sec:Learning}.

\vspace{.1cm}\noindent\textbf{Multi-\layer~saliency fusion.}
After conducting region saliency computation,
each region has a saliency value.
For each \layer, we assign the saliency value of each region to its contained pixels.
As a result,
we generate $M$ saliency maps $\{\mathbf{A}_1, \mathbf{A}_2, \cdots, \mathbf{A}_M\}$,
and then fuse them together,
$\mathbf{A} = g(\mathbf{A}_1, \cdots, \mathbf{A}_M)$,
to get the final saliency map $\mathbf{A}$, where $g$ is a combinator function
introduced in~\secref{sec:LearnSaliencyFusor}.

\section{Regional Saliency Features}
\label{sec:RegionalFeatures}
In this section, we present three types of regional saliency features, leading to a 93-dimensional feature vector for each region.

\subsection{Regional contrast descriptor}
A region is likely thought to be salient
if it is different from others.
Unlike most existing approaches
that compute
the contrast values,
\eg,
the distances of region features like color and texture,
and then combine them together directly
forming a saliency score,
our approach computes a contrast descriptor,
a vector representing the differences of feature vectors
of regions.
%which will be combined with other features
%via a discriminative feature integration scheme.

\newcommand{\vv}{\mathbf{v}}
\newcommand{\vp}{\mathbf{p}}

To compute the contrast descriptor,
we describe each region $R_i\in\mathcal{S}_m$
by a feature vector,
including color and texture features,
denoted by $\vv^{R_i}$.
The detailed description is given
in \tabref{tab:RegionContrastBackgroundness}. For color features, we consider RGB, HSV, and L*a*b* color spaces. For texture features, we adopt the LBP feature~\cite{DBLP:journals/pr/HeikkilaPS09} and the responses of the LM filter bank~\cite{DBLP:journals/ijcv/LeungM01}.

As suggested in previous works~\cite{ChengZMHH11,PerazziKPH12}, the regional contrast value $x_k^c$ derived from the $k$-th feature channel is computed by checking $R_i$ against all other regions,
\begin{eqnarray}
x_k^c(R_i) = \sum_{j=1}^{N_m} \alpha_j w_{ij} D_k(\vv^{R_i}, \vv^{R_j}),
\end{eqnarray}
where $D_k(\vv^{R_i}, \vv^{R_j})$ captures the difference of the $k$-th channel of the feature vectors $\vv^{R_i}$ and $\vv^{R_j}$. 
Specifically, 
the difference of the histogram feature
is computed as the $\chi^2$ distance 
and as their absolute differences for other features.
% the differences of other features
% are computed as their absolute differences. 
$w_{ij}=e^{-\frac{||\vp_i^m - \vp_j^m||^2}{2\sigma_s^2}}$ is a spatial weighting term, where $\vp_i$ and $\vp_j$ are the mean positions of $R_i$ and $R_j$, respectively. $\sigma_s$ controls the strength of the spatial weighting effect. We empirically set it as 1.0 in our implementation. $\alpha_j$ is introduced to account for the irregular shapes of regions, defined as the normalized area of the region $R_j$. $N_m$ is the number of regions in $\mathcal{S}_m$. 
As a result,
we get a $29$-dimensional feature vector.
The details of the regional contrast descriptor are given
in~\tabref{tab:RegionContrastBackgroundness}.

\subsection{Regional backgroundness descriptor}
There exist a few algorithms attempting
to make use of the characteristics of the background (\eg, homogeneous color or
textures)
to heuristically determine if one region is background,
\eg,~\cite{WeiWZ012}.
In contrast,
our algorithm extracts a set of features and adopts the supervised learning
approach to determine the background degree (accordingly the saliency degree) of a
region.

It has been observed
that the background identification depends on
the whole image context.
Image regions with similar appearances
might belong to the background in one image
but belong to the salient object
in some other images.
It is not enough to merely use the property features to
check if one region is in the background
or the salient object.

Therefore,
we extract the pseudo-background region
and compute the backgroundness descriptor for each region with the
pseudo-background region as a reference.
The pseudo-background region $B$ is defined as
the 15-pixel wide narrow border region of the image.
To verify such a definition,
we made a simple survey
on the MSRA-B data set with $5000$ images
and found that $98\%$ of pixels in the border area belongs to the background.
The backgroundness value $x_k^b$ of the region $R_i$ on the $k$-th feature is then defined as
\begin{eqnarray}
x_k^b(R_i) = D_k(\vv^{R_i}, \vv^B).
\end{eqnarray}
We get a $29$-dimensional feature vector. See details in
\tabref{tab:RegionContrastBackgroundness}.

% computed
% as the differences $\operatorname{diff}(\mathbf{v}^R, \mathbf{v}^B)$
% between its color and texture features $\mathbf{v}^R$
% and the color and texture features $\mathbf{v}^B$
% of the pseudo-background region,
% resulting a $29$-dimensional feature vector. See detailes in
% \tabref{tab:RegionFeatures}.

%The properties like the size and location of a region may also influence its saliency score. For example, a region far away from the image center is less likely to be considered as salient.
\subsection{Regional property descriptor}
% In addition to regional contrast, 
Additionally, we consider the generic properties of a region,
including appearance and geometric features.
These two features are
extracted independently from each region
like the feature extraction algorithm in image labeling~\cite{HoiemEH05}.
The appearance features attempt to
describe the distribution of colors and textures in a region,
which can characterize their common properties for the salient object and the background.
For example, the background usually has homogeneous color distribution or similar texture pattern.
The geometric features include
the size and position of a region
that may be useful to describe the spatial distribution of the salient object and the background.
For example, the salient object tends to be placed near the center of the image while the background usually scatters over the entire image.
Finally,
we obtain a $35$-dimensional
regional property descriptor.
The details are given in~\tabref{tab:RegionProperty}.

\vspace{.05in}In summary, we obtain a 93-dimensional ($2\times29 + 35$) feature vector for each region.~\Figref{fig:impFeat} demonstrates visualizations of the most important features for each kind of regional feature descriptor.

\begin{figure}
    \centering
    \small
    \renewcommand{\arraystretch}{1}
    \renewcommand{\tabcolsep}{.7mm}
    \begin{tabular}{|l|c|c|}
    \hline
    ~~~~~~~~~~~~~~~~~description & notation & dim \\
    \hline
    average normalized $x$ coordinates & $p_1$ & $1$ \\
    average normalized $y$ coordinates &  $p_2$ & $1$ \\
    $10th$ percentile of the normalized $x$ coord. & $p_3$ & $1$ \\
    $10th$ percentile of the normalized $y$ coord. & $p_4$ & $1$ \\
    $90th$ percentile of the normalized $x$ coord. & $p_5$ & $1$ \\
    $90th$ percentile of the normalized $y$ coord. & $p_6$ & $1$ \\
    normalized perimeter & $p_7$ & $1$ \\
    aspect ratio of the bounding box &  $p_8$ & $1$  \\
    variances of the RGB values & $p_9\/\sim p_{11}$ & $3$ \\
    variances of the L*a*b* values & $p_{12}\/\sim p_{14}$ & $3$ \\
    variances of the HSV values & $p_{15}\/ \sim p_{17}$ & $3$ \\
    variance of the response of the LM filters &  $p_{18} \sim p_{32}$ & $15$\\
    variance of the LBP feature & $p_{33}$ & 1 \\
    normalized area & $p_{34}$ & $1$ \\
    normalized area of the neighbor regions &  $p_{35}$ & $1$ \\
    \hline
    \end{tabular}\\
    \caption{The regional property descriptor. (The abbreviation
        coord. indicates coordinates.)
    }\label{tab:RegionProperty}
\end{figure}

\section{Learning}
\label{sec:Learning}
% \vspace{.1cm}\noindent\textbf{Generating training samples.}
In this section, we introduce how to learn a \rforest~to map the feature vector of each region to a saliency score. Learning the multi-\layer~saliency fusion weight is also presented.

\subsection{Generating training samples}
\label{sec:TrainingSamples}
% The success of the supervised approaches mostly lies in the availability of the huge amount of training data, \eg~\cite{DBLP:conf/cvpr/ShottonFCSFMKB11}, where
% the training samples cover the variations of the data. In our scenario, such
% variations lie in the scales of the region (\ie, over or
% under segmentation of the object). 
% % While the backgroundness descriptor is
% % invariant to the scale of the region, 
% During the training process, we hope to learn the invariance from the
% training data to robustly estimate the saliency under different parameter
% setting of segmentations. In our
% experiments, however, we find that it is difficult to choose appropriate
% parameters to generate training samples with large variations based on the multi-\layer~
% segmentation introduced in~\secref{sec:imgSaliencyComputation} since the
% cues (\ie, pixels' difference) used for clustering pixels are low-level. In
% most cases, the regions are over-segmentation of the object. It is
% hard to collect big regions covering large portions of the object. Therefore, we adopt a
% region-based supervised approach to predict the similarity of two adjacent
% regions for more accurate grouping.

\begin{figure}[t]
    \centering
    \renewcommand{\arraystretch}{.9}
    \renewcommand{\tabcolsep}{.5mm}
	\begin{tabular}{cccc}
		\includegraphics[height=0.1\textwidth,keepaspectratio]{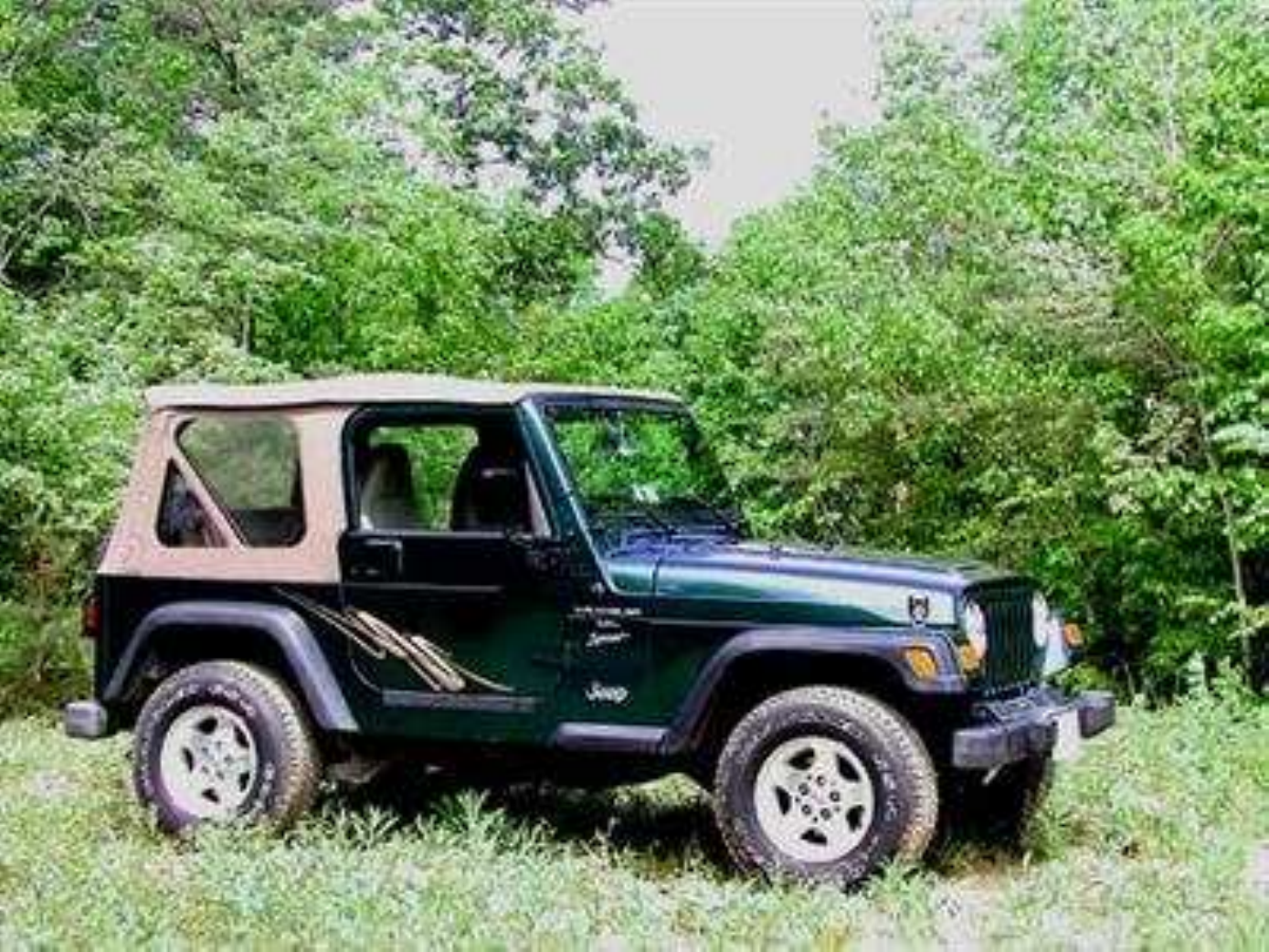}&
		\includegraphics[height=0.1\textwidth,keepaspectratio]{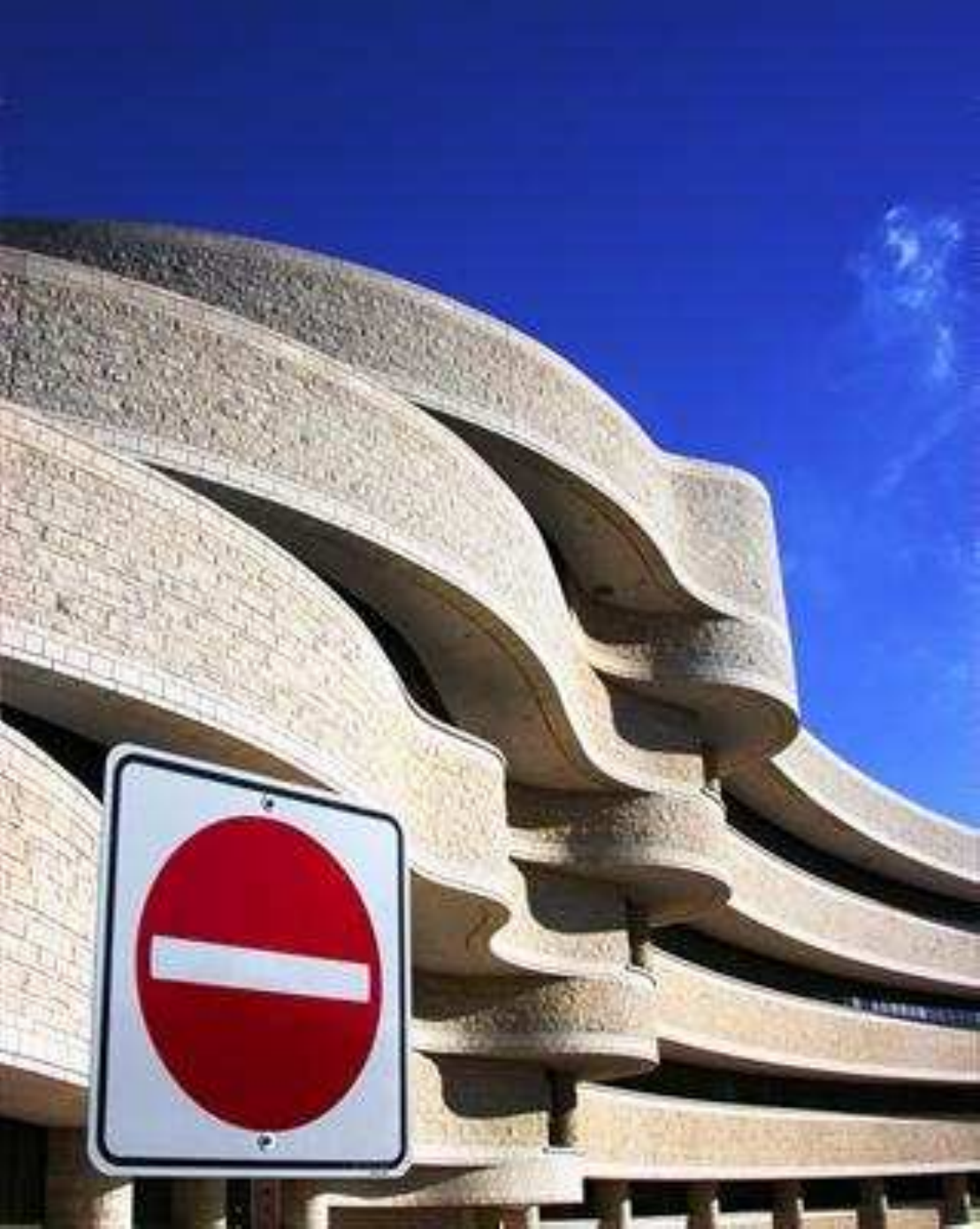}&
		\includegraphics[height=0.1\textwidth,keepaspectratio]{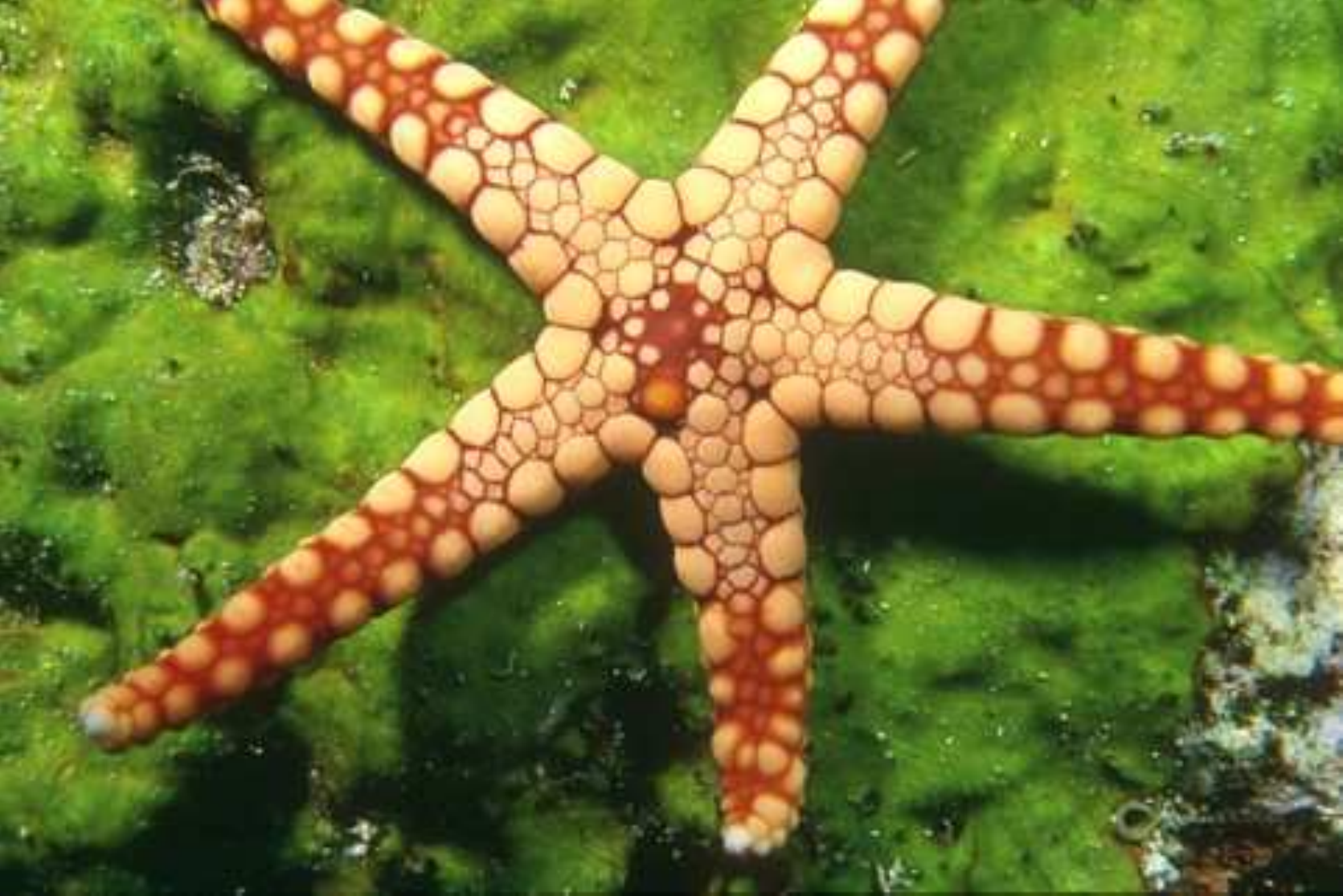}&
		\includegraphics[height=0.1\textwidth,keepaspectratio]{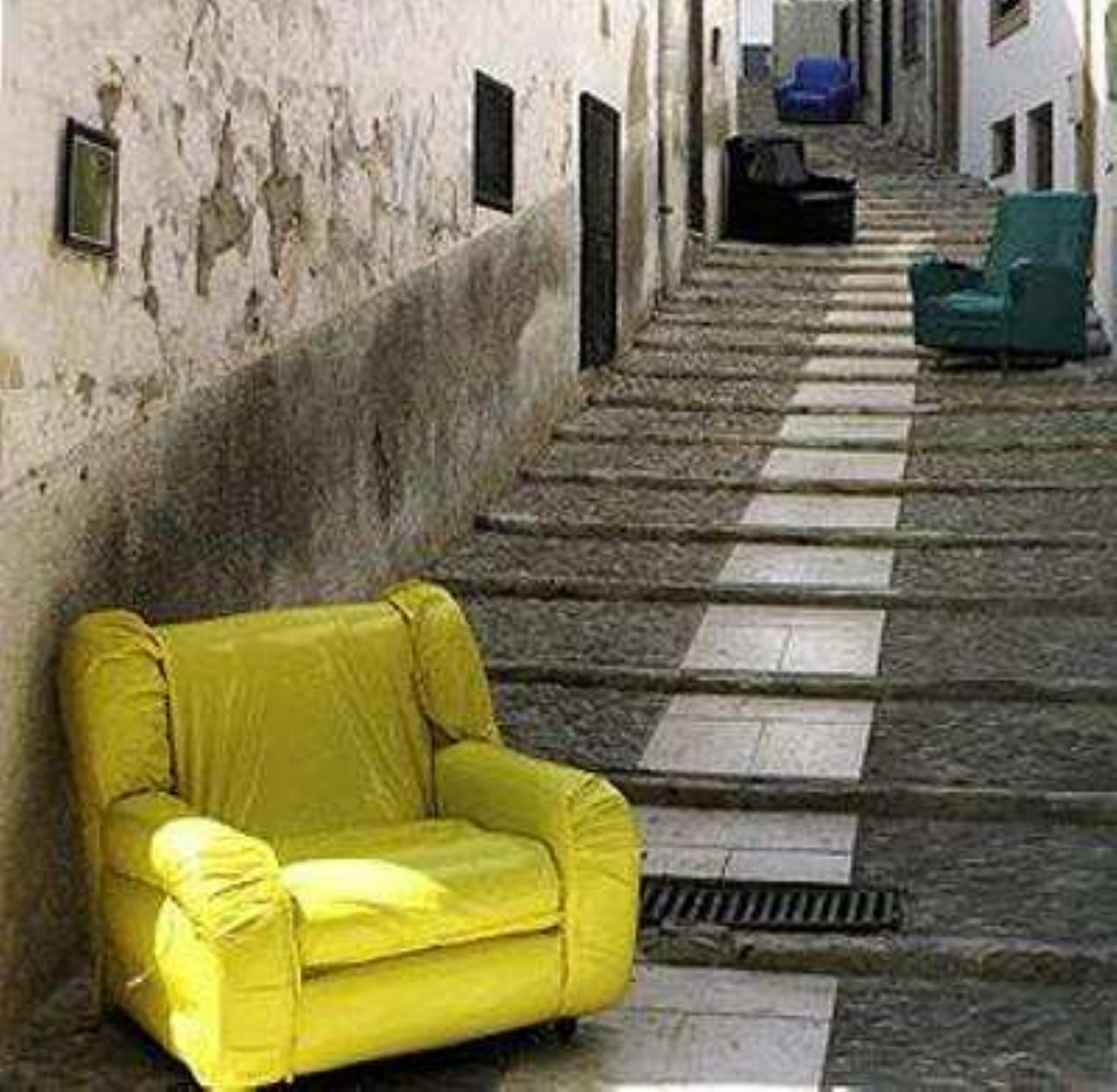}\\
		\includegraphics[height=0.1\textwidth,keepaspectratio]{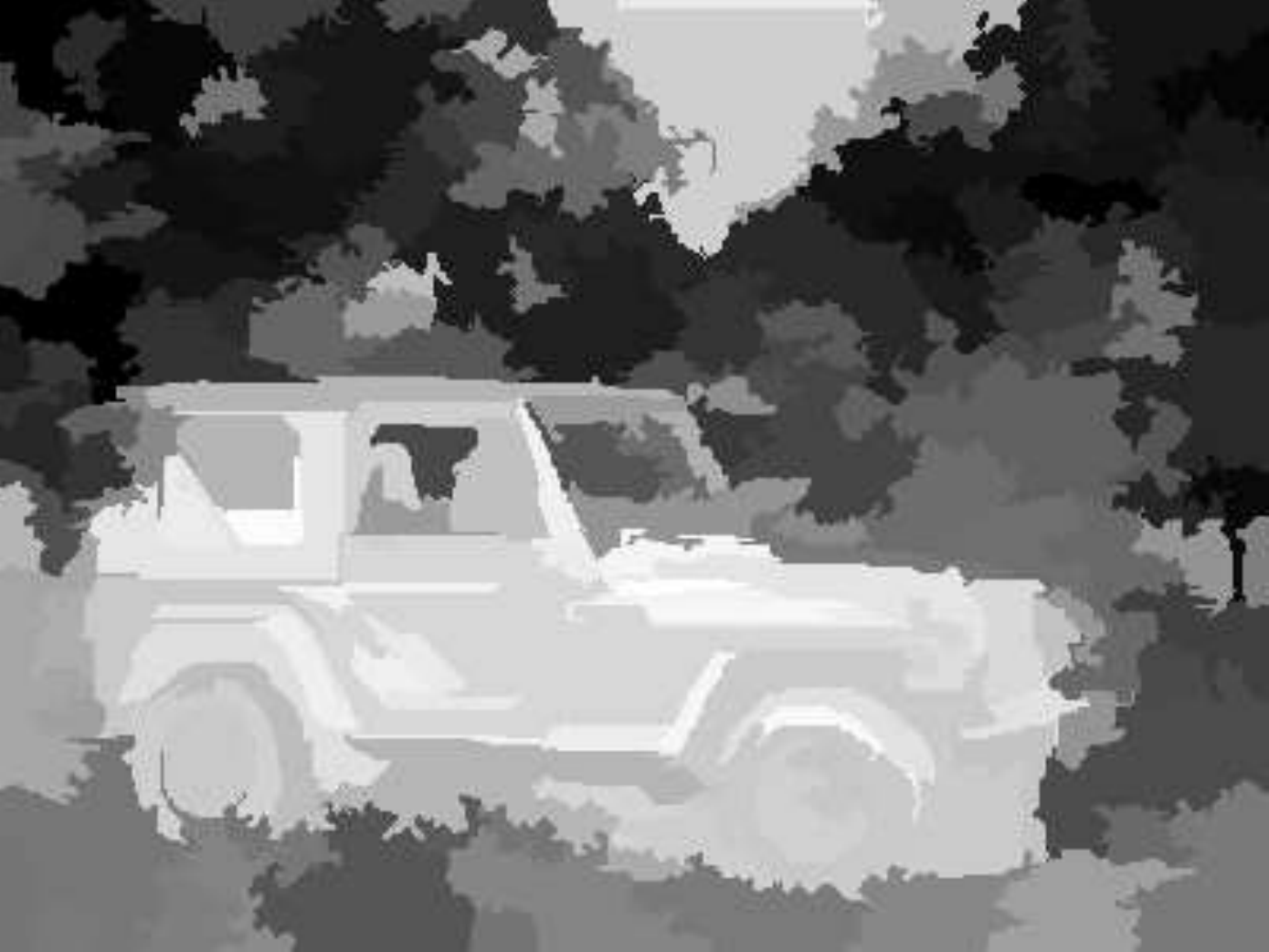}&
		\includegraphics[height=0.1\textwidth,keepaspectratio]{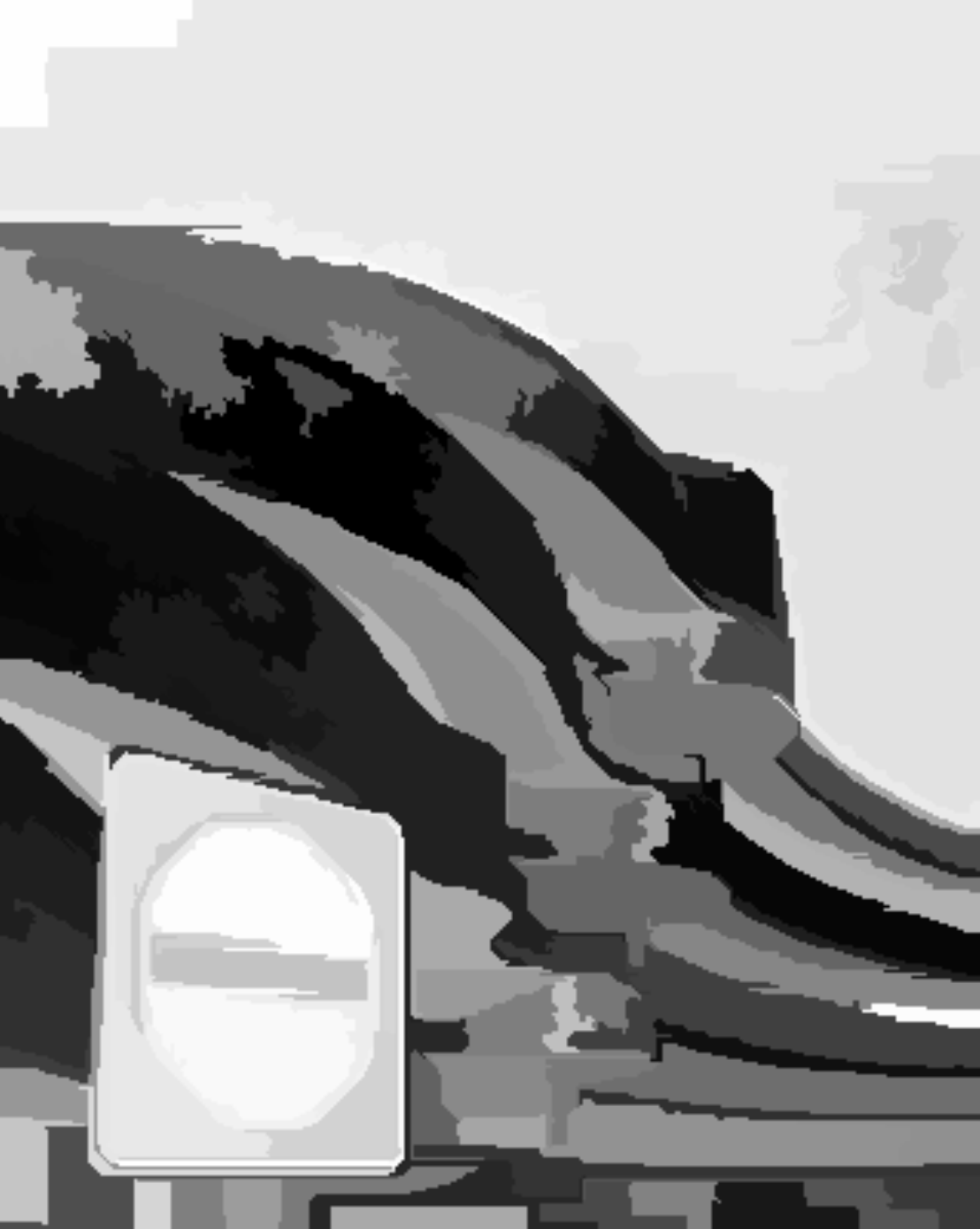}&
		\includegraphics[height=0.1\textwidth,keepaspectratio]{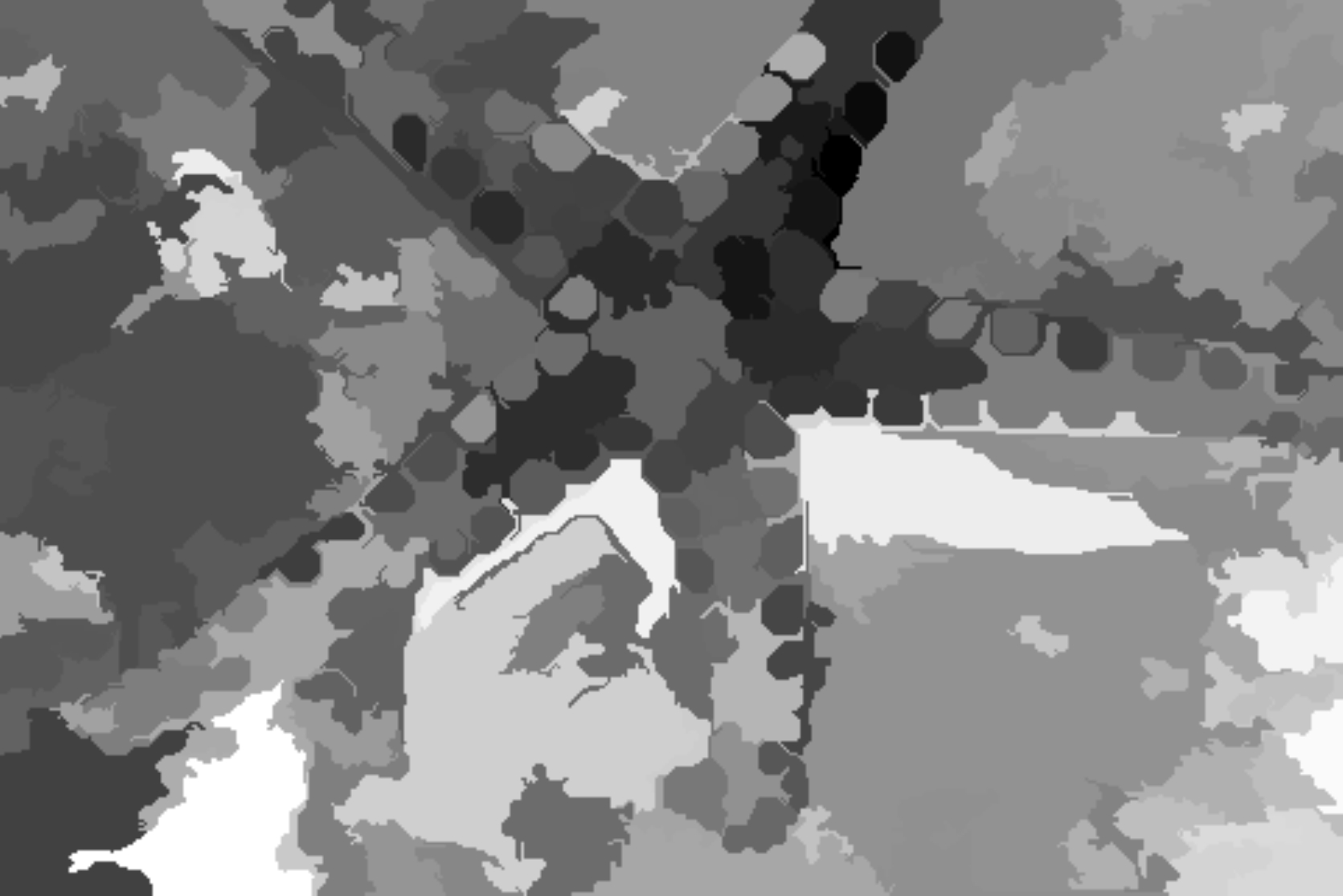}&
		\includegraphics[height=0.1\textwidth,keepaspectratio]{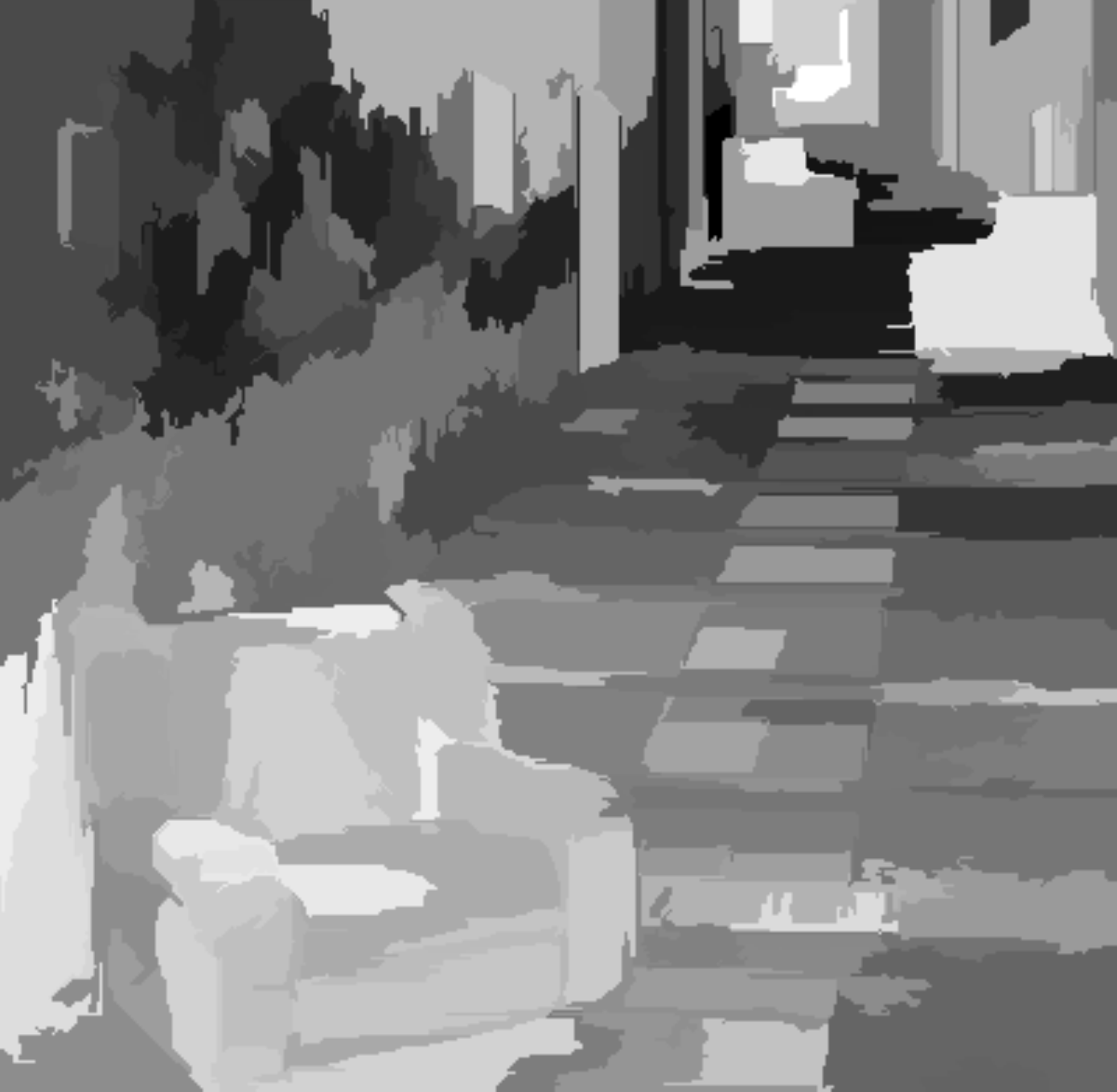}\\
		\includegraphics[height=0.1\textwidth,keepaspectratio]{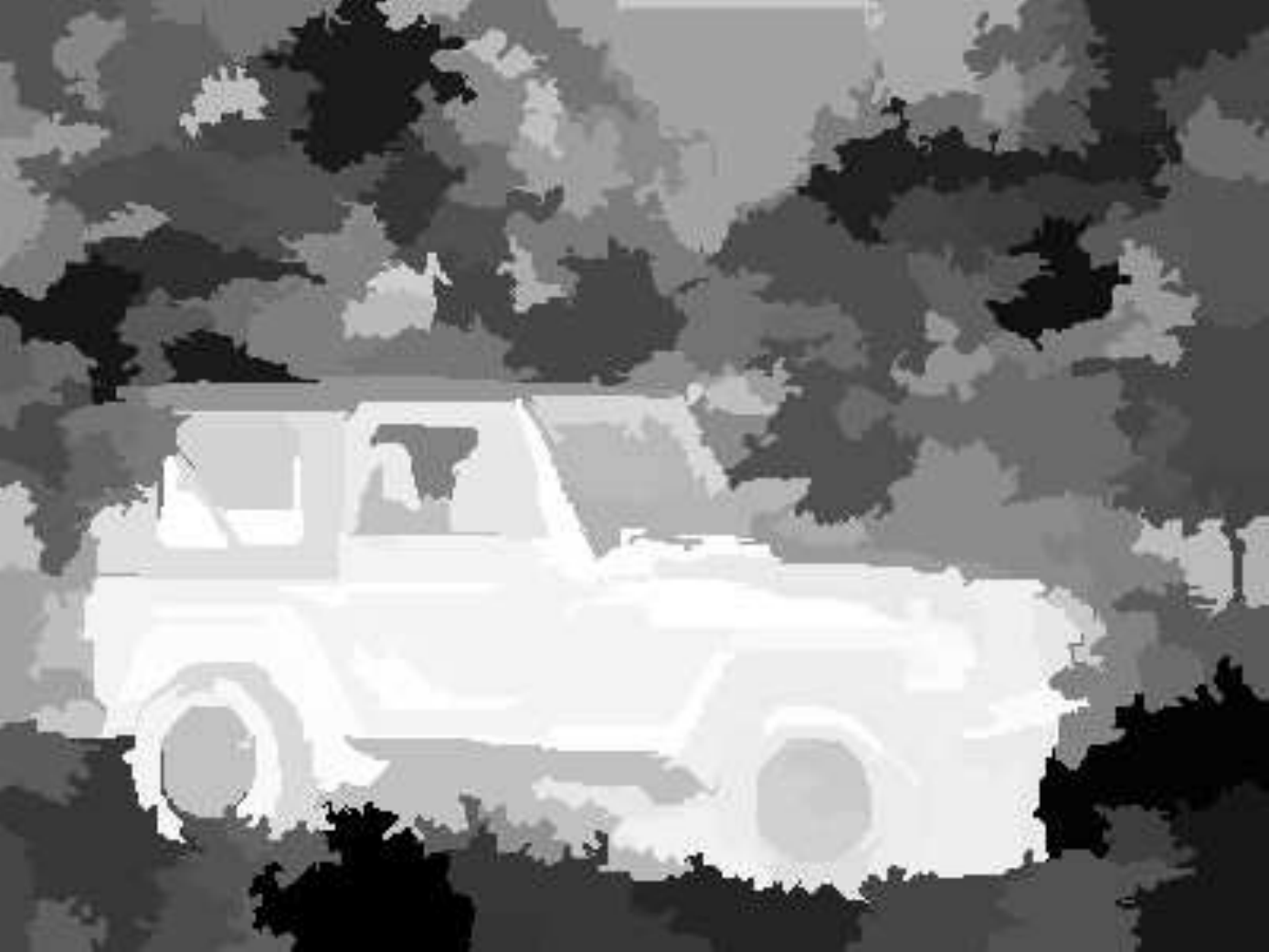}&
		\includegraphics[height=0.1\textwidth,keepaspectratio]{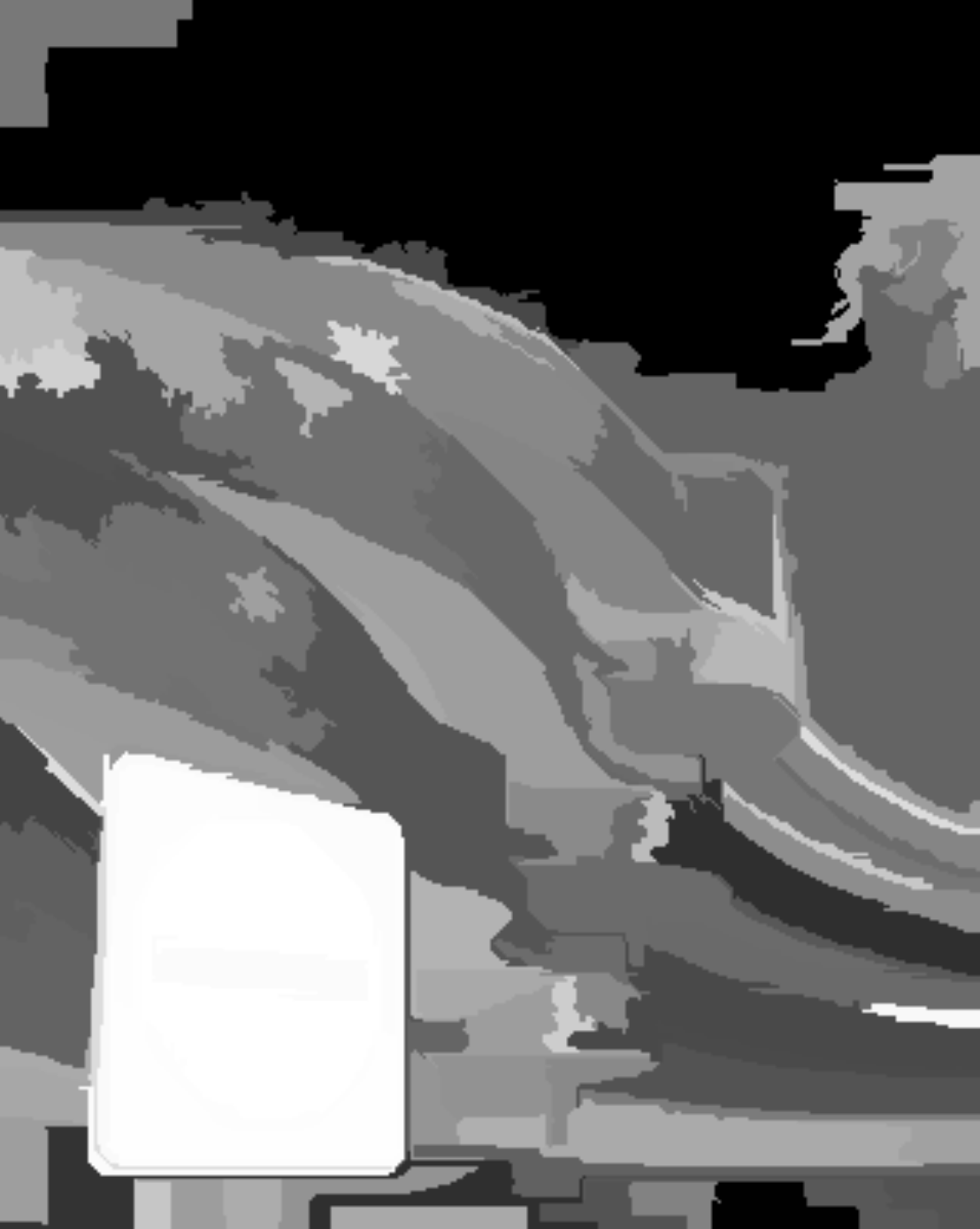}&
		\includegraphics[height=0.1\textwidth,keepaspectratio]{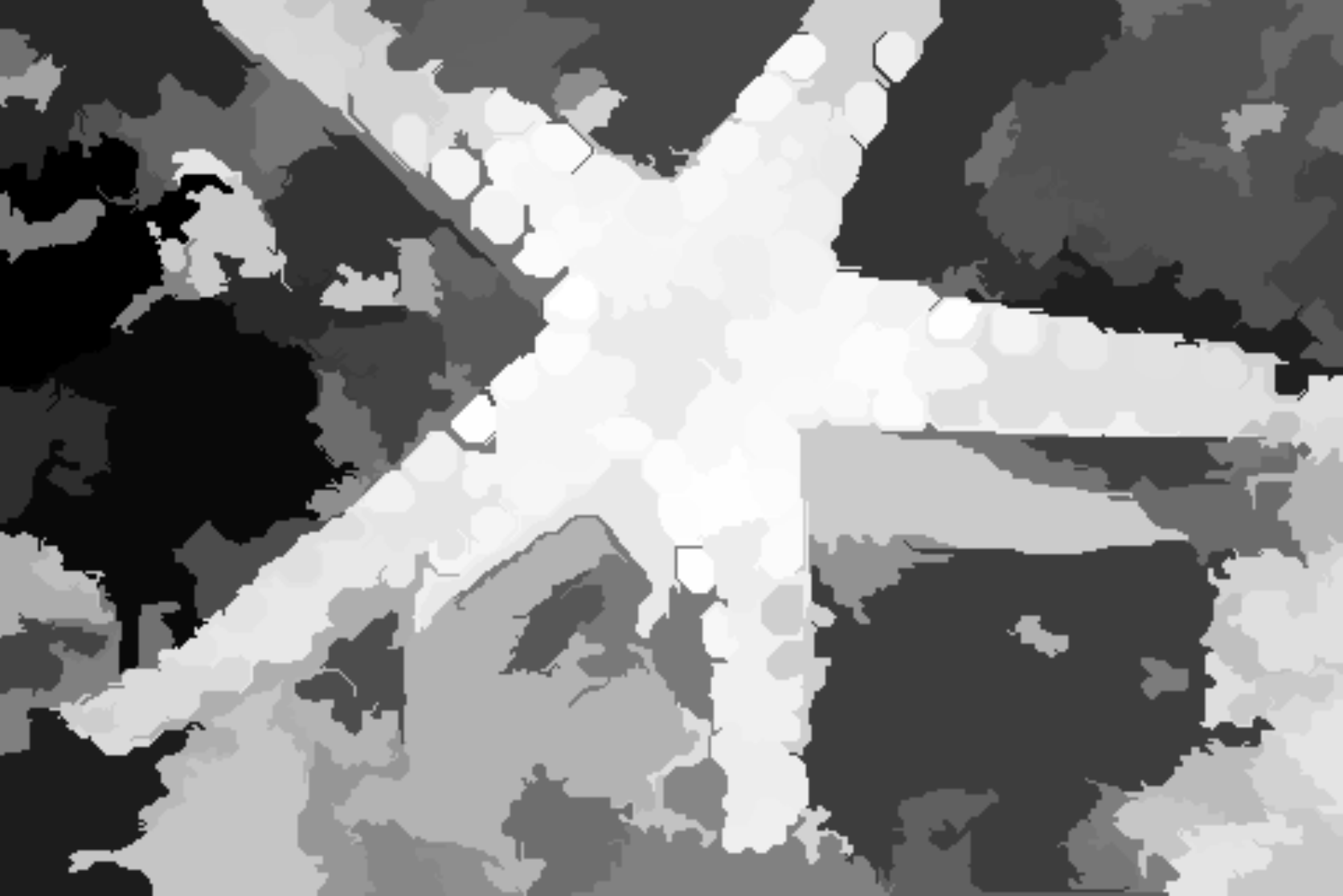}&
		\includegraphics[height=0.1\textwidth,keepaspectratio]{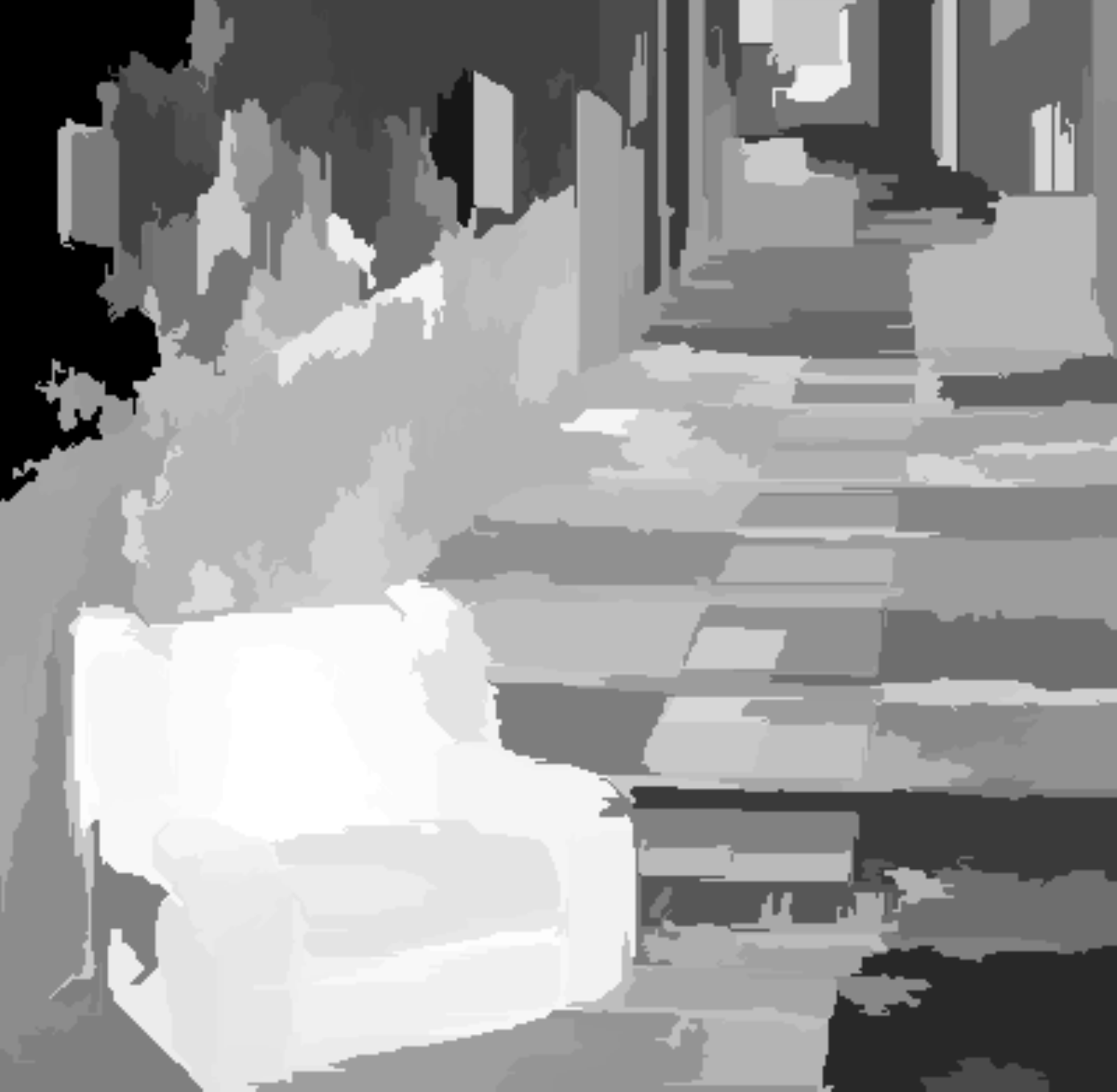}\\
		\includegraphics[height=0.1\textwidth,keepaspectratio]{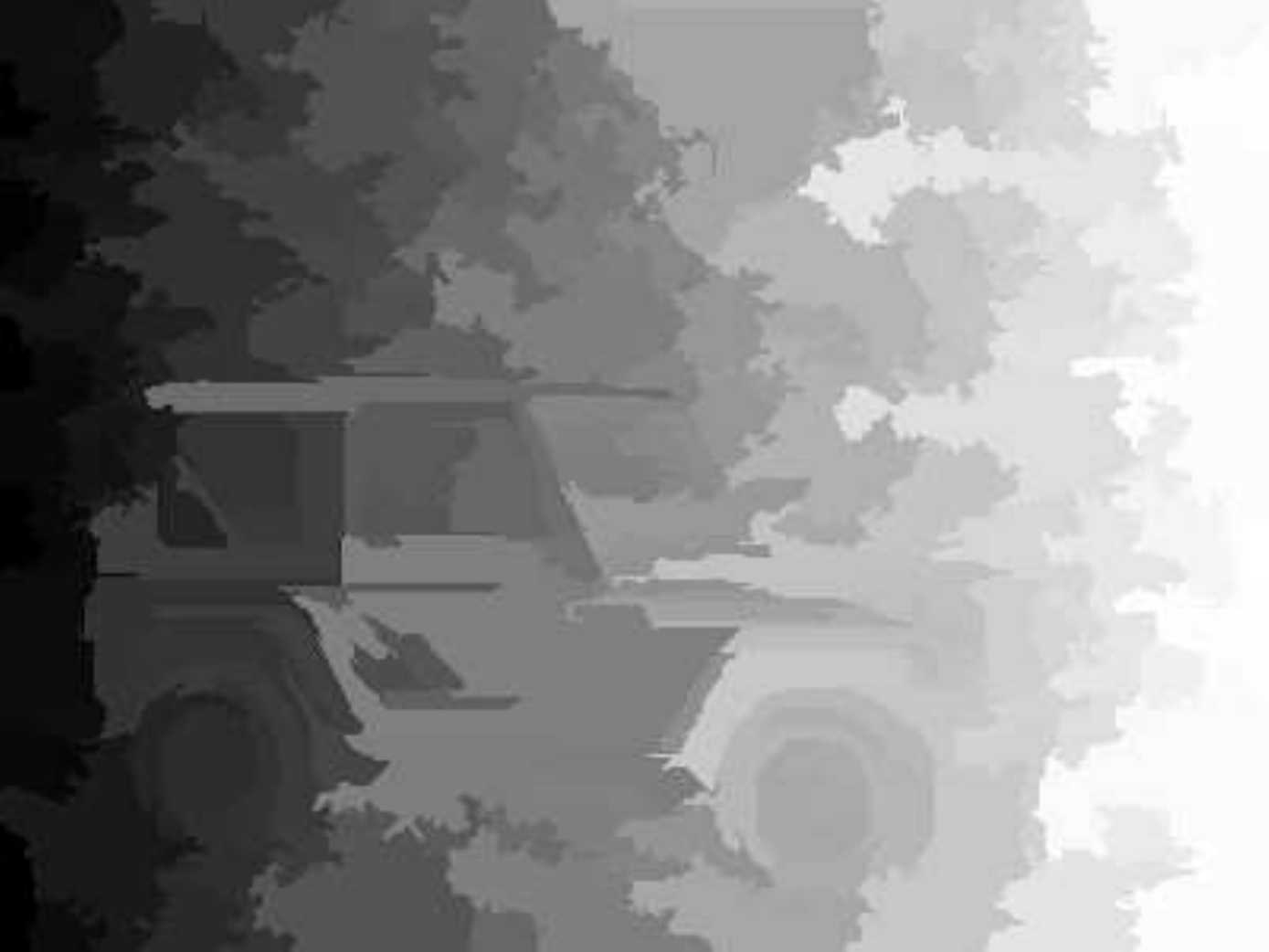}&
		\includegraphics[height=0.1\textwidth,keepaspectratio]{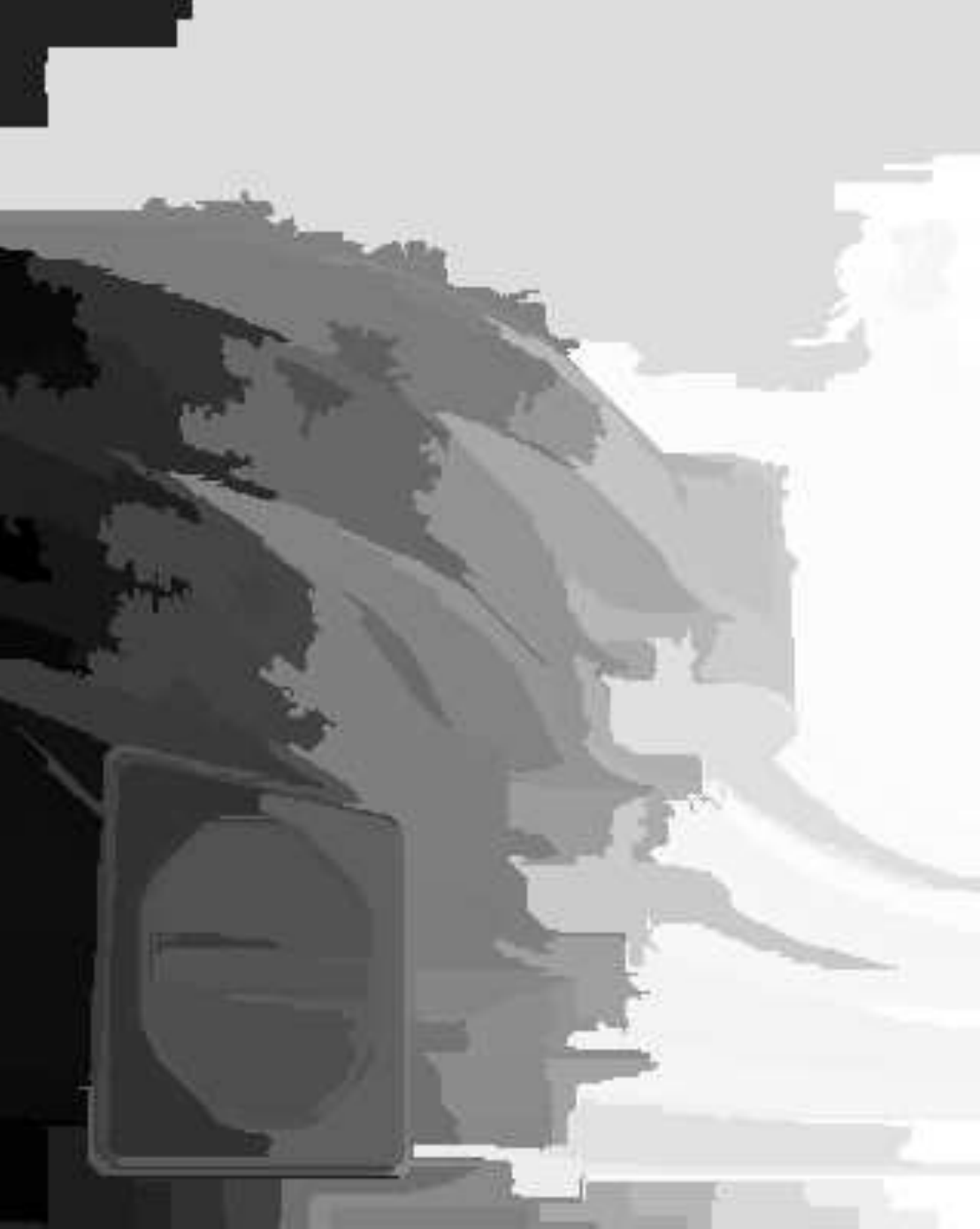}&
		\includegraphics[height=0.1\textwidth,keepaspectratio]{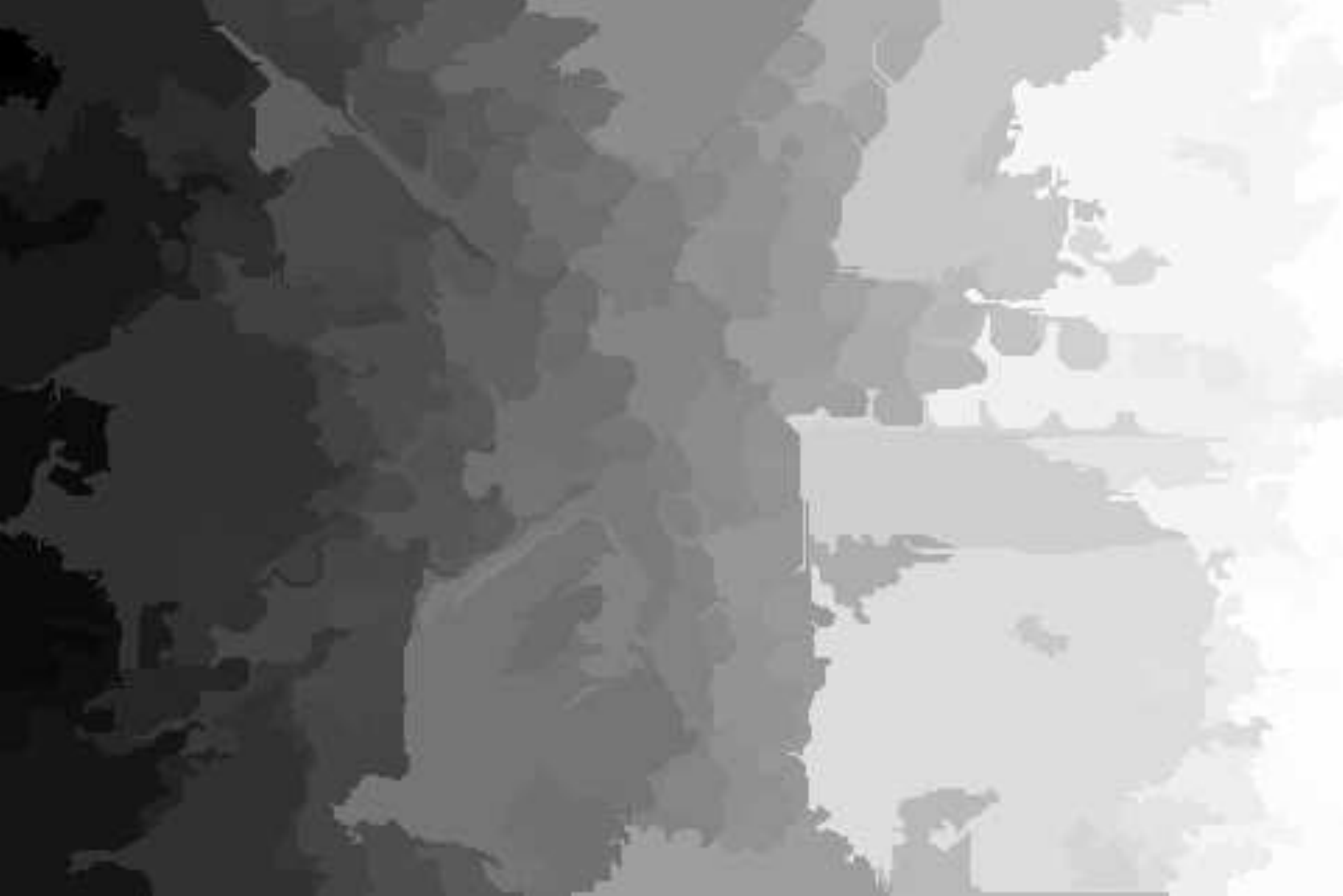}&
		\includegraphics[height=0.1\textwidth,keepaspectratio]{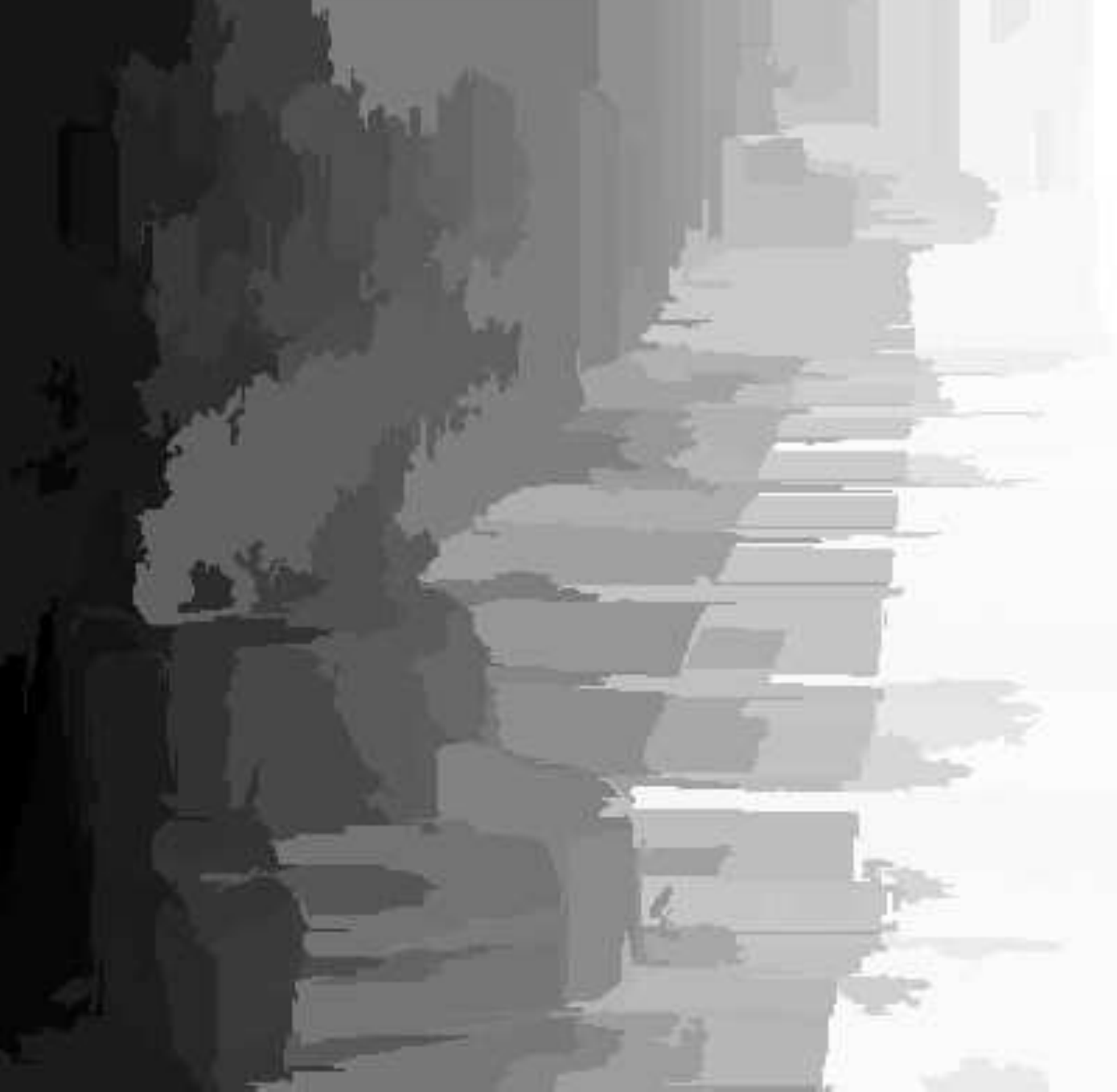}\\
		\includegraphics[height=0.1\textwidth,keepaspectratio]{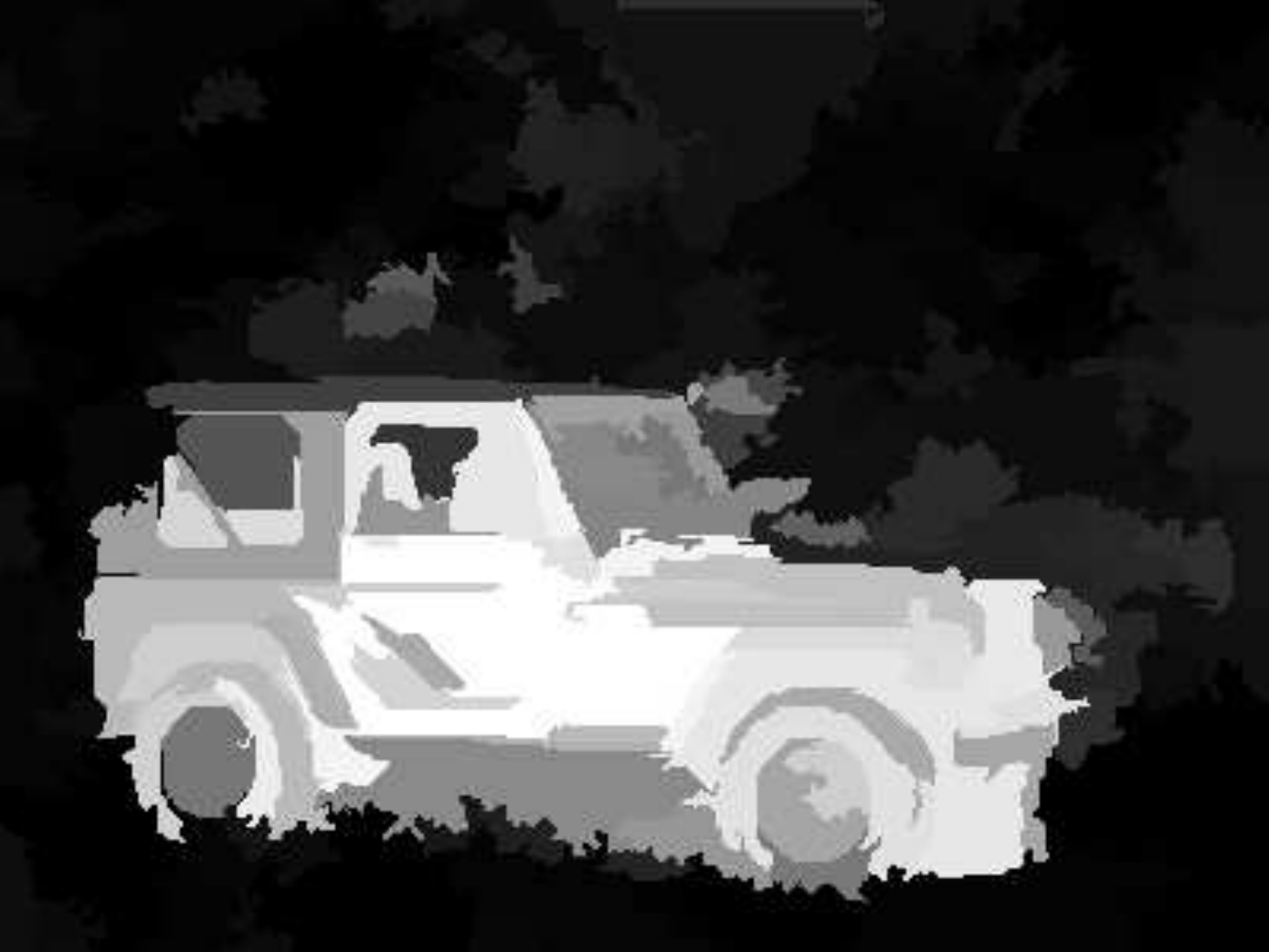}&
		\includegraphics[height=0.1\textwidth,keepaspectratio]{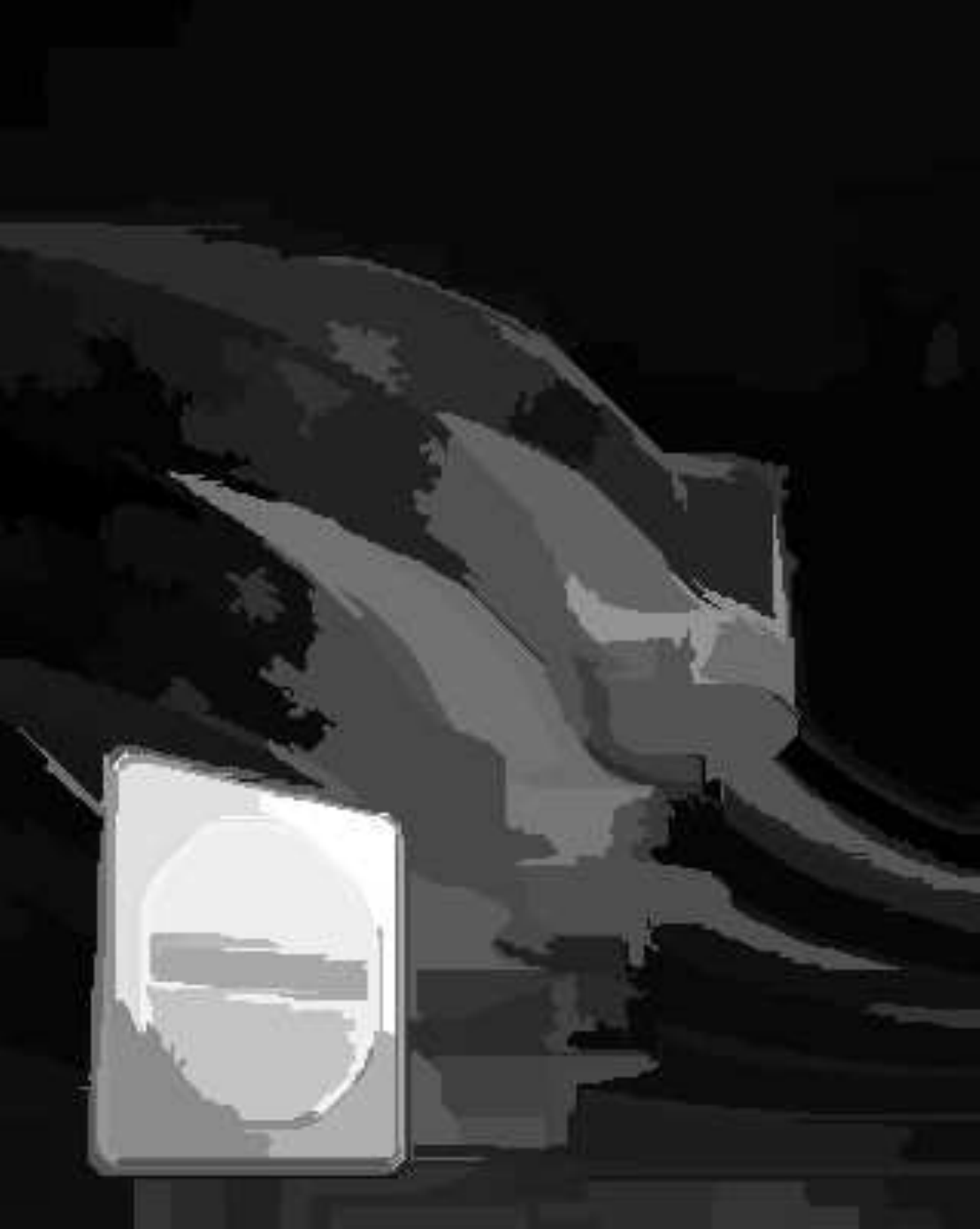}&
		\includegraphics[height=0.1\textwidth,keepaspectratio]{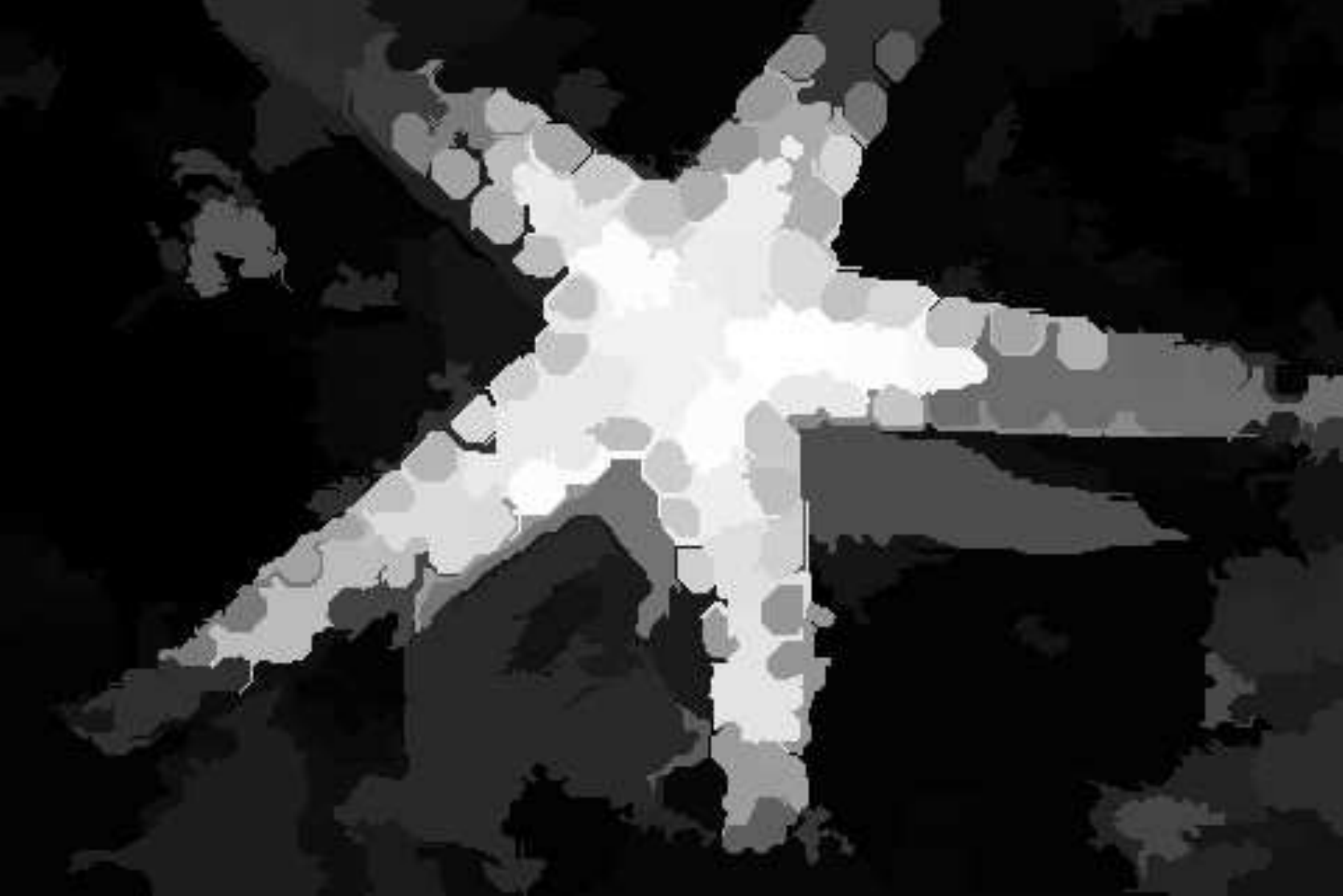}&
		\includegraphics[height=0.1\textwidth,keepaspectratio]{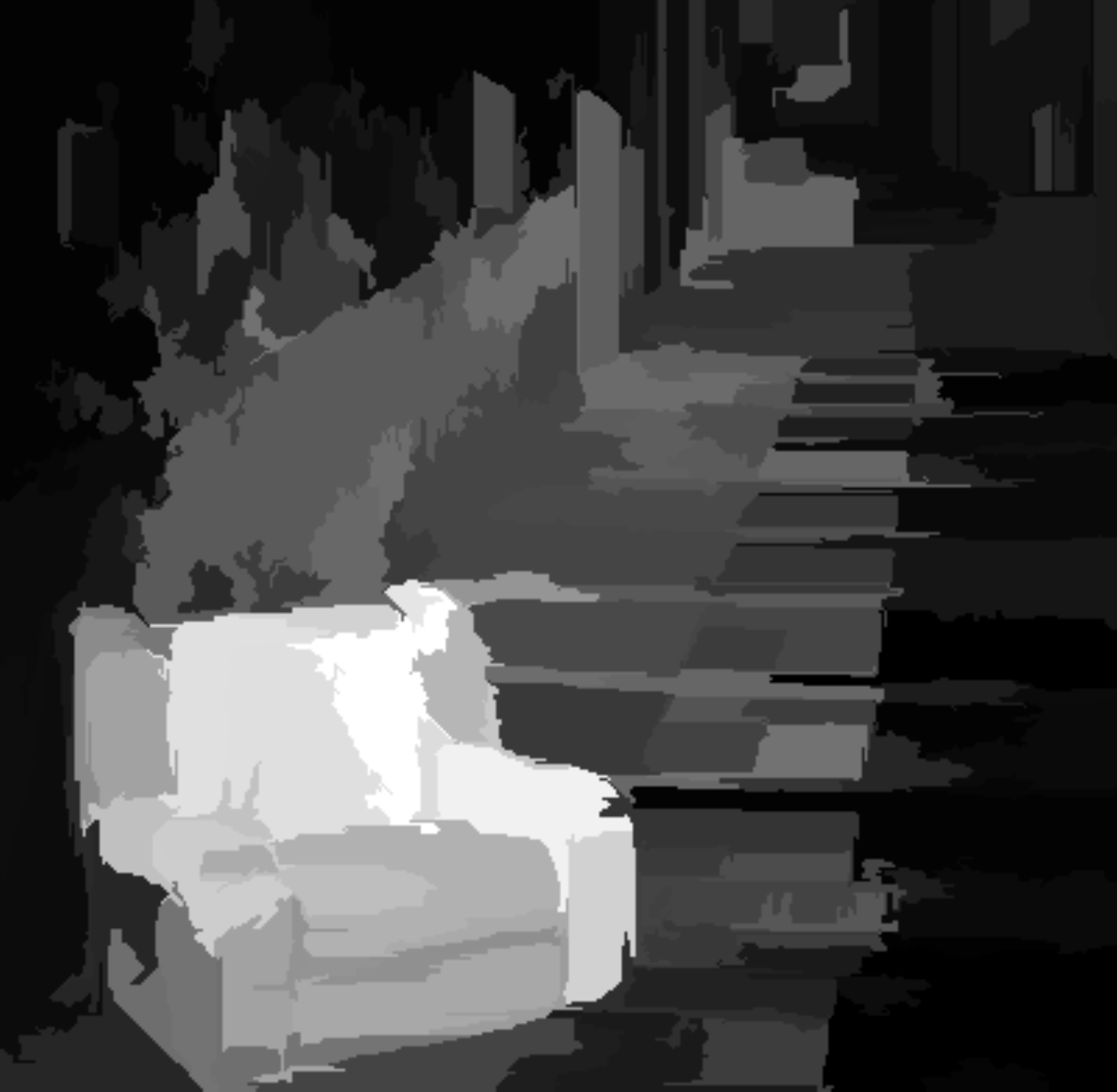}\\
	\end{tabular}
    \caption{Illustration of the most important features. From top to bottom: 
        input images, the most important contrast feature ($c_{12}$), 
        the most important backgroundness feature ($b_{12}$), the most important property feature ($p_5$), 
        and the saliency map of our approach (DRFIs) produced on a single-\layer~segmentation. Brighter area indicates larger feature value (thus larger saliency value according to $c_{12}, b_{12}$ and the saliency map).
    }\label{fig:impFeat}
\end{figure}

% To generate training samples, we first learn the similarity score of each adjacent regions. Similar regions will be grouped together in a hierarchical way. Training samples of the saliency regressor are those \emph{confident} regions in the grouping hierarchy.

We use supervised multi-level segmentation to generate training samples. We first learn the similarity score of each adjacent regions, to show the probability that the adjacent regions both belong to the salient region or the background. Similar regions will be grouped together in a hierarchical way. Training samples of the saliency regressor are those confident regions in the grouping hierarchy.

%\vspace{.1cm}\noindent\textbf{Learning the co-salient and co-background
%degree.}
By learning the similarity score, we hope that those regions from the object (or
background) are more likely to be grouped together. In specific, given an over-segmentation of an image, we connect each region
and its spatially-neighboring regions forming a set of pairs $\mathcal{P} = \{(R_i, R_j)\}$ and learn the
probability $p(a_i=a_j)$, where $a_i$ is the saliency label of the region $R_i$.
% \ie, if they are both from the salient object or the
% background. Based on such similarity metric,  
Such a
set of pairs
into two parts:
a positive part $\mathcal{P}^{+} = \{(R_i, R_j) | a_i = a_j\}$
and
a negative part $\mathcal{P}^{-} = \{(R_i, R_j) | a_i \neq a_j\}$.
Following~\cite{jiang2013probabilistic}, each region pair is described by a set of features including the regional
saliency of two regions ($2\times$93 features), the feature contrast of two
regions (similar to the regional contrast descriptor, 29 features), and the
geometry features of the superpixel boundary of two regions (similar to $p_1\sim
p_7$ in~\tabref{tab:RegionProperty}, 7 features). 
% If $R_i$ is
% likely to be salient while $R_j$ be non-salient, they tend to share low
% similarity. The feature contrast is also helpful since regions with high
% similarity are more likely to have homogeneous appearance. And the geometric
% property of the superpixel boundary may influence their simialrity as well. For
% example, those region pairs near the image border are more likely to have higher
% similarity since they are all probabily from the background. 
Given these
222-dimensional feature, we learn a boosted decision tree classifier to estimate
the similarity score of each adjacent region pair.

\newcommand{\calS}{\mathcal{S}}

Based on the learned similarity of two adjacent regions, we produce multi-\layer~segmentation $\{\calS_1^t, \calS_2^t,\ldots, \calS_M^t\}$ to gather a large mount of training samples. Specifically, denote $\mathcal{S}_0^t$ as the over-segmentation of the image generated using the graph-based image segmentation algorithm~\cite{FelzenszwalbH04}. The regions in $\calS_0^t$ are represented
by a weighted graph, which connects the spatially neighboring regions. The weight of each edge is the learned similarity of two adjacent superpixels. Similar to the pixel-wise grouping in~\cite{FelzenszwalbH04}, pairs of regions are sequentially merged
in the order of decreasing the weights of edges.
%Different from computing the visual similarity
%between regions as the weight~\cite{FelzenszwalbH04},
%our algorithm computes
%the co-salient and co-background degree,
%which is defined as the degree
%that two neighboring regions both belong to
%the salient object or the background,
%as the weight of the corresponding edge.
We change the tolerance degree of small regions,
\ie, the parameter $k$ of the approach~\cite{FelzenszwalbH04}
(see the details in~\cite{FelzenszwalbH04}) to generate the segmentations from $\calS_1^t$ to $\calS_M^t\}$. To avoid too fine groupings, we discard $\calS_i^t$ if $\frac{|\calS_0^t|}{|\calS_i^t|} > 0.6$, where $|\cdot|$ denotes the number of superpixels.

Given a set of training images with ground truth annotations and their multi-\layer~segmentation, we can collect lots of confident regions $\mathcal{R} = \{R^{(1)}, R^{(2)}, \cdots, R^{(Q)}\}$ and the corresponding saliency scores $\mathcal{A} = \{a^{(1)}, a^{(2)}, \cdots, a^{(Q)}\}$ to learn a \rforest~saliency regressor. Only confident regions are kept for training since some regions may contain pixels from both the salient object and background.
% , though we adopt the supervised approach to learn the similarity to regions to guide the grouping. 
A region is considered to be confident
if the number of pixels belonging to the salient object or the background
exceeds $80\%$ of the total number of pixels in the region.
Its saliency score is set as $1$ or $0$ accordingly. In experiments we find
that few regions of all the training examples, around $6\%$, are unconfident
and we discard them from training.

One benefit to generate multi-level segmentation is that a large amount of training samples can be gathered. In the~\secref{sec:ParaAnalysis}, we empirically analyze different settings of $M_t$ and validate our motivation to generate training samples based on multi-\layer~image segmentation. Additionally, with the guiding of learned similarity metric, there might be only a few large regions left in the high-level segmentation, which are helpful for the \rforest~regressor to learn object-level properties. However, the learned similarity is hard to generalize across datasets. This is why our approach did not perform the best on SED2 dataset in our pervious version~\cite{jiang2013salient}. To this end, we adopt the unsupervised multi-level segmentation in the testing phrase, which is also more efficient without learning the similarity score.
% Compared with the approach introduced in~\secref{sec:imgSaliencyComputation}, such a supervised approach is more computationally expensive to generate multi-\layer~segmentation. For efficiency, it is adopted for training only.

% \vspace{.1cm}\noindent\textbf{Learning the regional saliency regressor.}
\subsection{Learning the regional saliency regressor}
\label{sec:LearnSaiencyRegressor}
Our aim is to learn the regional saliency estimator
from a set of training examples. As aforementioned,
each region is described
by a feature vector $\mathbf{x}\in\mathbb{R}^d$,
composed of the regional contrast,
regional property, and regional backgroundness descriptors (\ie, $d=93$).
From the training data
$\mathcal{X} = \{\mathbf{x}^{(1)}, \mathbf{x}^{(2)}, \cdots, \mathbf{x}^{(Q)}\}$
and the saliency scores
$\mathcal{A} = \{a^{(1)}, a^{(1)}, \cdots, a^{(Q)}\}$, we learn a random forest regressor $f_r: \mathbb{R}^d\rightarrow \mathbb{R}$ which maps the feature vector of each region to a saliency score.

A \rforest~saliency regressor is an ensemble of $T$ decision trees, where each tree consists of split and leaf nodes. Each split node stores a feature index $f$ and a threshold $\tau$. Given a
feature vector $\mathbf{x}$, each split node in the tree makes a decision
based on the feature index and threshold pair $(f, \tau)$. If
$\mathbf{x}(f)<\tau$ it traverses to the left child, otherwise to the right
child. When reaching a leaf node, its stored prediction value will be given. The
final prediction of the forest is the average of the predictions over all the
decision trees.

Training a \rforest~regressor is to independently build each decision tree. For each tree, the training samples are randomly drawn \emph{with replacement}, $\mathcal{X}_t=\{\mathbf{x}^{(t_1)}, \mathbf{x}^{(t_2)}, \cdots, \mathbf{x}^{(t_Q)}\}$, $\mathcal{A}_t=\{a^{(t_1)}, a^{(t_2)}, \cdots, a^{(t_Q)}\}$, where $t_i\in[1, Q], i\in[1, Q]$. Constructing a tree is to find the pair $(f, \tau)$ for each split node and the prediction value for each leaf node. Starting
from the root node, $m$ features $\mathcal{F}_m$ are randomly chosen \emph{without replacement} from
the full feature vector. The best split will be found among these features $\mathcal{F}_m$ to maximize the splitting criterion
\begin{align}
(f^*, \tau^*) = \max_{f\in\mathcal{F}_m, \tau} &
\left(\frac{\sum_{t_i\in \mathcal{D}_l}\left(a^{(t_i)}\right)^2}{|\mathcal{D}_l|} + \frac{\sum_{t_i\in
\mathcal{D}_r}\left(a^{(t_i)}\right)^2}{|\mathcal{D}_r|}\right. \notag \\
& \left.- \frac{\sum_{t_i\in\mathcal{D}}\left(a^{(t_i)}\right)^2}{|\mathcal{D}|}\right),
\end{align}
where $\mathcal{D}_l=\{(\mathbf{x}^{(t_i)}, a^{(t_i)}) | \mathbf{x}^{(t_i)}(f)<\tau\}$,\\
$\mathcal{D}_r=\{(\mathbf{x}^{(t_i)}, a^{(t_i)}) | \mathbf{x}^{(t_i)}(f)\geq\tau\}$, and
$\mathcal{D}=\mathcal{D}_l\cup\mathcal{D}_r$. Such a spliting procedure is repeated until $|D|<5$ and a leaf node is created. The prediction value of the
leaf node is the average saliency scores of the training samples falling in it.
We will empirically examine the settings of parameters $T$ and $m$ in~\secref{sec:ParaAnalysis}.
% \begin{eqnarray*}
%  \mathcal{S}_L=\{(\mathbf{x}_i, a_i) | \mathbf{x}_i(f)<\tau\},\\
%  \mathcal{S}_R=\{(\mathbf{x}_i, a_i) | \mathbf{x}_i(f)\geq\tau\},\\
%  \mathcal{S}_r=\mathcal{S}_L\cup\mathcal{S}_R.
% \end{eqnarray*}

Learning a saliency regressor
can automatically integrate the features and
discover the most discriminative ones. Additionally, in the training procedure of the random forest,
the feature importance can be estimated simultaneously. Refer to the supplementary for more details.
\Figref{fig:feature_rank}
presents the most important $60$ features.

% \vspace{.1cm}\noindent\textbf{Learning the multi-\layer saliency fusor.}
\subsection{Learning the multi-\layer~saliency fusor}
\label{sec:LearnSaliencyFusor}
Given the multi-\layer~saliency maps
$\{\mathbf{A}_1, \mathbf{A}_2, \cdots, \mathbf{A}_M\}$ for an image,
our aim is to learn a combinator
$g(\mathbf{A}_1, \mathbf{A}_2, \cdots, \mathbf{A}_M)$
to fuse them together to form the final saliency map $\mathbf{A}$.
Such a problem has been already addressed
in existing methods, such as the conditional random field solution~\cite{LiuYSWZTS11}.
In our implementation,
we find that a linear combinator,
$\mathbf{A} = \sum_{m=1}^M w_m\mathbf{A}_m$,
performs well by learning the weights
using a least square estimator,
\ie, minimizing the sum of the losses ($\|\mathbf{A} - \sum_{m=1}^M w_m\mathbf{A}_m\|_F^2$)
over all the training images.

In practice, we found that the average of multi-\layer~saliency maps performs as nearly well as a weighted average.

\section{Experiments}
\label{sec:Experiment}
In this section, we empirically analyze our proposed approach. Comparisons with state-of-the-art methods on benchmark data sets are also demonstrated.
%\begin{table}[t]
%\begin{tabular}{|c|c|c|}
%\hline
%\textbf{Name} & \textbf{No. of Images} & \textbf{Remarks} \\
%\hline
%MSRA B~\cite{LiuYSWZTS11} & 5000 & One Object \\
%\hline
%SED1~\cite{AlpertGBB07} & 100 & One Object \\
%\hline
%SED2~\cite{AlpertGBB07} & 100 & Multiple Objects \\
%\hline
%SOD~\cite{bb34461} & 300 & One Object \\
%\hline
%iCoSeg~\cite{conf/cvpr/BatraKPLC10} & 643 & Multiple Objects \\
%\hline
%\end{tabular}
%\caption{Datasets we adopted for experiments.}
%\label{tab:dataset}
%\end{table}

\subsection{Data sets}
We evaluate the performance
over five data sets
that are widely used in salient object detection and segmentation.

\myPara{MSRA-B}\footnote{\href{http://research.microsoft.com/en-us/um/people/jiansun/}{http://research.microsoft.com/en-us/um/people/jiansun/}}
This data set~\cite{LiuYSWZTS11} includes $5000$ images,
originally containing labeled rectangles
from nine users drawing a bounding box
around what they consider the most salient object.
There is a large variation among images
including natural scenes,
animals,
indoor,
outdoor, etc.
We manually segment the salient object (contour)
within the user-drawn rectangle to
obtain binary masks.
The ASD data set~\cite{AchantaHES09}
is a subset (with binary masks) of MSRA-B, and
thus we no longer make evaluations on it.
%Link:\text{http://research.microsoft.com/en-us/um/people/jiansun/SalientObject/salient_object.htm}.

\myPara{iCoSeg}\footnote{\href{http://chenlab.ece.cornell.edu/projects/touch-coseg}{http://chenlab.ece.cornell.edu/projects/touch-coseg}}
This is a publicly available co-segmentation data set~\cite{conf/cvpr/BatraKPLC10},
including $38$ groups
of totally $643$ images.
Each image is along with a pixel-wise ground-truth annotation, which may contain one or multiple salient objects.
In this paper, we use it to evaluate the performance
of salient object detection.
% Link:~\url{http://chenlab.ece.cornell.edu/projects/touch-coseg/CMU_Cornell_iCoseg_dataset.zip}.

\myPara{SED}\footnote{\fontsize{.92em}{.92em}\selectfont
        \href{http://www.wisdom.weizmann.ac.il/~vision/Seg_Evaluation_DB/}
        {http://www.wisdom.weizmann.ac.il/$\sim$vision/Seg\_Evaluation\_DB/}} This data set~\cite{AlpertGBB07} contains two subsets:
SED1 that has $100$ images containing only one salient object
and SED2 that has $100$ images containing exactly two salient objects. Pixel-wise groundtruth annotations for the salient objects in both SED1 and SED2 are provided. We only make evaluations on SED2. Similar to the larger MSRA-B dataset, only single one salient object exists in each image in SED1, where state-of-the-art performance was reported in our previous version~\cite{jiang2013salient}. Additionally, evaluations on SED2 may help us check the adaptability of salient object detection algorithms on multiple-object cases.
% Link:~\url{http://www.wisdom.weizmann.ac.il/~vision/Seg_Evaluation_DB/index.html}

\myPara{ECSSD}\footnote{\href{http://www.cse.cuhk.edu.hk/leojia/projects/hsaliency}{http://www.cse.cuhk.edu.hk/leojia/projects/hsaliency}} To overcome the weakness of existing data set such as ASD, in which background structures are primarily simple and smooth, a new data set denoted as Extended Complex Scene Saliency Dataset (ECSSD) is proposed recently in~\cite{yan2013hierarchical}. It contains 1000 images with diversified patterns in both foreground and background, where many semantically meaningful but structurally complex images are available. Binary masks for salient objects are produced by 5 subjects.

\myPara{DUT-OMRON}\footnote{\href{http://ice.dlut.edu.cn/lu/DUT-OMRON/Homepage.htm}{http://ice.dlut.edu.cn/lu/dut-omron/homepage.htm}} Similarly, this dataset is also introduced to evaluate salient object detection algorithms on images with more than a single salient object and relatively complex background. It contains 5,168 high quality natural images, where each image is resized to have a maximum side length of 400 pixels. Annotations are available in forms of both bounding boxes and pixel-wise binary object masks. Furthermore, eye fixation annotations are also provided makeing this dataset suitable for simultaneously evaluating salient object localization and detection models as well as fixation prediction models.

\begin{figure*}
    \centering
    \renewcommand{\arraystretch}{.9}
    \renewcommand{\tabcolsep}{.5mm}
    \begin{tabular}{cccc}
      \includegraphics[width=0.245\textwidth,keepaspectratio]{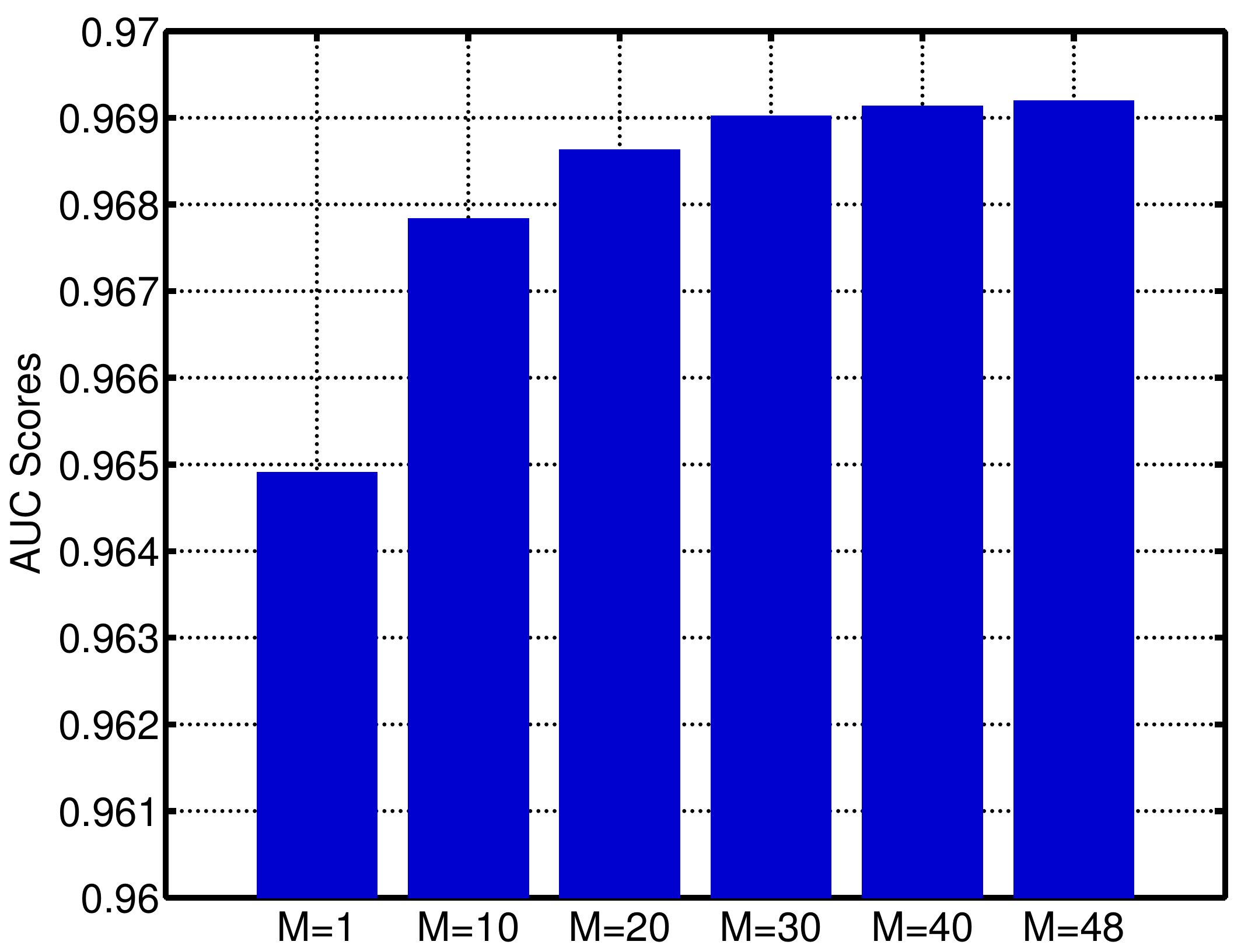} &
      \includegraphics[width=0.245\textwidth,keepaspectratio]{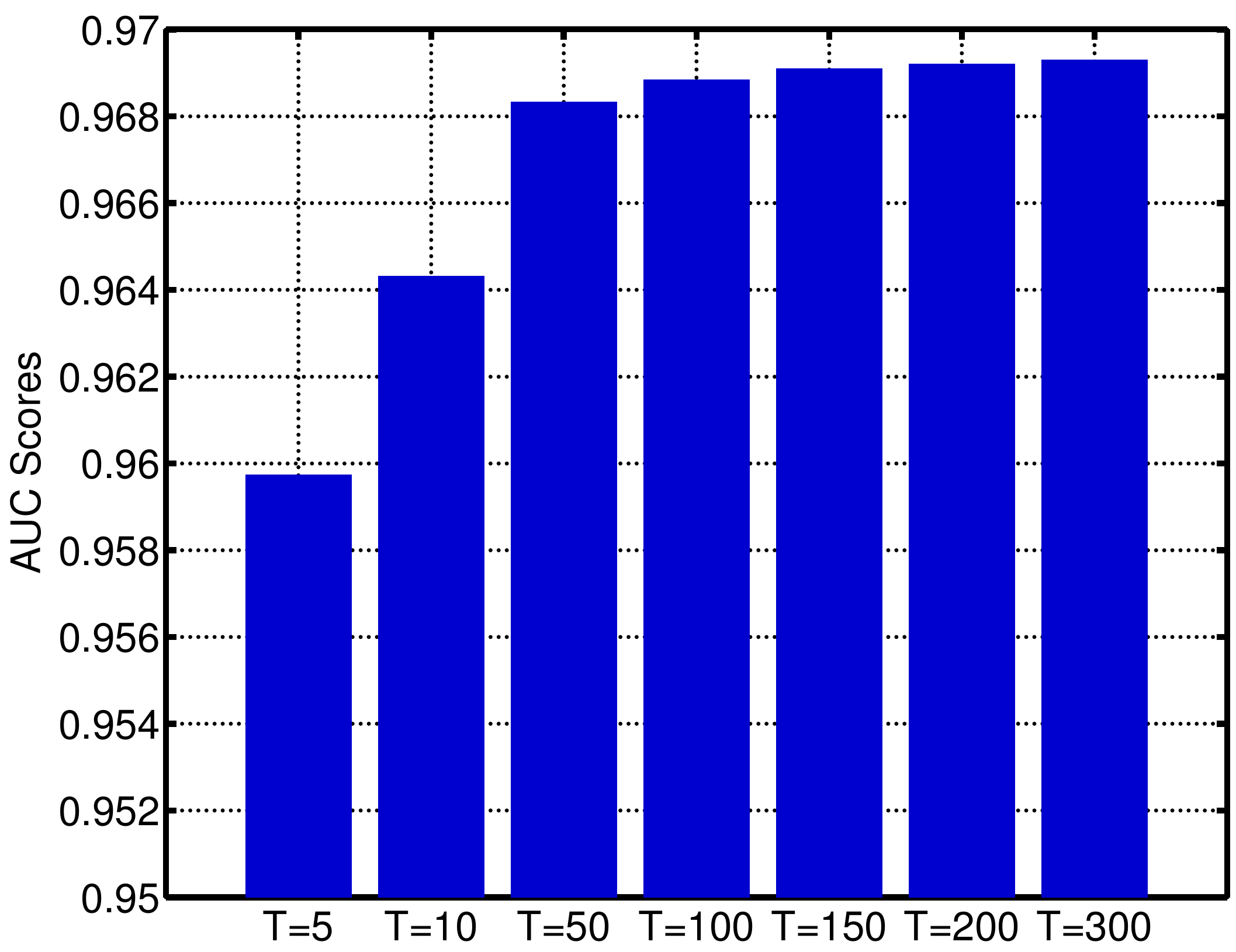} &
      \includegraphics[width=0.245\textwidth,keepaspectratio]{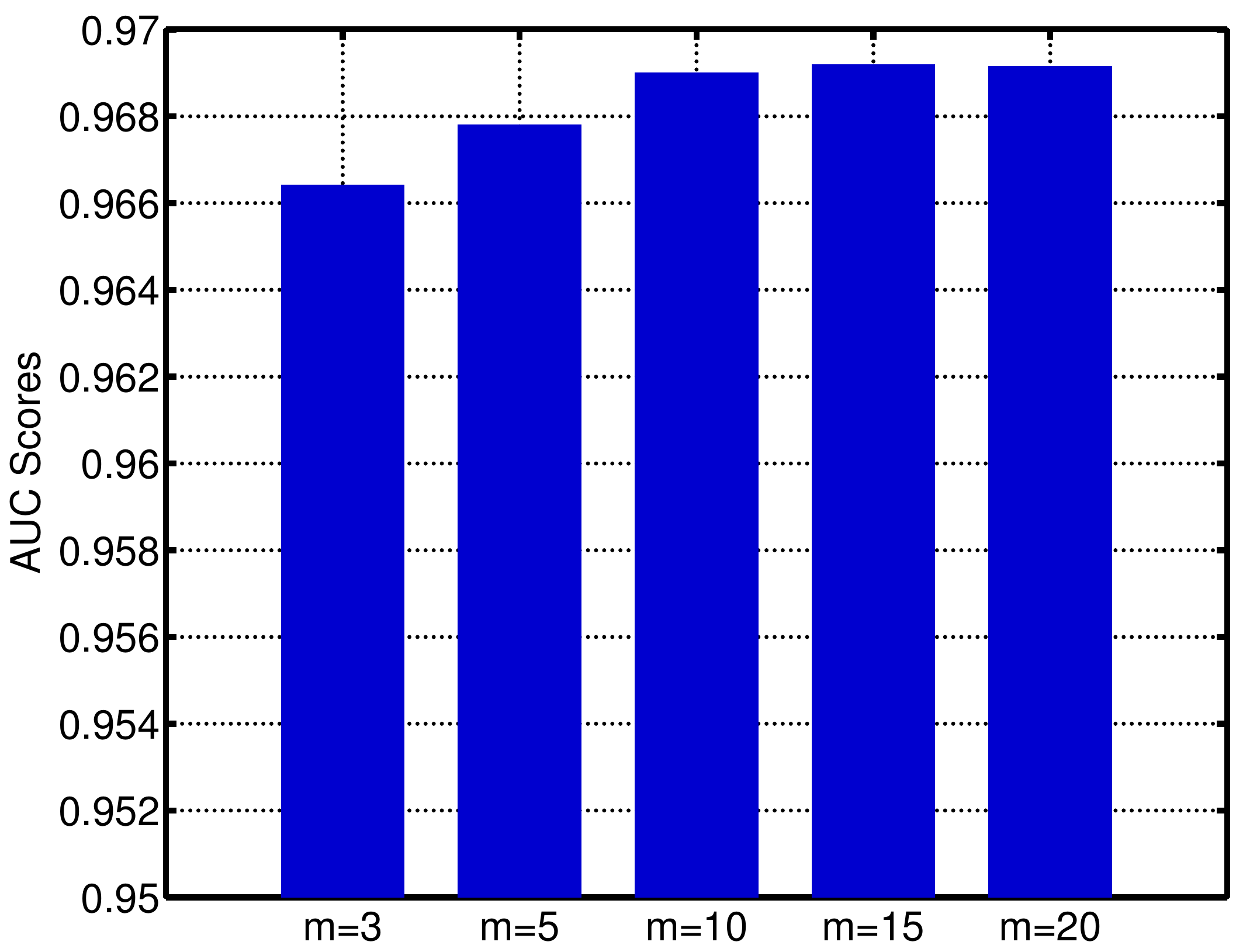} &
      \includegraphics[width=0.245\textwidth,keepaspectratio]{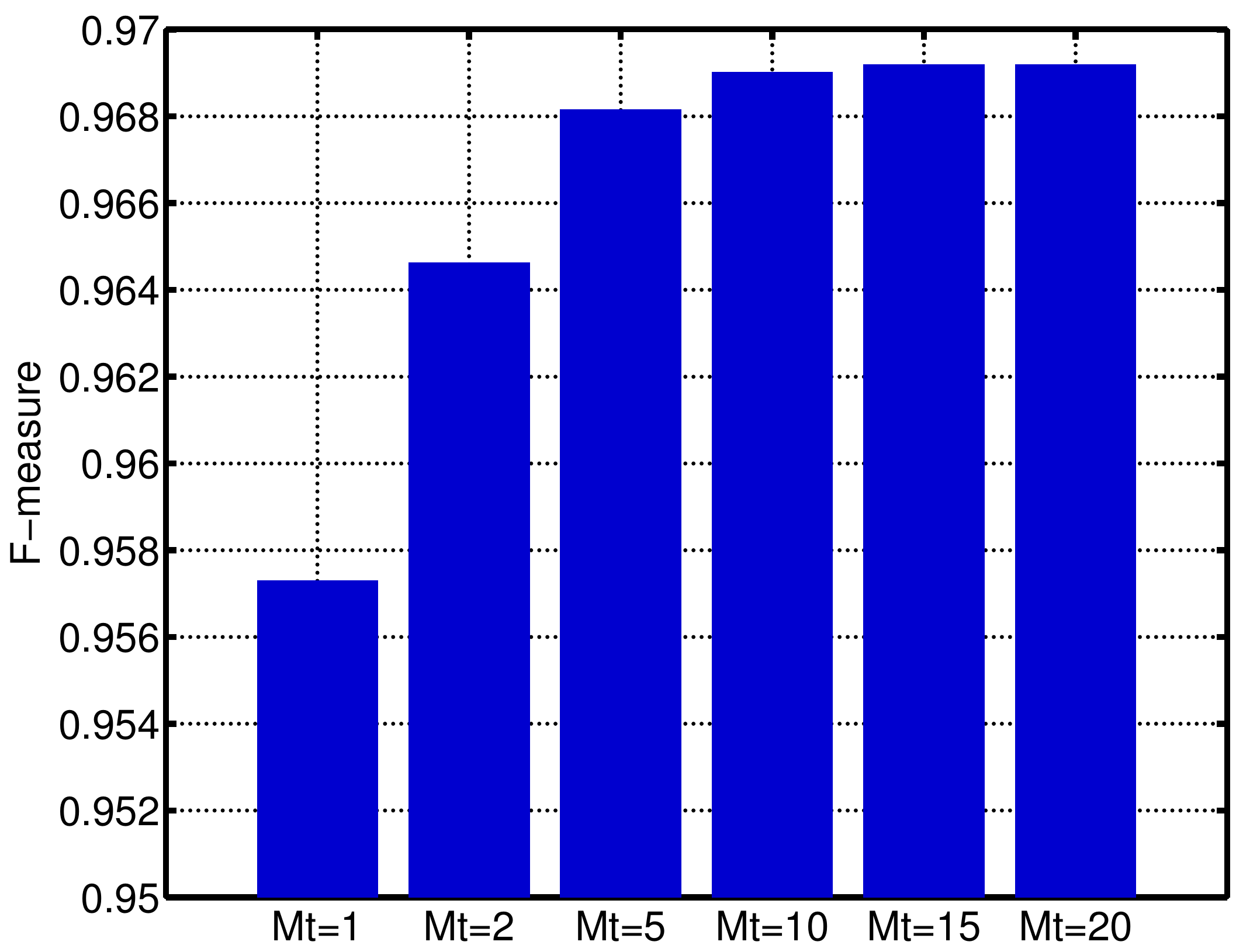} \\
      (a) & (b) & (c) & (d) \\
    \end{tabular}
\caption{Empirical analysis of parameters in terms of the AUC scores based on five-fold cross-validation of the training set. From left to right, (a) AUC scores versus the number of segmentations to generate training samples, (b)(c) AUC scores versus number of decision trees and number of randomly chosen features at each node in the Random Forest saliency regressor, and (d) number of segmentations in generating saliency maps.}
\label{fig:paraAnalysis}
\end{figure*}

\begin{figure*}
    \centering
    \renewcommand{\arraystretch}{.9}
    \renewcommand{\tabcolsep}{.5mm}
    \includegraphics[width=0.475\textwidth,keepaspectratio]{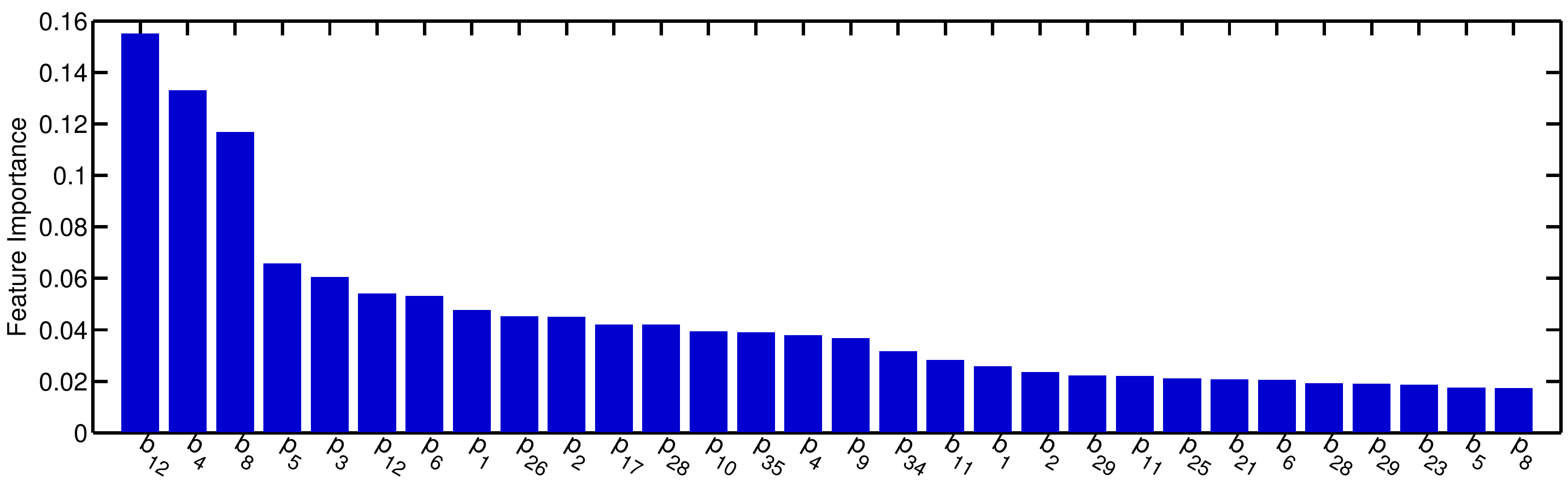}
    \includegraphics[width=0.475\textwidth,keepaspectratio]{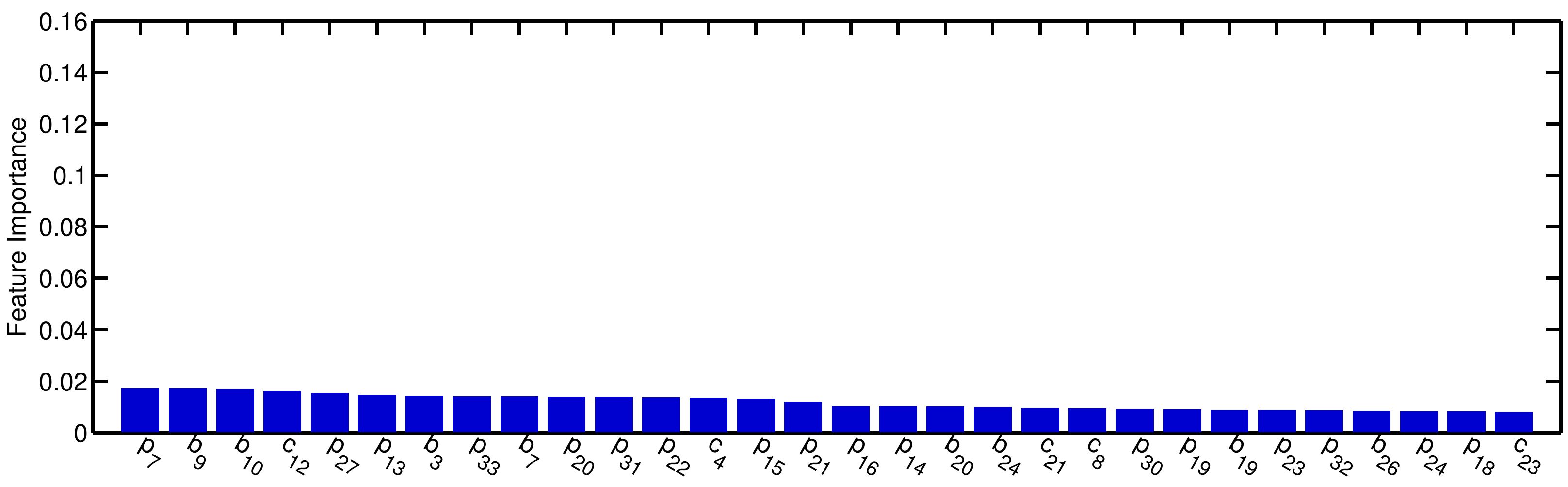}
    \caption{The most important 60 regional saliency features given by \rforest~regressor, occupying around $90\%$ of the energy of total features. There are 5 contrast features, 20 backgroundness features, and 35 property features. From left to right: the first and second 30 important features, respectively. See \tabref{tab:RegionContrastBackgroundness} and \tabref{tab:RegionProperty} for the description of the features. }
    \label{fig:feature_rank}
\end{figure*}

We randomly sample $3000$ images from the MSRA-B data set
to train our model. Five-fold cross validation is run to select the parameters.
The remaining $2000$ images are used for testing.
Rather than training a model for each data set,
we use the model trained from the MSRA-B data set
and test it over others. This can help test the adaptability to other different data sets
of the model trained from one data set
and avoid the model overfitted to a specific one.

\subsection{Evaluation Metrics}
% \vspace{.1cm}
We evaluate the performance using the measures used in~\cite{BorjiSI12b} based on the overlapping area between groundtruth annotation
and saliency prediction, including the PR (precision-recall) curve,
the ROC (receiver operating characteristic) curve and the AUC (Area Under ROC Curve) score.
Precision corresponds to the percentage
of salient pixels
correctly assigned,
and recall is the fraction
of detected salient pixels belonging to the salient object in the ground truth.

For a grayscale saliency map, whose pixel values are in the range $[0, 255]$, we vary the threshold from 0 to 255 to obtain a series of salient object segmentations. 
The PR curve is created by computing the precision and recall values at each threshold. The ROC curve can also be generated based on true
positive rates and false positive rates obtained during the calculation of the PR curve. %The AUC score is simply calculated as the area under the ROC curve.
%In addition to these overlap-based metrics, we also report the MAE (mean absolute error) score between a saliency map and the groundtruth annotation to directly measure their difference.

\subsection{Parameters Analysis}
\label{sec:ParaAnalysis}
In this section, we empirically analyze the performance of salient object detection against the settings of parameters during both training and testing phrases. Since we want to test the cross-data generalization ability of our approach, we run five-fold cross-validation on the training set. Settings of parameters are thus blind to other testing data and fair comparisons with other approaches can be conducted. Average AUC scores resulting from cross-validation under different parameter setting are plotted in Fig.~\ref{fig:paraAnalysis}.

\myPara{Training parameters analysis.} There are three parameters during training, number of segmentations $M_t$ to generate training samples, number of trees $T$ and number of randomly chosen features $m$ when training the Random Forest regressor.

Larger number of segmentations lead to larger amount of training data. As a classifier usually benefits more from greater quantity of training samples, we can observe from ~\Figref{fig:paraAnalysis}(a) that the performance steadily increase when $M_t$ becomes larger. We finally set $M_t=48$ to generate around 1.7 million samples to train our regional \rforest~saliency regressor.

As shown in~\Figref{fig:paraAnalysis}(b), the performance of our approach with more trees in the Random Forest saliency regressor is higher.
The more trees there are, the less variances are among the decision trees, and
thus the better performance can be achieved. Though the performance keeps increasing as more trees adopted, we choose to set $T=200$ trees to
train the regressor to balance the efficiency and the effectiveness.

When splitting each node during the construction a decision tree, only $m$ randomly chosen features can be observed . Intuitively, on one hand, increasing $m$ will give the node greater chance to select more discriminative features. On the other hand, however, larger $m$ will bring smaller variances between decision trees. For instance, suppose $m$ is set to the dimension of the feature vector, \ie, all of the features can be seen during the splitting, most likely the same most discriminative feature will be chosen at each split node. Consequently, nearly identical decision trees are built that can not complement each other well and thus may result in inferior performance. According to \Figref{fig:paraAnalysis}(c), we empirically set $m=15$ since the performance is the best.

\vspace{.1cm}\noindent\textbf{Testing parameters analysis.} One can see in~\Figref{fig:paraAnalysis}(d)
that the AUC scores of the saliency maps increase
when more layers of segmentations are adopted.
The reason is
that there may exist some confident regions
that cover the most (even entire) part of an object
in more layers of segmentations.
However, a larger number of segmentations introduce more computational burden.
Therefore,
to balance the efficiency and the effectiveness,
we set $M$ to $15$ segmentations in our experiments.

% \vspace{.1cm}\noindent\textbf{Feature importance.}
\subsection{Feature Importance}
Our approach
uses a wide variety of features.
In this section, we empirically analyze
the usefulness
of these regional saliency features.
\Figref{fig:feature_rank} shows the rank of the most important 60 regional features produced during the training of \rforest~regressor, which occupy around $90\%$ of the energy of total features.

The feature rank indicates that
the property descriptor is the most critical one in our feature set (occupies 35 out of top 60 features).
The reason might be that salient objects share some common properties, validating our motivation to exploit the generic regional properties for salient object detection that are widely studied in other tasks such as image classification~\cite{DBLP:journals/ijcv/HoiemEH07}.
For example, high rank of geometric features $p_5, p_3$ and $p_6$ might correspond to the compositional bias of salient objects. The importance of variance features $p_{12}$ and $p_{26}$, on the other hand, might be related with the background properties. Among the contrast-based descriptors, regional contrast descriptor is the least important. Since it might be affected by cluttered scenes and less important compared with the regional backgroundness descriptor which is in some sense more robust. Moreover, we also observe that color features are much more discriminative than texture features.

\begin{figure}[t]
\centering
    \includegraphics[width=0.45\textwidth,keepaspectratio]{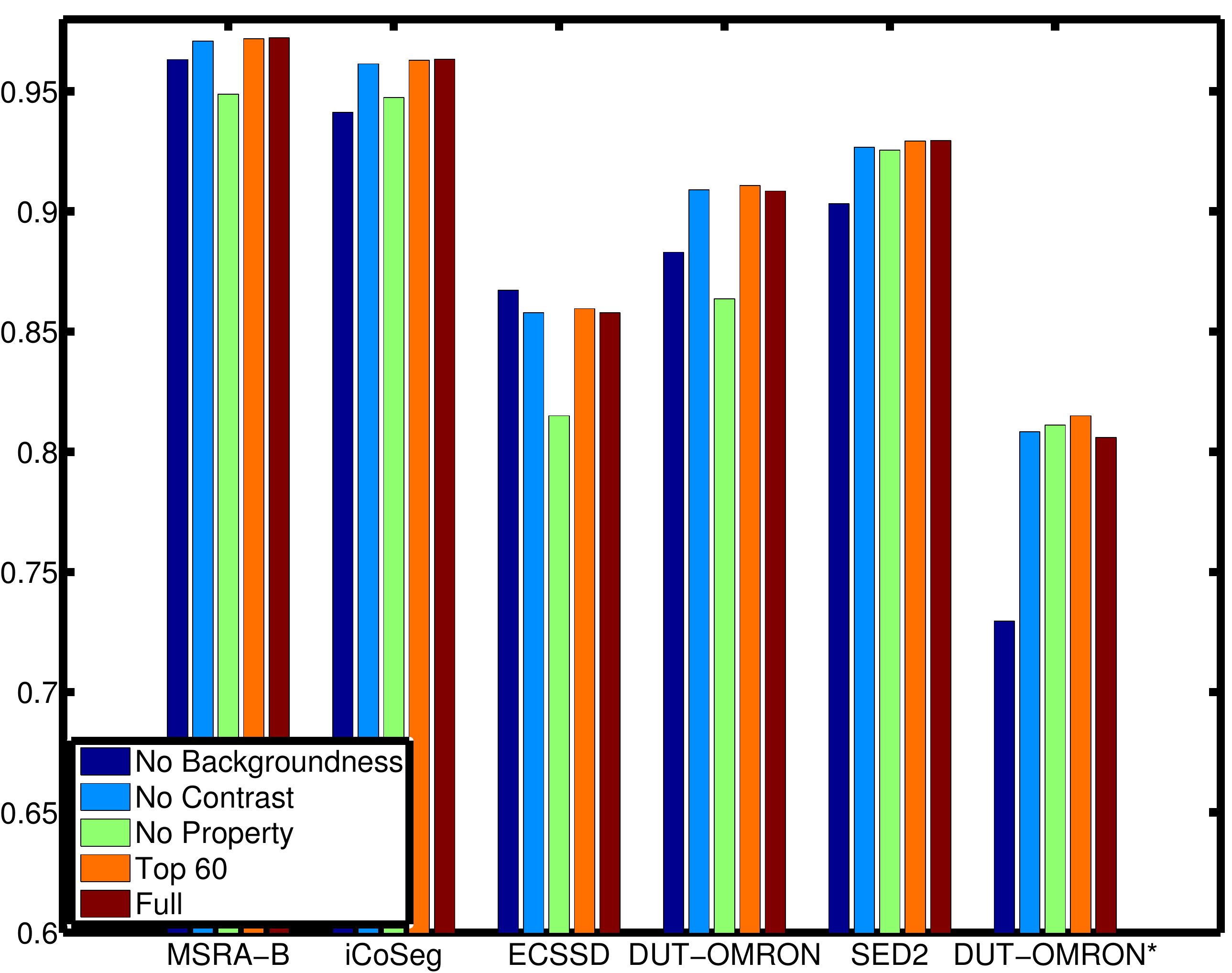}
\caption{Feature importance across different data sets. For each data set, we report the AUC scores of saliency maps by removing each kind of descriptor to see the performance drop. Additionally, we also demonstrate the performance exploiting only the top 60 features shown in~\Figref{fig:feature_rank}.}
\label{fig:featImpDataset}
\end{figure}

To further validate the importance of features across different data sets, we train classifiers by removing each kind of feature descriptor on each benchmark data set (testing set of MSRA-B). AUC scores of saliency maps are demonstrated in ~\Figref{fig:featImpDataset}. As can be seen, removing some feature descriptors does not necessarily lead to performance decrease. Consistent with the feature rank given by the \rforest regressor, regional contrast descriptor is the least important one on most of the benchmark data sets as least performance drop are observed with its removal on most of the data sets. Regional property descriptor still plays the most important role on MSRA-B, ECSSD and DUT-OMRON. Since there are multiple salient objects in an image in SED2 and iCoSeg, the common properties learned from the training data, where only a single salient object exists in most of the images, may not perform the best. The backgroundness descriptor performs well on MSRA-B, SED2, and iCoSeg. However, it plays the least important role on ECSSD data set. Its removal even leads to better performance, which indicates the pseudo-background assumption might not always hold well. Finally, instead of considering all of the 93 features, we also adopt only the top 60 features for training. Surprisingly, this feature vector performs as well as the entire feature descriptor, even slightly better on DUT-OMRON, implying some features contribute little.

We visualize the most important features of each descriptor in ~\Figref{fig:impFeat}. As we can see, even the most powerful backgroundness feature provide far less accurate information of salient objects. By integrating all of the weak information, much better saliency maps can be achieved. Note that we do not adopt the multi-\layer~fusion enhancement. Another advantages of our approach is the automatic fusion of features. For example, the rules to employ geometric features are discovered from the training data instead of heuristically defined as previous approaches~\cite{JiangWYLZL11}, which might be poor to generalize.

% \newcommand{\addPlots}[1]{\includegraphics[width=0.32\linewidth]{#1}}
% \begin{figure*}[t!]
%     \centering
%     \renewcommand{\arraystretch}{.1}
%     \renewcommand{\tabcolsep}{.1mm}
%     \small
%     \begin{tabular*}{\linewidth}{cccc}
%         \begin{sideways} ~~~~~~~~~~~~~ (a) MSRA-B \end{sideways} &
%         \addPlots{smap_comp_pr_MSRA} &
%         \addPlots{smap_comp_roc_MSRA} &
%         \addPlots{smap_comp_auc_MSRA} \\
%         \begin{sideways} ~~~~~~~~~~~~~ (b) iCoSeg \end{sideways} &
%         \addPlots{smap_comp_pr_iCoSeg} &
%         \addPlots{smap_comp_roc_iCoSeg} &
%         \addPlots{smap_comp_auc_iCoSeg} \\
%         \begin{sideways} ~~~~~~~~~~~~~ (c) ECSSD \end{sideways} &
%         \addPlots{smap_comp_pr_ECSSD} &
%         \addPlots{smap_comp_roc_ECSSD} &
%         \addPlots{smap_comp_auc_ECSSD} \\
%         \begin{sideways} ~~~~~~~~~~ (d) DUT-OMRON \end{sideways} &
%         \addPlots{smap_comp_pr_DUT-OMRON} &
%         \addPlots{smap_comp_roc_DUT-OMRON} &
%         \addPlots{smap_comp_auc_DUT-OMRON} \\
%         \begin{sideways} ~~~~~~~~~~~~~~~ (e) SED2 \end{sideways} &
%         \addPlots{smap_comp_pr_SED2} &
%         \addPlots{smap_comp_roc_SED2} &
%         \addPlots{smap_comp_auc_SED2} \\
%         & (i) Precision-Recall curves   & (ii) ROC curves & (iii) AUC scores
%     \end{tabular*}
%     \caption{Quantitative comparison of saliency maps produced
%         by different approaches on different data sets.
%         See supplemental materials for more evaluations.
%     }\label{fig_smap_comp}
% \end{figure*}

\begin{figure}
\centering
    % \fontsize{8.5}{1em}\selectfont
    % \renewcommand{\arraystretch}{.5}
    \renewcommand{\tabcolsep}{.8mm}
    \begin{tabular}{|l||c|c|c|c|c|c|} 
	\hline
	 & \scriptsize{MSRA-B} & \scriptsize{iCoSeg} & \scriptsize{ECSSD} & \scriptsize{DUT-OMRON} & \scriptsize{SED2} & \scriptsize{DUT-OMRON*}\\
	\hline \hline
	\textbf{SVO} & 0.899 & 0.861 & 0.799 & 0.866 & 0.834 & \color{blue}{0.793}\\
	\textbf{CA} & 0.860 & 0.837 & 0.738 & 0.815 & 0.854 & 0.760\\
	\textbf{CB} & 0.930 & 0.852 & 0.819 & 0.831 & 0.825 & 0.624\\
	\textbf{RC} & 0.937 & 0.880 & 0.833 & 0.859 & 0.840 & 0.679\\
	\textbf{SF} & 0.917 & 0.911 & 0.777 & 0.803 & 0.872 & 0.715\\
	\textbf{LRK} & 0.925 & 0.908 & 0.810 & 0.859 & 0.881 & 0.758\\
	\textbf{HS} & 0.930 & 0.882 & 0.829 & 0.860 & 0.820 & 0.735\\
	\textbf{GMR} & 0.942 & 0.902 & 0.834 & 0.853 & 0.831 & 0.646\\
	\textbf{PCA} & 0.938 & 0.895 & 0.817 & 0.887 & \color{green}{0.903} & 0.776\\
	\textbf{MC} & 0.951 & 0.898 & 0.849 & 0.887 & 0.863 & 0.715\\
	\textbf{DSR} & \color{green}{0.956} & 0.921 & \color{blue}{0.856} & \color{blue}{0.899} & 0.895 & 0.776\\
	\textbf{RBD} & 0.945 & \color{blue}{0.941} & 0.840 & 0.894 & 0.873 & 0.779\\
	\textbf{DRFIs} & \color{blue}{0.954} & \color{green}{0.944} & \color{green}{0.858} & \color{green}{0.910} & \color{blue}{0.902} & \color{green}{0.804}\\
	\textbf{DRFI} & \color{red}{0.971} & \color{red}{0.968} & \color{red}{0.875} & \color{red}{0.931} & \color{red}{0.933} & \color{red}{0.822}\\
\hline
\end{tabular}

    \caption{AUC: area under ROC curve (larger is better). The best three results are highlighted with {\color{red}red}, {\color{green}green}, and {\color{blue}blue} fonts, respectively.}
    \label{tab:AUC}
\end{figure}

% \begin{figure*}[t]
%     \renewcommand{\arraystretch}{.9}
%     \renewcommand{\tabcolsep}{.5mm}
%     \centering
%     \begin{tabular}{ccccc}
%         \includegraphics[height=0.16\textwidth,keepaspectratio]{smap_comp_pr_DUT-OMRON-hard}&
%         \includegraphics[height=0.16\textwidth,keepaspectratio]{smap_comp_roc_DUT-OMRON-hard}&
%         \includegraphics[height=0.16\textwidth,keepaspectratio]{smap_comp_auc_DUT-OMRON-hard}&
%         \includegraphics[height=0.16\textwidth,keepaspectratio]{smap_comp_mae_DUT-OMRON-hard}&
%         \includegraphics[height=0.16\textwidth,keepaspectratio]{avgDutOmronHard}\\
%     \end{tabular}
%     \caption{Quantatitive comparions on selected images from DUT-OMRON. 
%         From left to right: PR curves, AUC curves, AUC scores, MAE scores, 
%         and average annotation map over all the images.
%     }\label{fig:quanCompHard}
% \end{figure*}

% \begin{table*}
% \begin{minipage}[t]{0.45\linewidth}
%     \centering
%     % \fontsize{8.5}{1em}\selectfont
%     % \renewcommand{\arraystretch}{.5}
%     \renewcommand{\tabcolsep}{.8mm}
%     \input{auc.tex}
%     \caption{AUC: area under ROC curve (larger is better).}\label{fig:AUC}
% \end{minipage}
% \hspace{0.9cm}
% \begin{minipage}[t]{0.45\linewidth}
%     \centering
%     % \fontsize{8.5}{1em}\selectfont
%     % \renewcommand{\arraystretch}{.5}
%     \renewcommand{\tabcolsep}{.8mm}
%     \input{mae.tex}
%     \caption{MAE: Mean Absolute Error (smaller is better).}\label{fig:MAE}
% \end{minipage}
% \end{table*}

\subsection{Performance Comparison}
We report both quantitative and qualitative comparisons of our approach with
state-of-the-art methods. To save the space, we only
consider the top four models ranked in the
survey~\cite{BorjiSI12b}: SVO~\cite{ChangLCL11},
CA~\cite{GofermanZT10}, CB~\cite{JiangWYLZL11}, and
RC~\cite{ChengZMHH11} and recently-developed methods:
SF~\cite{PerazziKPH12}, LRK~\cite{ShenW12}, HS~\cite{YanXSJ2013},
GMR~\cite{YangZLRY2013}, PCA~\cite{MargolinTZ2013}, MC~\cite{Jiang2013Saliency}, DSR~\cite{li2013saliency}, RBD~\cite{zhu2014saliency}
that are not covered in~\cite{BorjiSI12b}. Note that we compare our approach with the extended version of RC. In total, we make comparisons with 12 approaches. Additionally, we also report the performance of our DRFI approach with a single layer (DRFIs).

\myPara{Quantitative comparison.} Quantitative comparisons are shown in~\tabref{tab:AUC},~\Figref{fig:smapCompPR} and~\Figref{fig:smapCompROC}.
As can be seen,
our approach (DRFI) consistently outperforms others on all benchmark data sets with large margins in terms of AUC scores, PR and ROC curves. In specific, it improves by $1.57\%$, $2.66\%$, $2.34\%$, $3.45\%$ and $3.21\%$ over the best-performing state-of-the-art algorithm according to the AUC scores on MSRA-B, iCoSeg, ECSSD, DUT-OMRON, and SED2, respectively.

Our single-\layer~version (DRFIs) performs best on iCoSeg, ECSSD, and DUT-OMRON as well. It improves by $0.32\%$, $0.23\%$, and $1.22\%$ over the best-performing state-of-the-art method in terms of AUC scores on these three data sets, respectively. It is slightly worse (but still one of the top 3 best models) on MSRA-B and SED2 data set. Such improvement is substantial by considering the already high performance of state-of-the-art algorithms. More importantly, though the \rforest~regressor is trained on MSRA-B, it performs best on other challenging data sets like ECSSD and DUT-OMRON.

With the multi-\layer~enhancement, performance of our approach can be further improved. For instance, it improves by $1.22\%$ on MSRA-B and $1.78\%$ on DUT-OMRON.

% Regarding MAE scores, our approach performs the best on four out of total six benchmark datasets. On both MSRA-B and DUT-OMRON, it ranks as third in terms of MAE scores. The reason might be that post-processing steps are adopted by other two best methods (Bayesian integration in DSR and quadratic optimization in RBD). Instead, our approach utilizes the linear combination without any post-processing. The last row of~\Figref{fig_smap_comp},
% corresponding to the SED2 data set,
% shows that the application of our approach is not limited to cases with a single salient object. It can also successfully detect multiple salient objects. This is also validated by the quantitative results on SOD and iCoSeg data sets, where an image may also contain one or multiple objects. On a recent benchmark data set ECSSD, our approach wins over all four metrics, which improves by $2.35\%$ and $3.08\%$ over the second and third best algorithms in terms of AUC scores, respectively. Additionally, it reduces the MAE scores by $0.14\%$ and $0.67\%$ over the second and third best approaches, respectively. Overall, the improvement of our proposed approach over state-of-the-arts are substantial when considering their performance and especially the adaptability of our model to different data sets.

\newcommand{\addPlots}[1]{\includegraphics[width=0.32\linewidth]{#1}}
\begin{figure*}[t!]
    \centering
    \renewcommand{\arraystretch}{.1}
    \renewcommand{\tabcolsep}{.1mm}
    \small
    \begin{tabular*}{\linewidth}{ccc}
        \addPlots{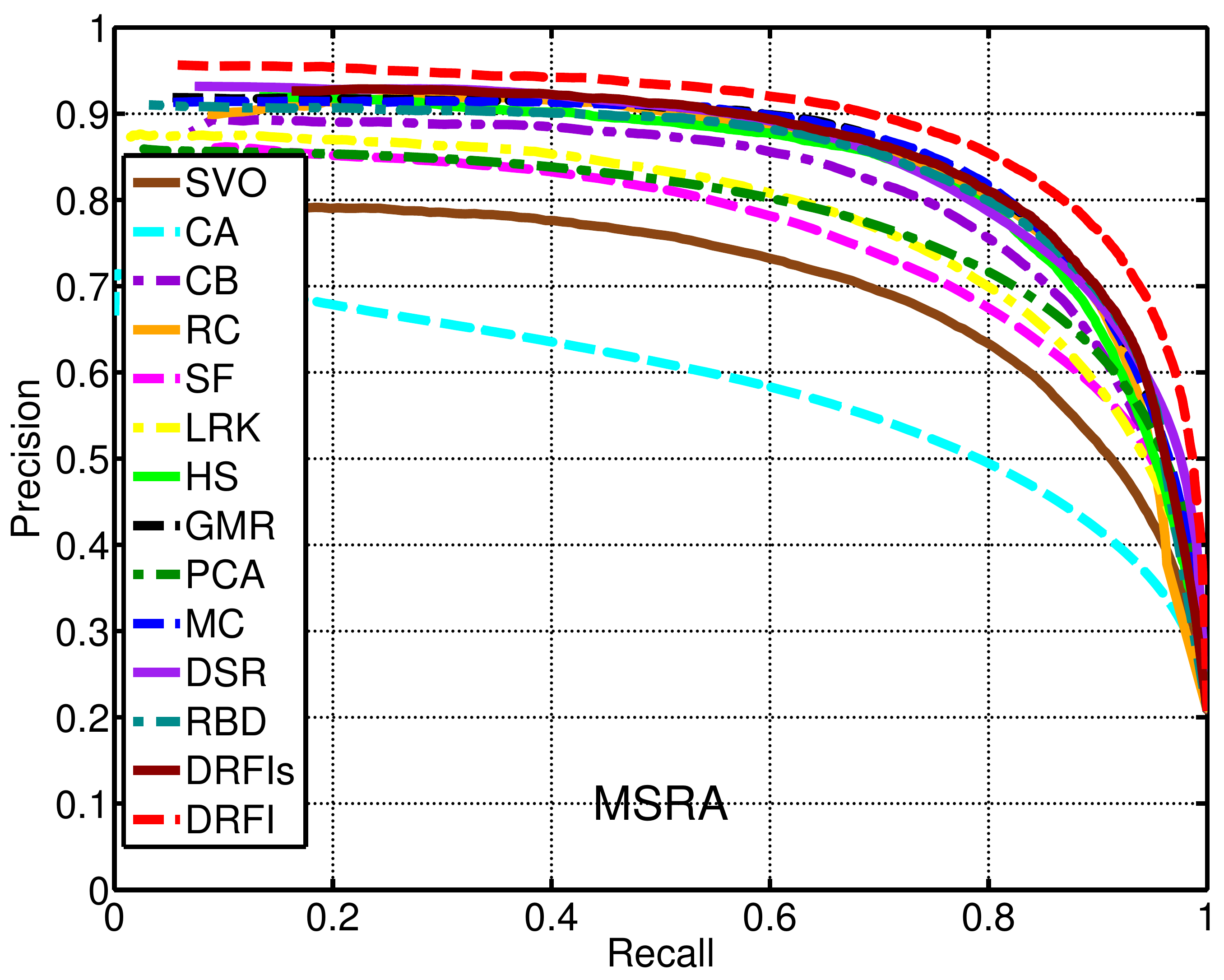} &
        \addPlots{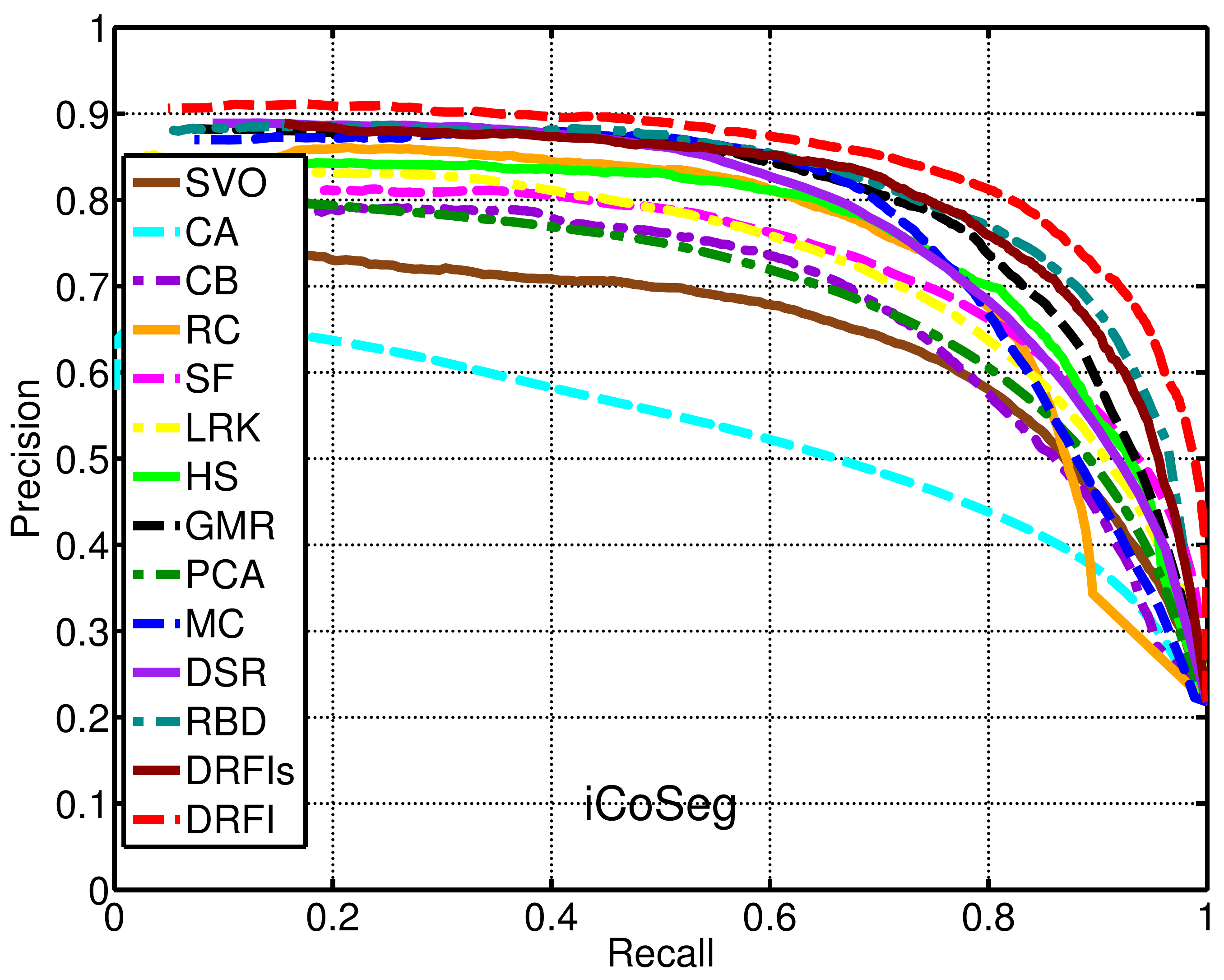} &
        \addPlots{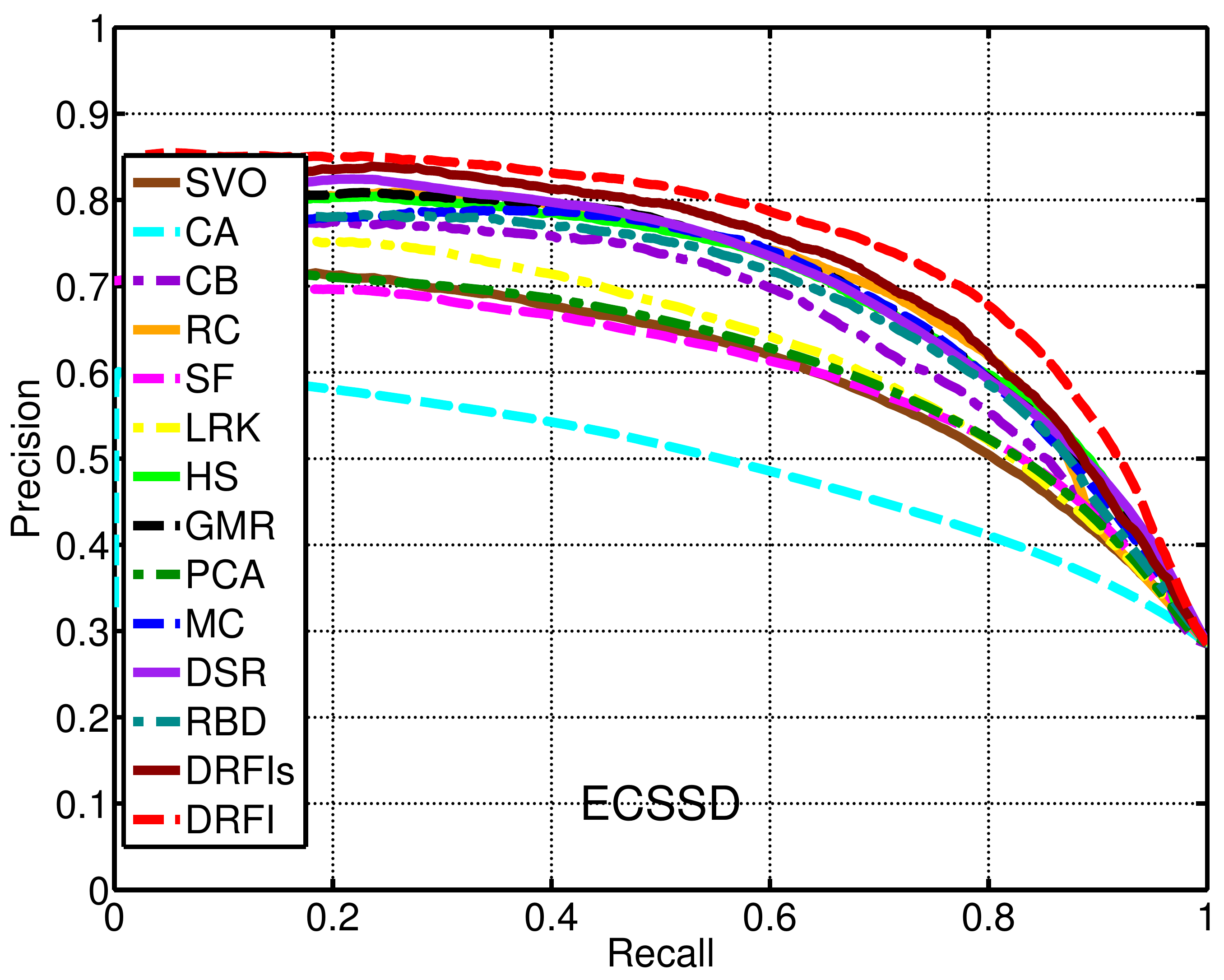} \\
        \addPlots{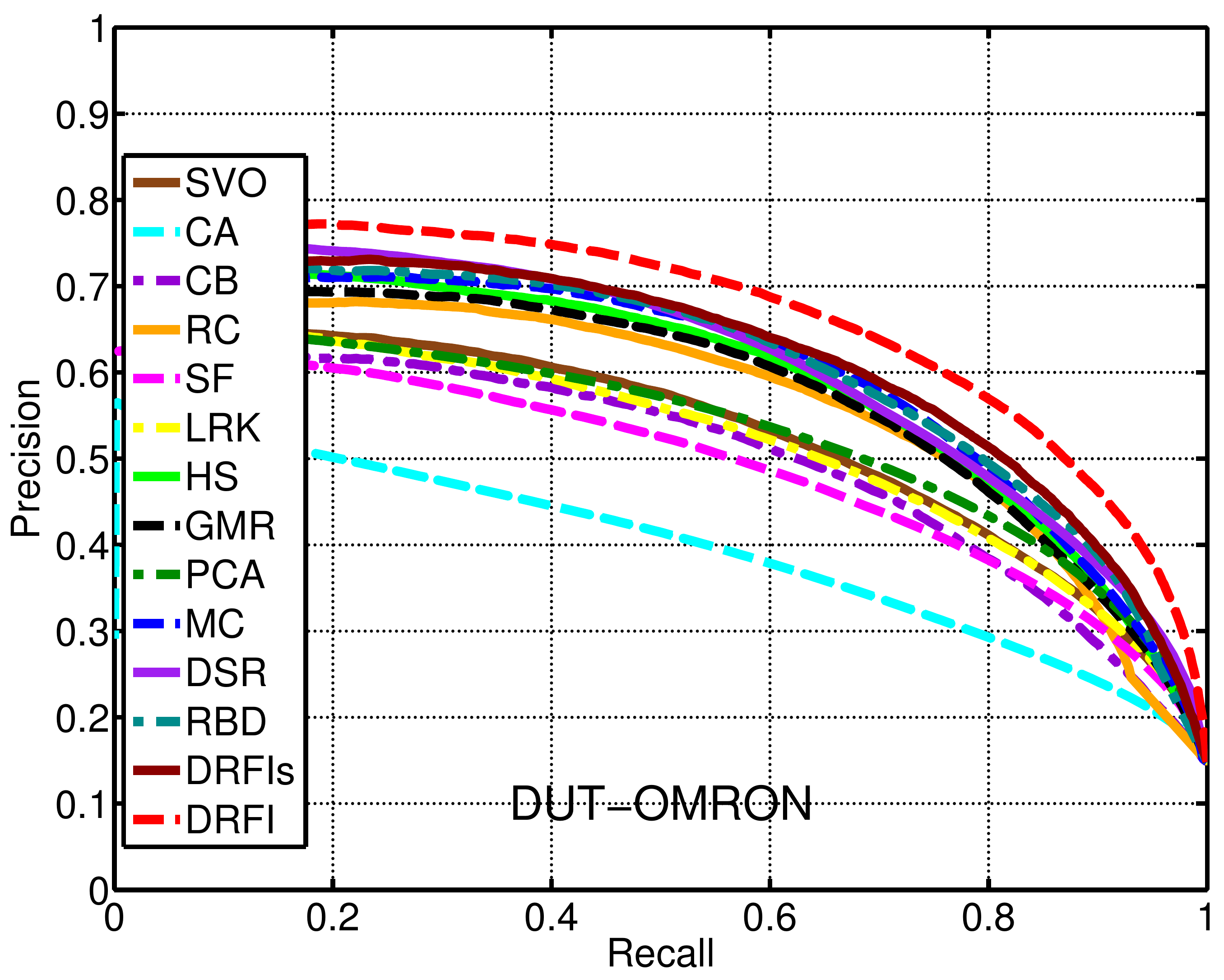} &
        \addPlots{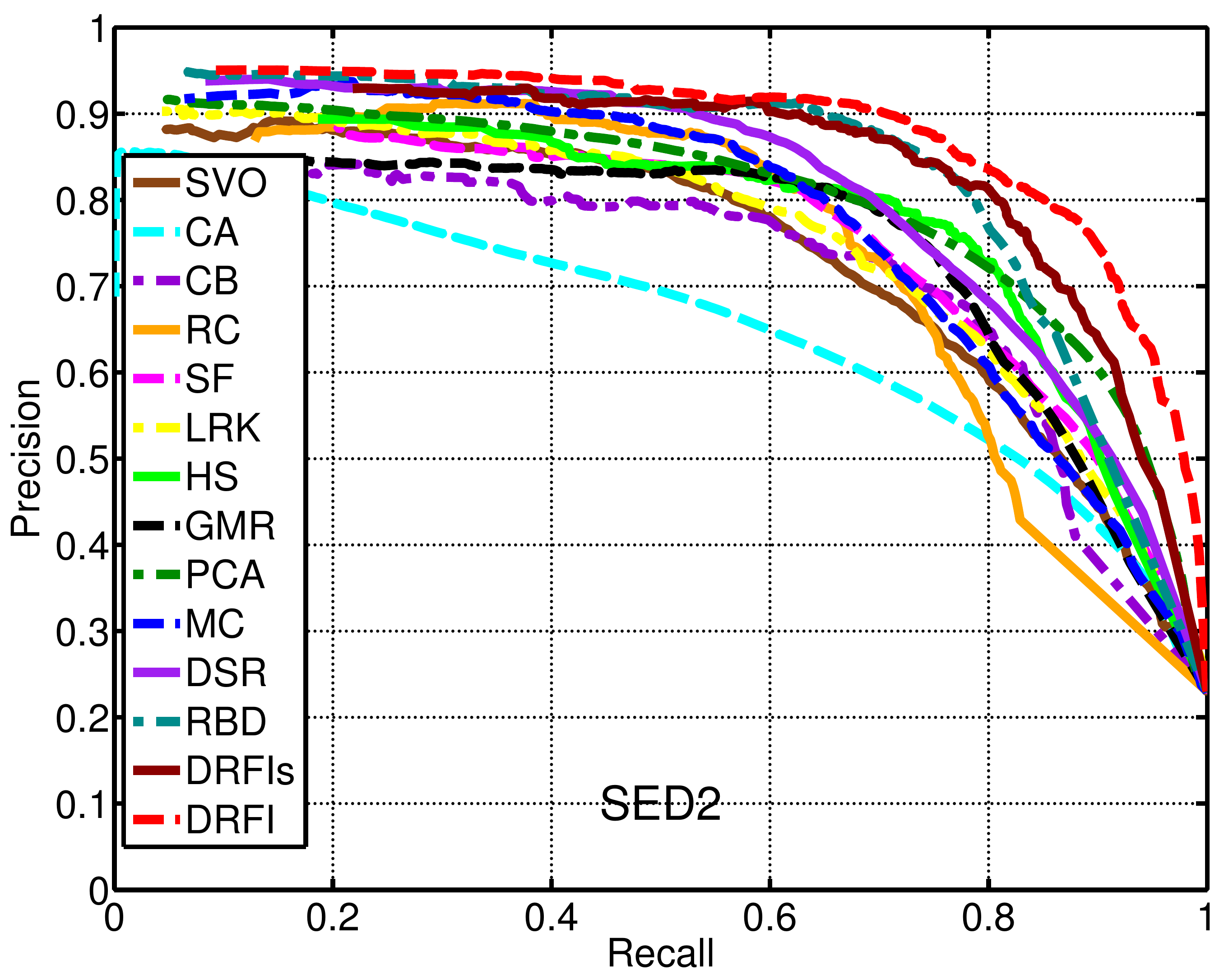} &
        \addPlots{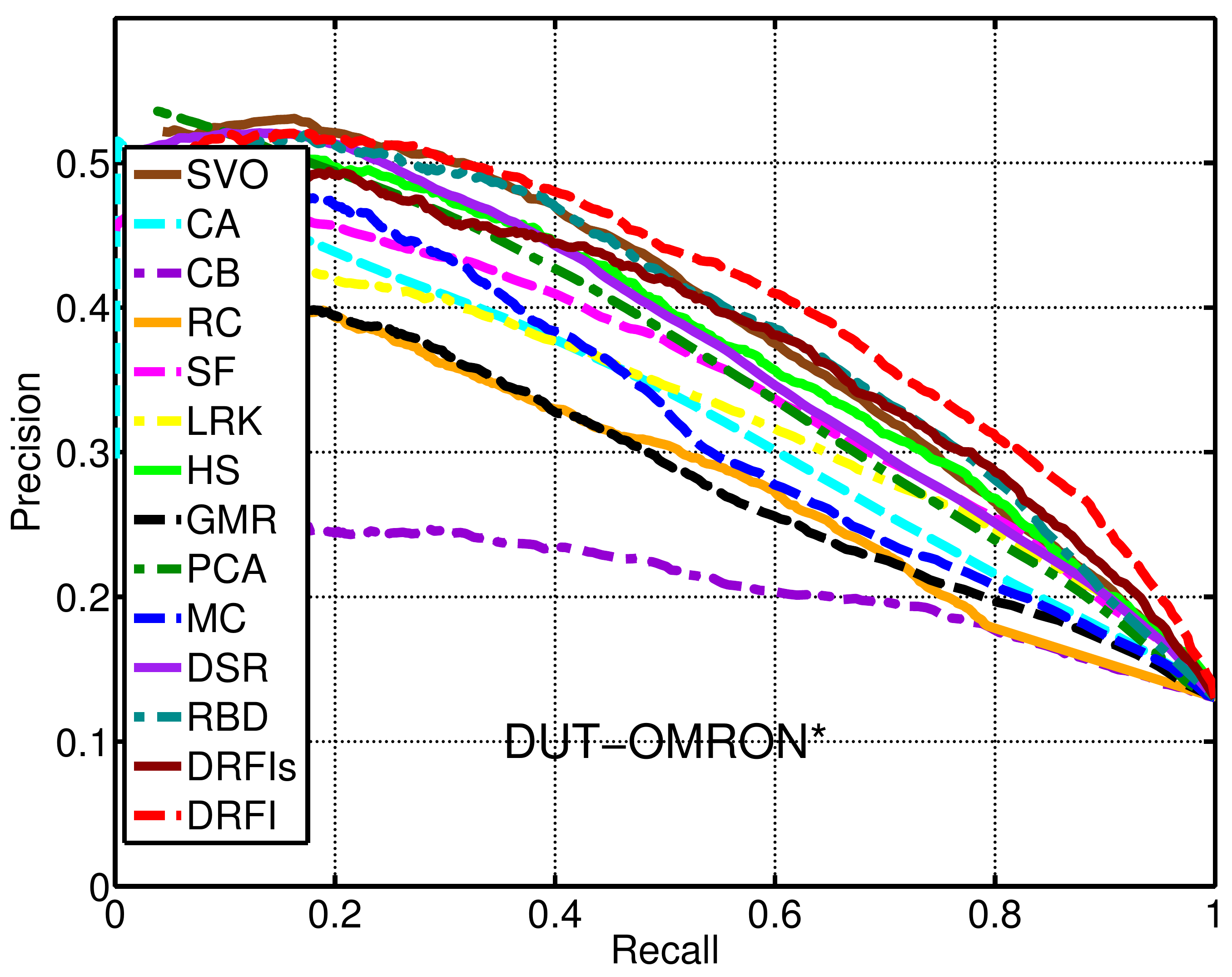} \\
    \end{tabular*}
    \caption{Quantitative comparisons of saliency maps produced
        by different approaches on different data sets in terms of PR curves.
        See supplemental materials for more evaluations.
    }\label{fig:smapCompPR}
\end{figure*}
\begin{figure*}[t!]
    \centering
    \renewcommand{\arraystretch}{.1}
    \renewcommand{\tabcolsep}{.1mm}
    \small
    \begin{tabular*}{\linewidth}{ccc}
        \addPlots{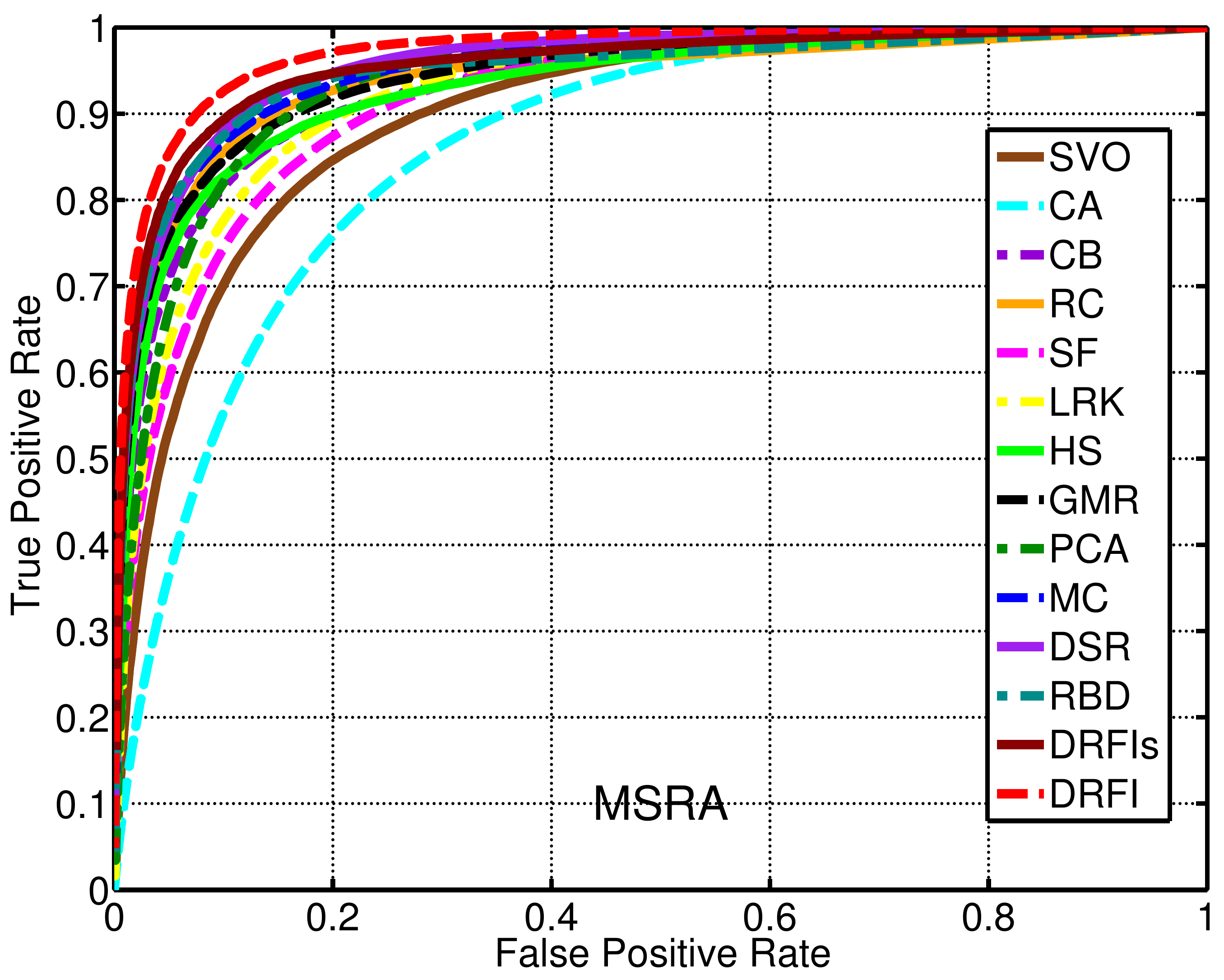} &
        \addPlots{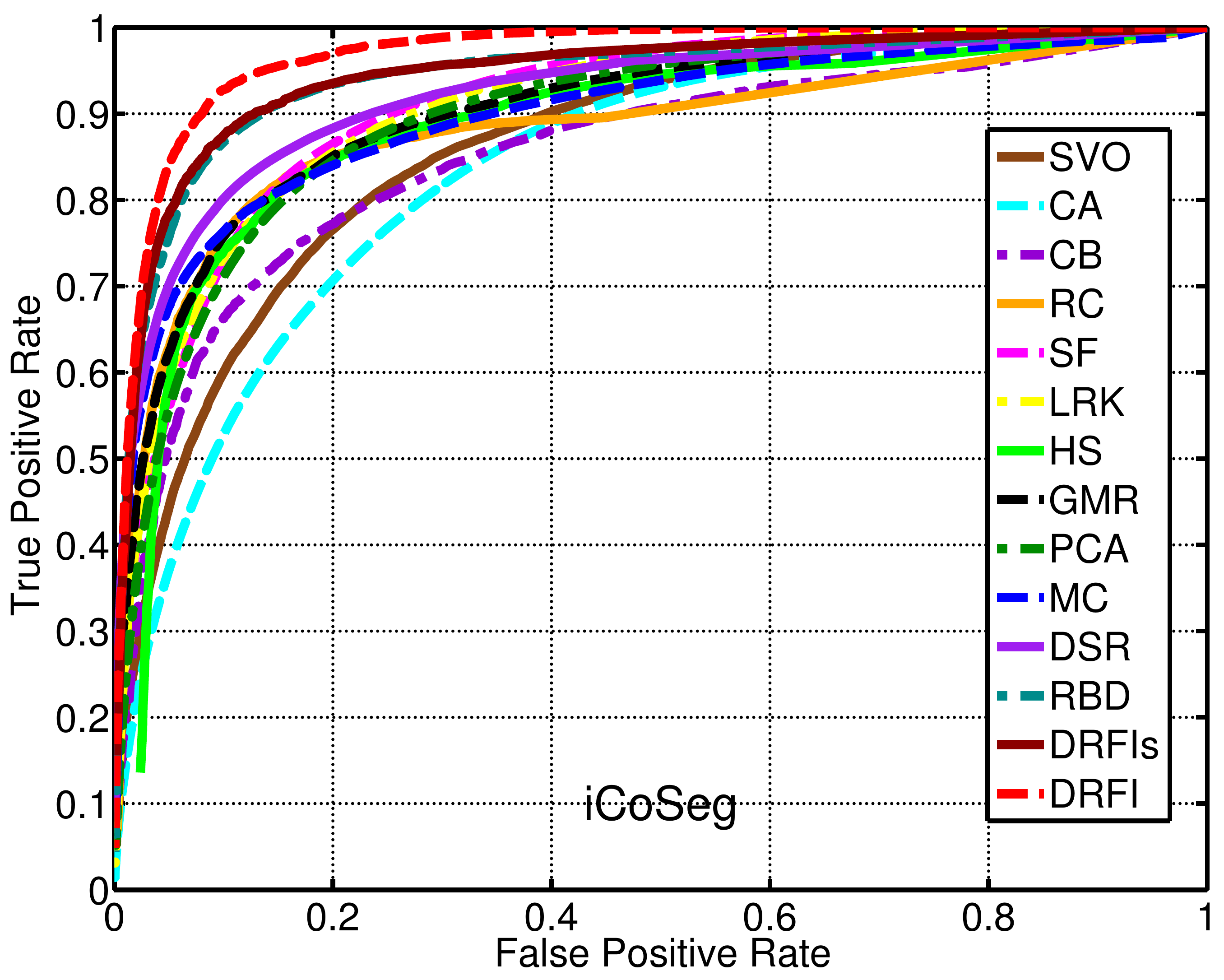} &
        \addPlots{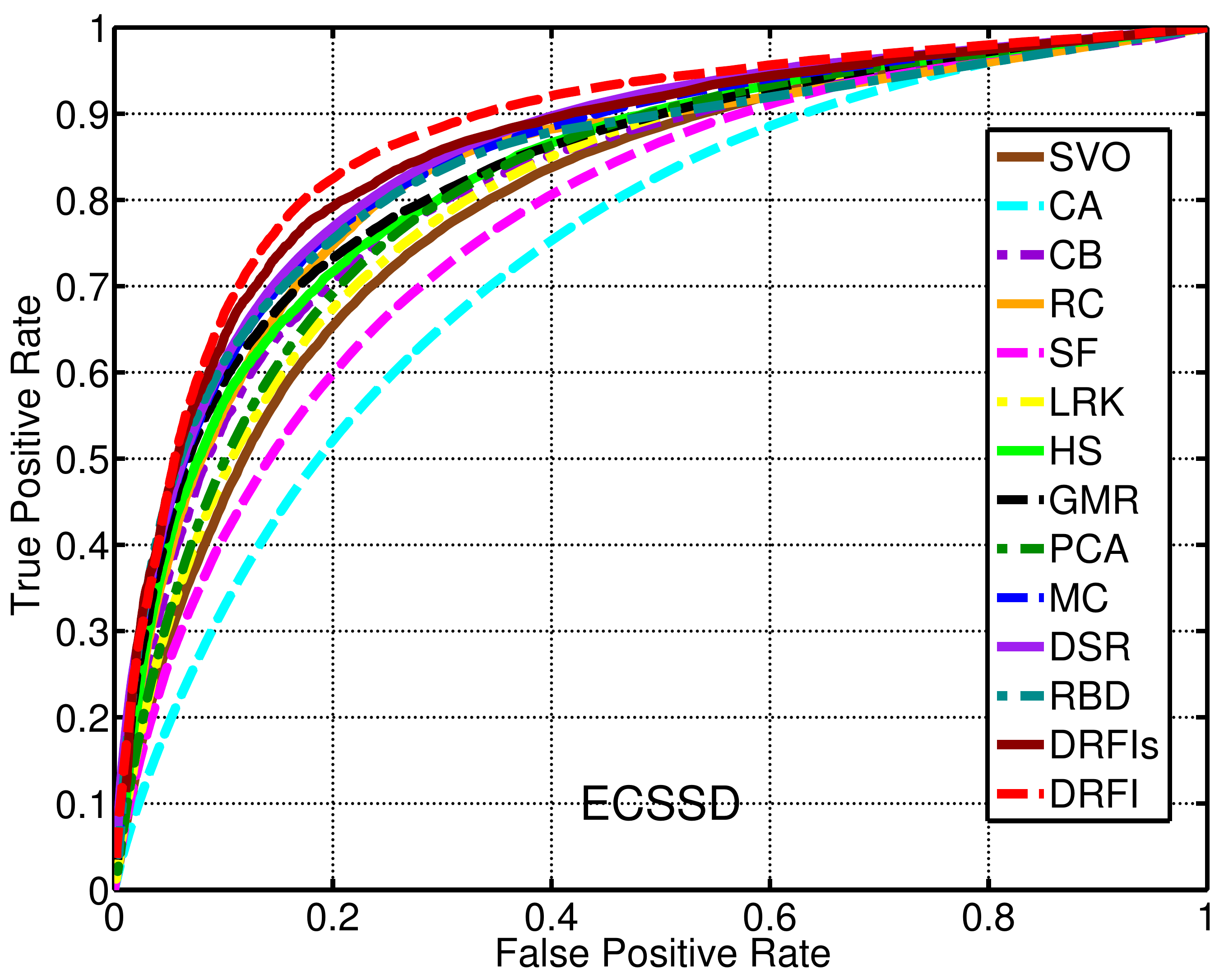} \\
        \addPlots{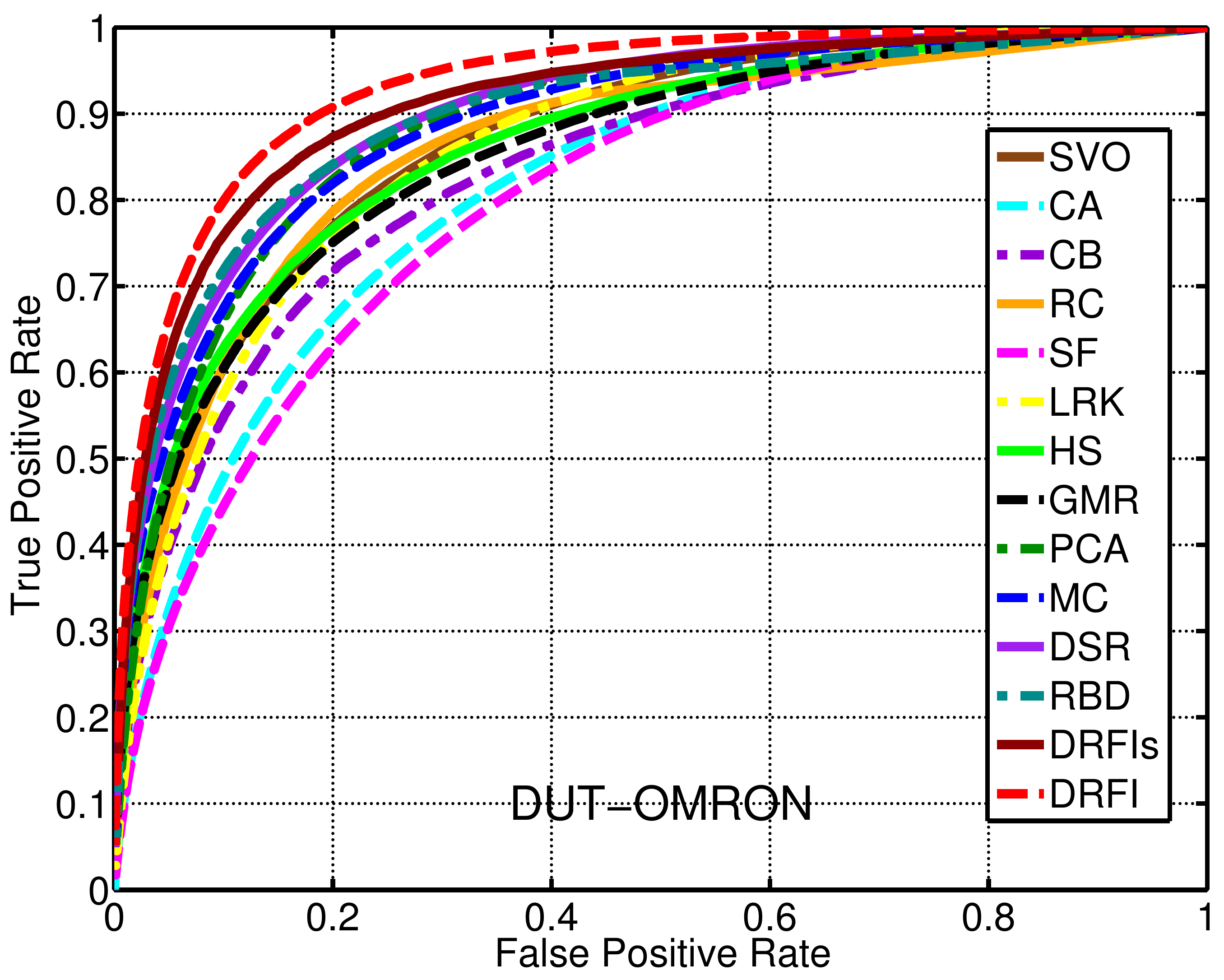} &
        \addPlots{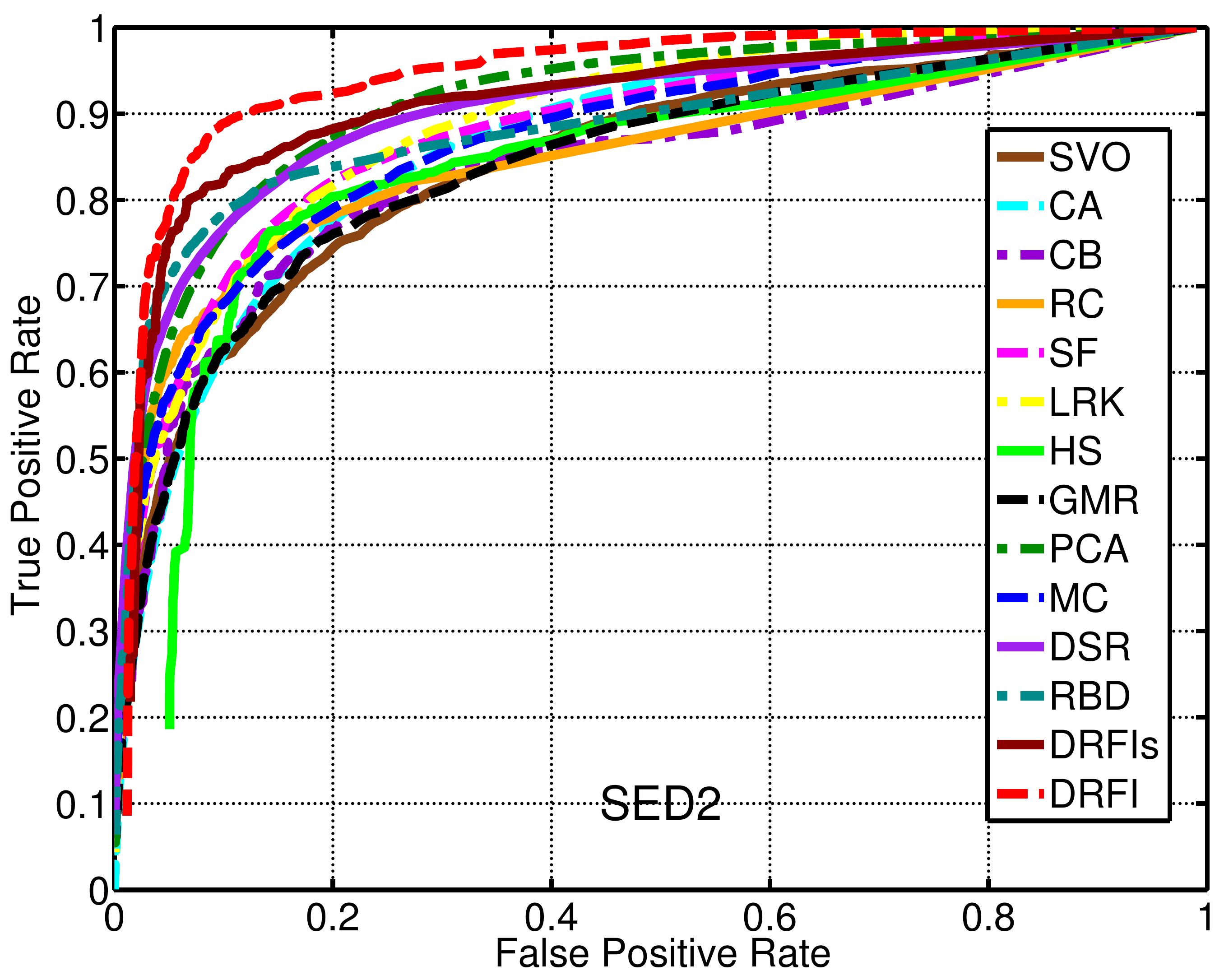} &
        \addPlots{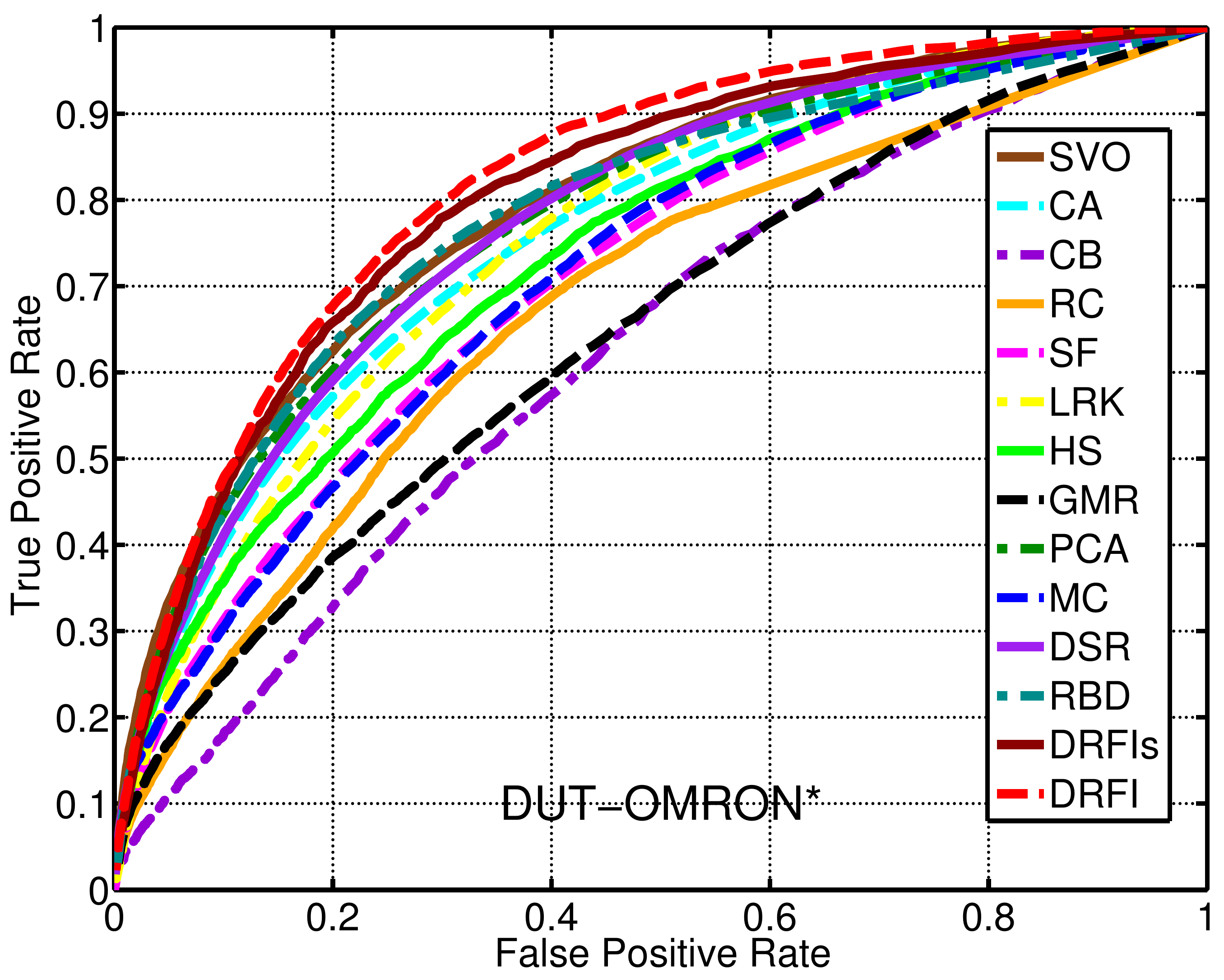} \\
    \end{tabular*}
    \caption{Quantitative comparison of saliency maps produced
        by different approaches on different data sets in terms of ROC curves.
        See supplemental materials for more evaluations.
    }\label{fig:smapCompROC}
\end{figure*}

\newcommand{\addExample}[1]{\includegraphics[width=0.06\textwidth,keepaspectratio]{#1}}

\begin{figure*}[!htbp]
    \renewcommand{\arraystretch}{0.8}
    \renewcommand{\tabcolsep}{.4mm}
    \centering
    \footnotesize
\begin{tabular}{ccccccccccccccc}
	\addExample{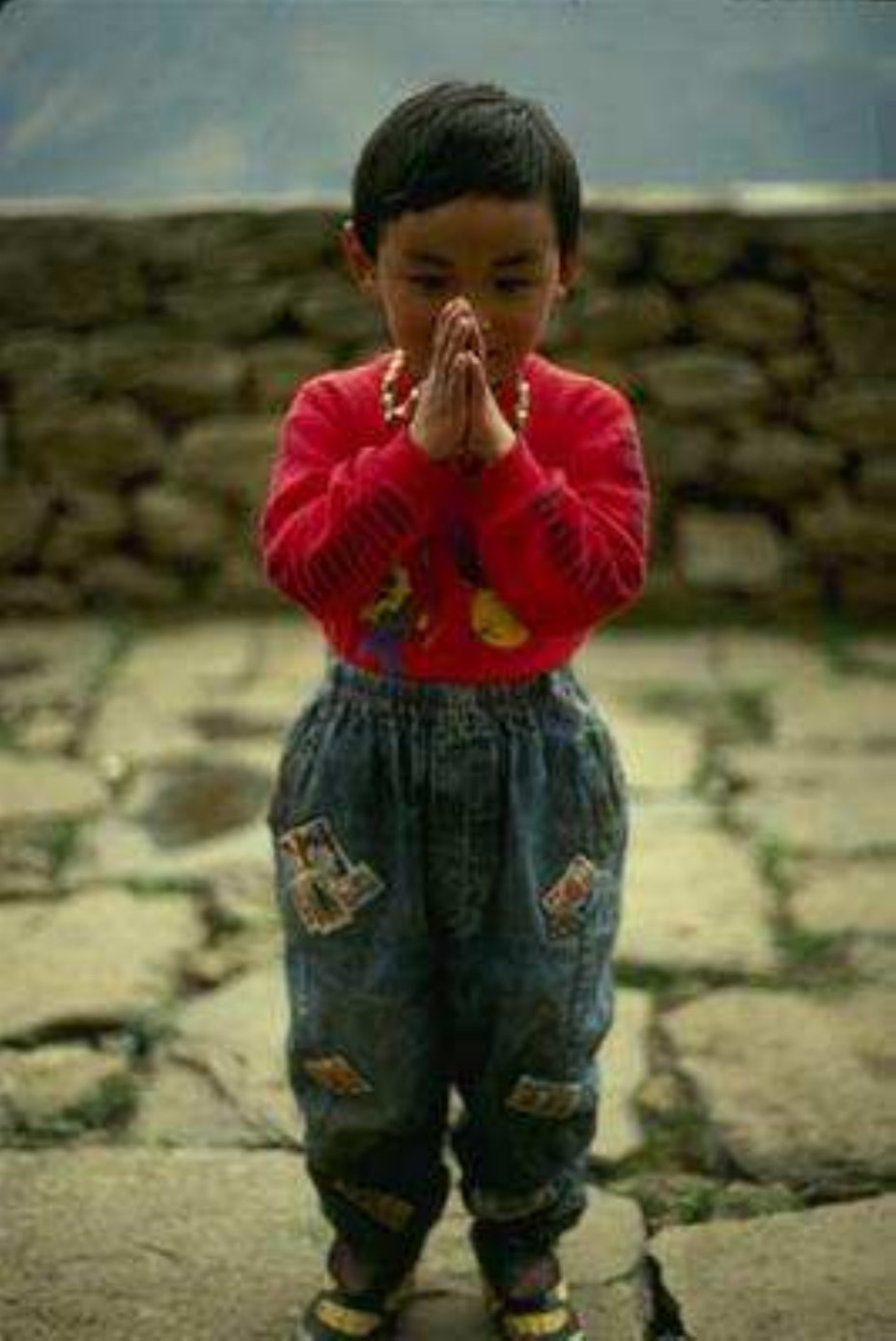}&
	\addExample{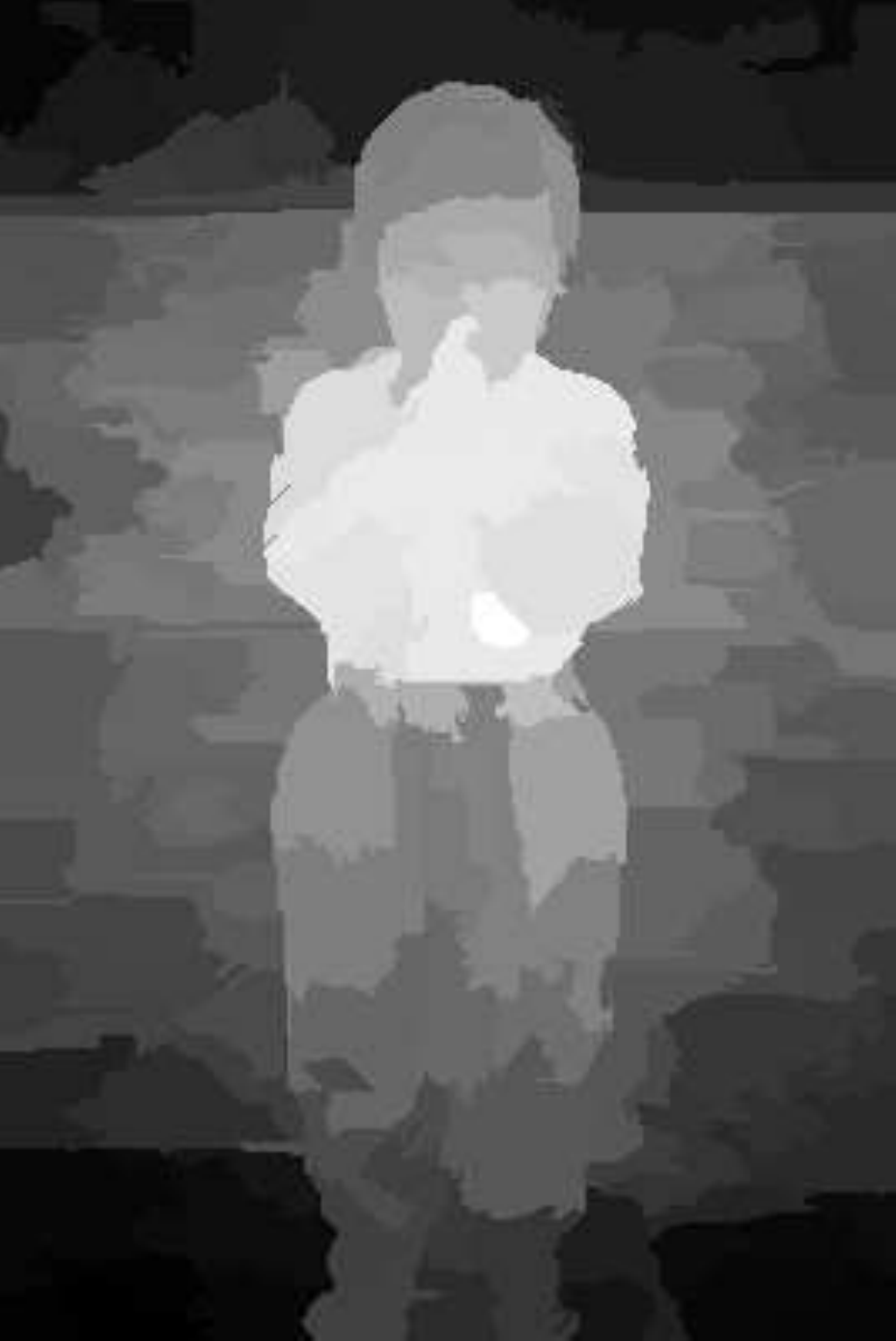}&
	\addExample{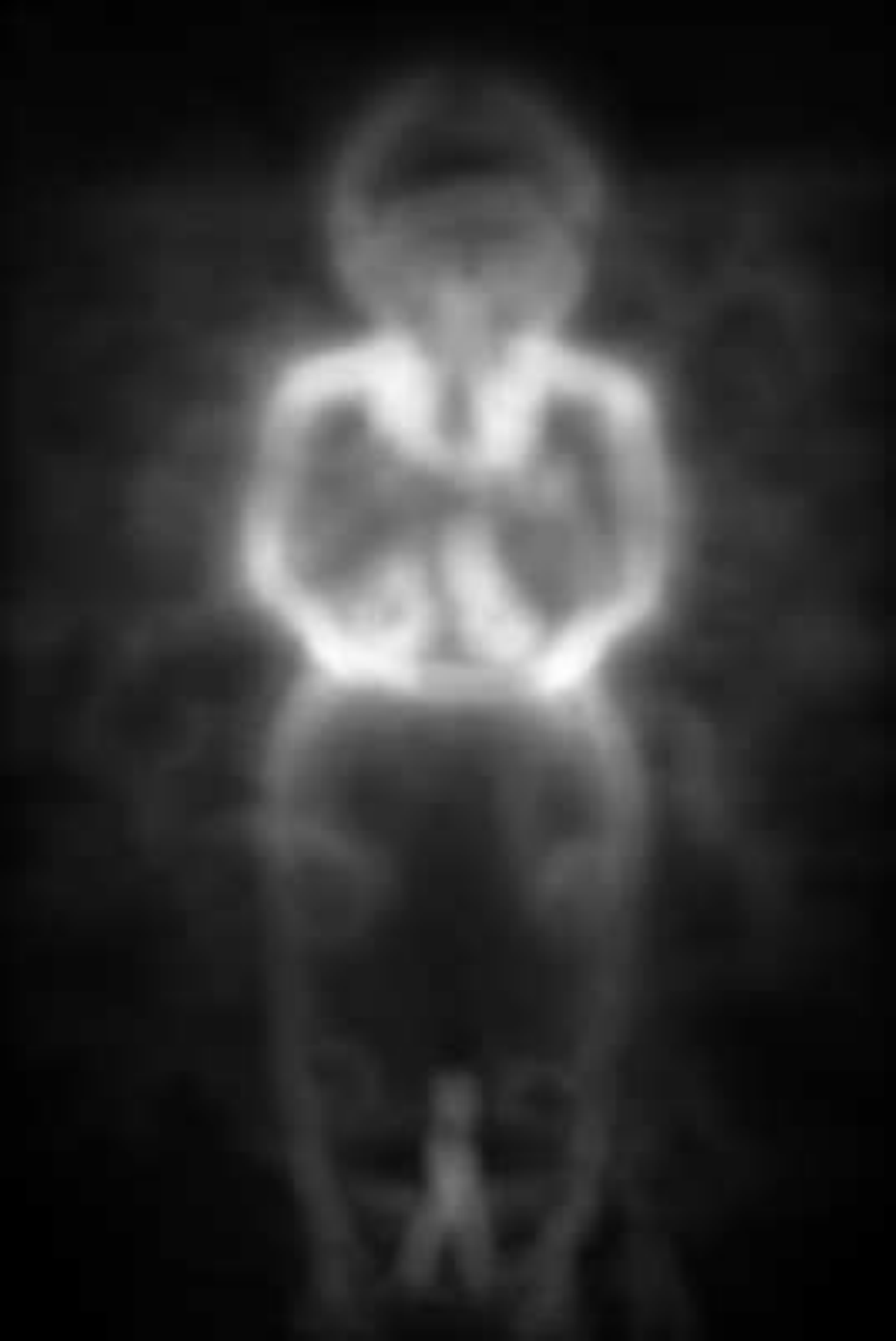}&
	\addExample{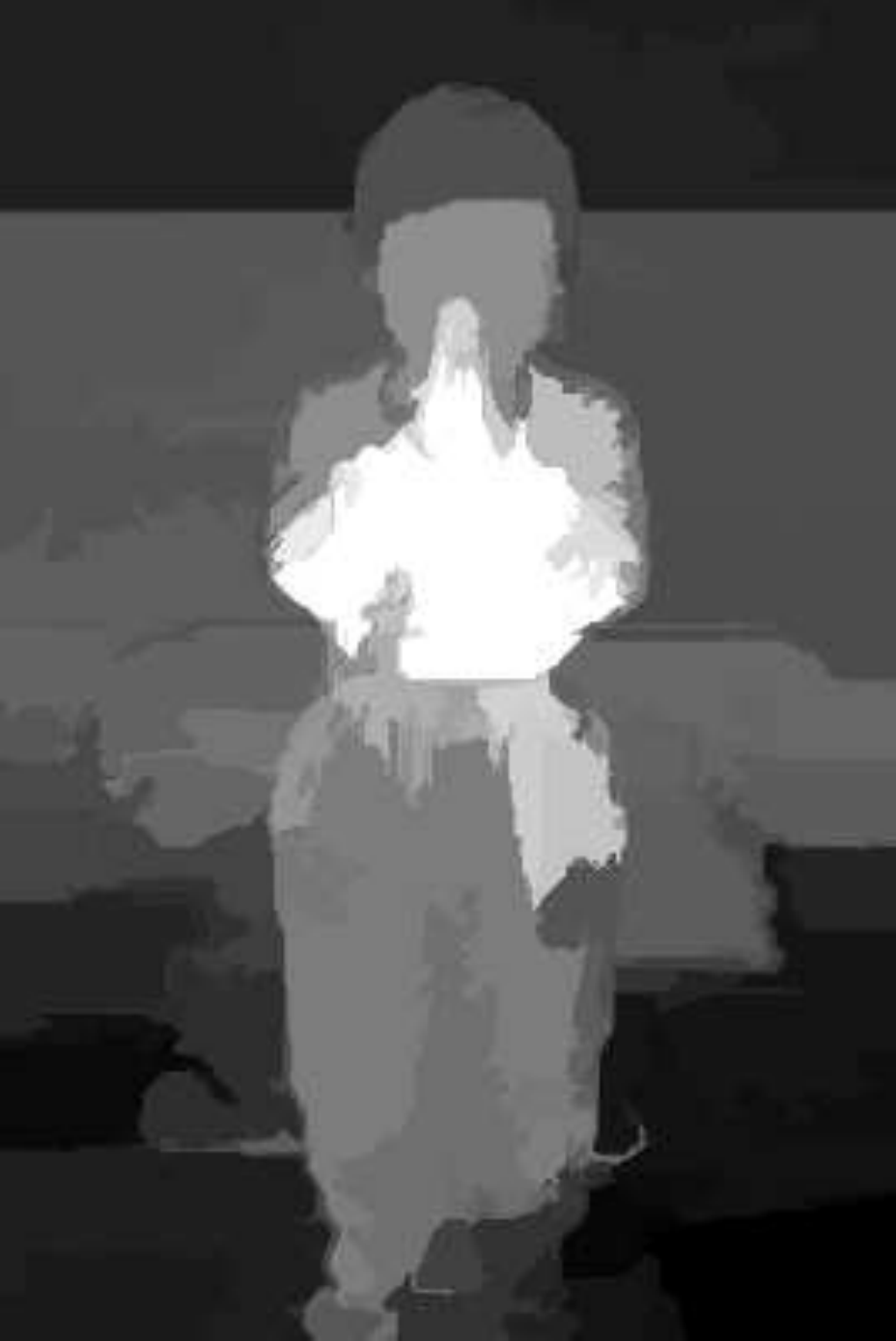}&
	\addExample{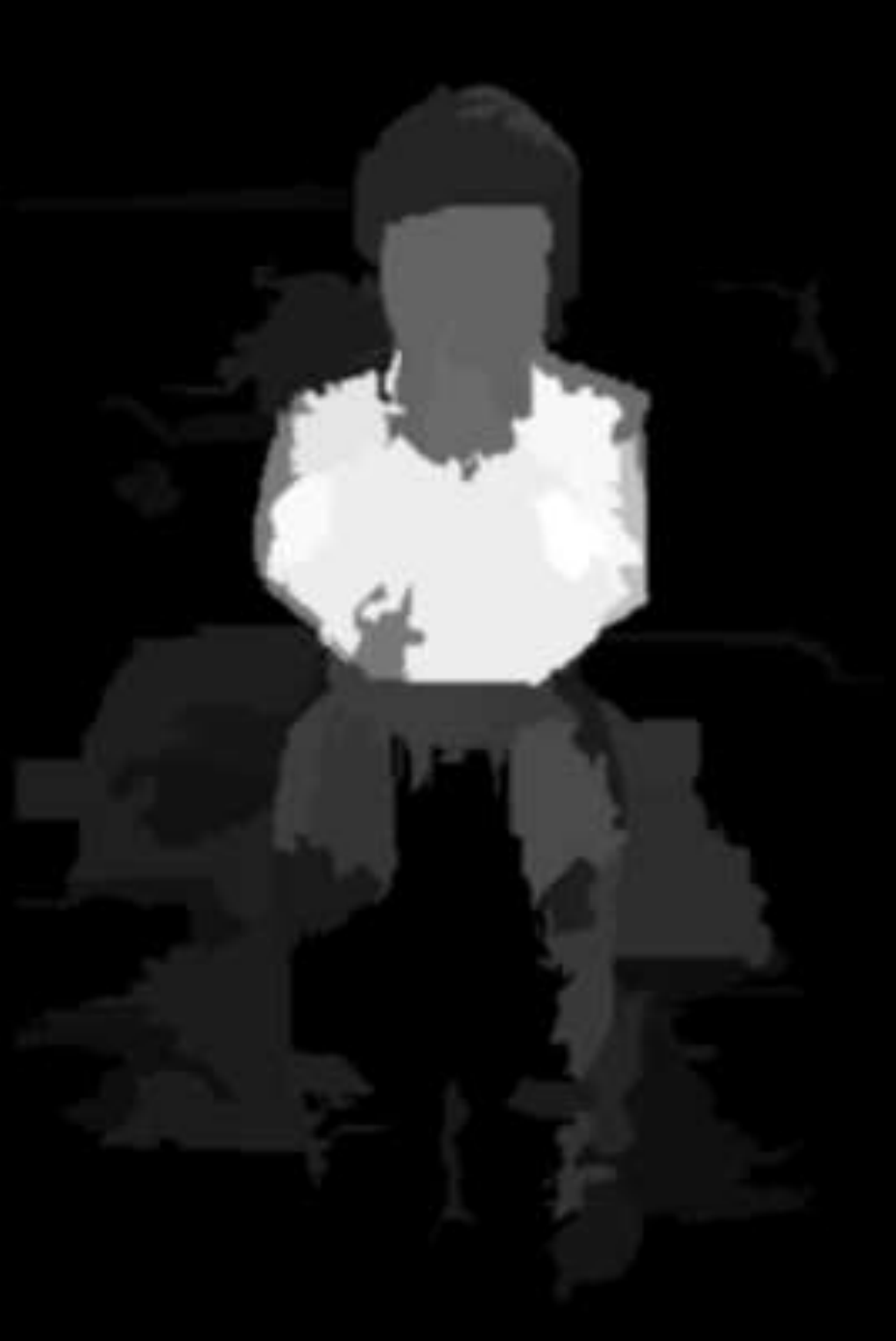}&
	\addExample{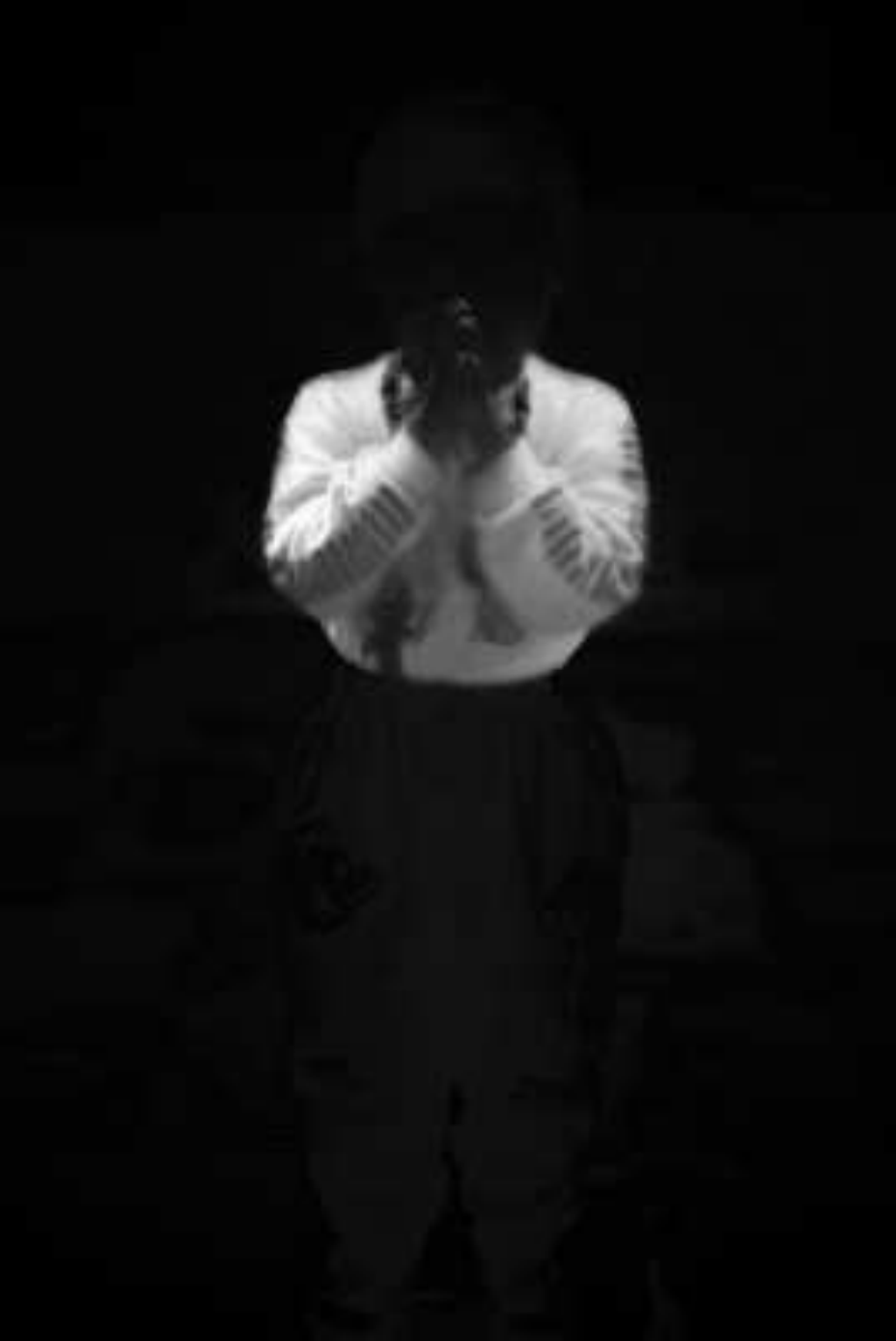}&
	\addExample{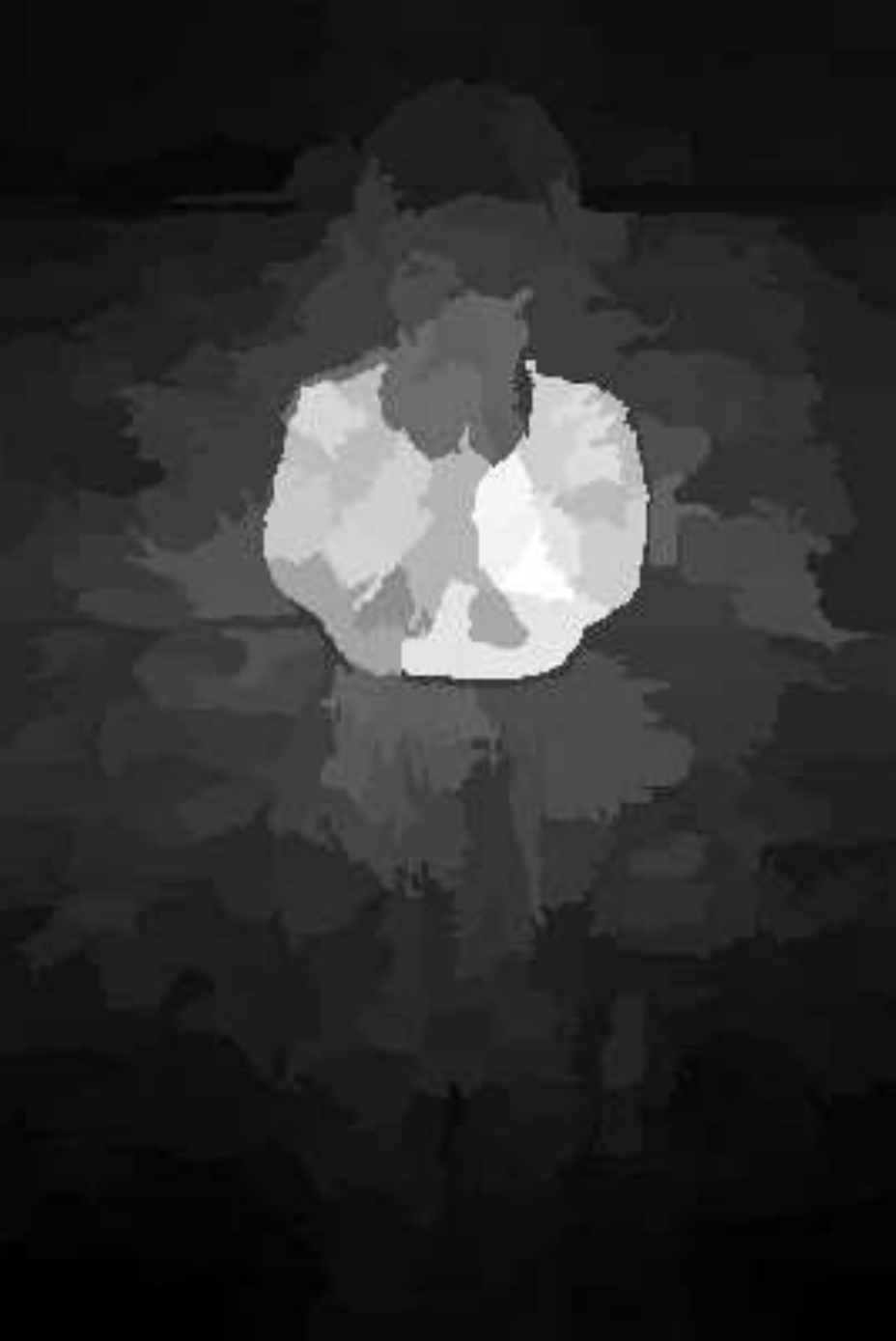}&
	\addExample{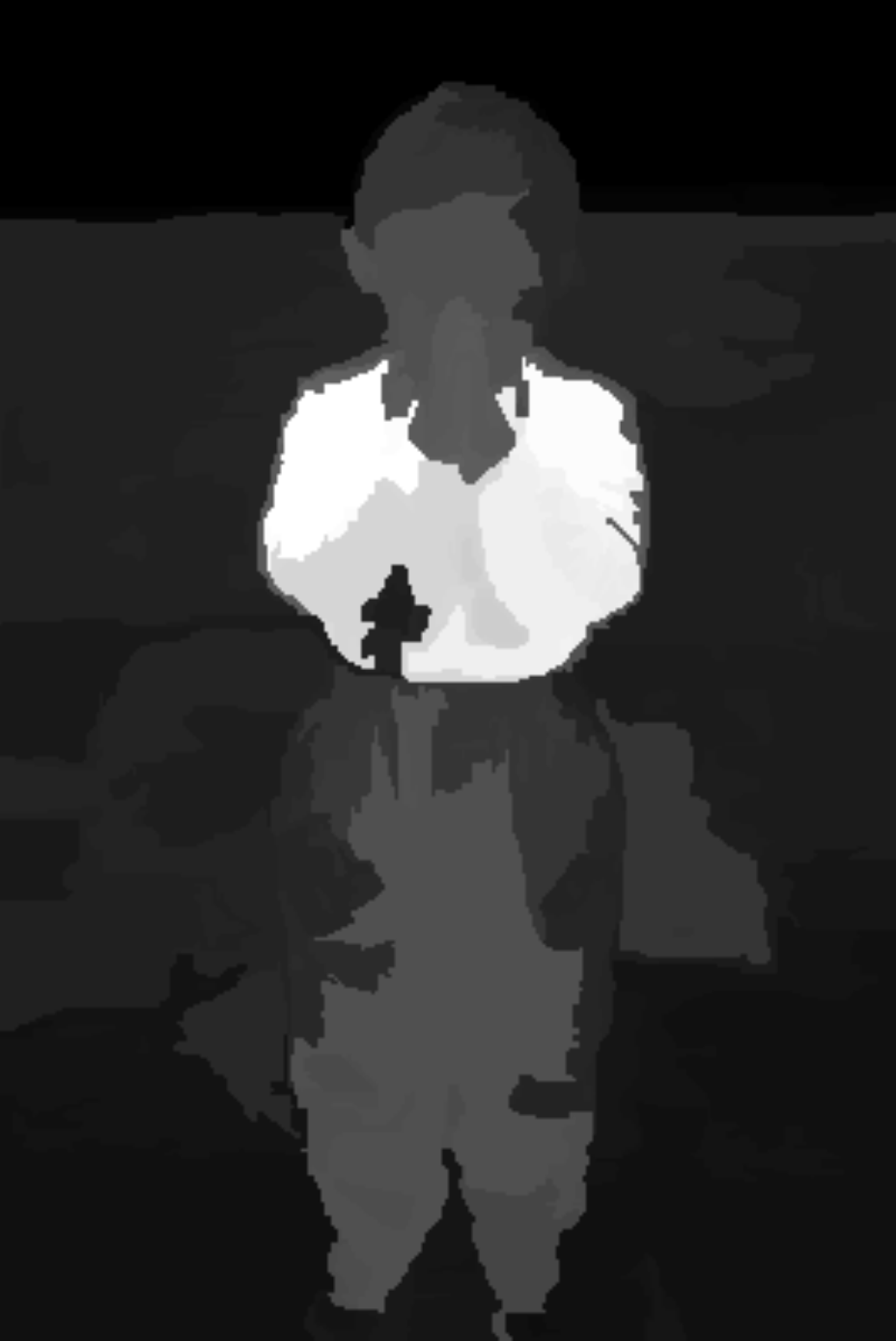}&
	\addExample{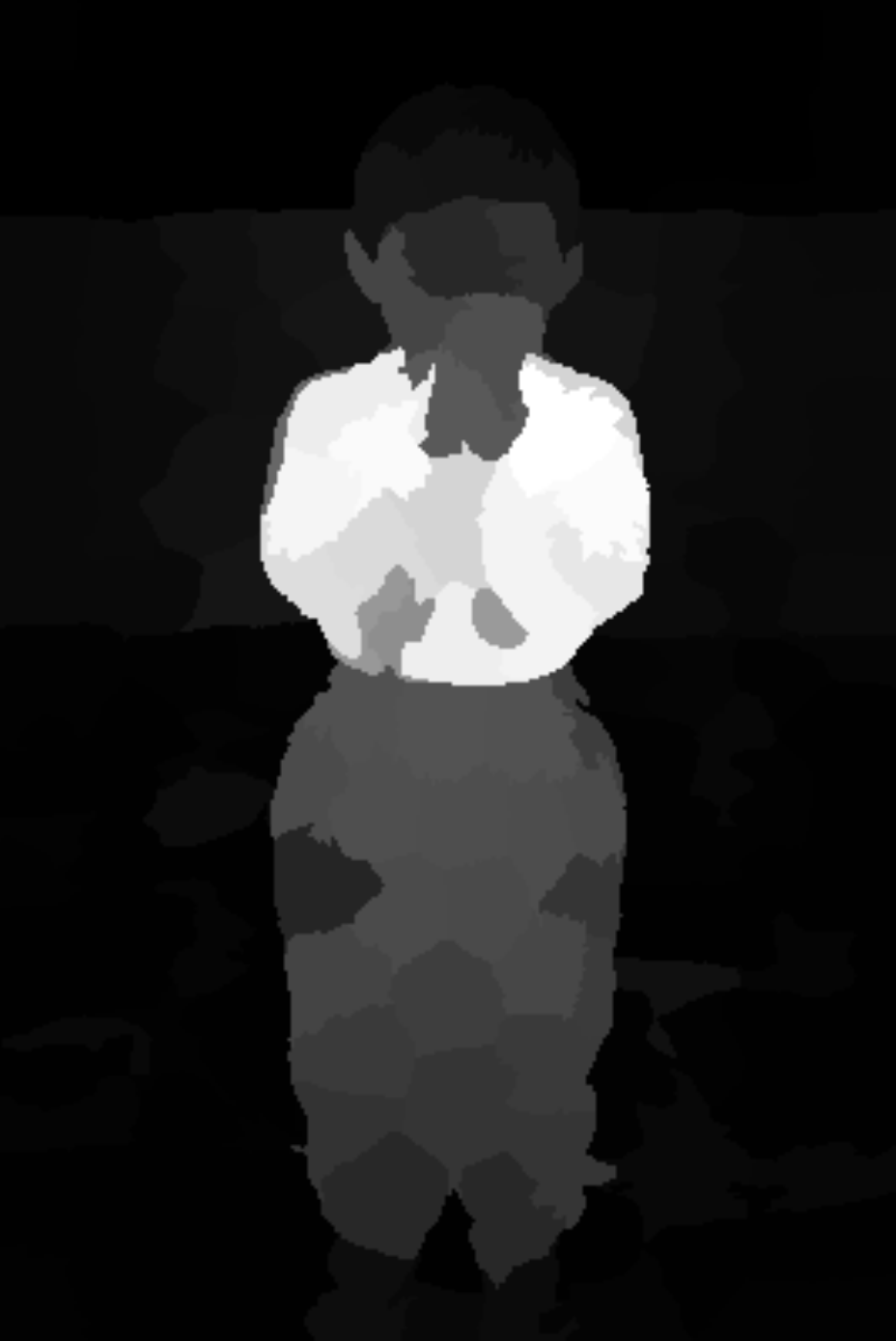}&
	\addExample{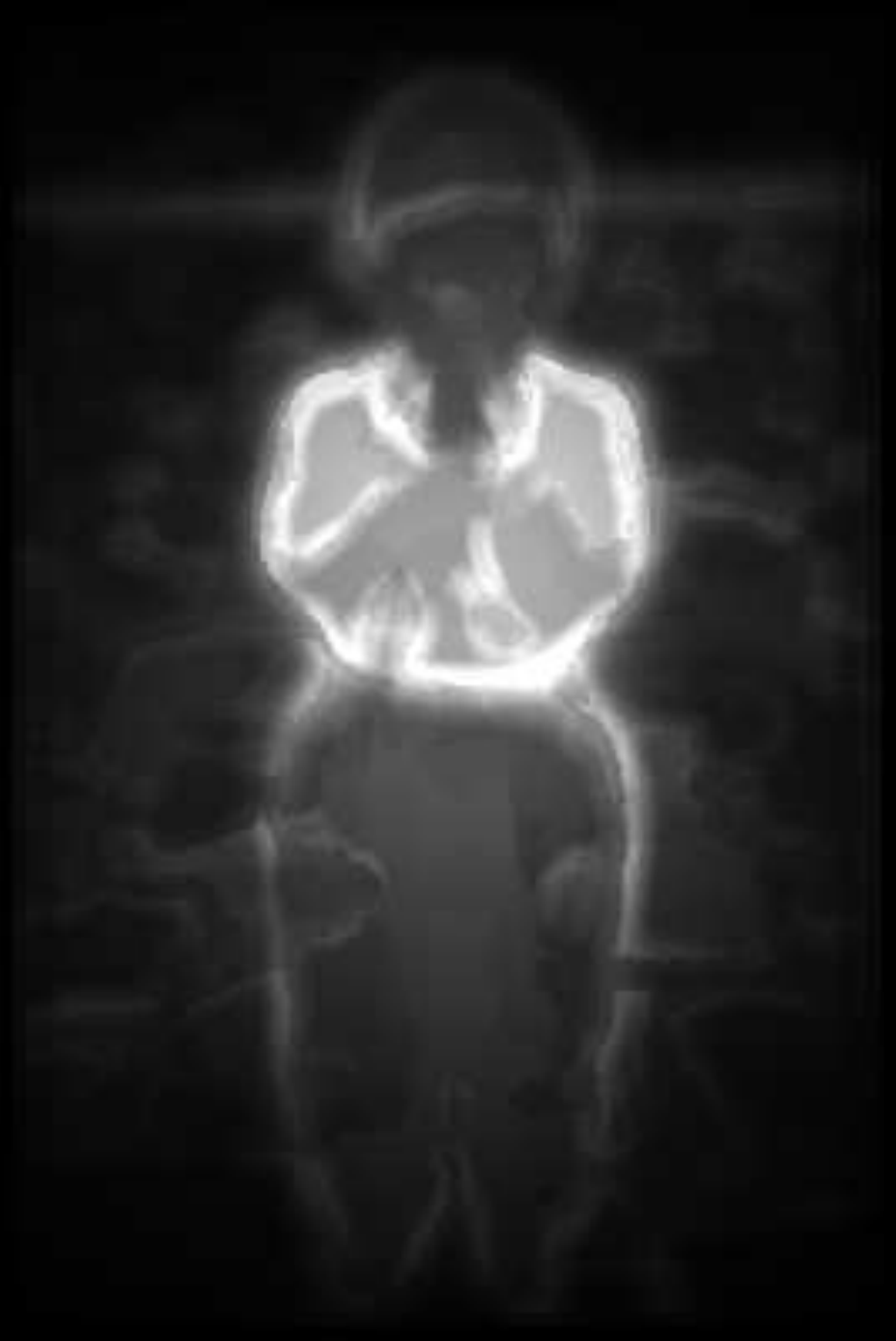}&
	\addExample{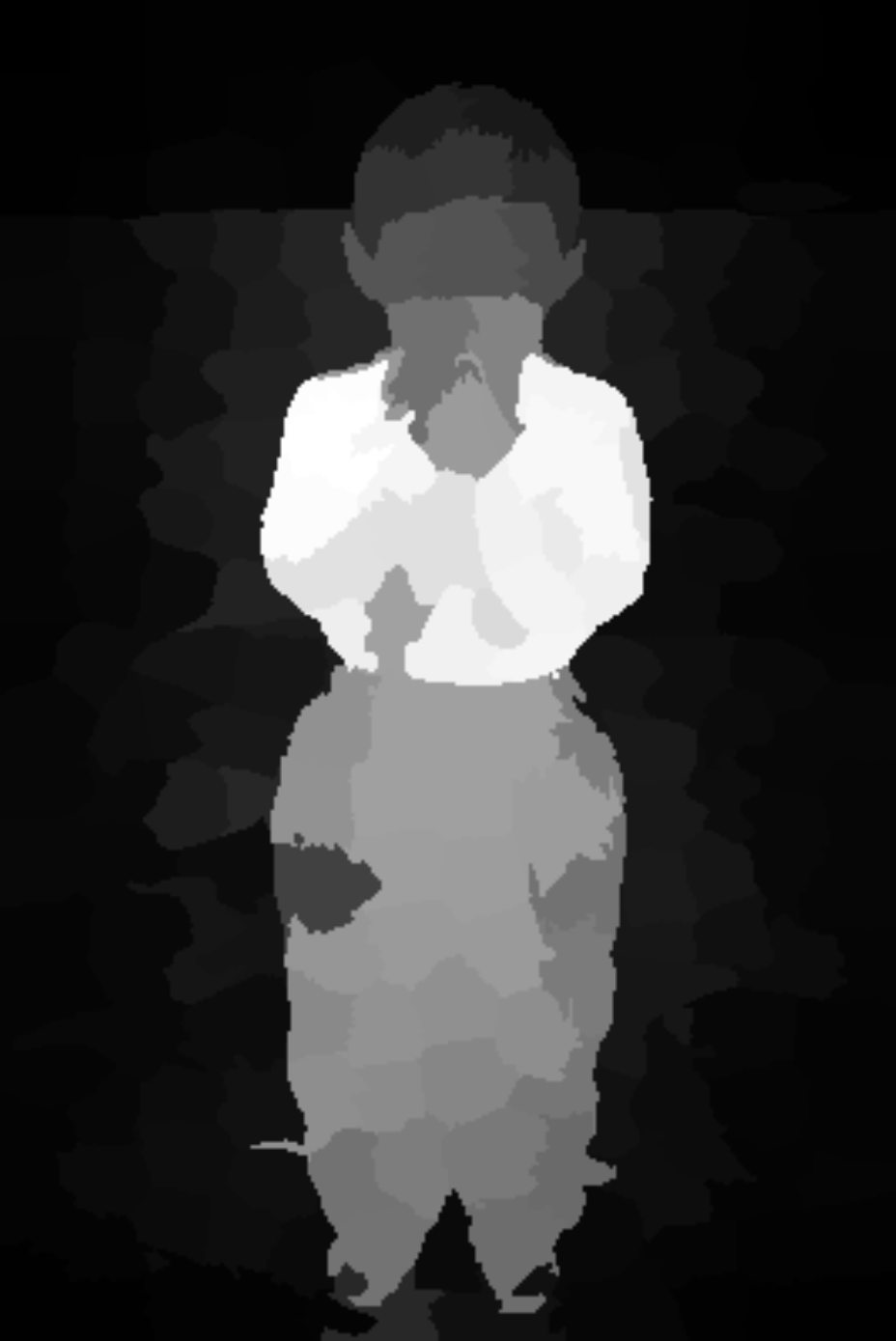}&
	\addExample{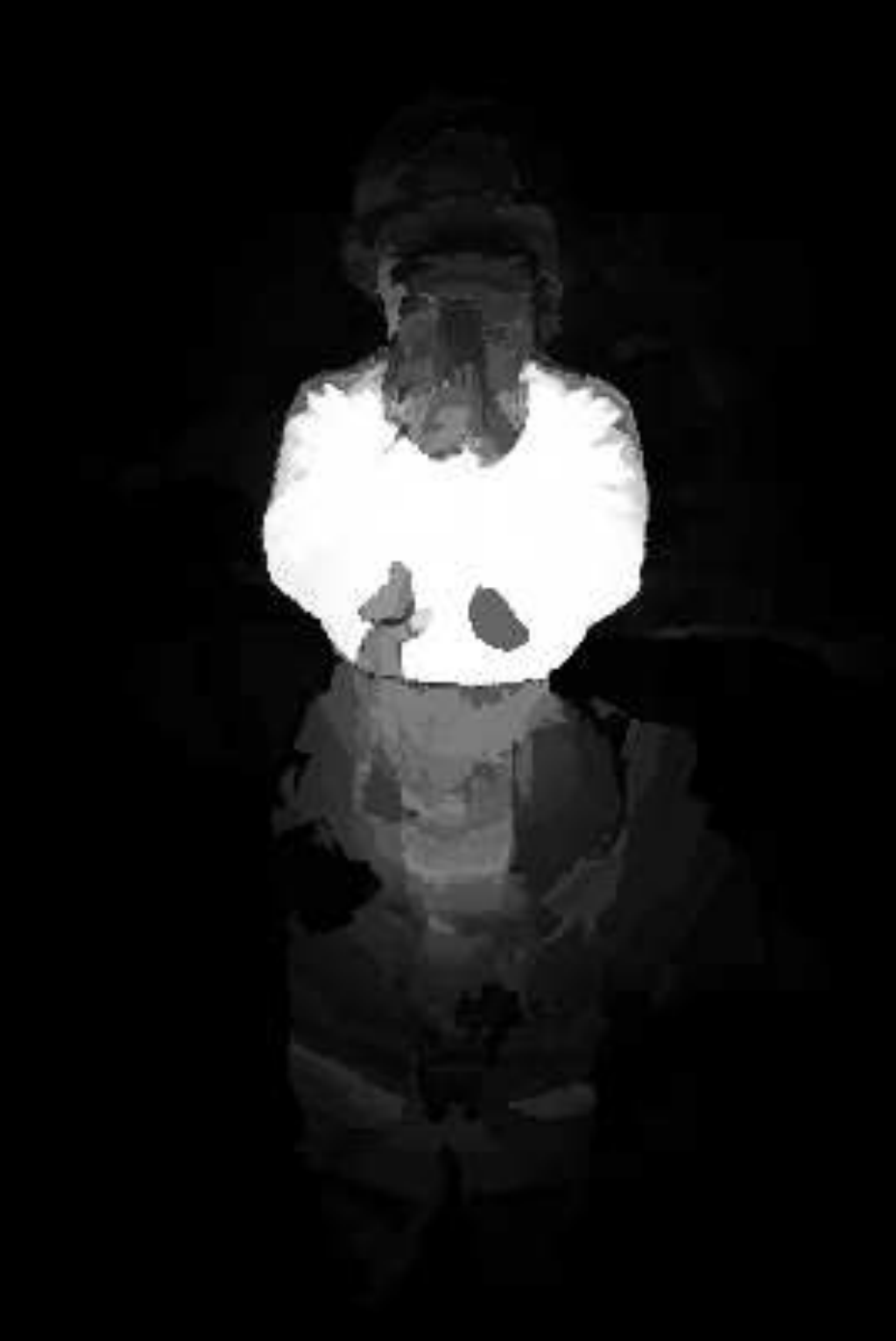}&
	\addExample{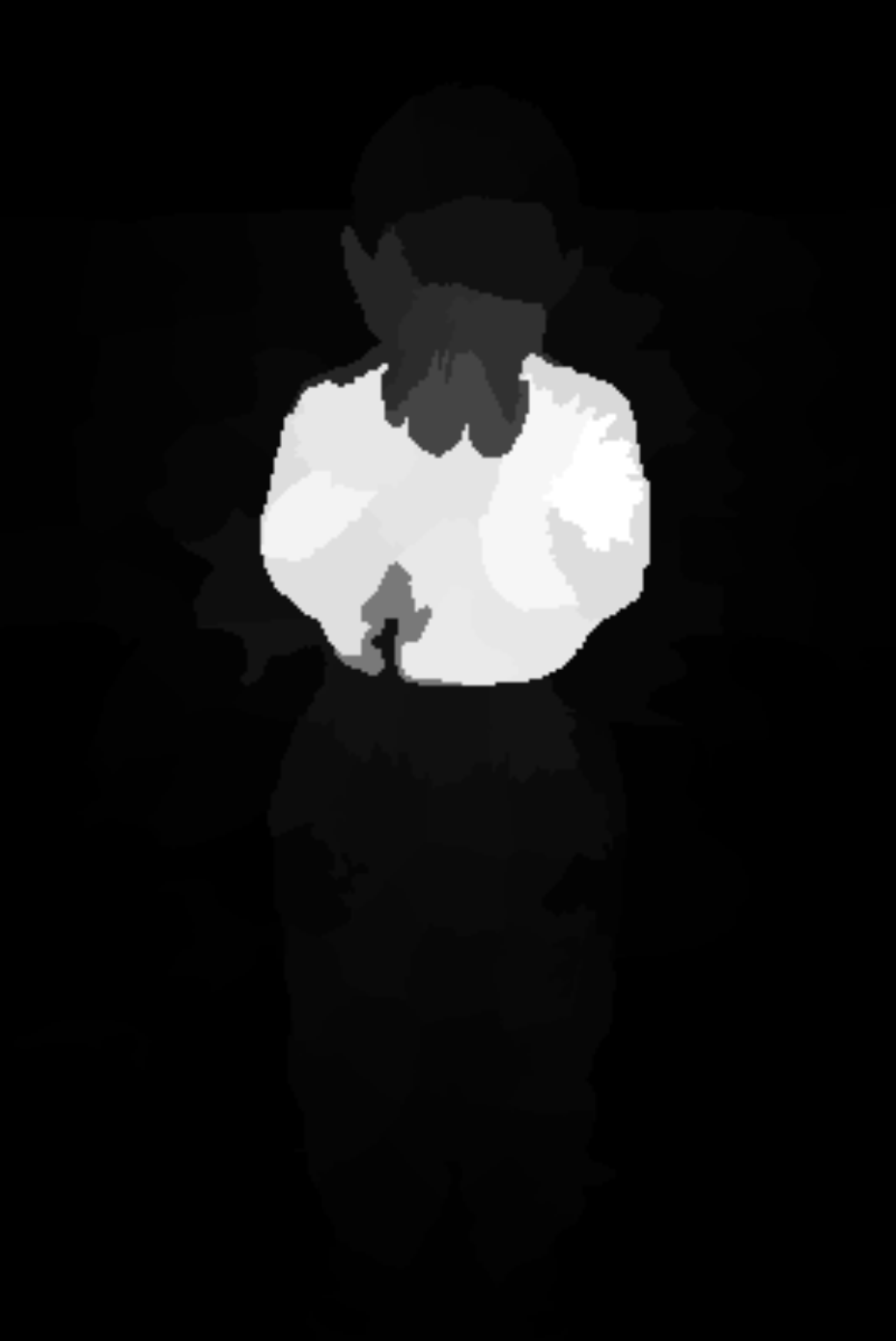}&
	\addExample{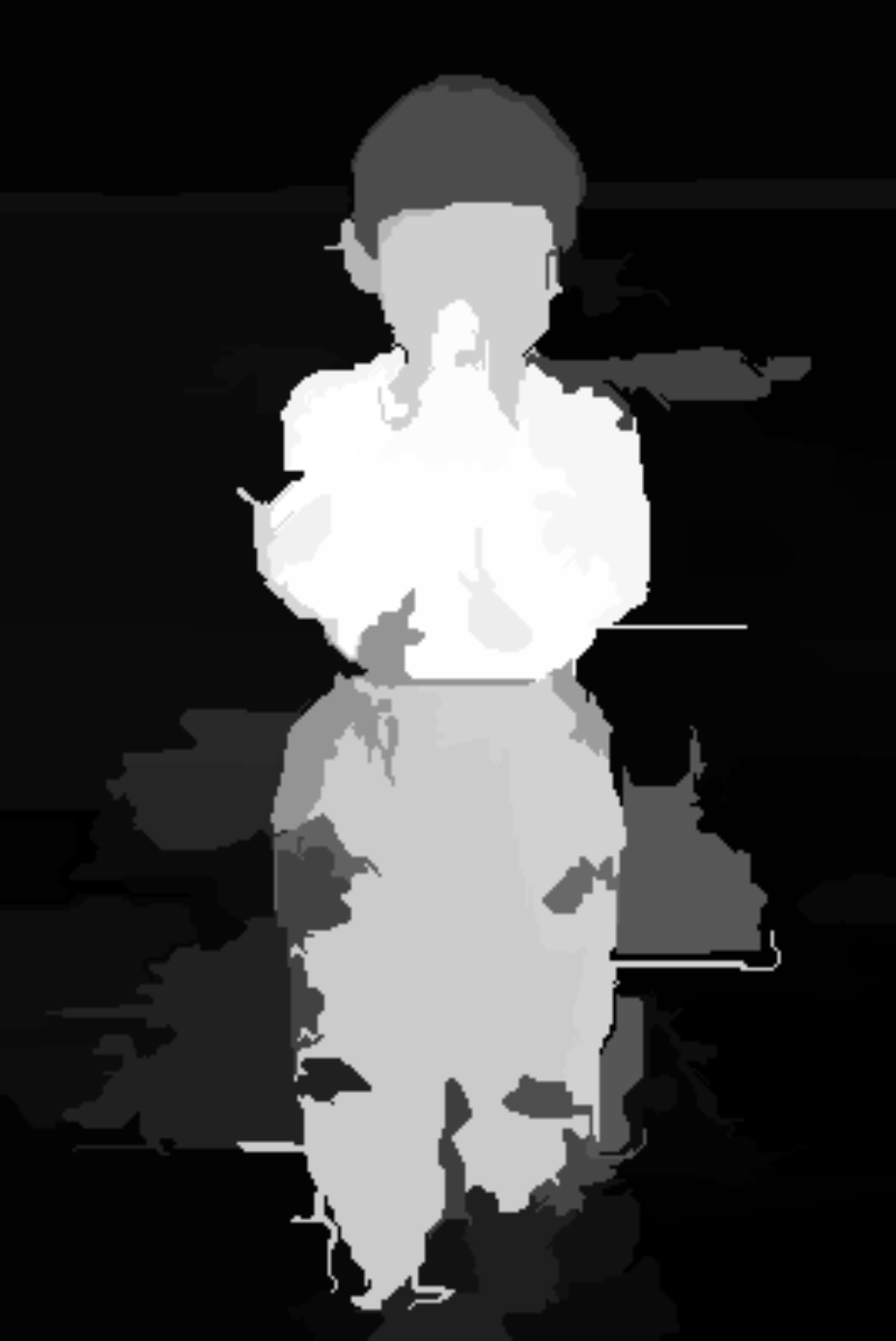}&
	\addExample{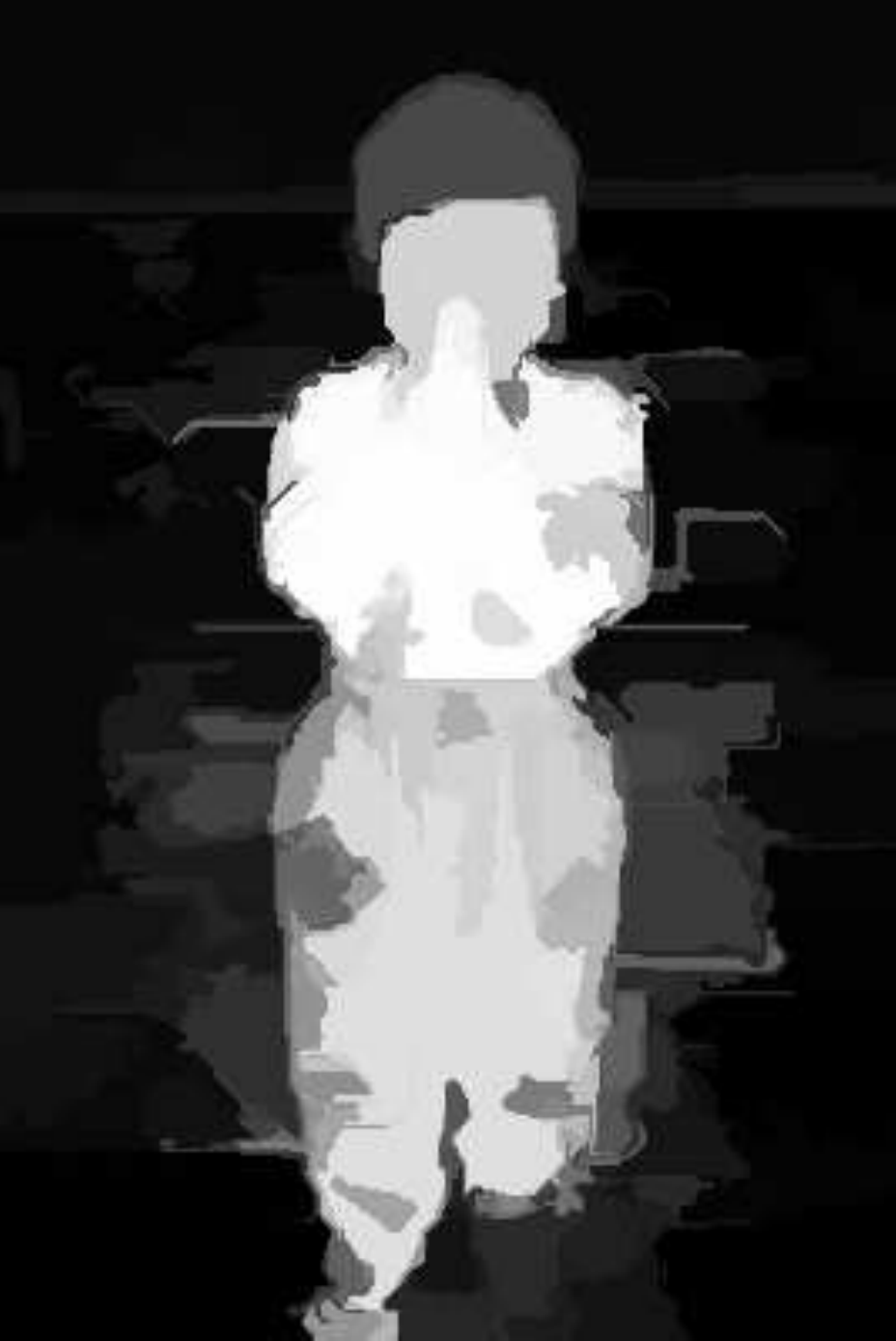}
    \\
	\addExample{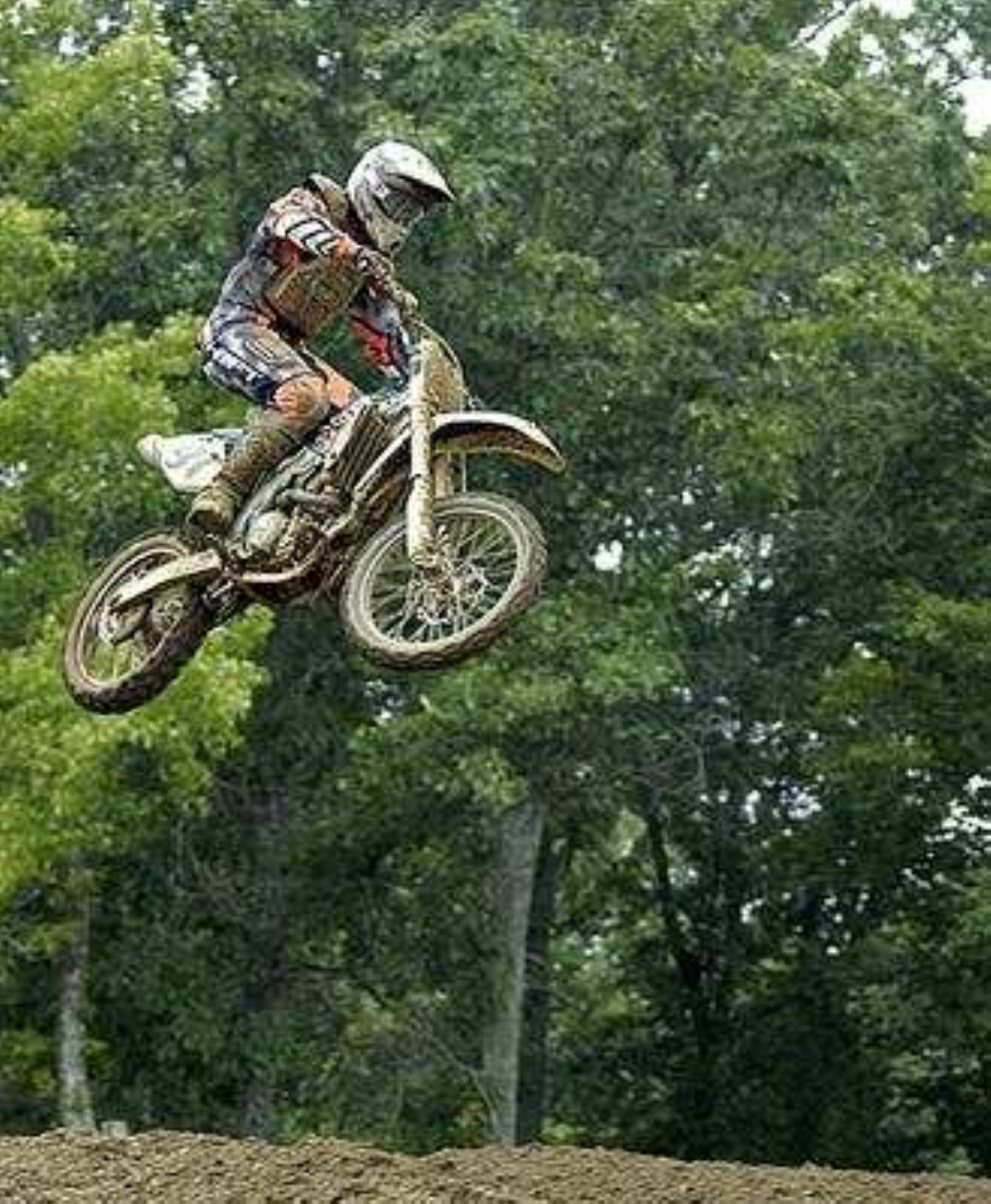}&
	\addExample{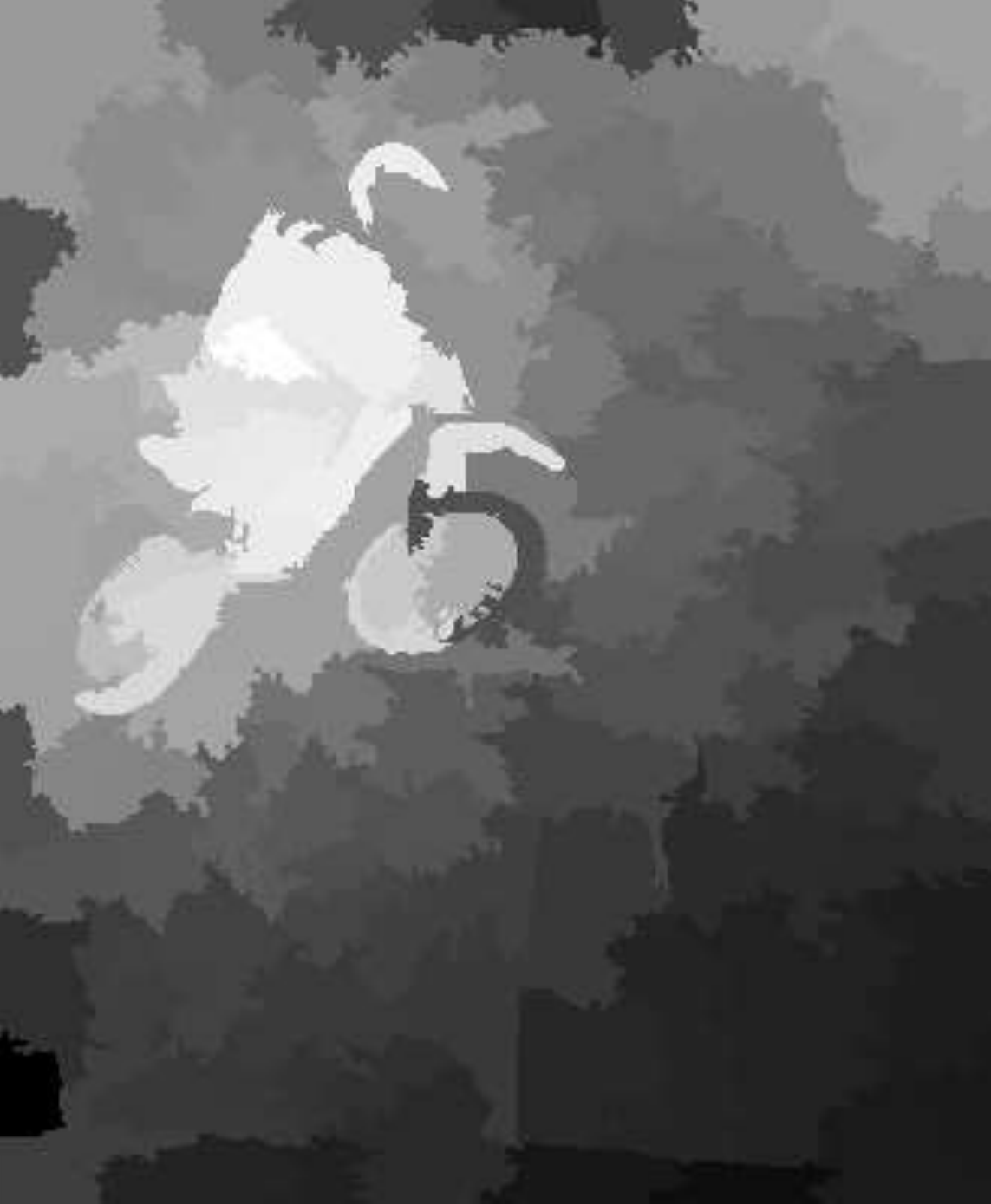}&
	\addExample{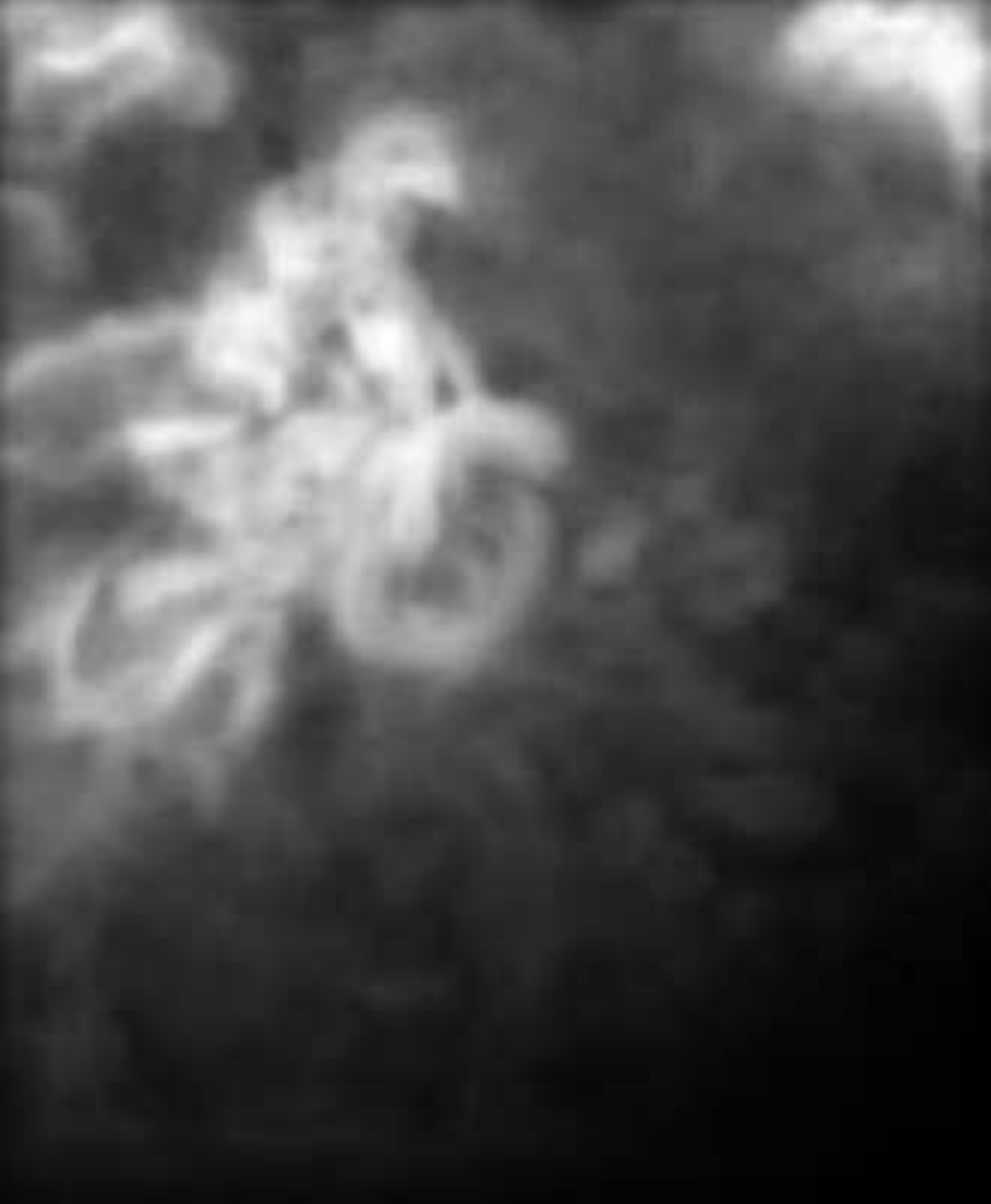}&
	\addExample{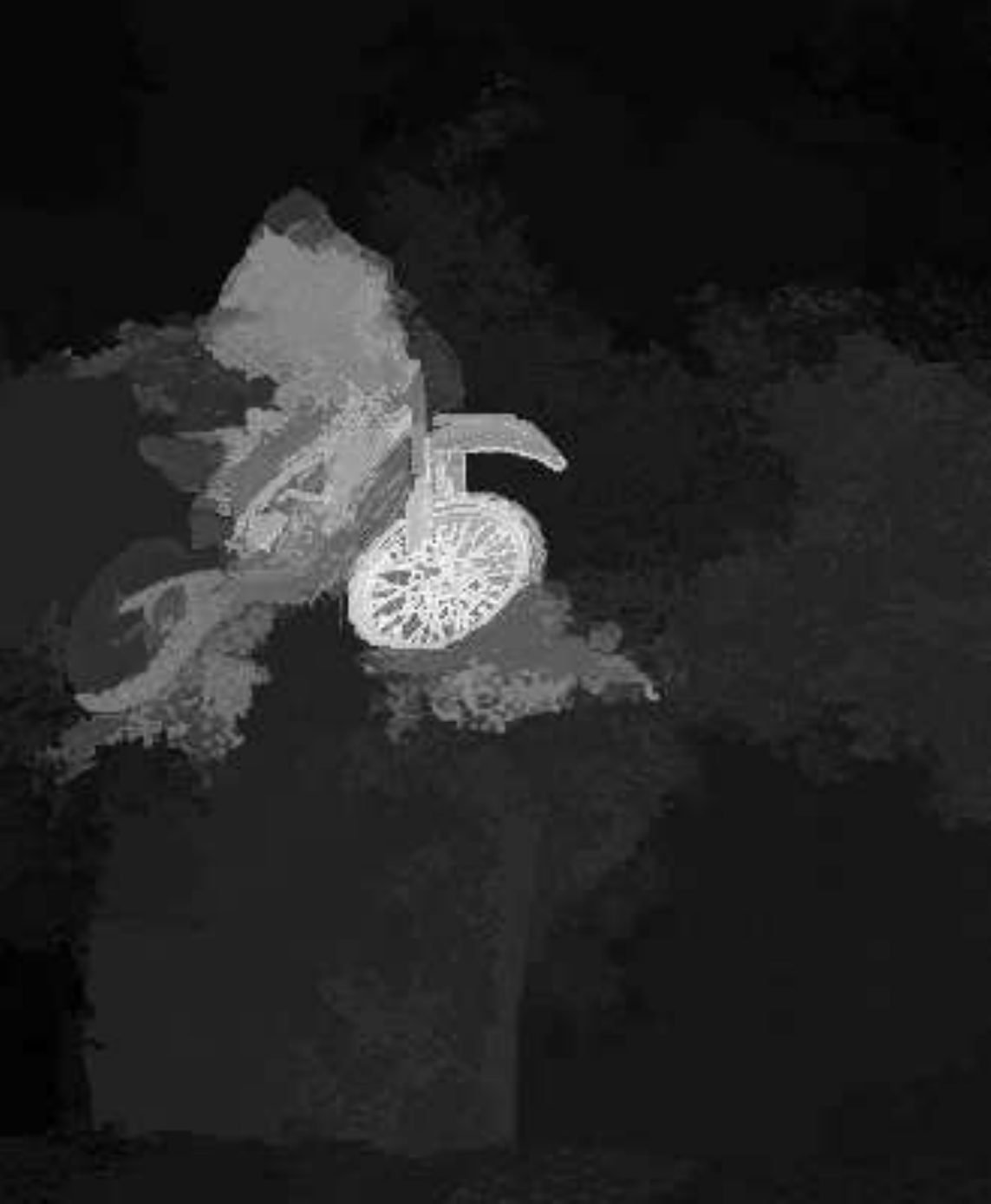}&
	\addExample{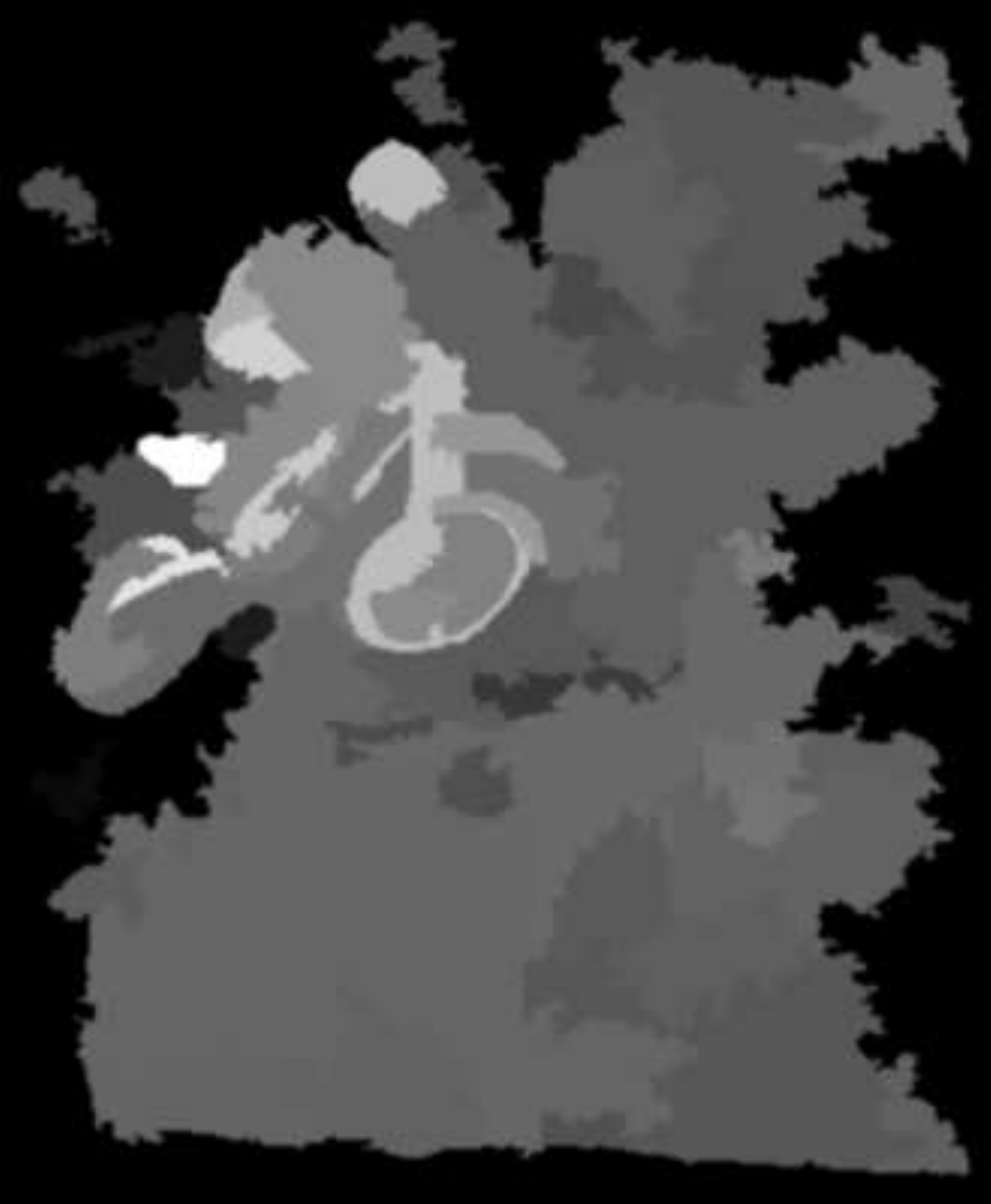}&
	\addExample{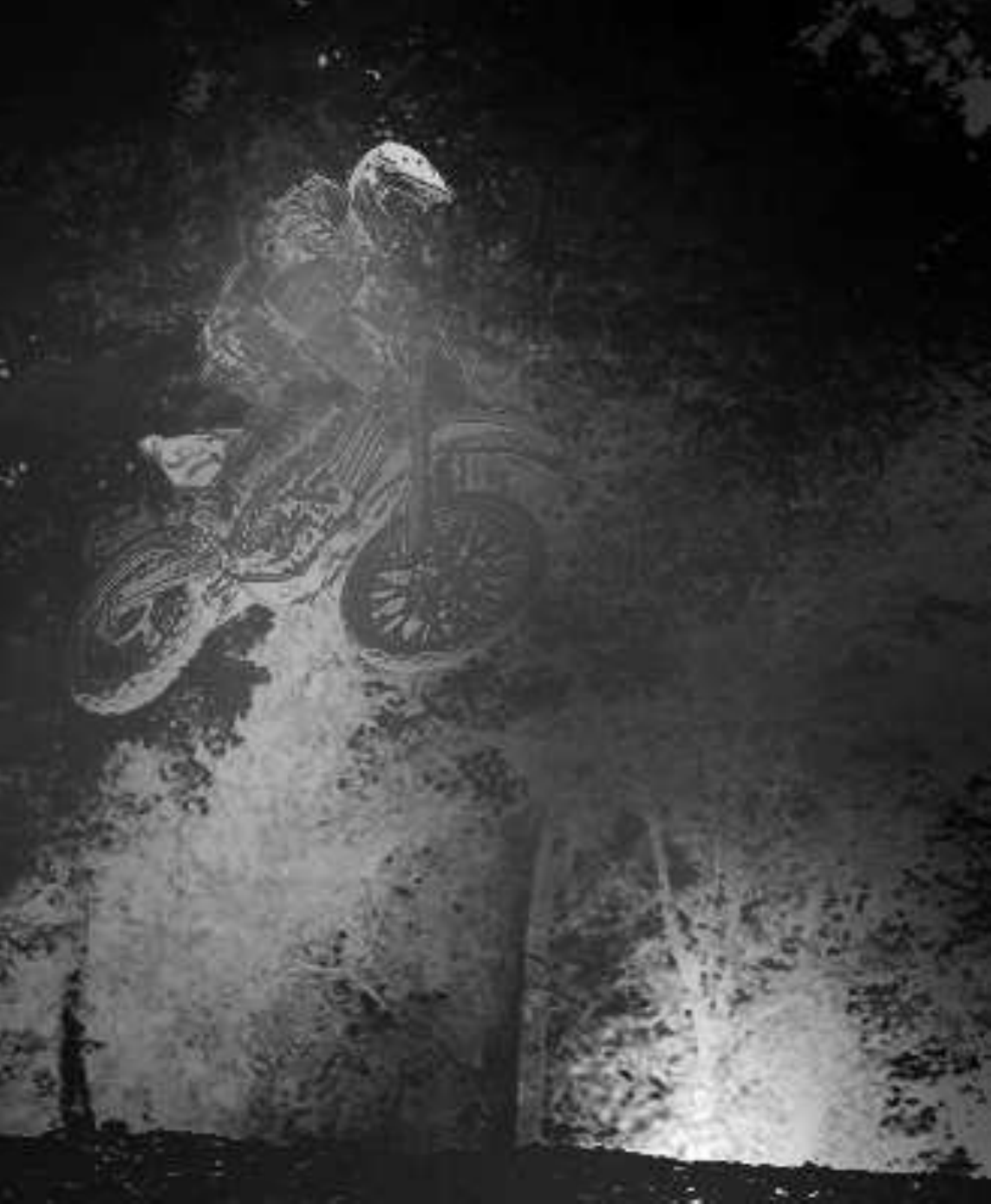}&
	\addExample{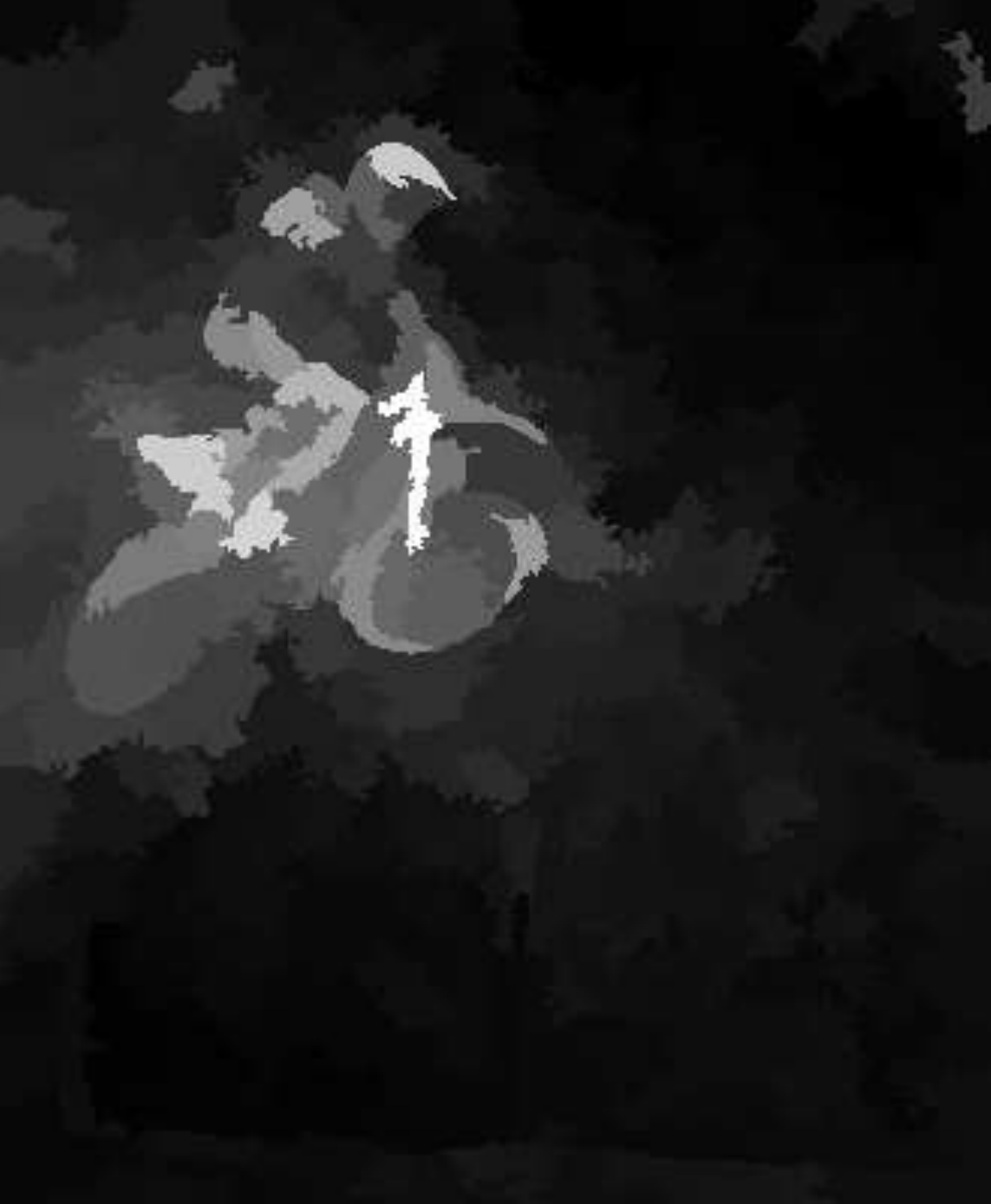}&
	\addExample{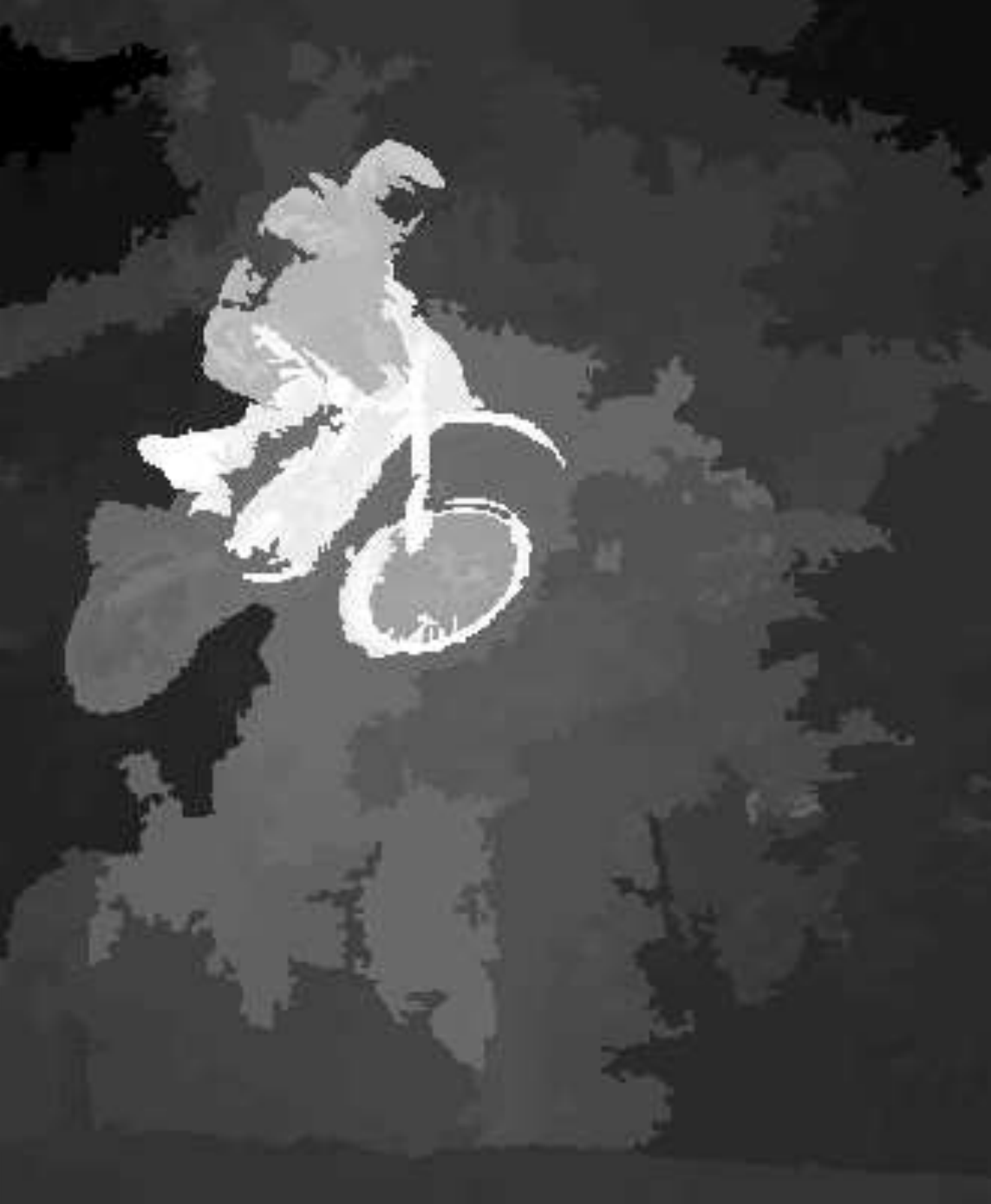}&
	\addExample{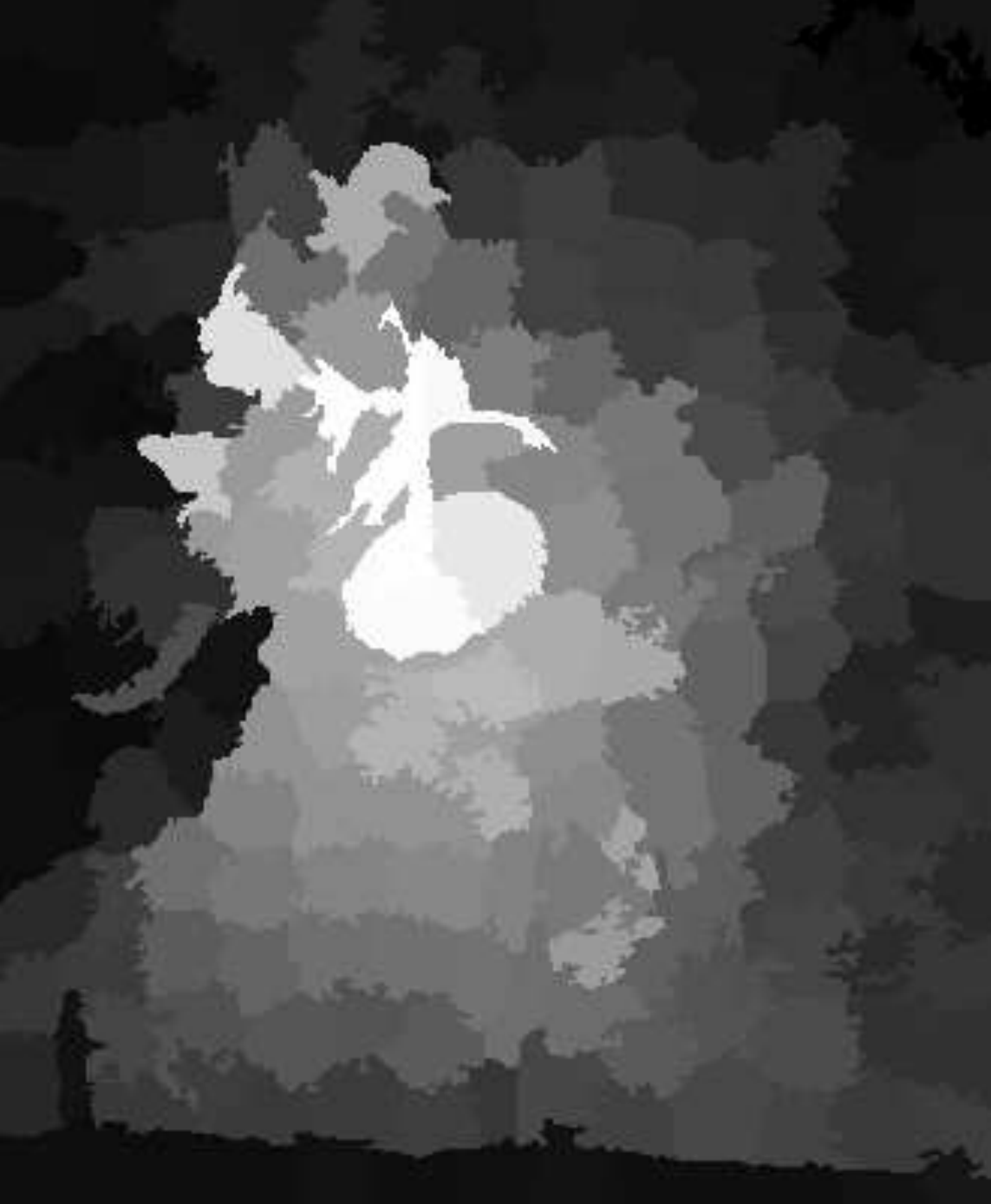}&
	\addExample{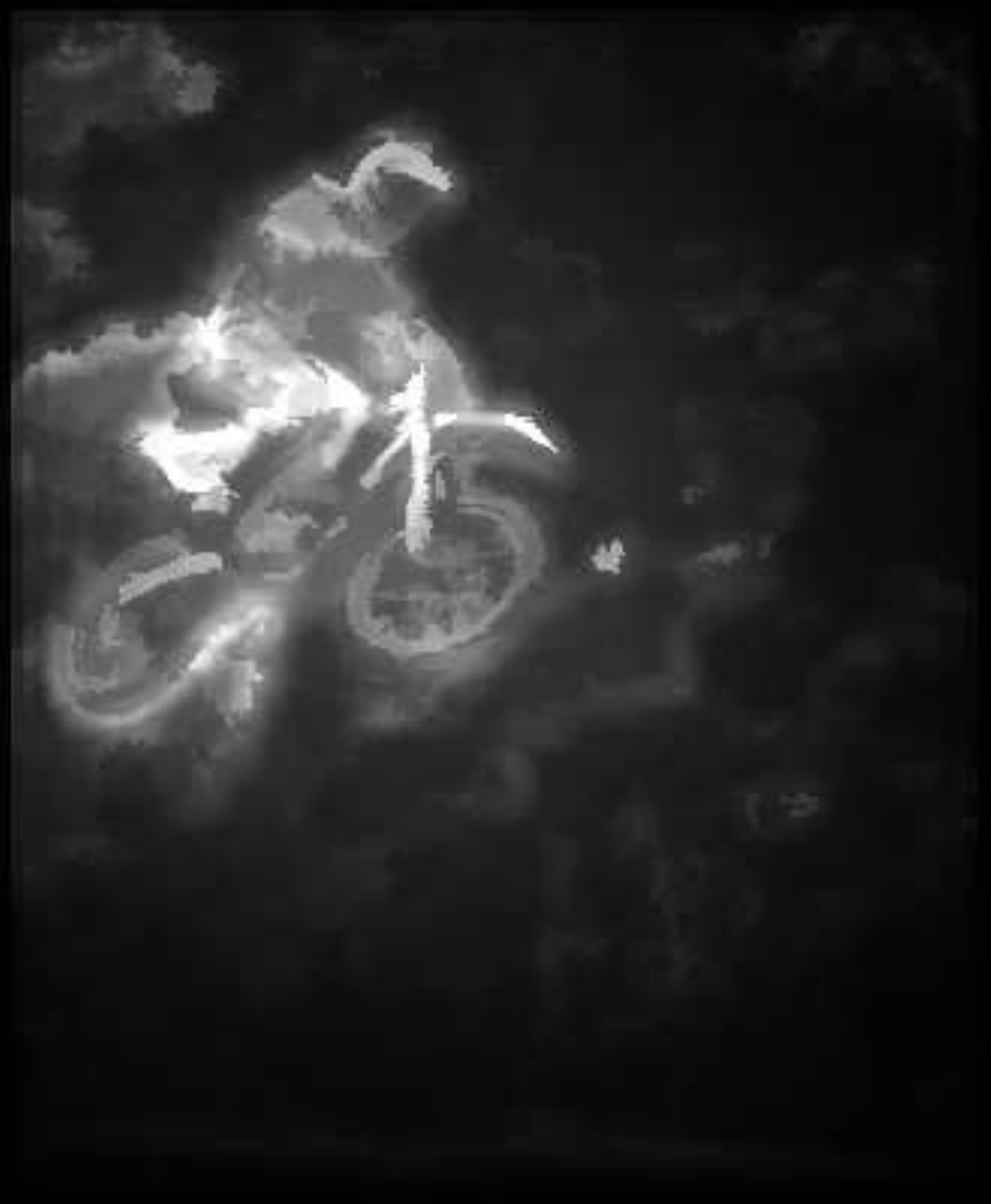}&
	\addExample{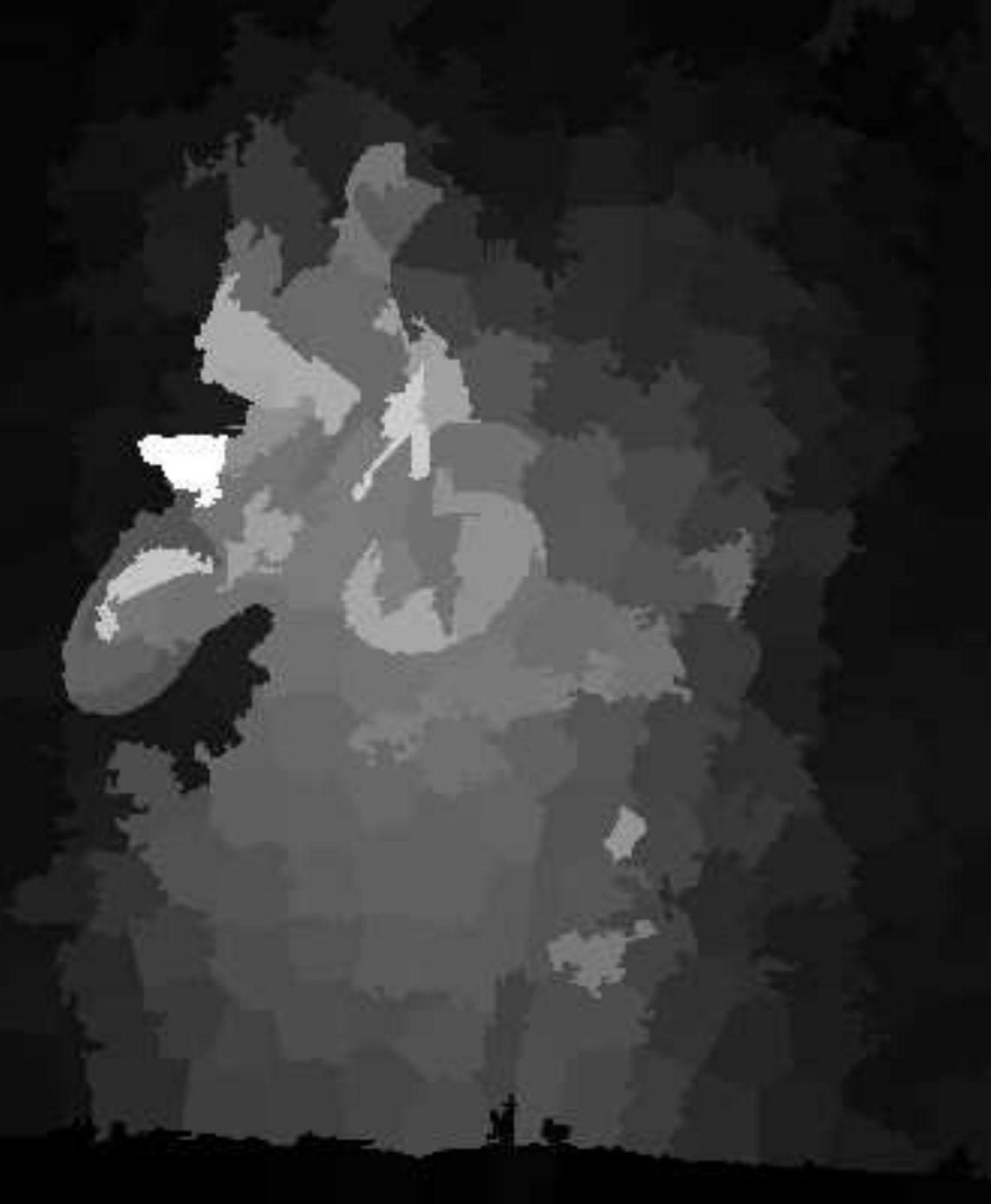}&
	\addExample{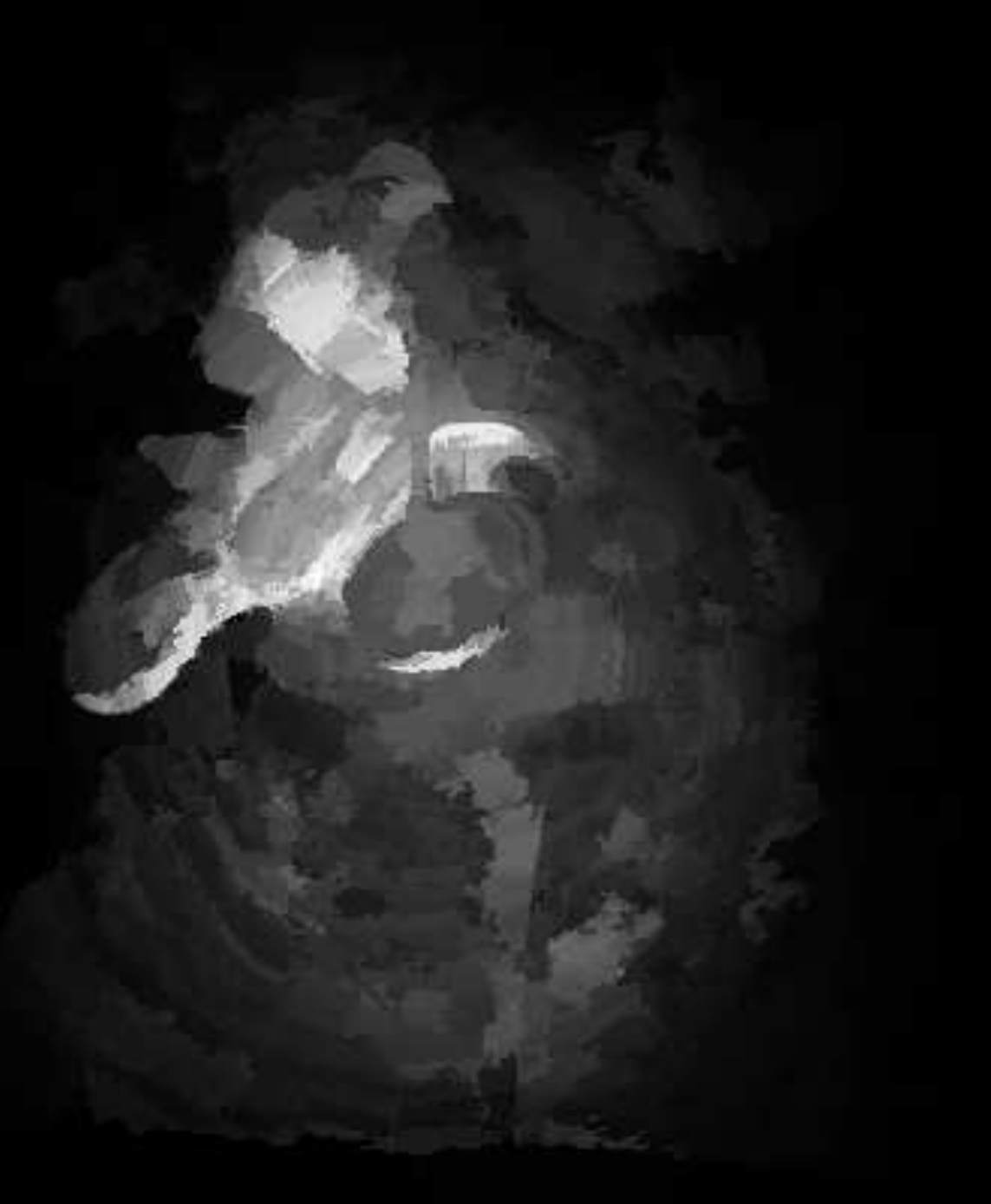}&
	\addExample{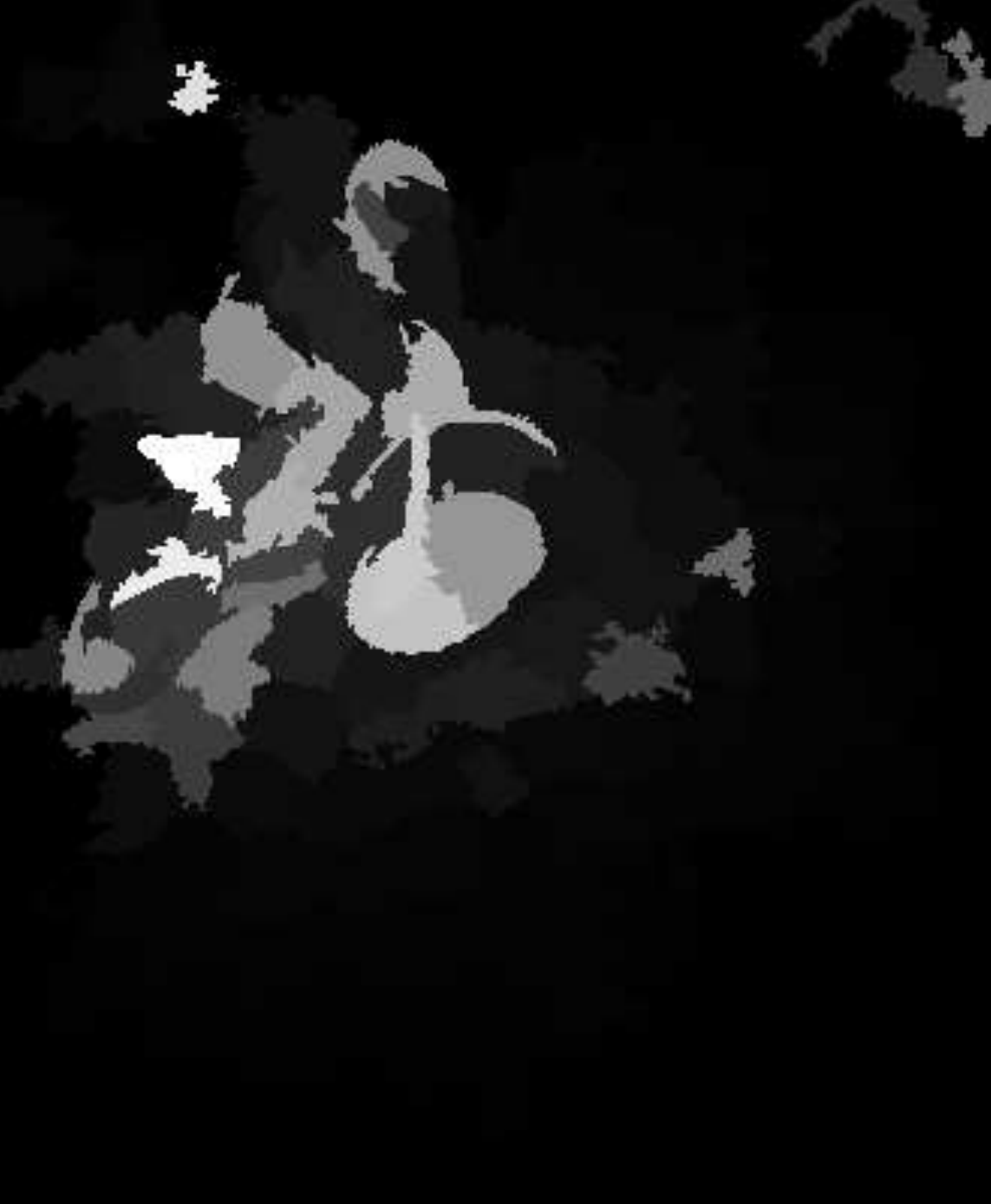}&
	\addExample{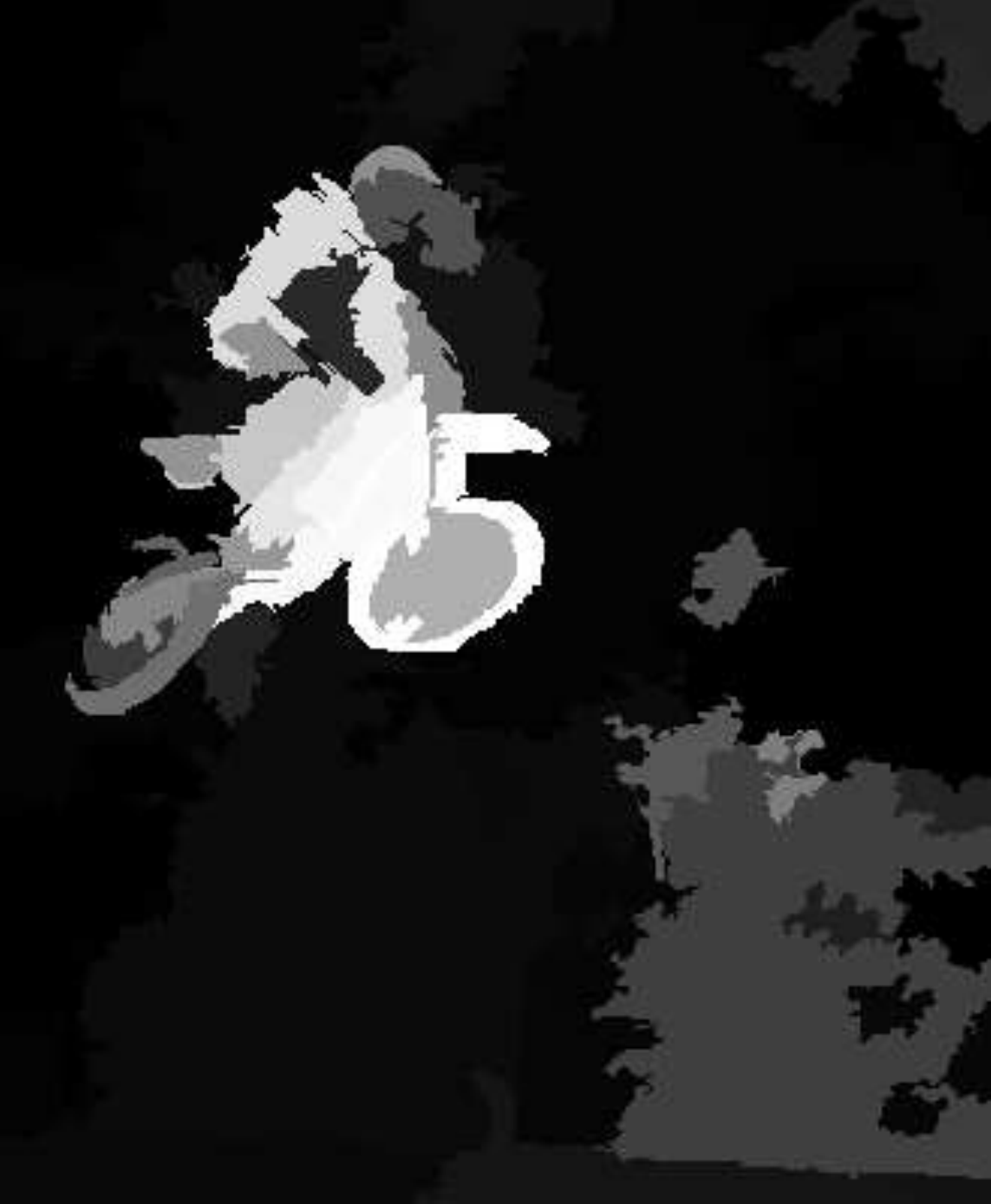}&
	\addExample{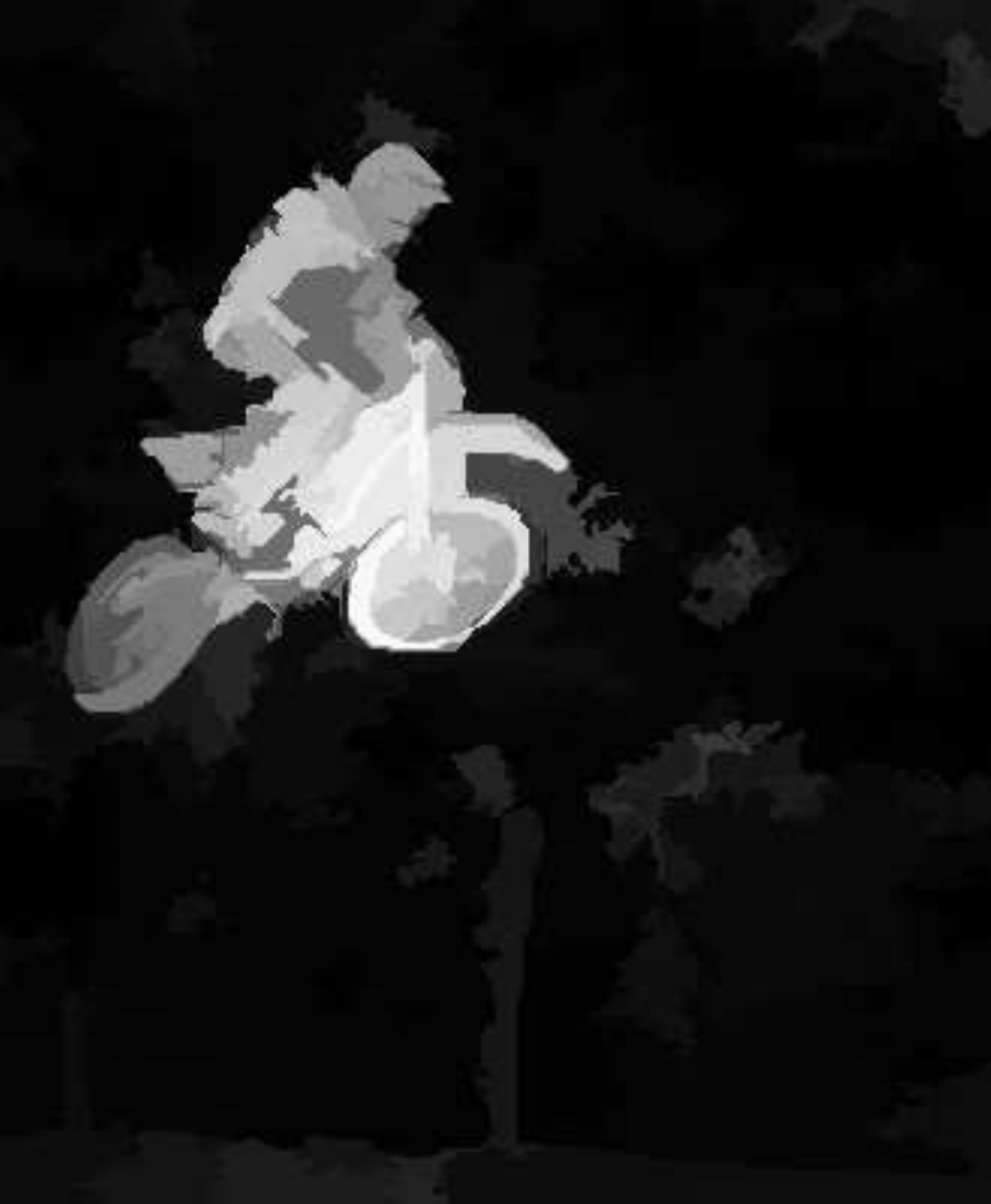}
    \\
	\addExample{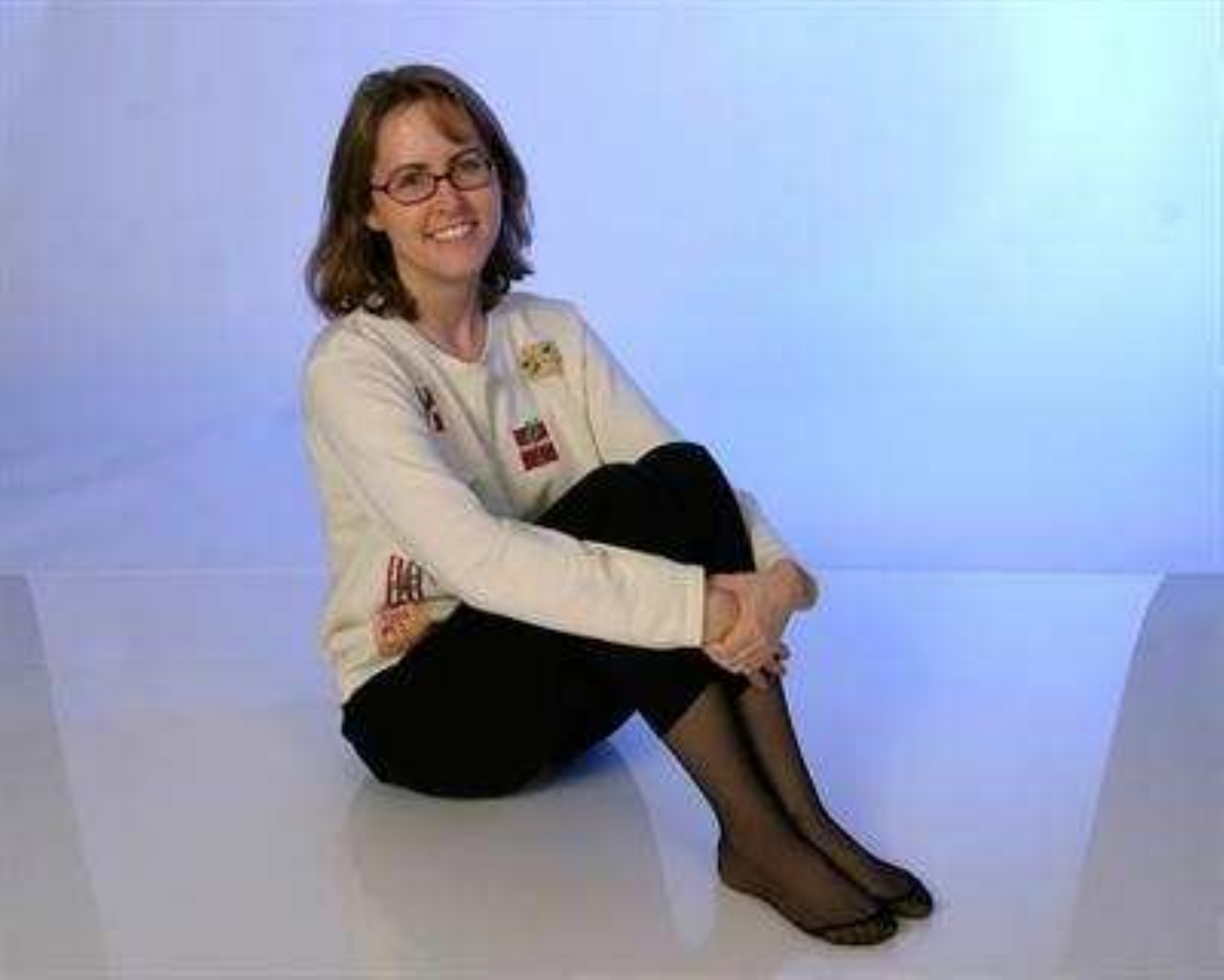}&
	\addExample{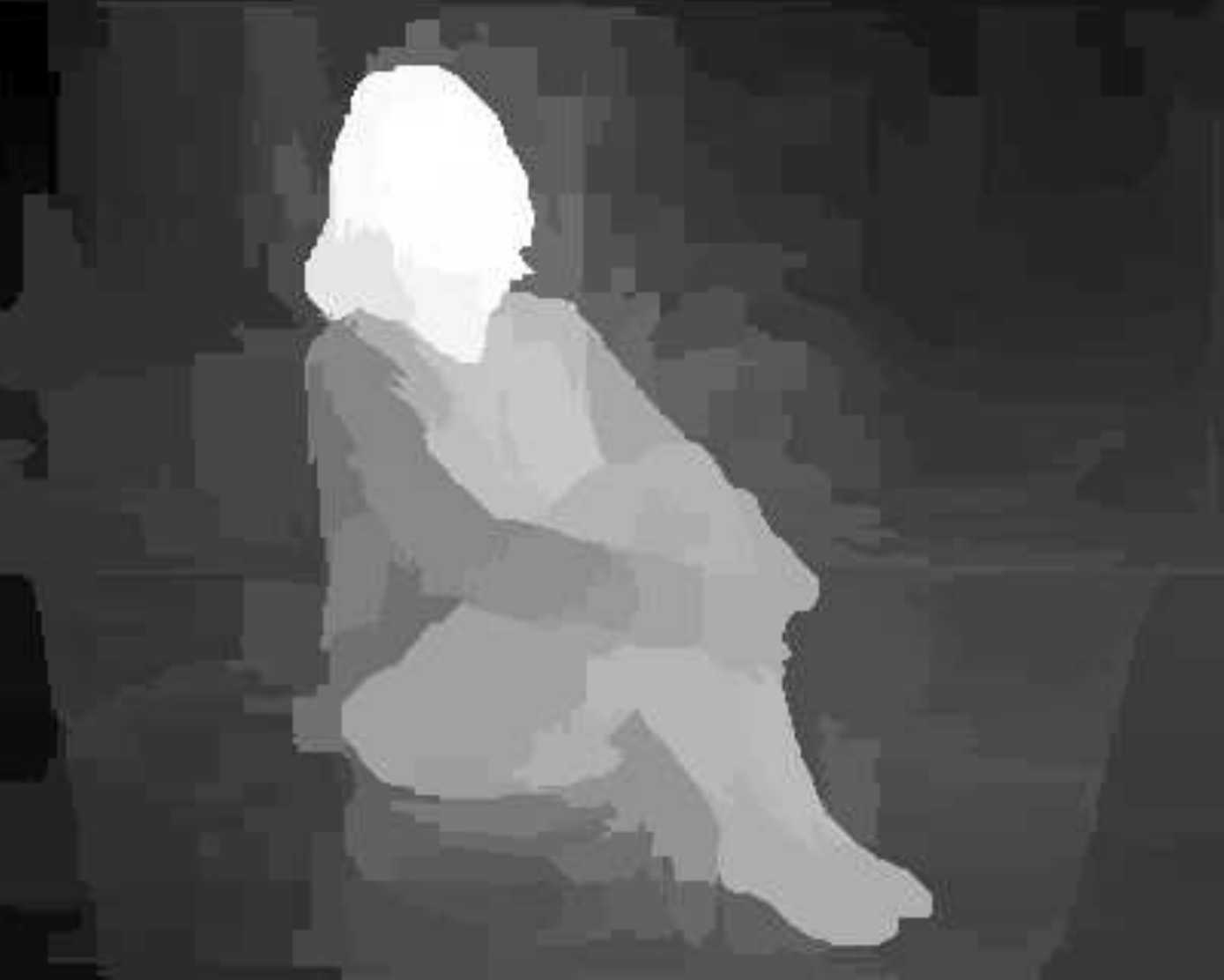}&
	\addExample{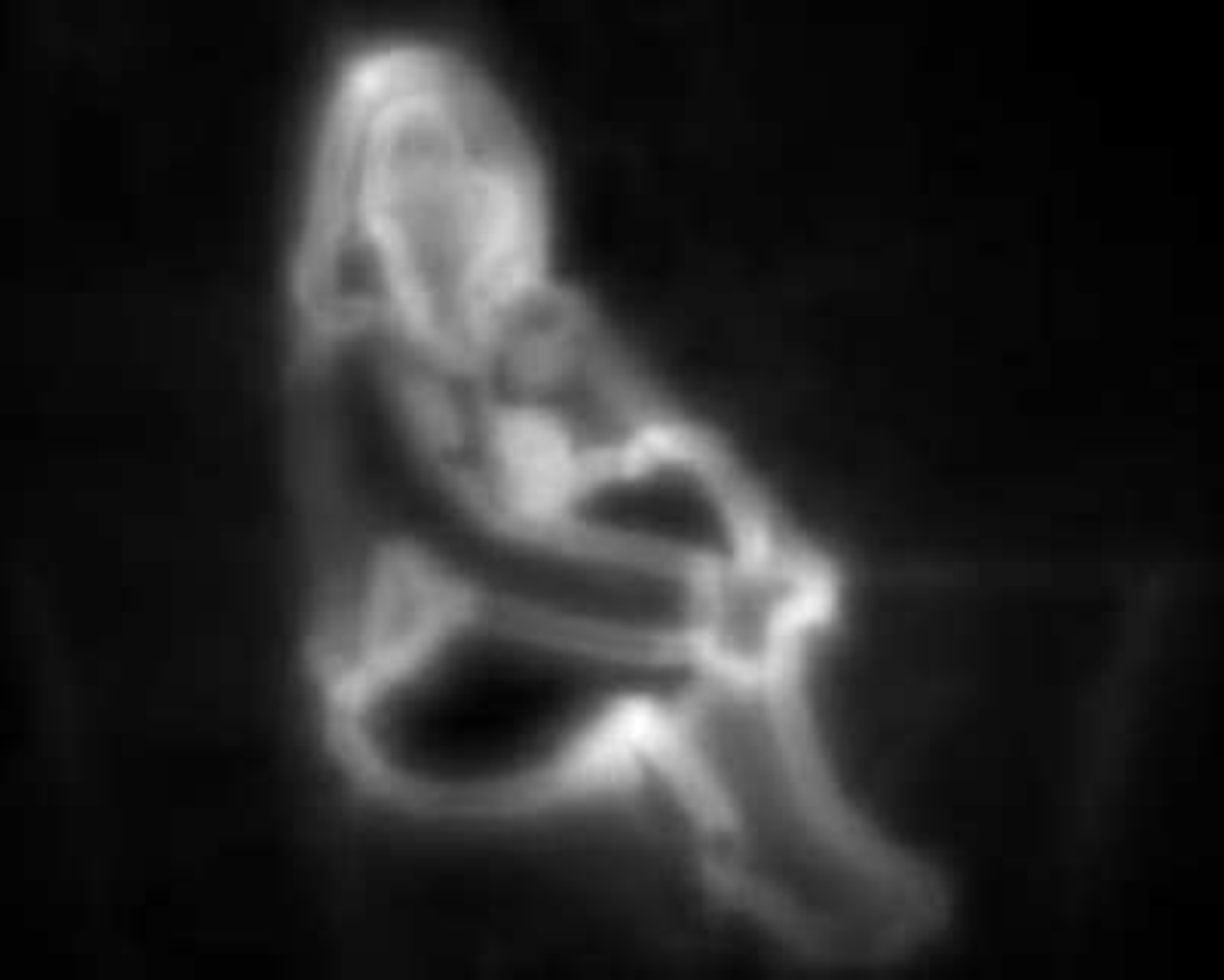}&
	\addExample{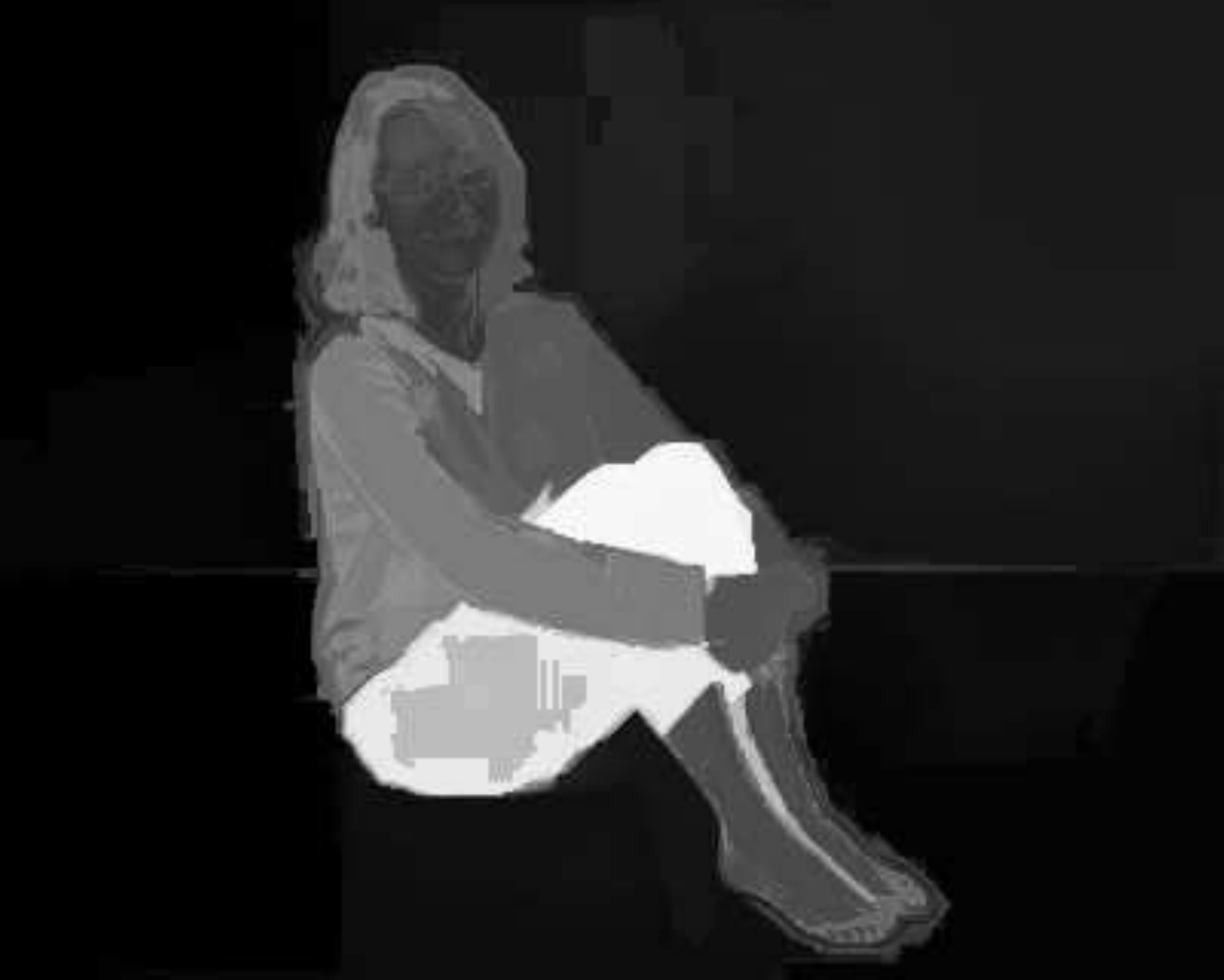}&
	\addExample{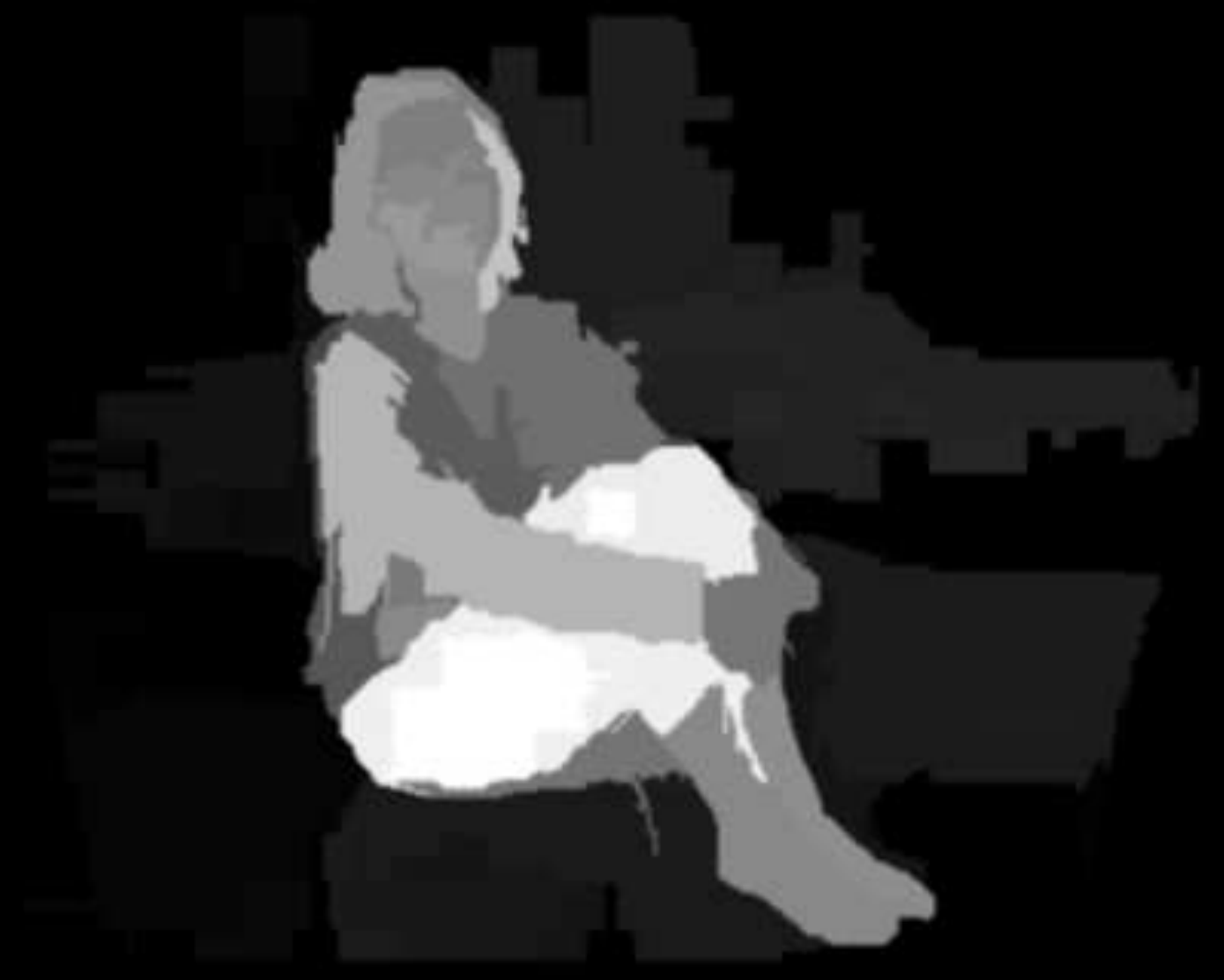}&
	\addExample{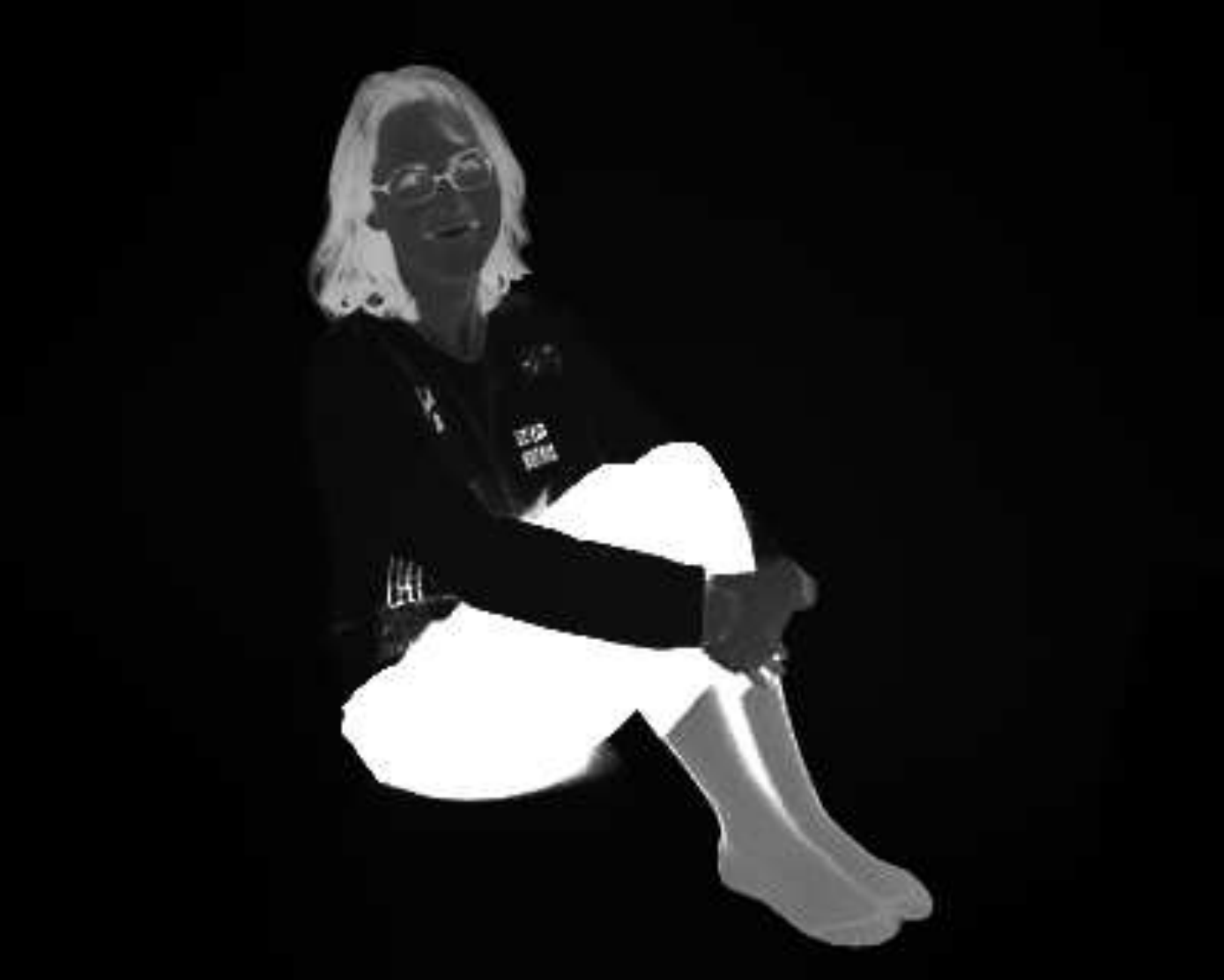}&
	\addExample{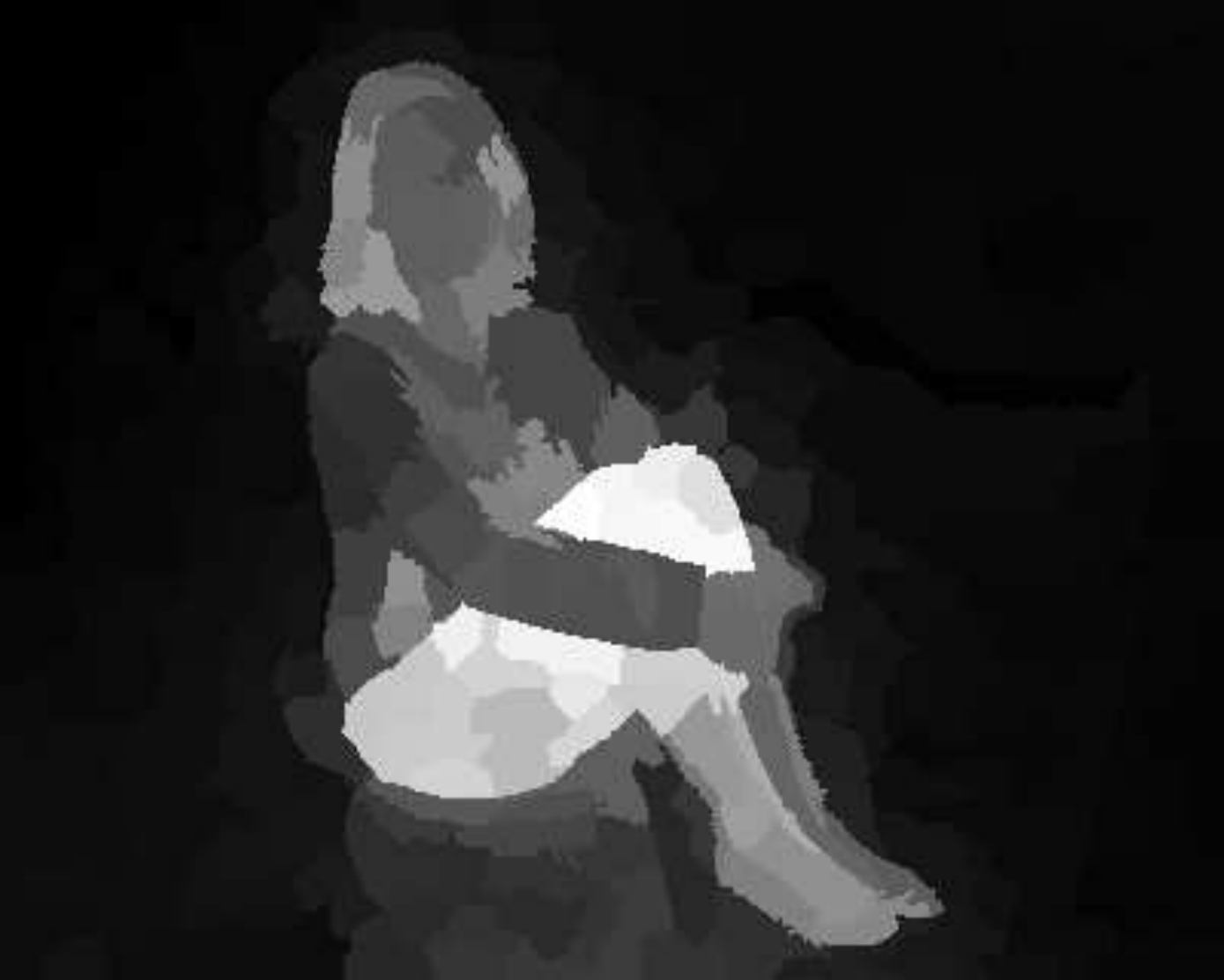}&
	\addExample{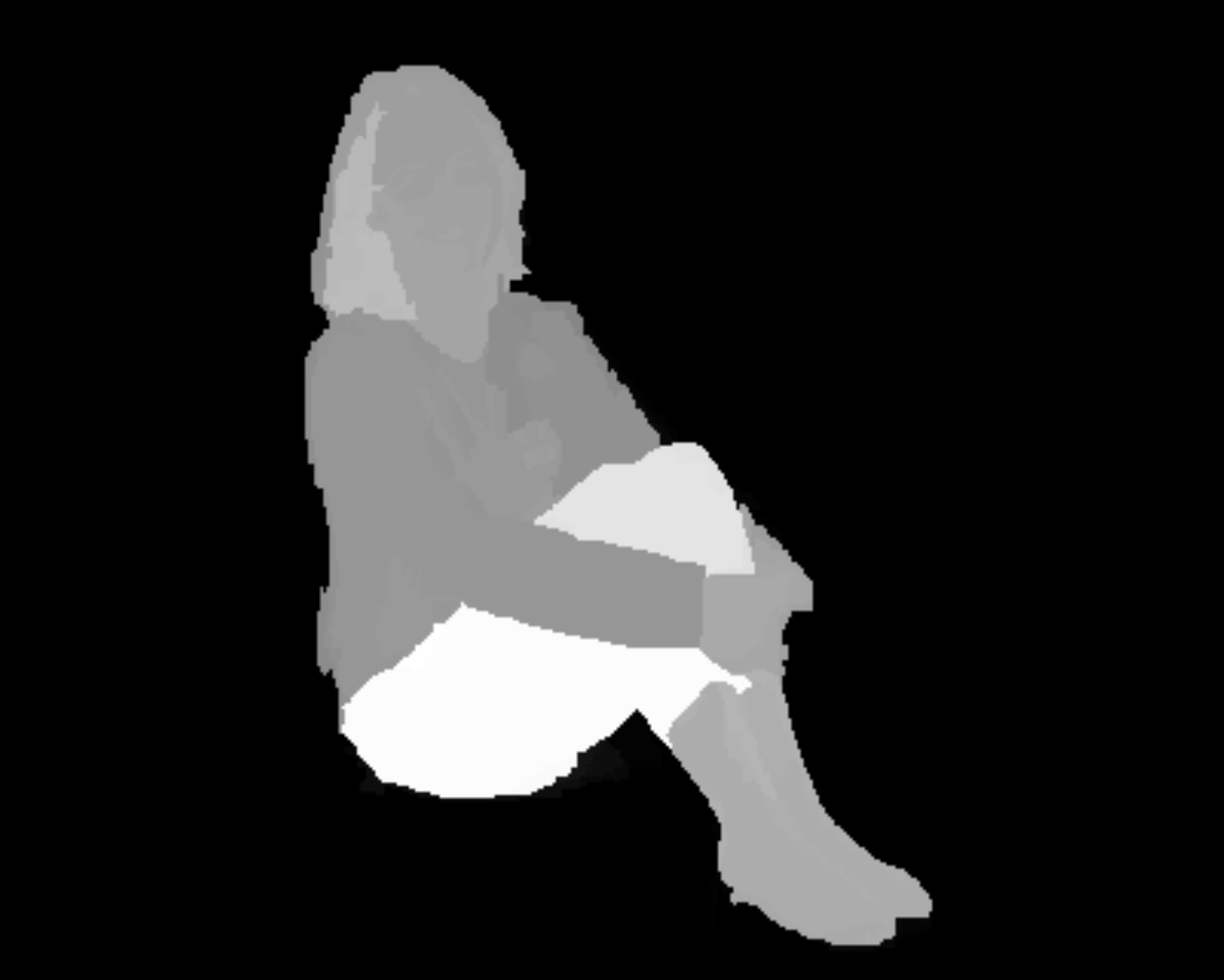}&
	\addExample{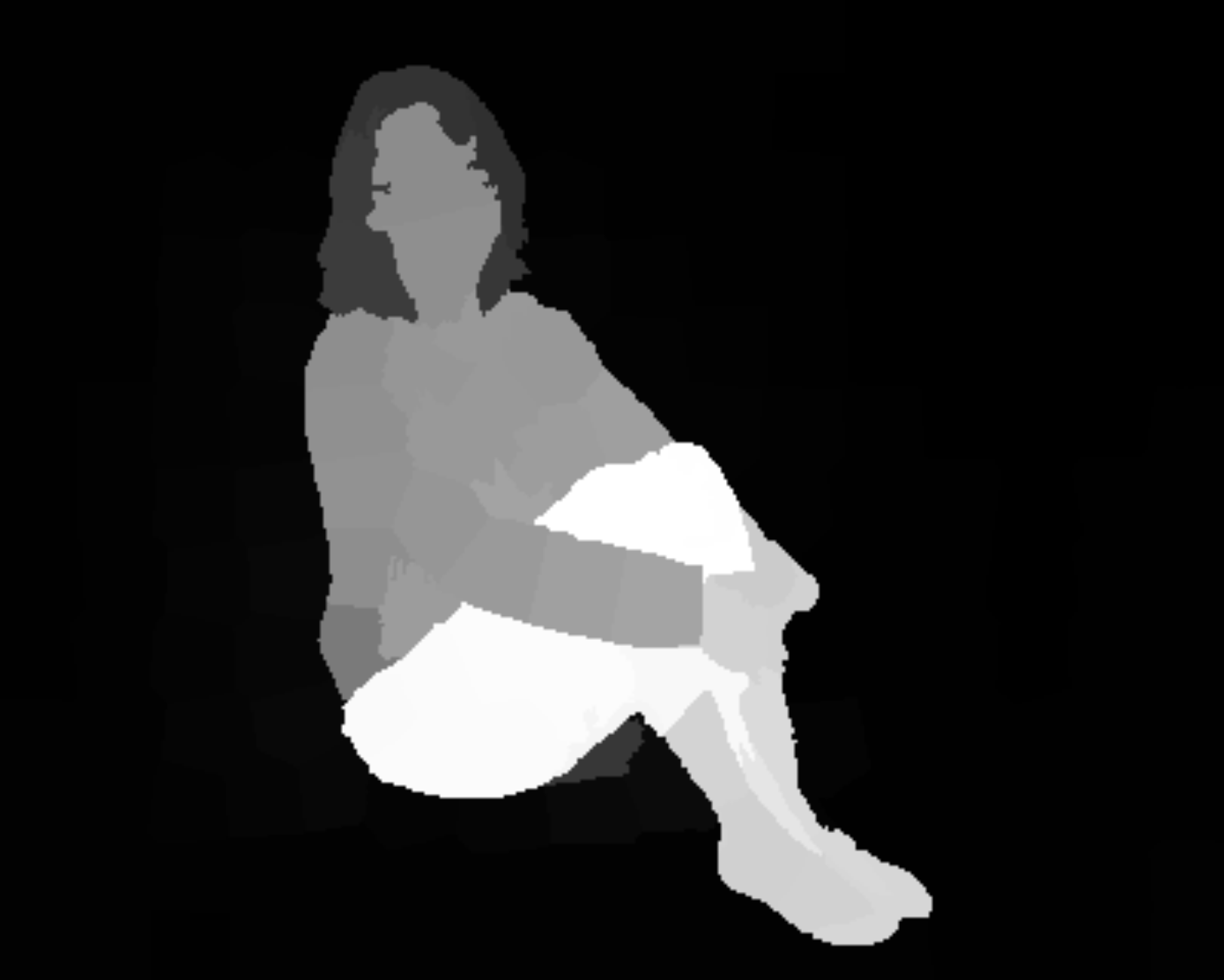}&
	\addExample{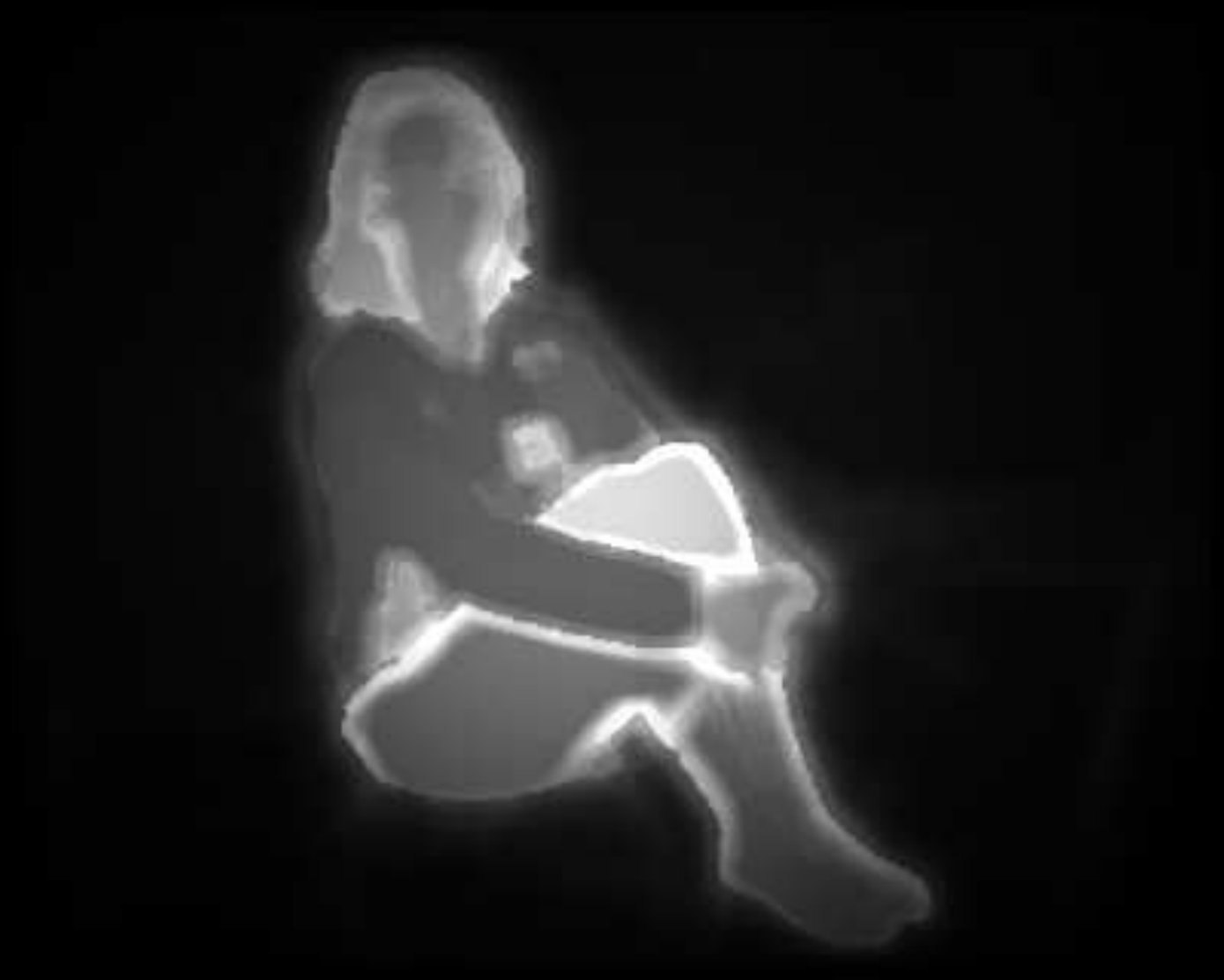}&
	\addExample{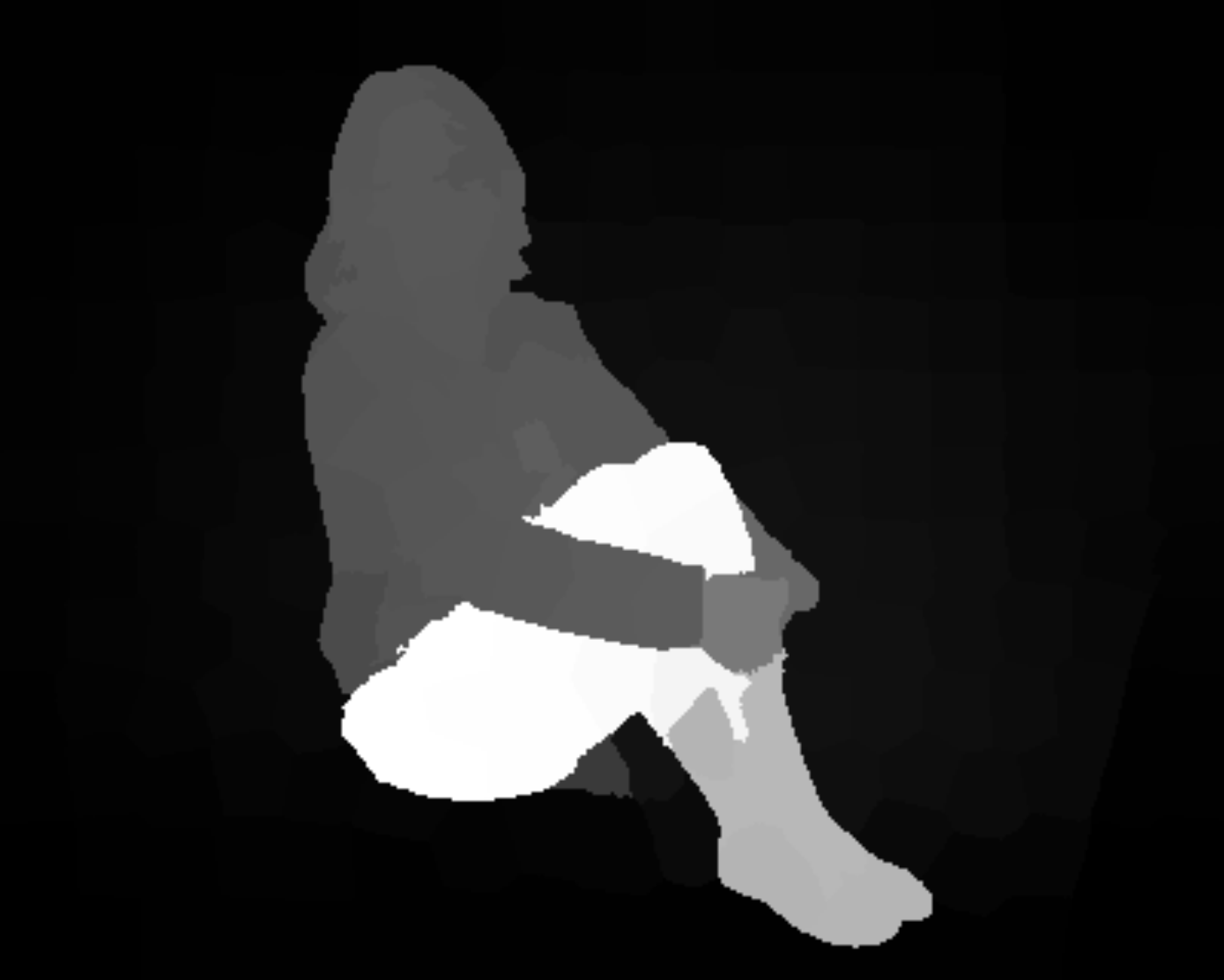}&
	\addExample{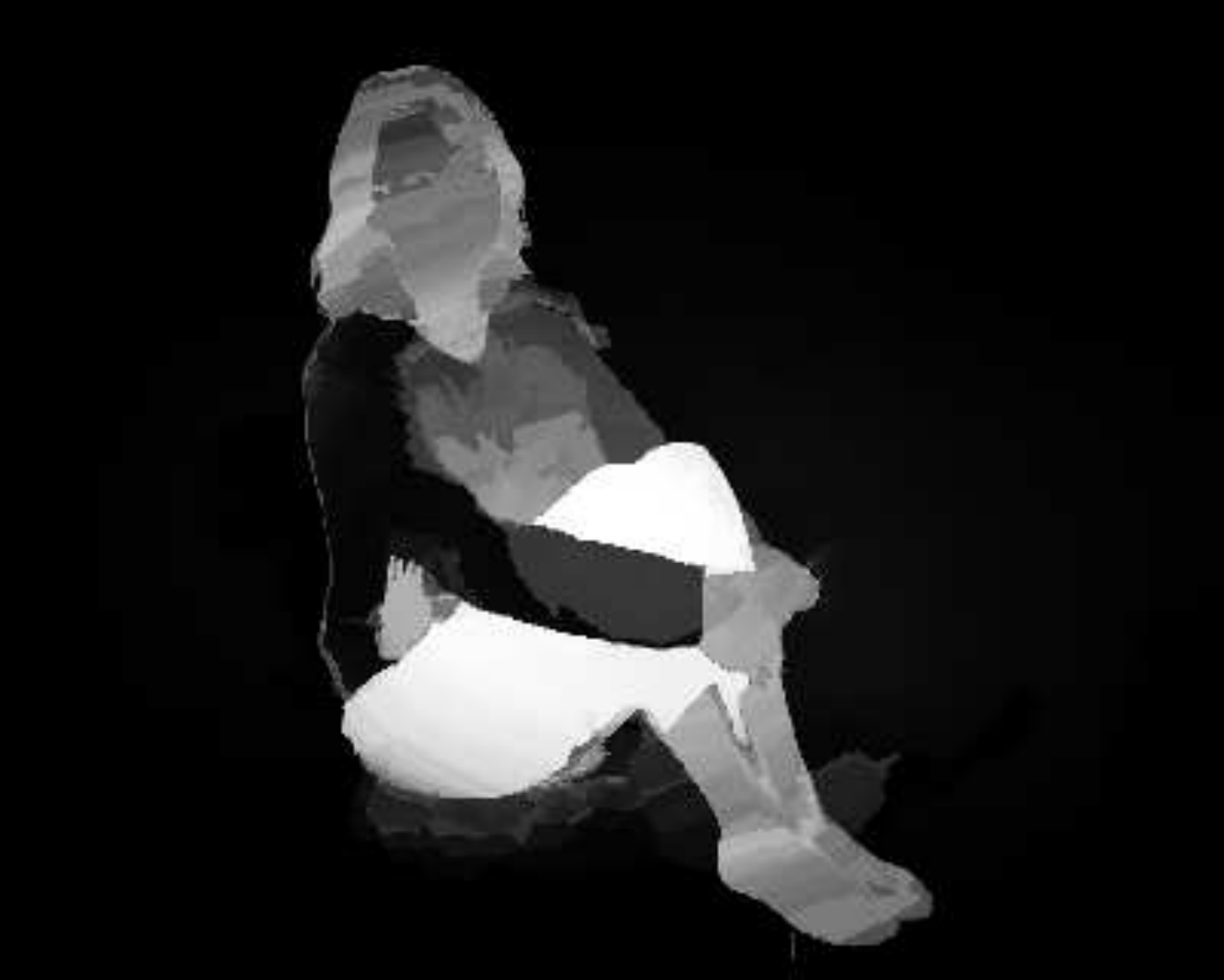}&
	\addExample{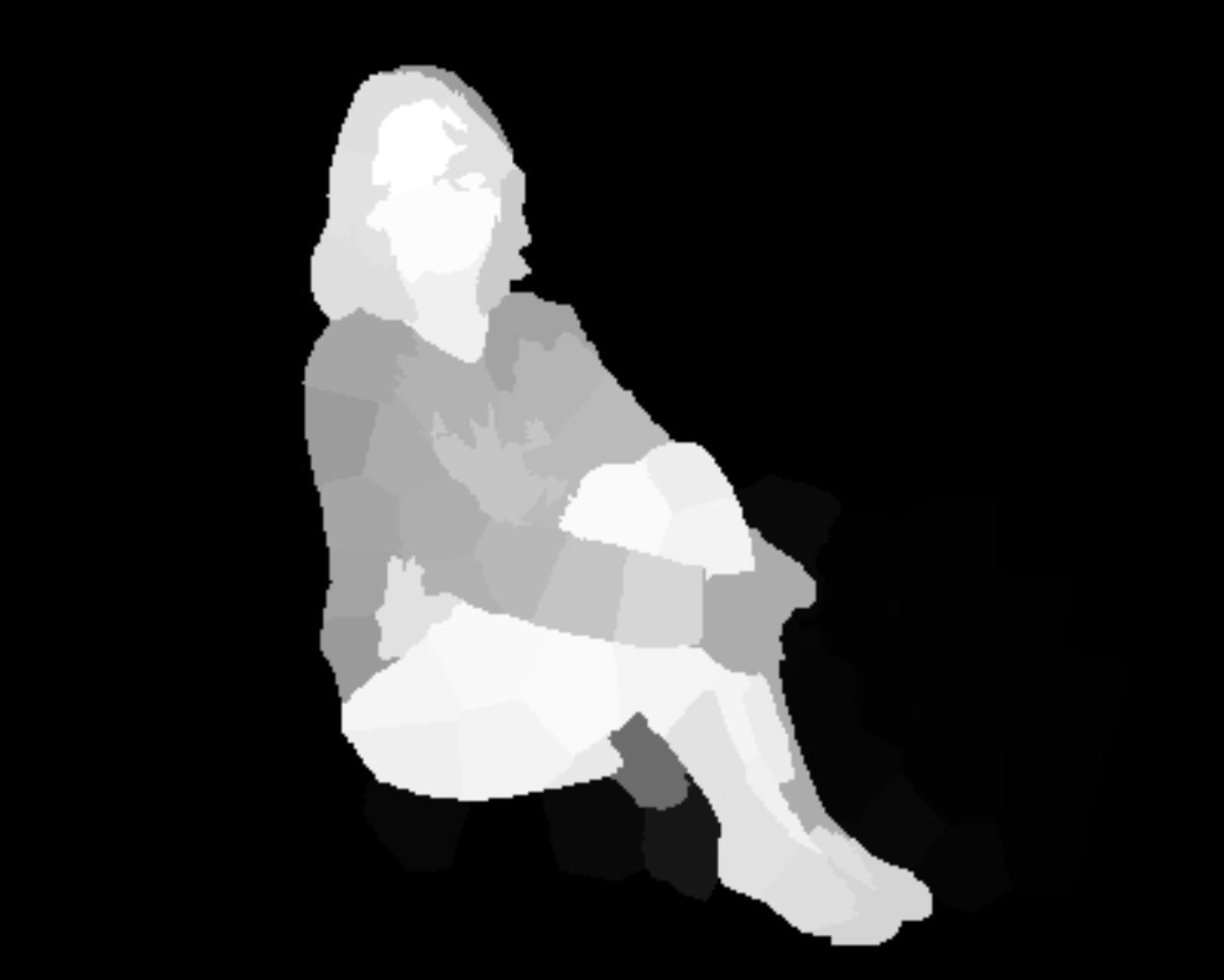}&
	\addExample{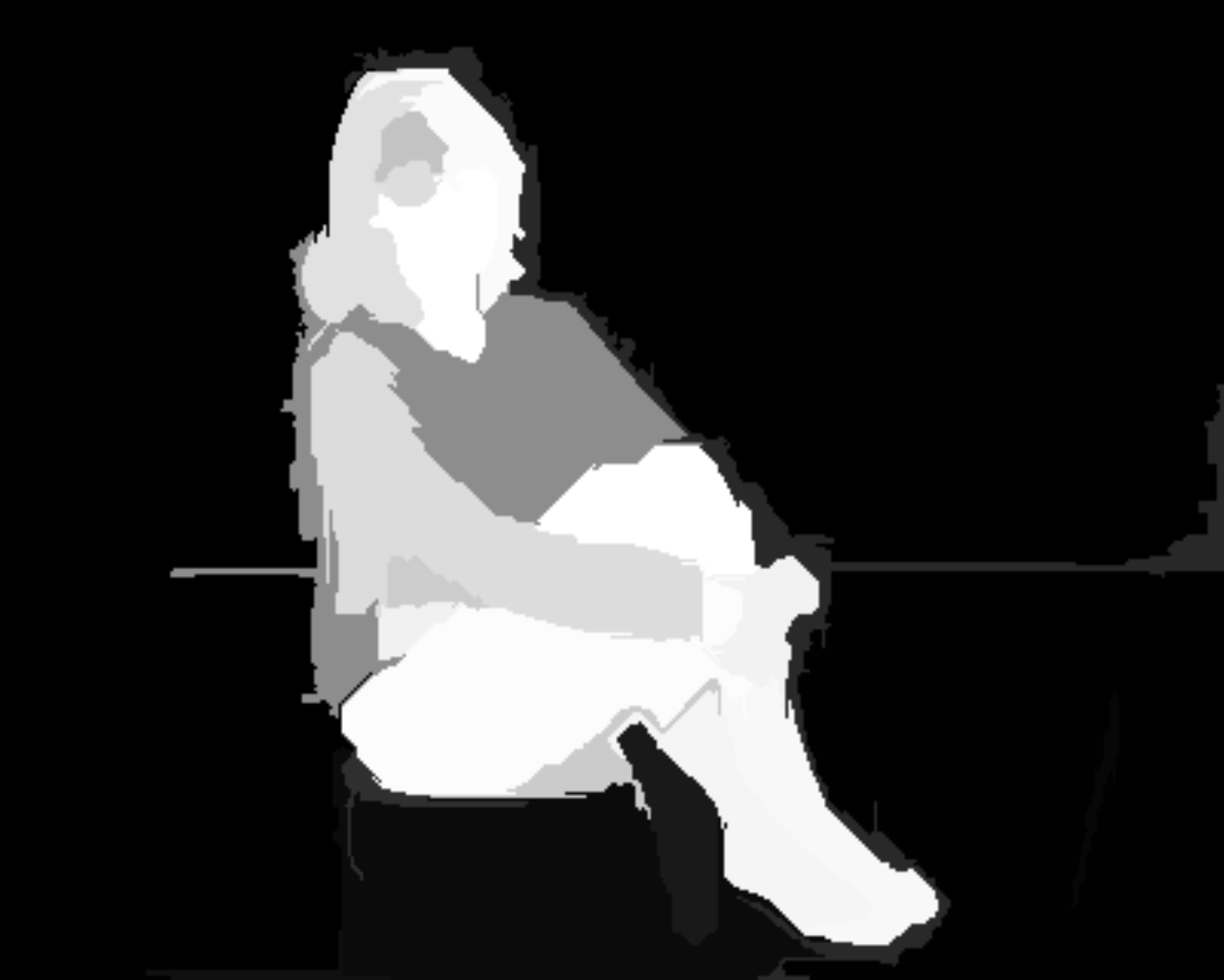}&
	\addExample{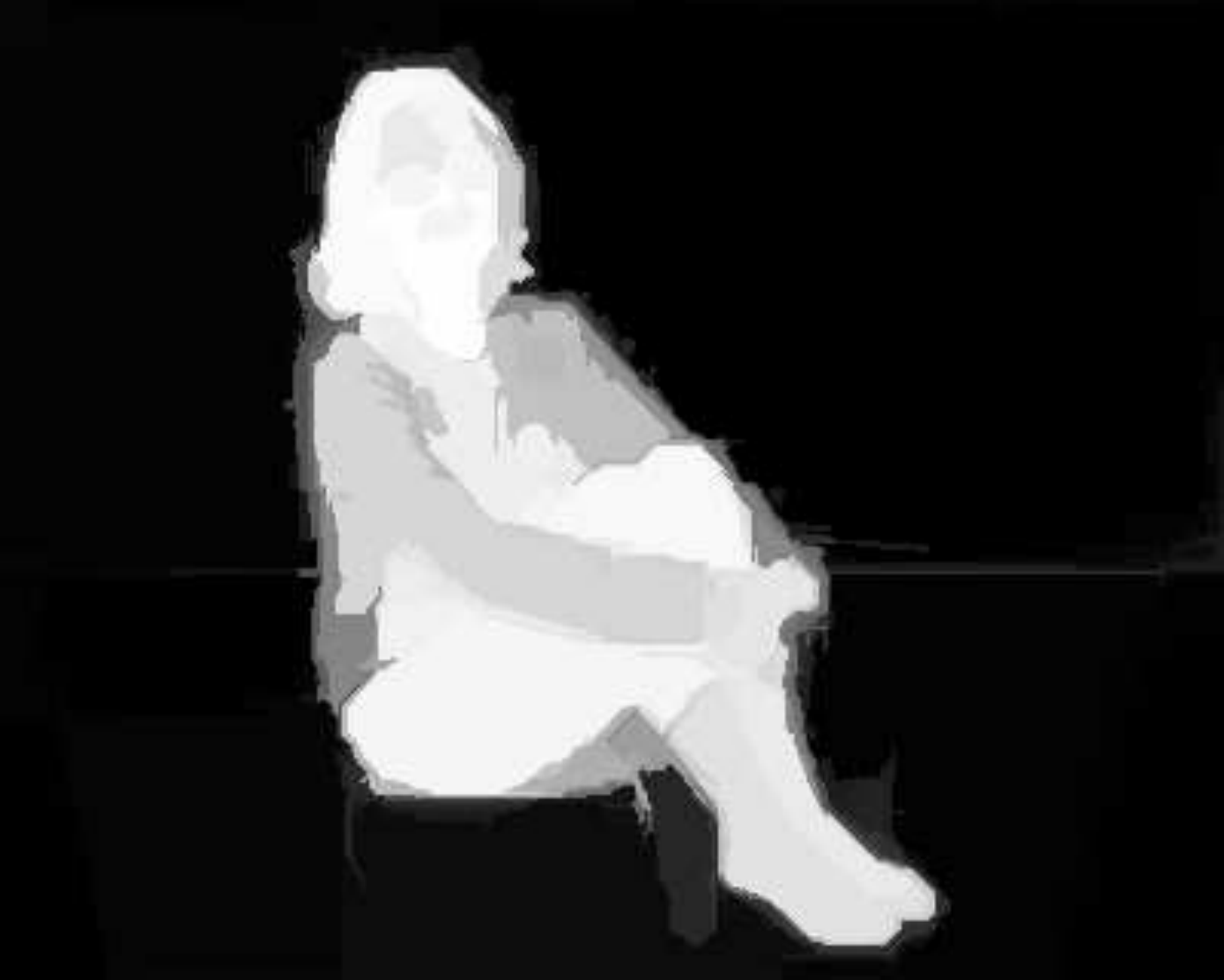}
    \\
	\addExample{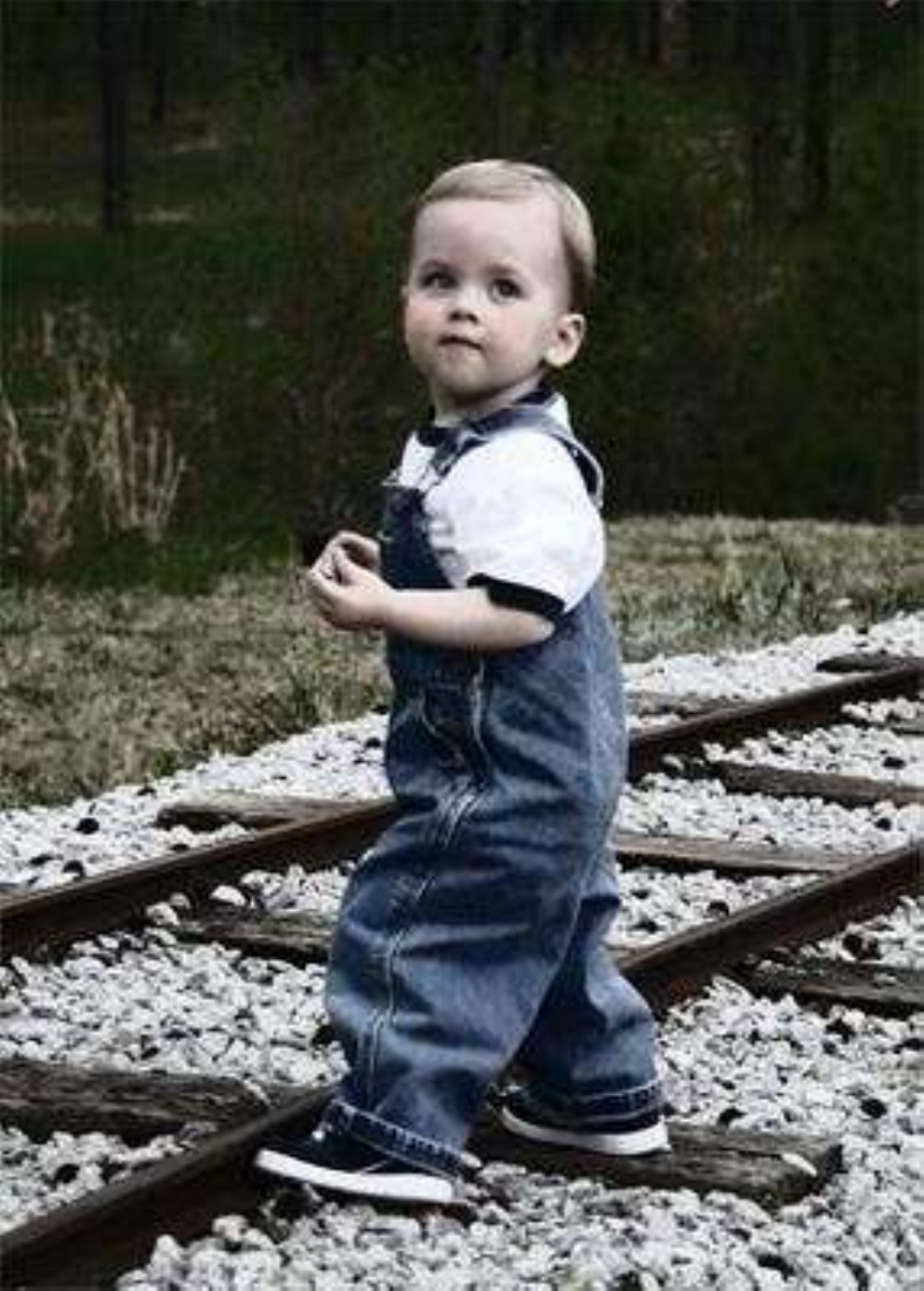}&
	\addExample{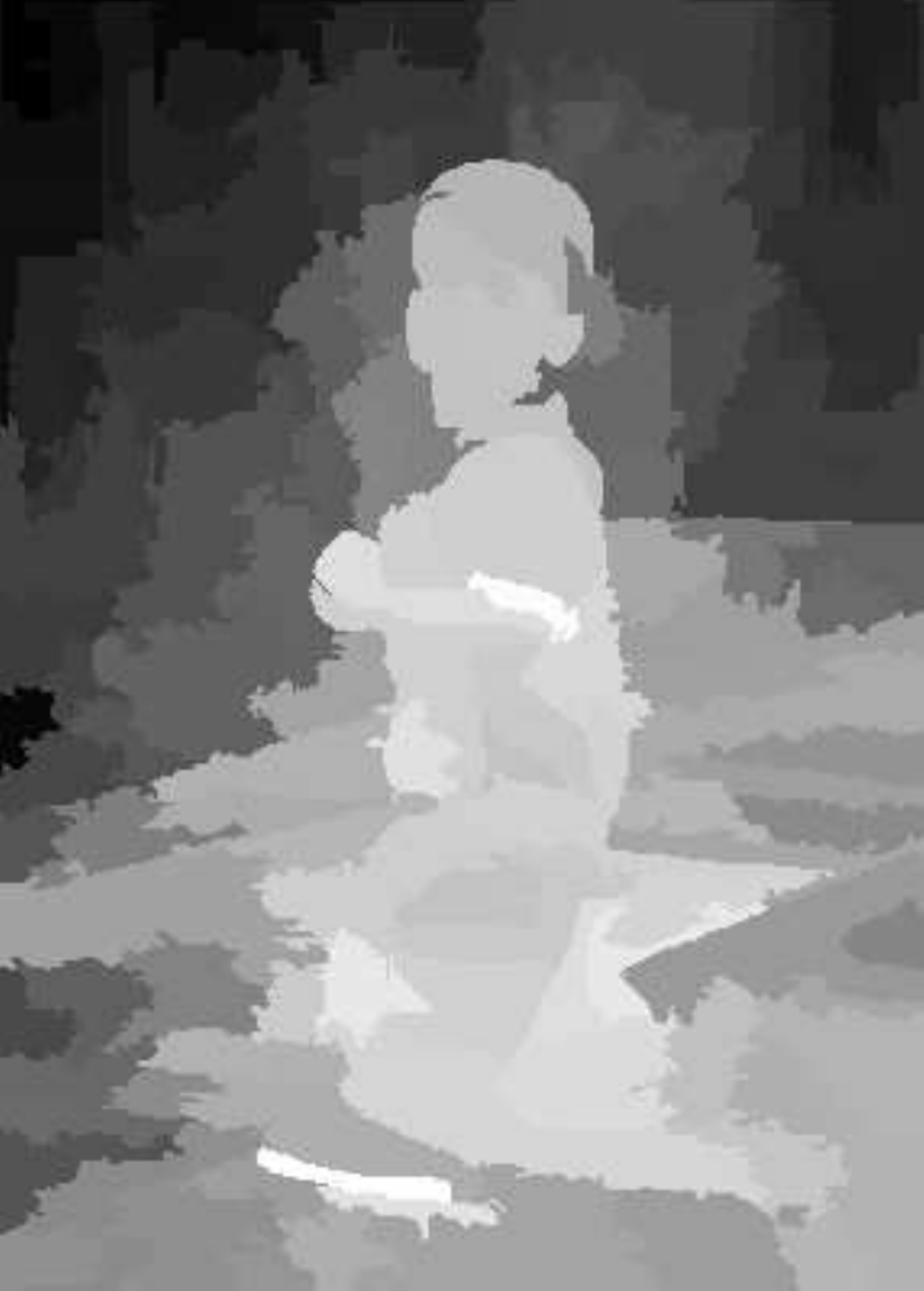}&
	\addExample{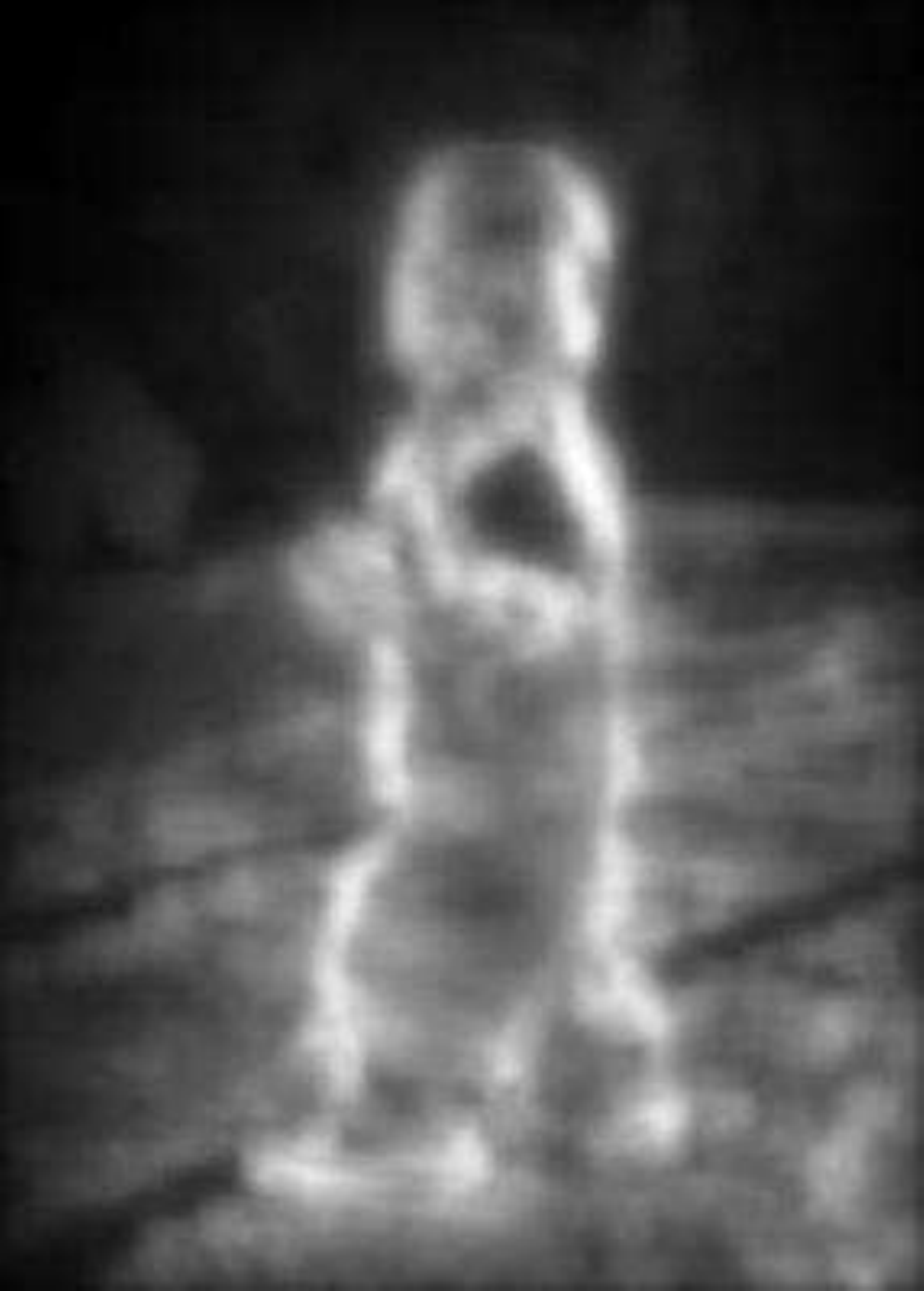}&
	\addExample{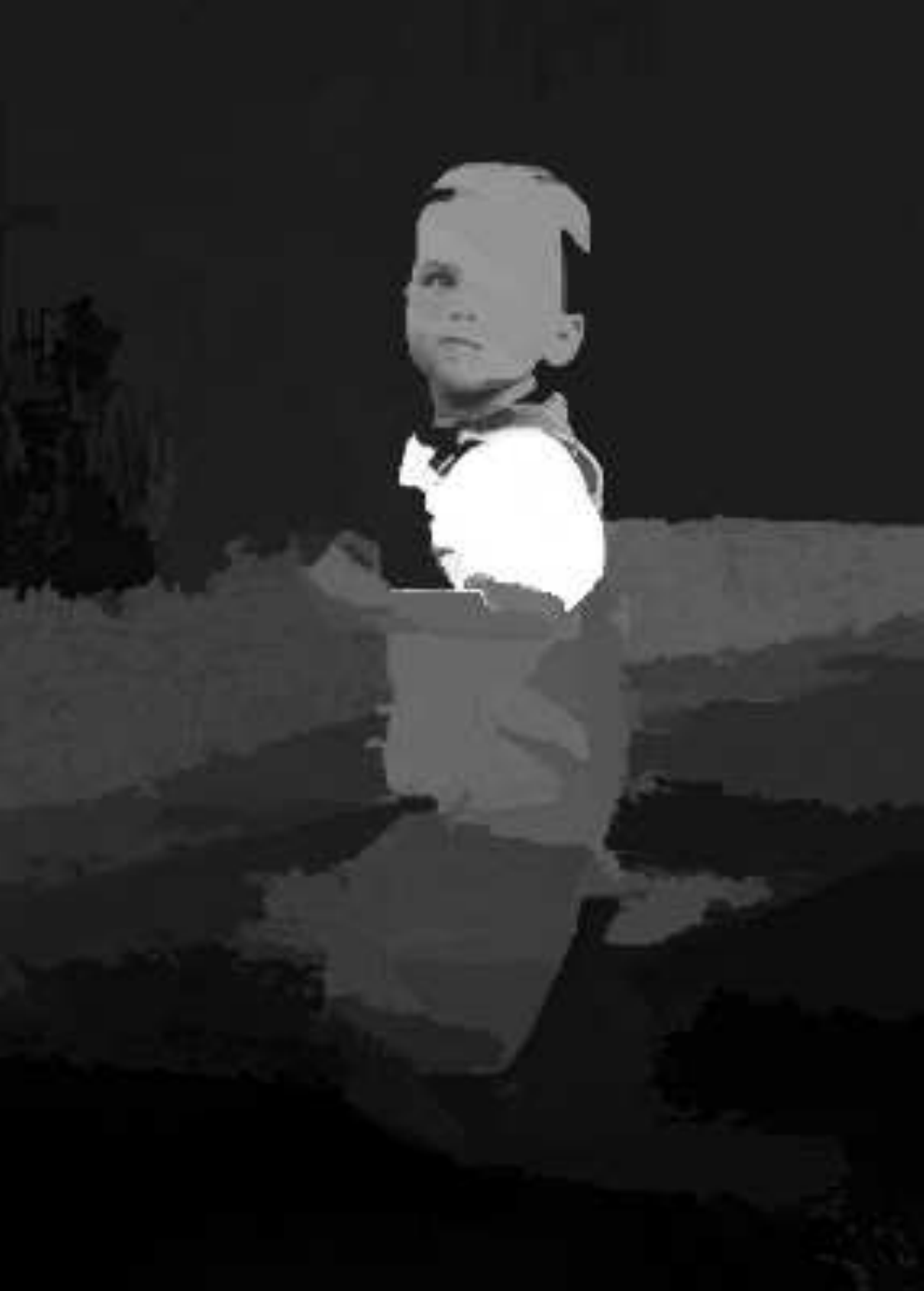}&
	\addExample{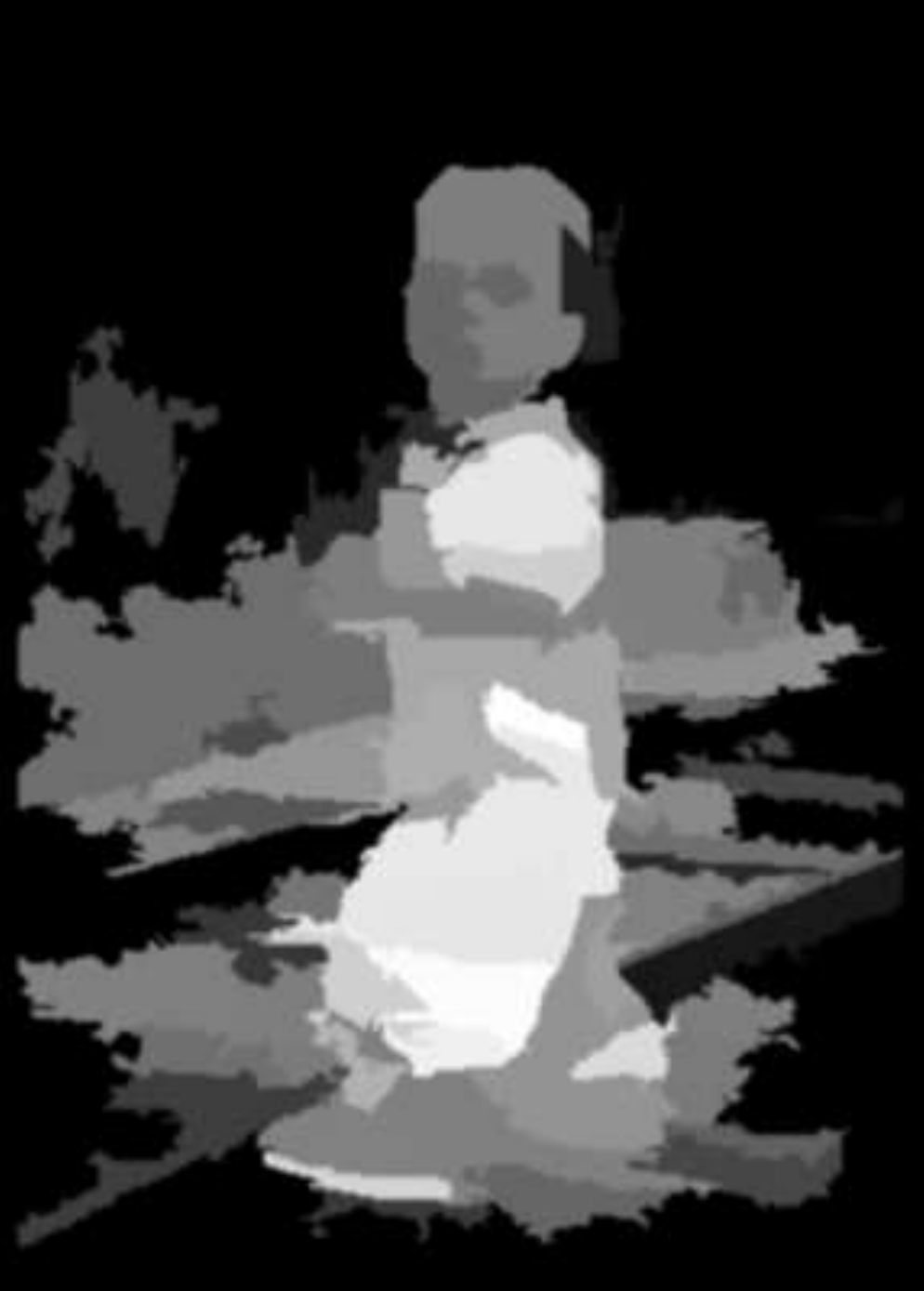}&
	\addExample{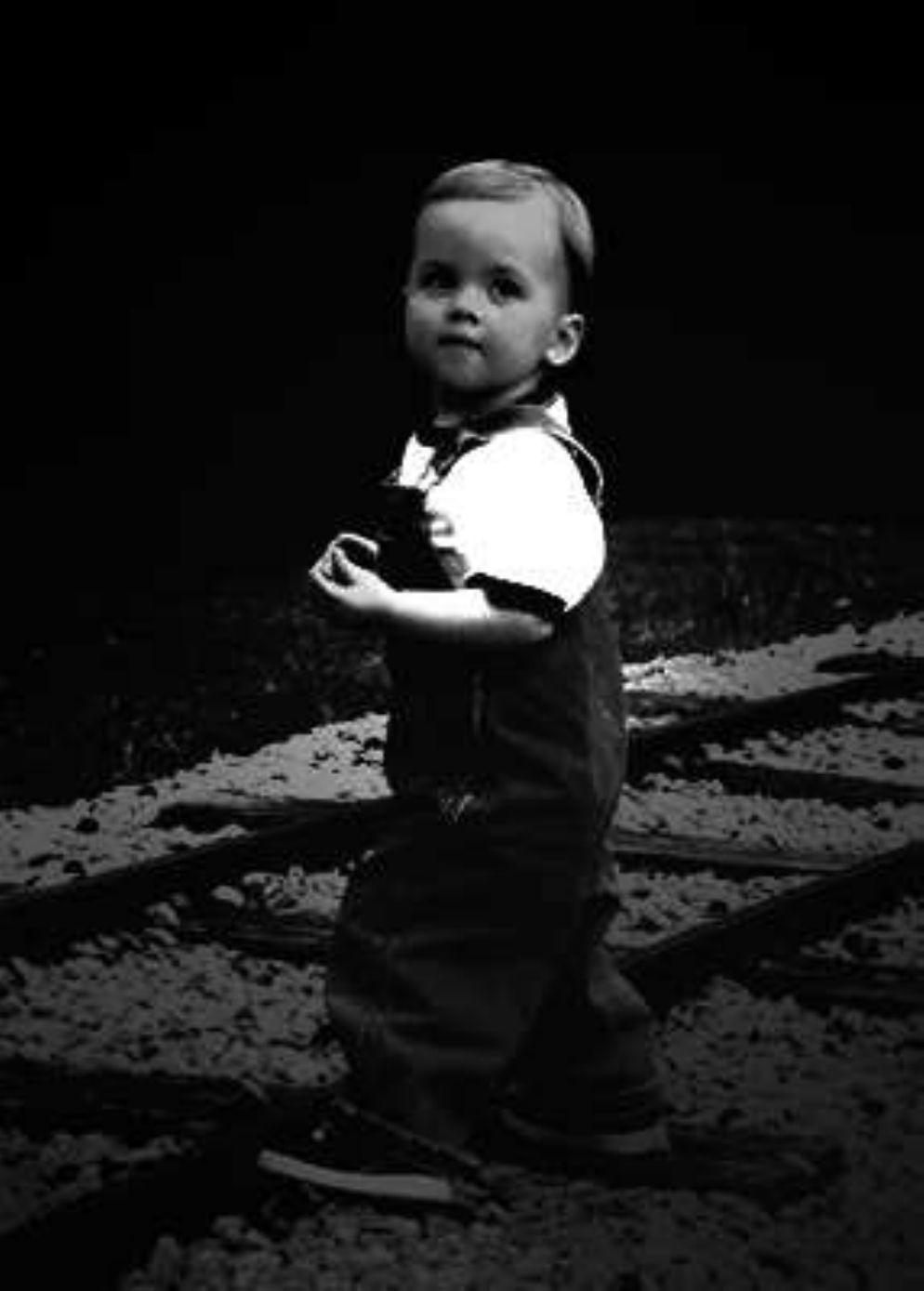}&
	\addExample{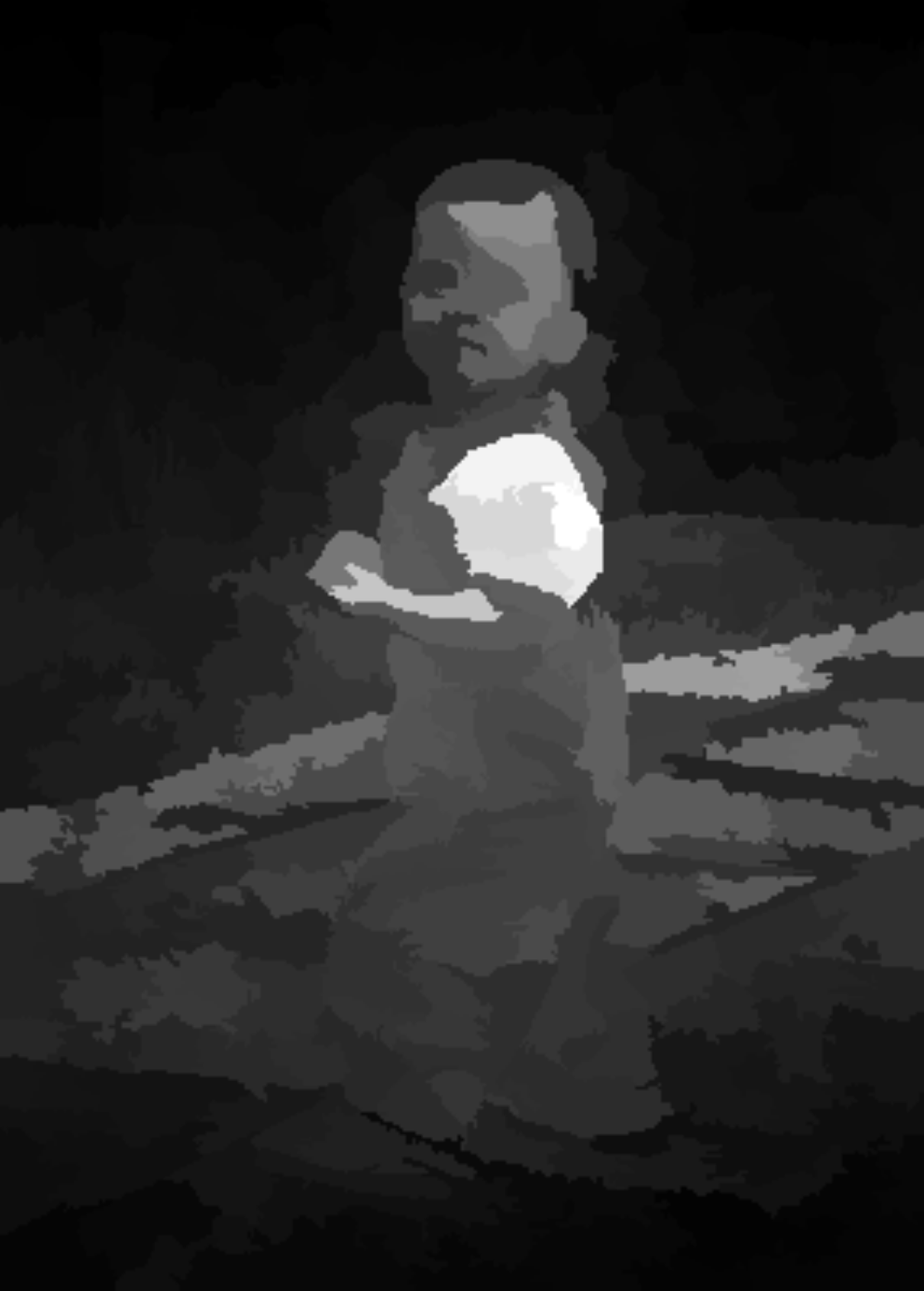}&
	\addExample{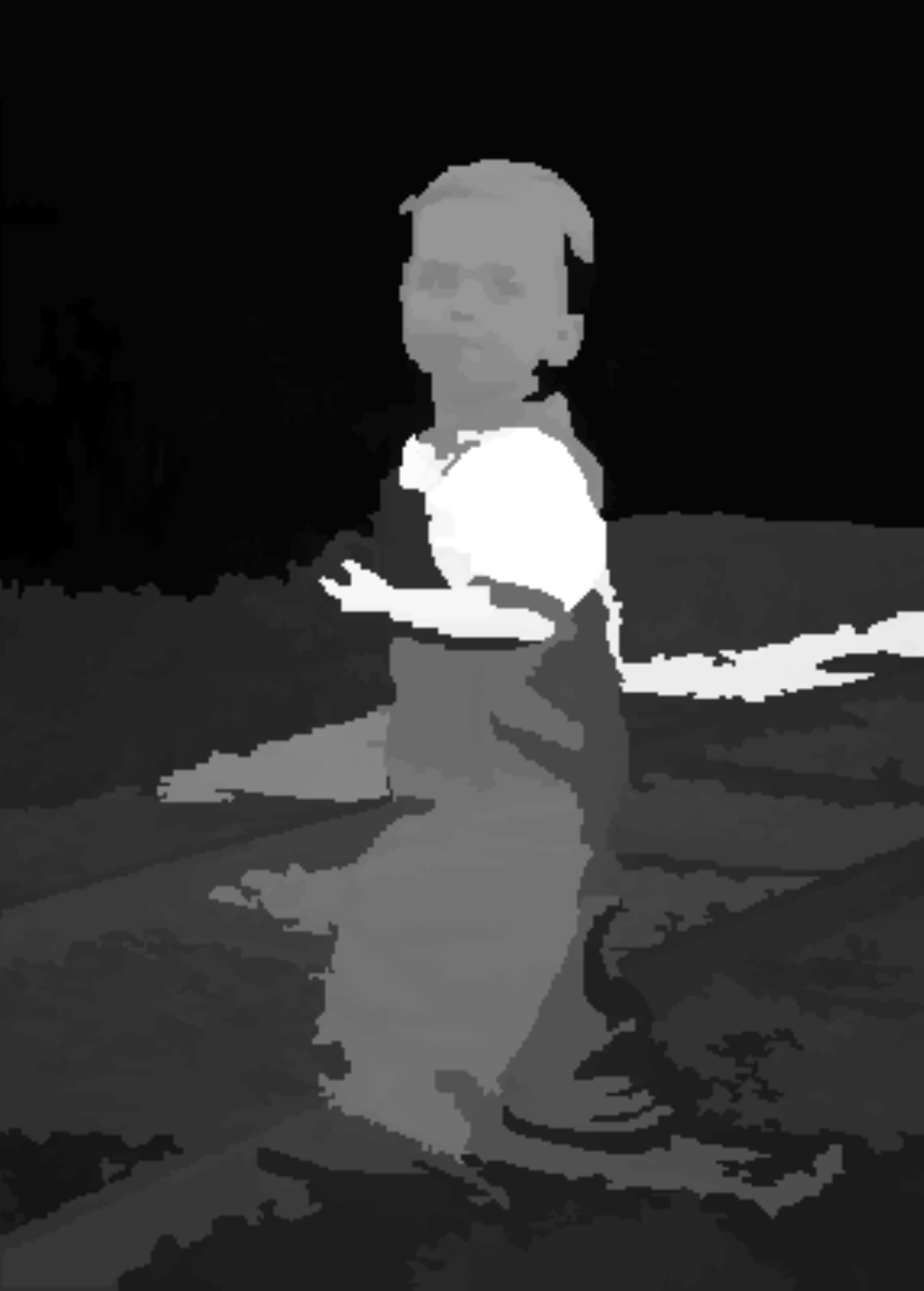}&
	\addExample{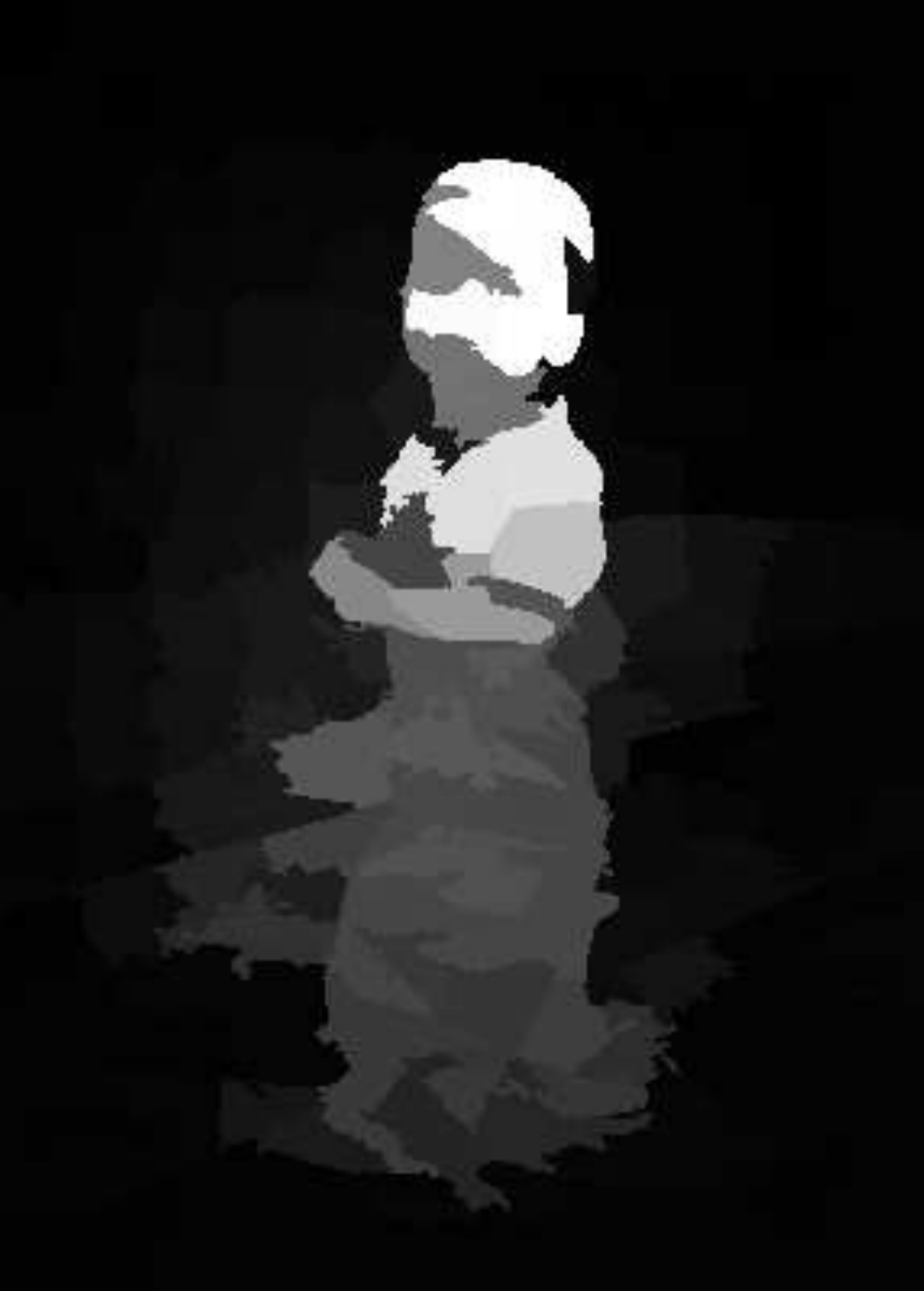}&
	\addExample{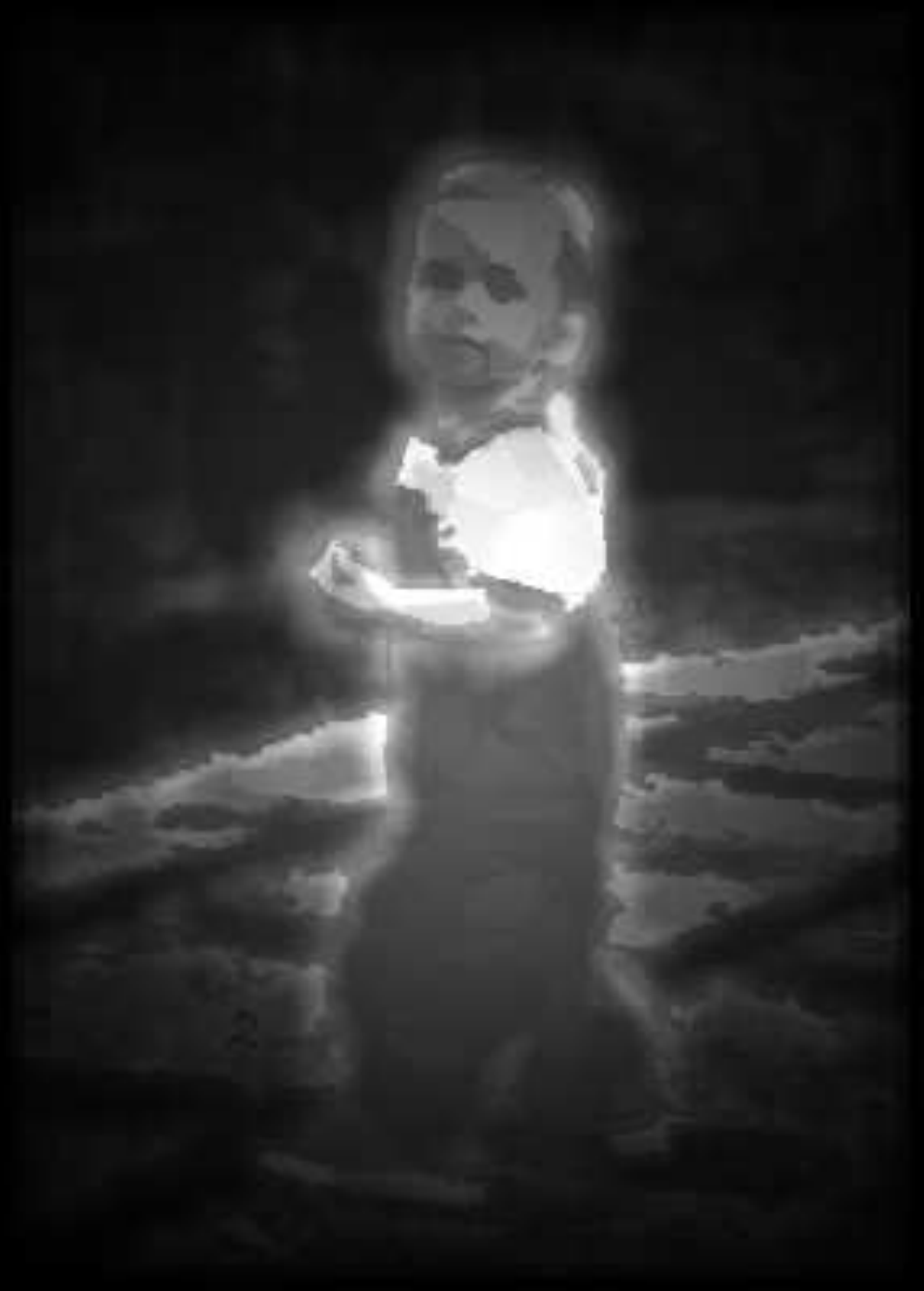}&
	\addExample{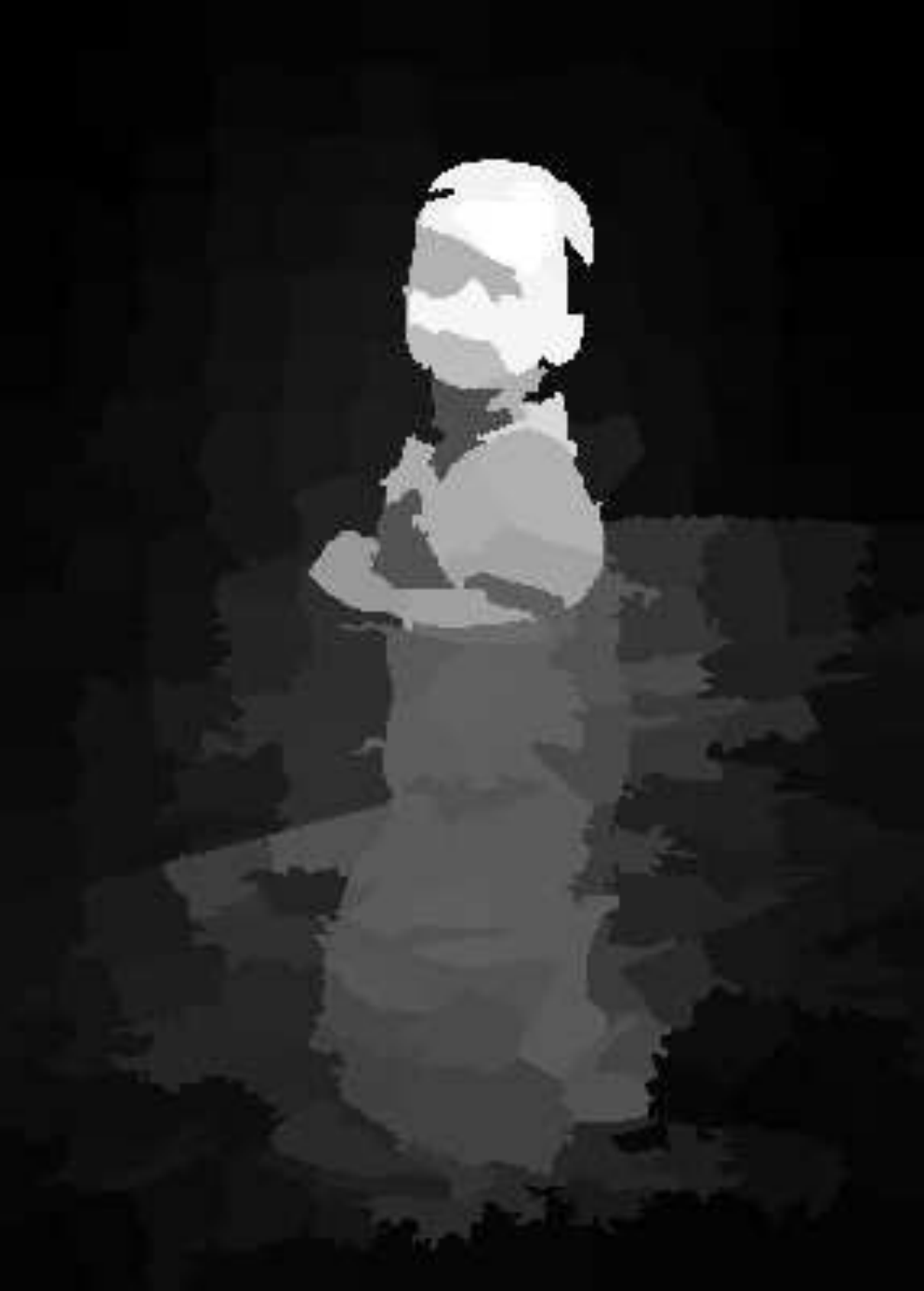}&
	\addExample{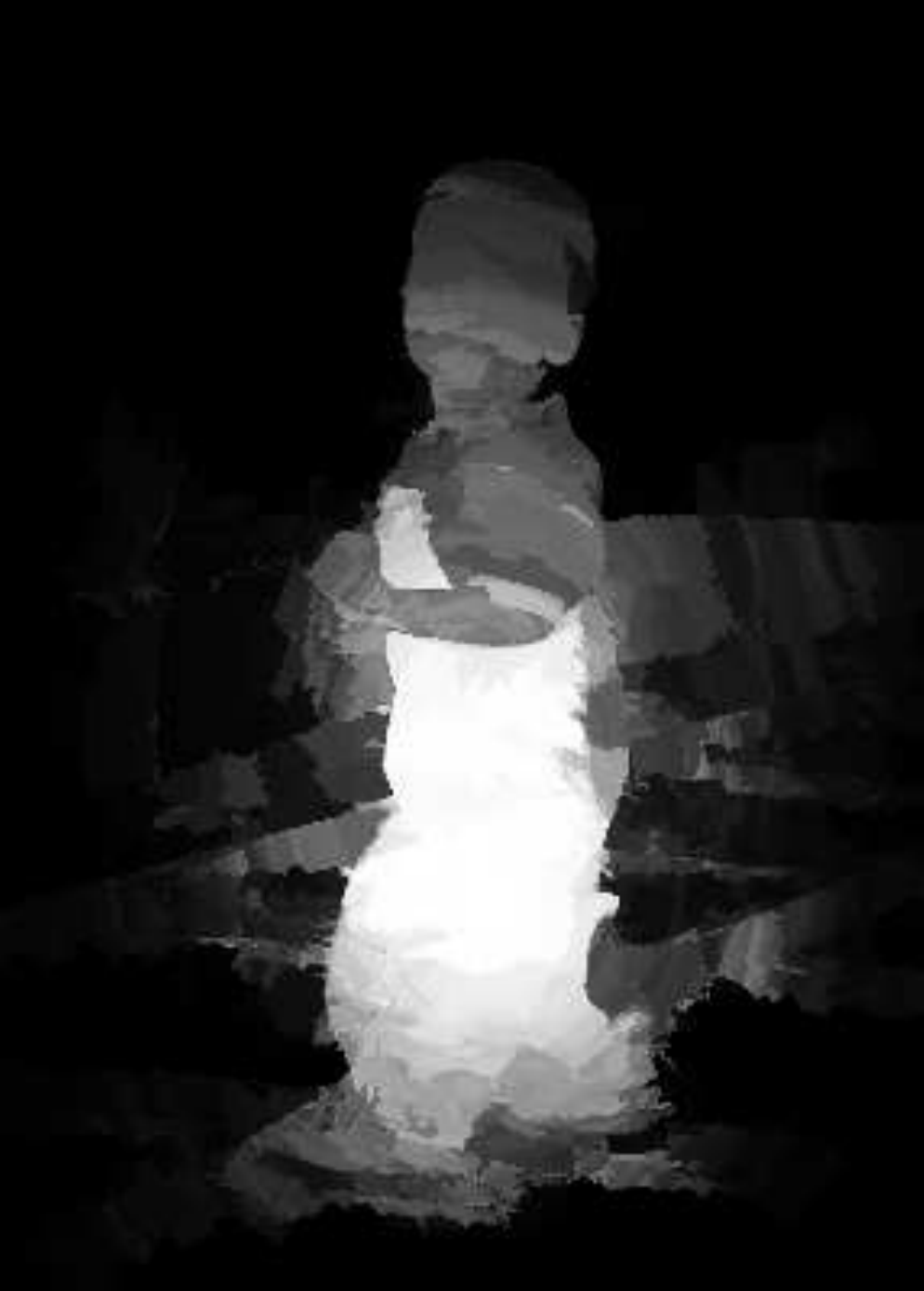}&
	\addExample{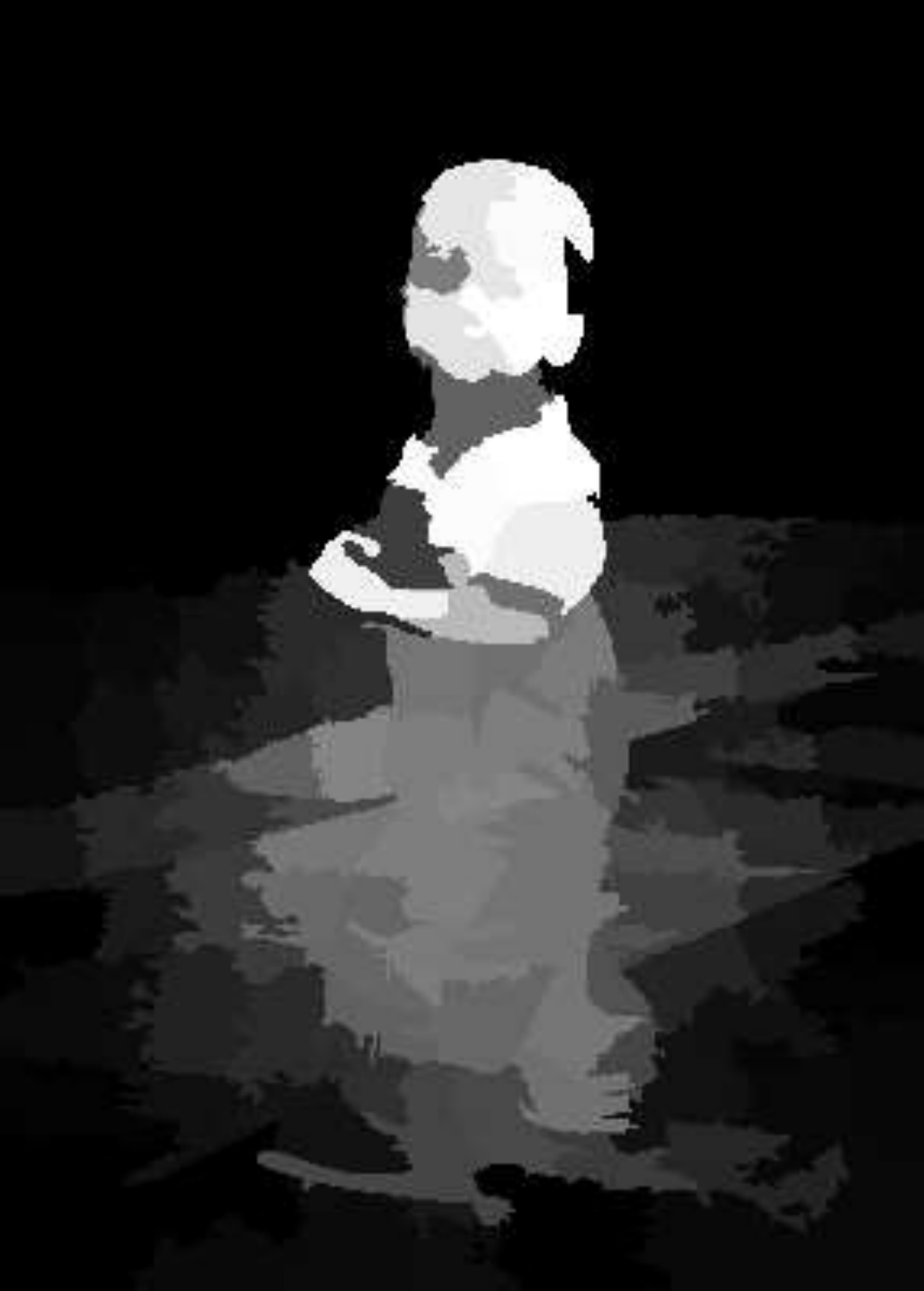}&
	\addExample{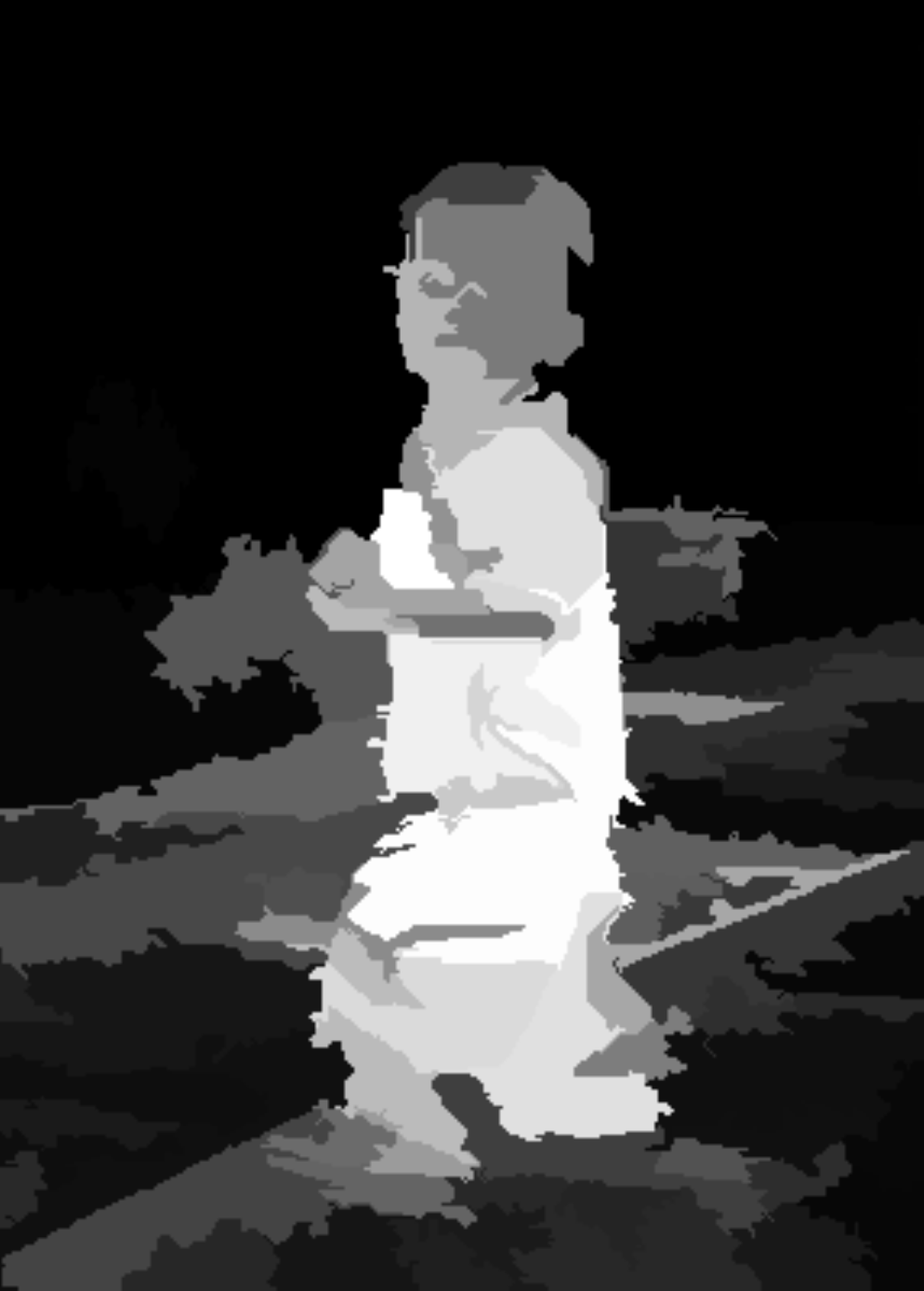}&
	\addExample{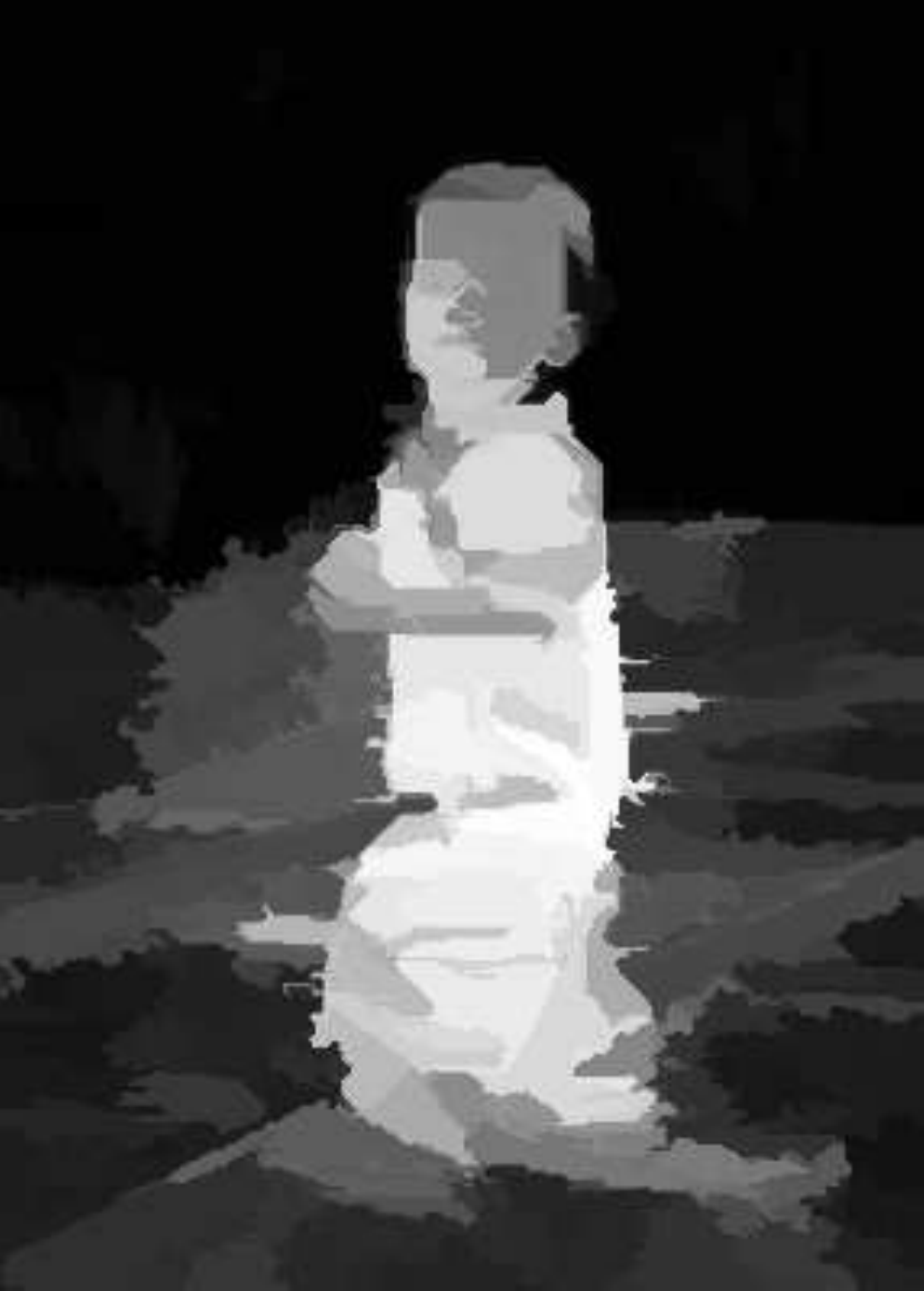}
    \\
	\addExample{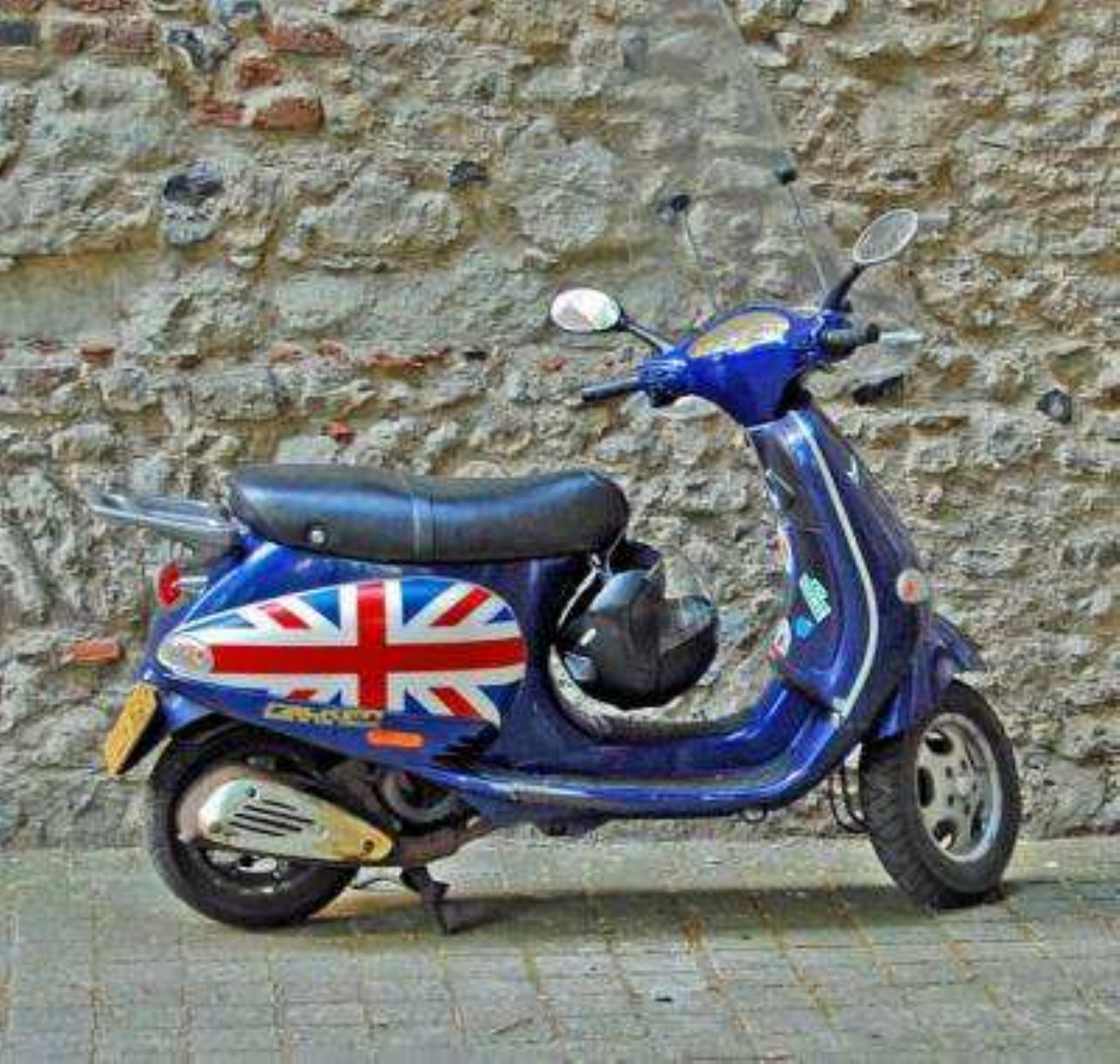}&
	\addExample{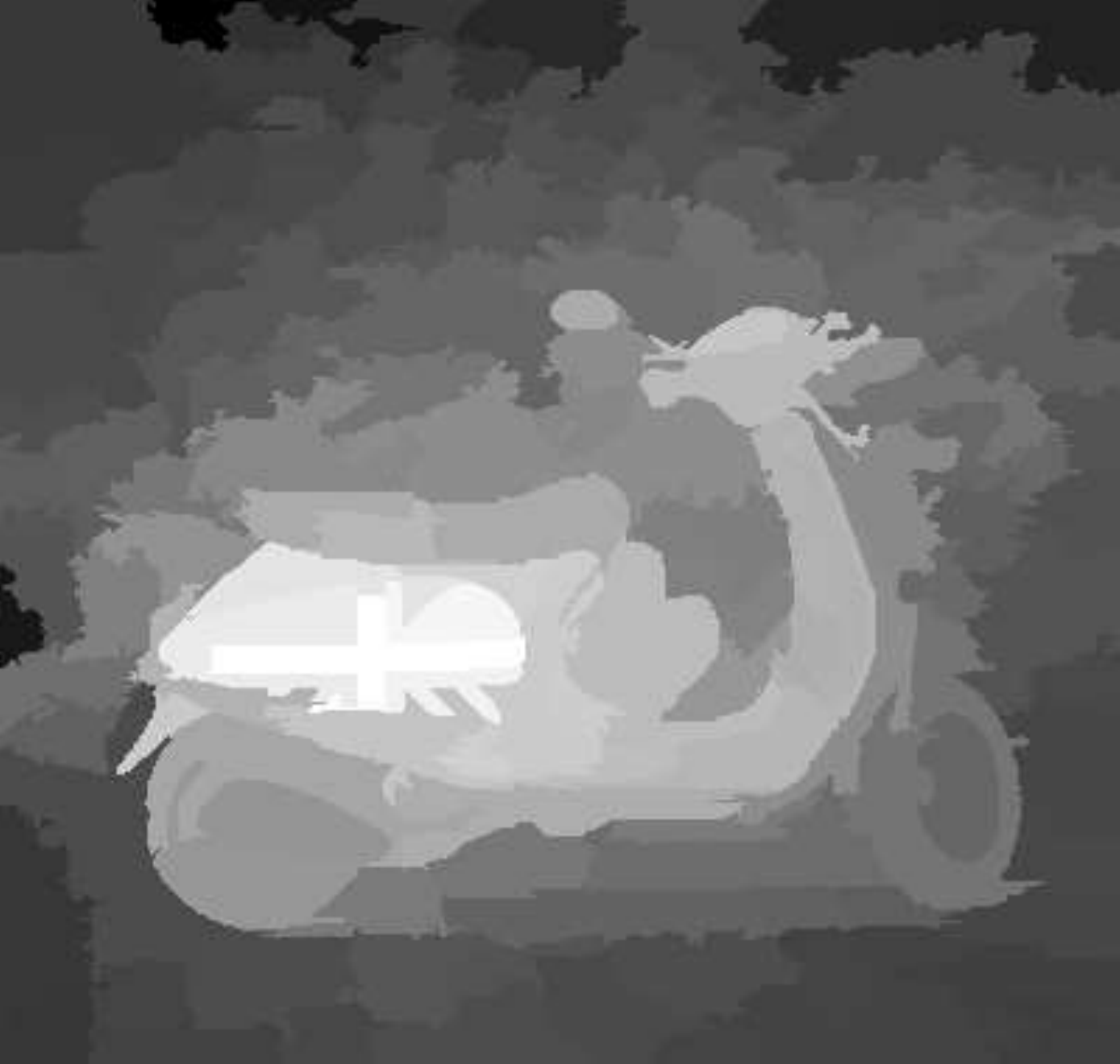}&
	\addExample{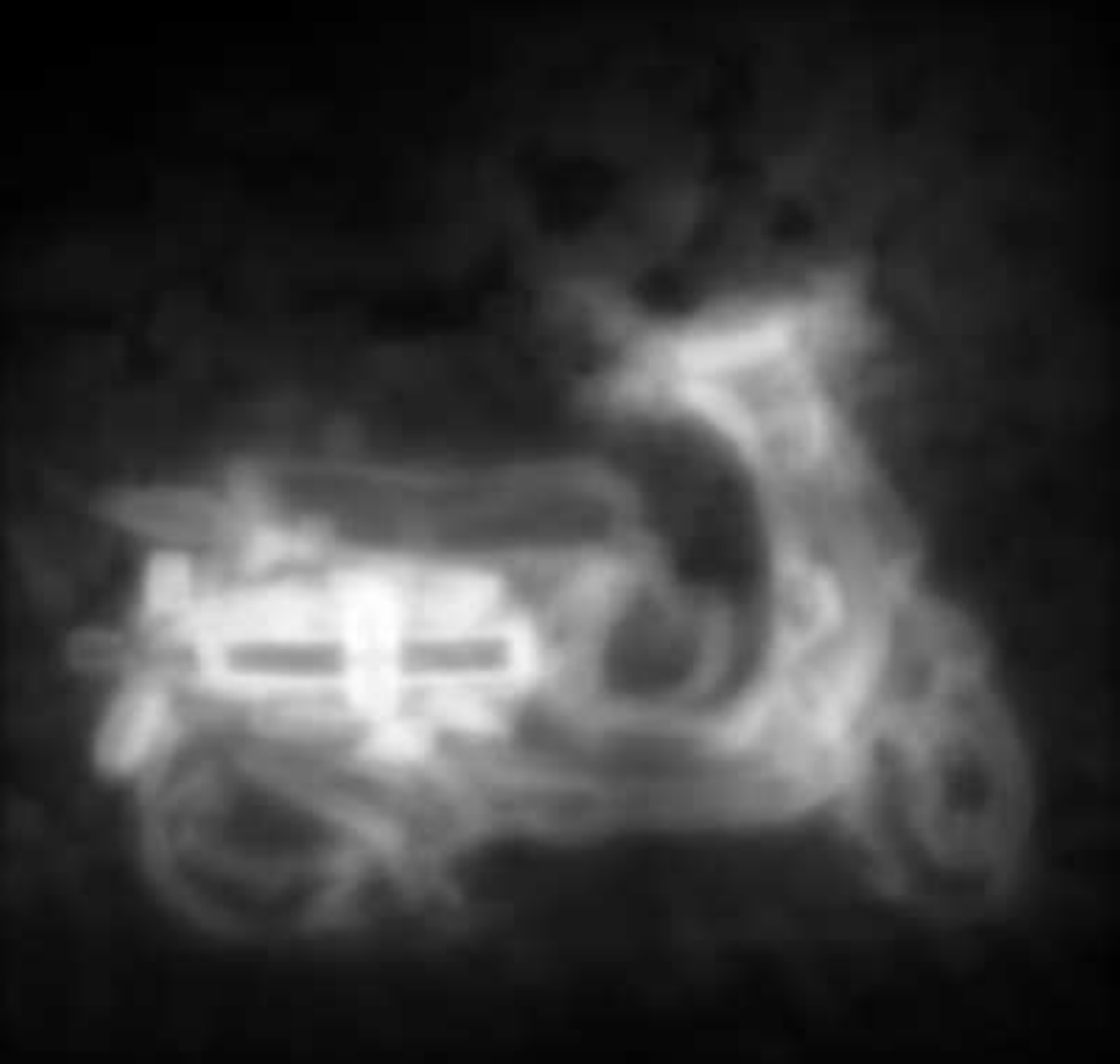}&
	\addExample{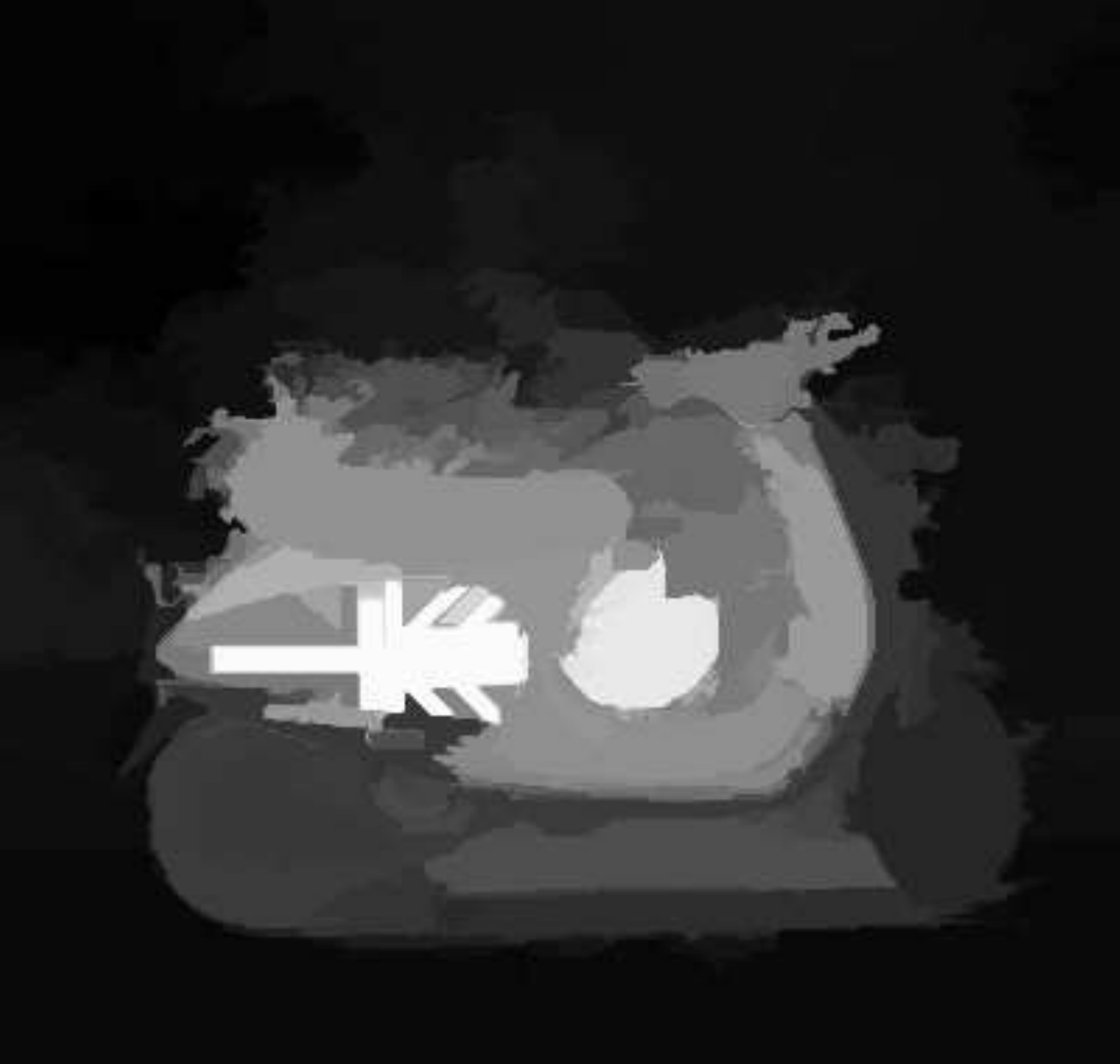}&
	\addExample{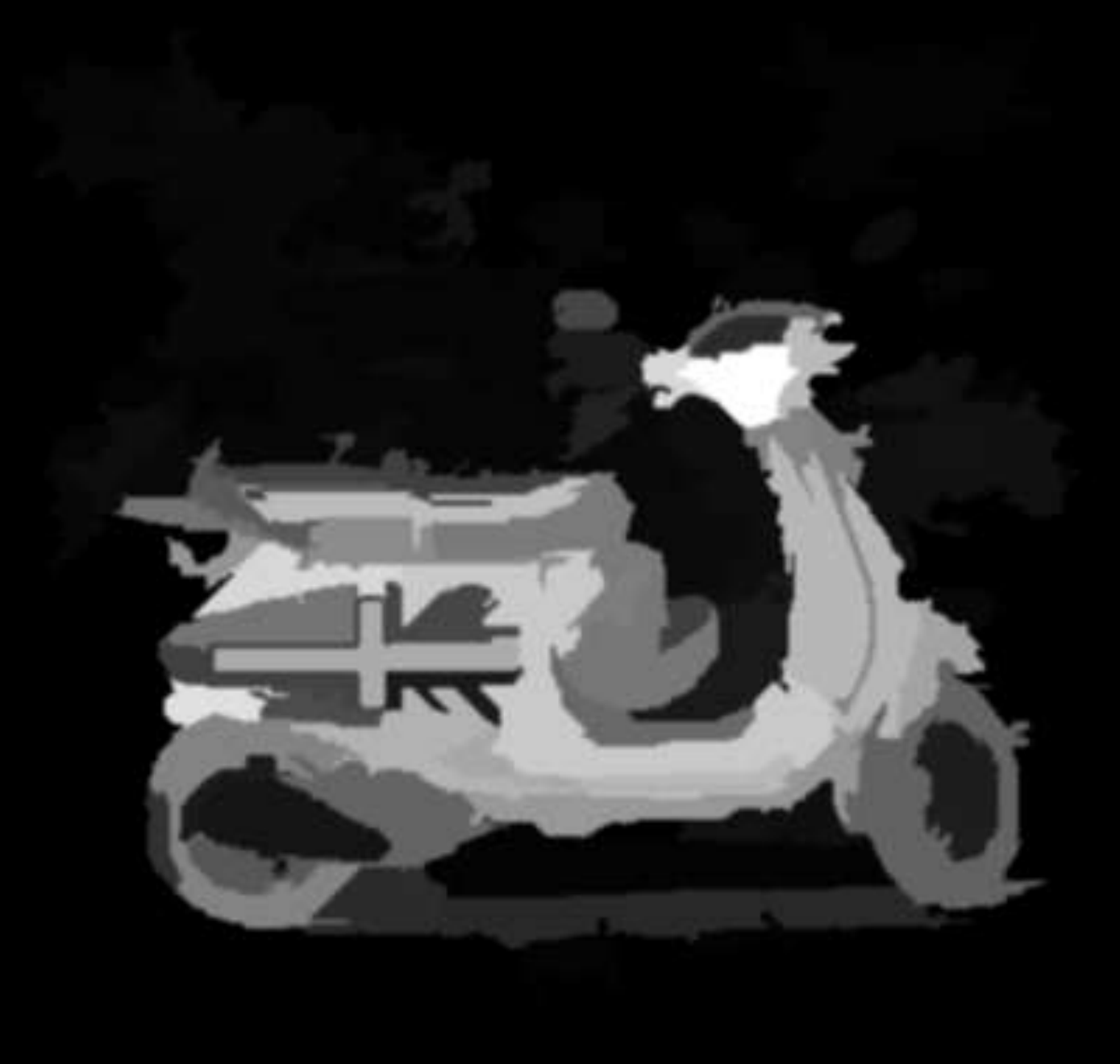}&
	\addExample{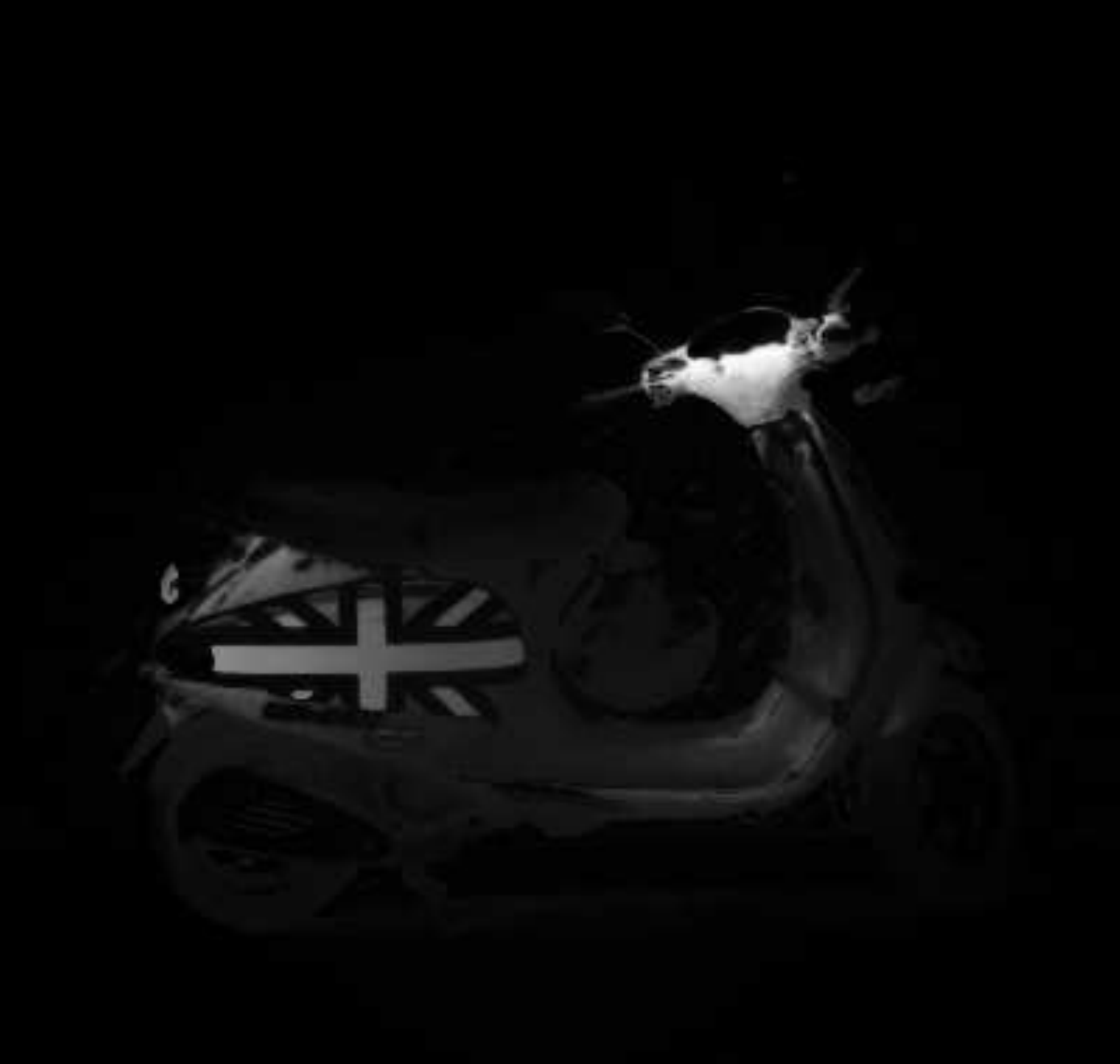}&
	\addExample{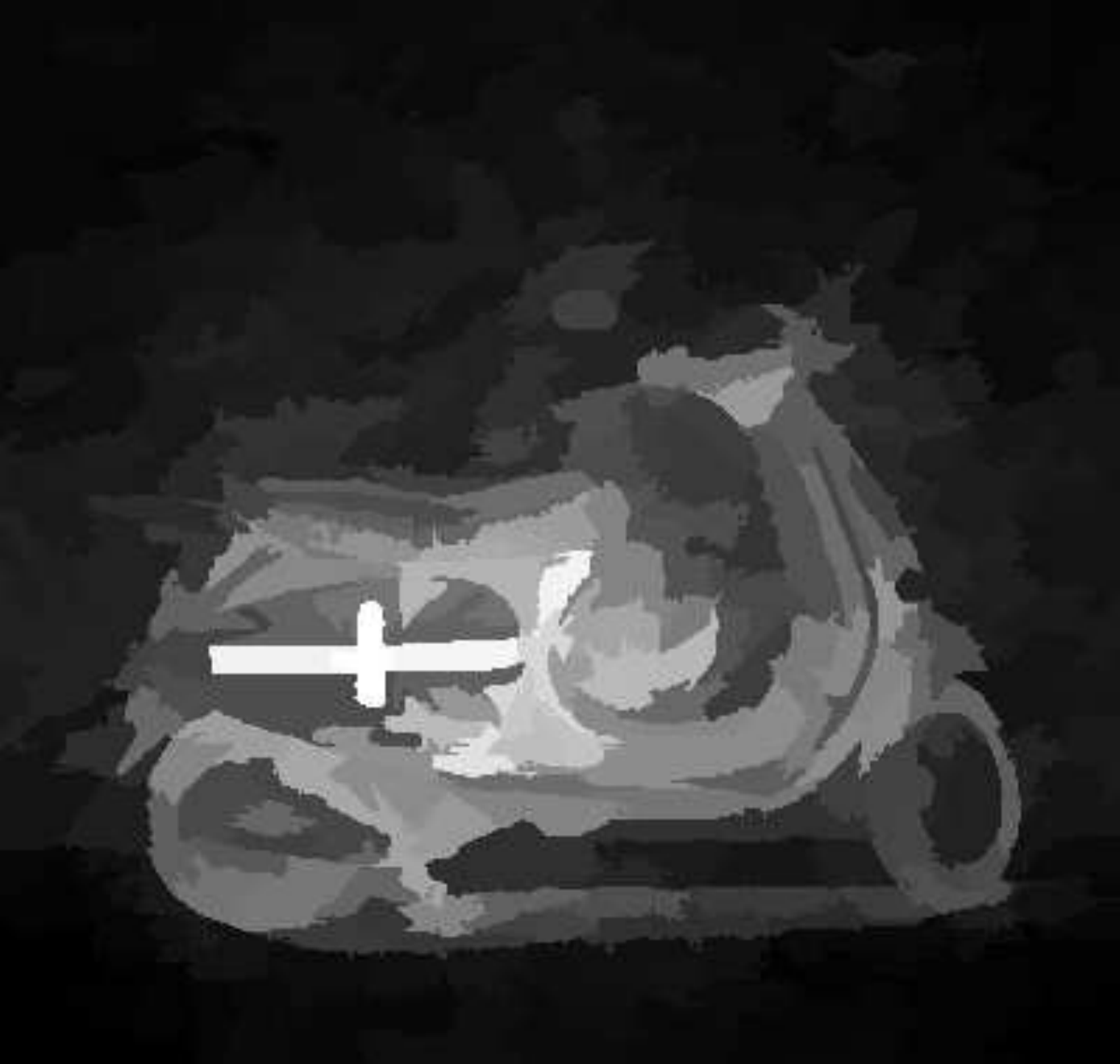}&
	\addExample{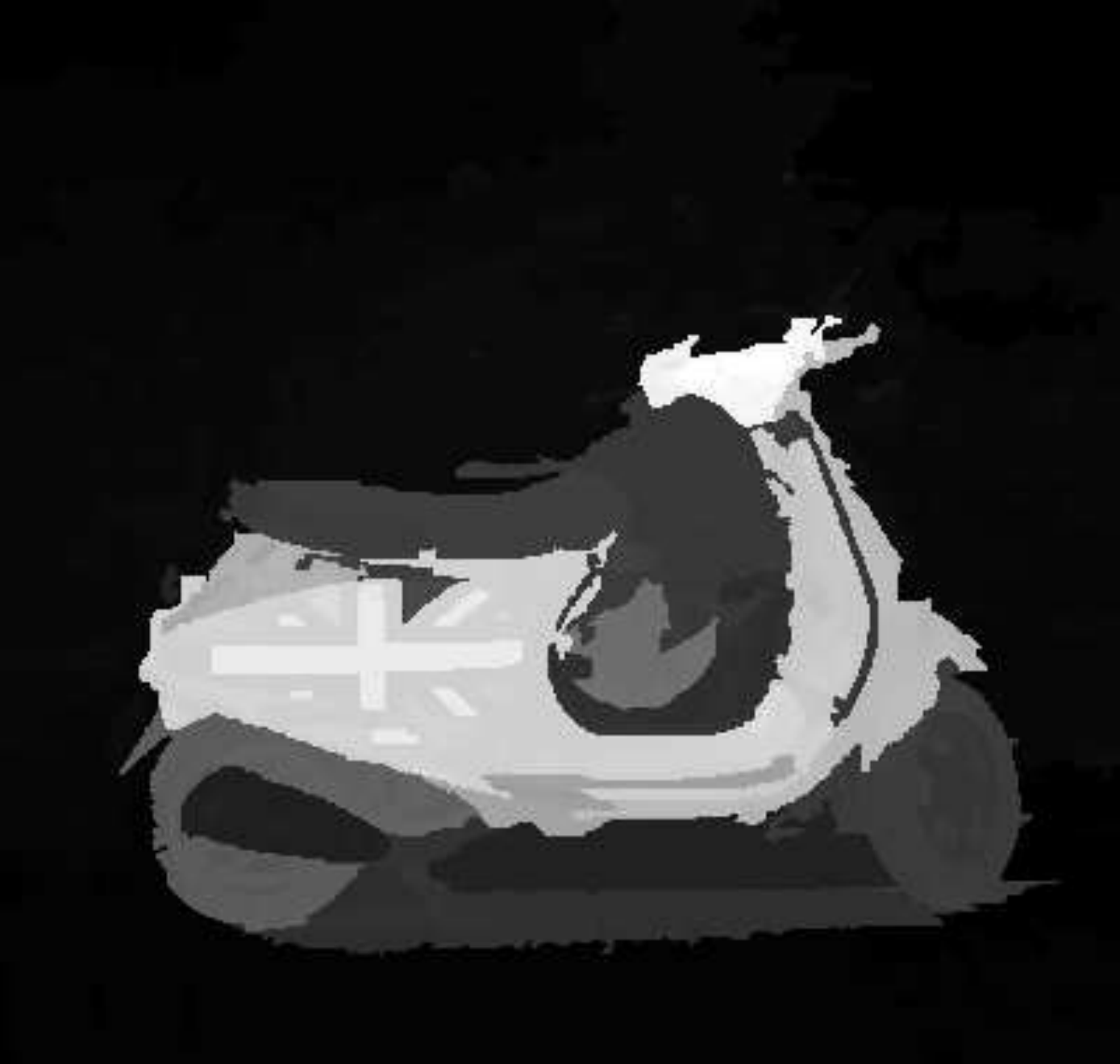}&
	\addExample{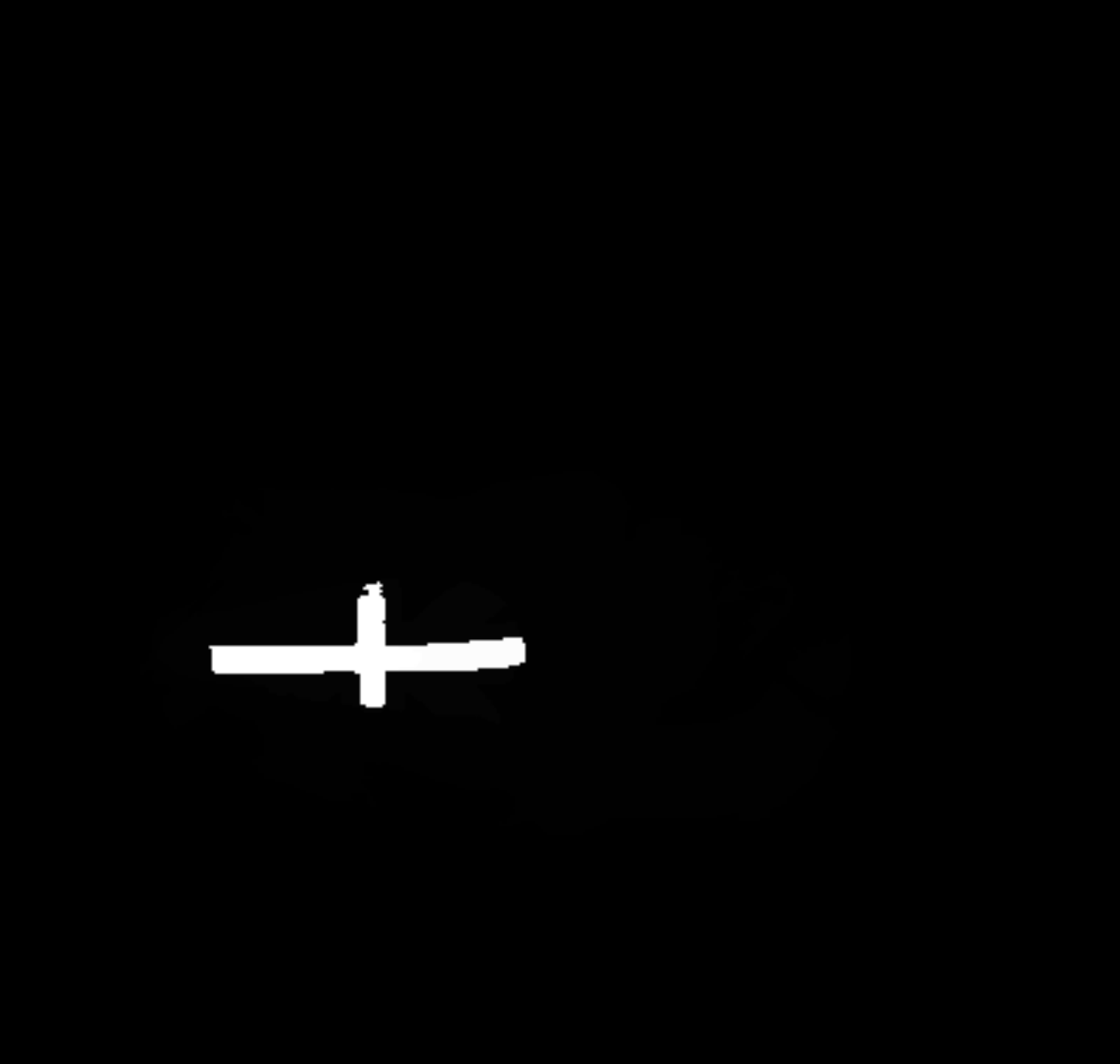}&
	\addExample{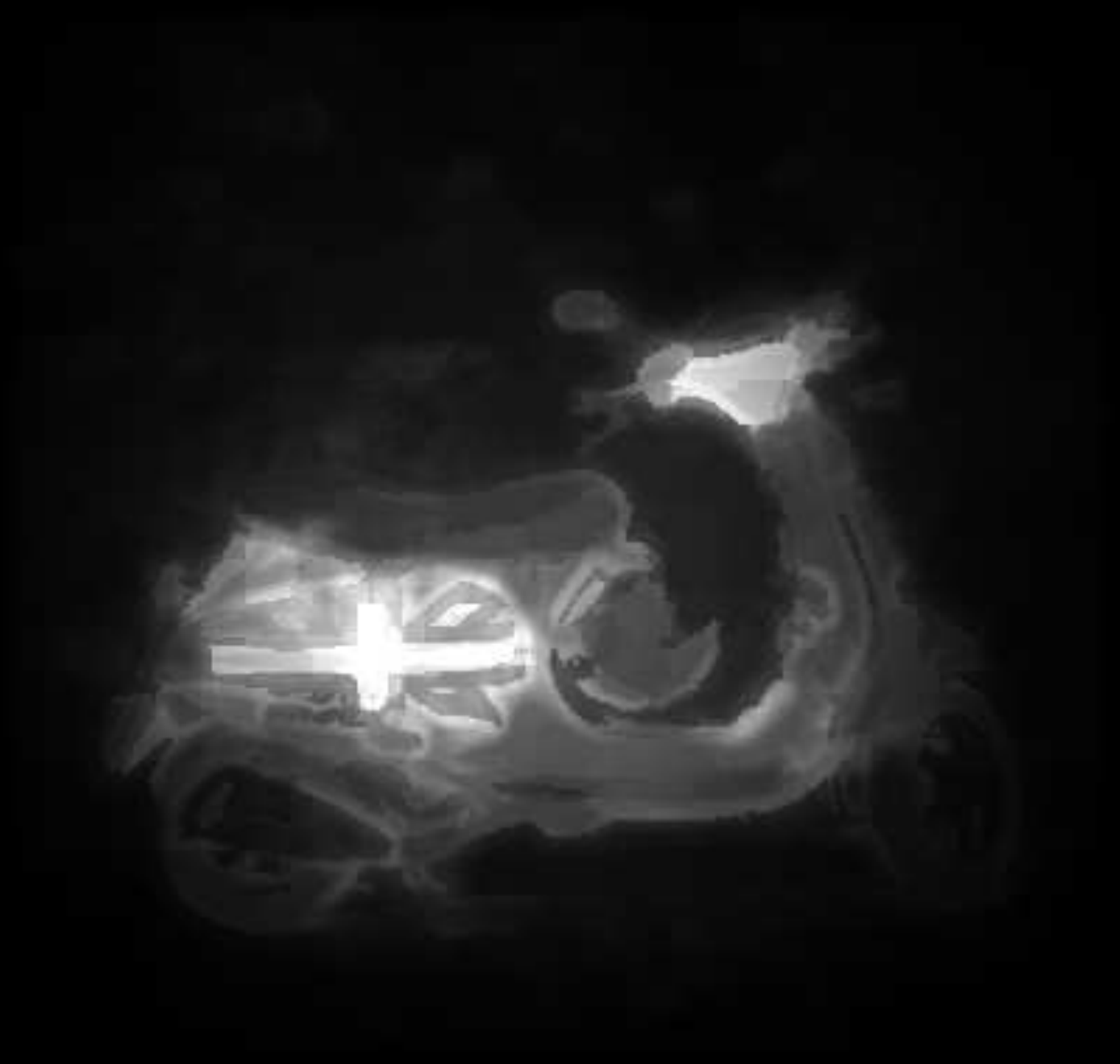}&
	\addExample{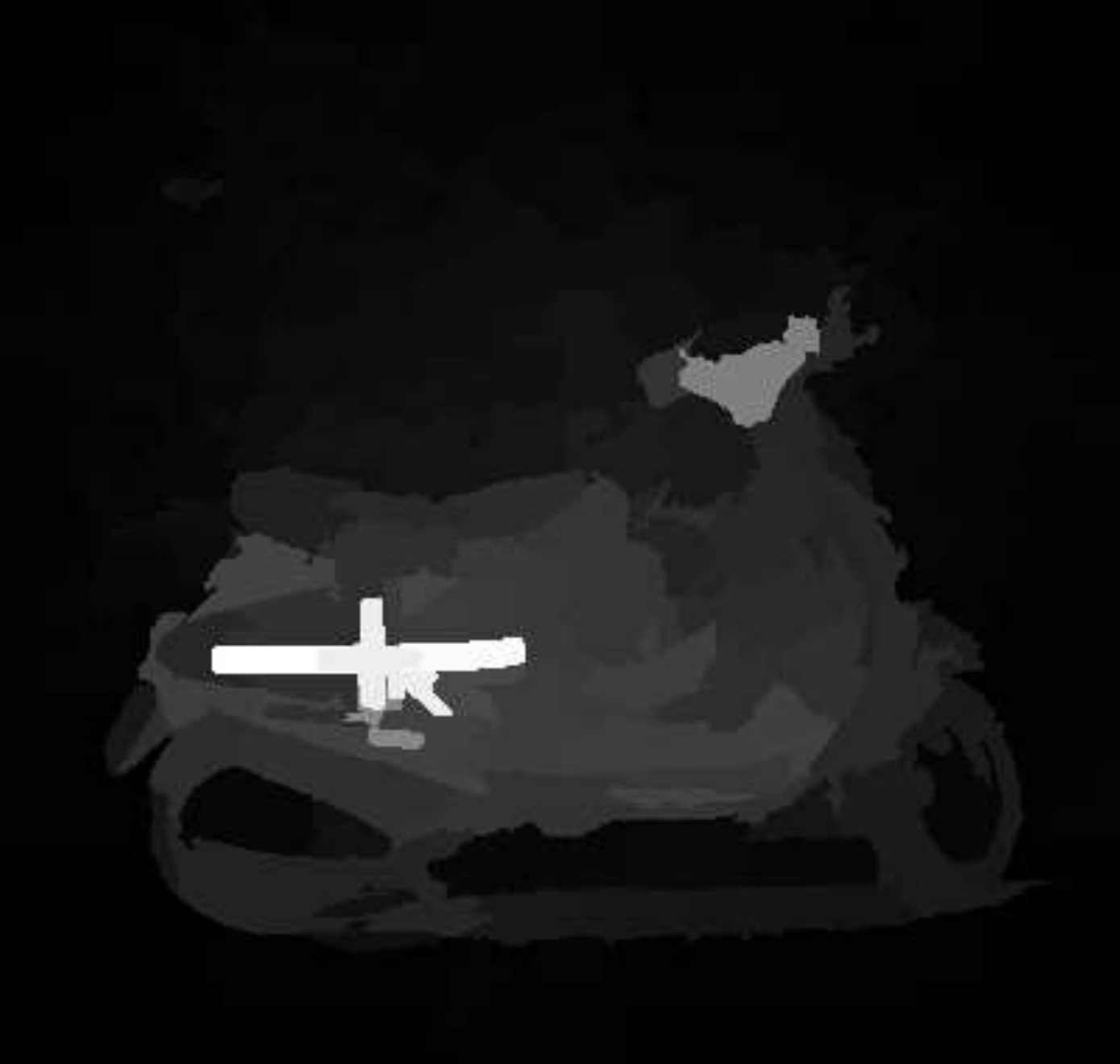}&
	\addExample{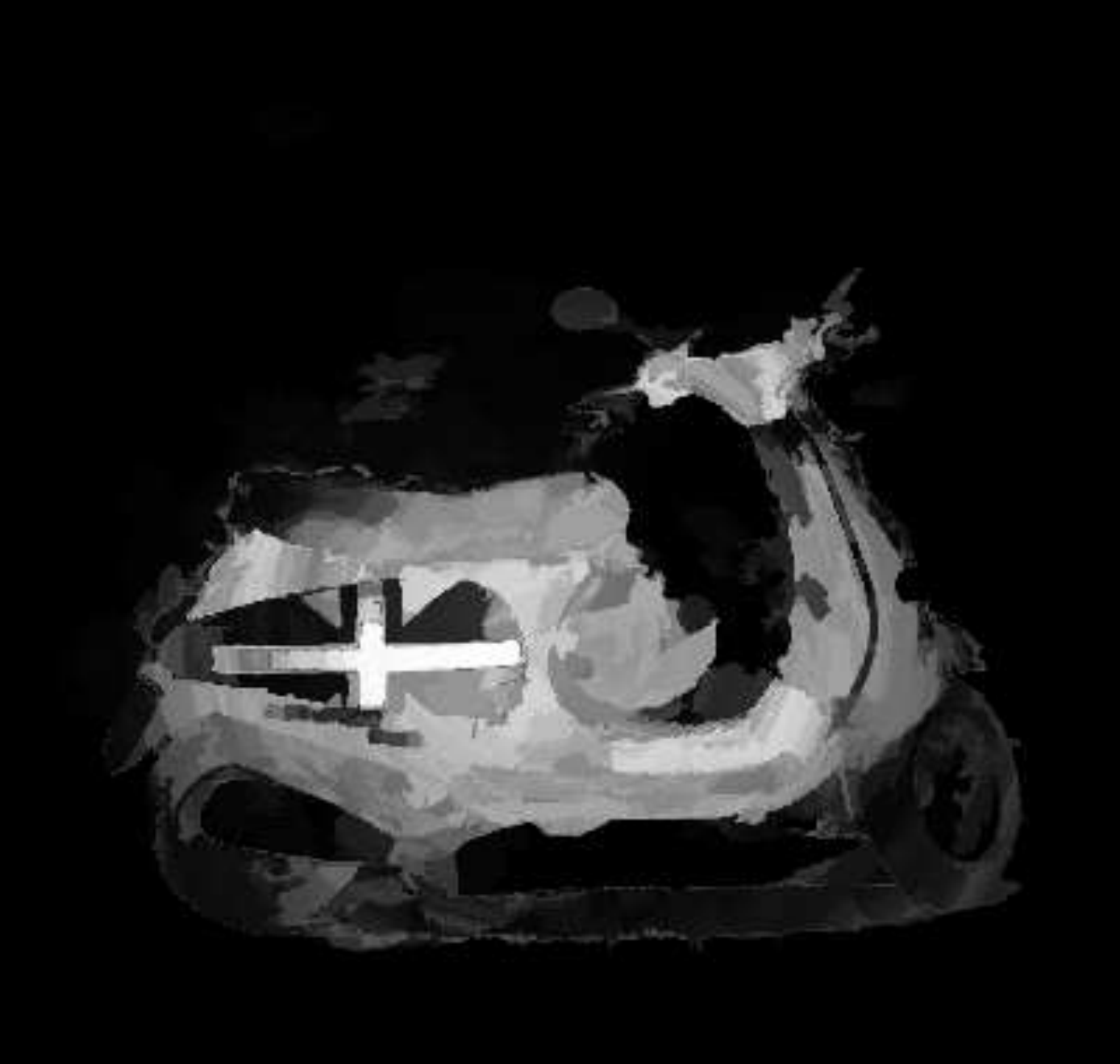}&
	\addExample{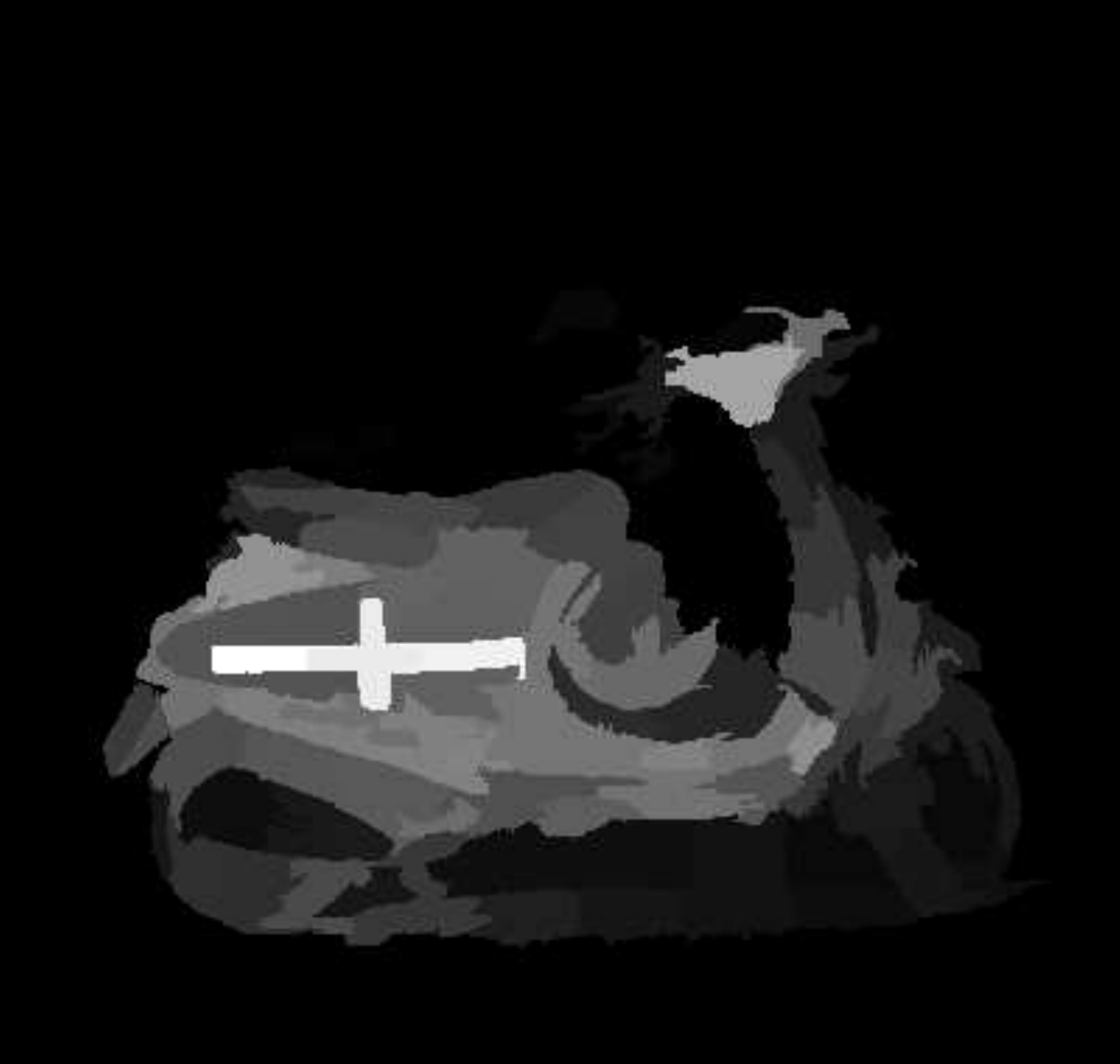}&
	\addExample{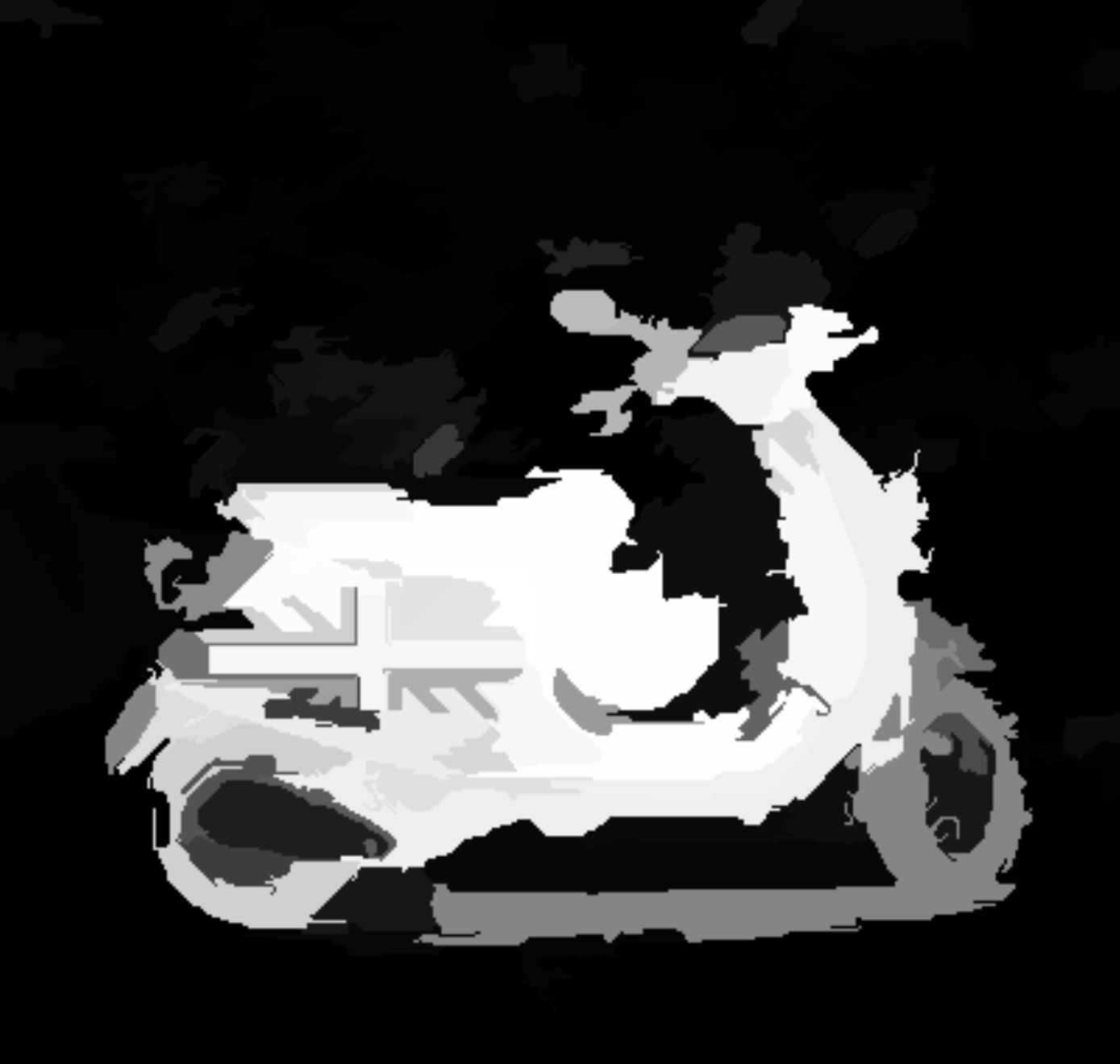}&
	\addExample{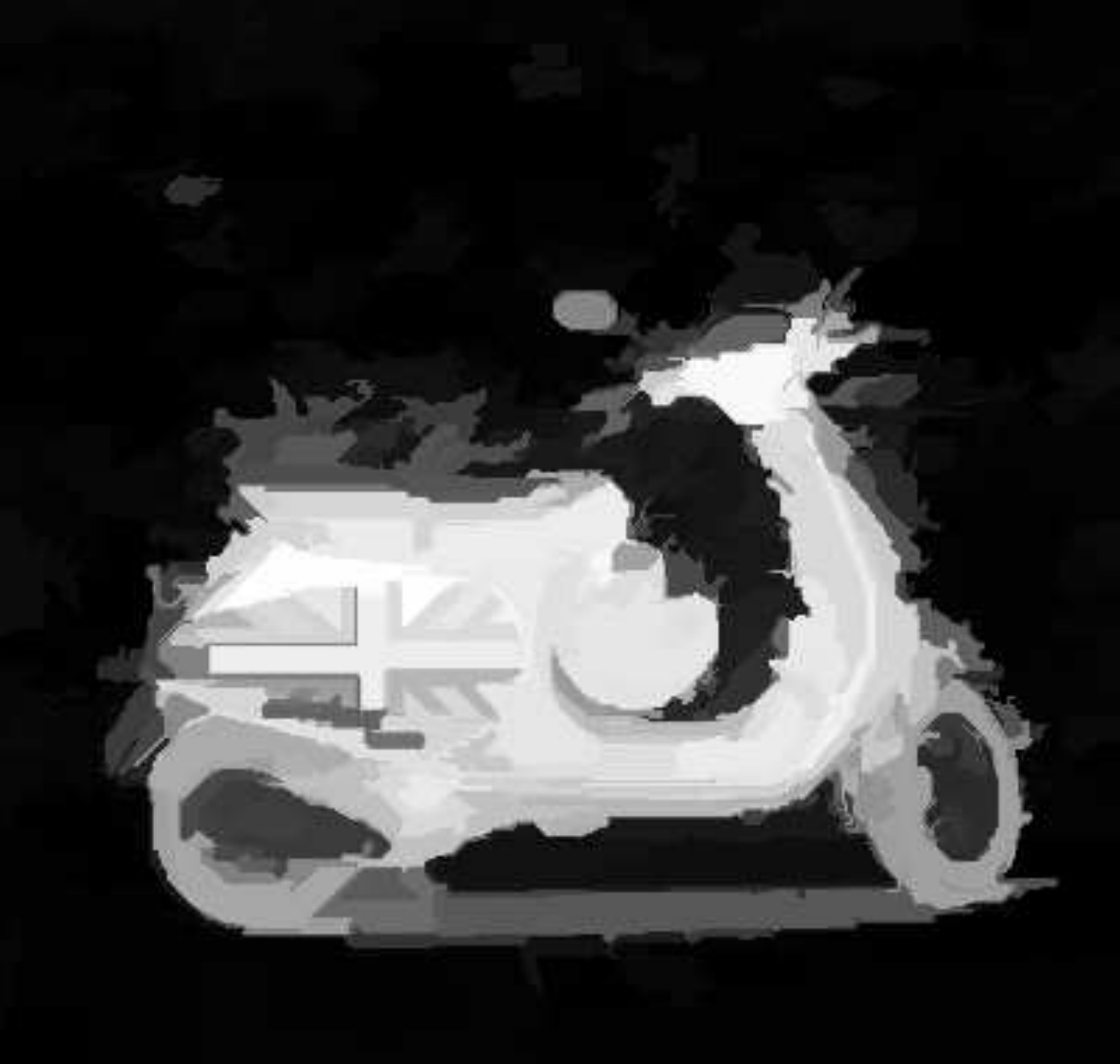}
    \\
	\addExample{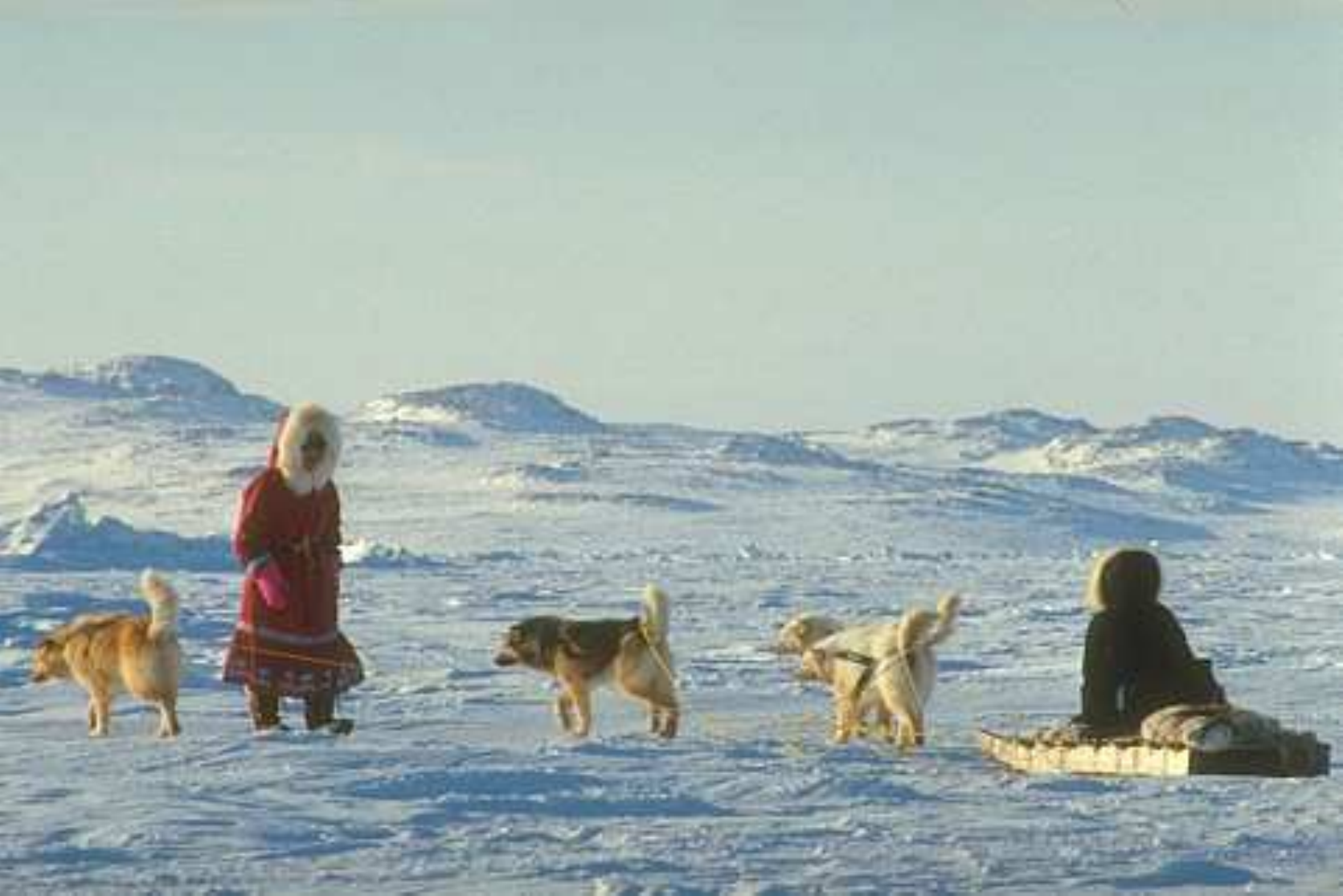}&
	\addExample{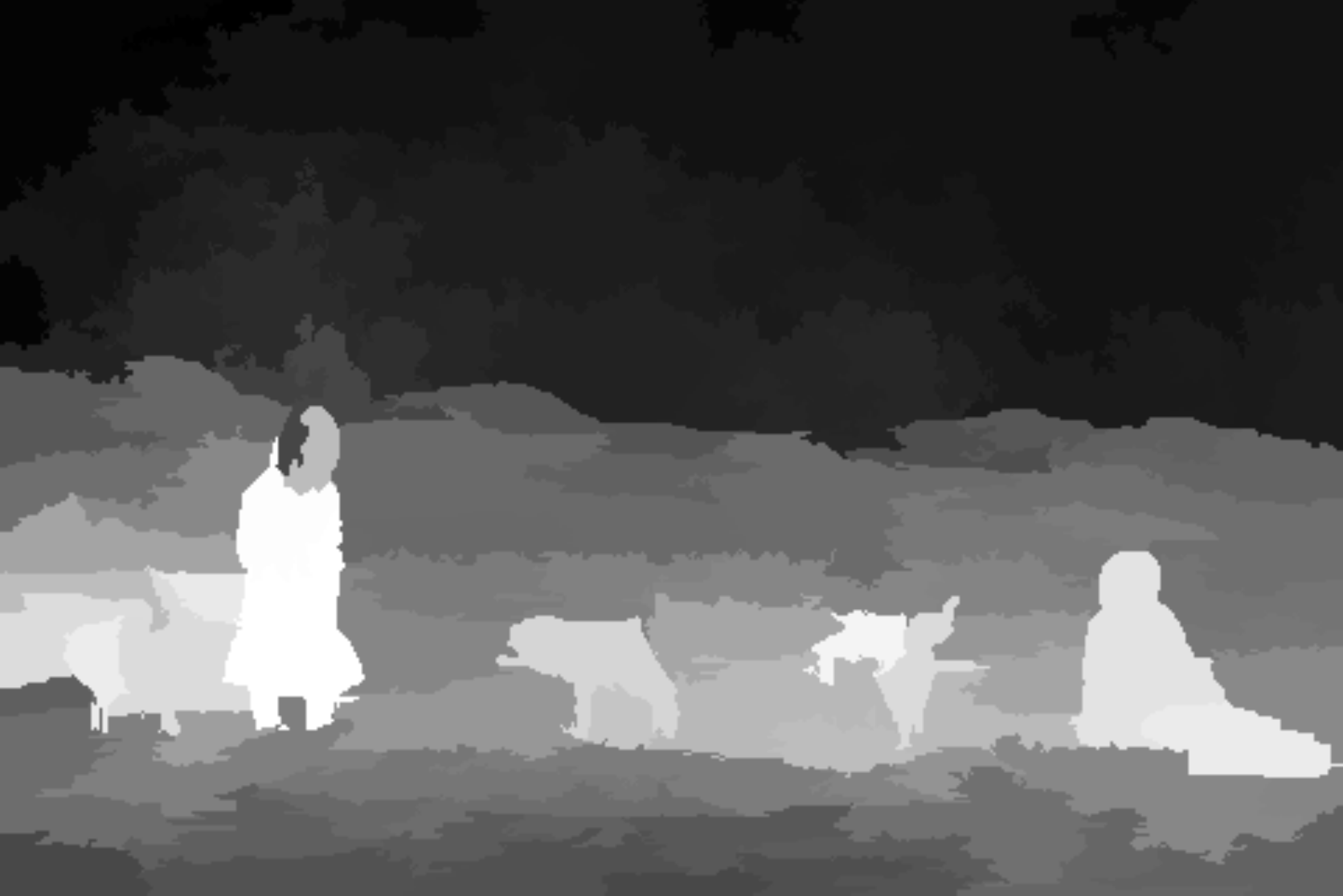}&
	\addExample{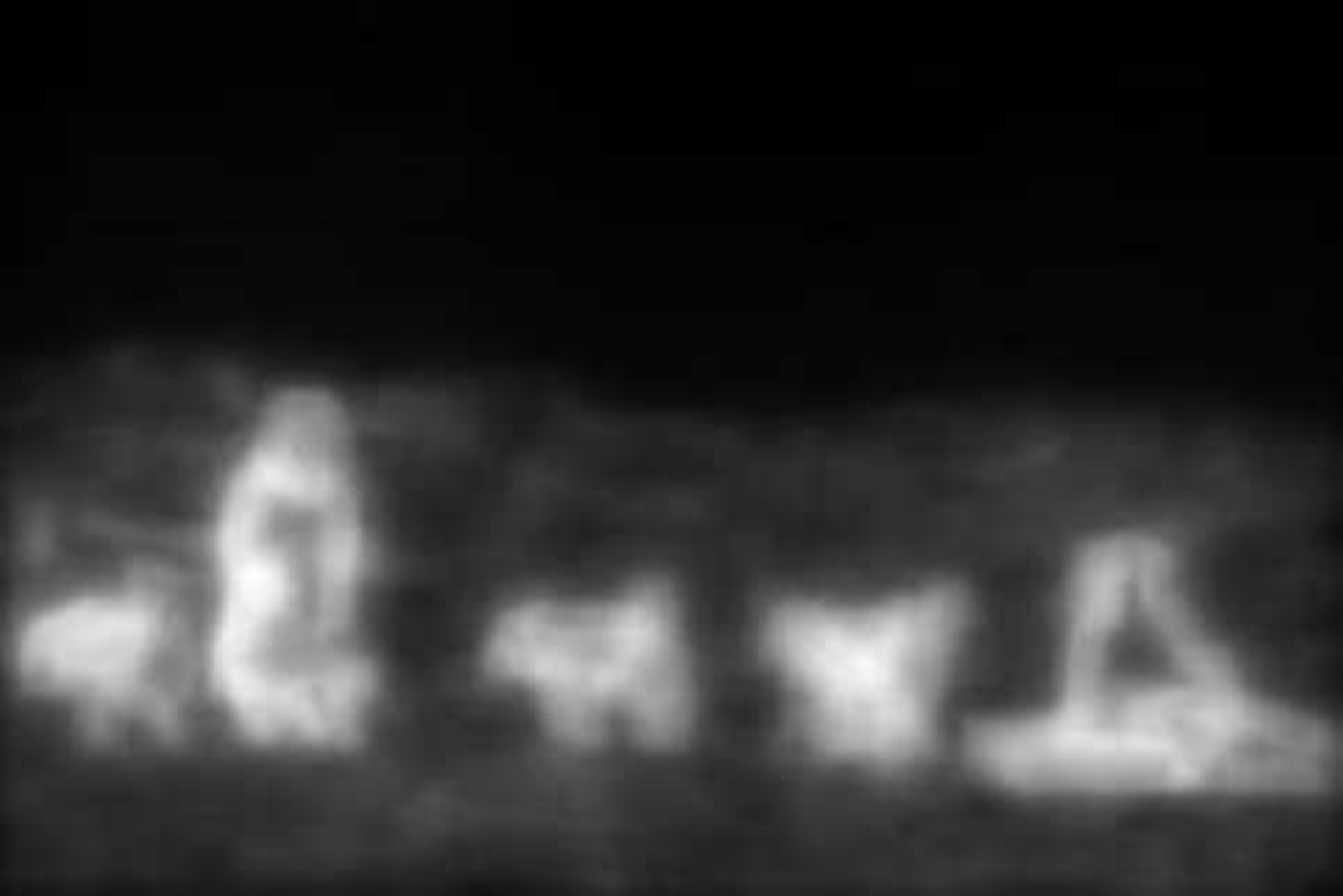}&
	\addExample{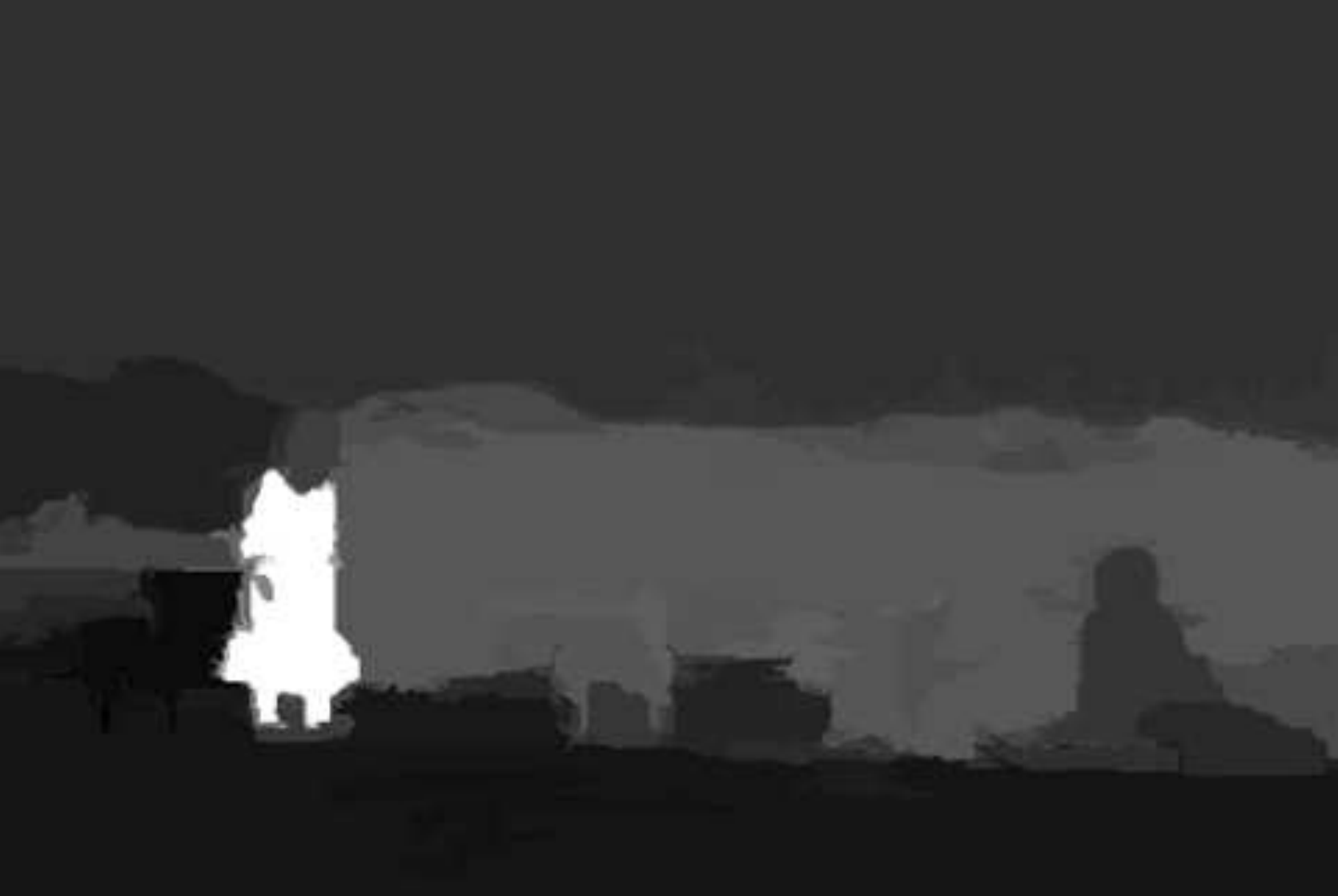}&
	\addExample{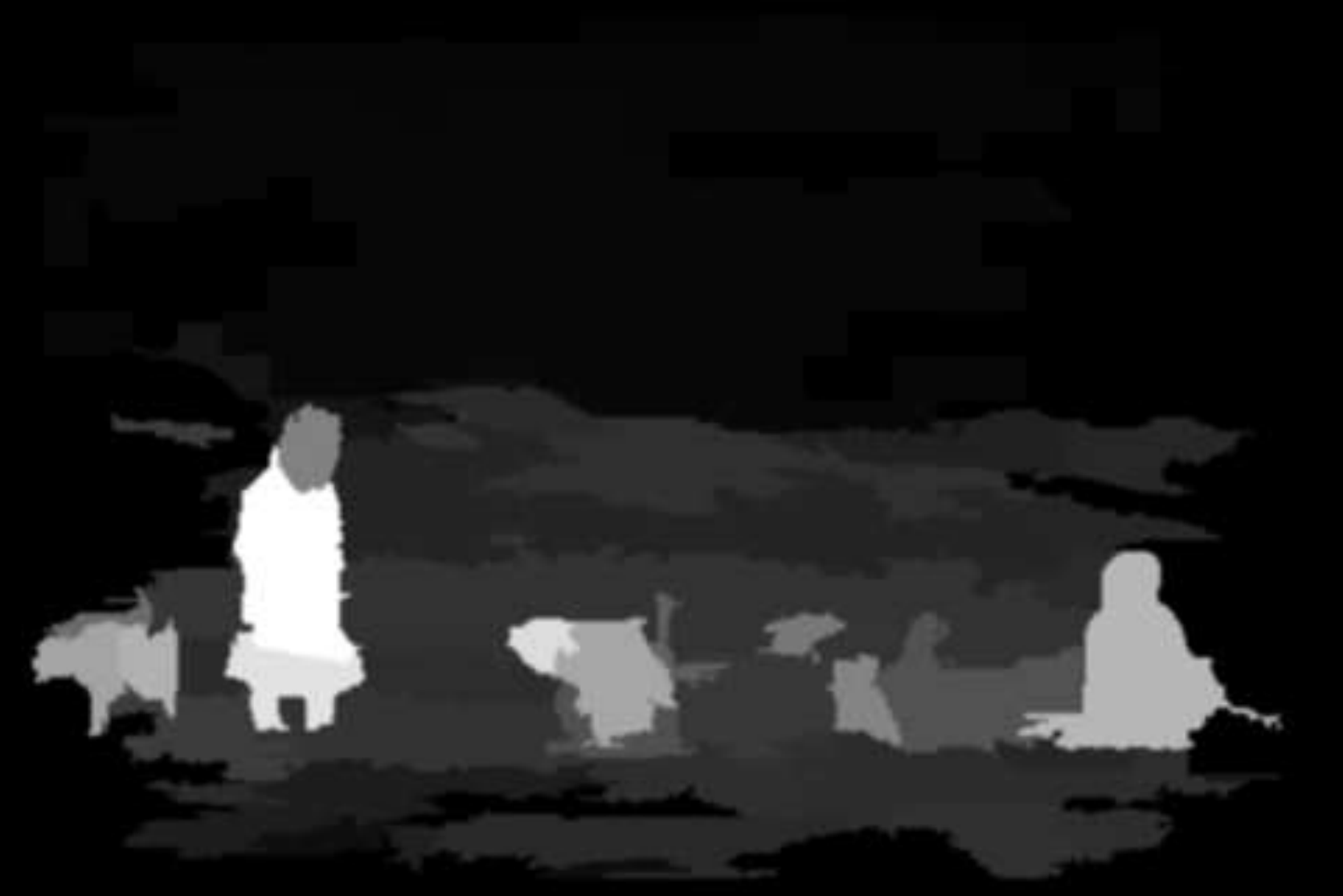}&
	\addExample{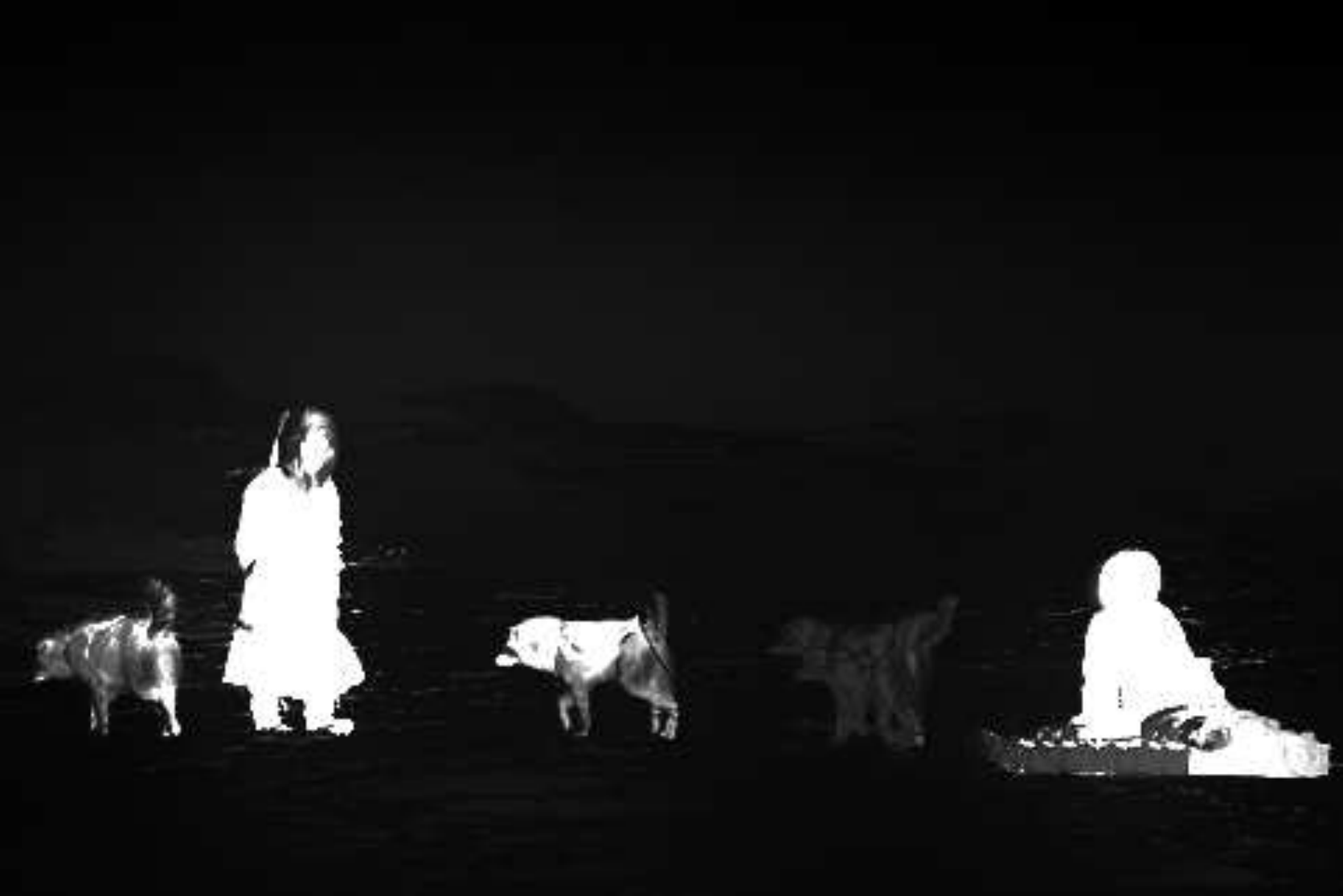}&
	\addExample{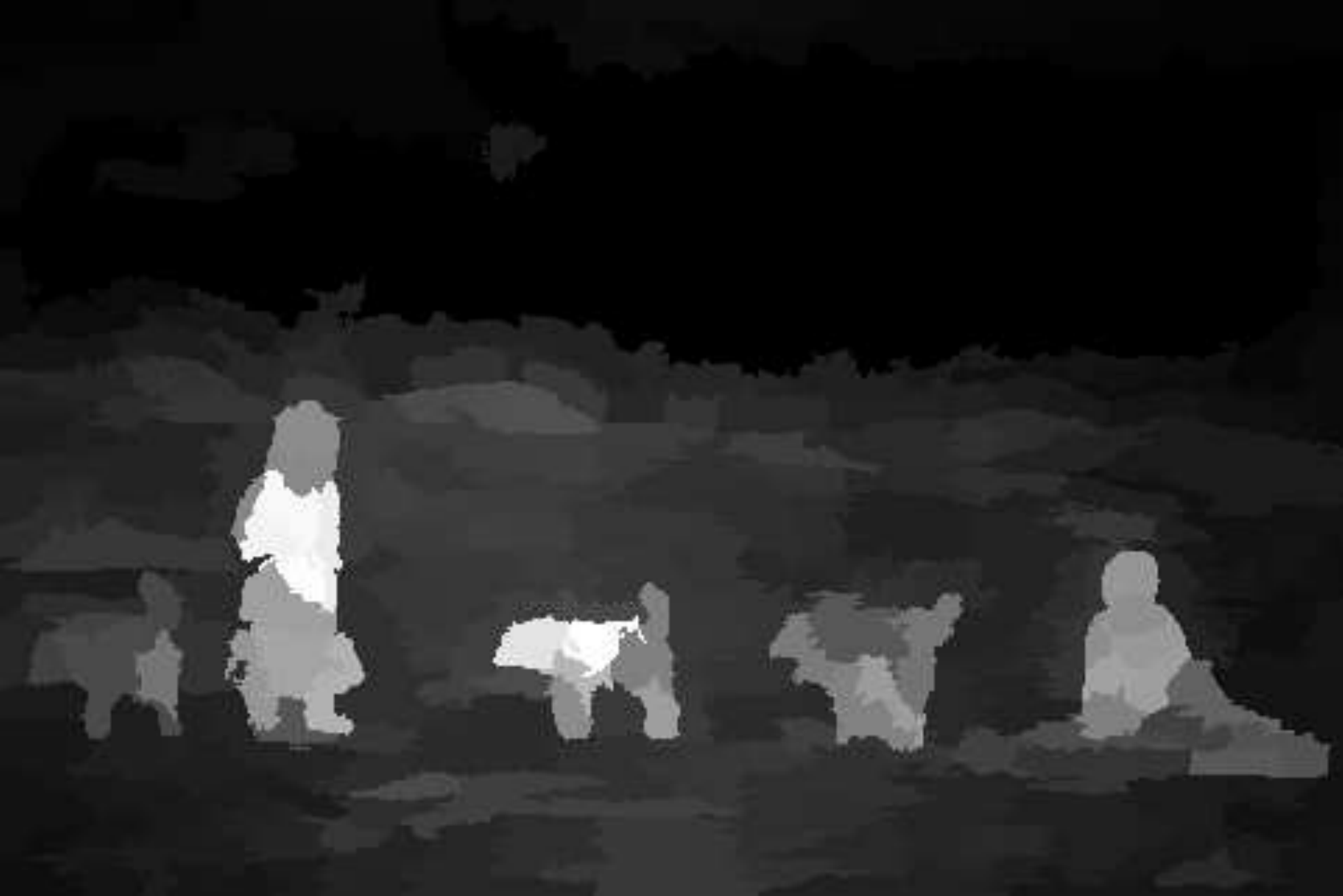}&
	\addExample{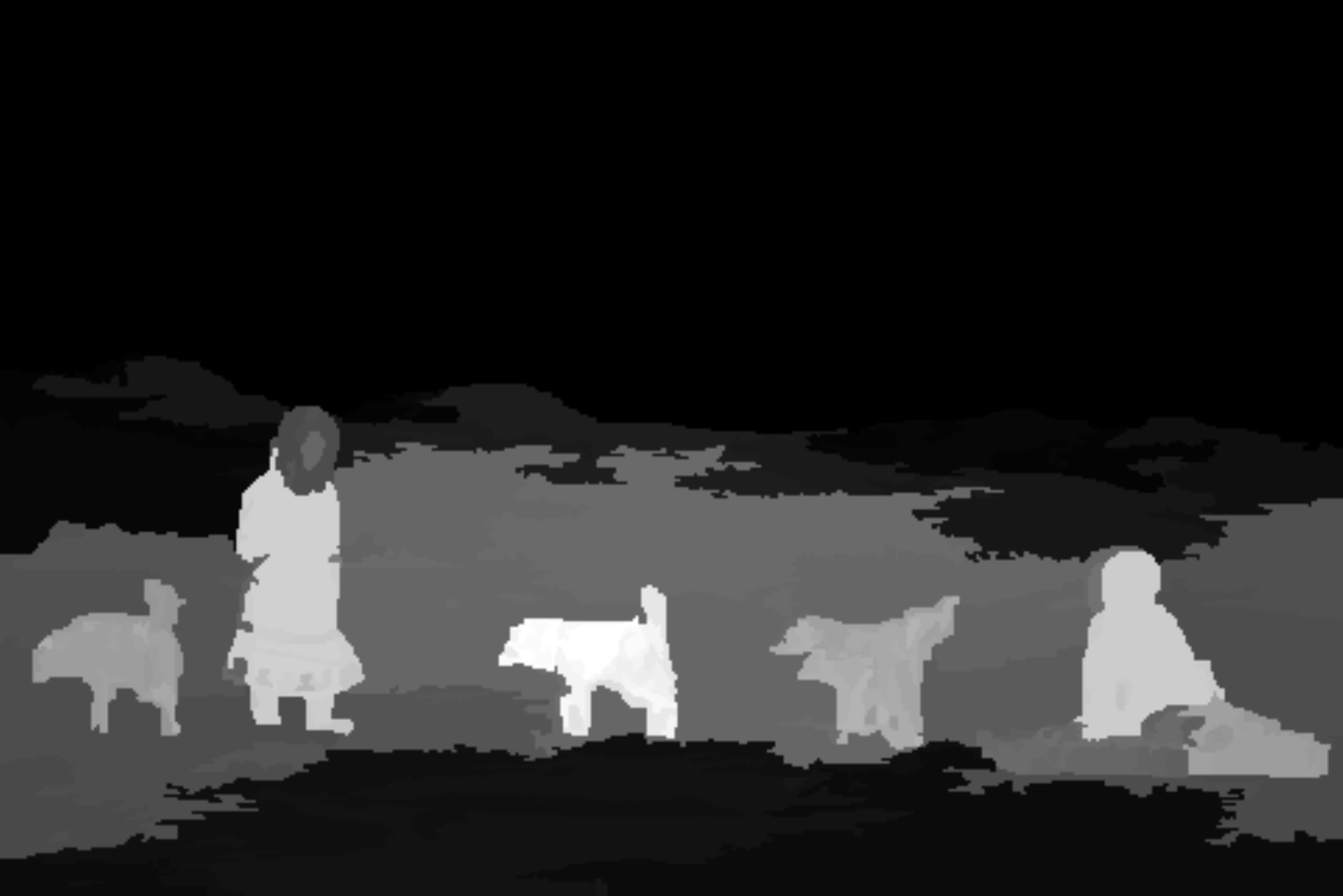}&
	\addExample{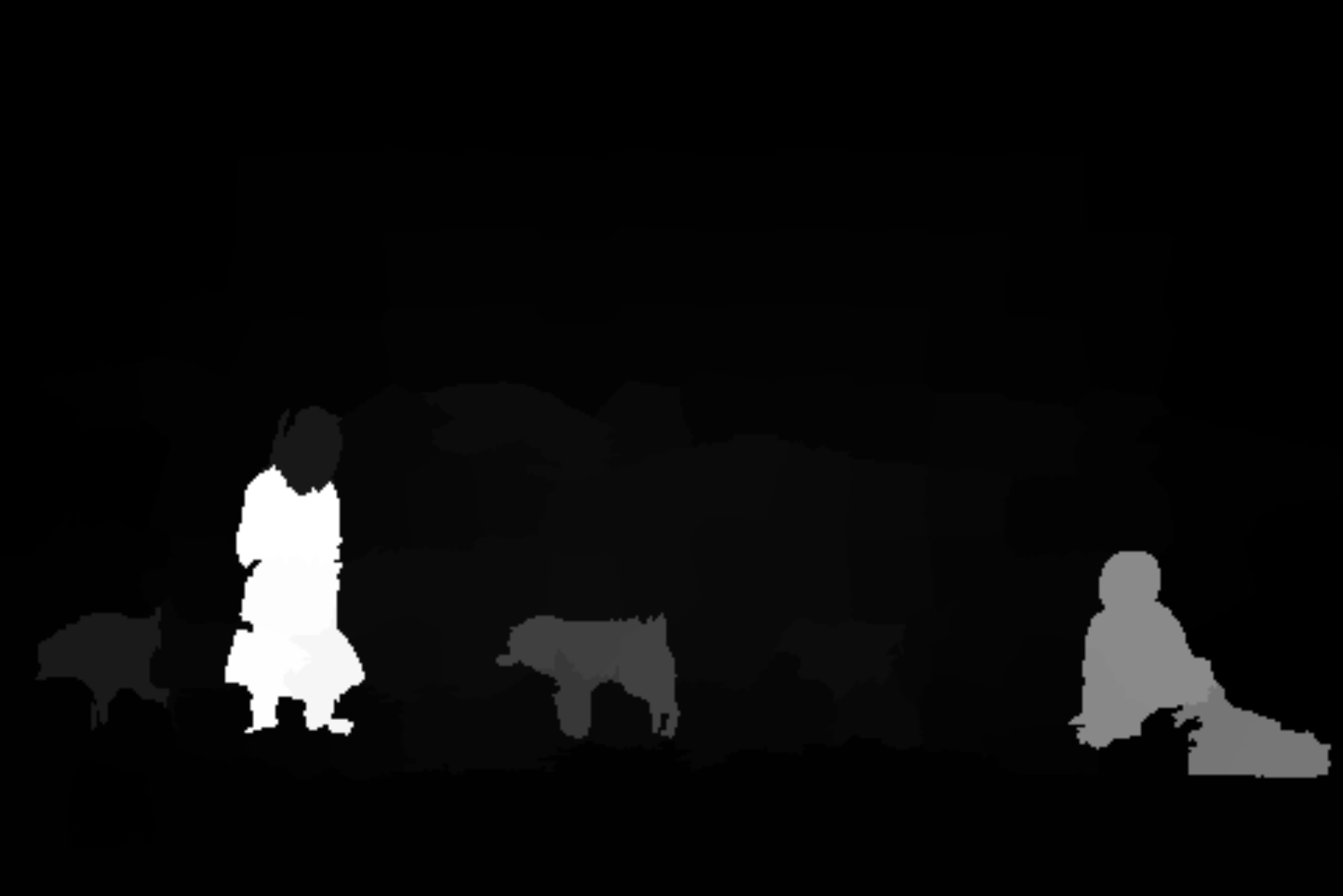}&
	\addExample{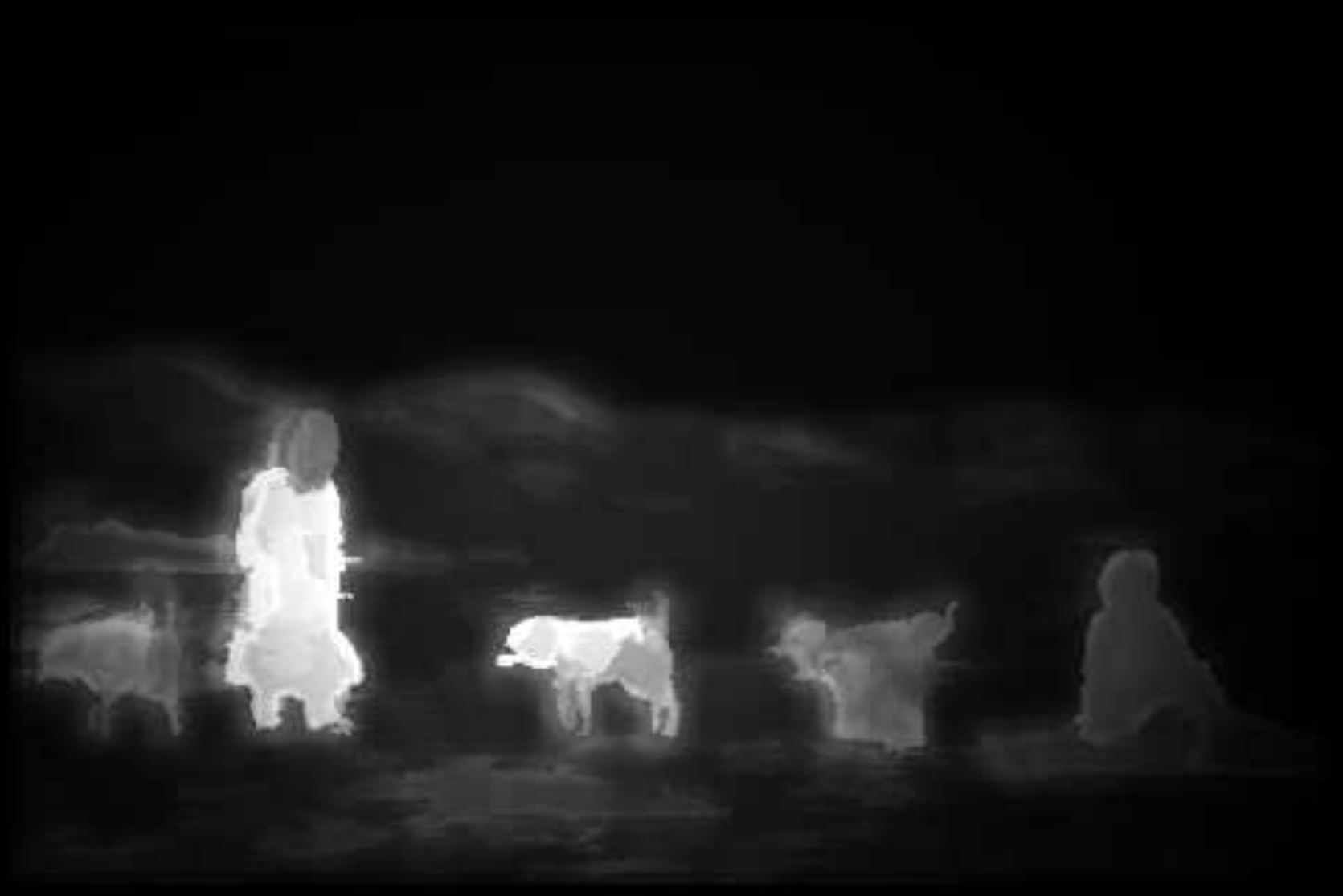}&
	\addExample{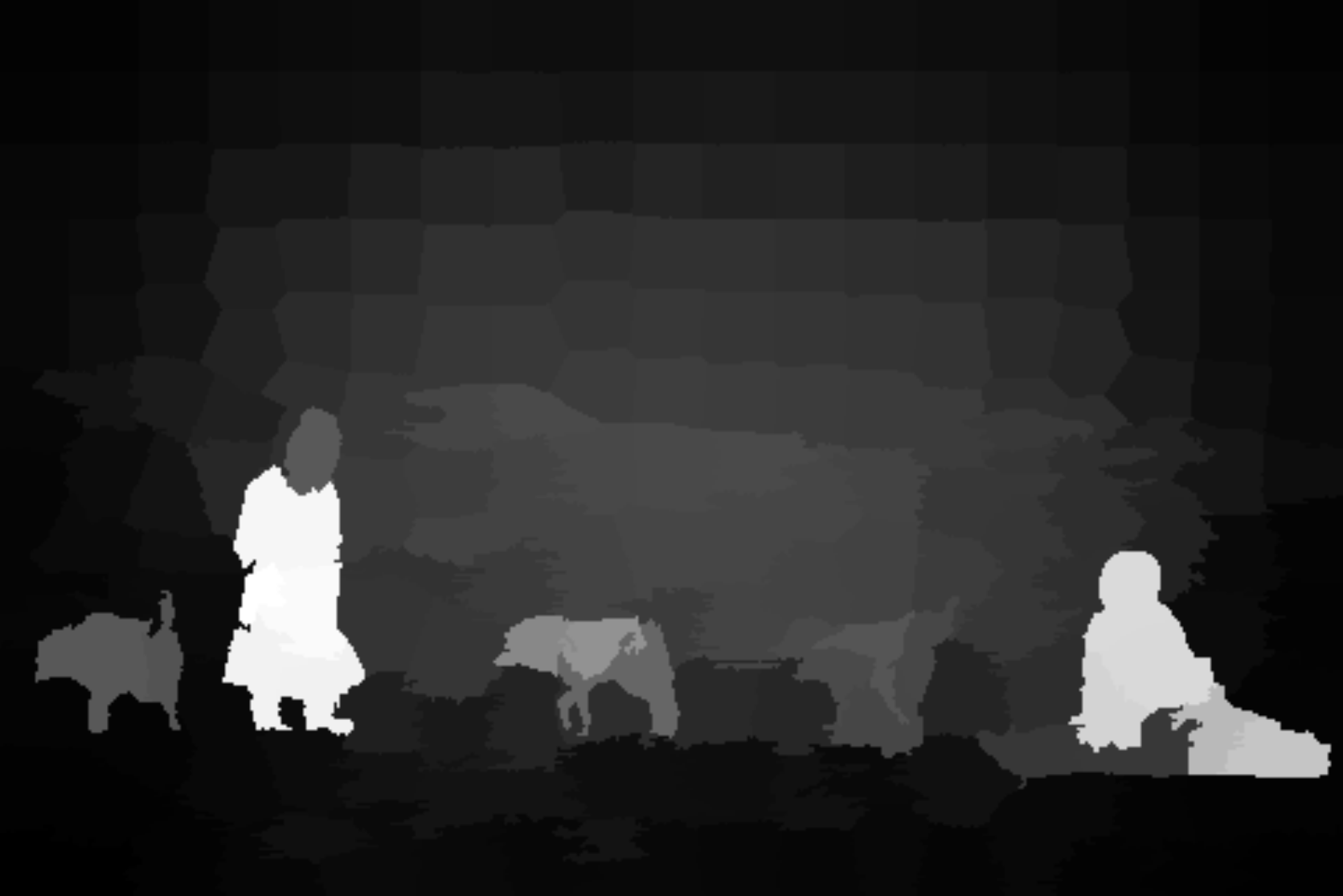}&
	\addExample{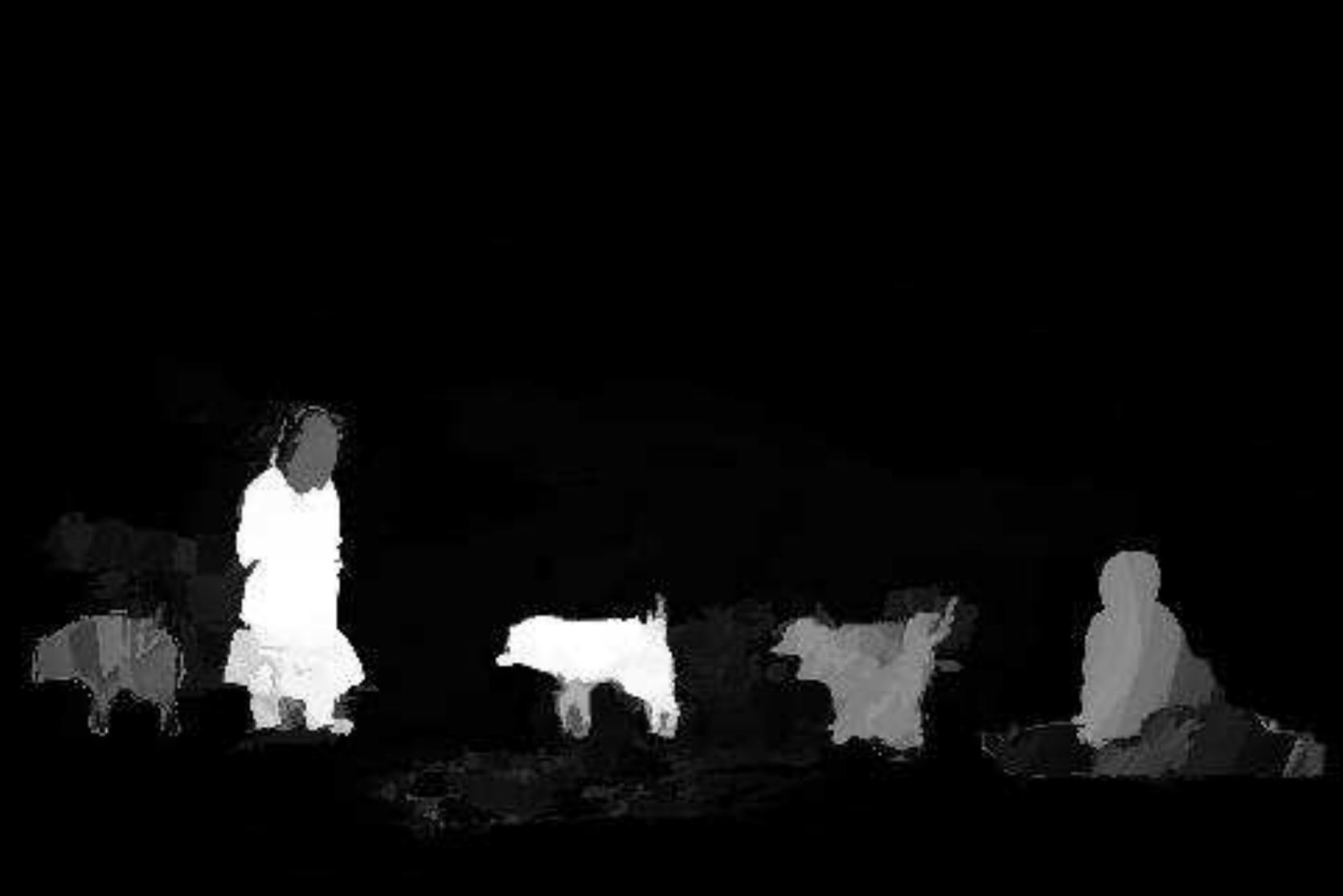}&
	\addExample{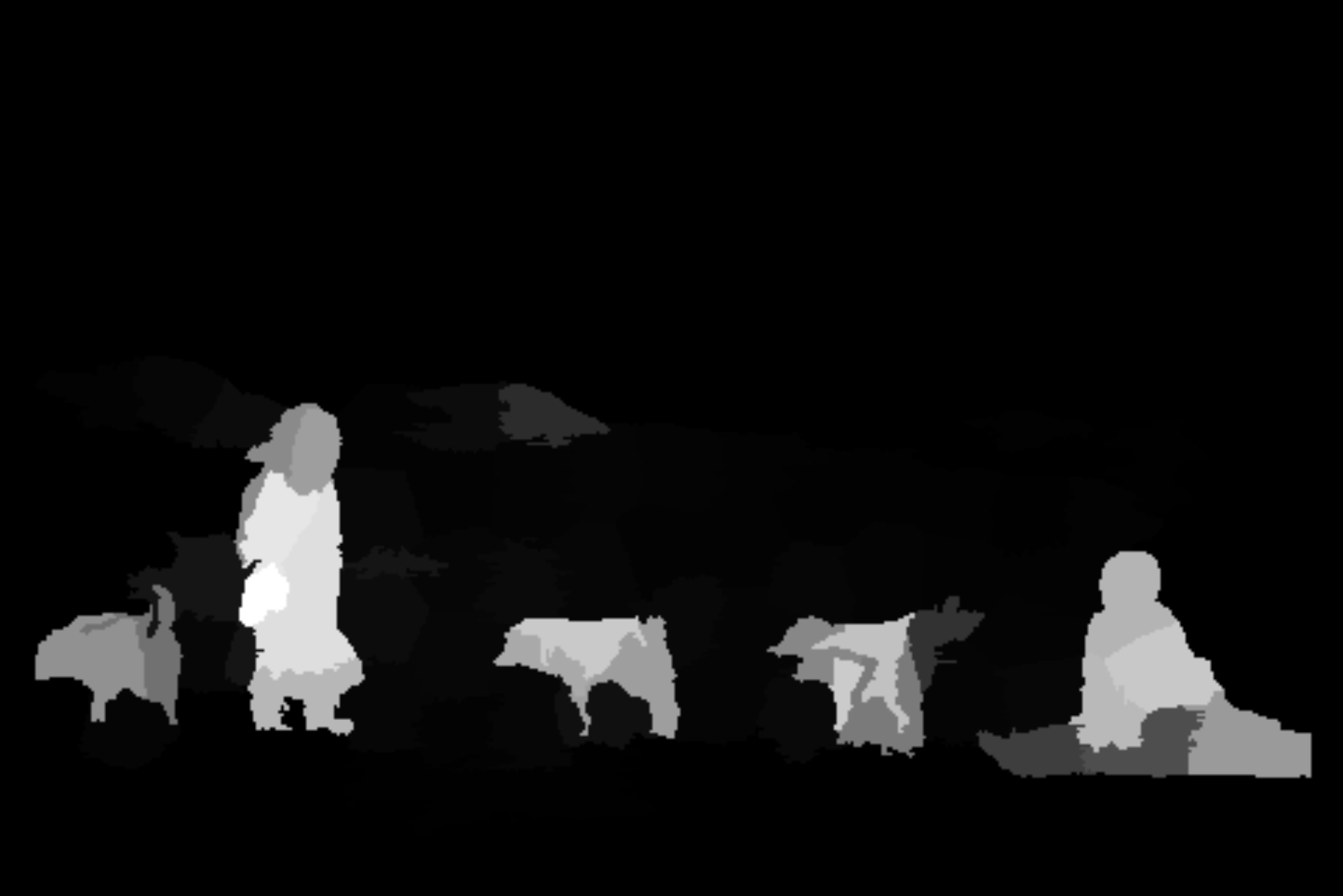}&
	\addExample{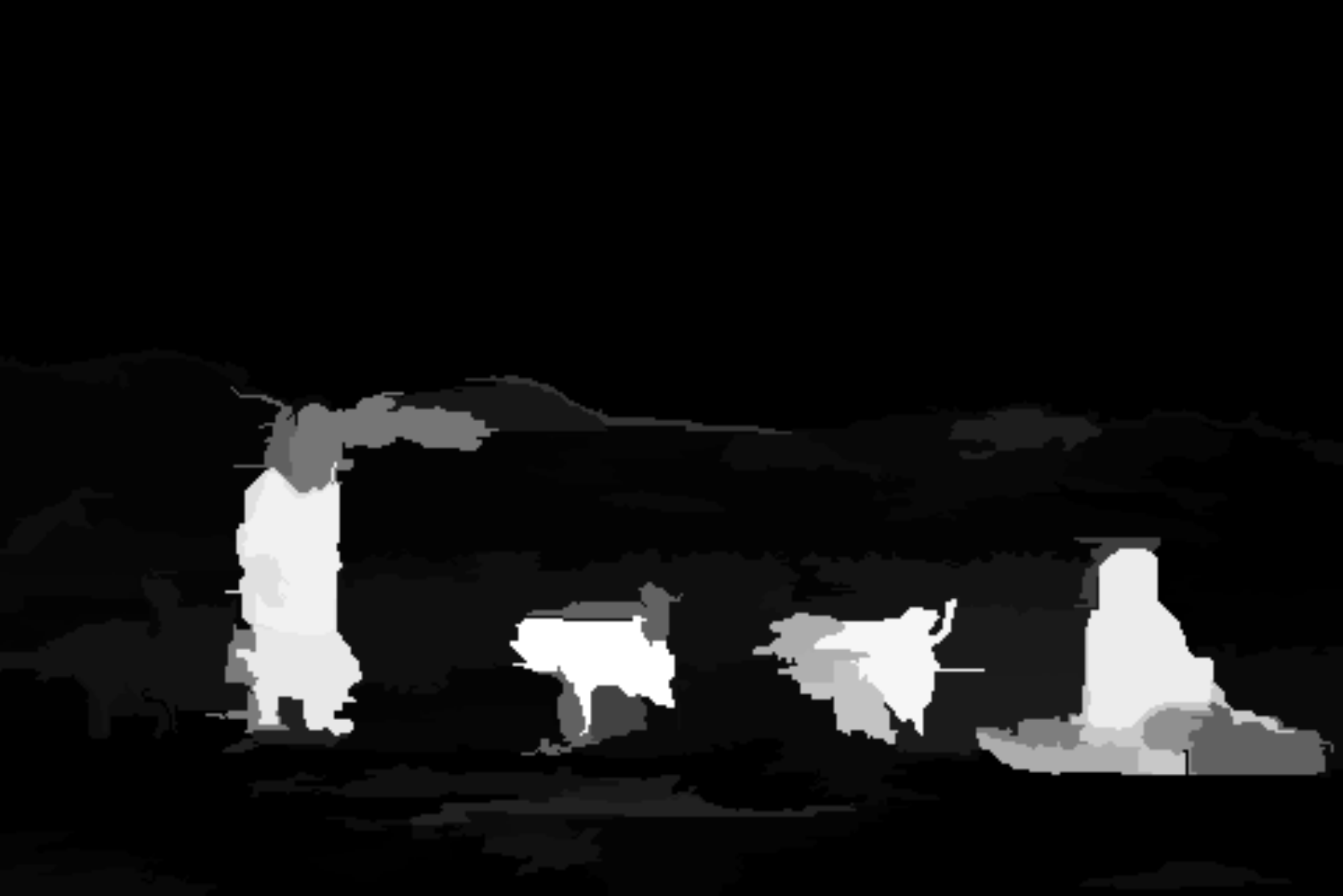}&
	\addExample{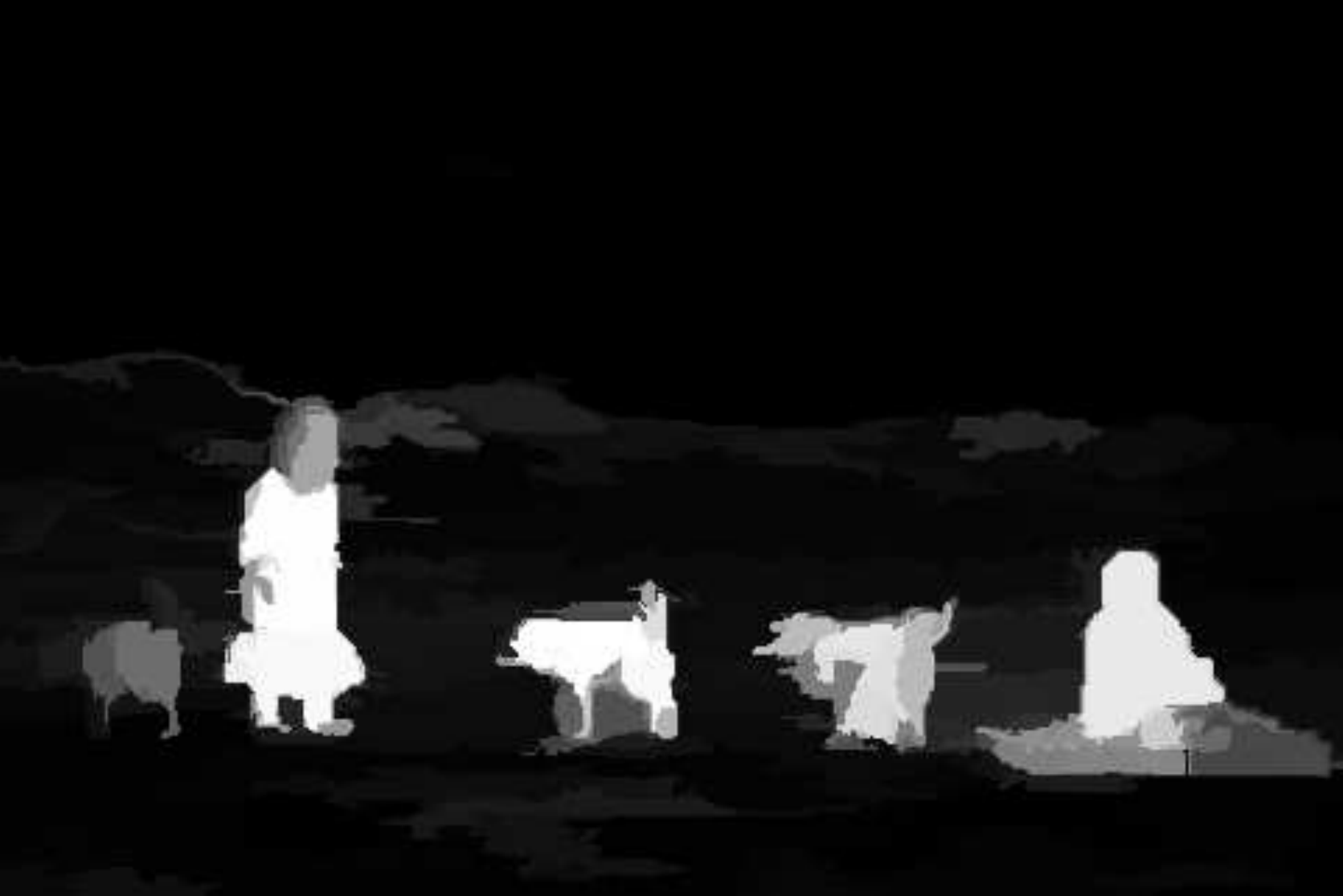}
    \\
	\addExample{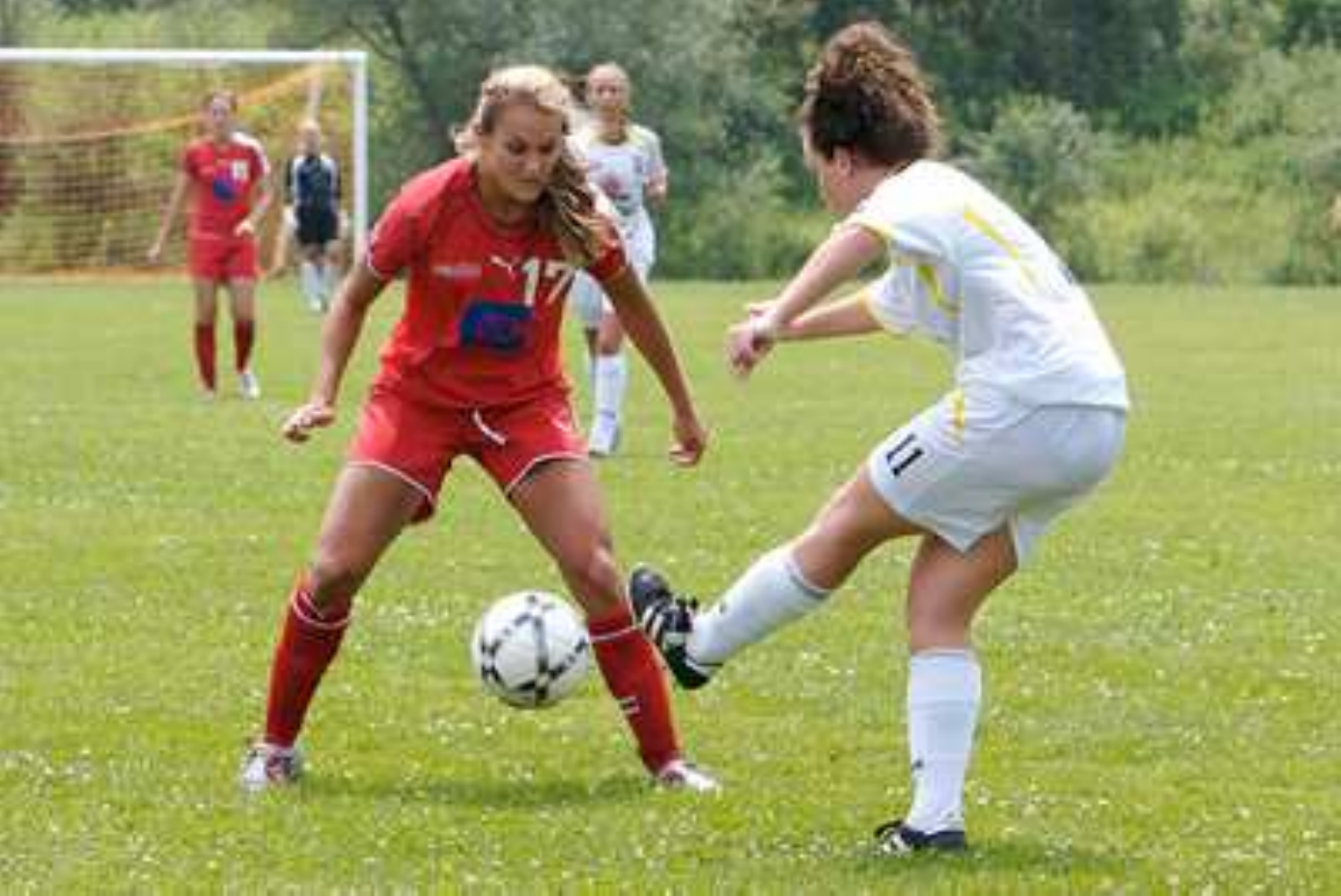}&
	\addExample{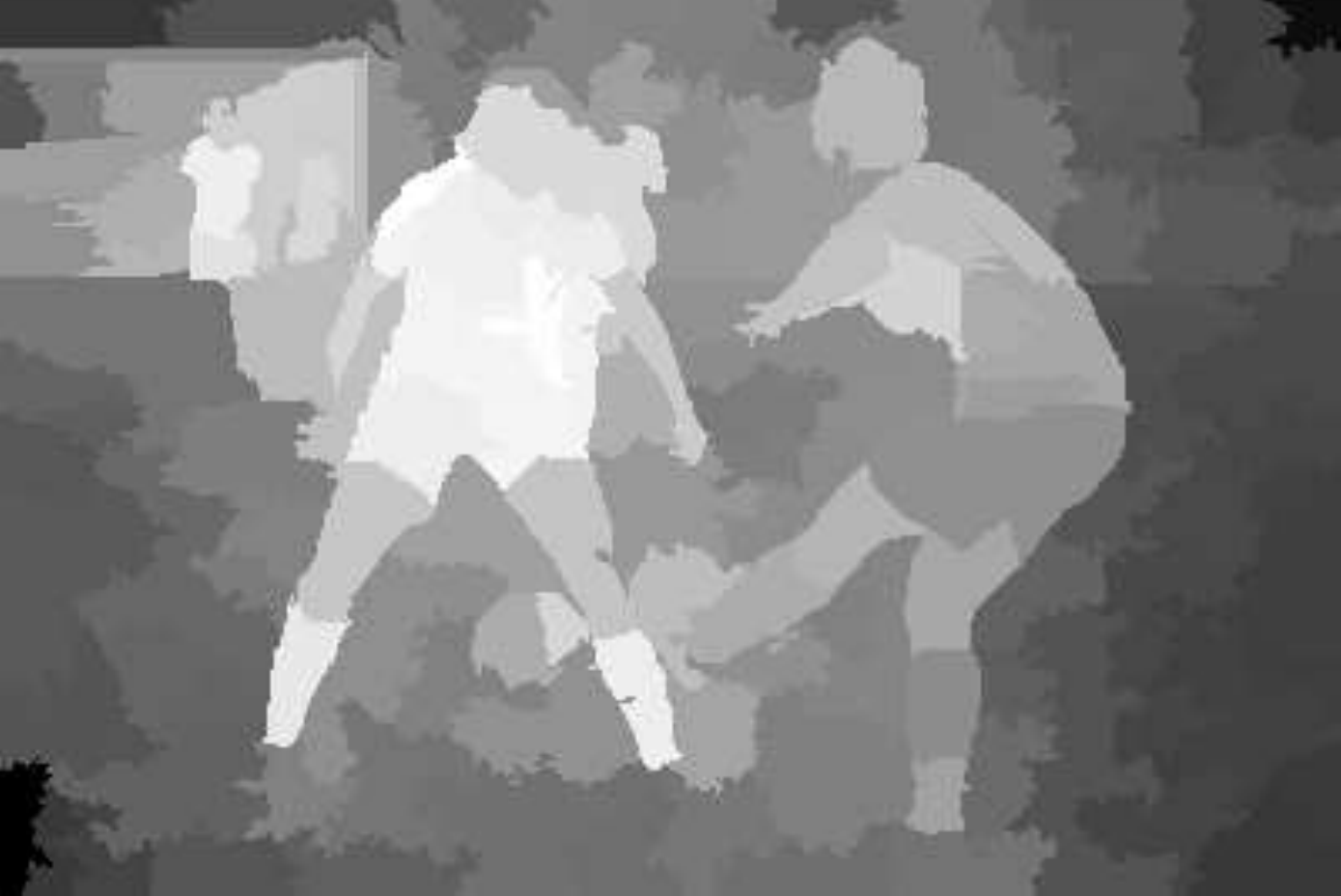}&
	\addExample{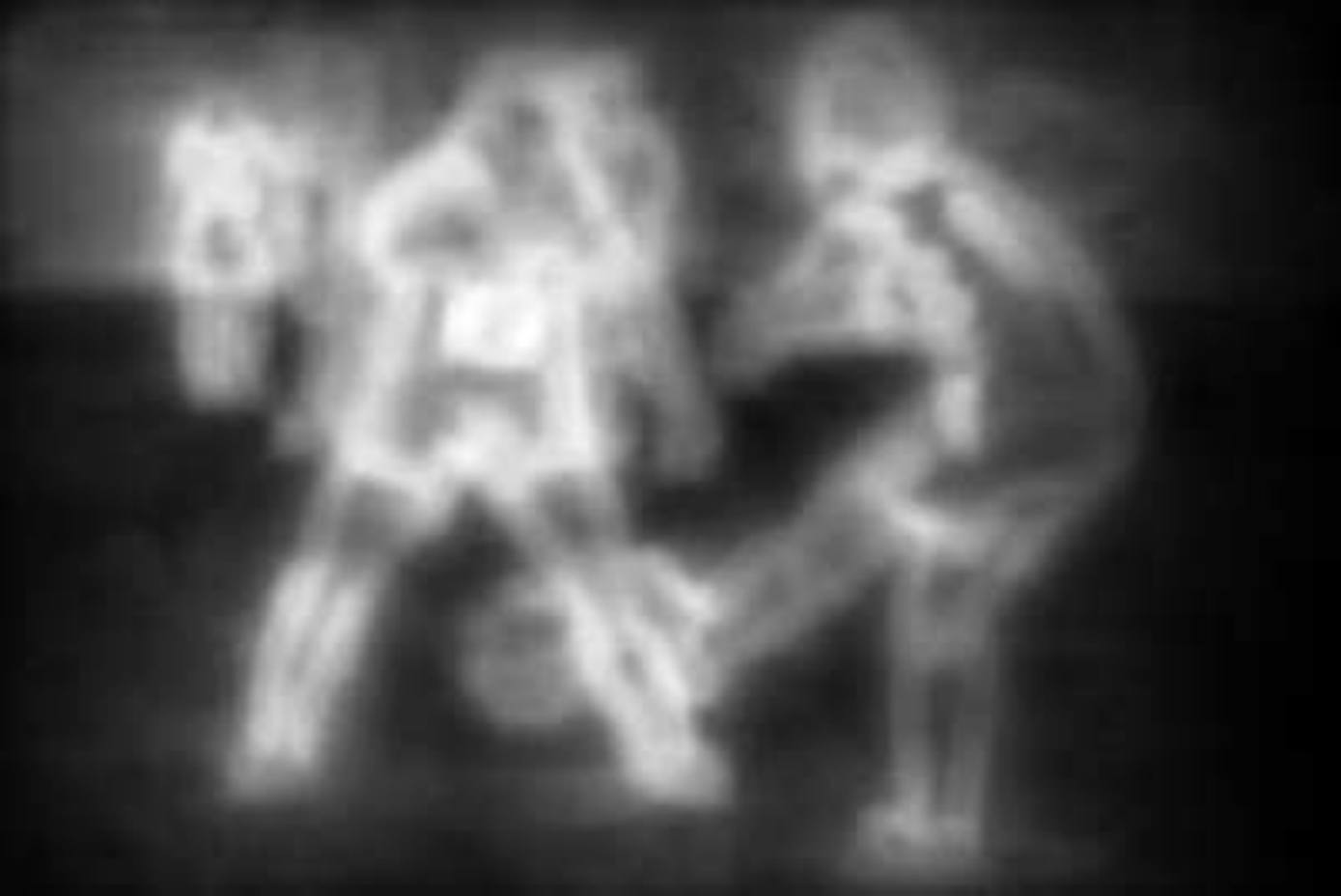}&
	\addExample{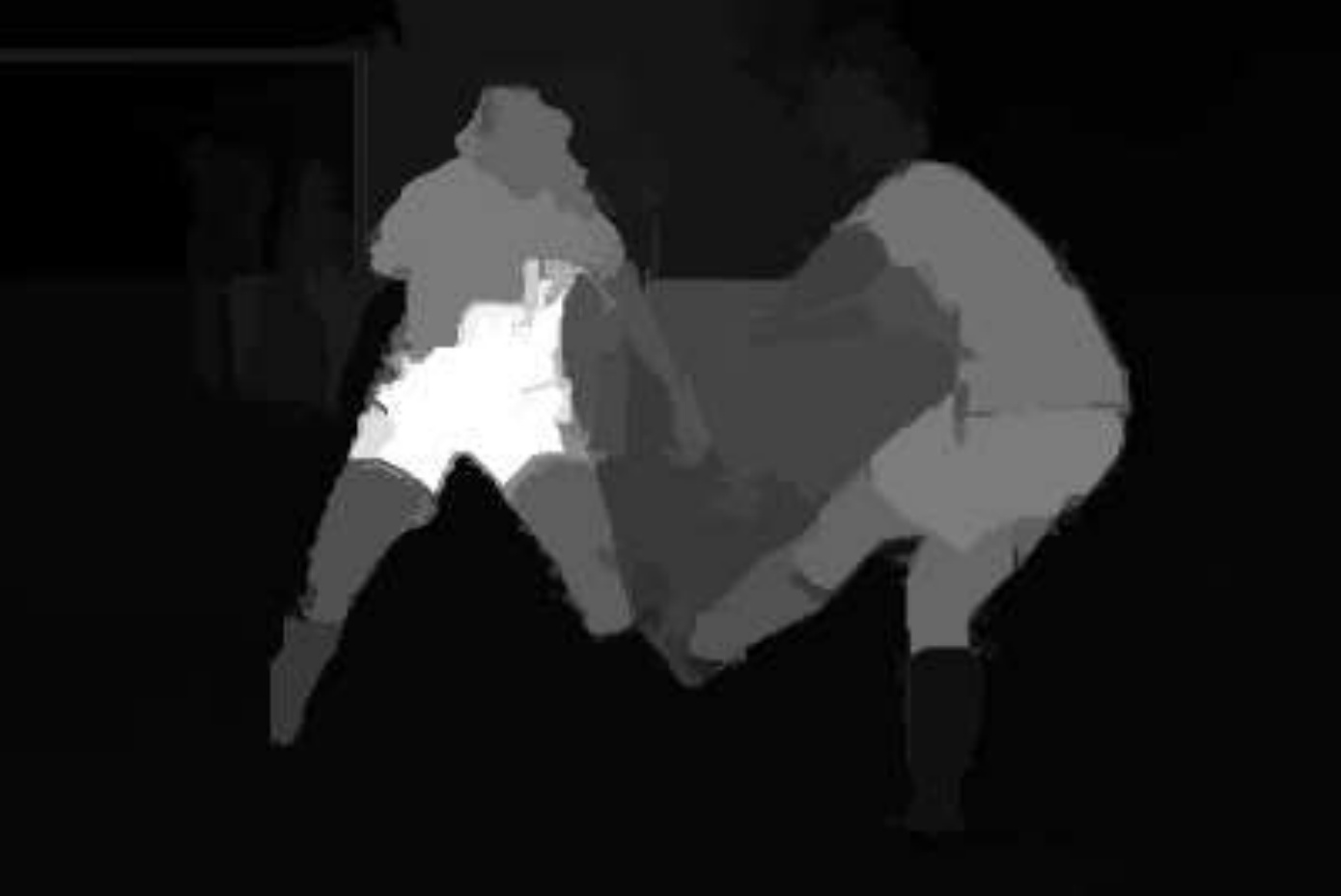}&
	\addExample{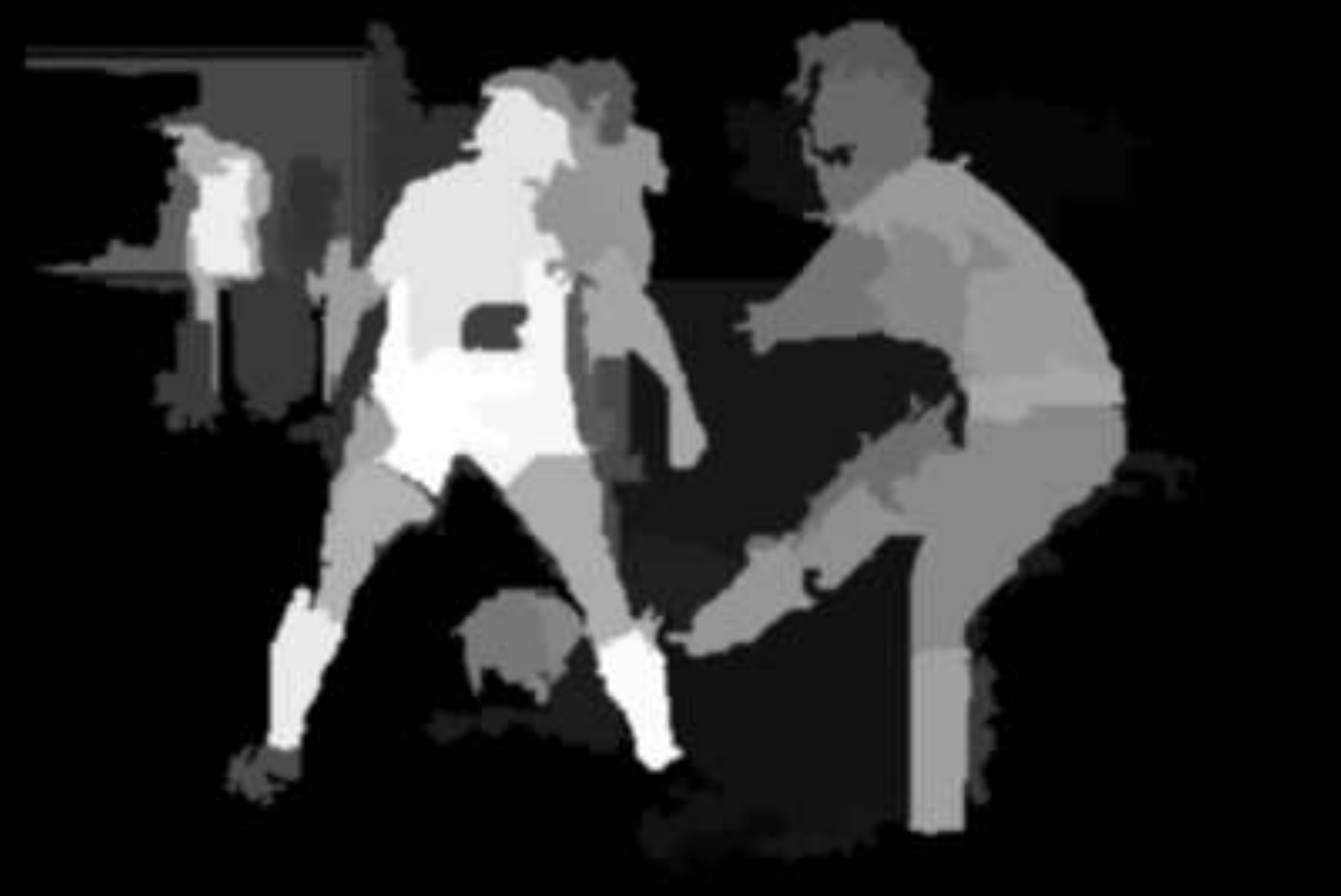}&
	\addExample{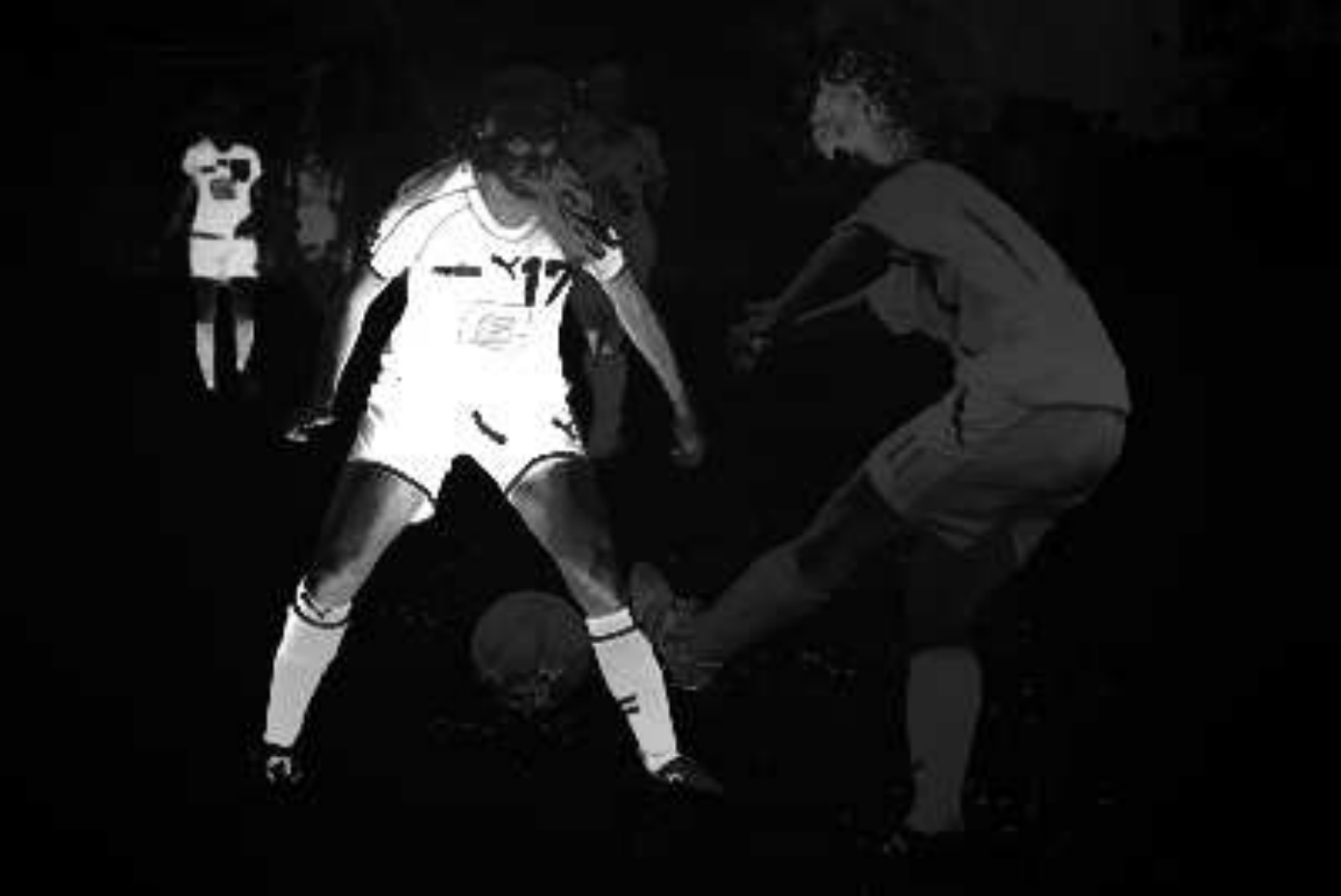}&
	\addExample{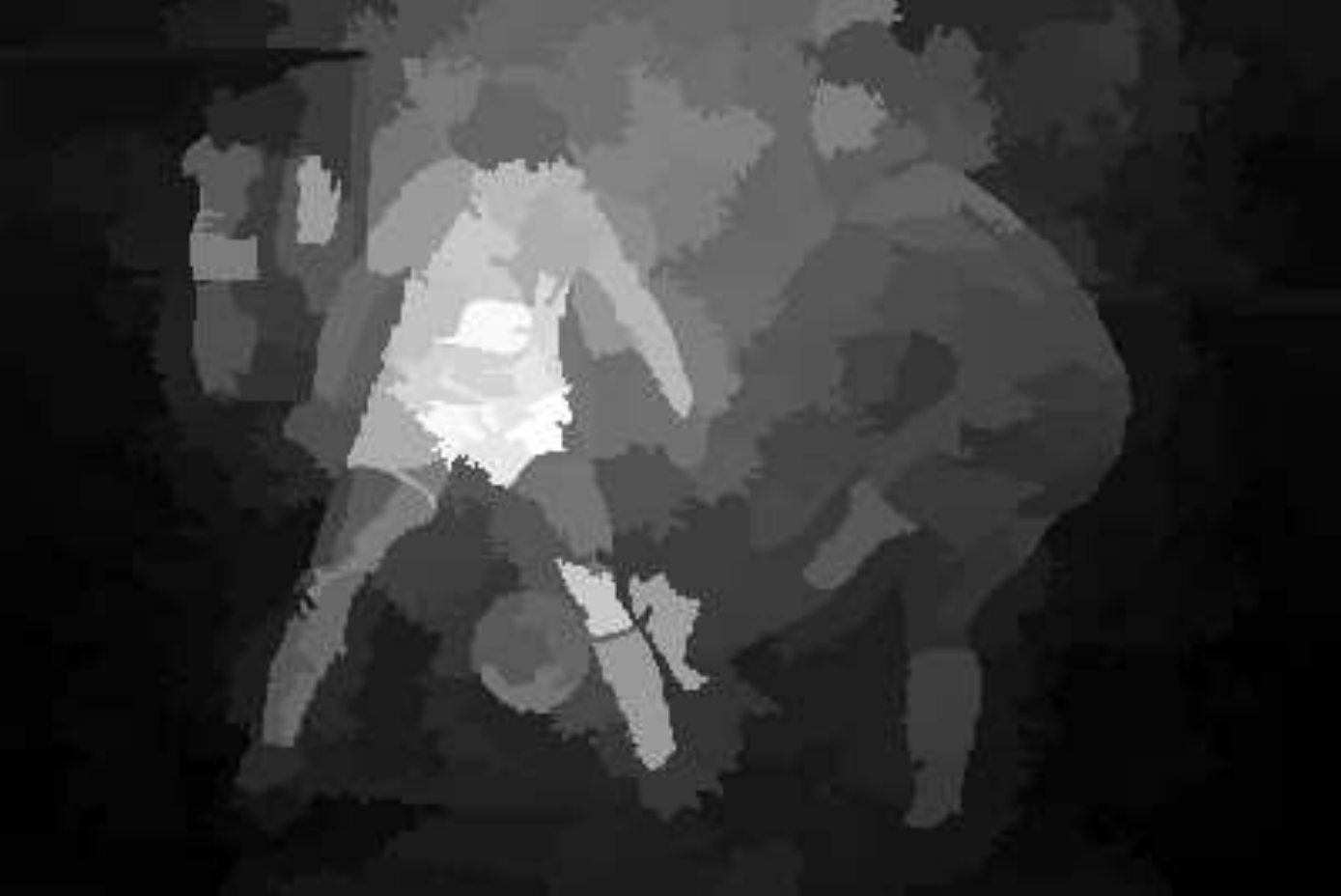}&
	\addExample{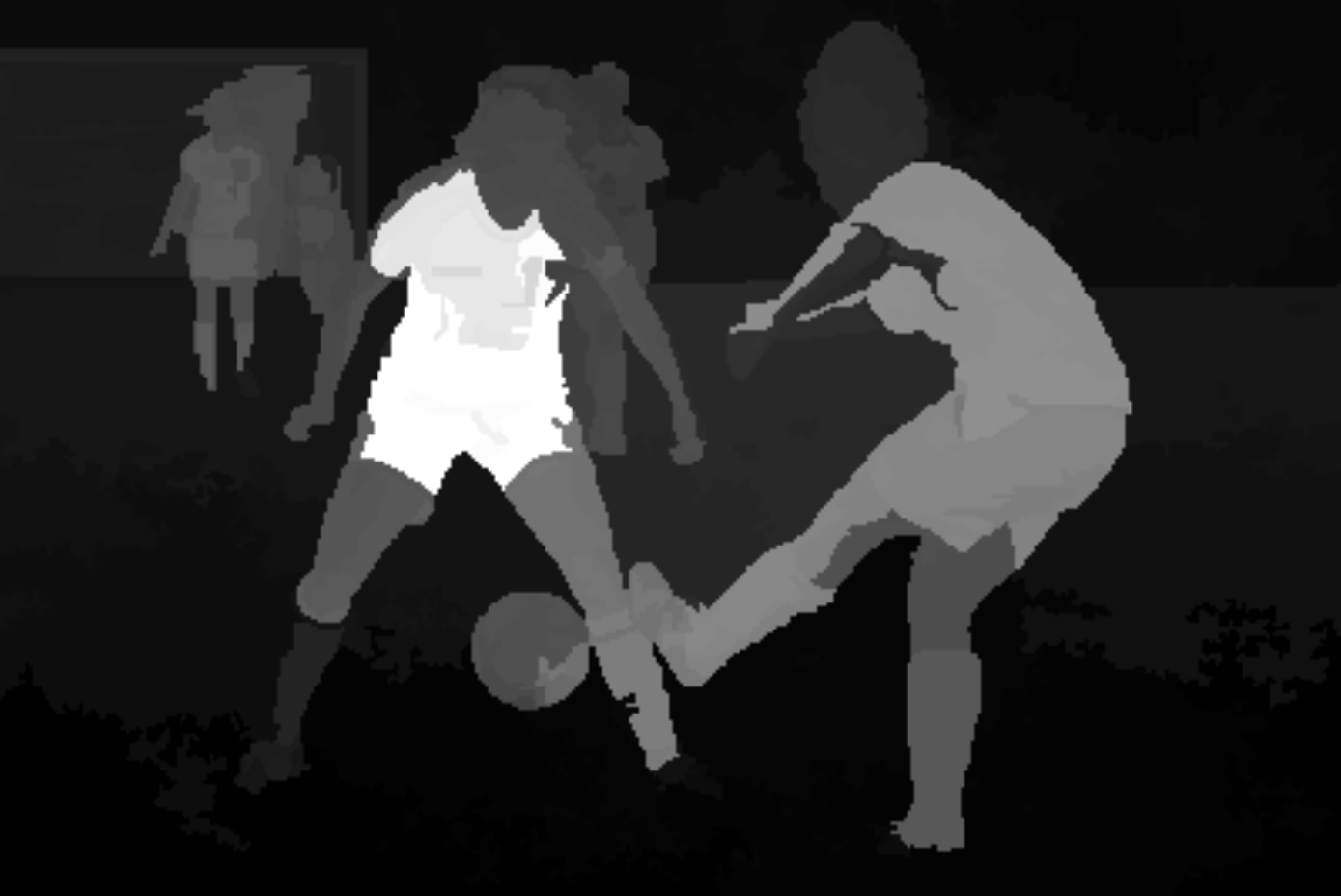}&
	\addExample{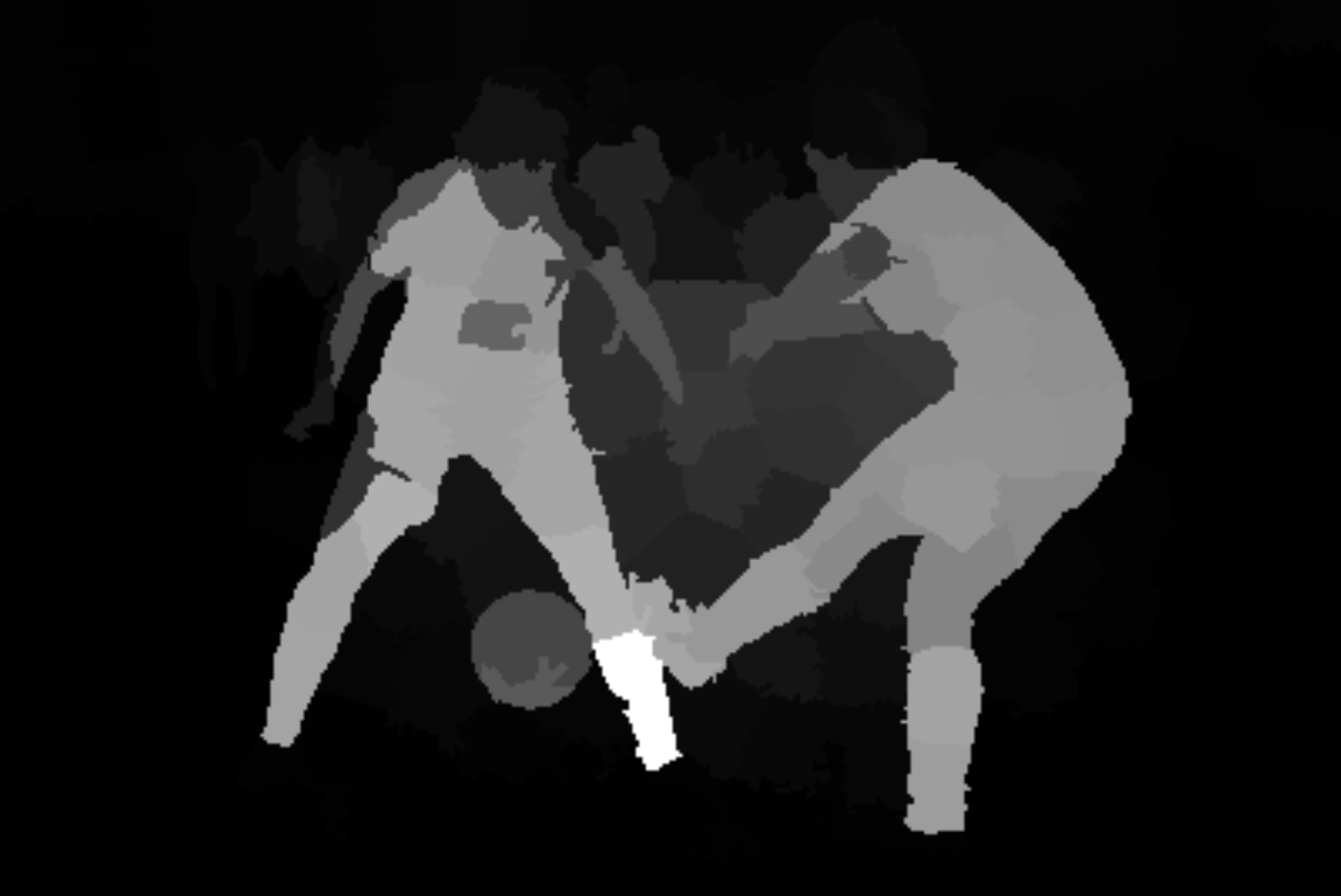}&
	\addExample{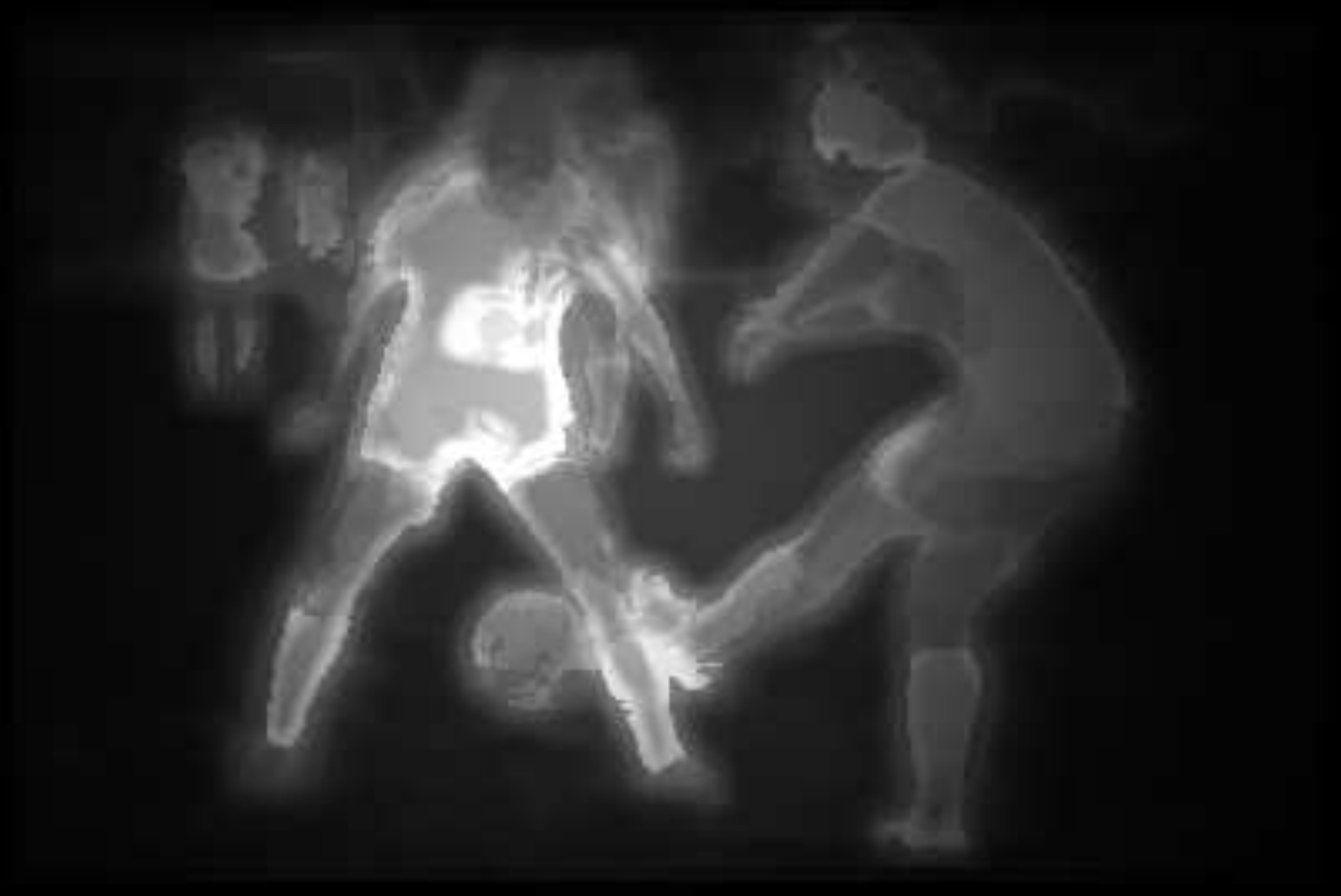}&
	\addExample{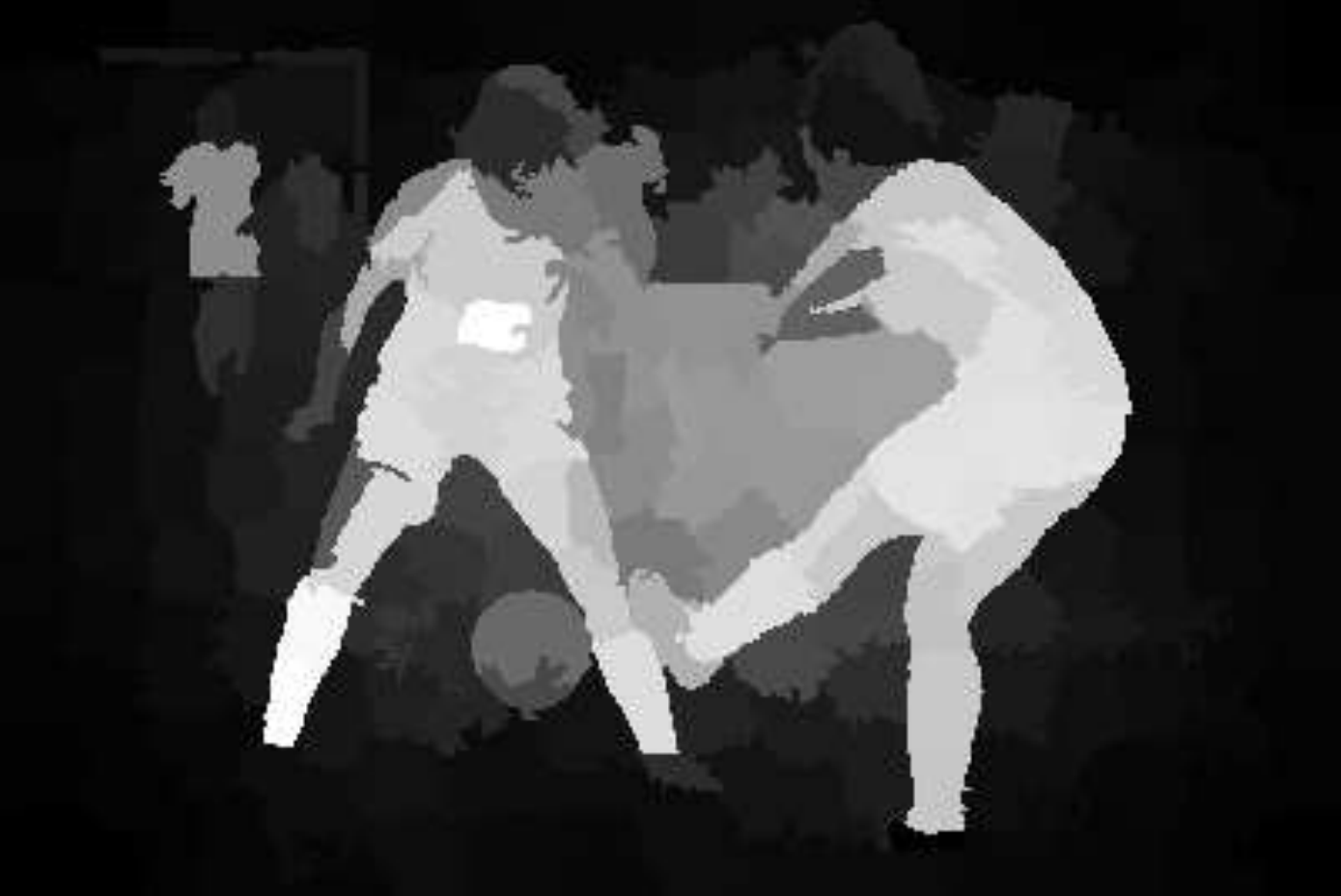}&
	\addExample{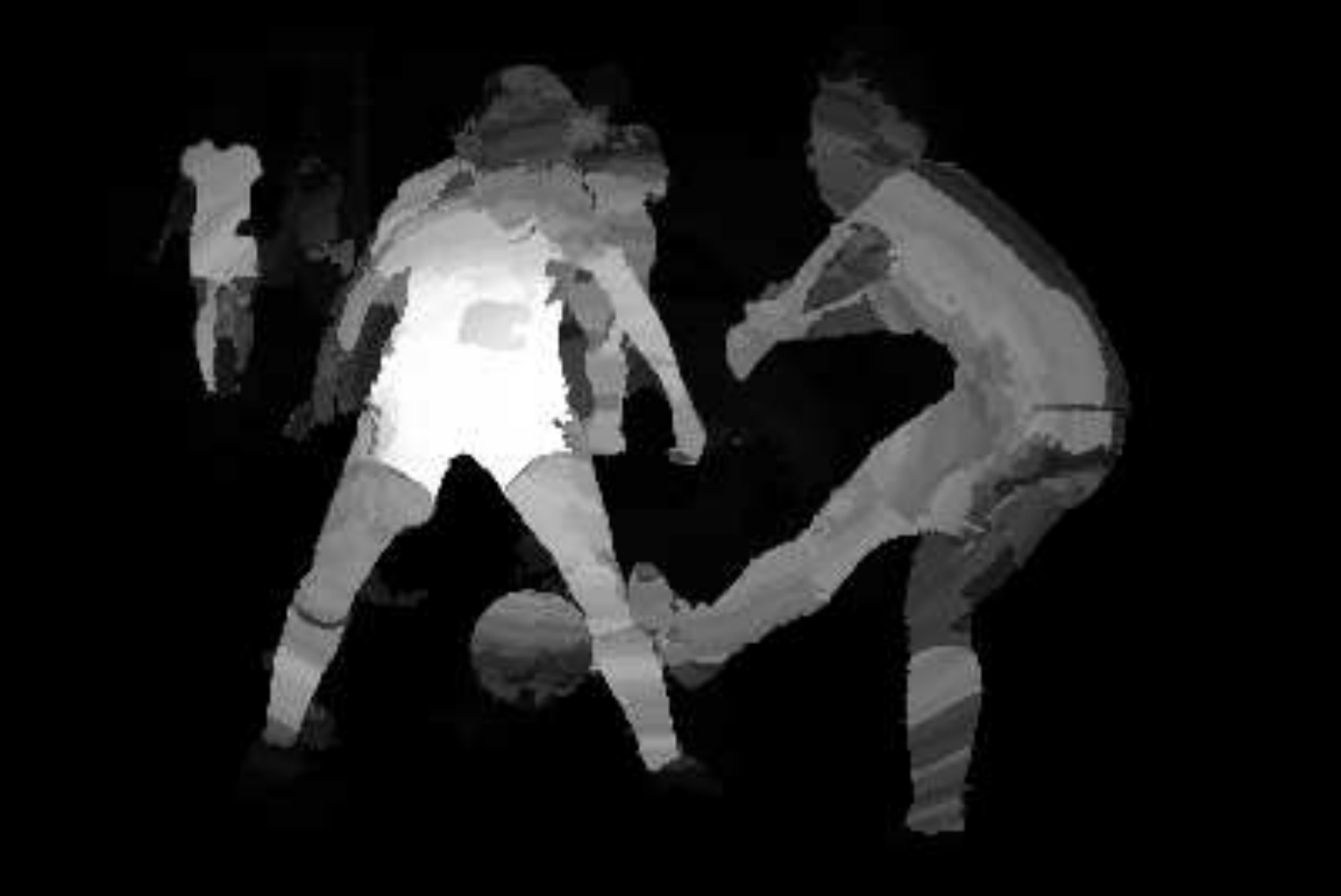}&
	\addExample{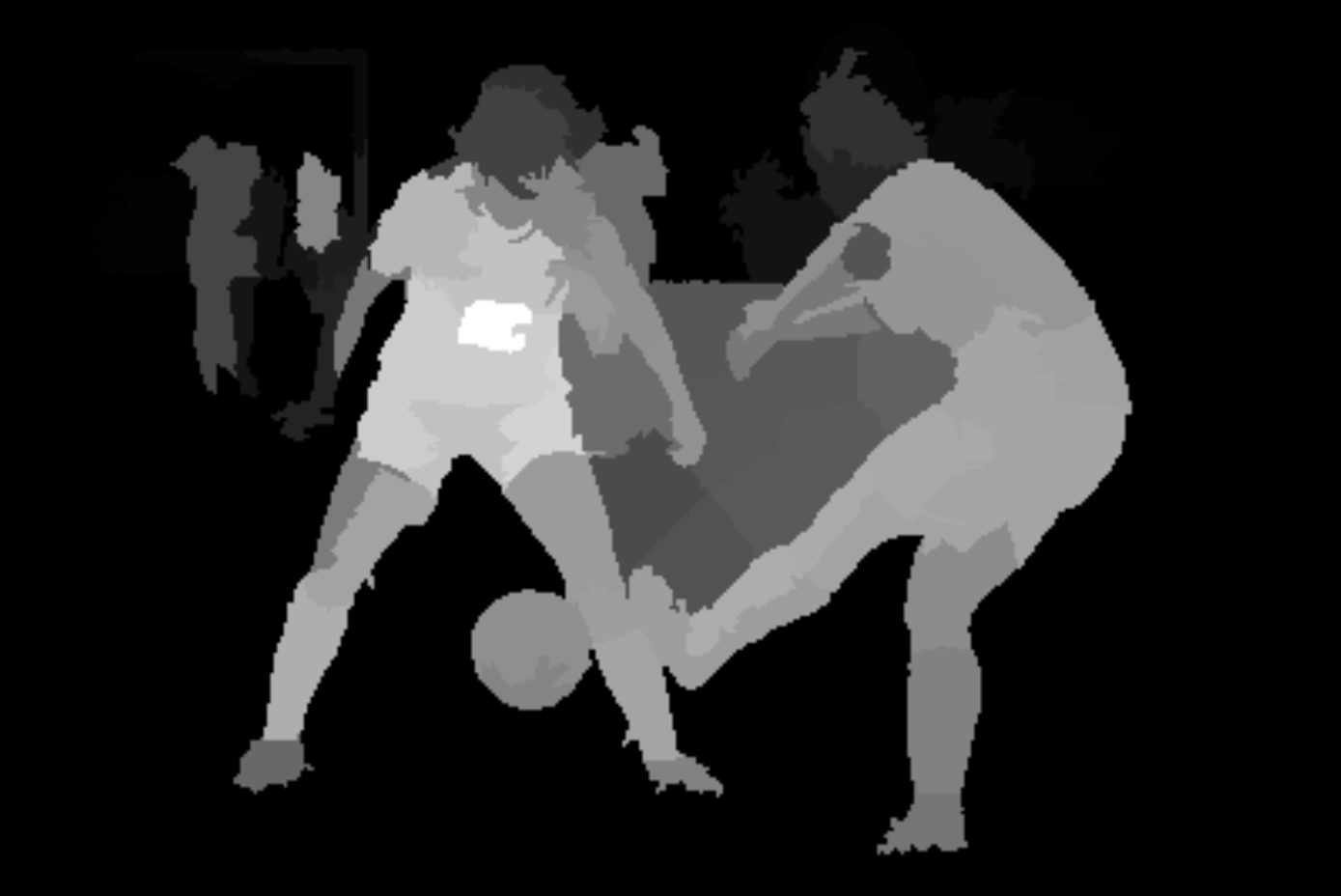}&
	\addExample{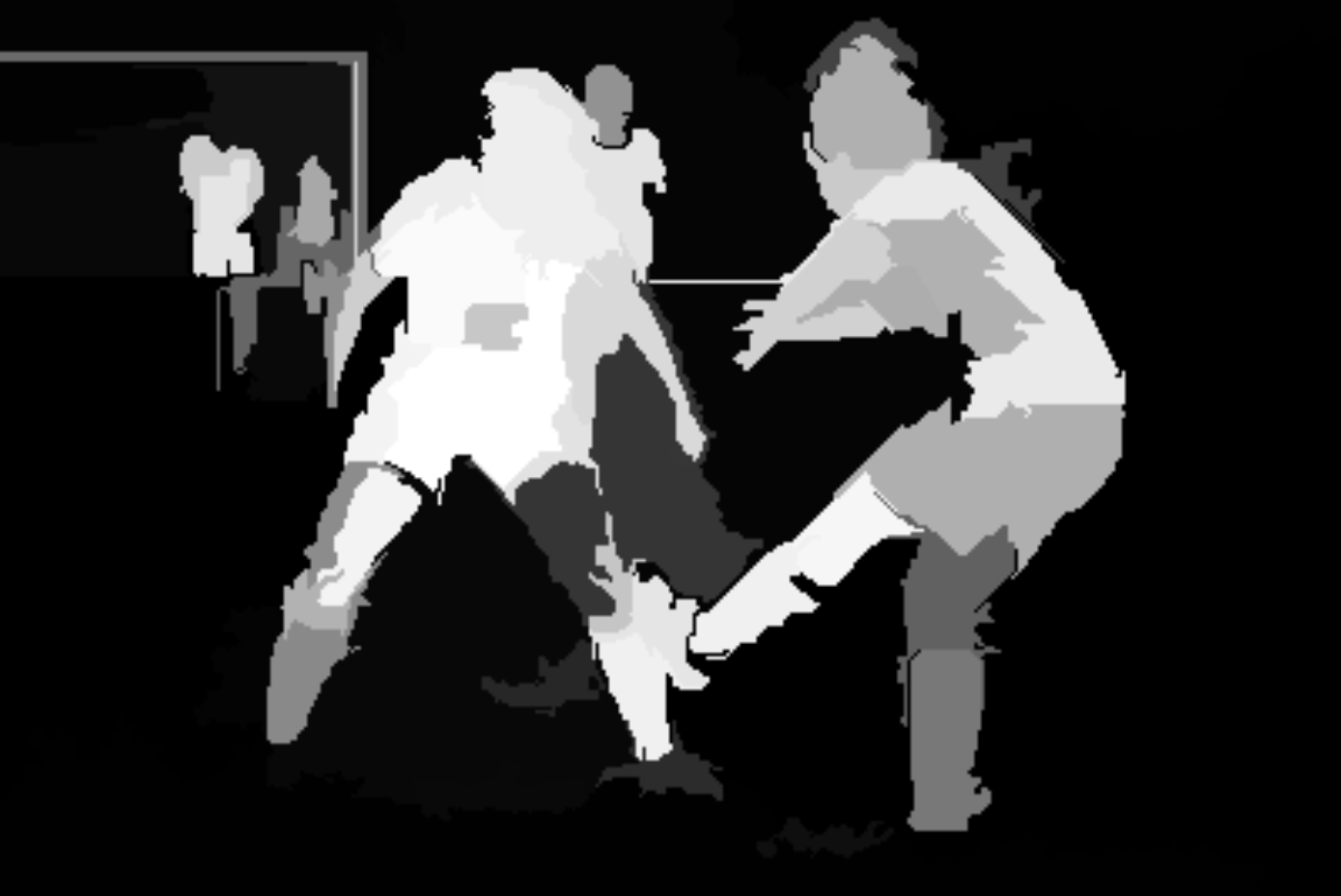}&
	\addExample{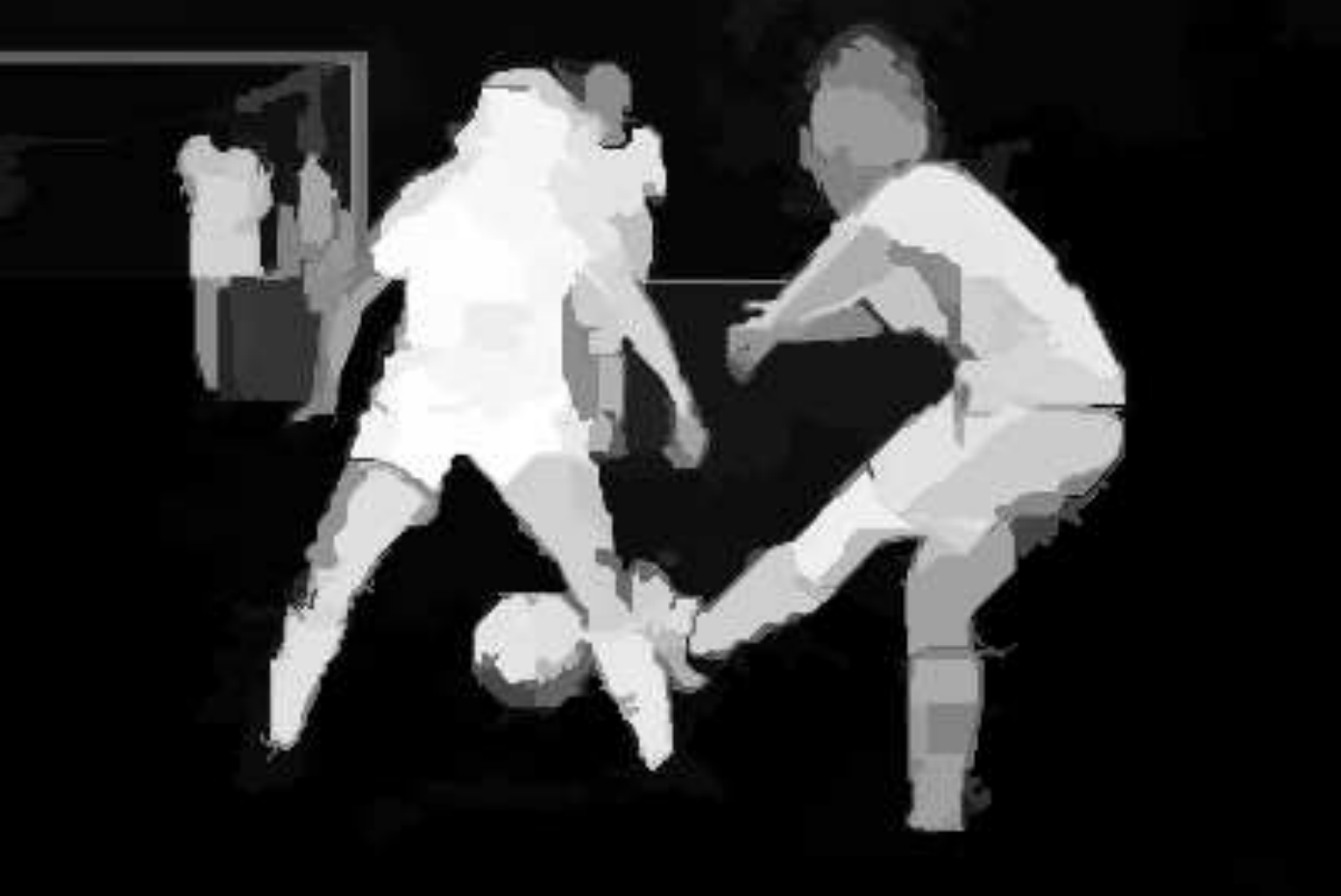}
    \\
	\addExample{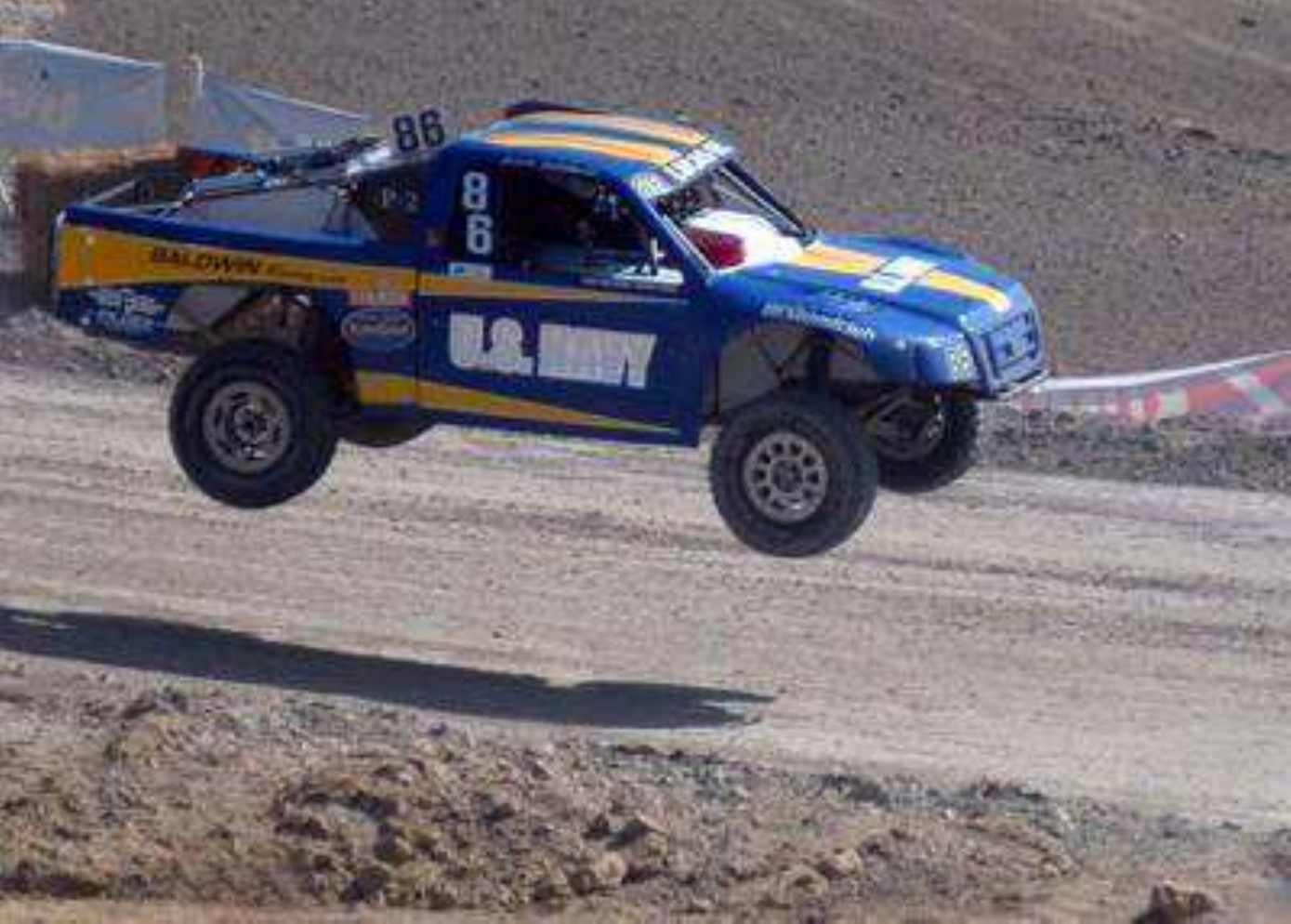}&
	\addExample{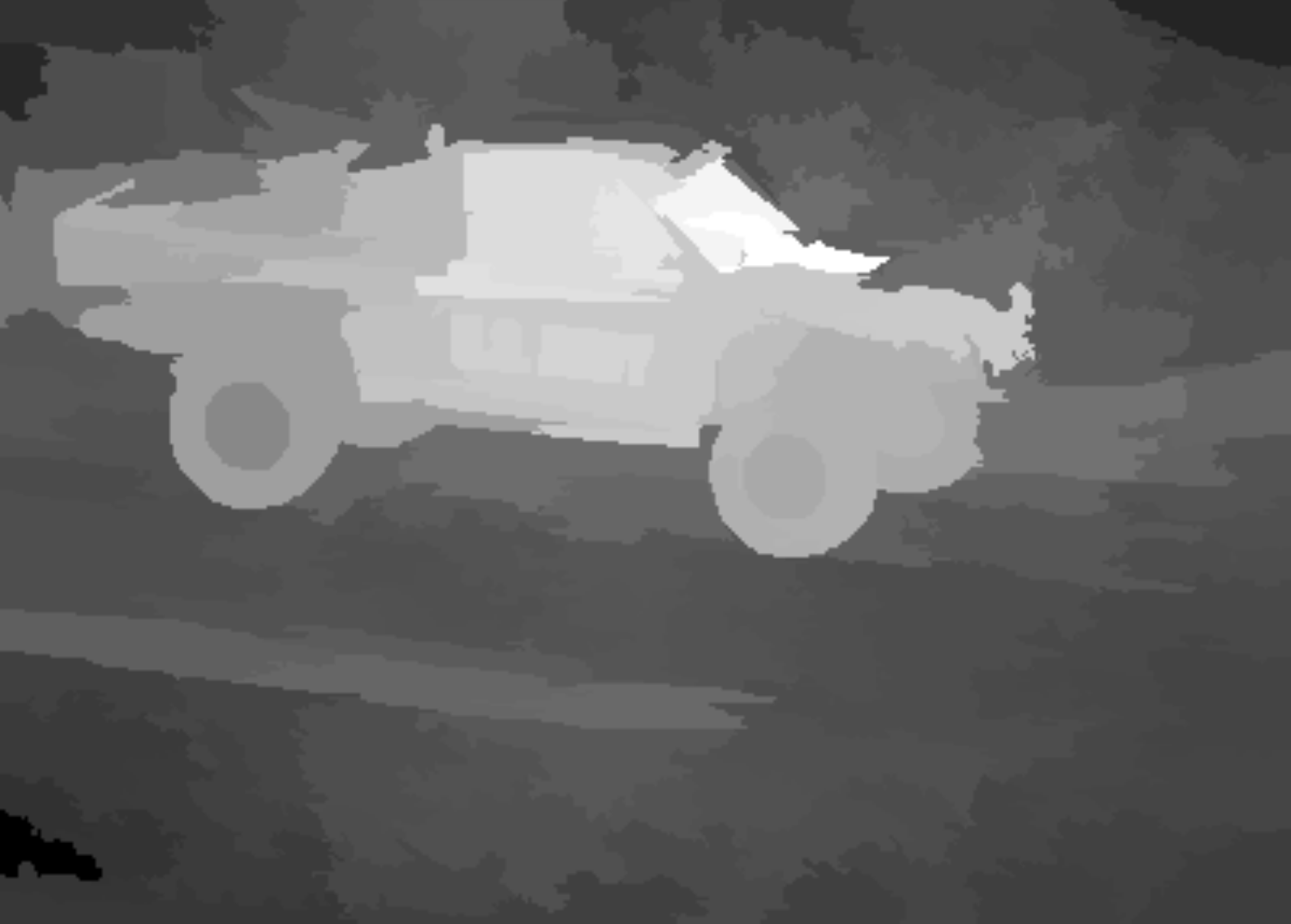}&
	\addExample{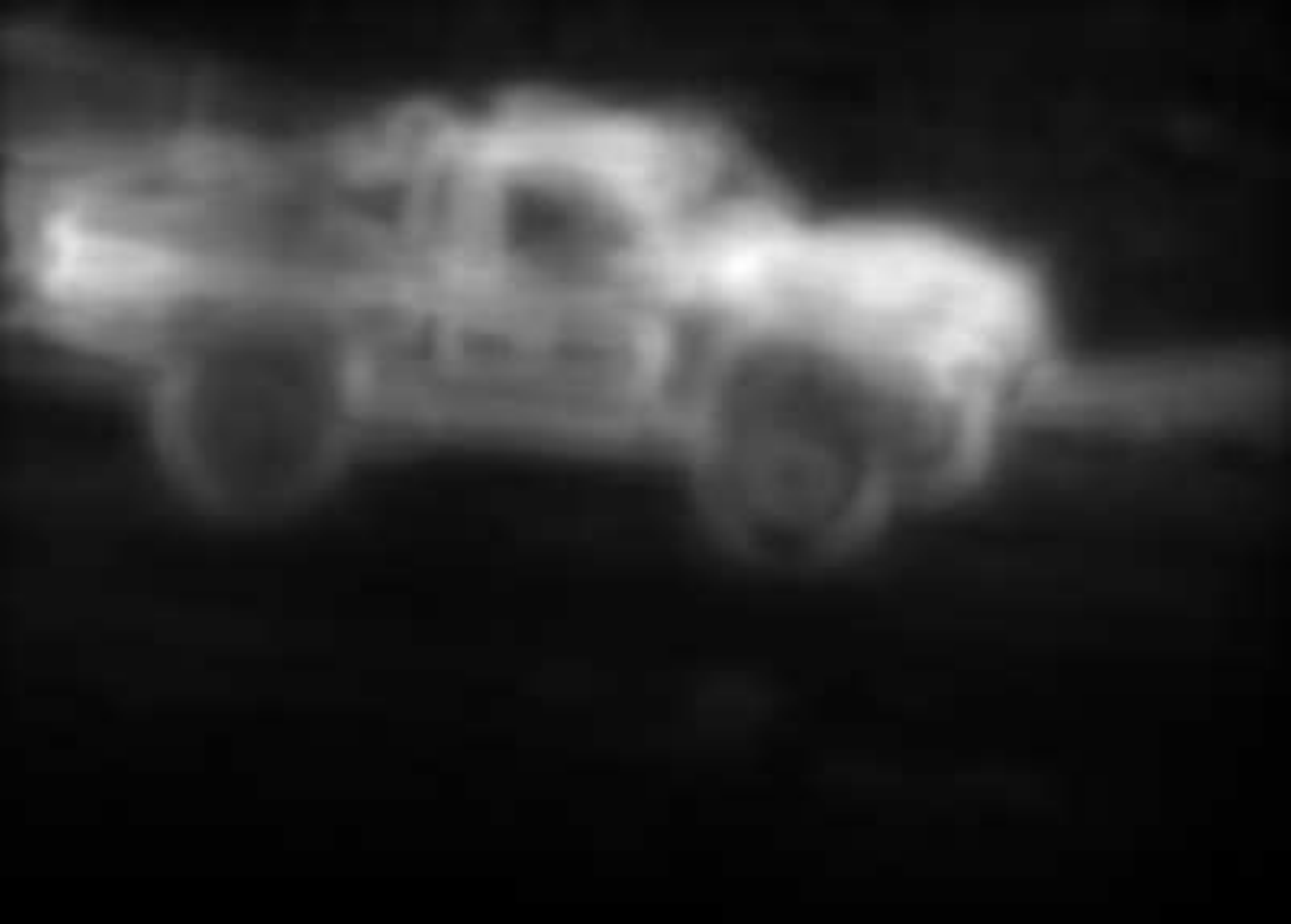}&
	\addExample{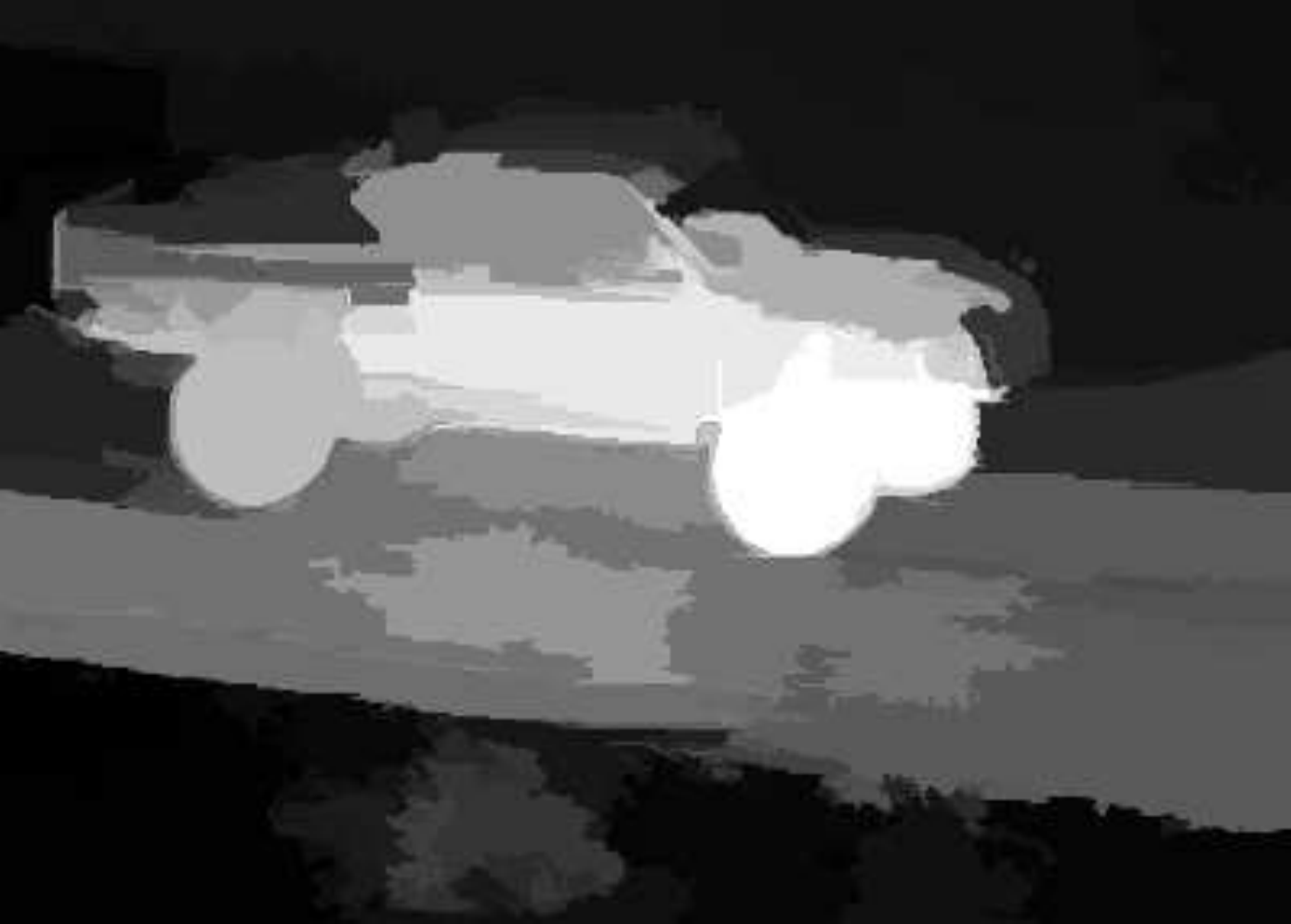}&
	\addExample{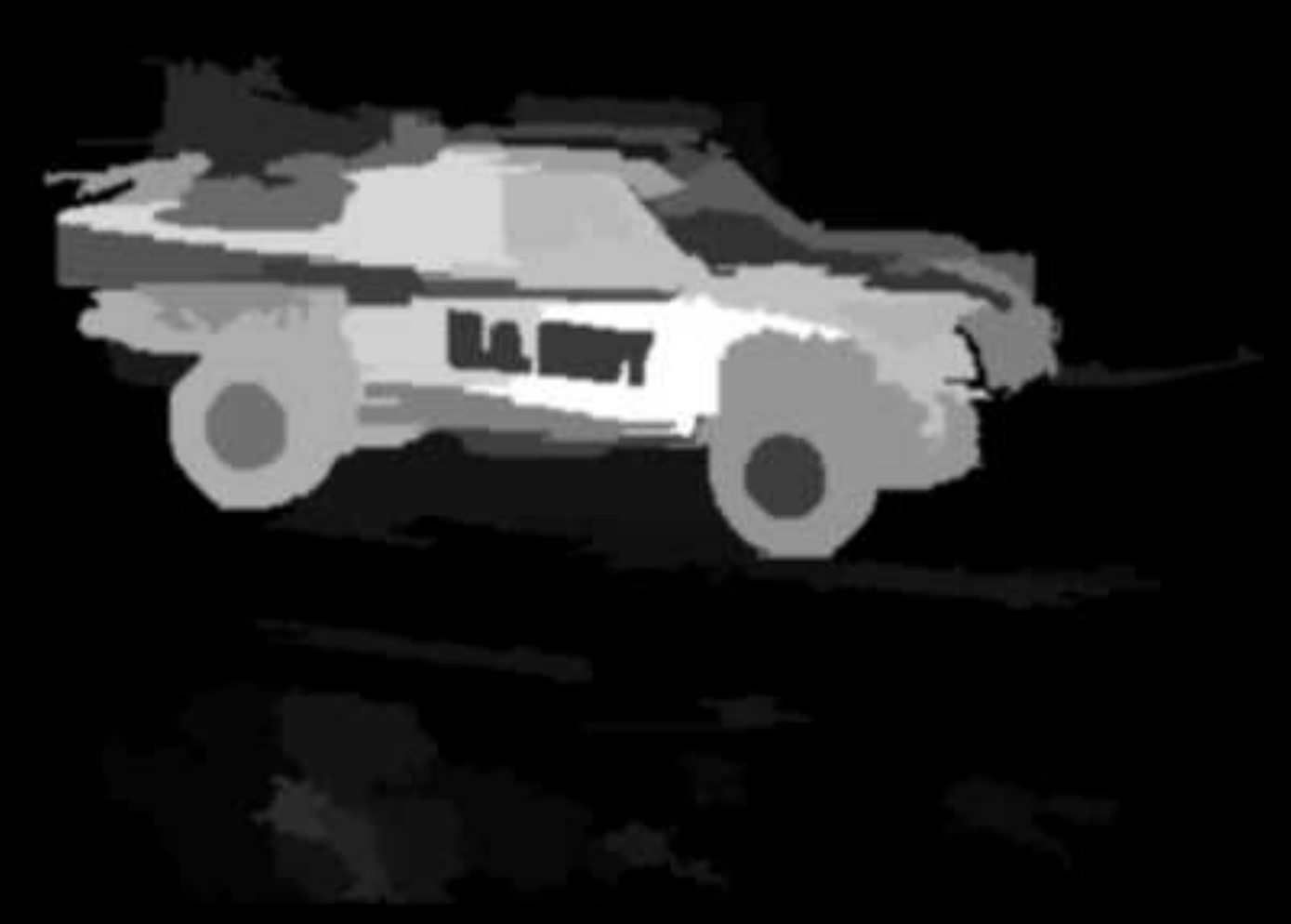}&
	\addExample{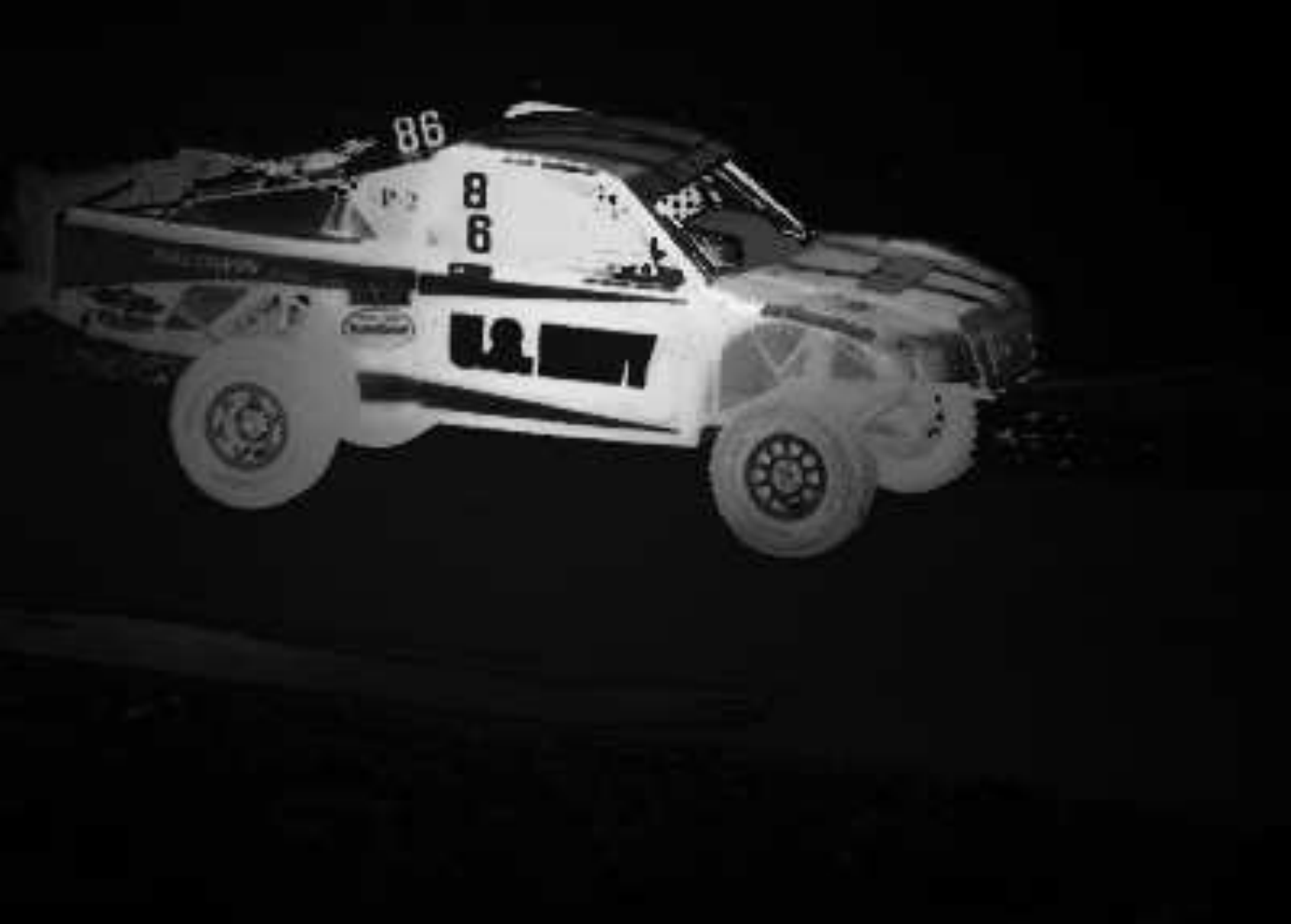}&
	\addExample{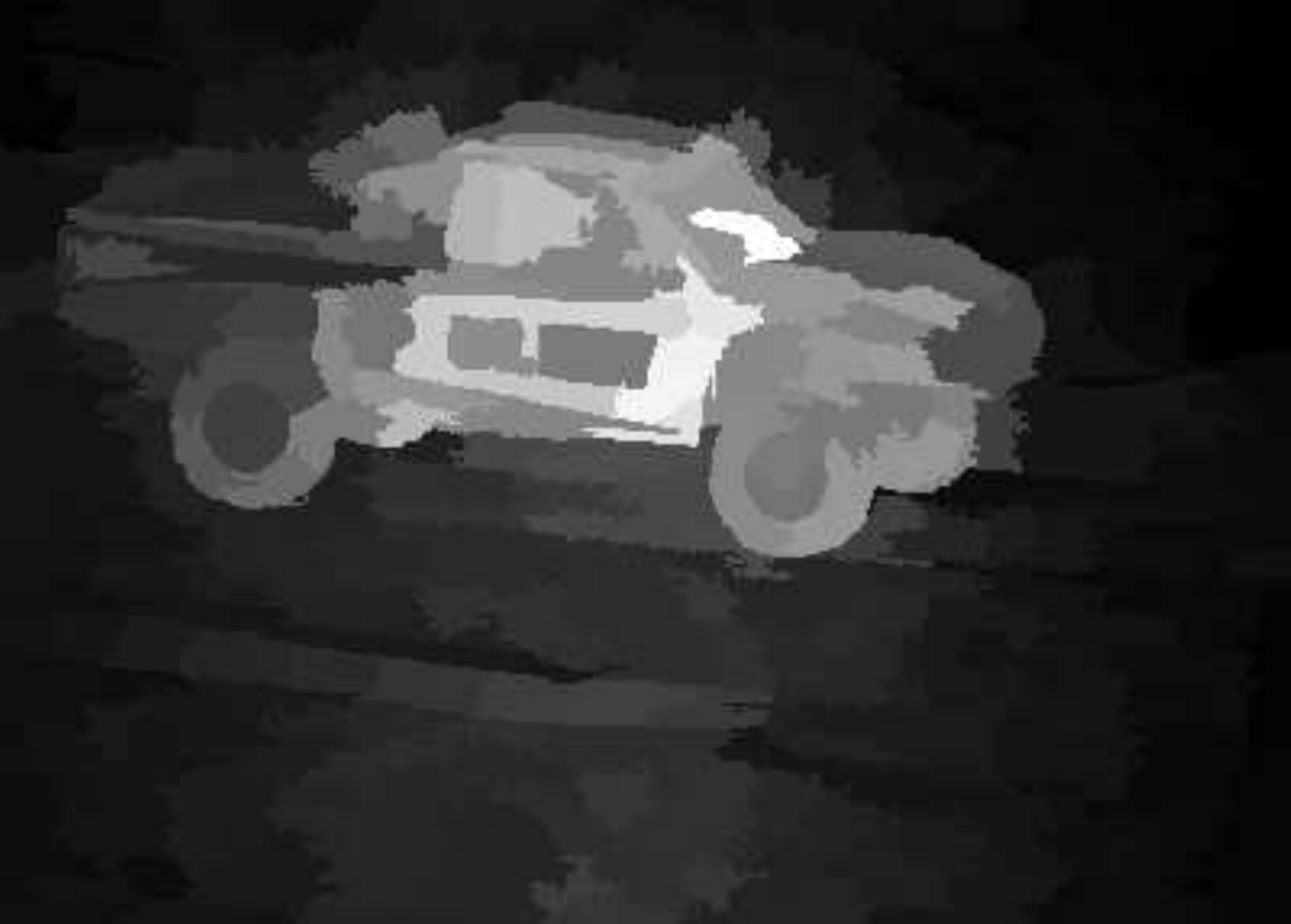}&
	\addExample{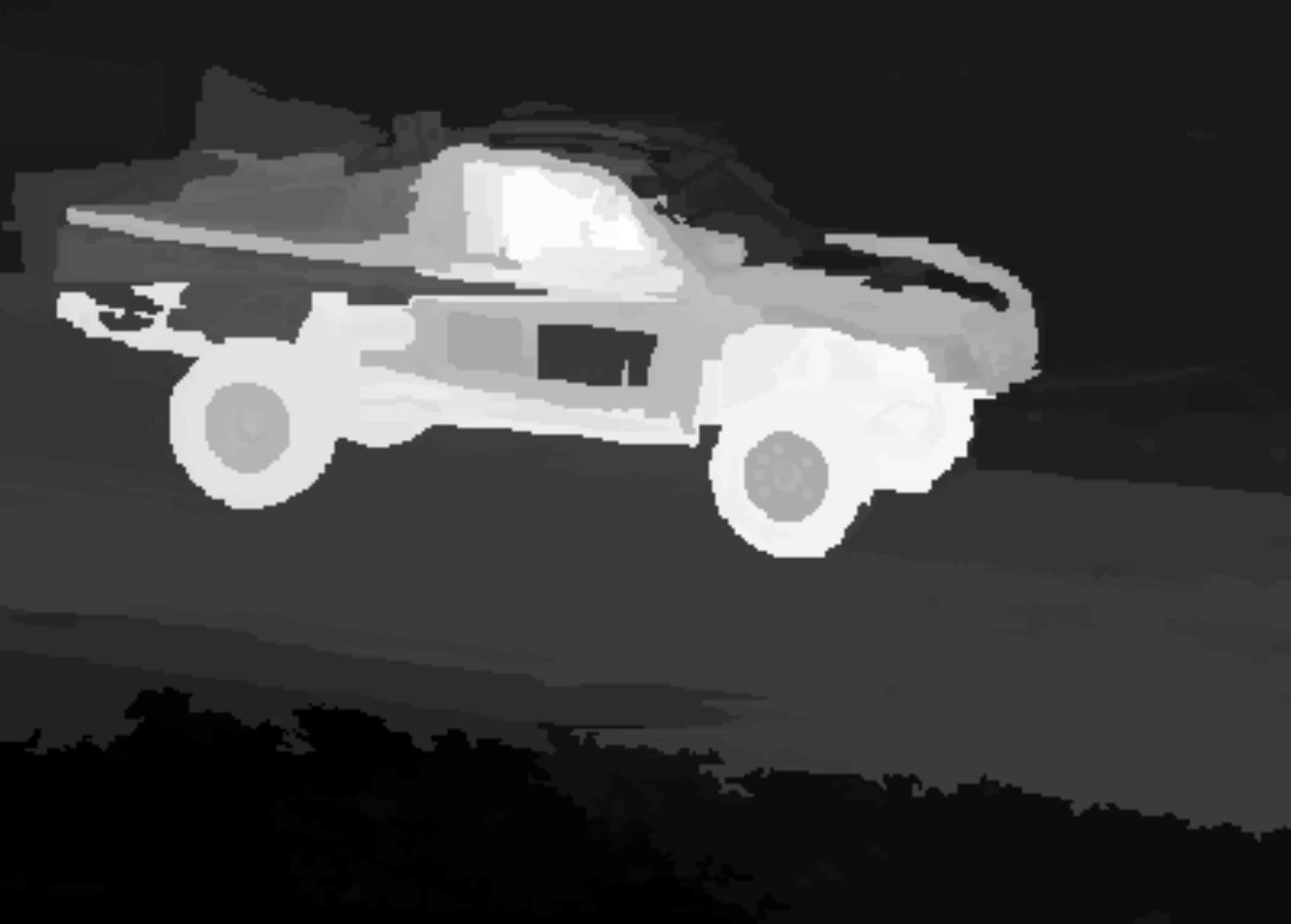}&
	\addExample{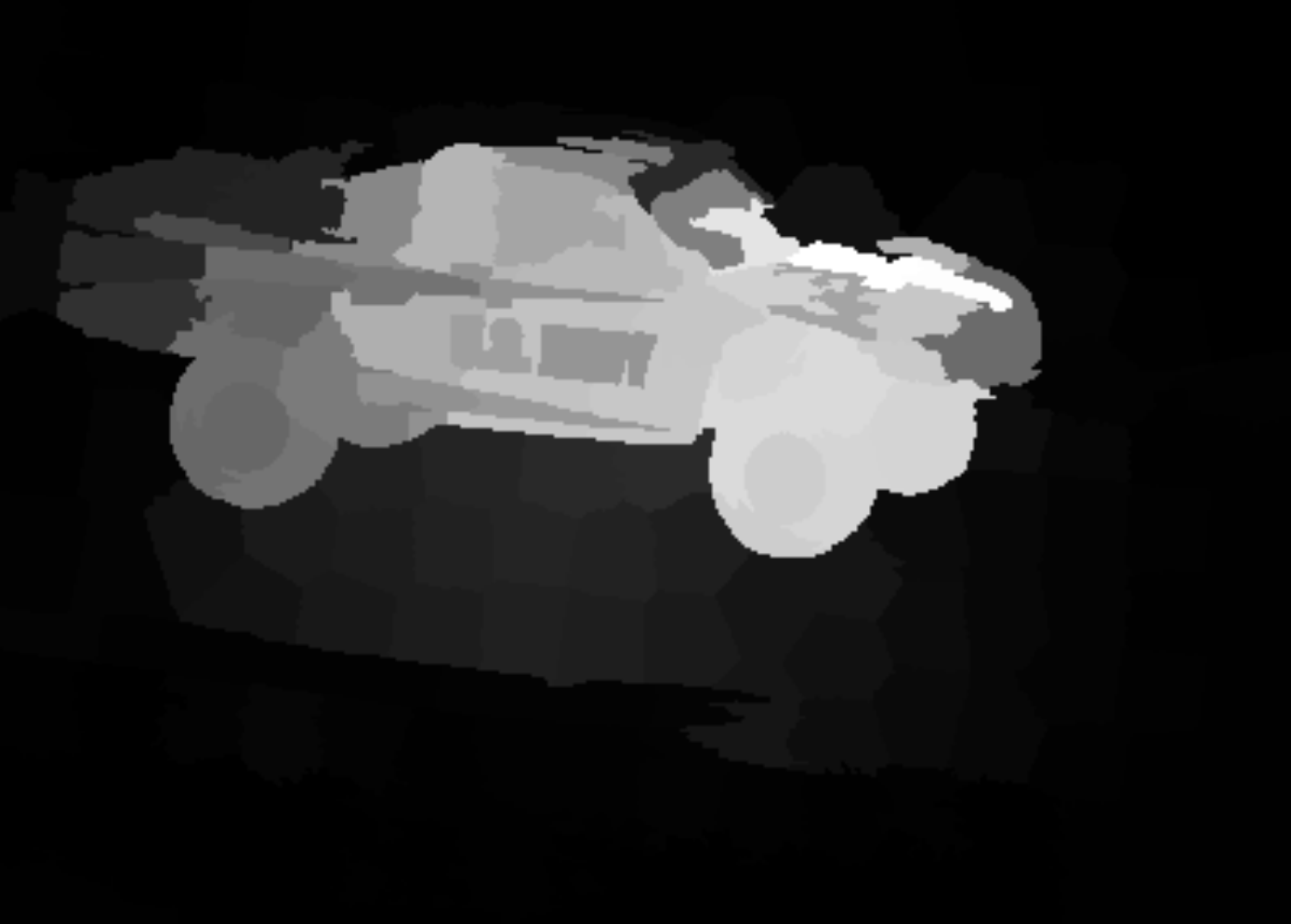}&
	\addExample{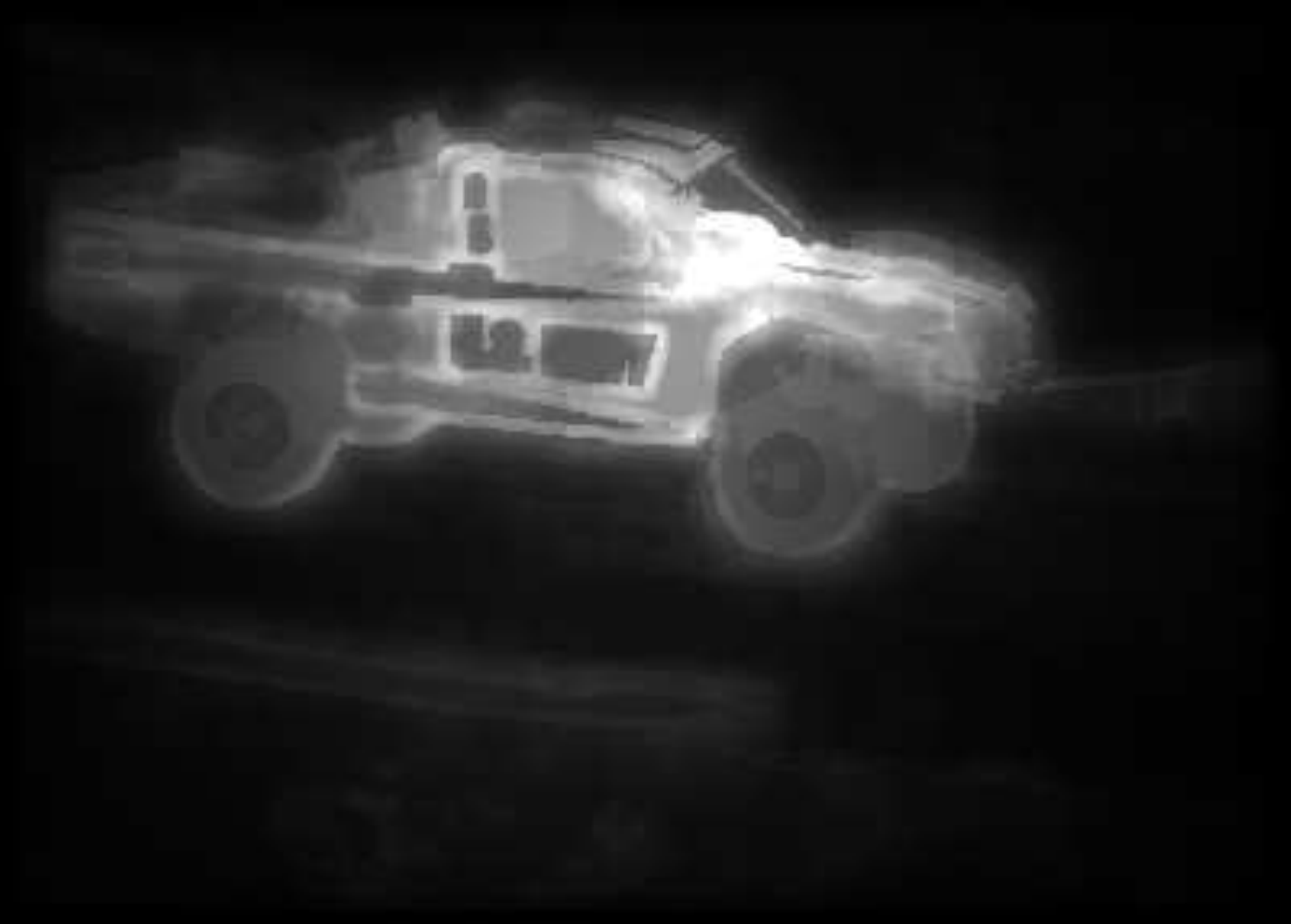}&
	\addExample{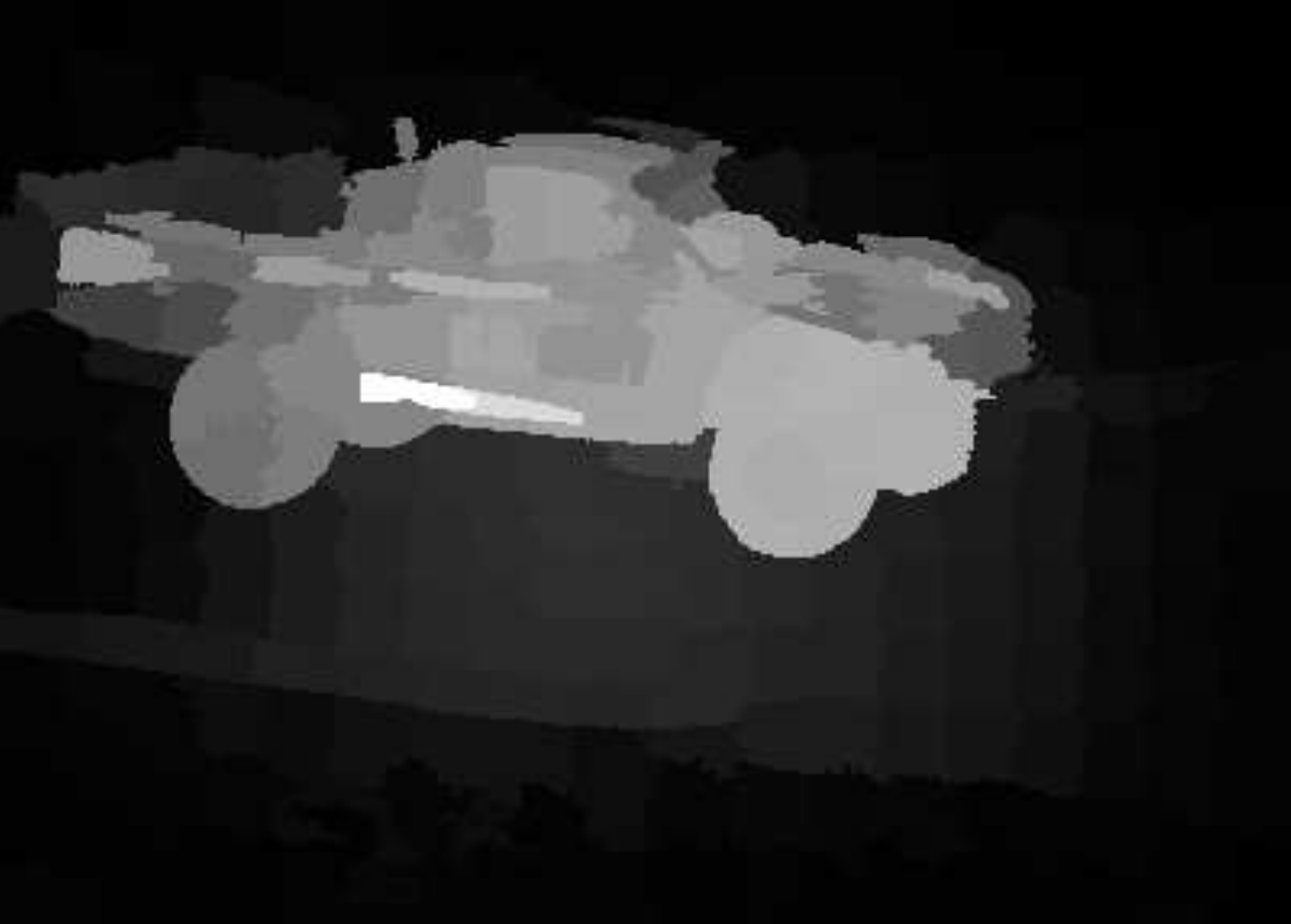}&
	\addExample{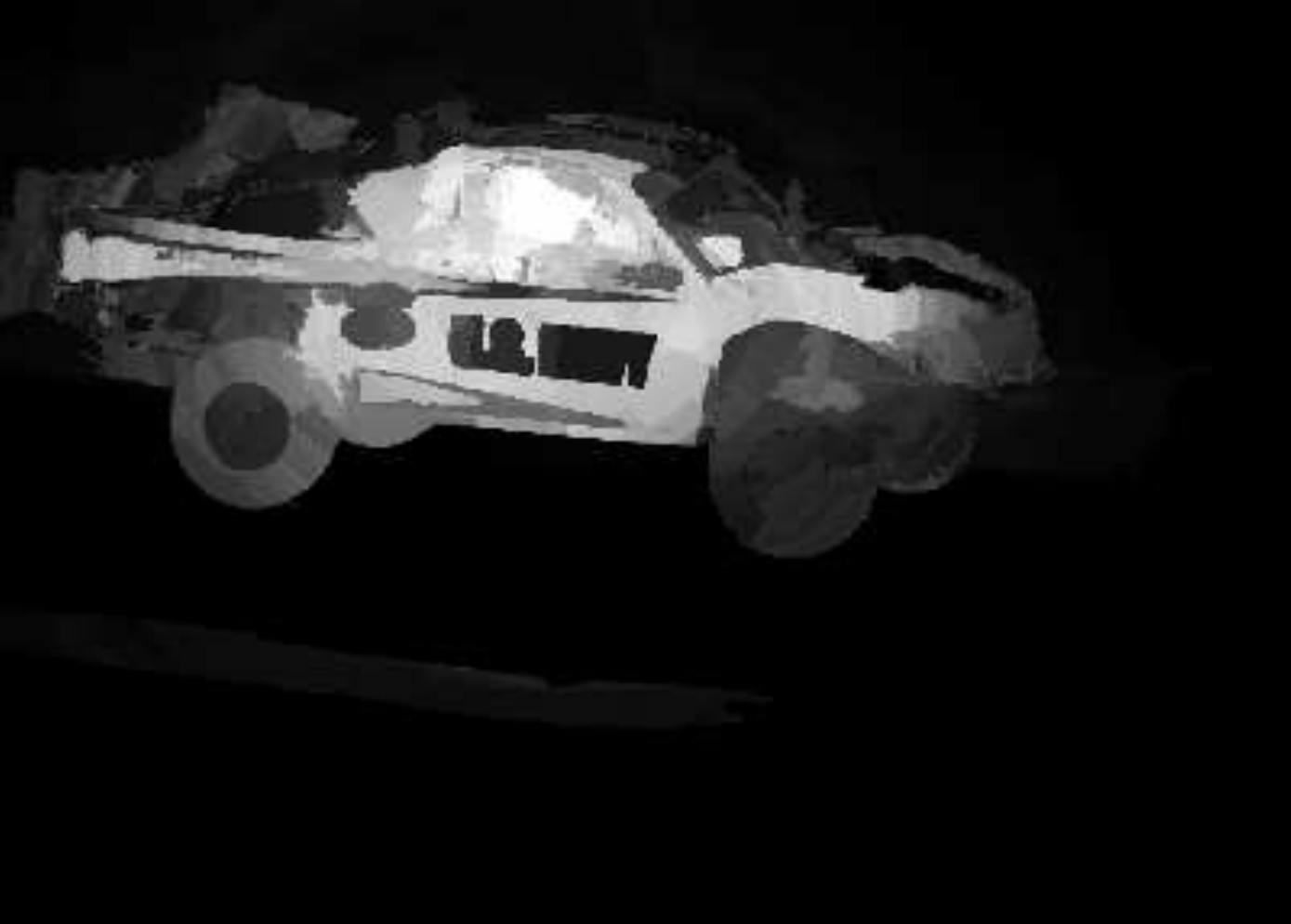}&
	\addExample{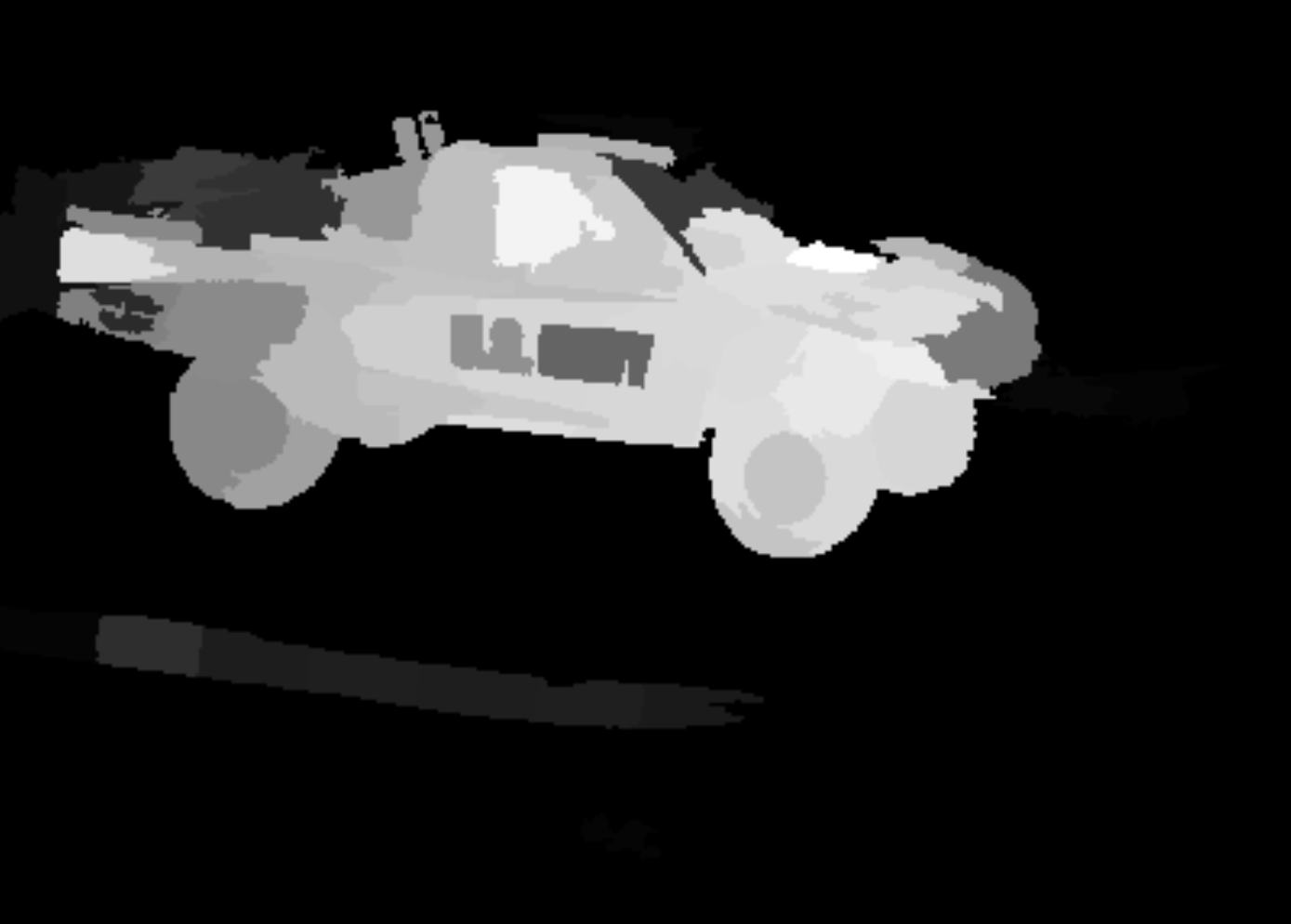}&
	\addExample{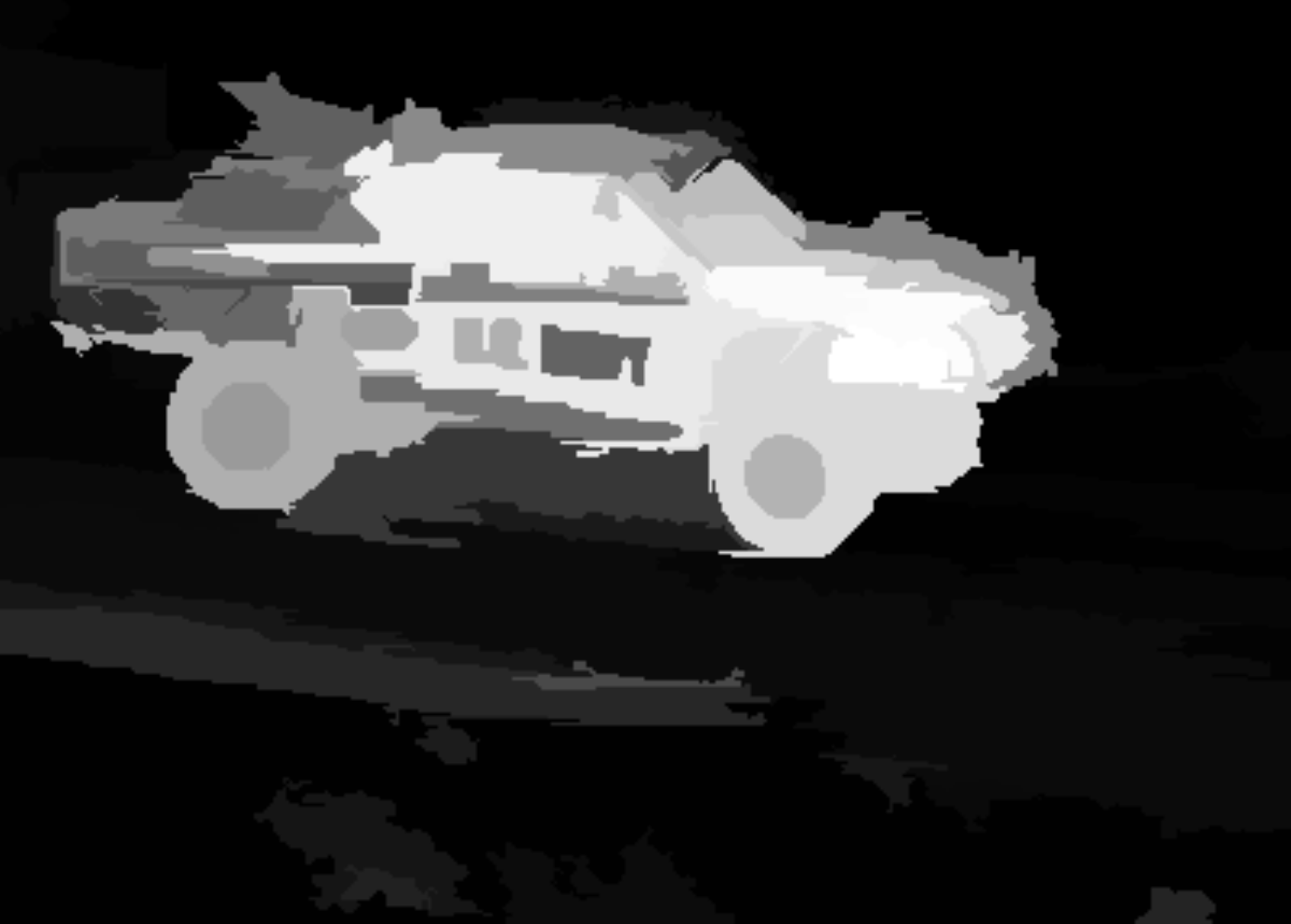}&
	\addExample{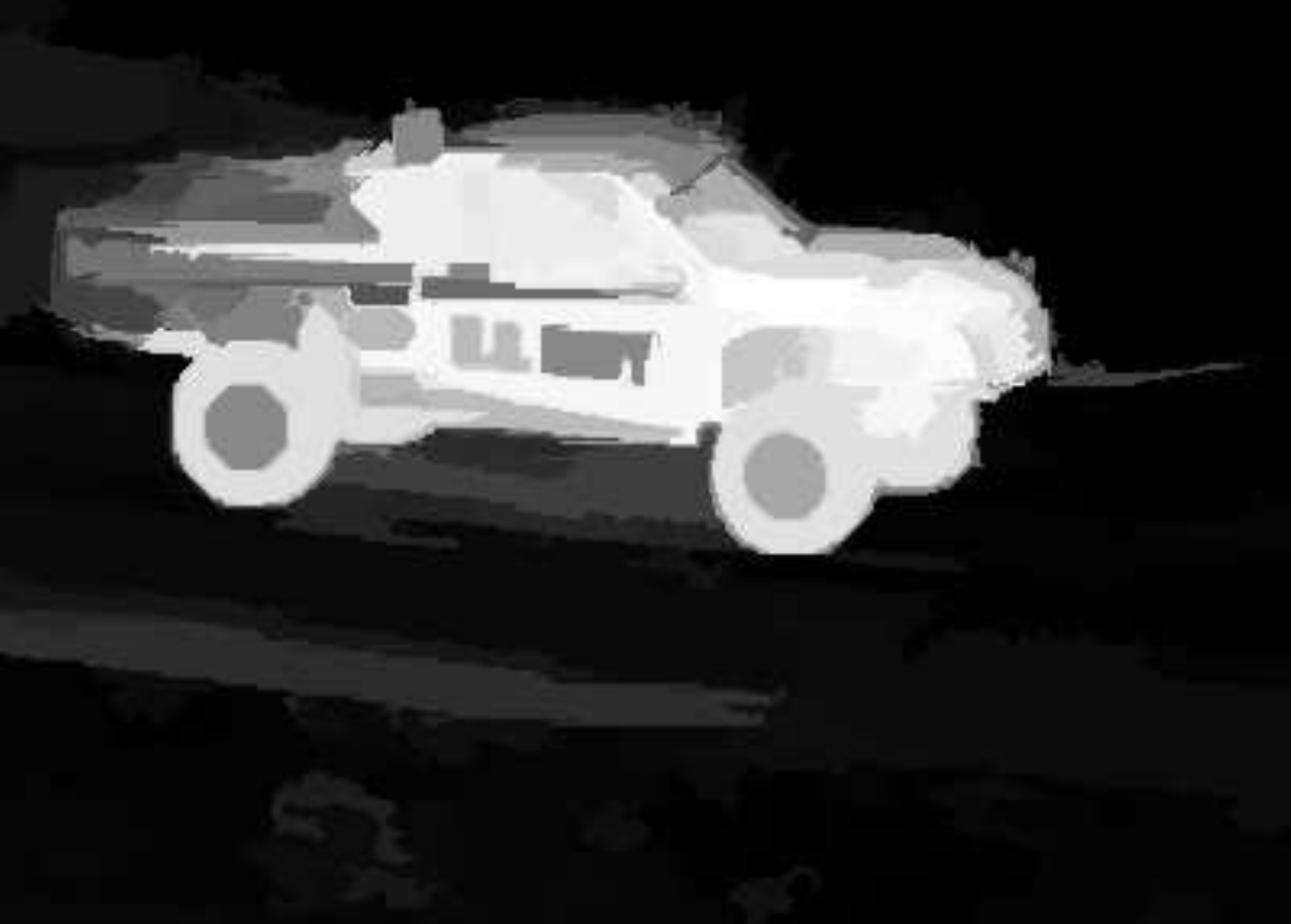}
    \\
	\addExample{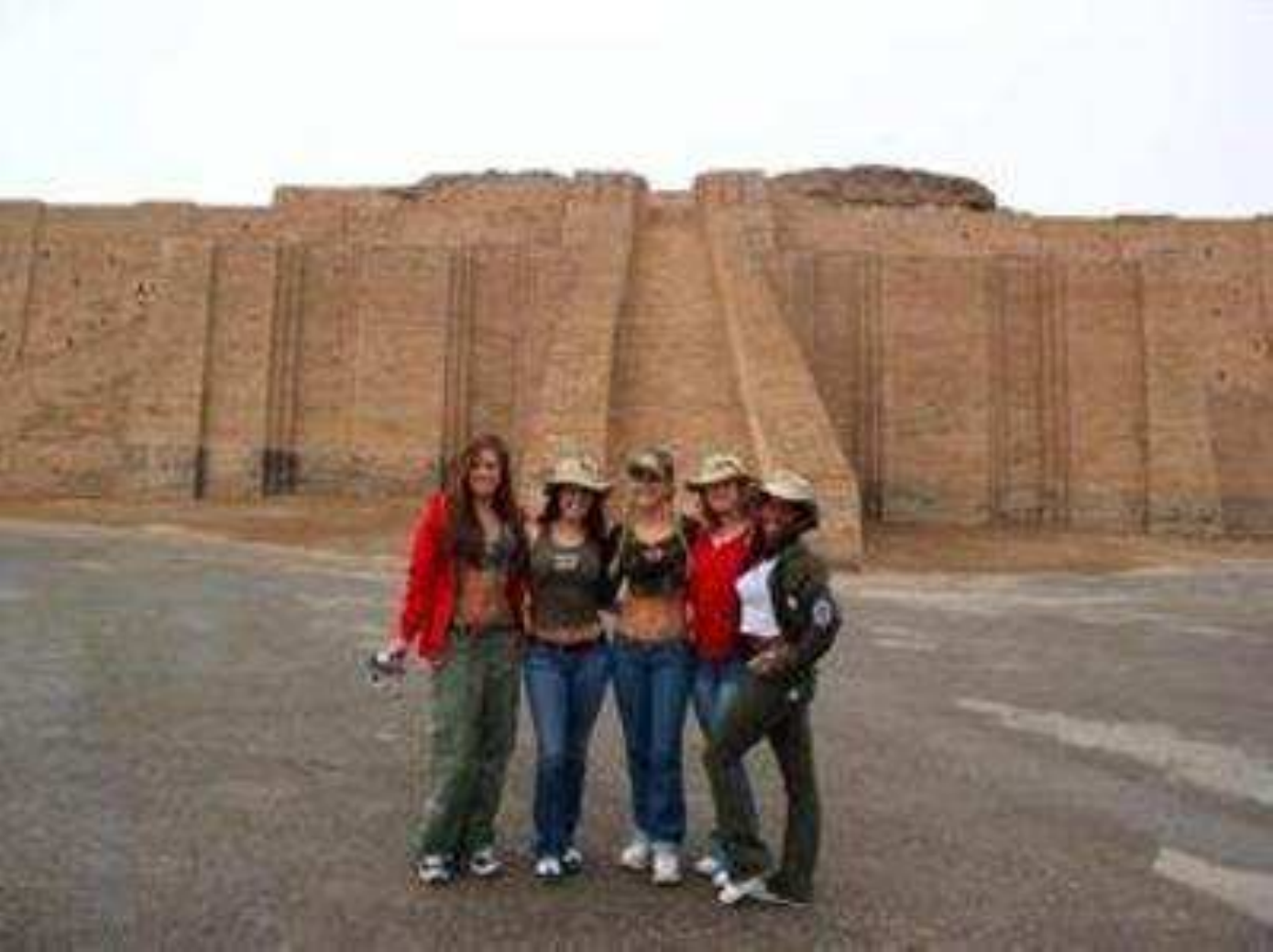}&
	\addExample{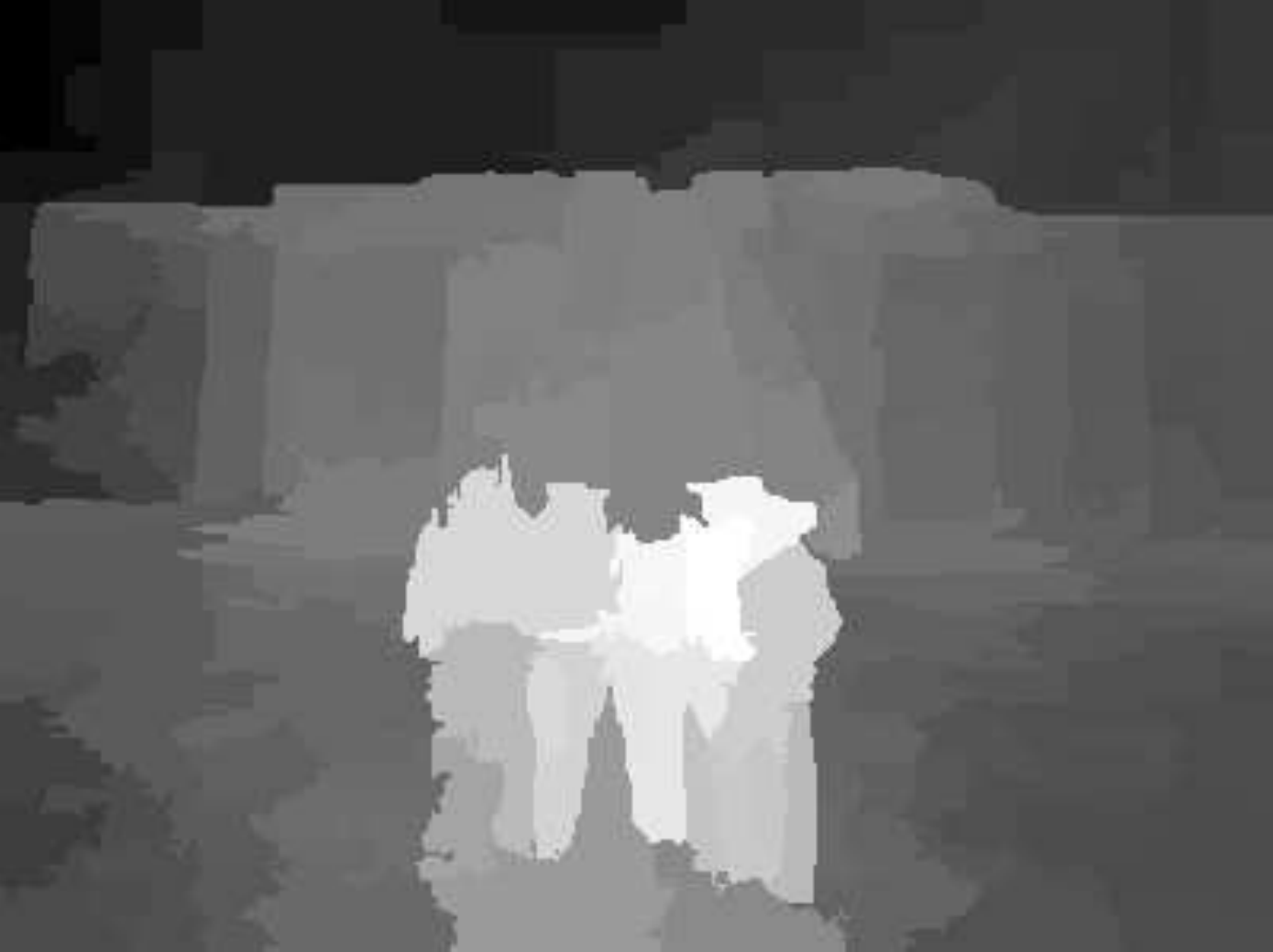}&
	\addExample{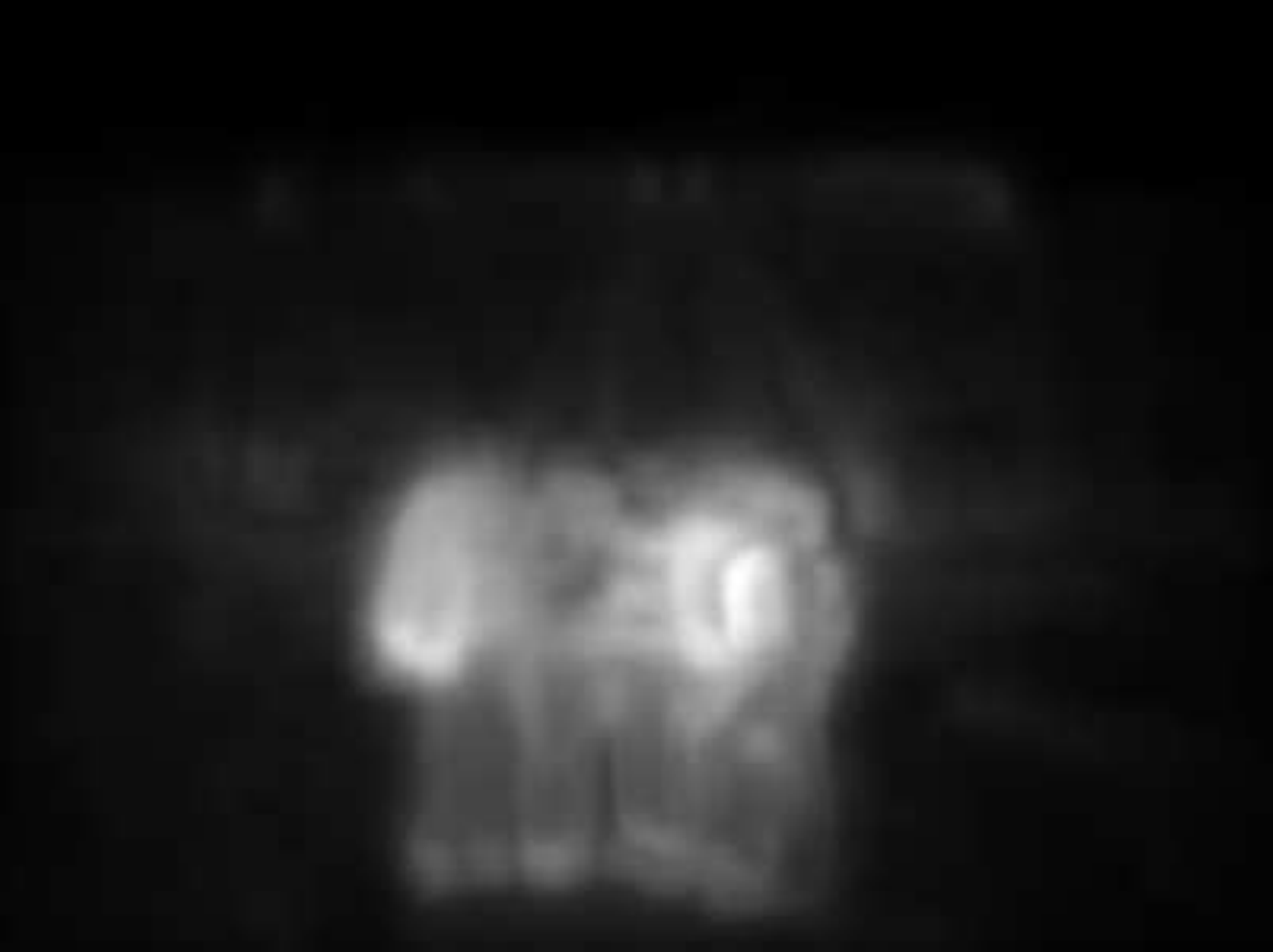}&
	\addExample{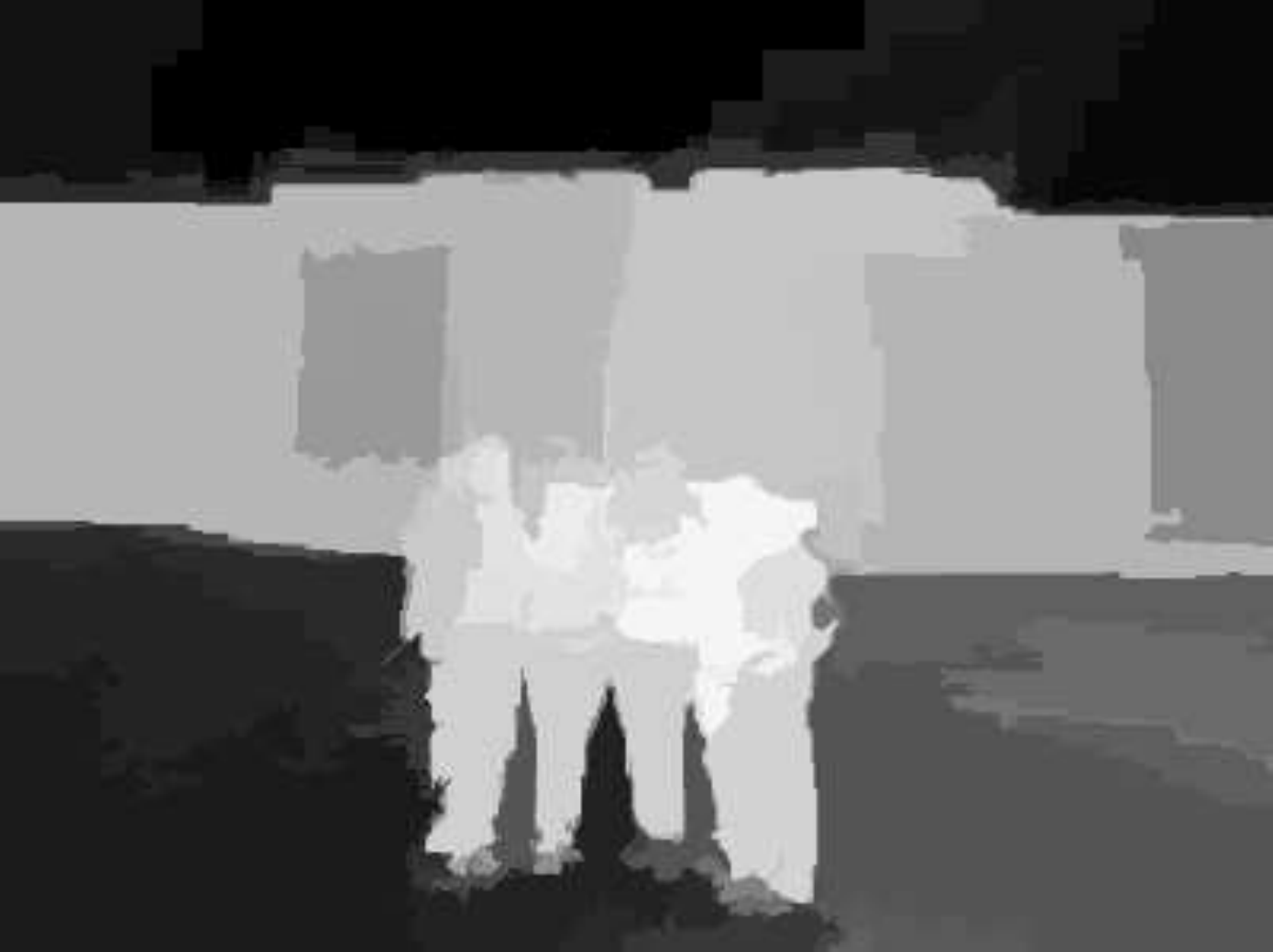}&
	\addExample{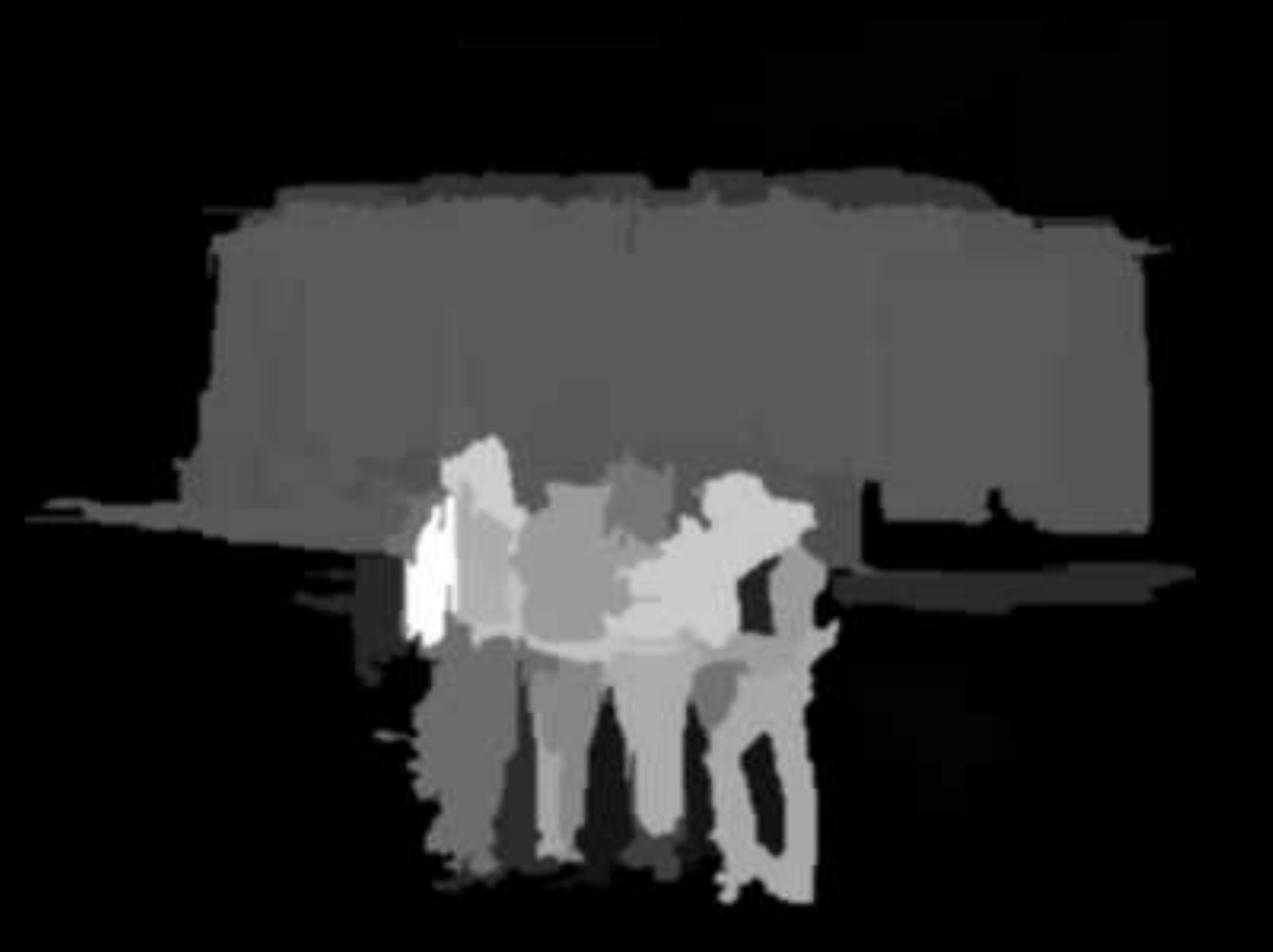}&
	\addExample{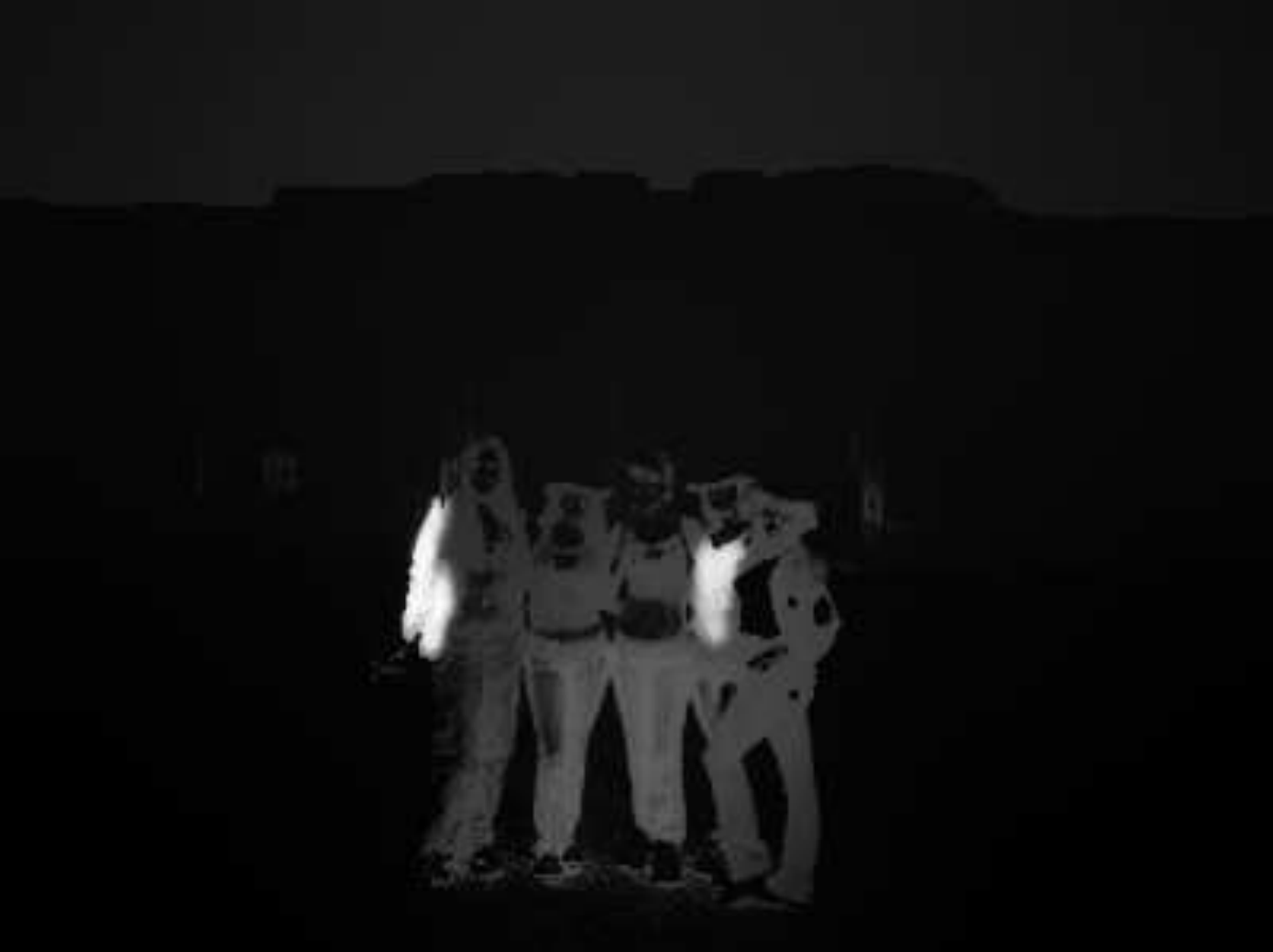}&
	\addExample{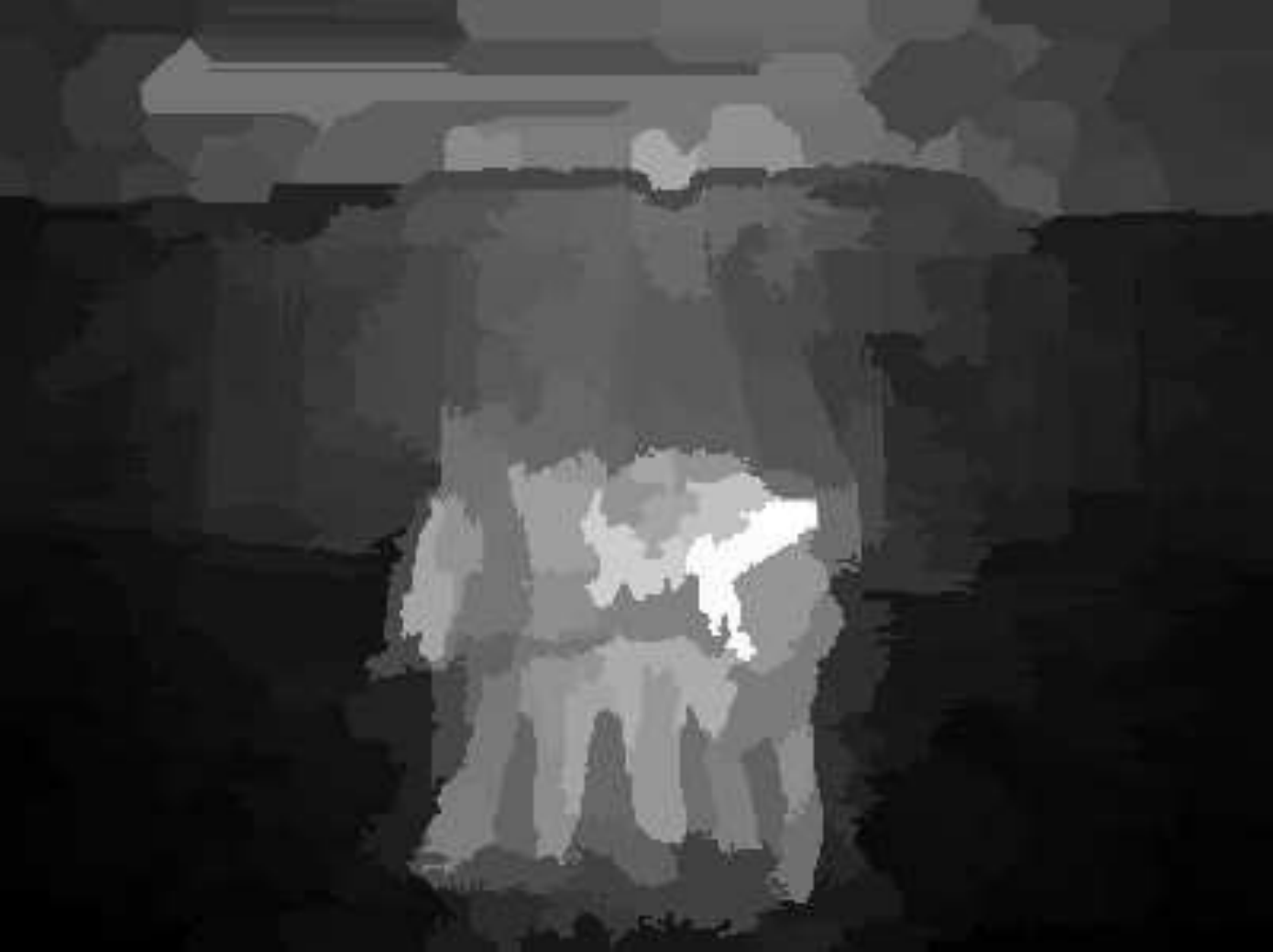}&
	\addExample{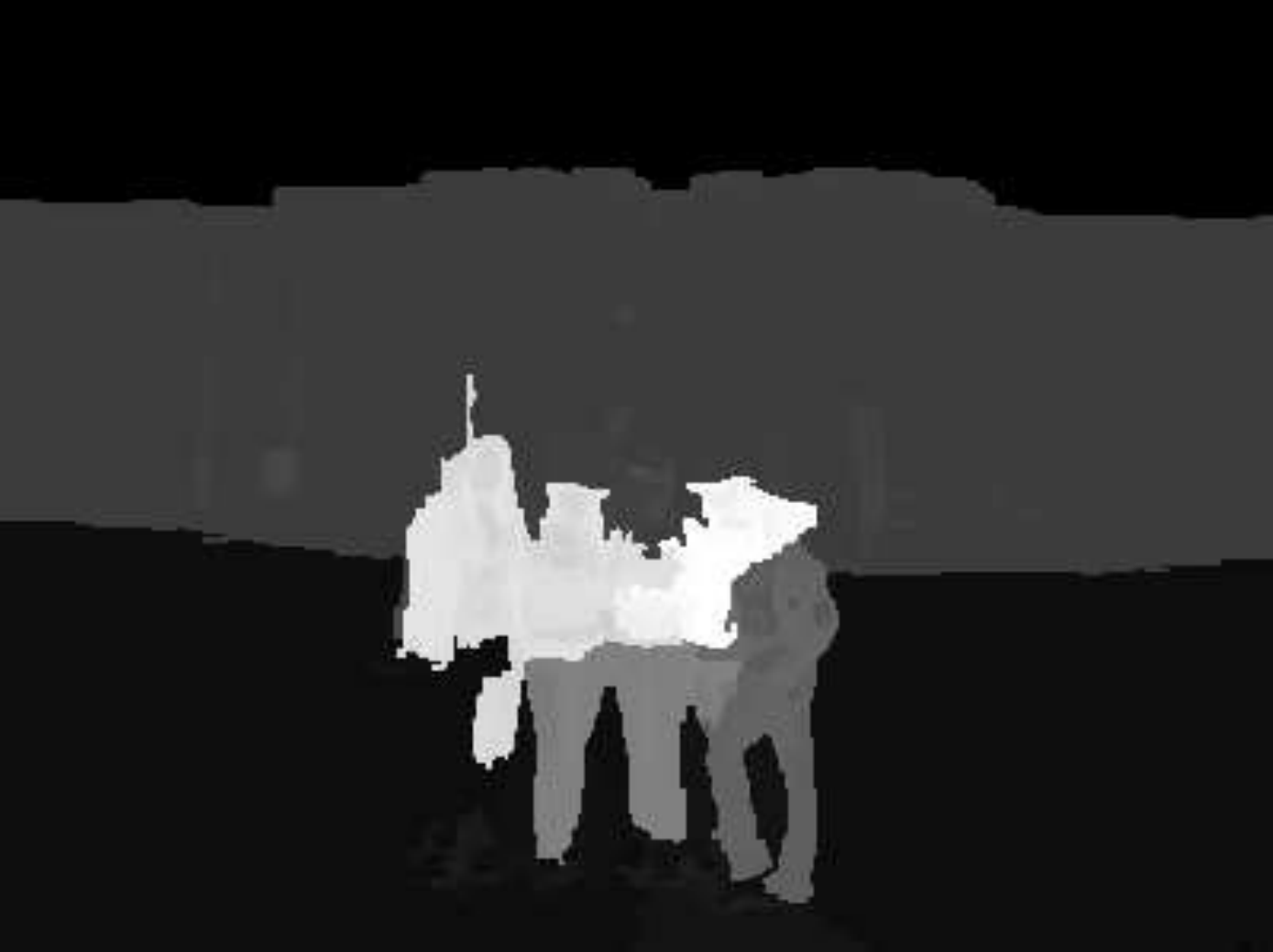}&
	\addExample{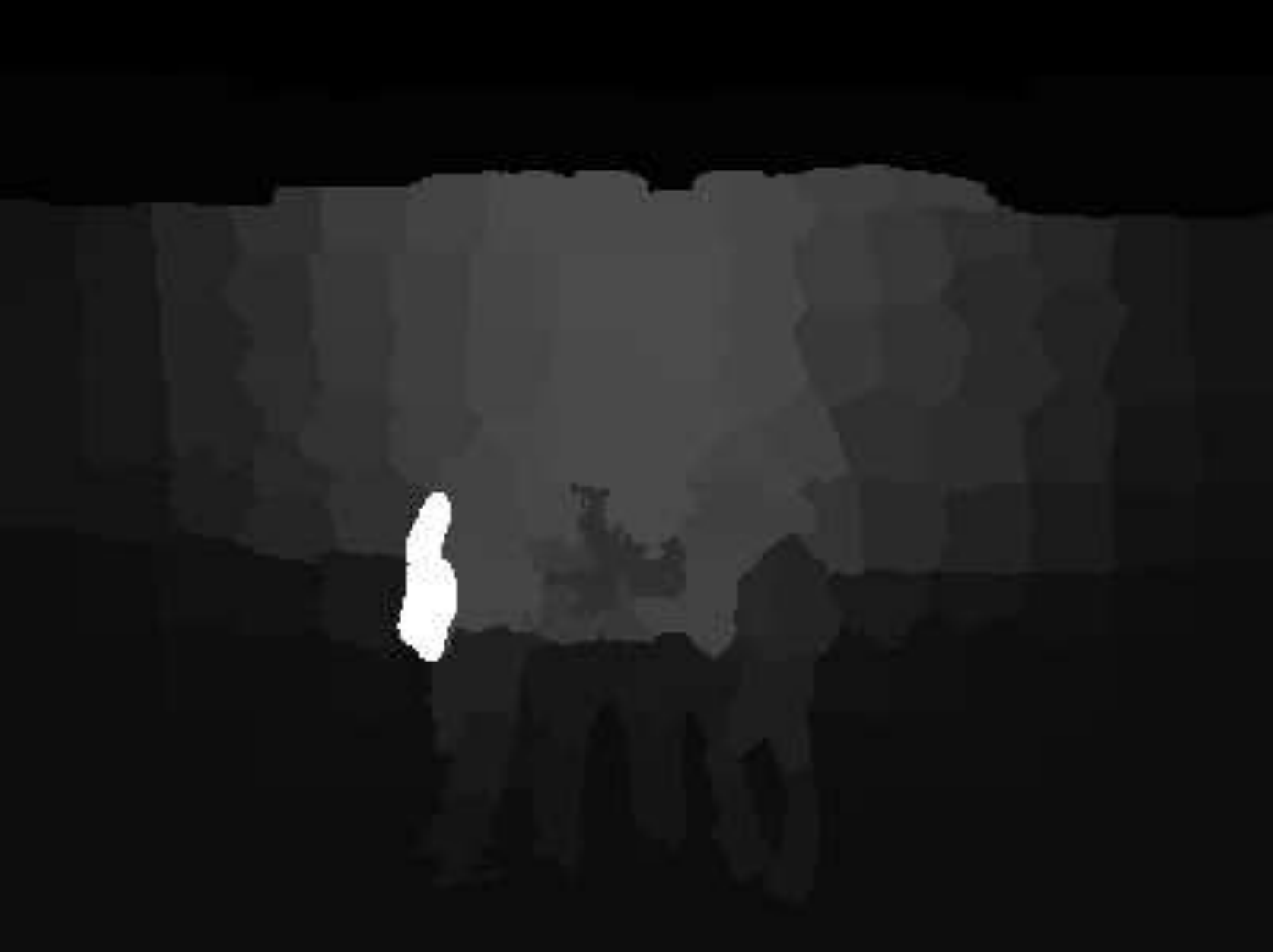}&
	\addExample{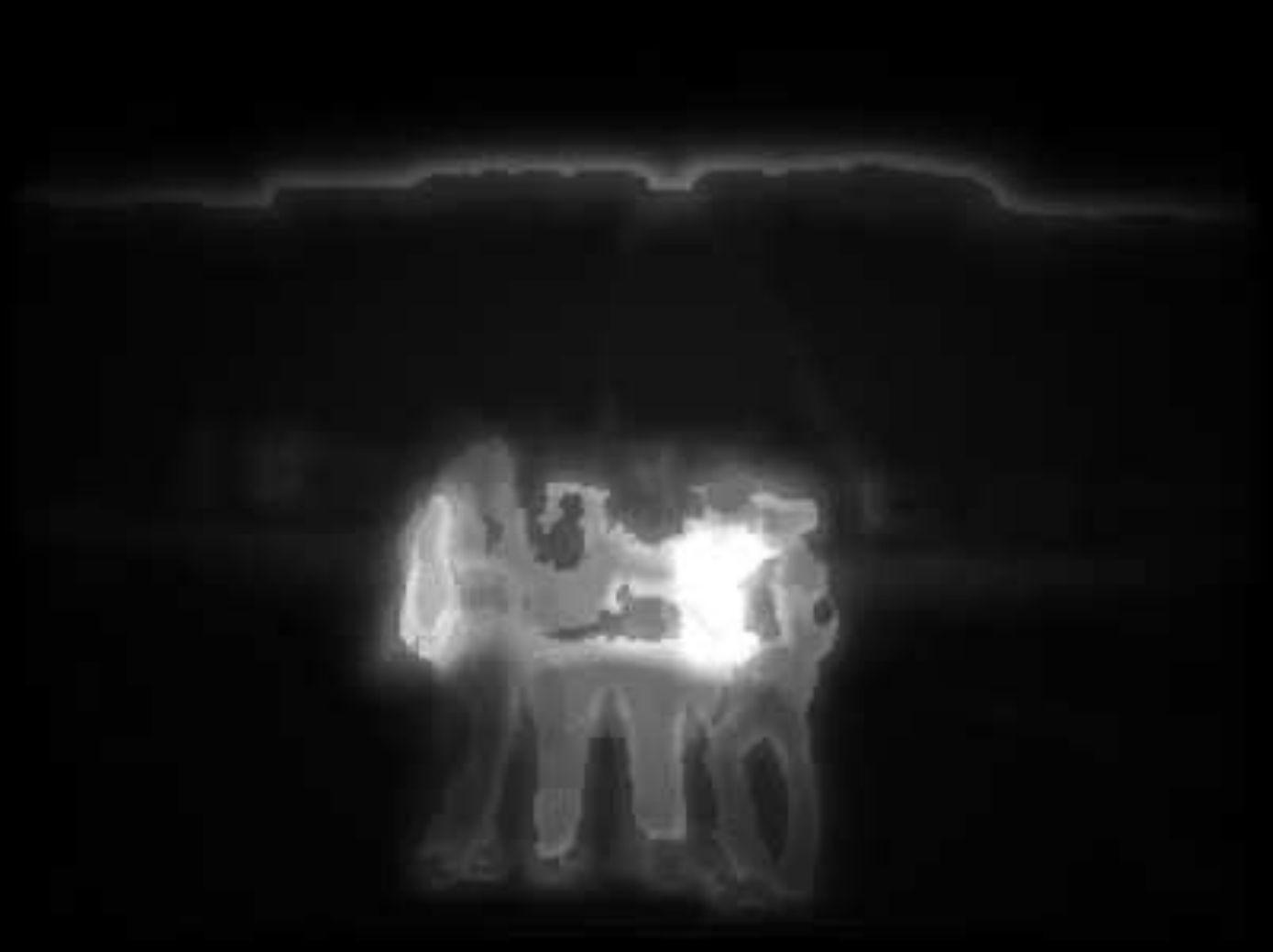}&
	\addExample{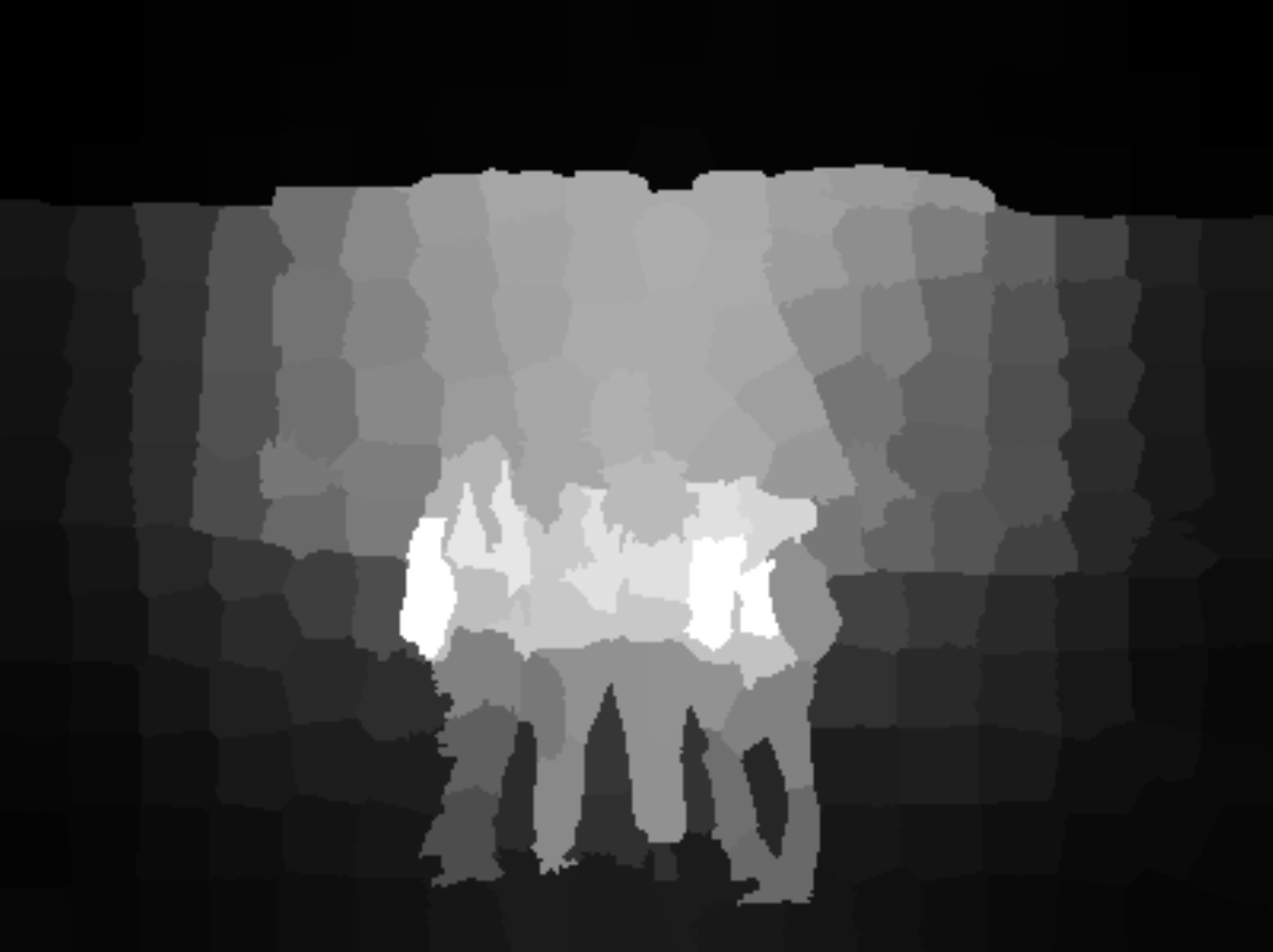}&
	\addExample{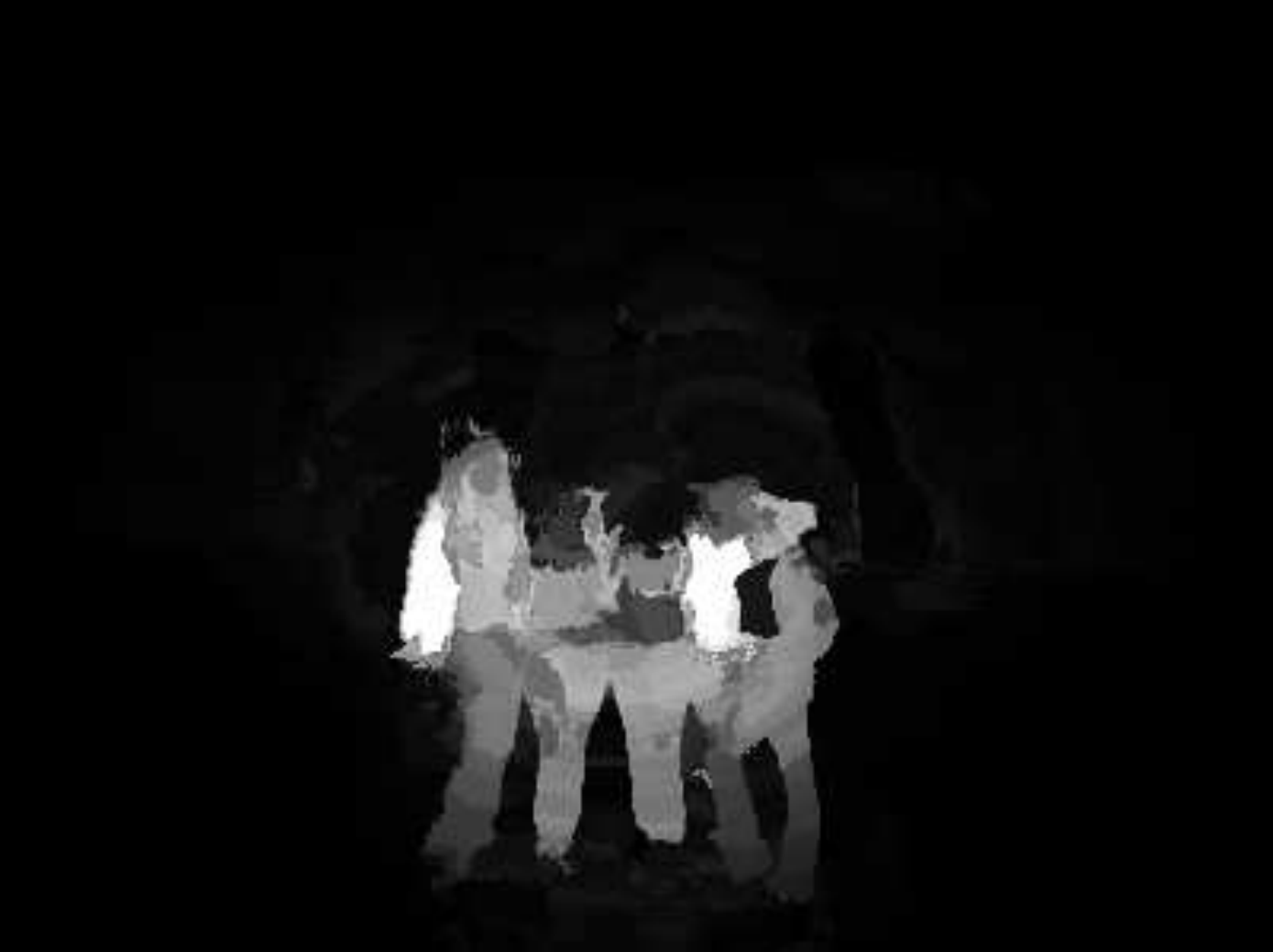}&
	\addExample{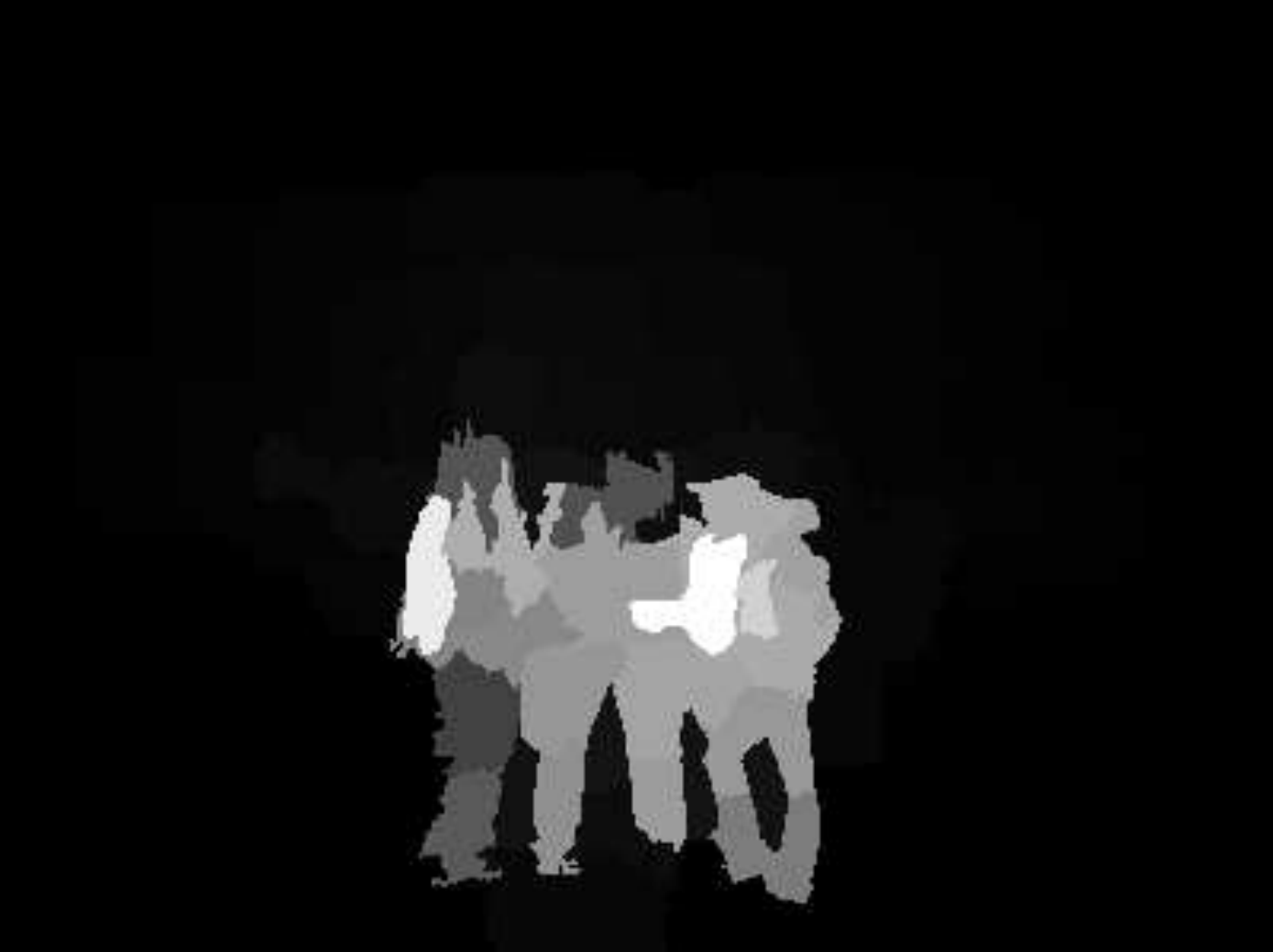}&
	\addExample{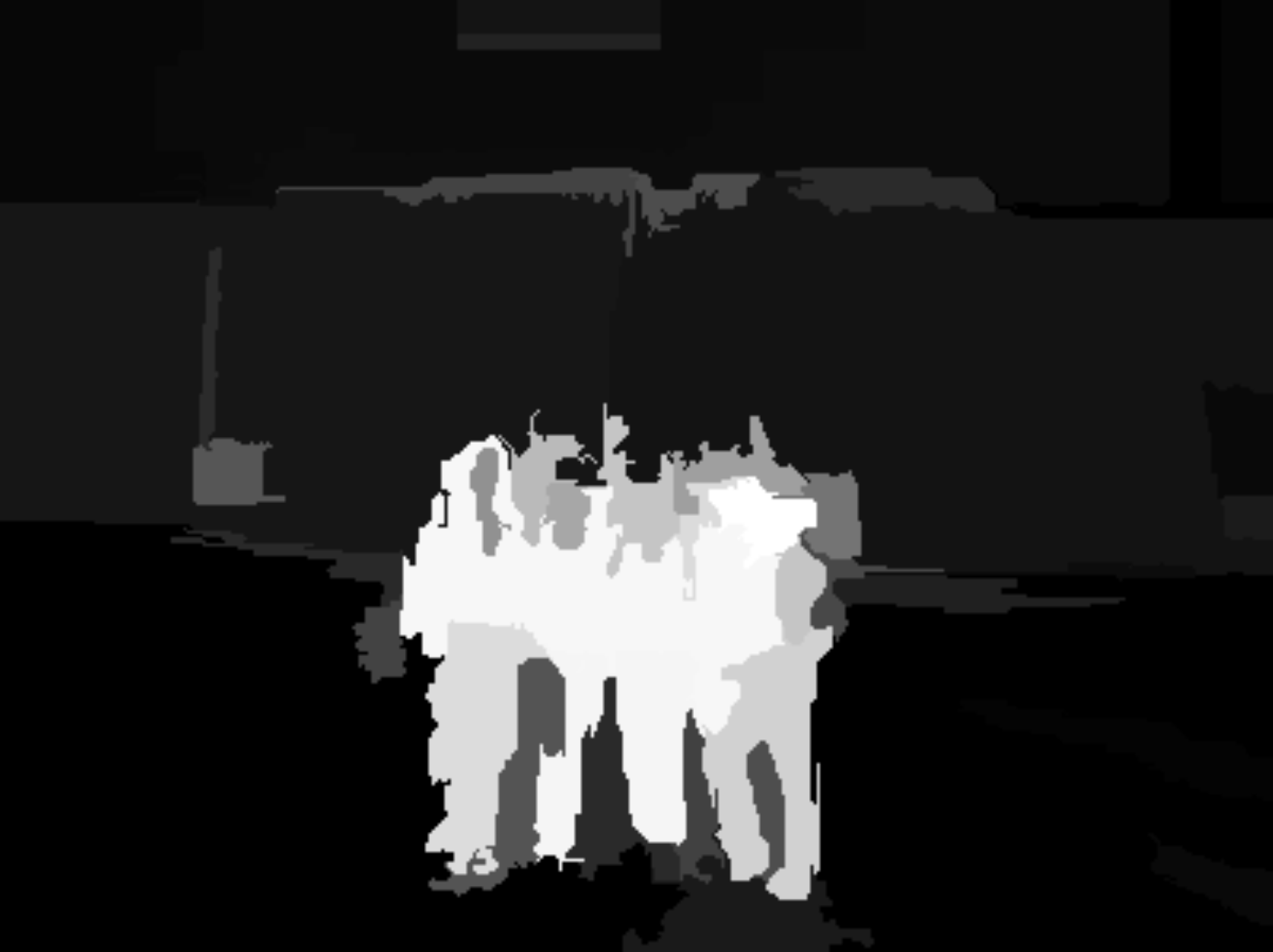}&
	\addExample{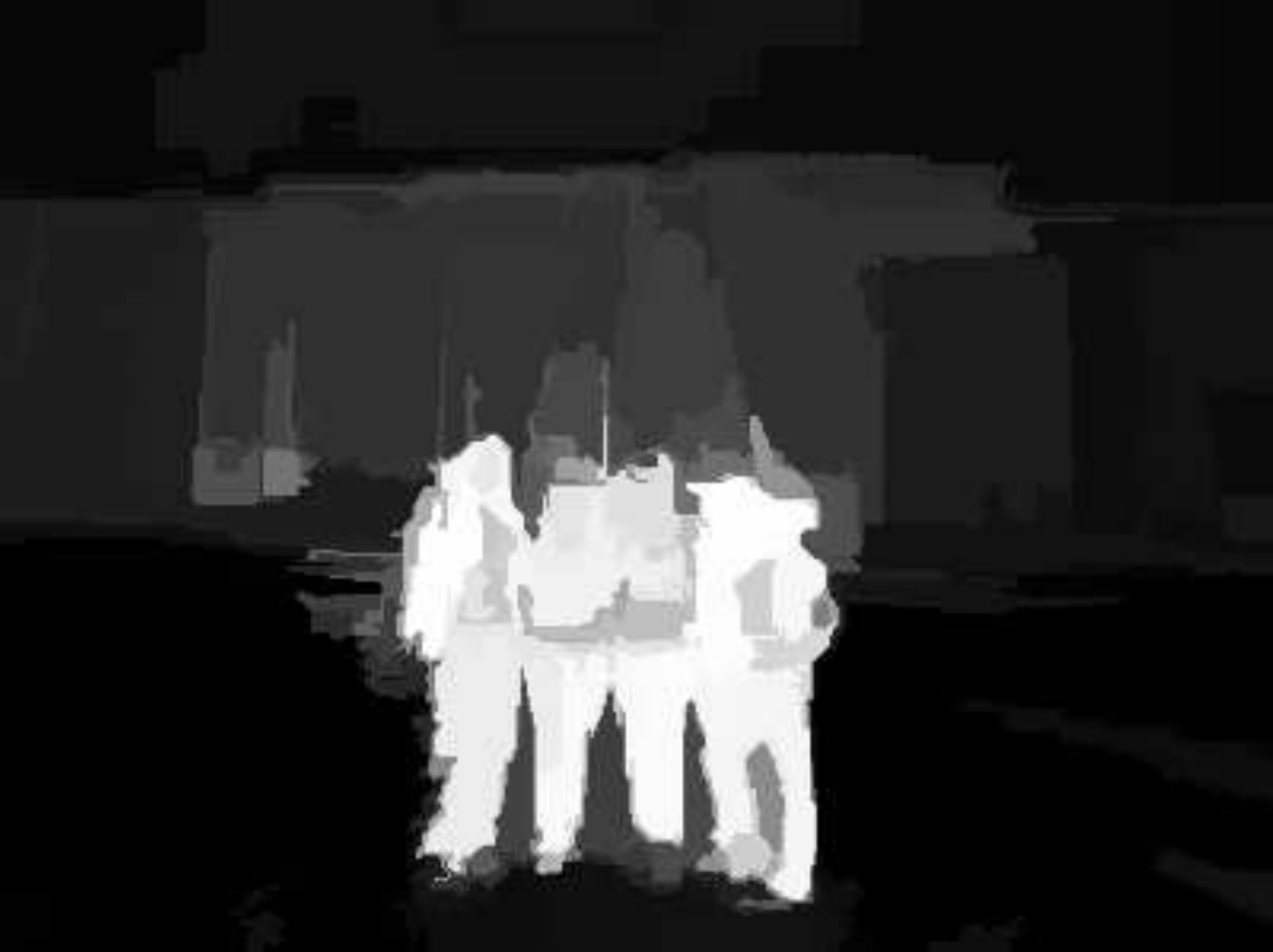}
    \\
	% \addExample{sun_adaherdkmjhhvuae}&
	% \addExample{sun_adaherdkmjhhvuae_svo}&
	% \addExample{sun_adaherdkmjhhvuae_ca}&
	% \addExample{sun_adaherdkmjhhvuae_cb}&
	% \addExample{sun_adaherdkmjhhvuae_rc}&
	% \addExample{sun_adaherdkmjhhvuae_sf}&
	% \addExample{sun_adaherdkmjhhvuae_lrk}&
	% \addExample{sun_adaherdkmjhhvuae_hs}&
	% \addExample{sun_adaherdkmjhhvuae_gmr}&
	% \addExample{sun_adaherdkmjhhvuae_pca}&
	% \addExample{sun_adaherdkmjhhvuae_mc}&
	% \addExample{sun_adaherdkmjhhvuae_dsr}&
	% \addExample{sun_adaherdkmjhhvuae_rbd}&
	% \addExample{sun_adaherdkmjhhvuae_drfi}\\
	% \addExample{0026}&
	% \addExample{0026_svo}&
	% \addExample{0026_ca}&
	% \addExample{0026_cb}&
	% \addExample{0026_rc}&
	% \addExample{0026_sf}&
	% \addExample{0026_lrk}&
	% \addExample{0026_hs}&
	% \addExample{0026_gmr}&
	% \addExample{0026_pca}&
	% \addExample{0026_mc}&
	% \addExample{0026_dsr}&
	% \addExample{0026_rbd}&
	% \addExample{0026_drfis}&
	% \addExample{0026_drfi}
 %    \\
	\addExample{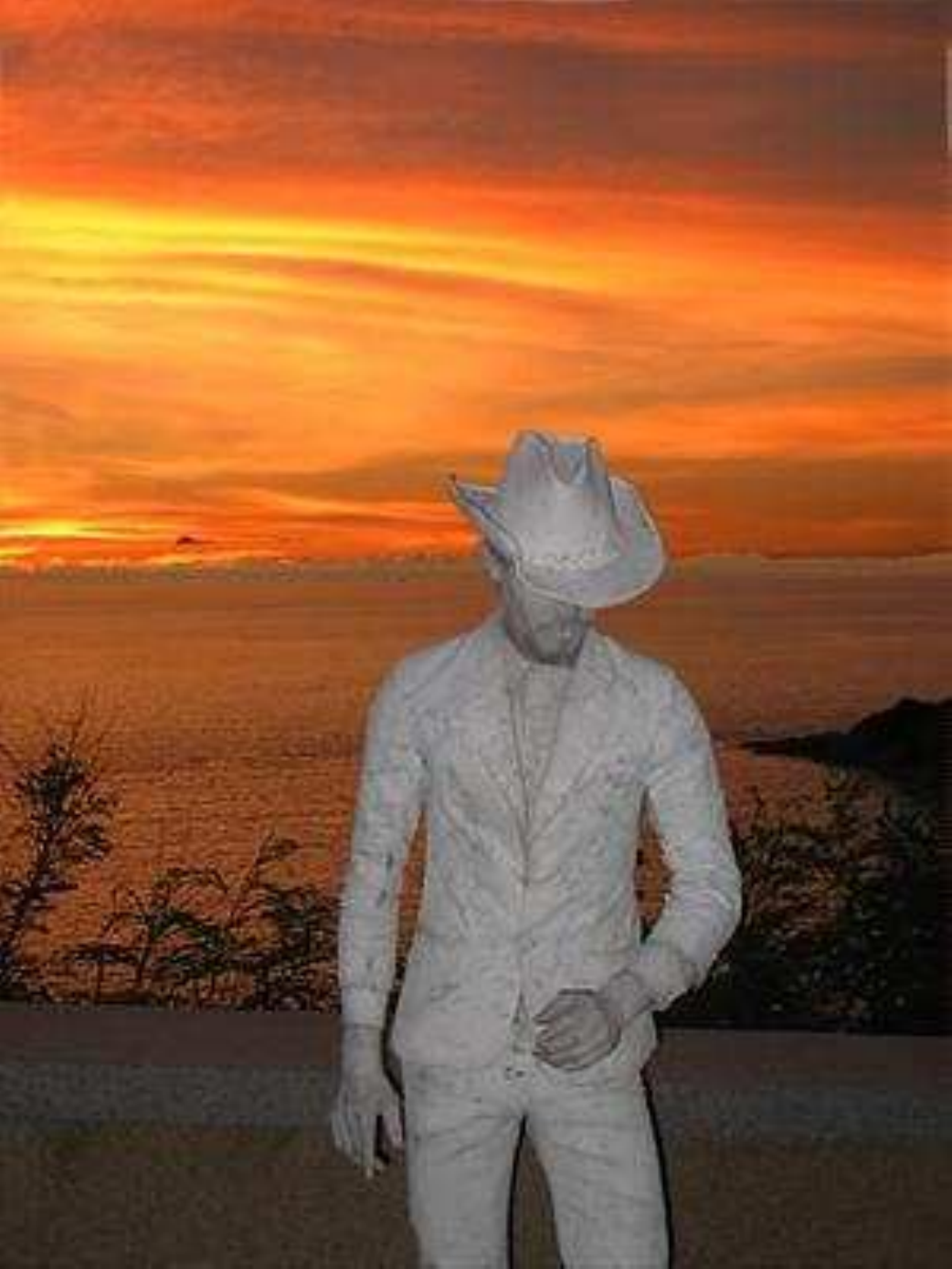}&
	\addExample{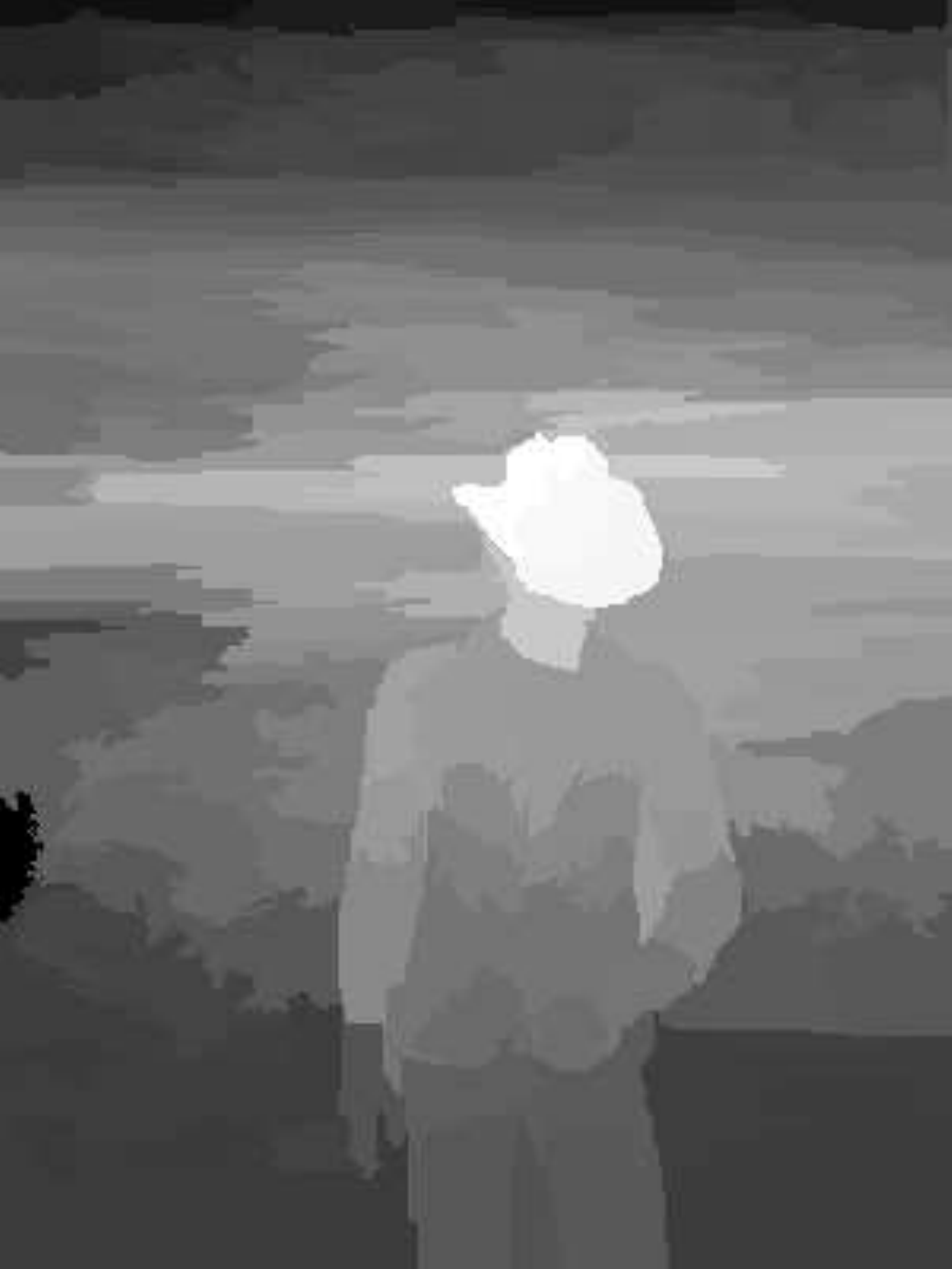}&
	\addExample{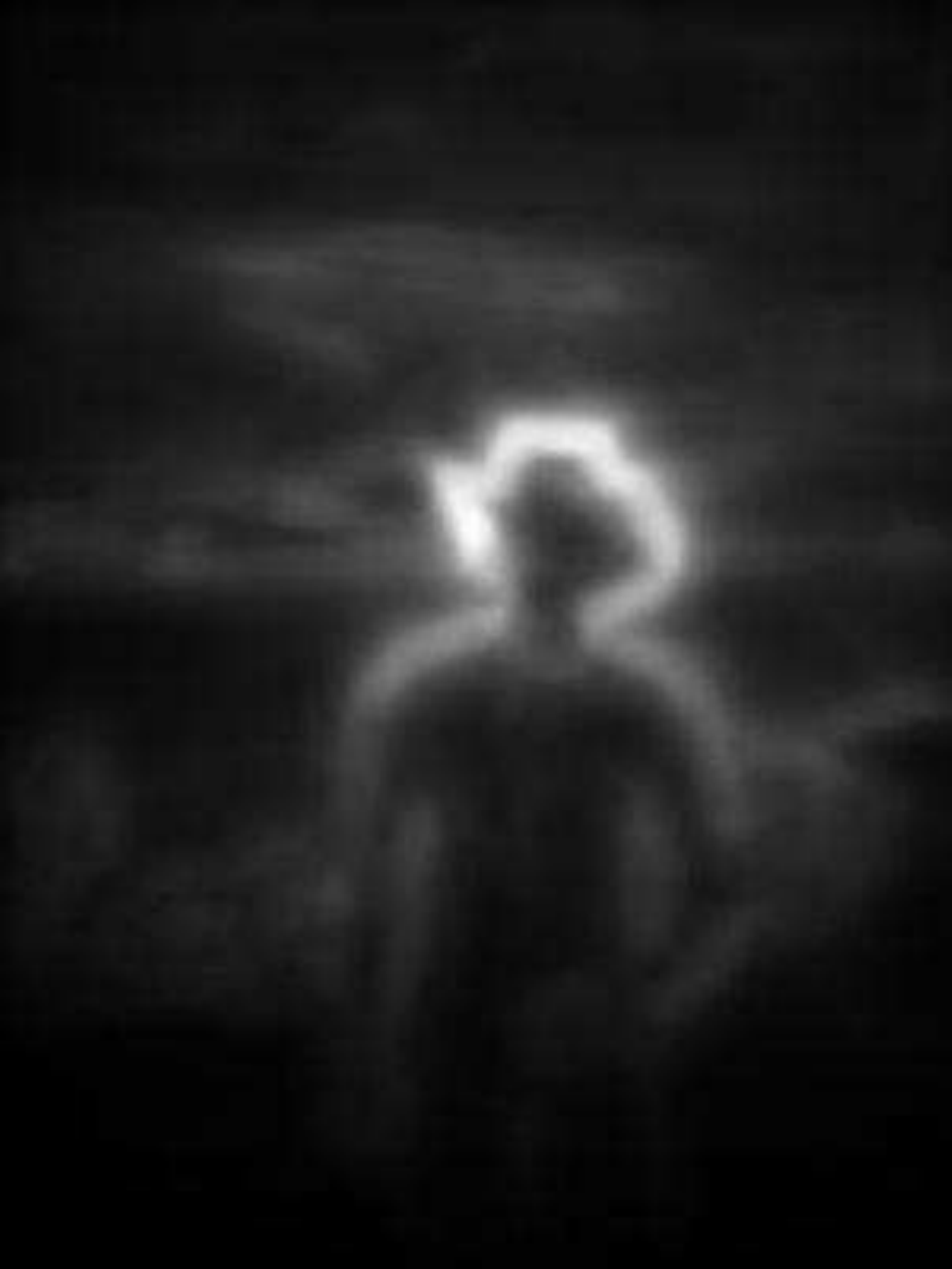}&
	\addExample{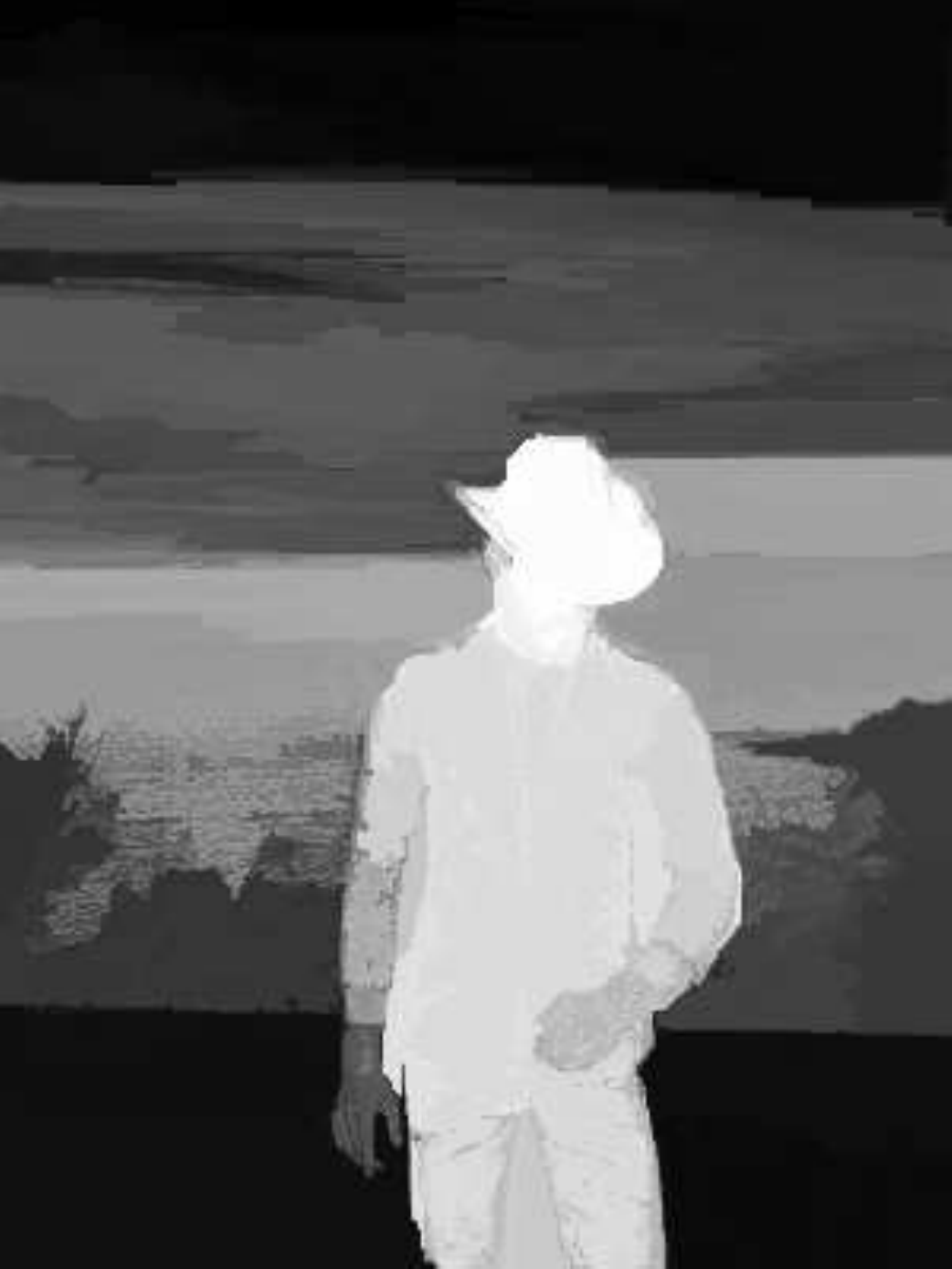}&
	\addExample{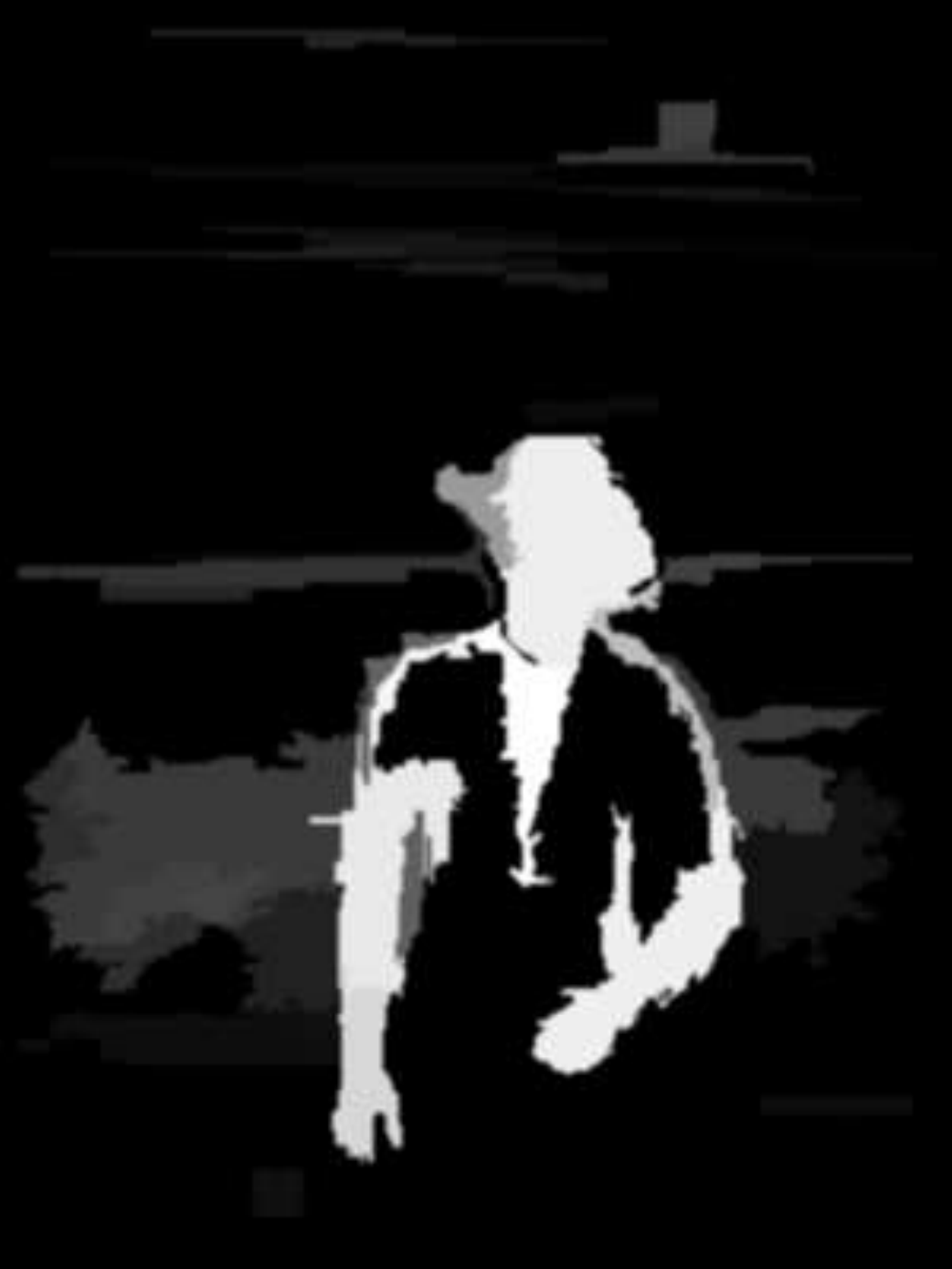}&
	\addExample{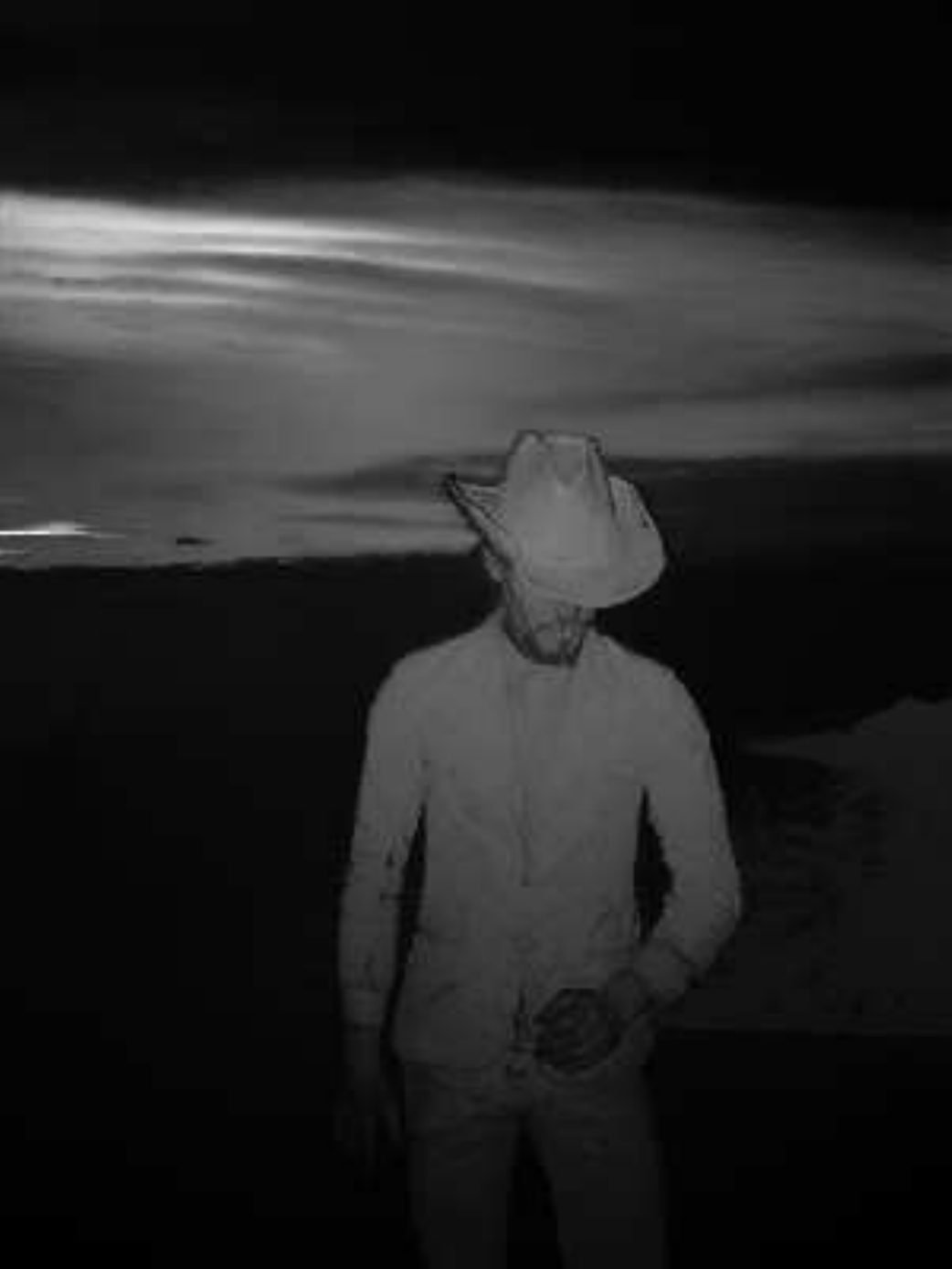}&
	\addExample{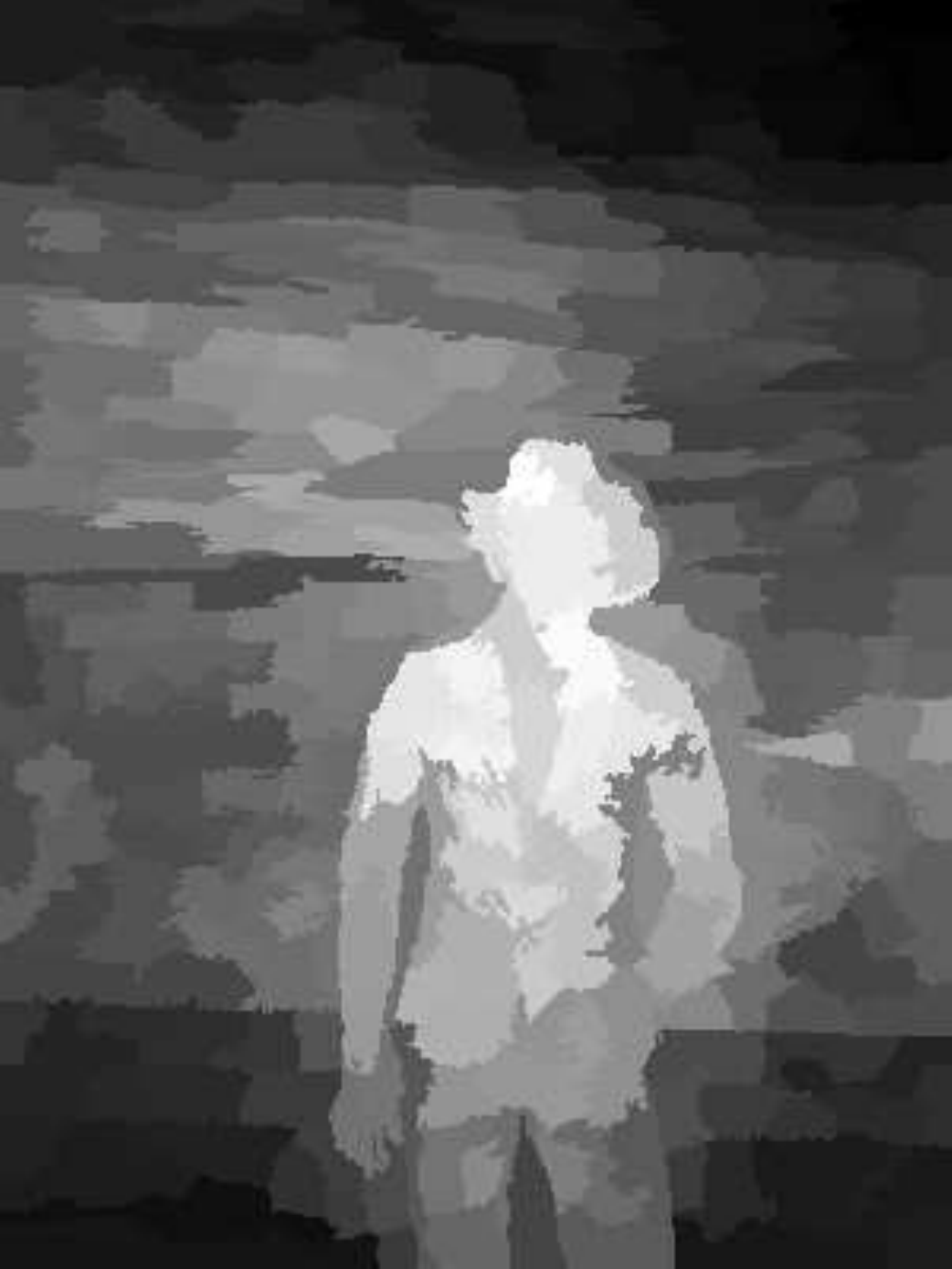}&
	\addExample{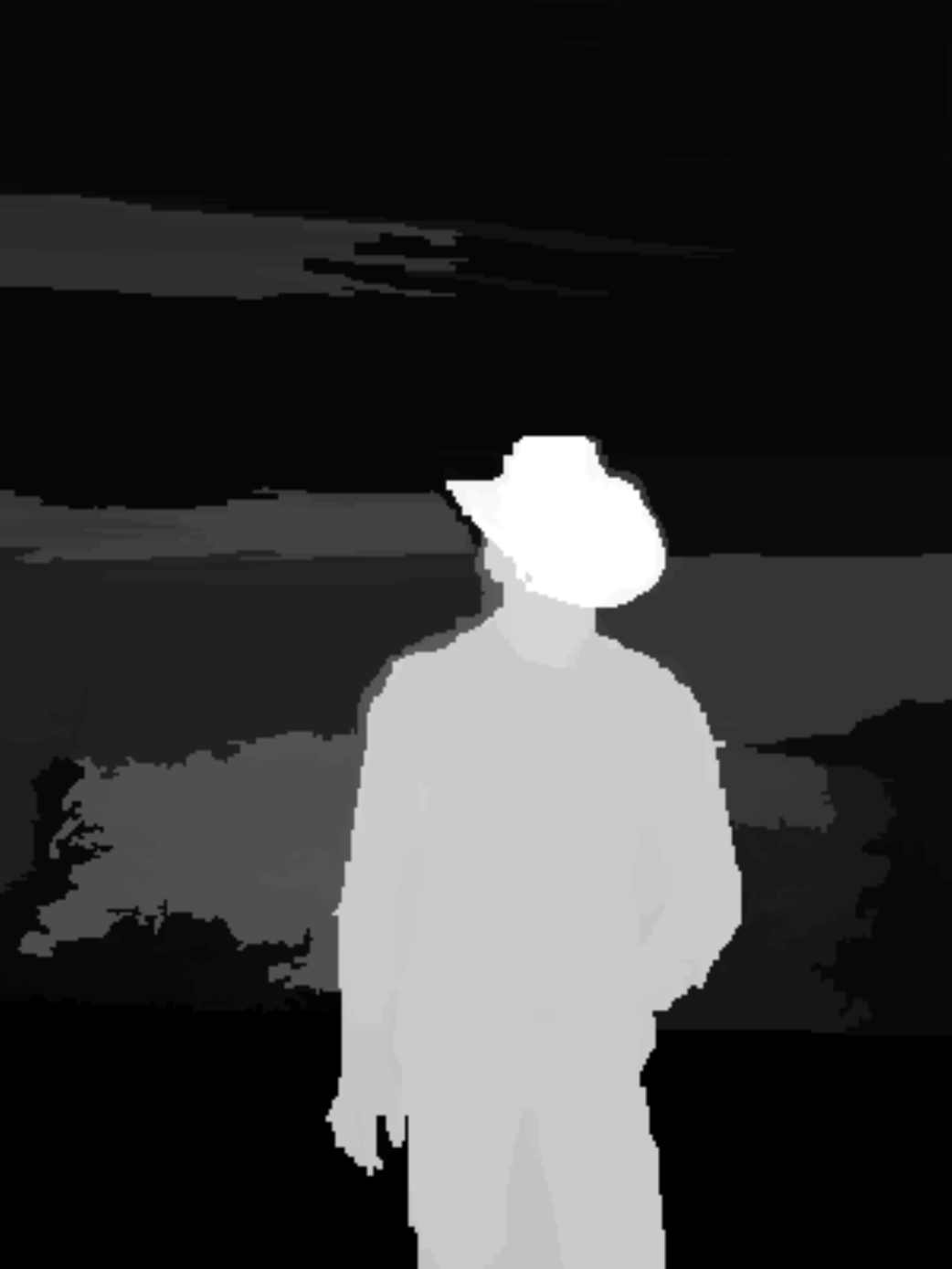}&
	\addExample{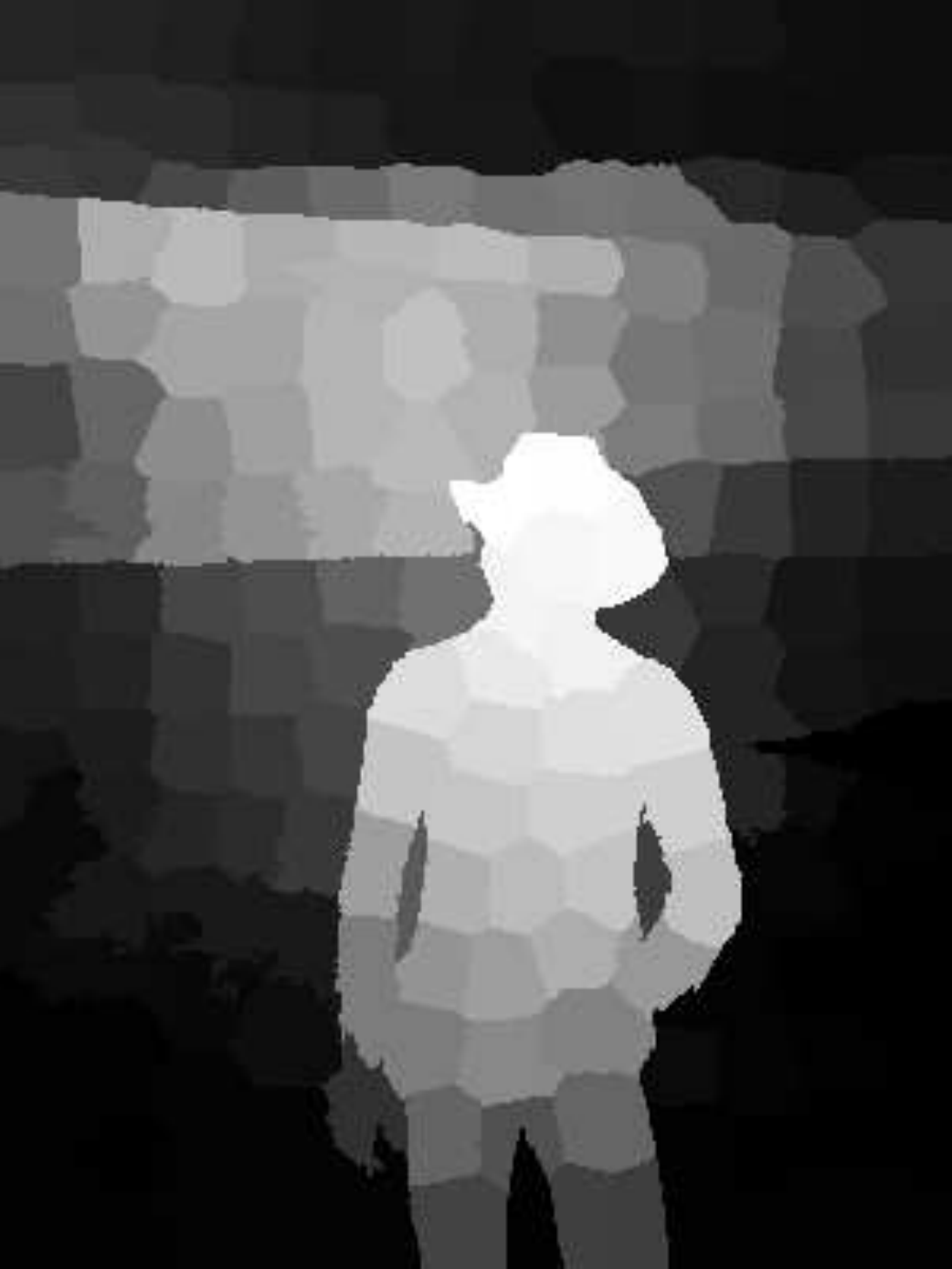}&
	\addExample{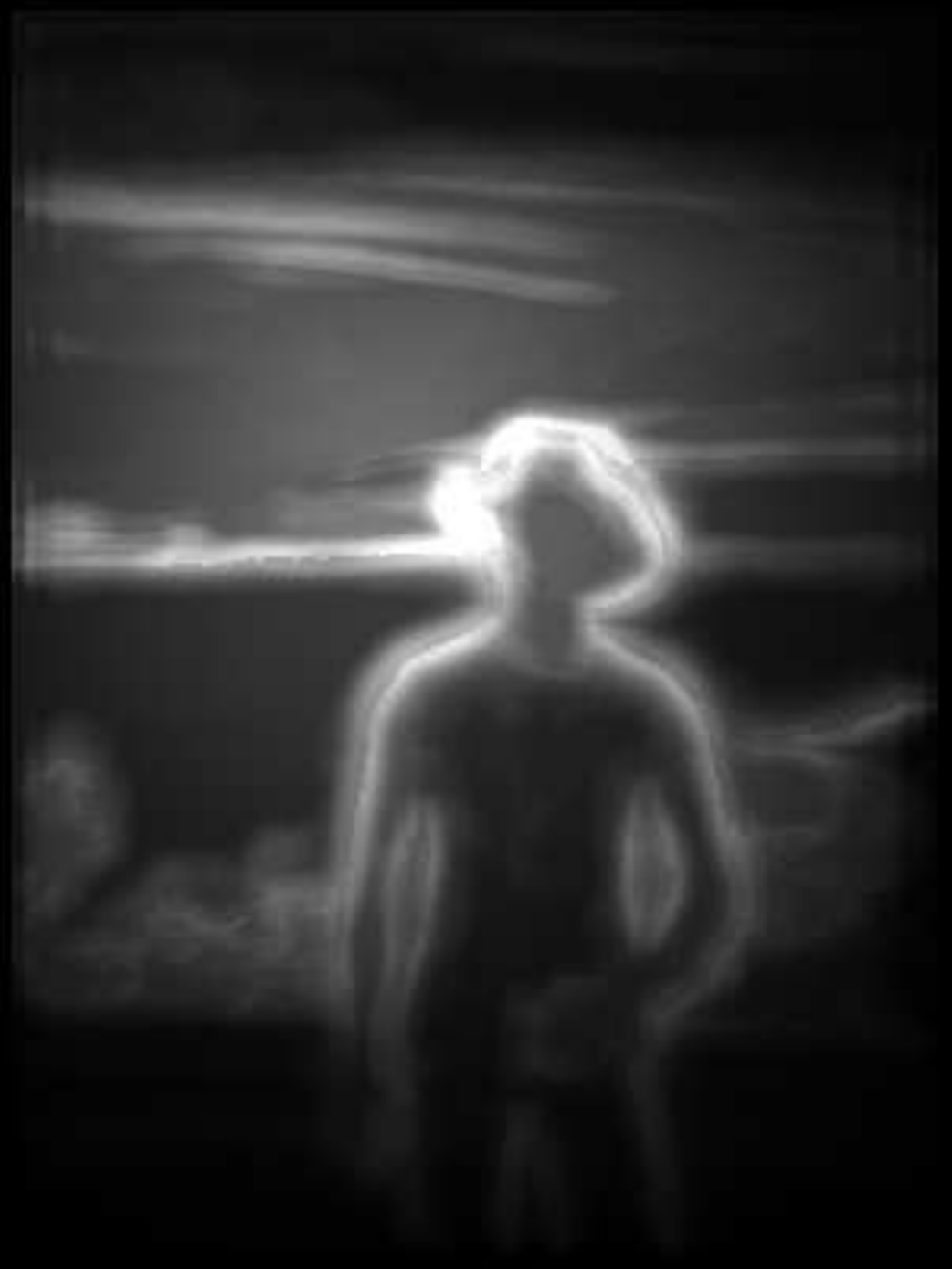}&
	\addExample{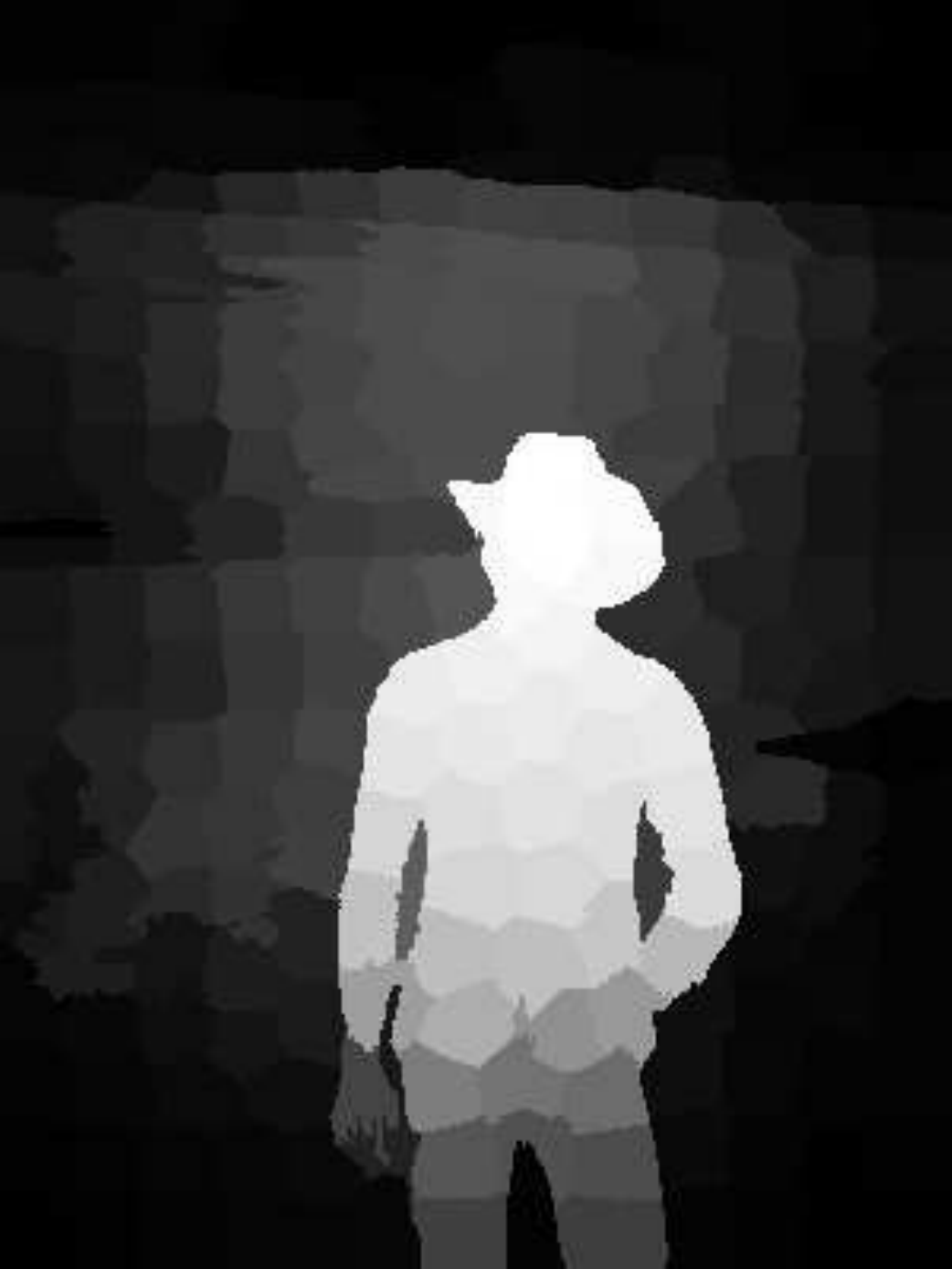}&
	\addExample{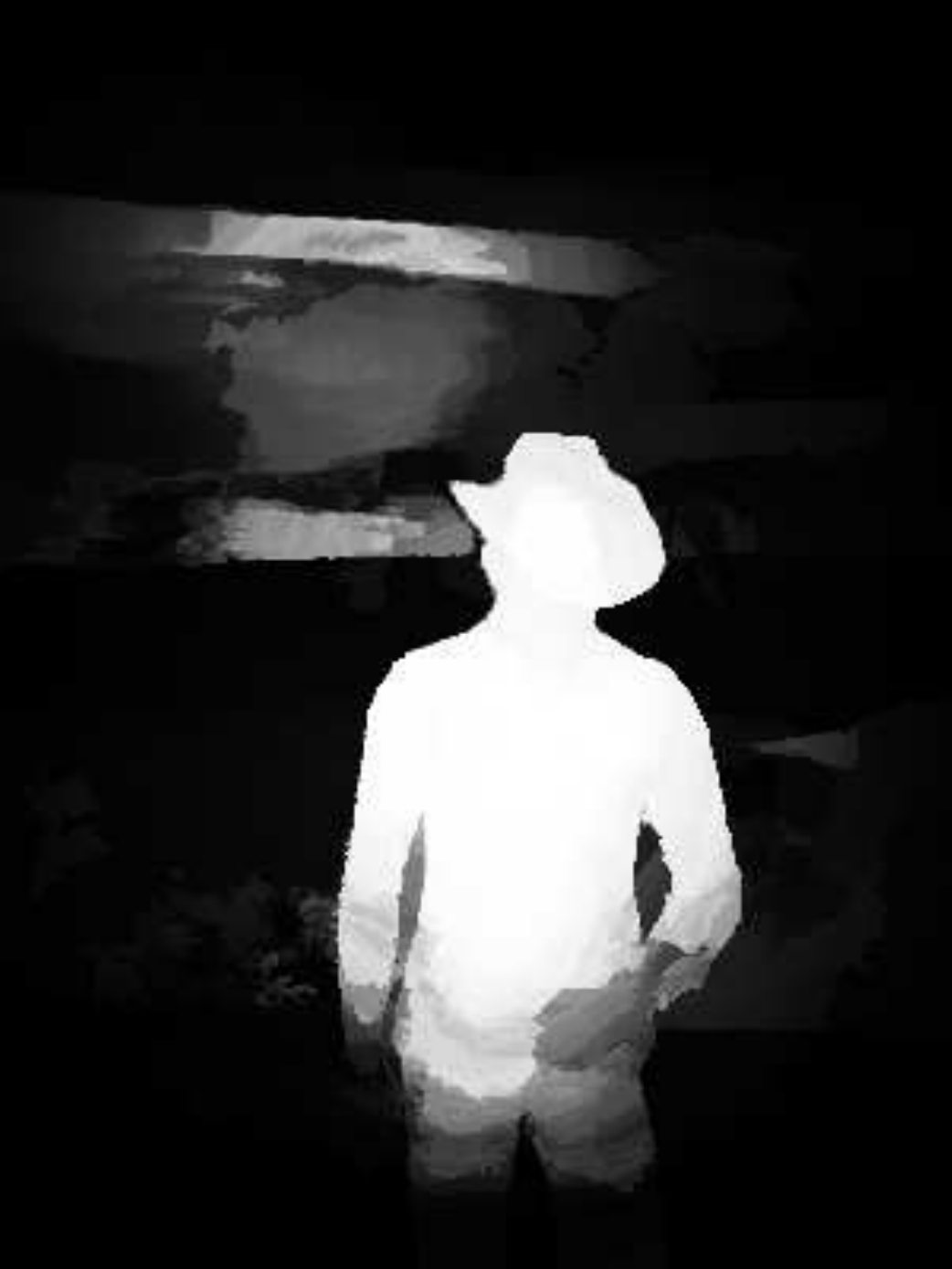}&
	\addExample{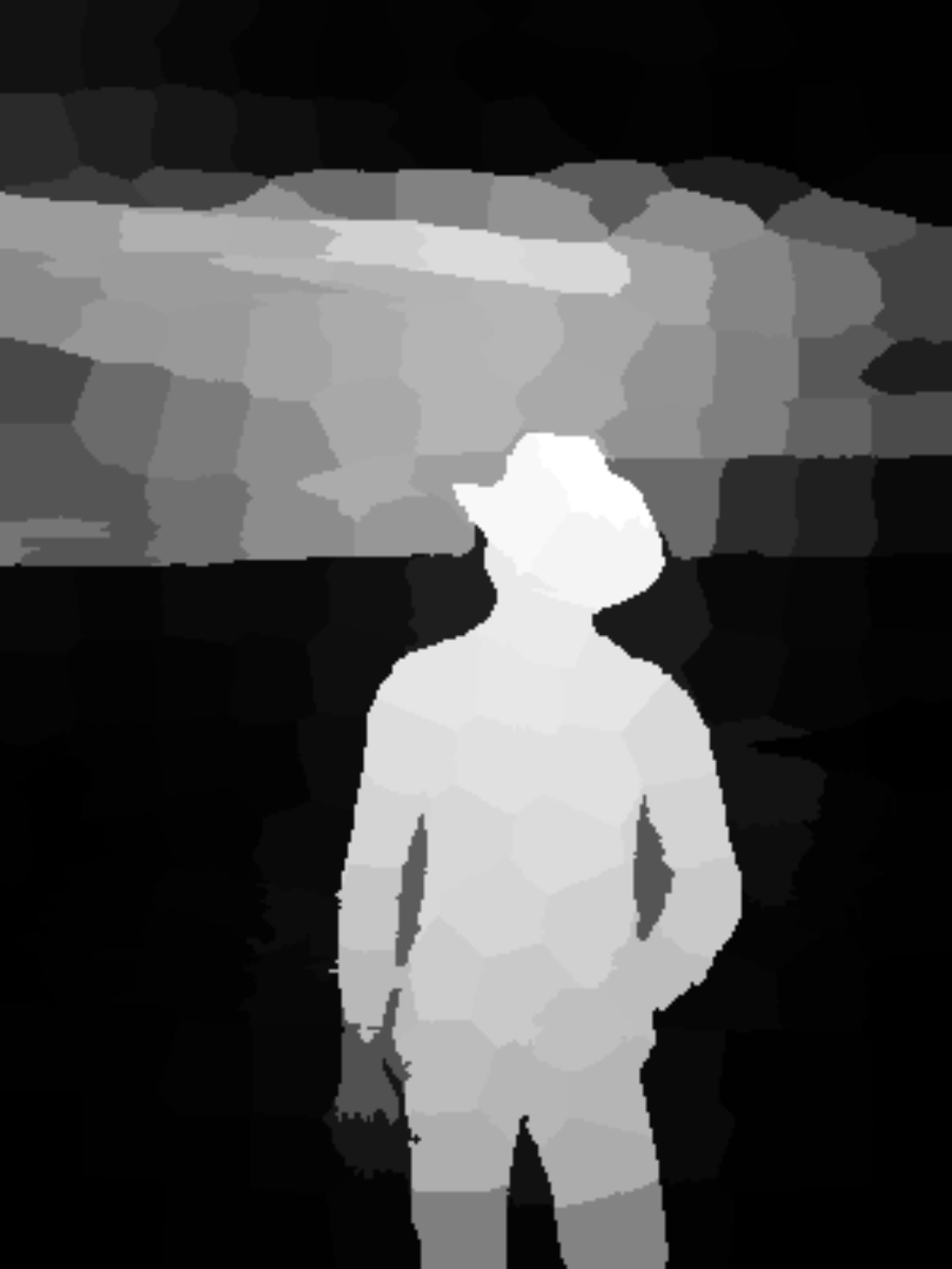}&
	\addExample{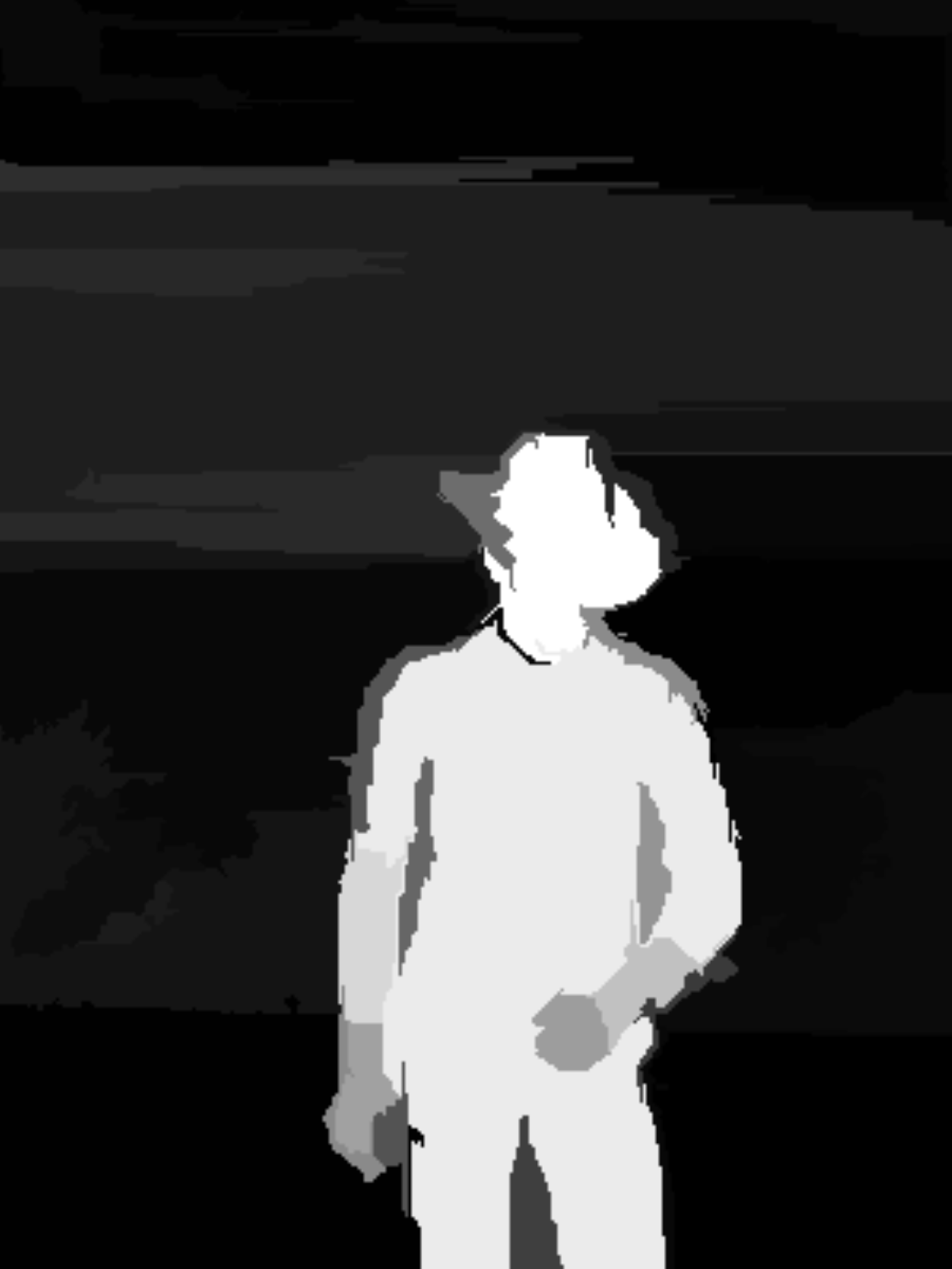}&
	\addExample{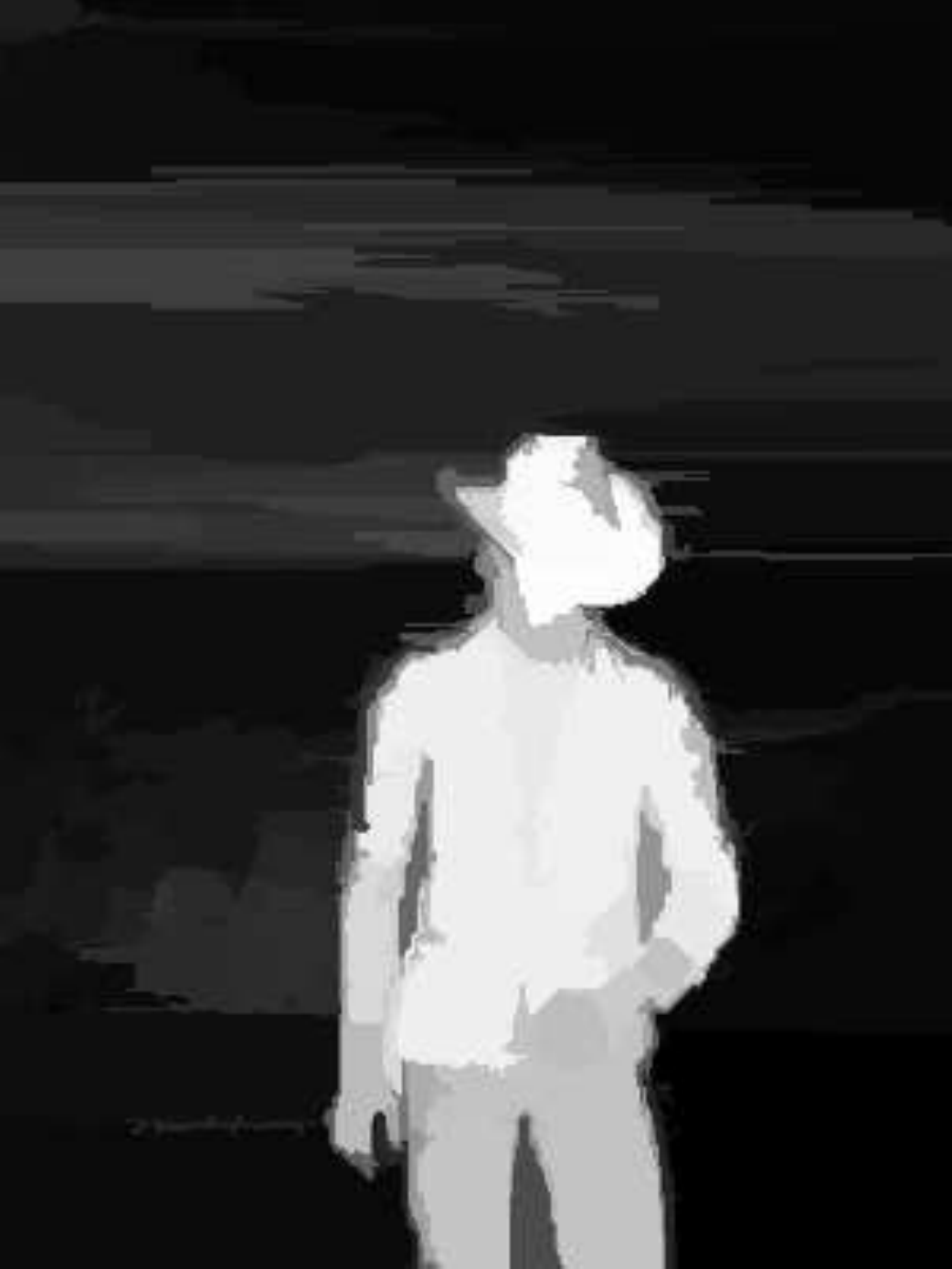}
    \\
	\addExample{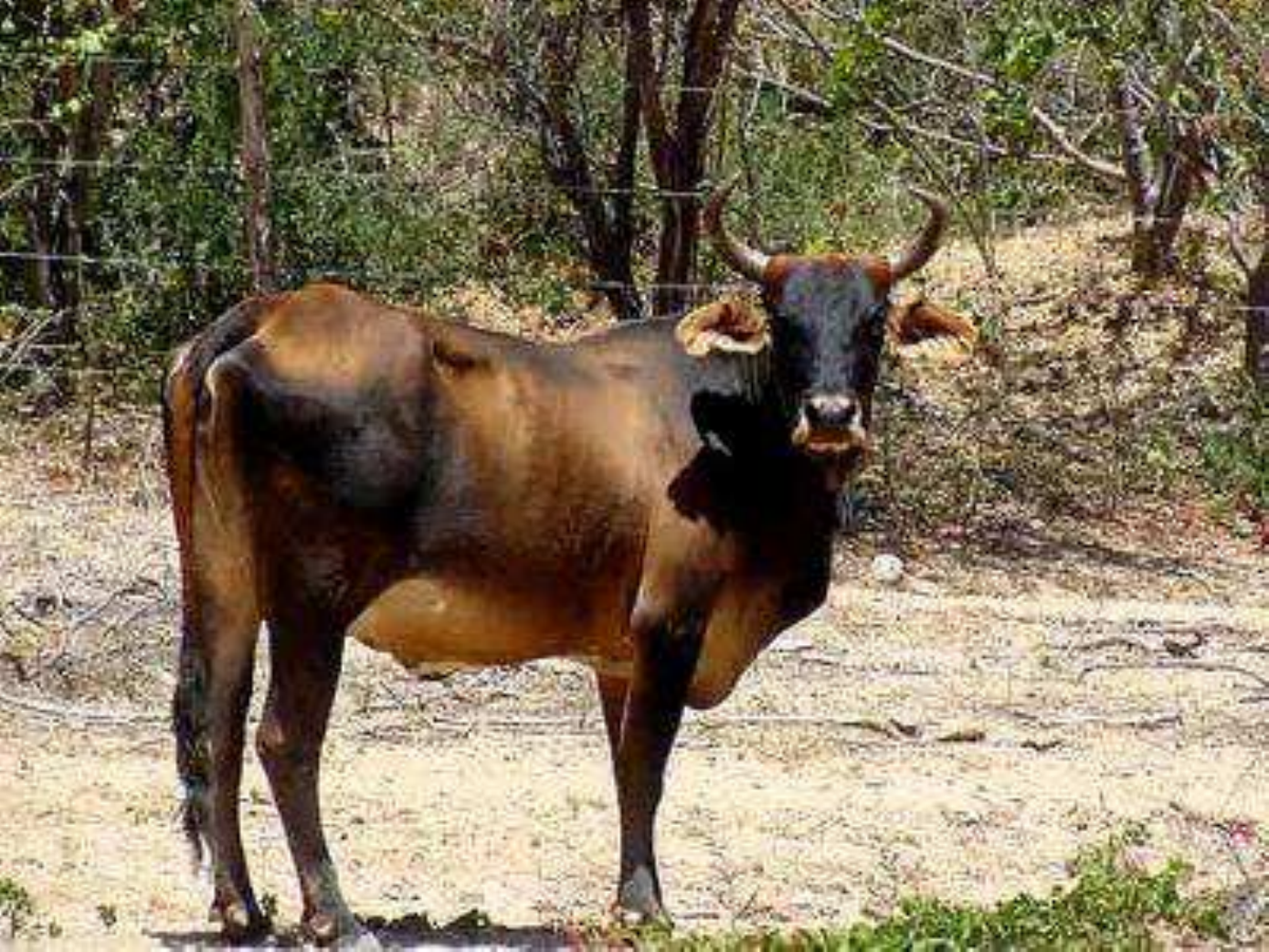}&
	\addExample{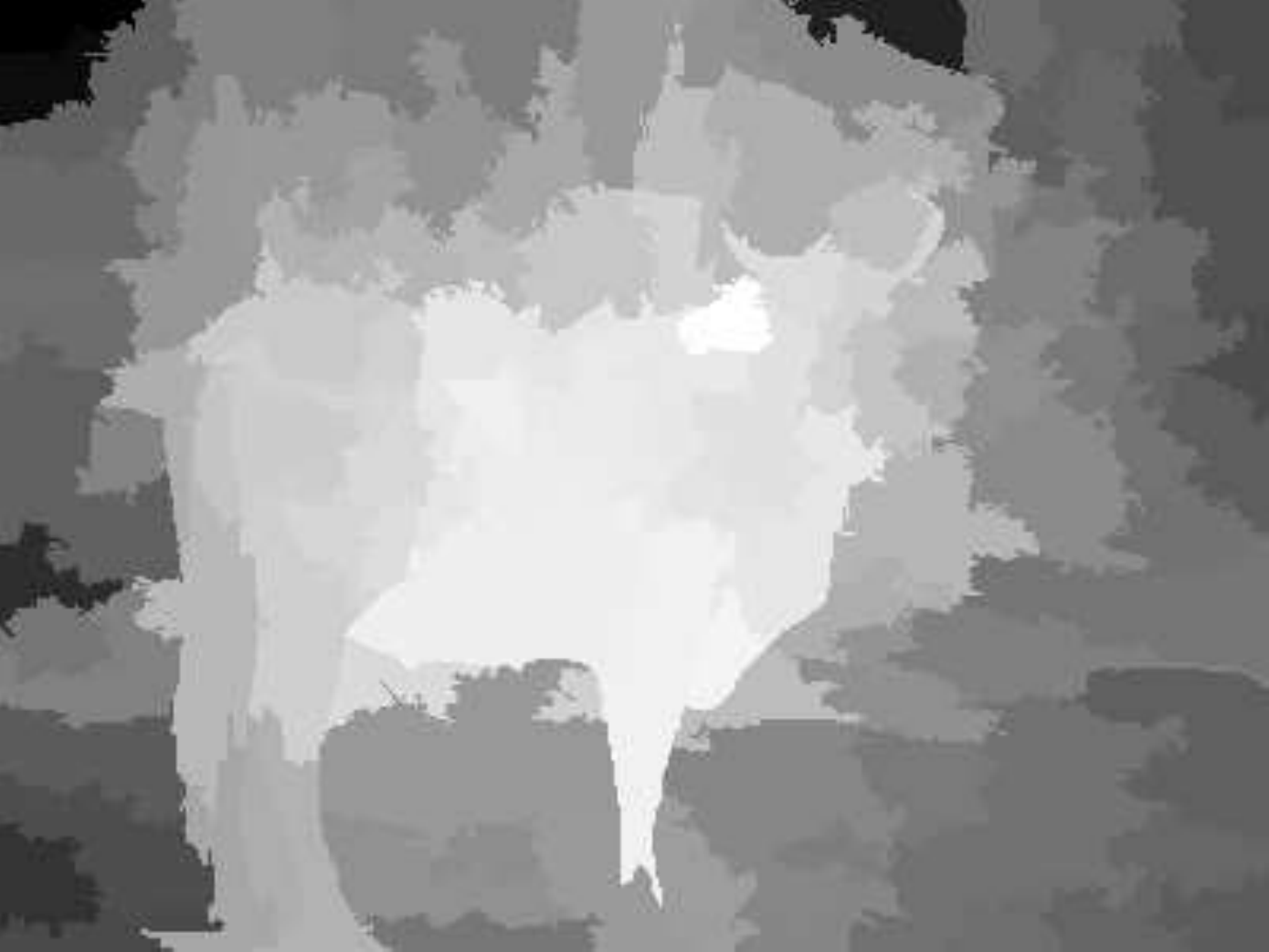}&
	\addExample{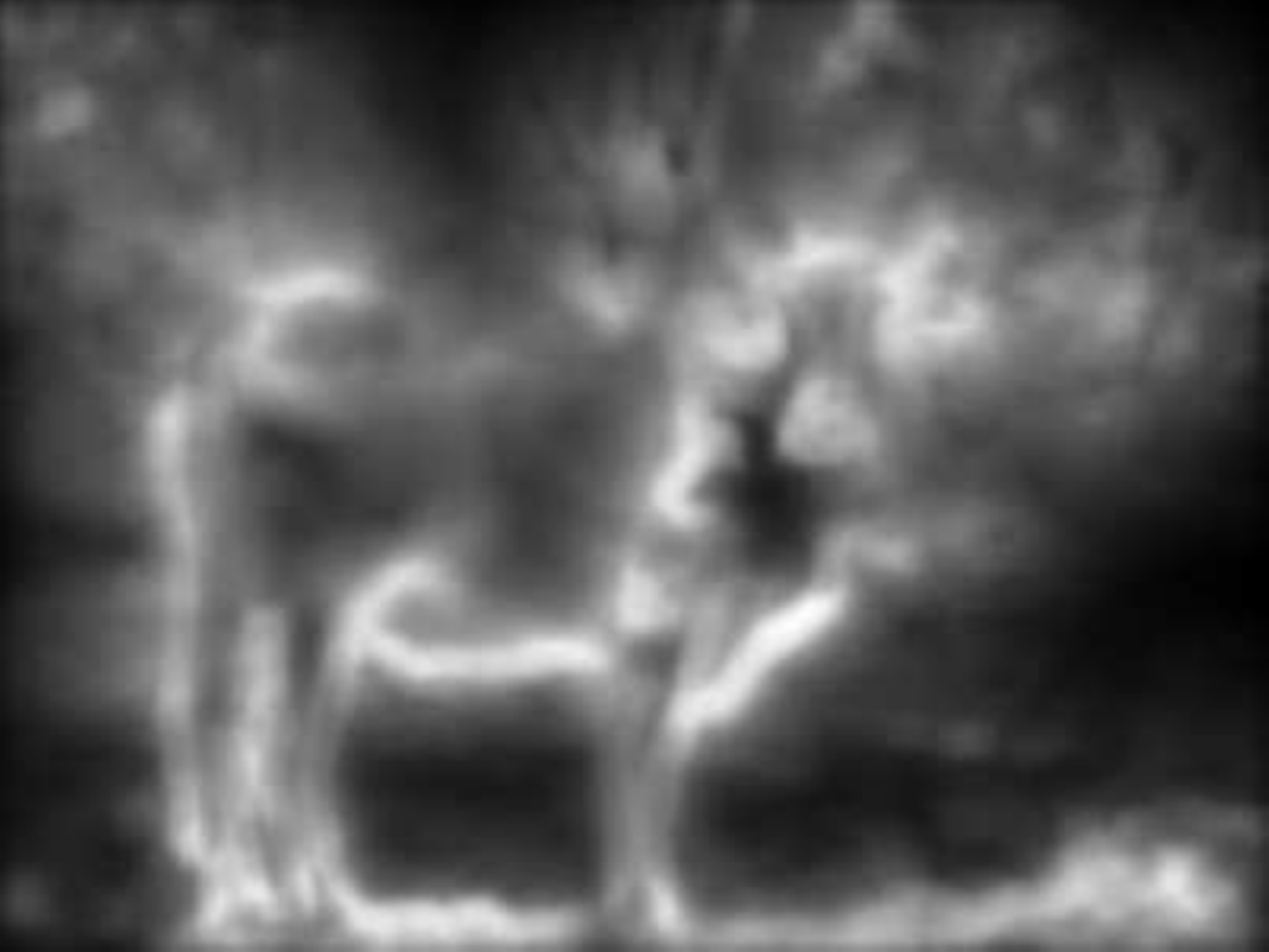}&
	\addExample{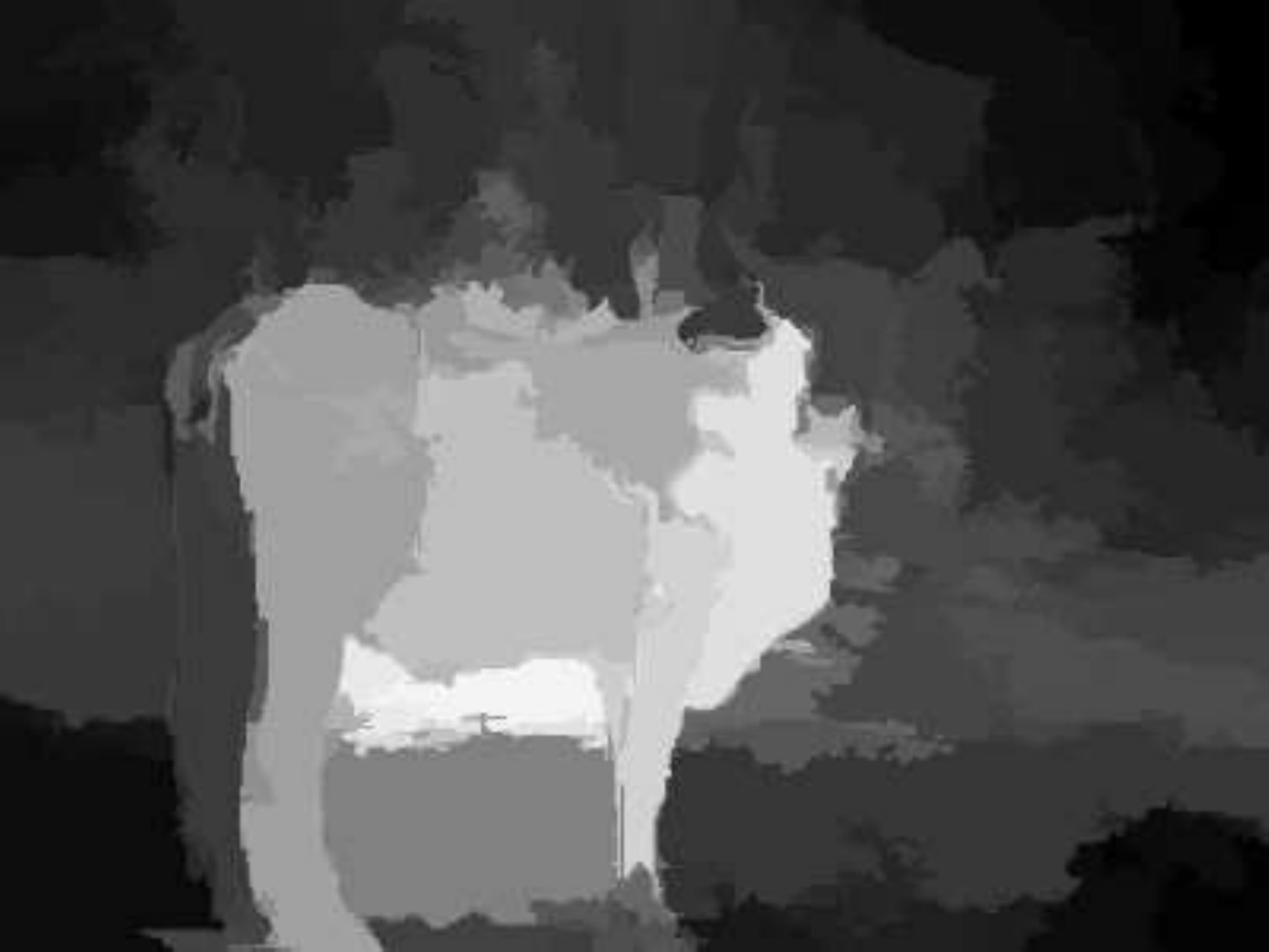}&
	\addExample{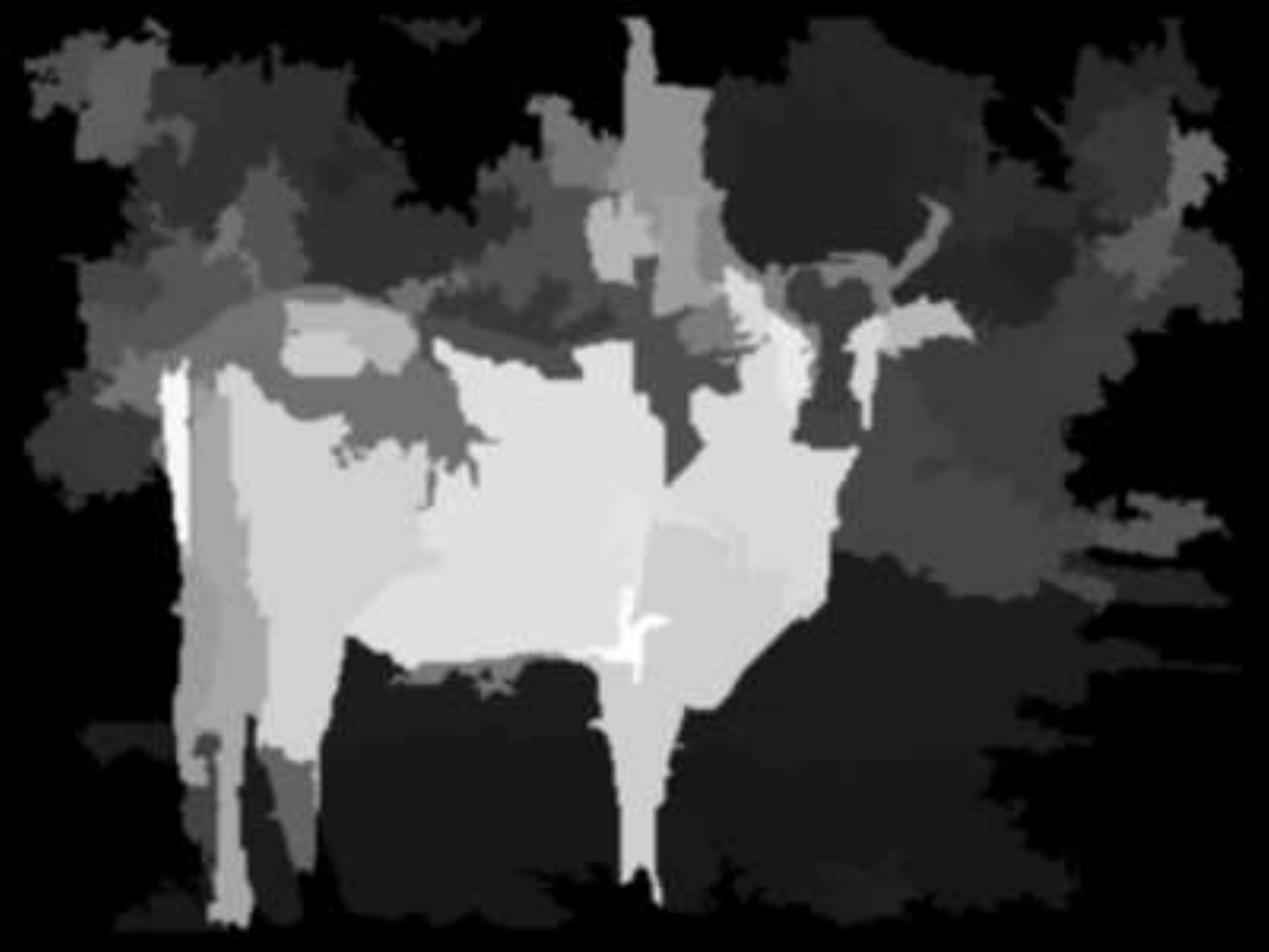}&
	\addExample{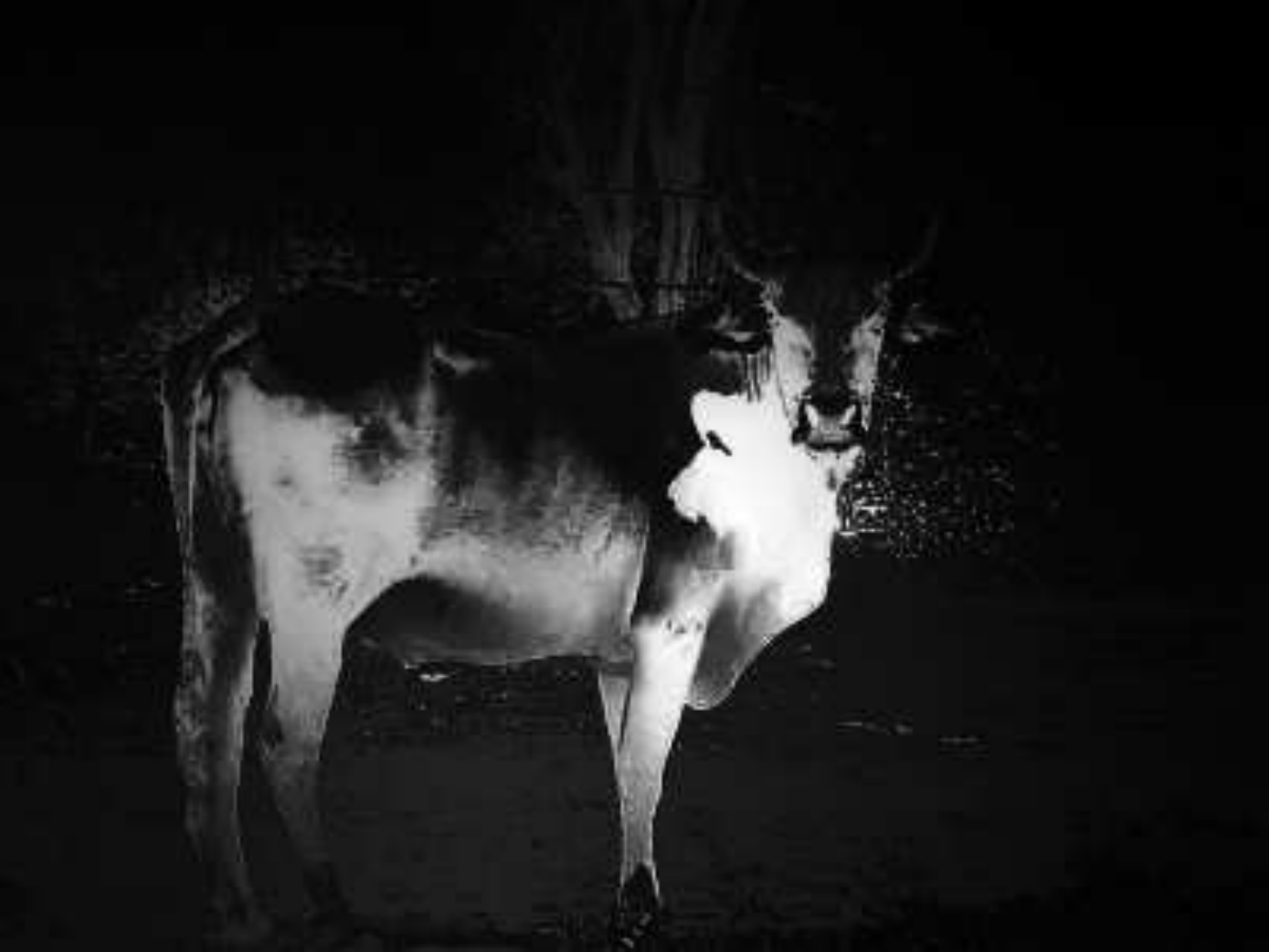}&
	\addExample{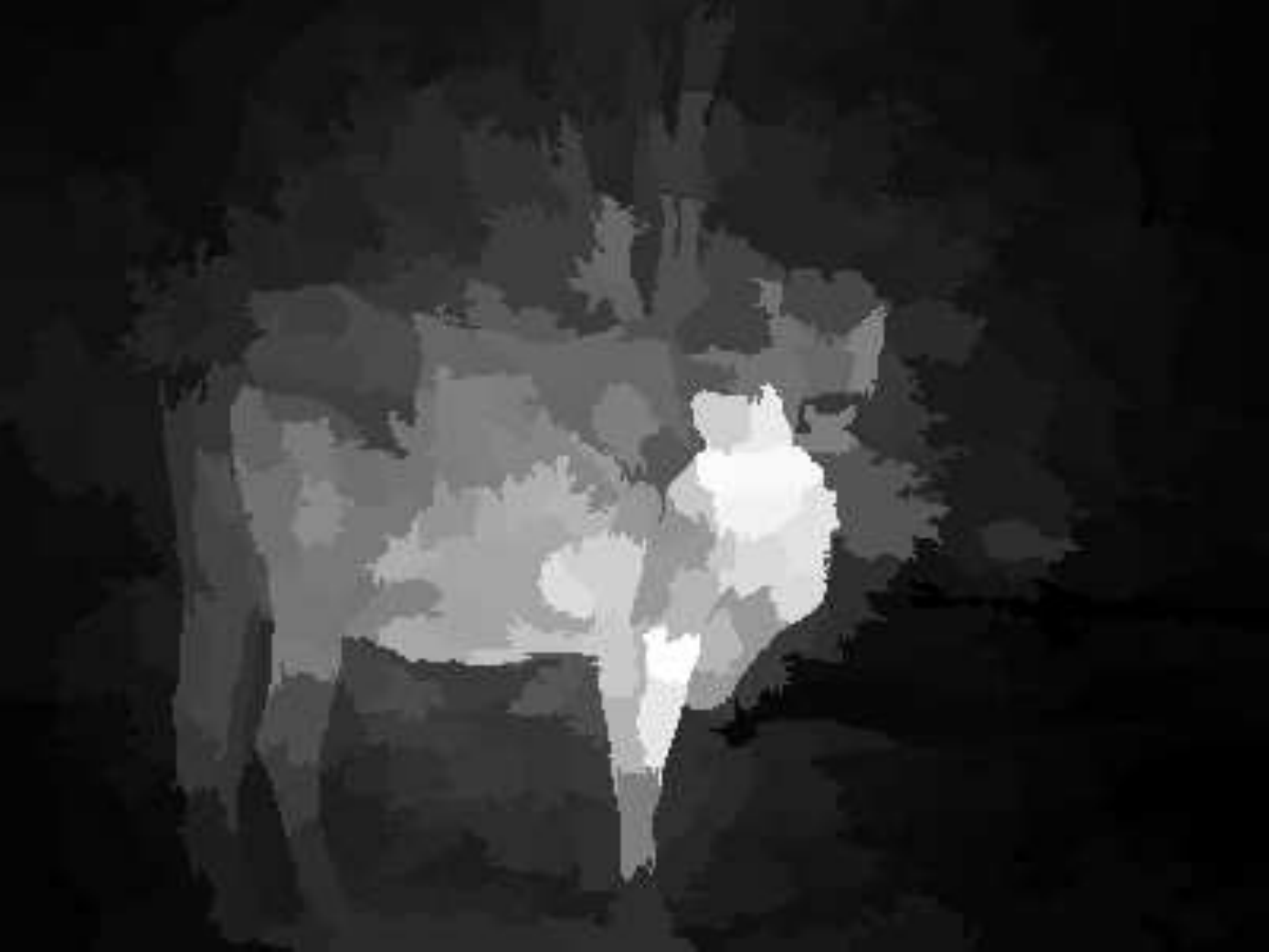}&
	\addExample{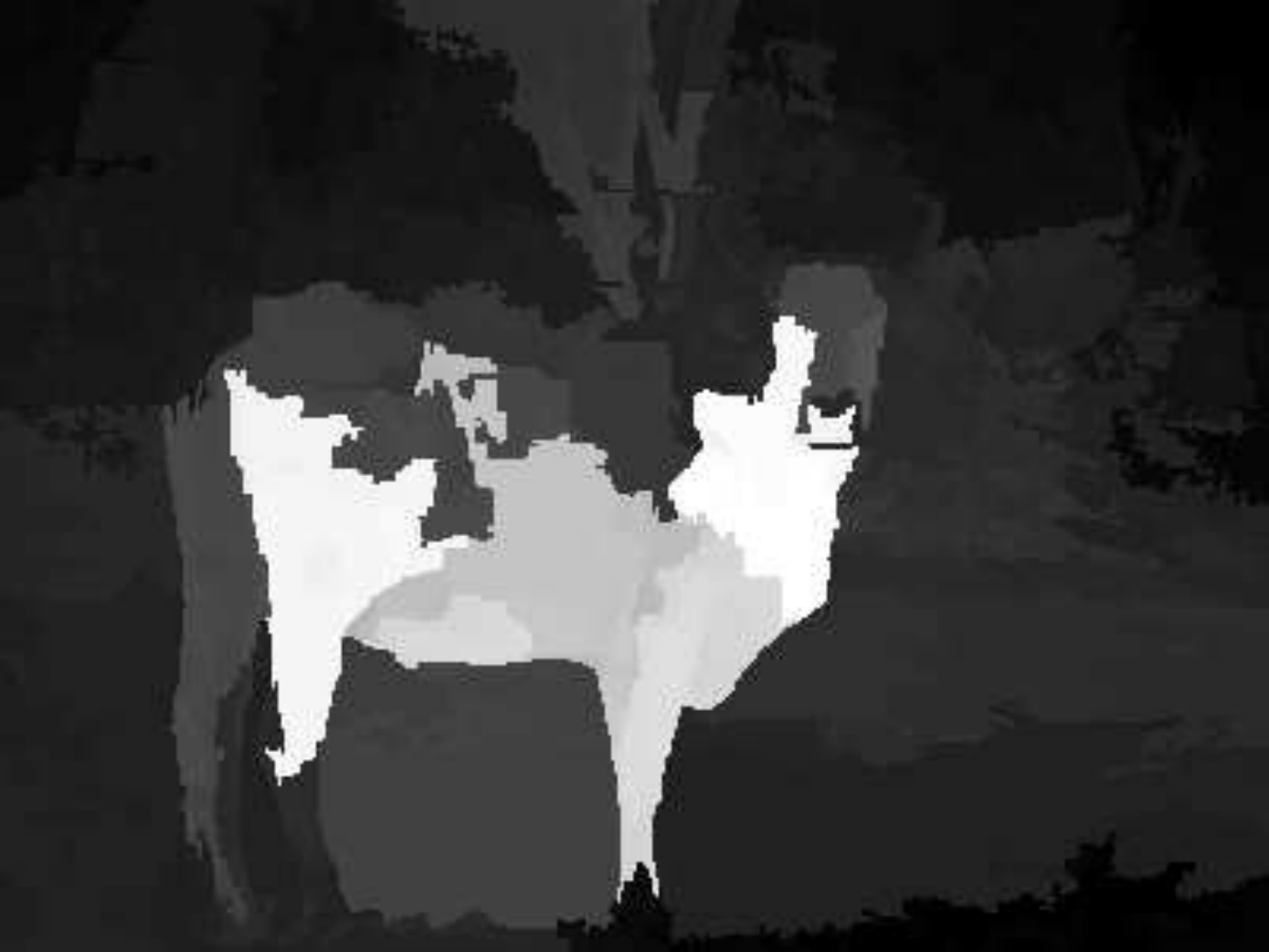}&
	\addExample{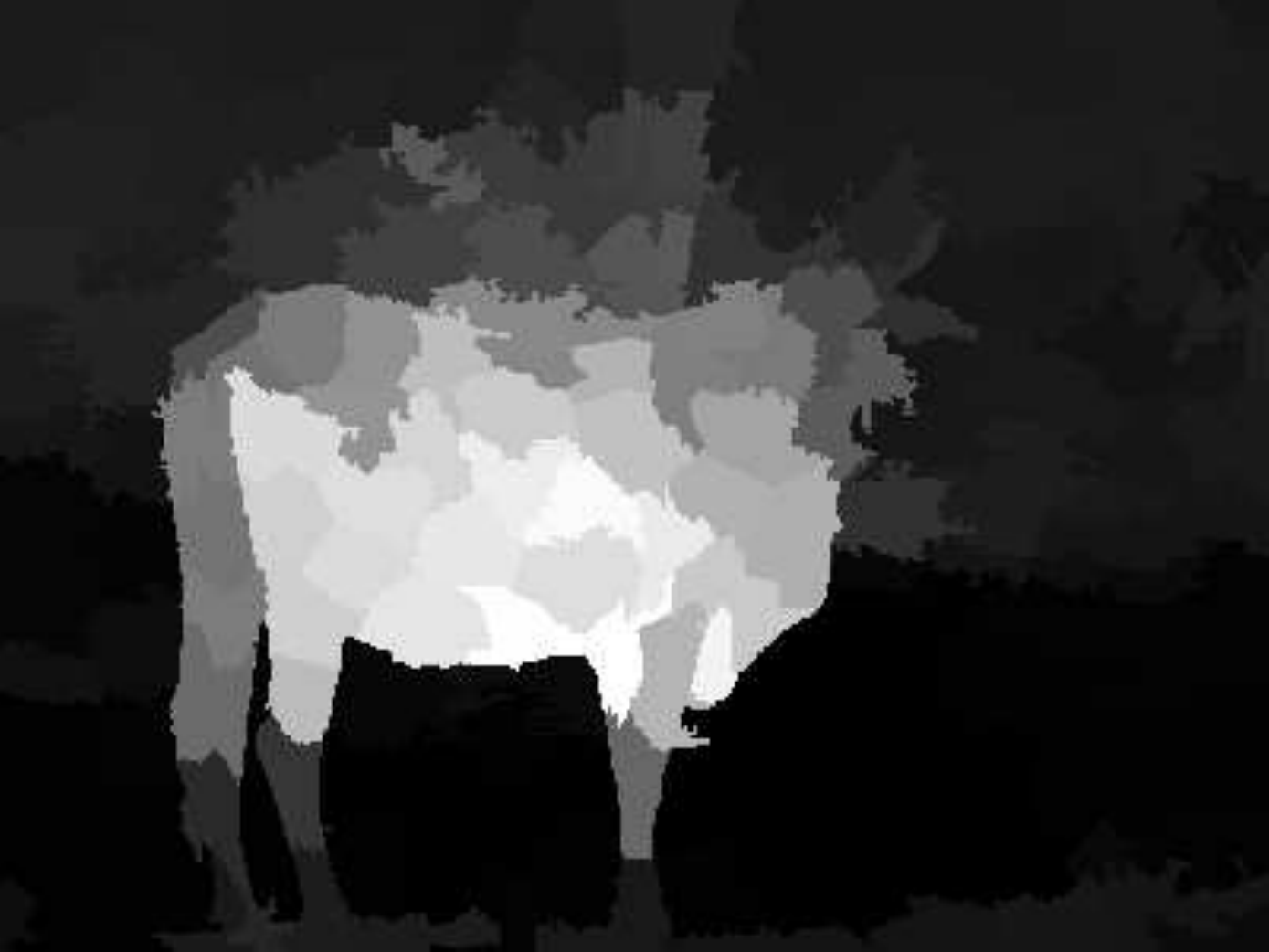}&
	\addExample{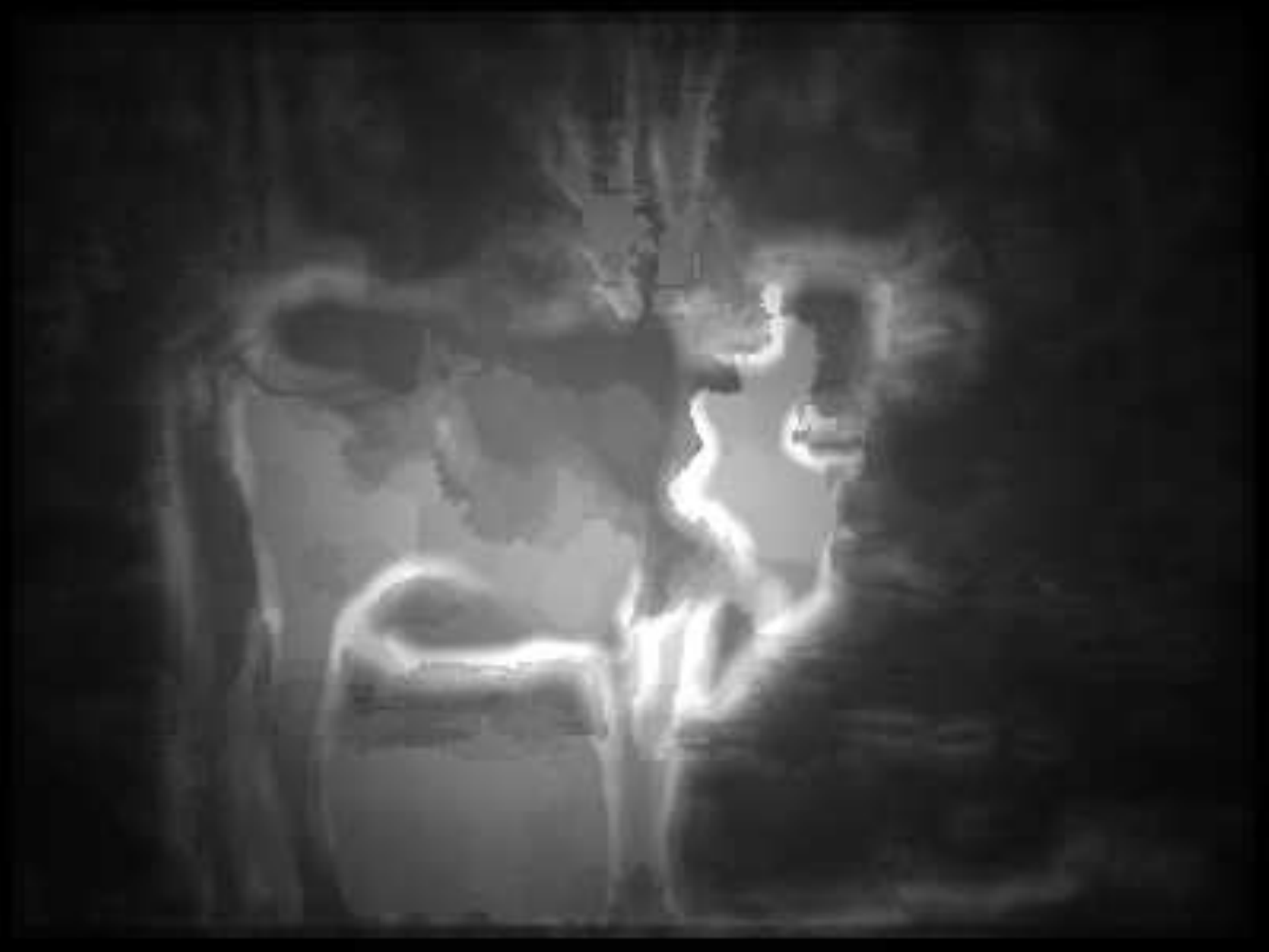}&
	\addExample{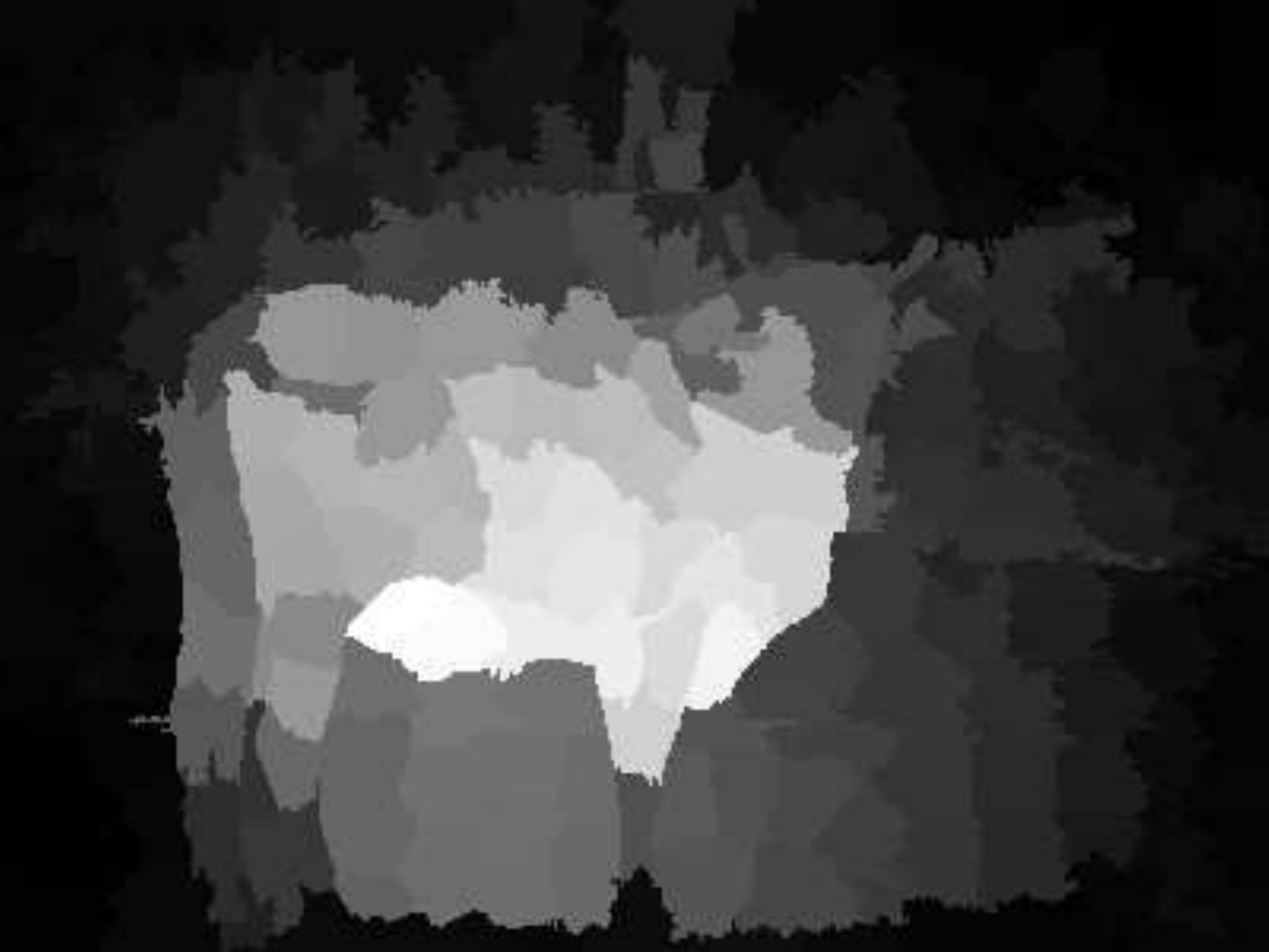}&
	\addExample{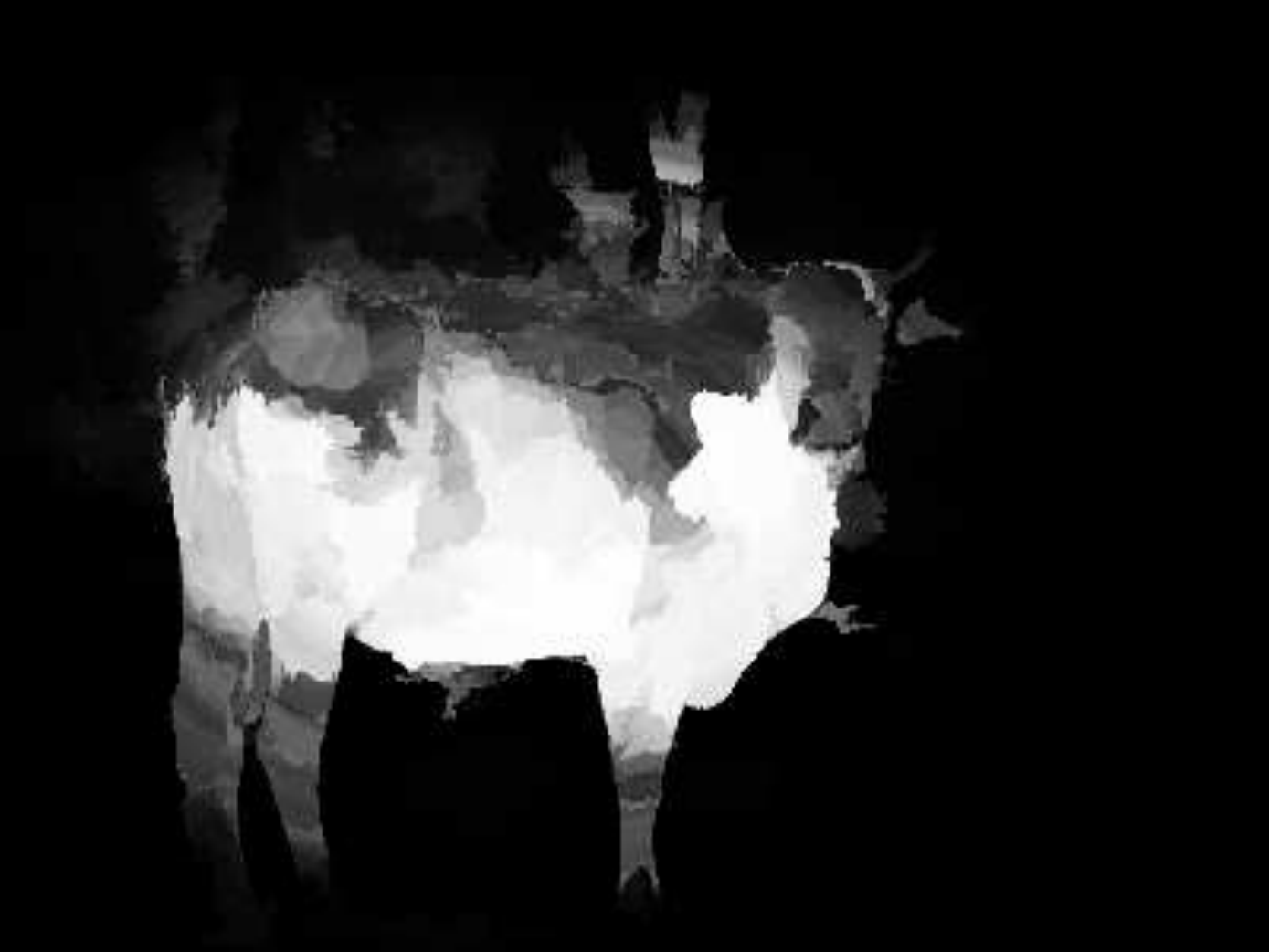}&
	\addExample{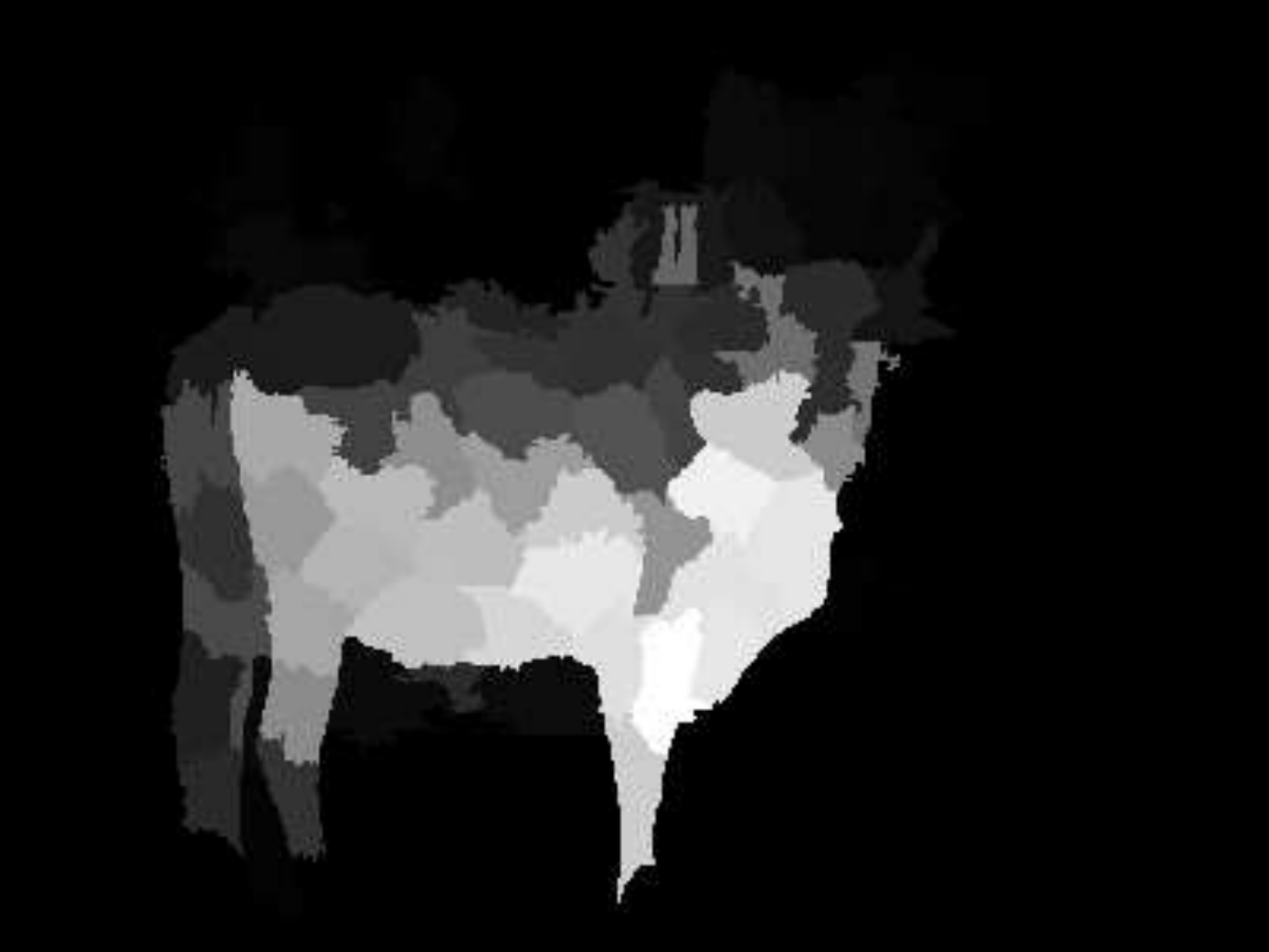}&
	\addExample{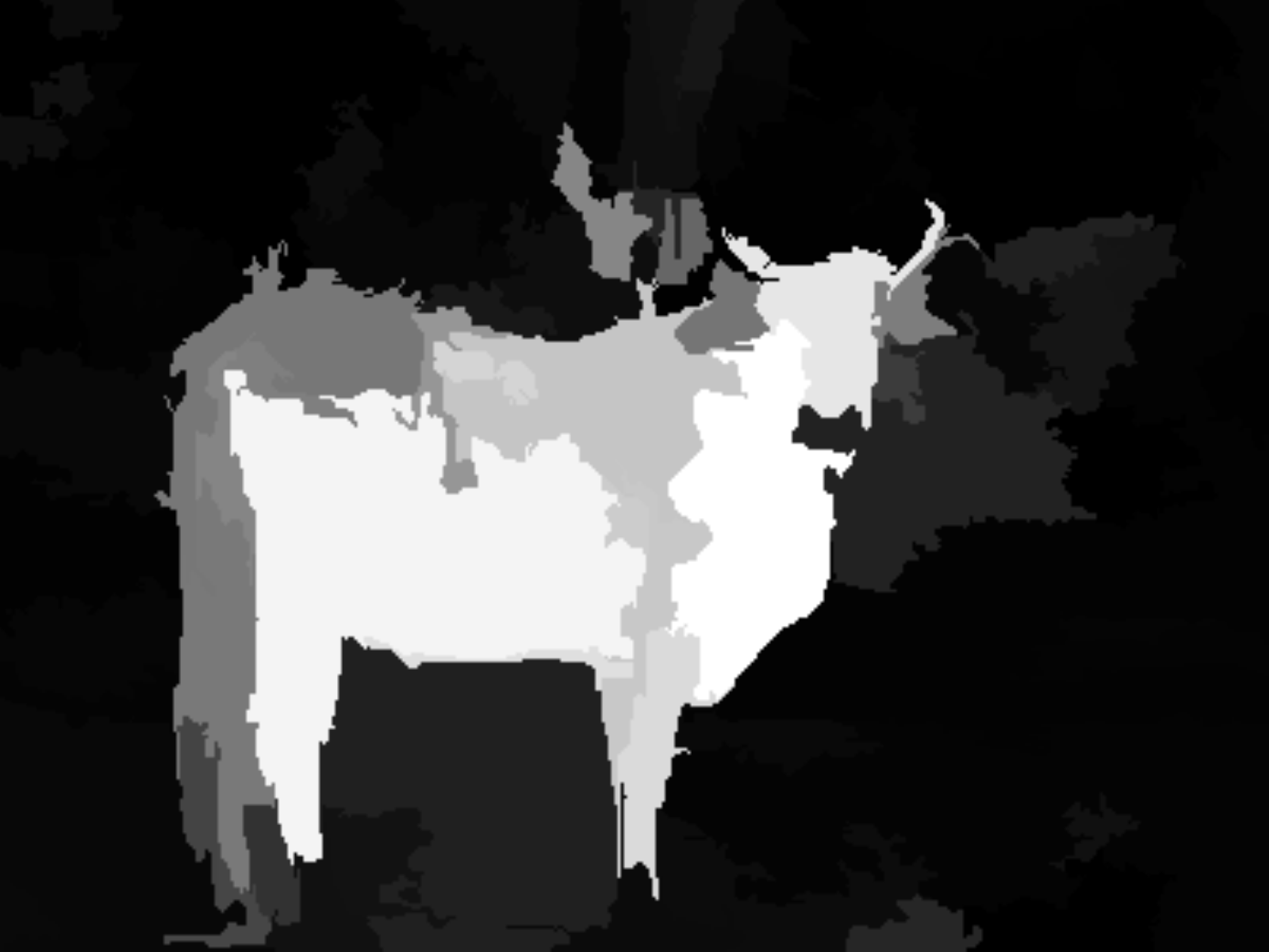}&
	\addExample{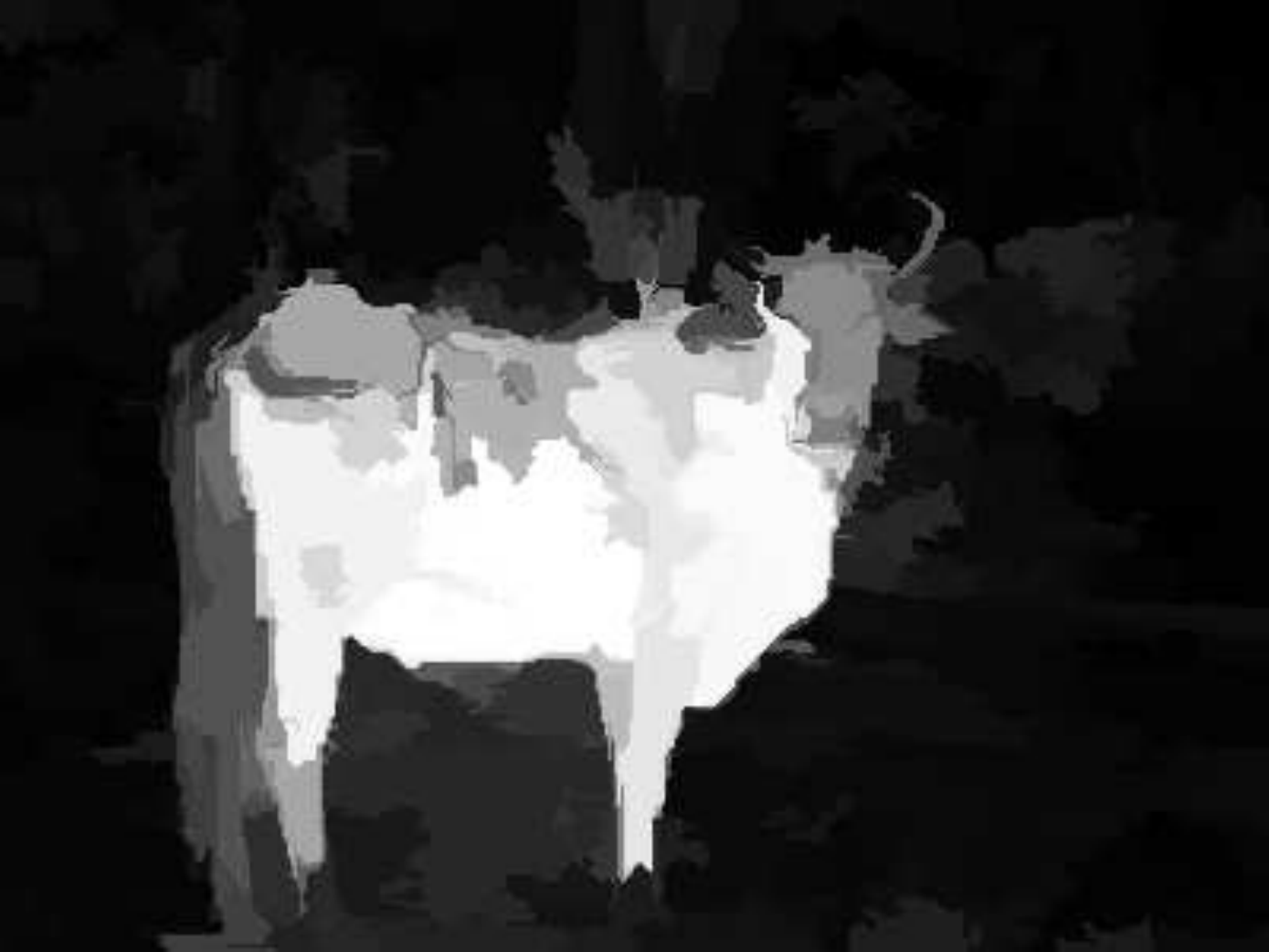}
    \\
	\addExample{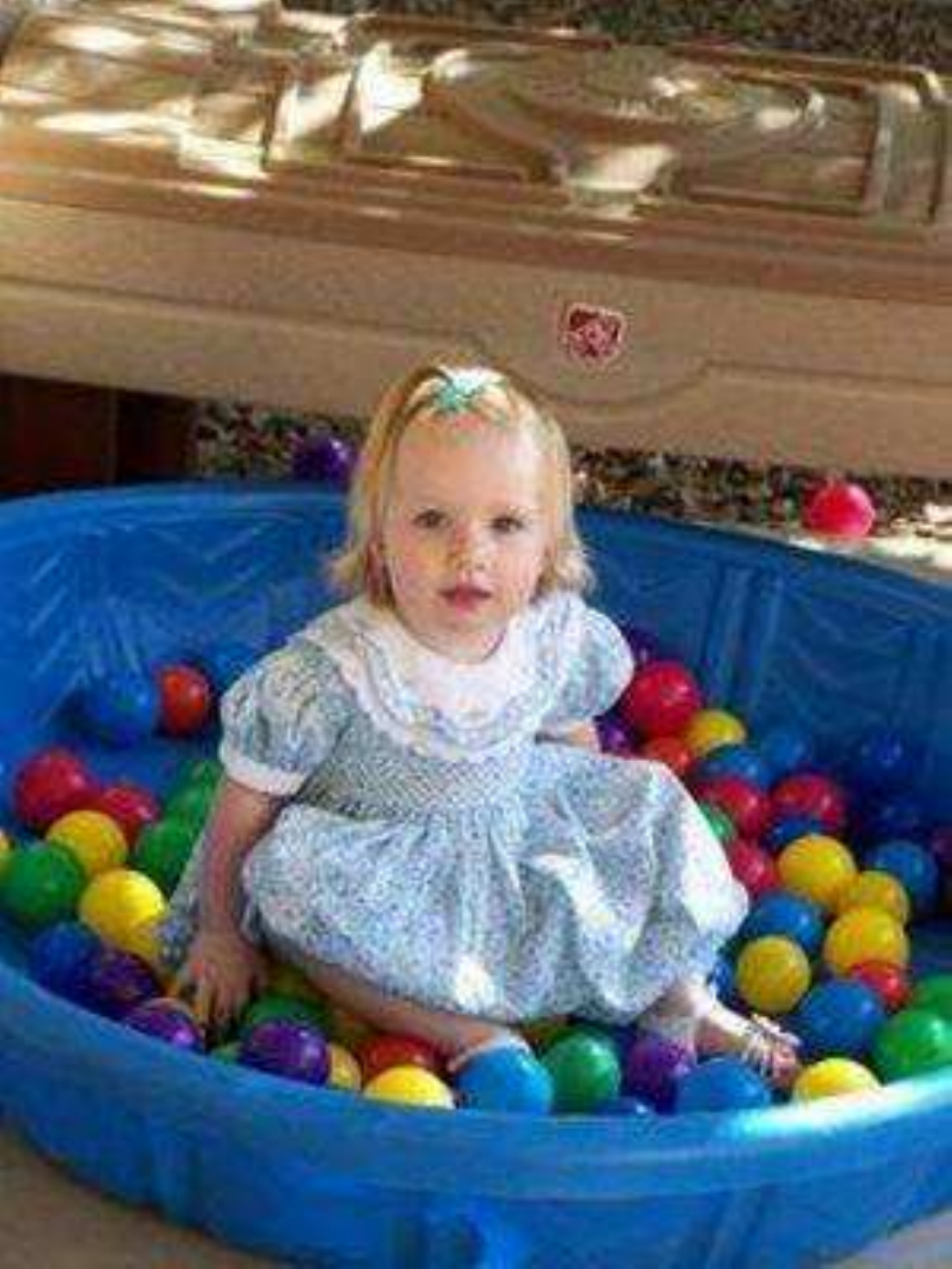}&
	\addExample{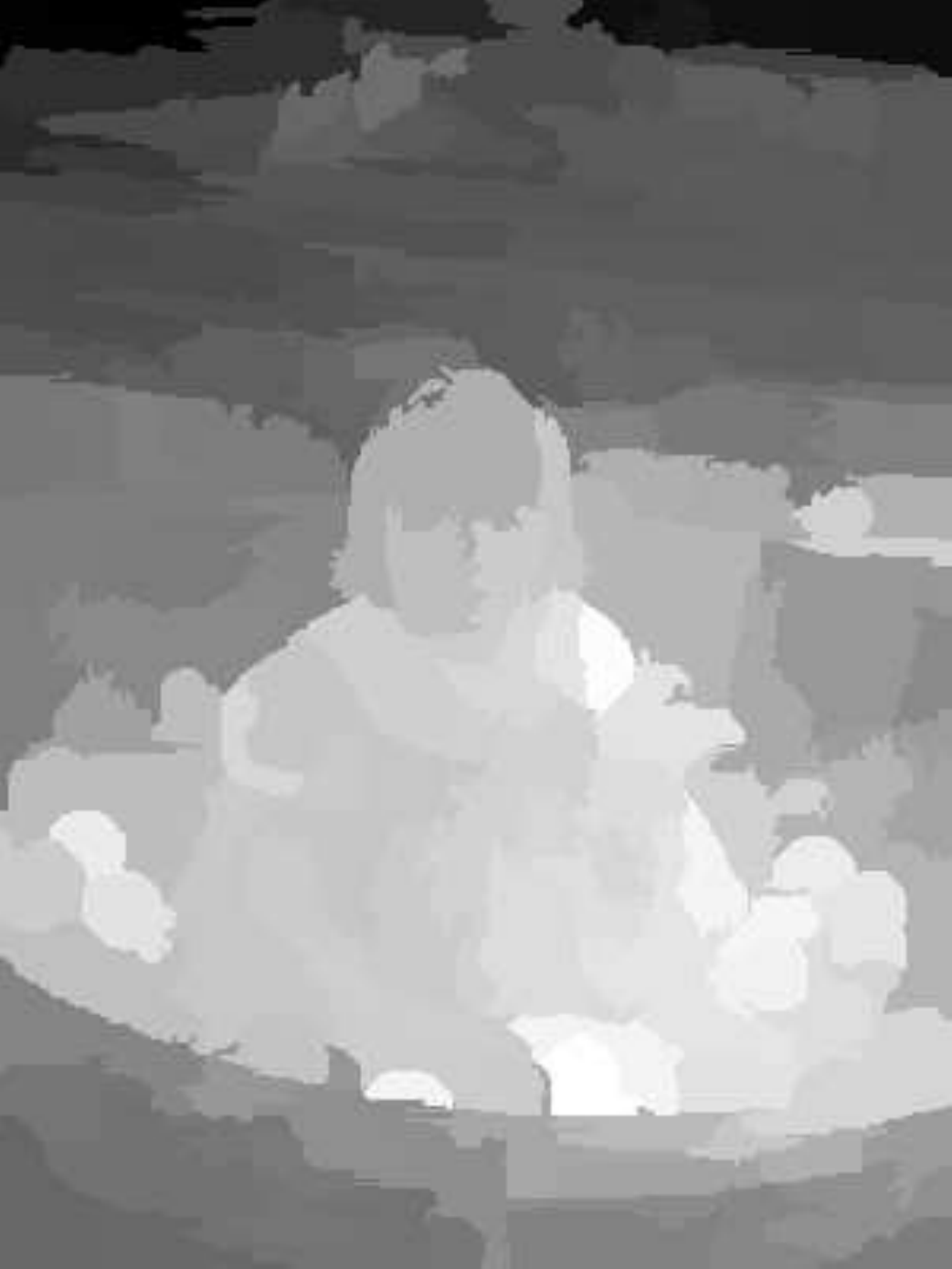}&
	\addExample{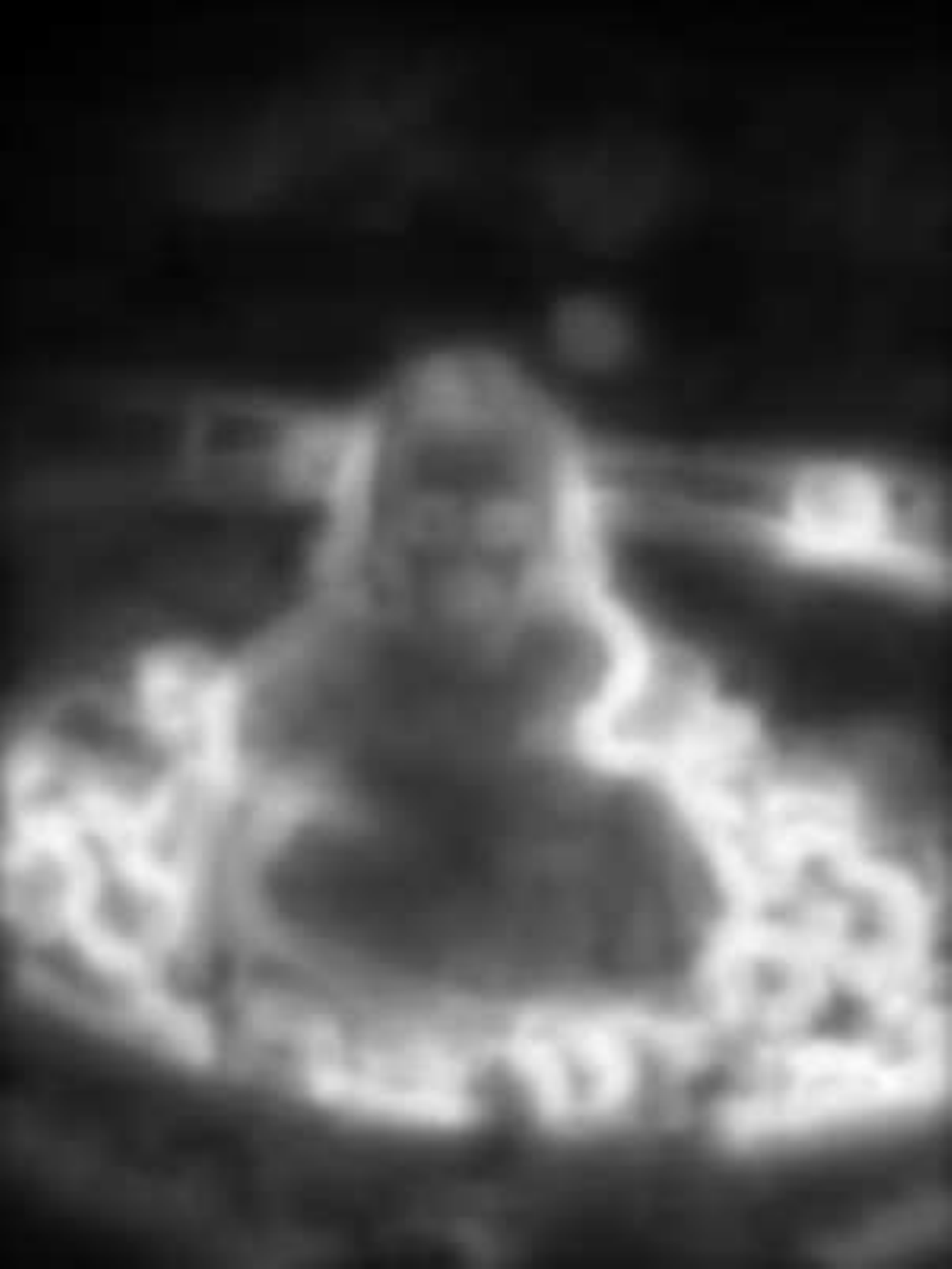}&
	\addExample{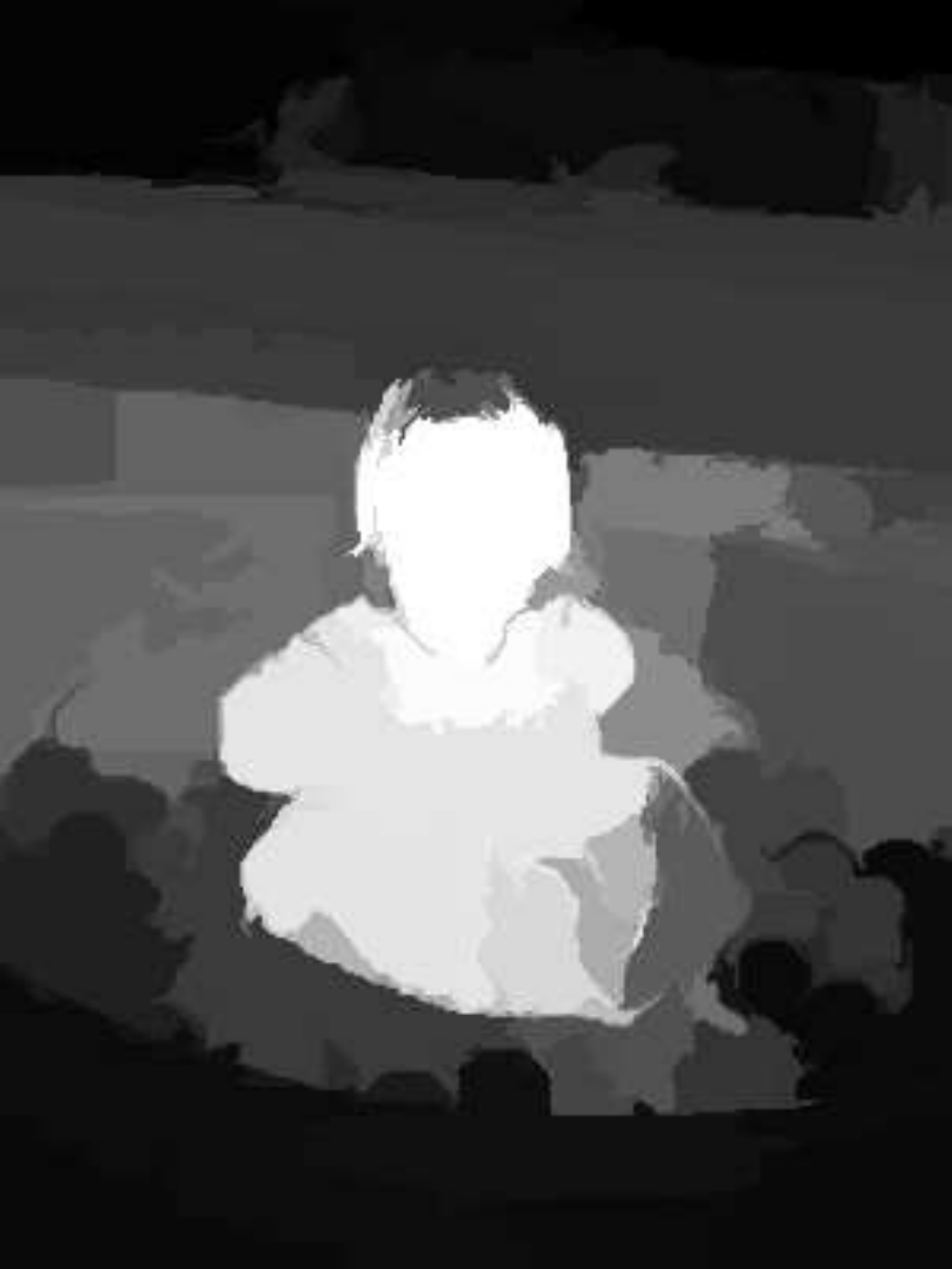}&
	\addExample{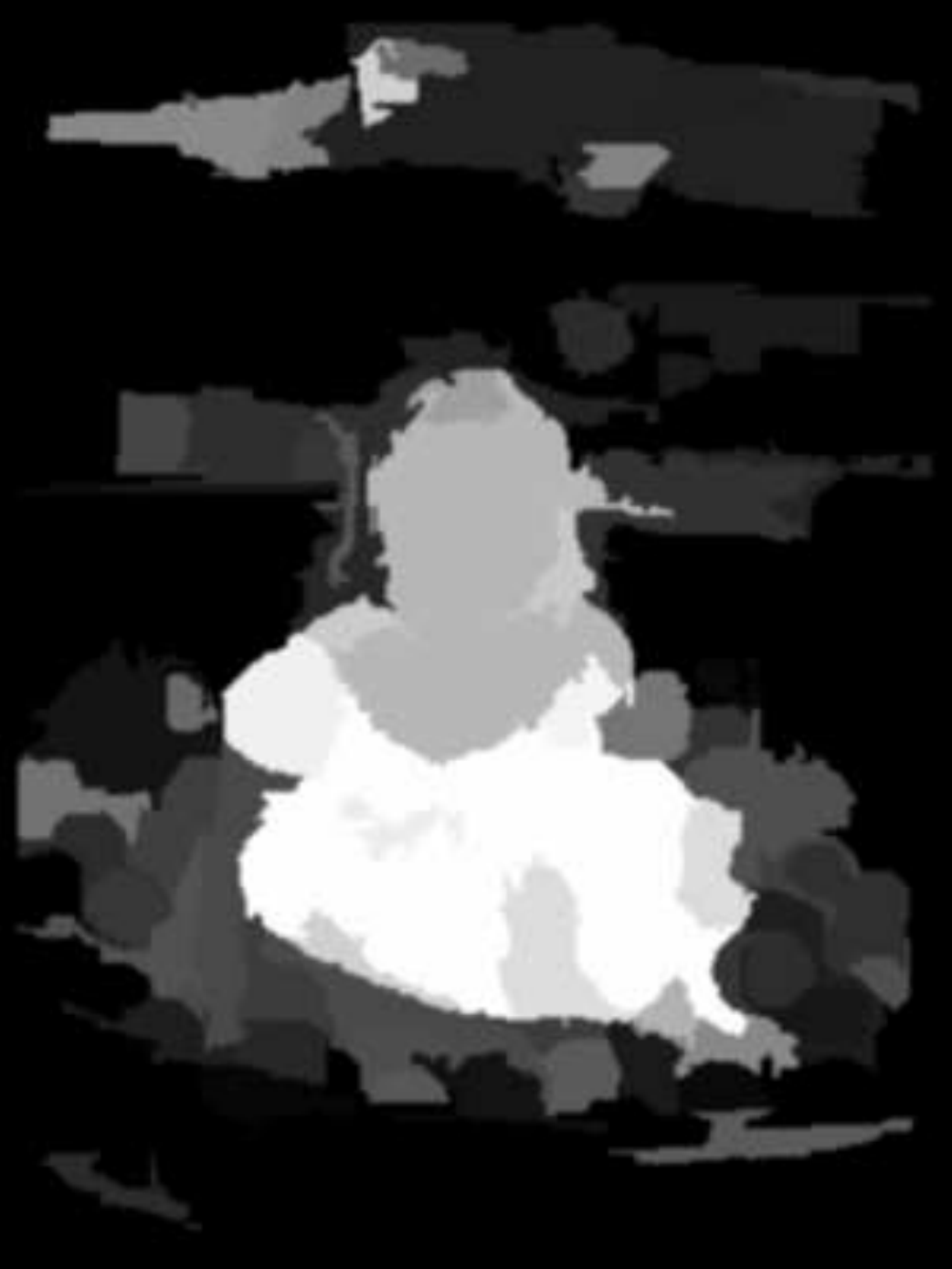}&
	\addExample{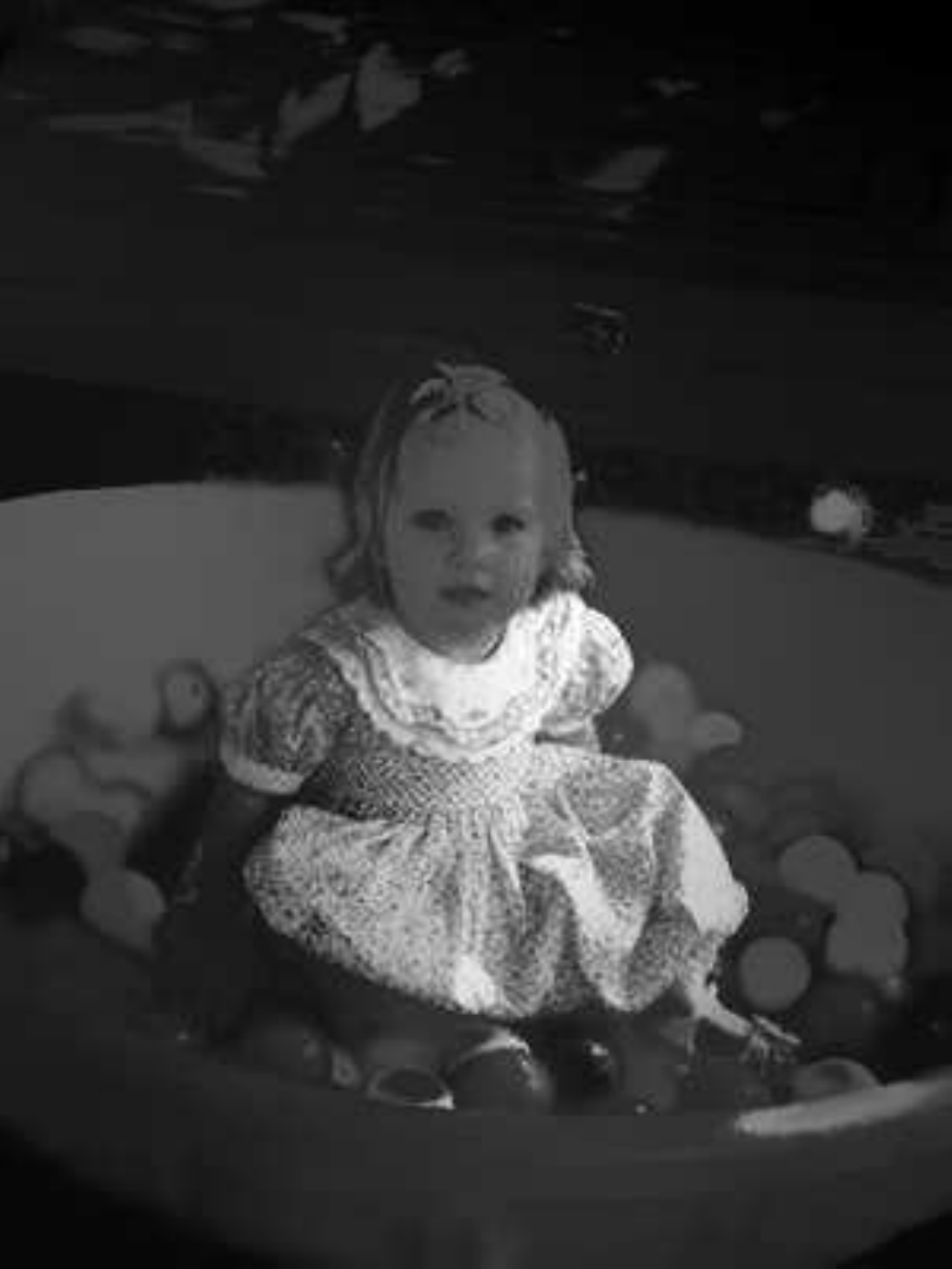}&
	\addExample{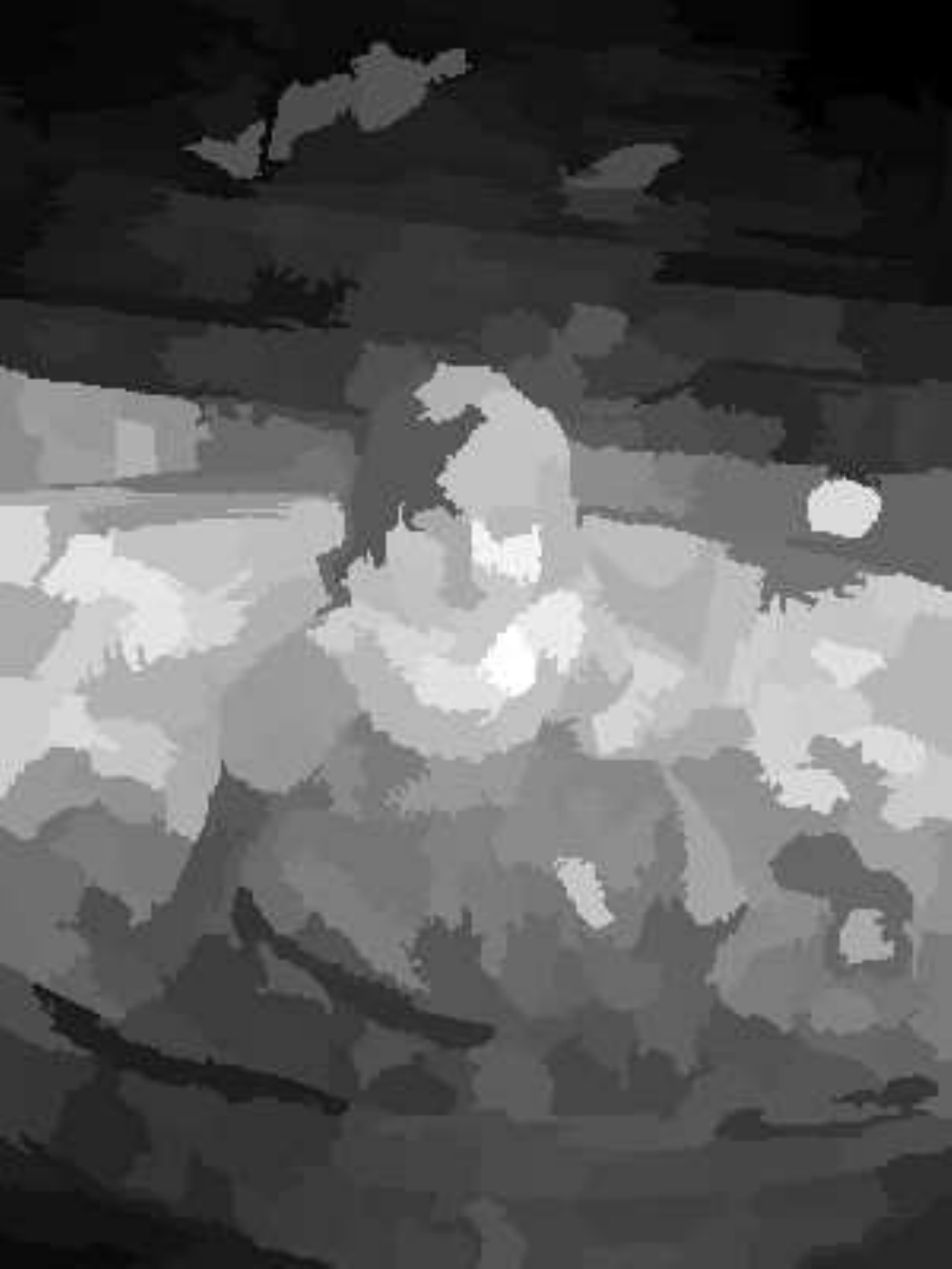}&
	\addExample{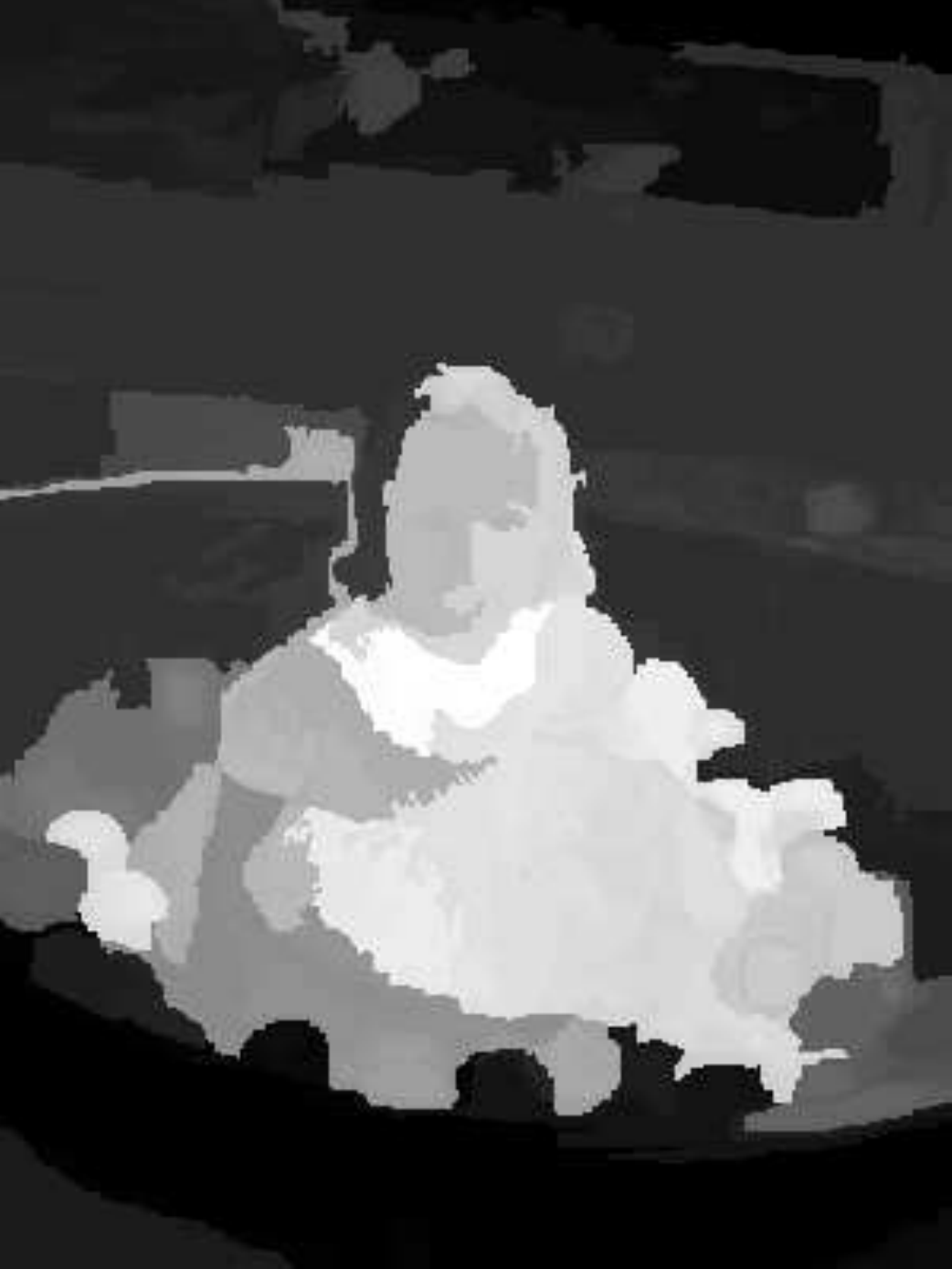}&
	\addExample{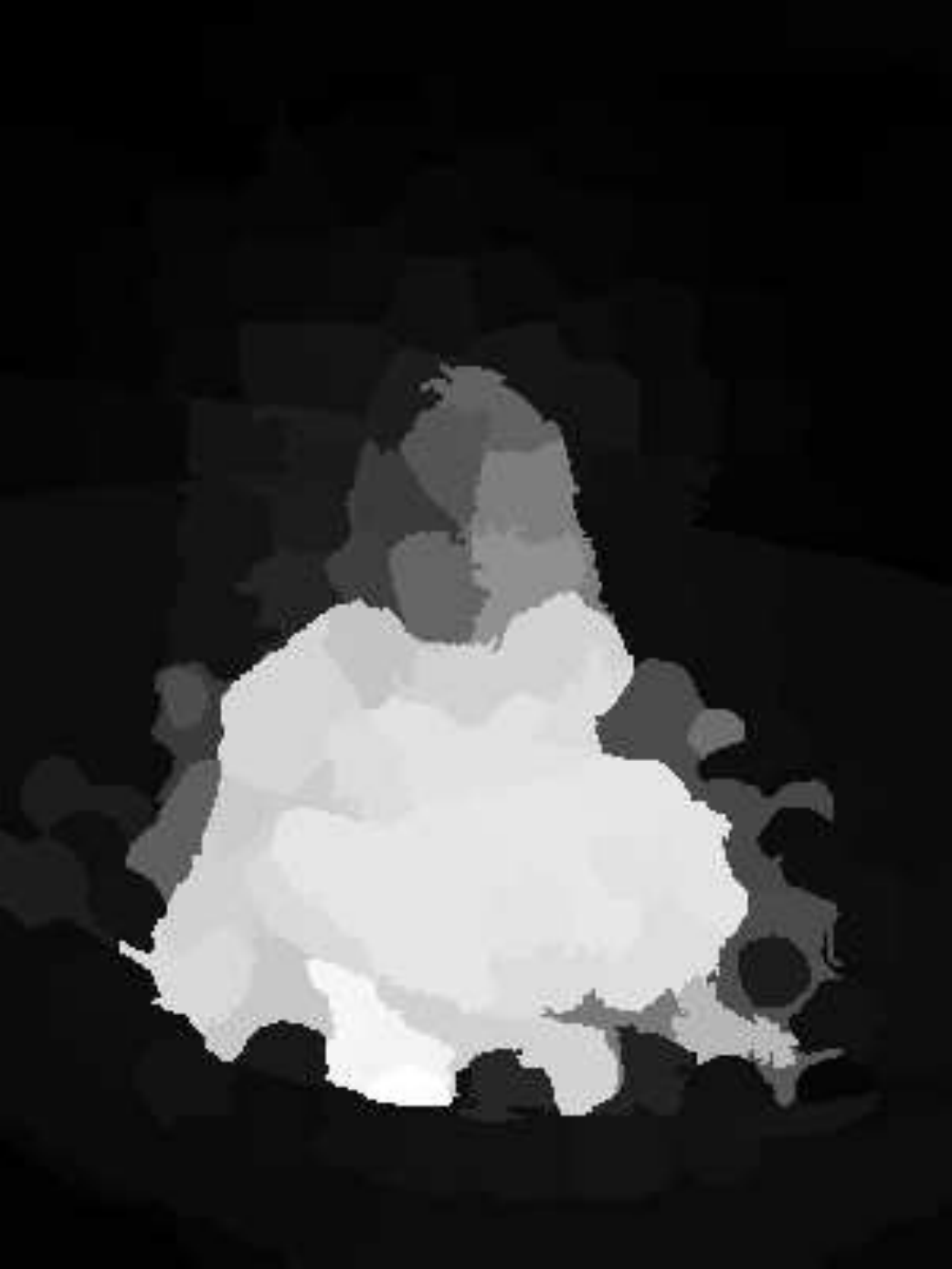}&
	\addExample{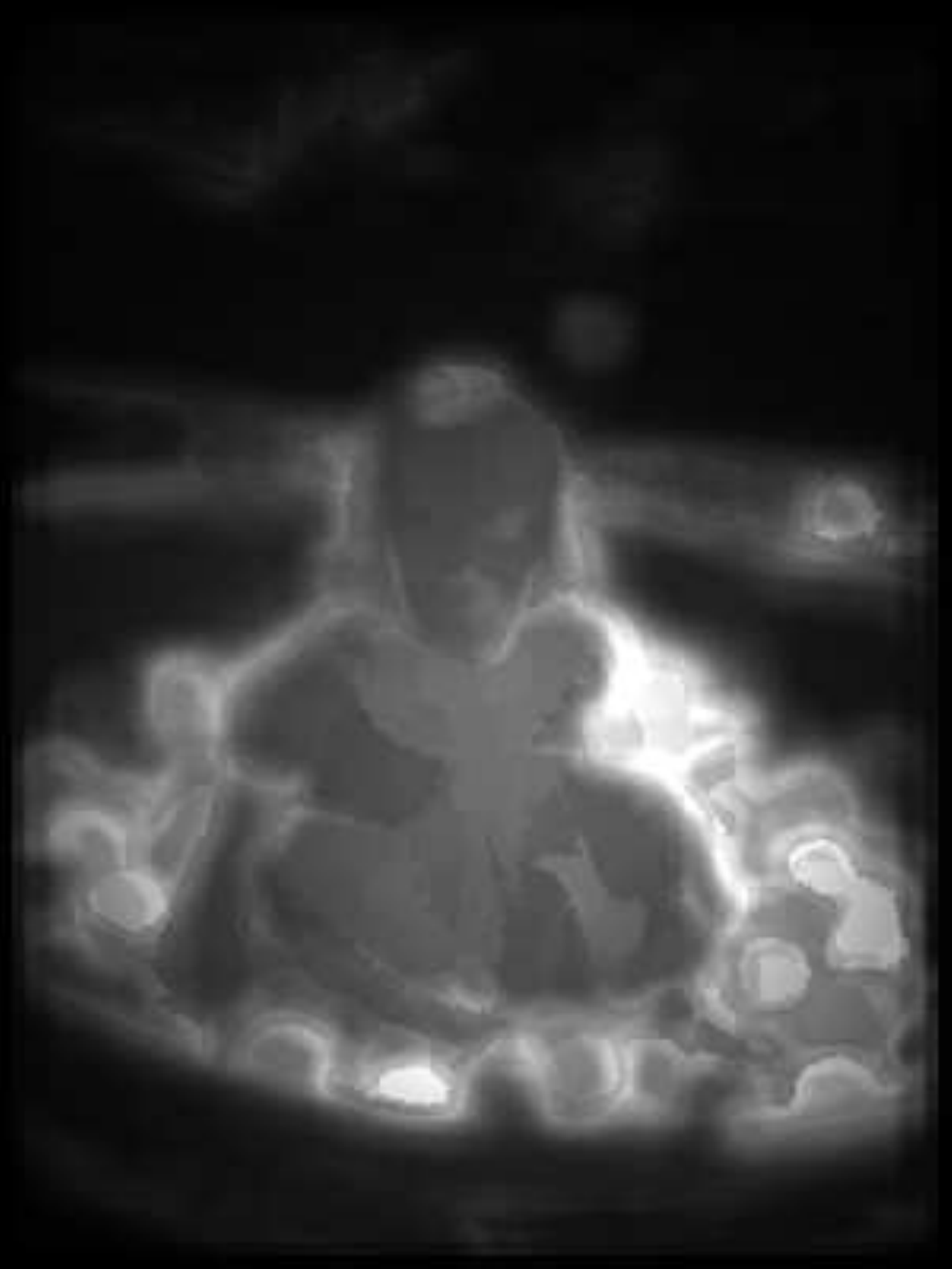}&
	\addExample{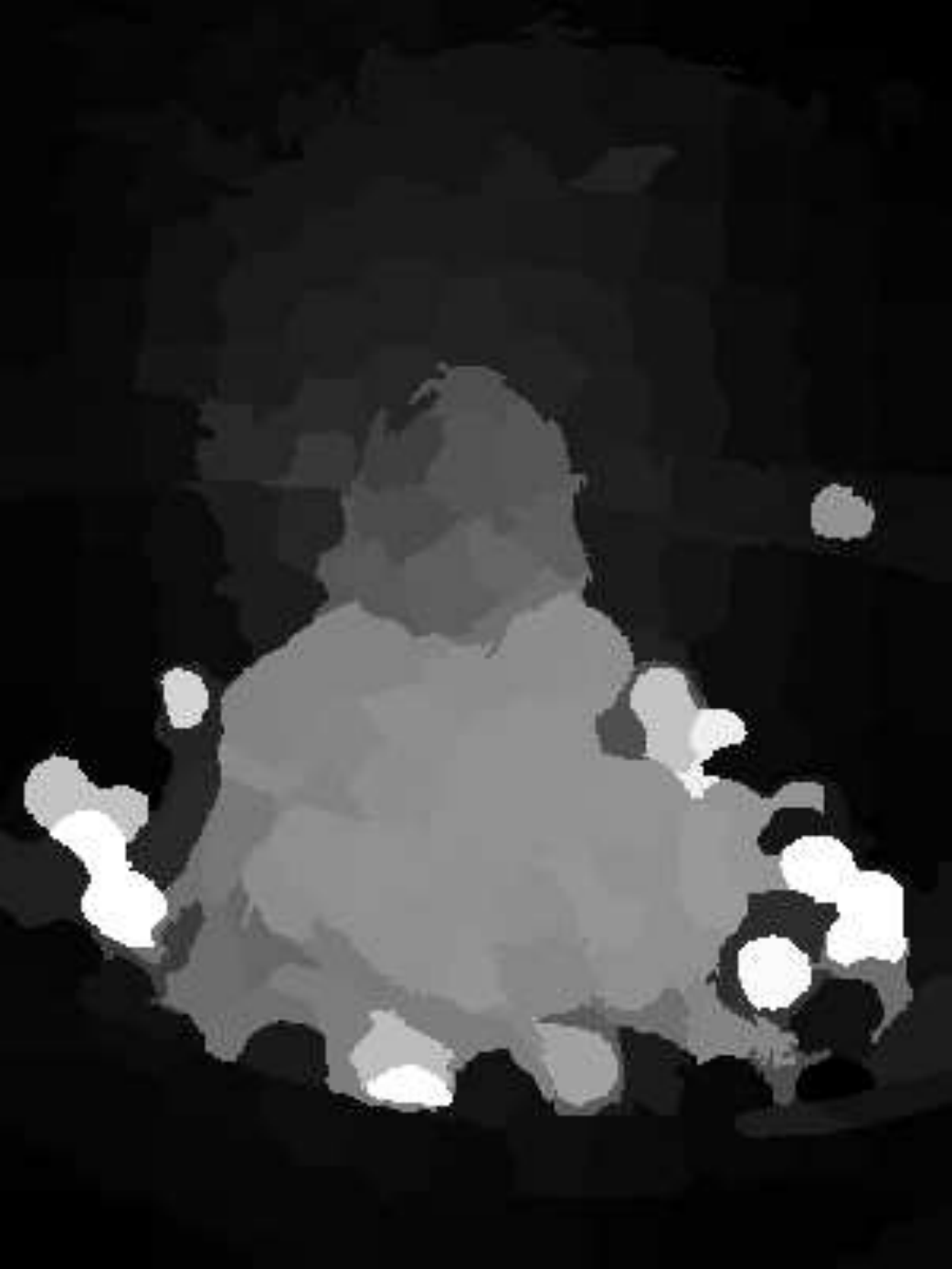}&
	\addExample{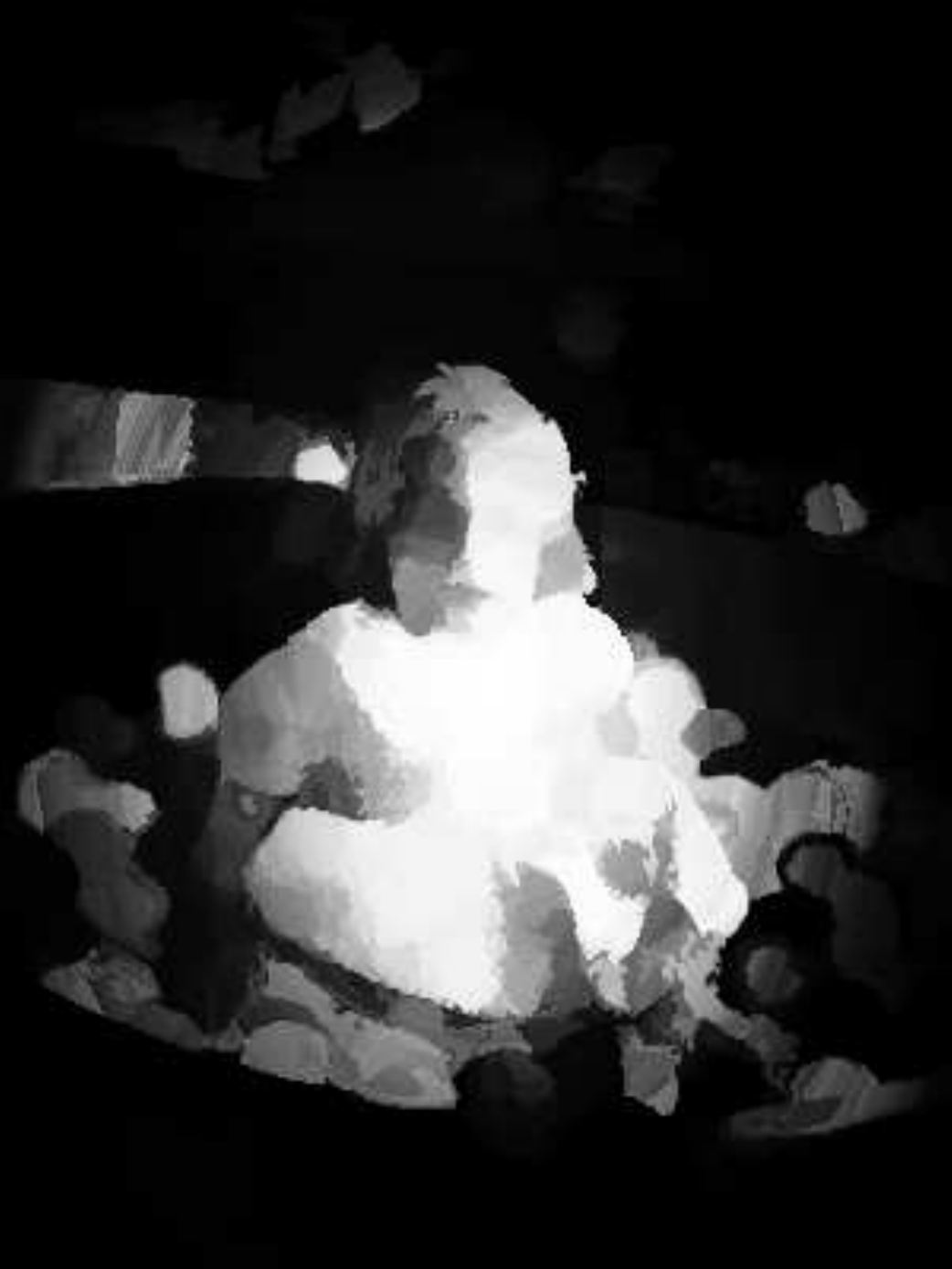}&
	\addExample{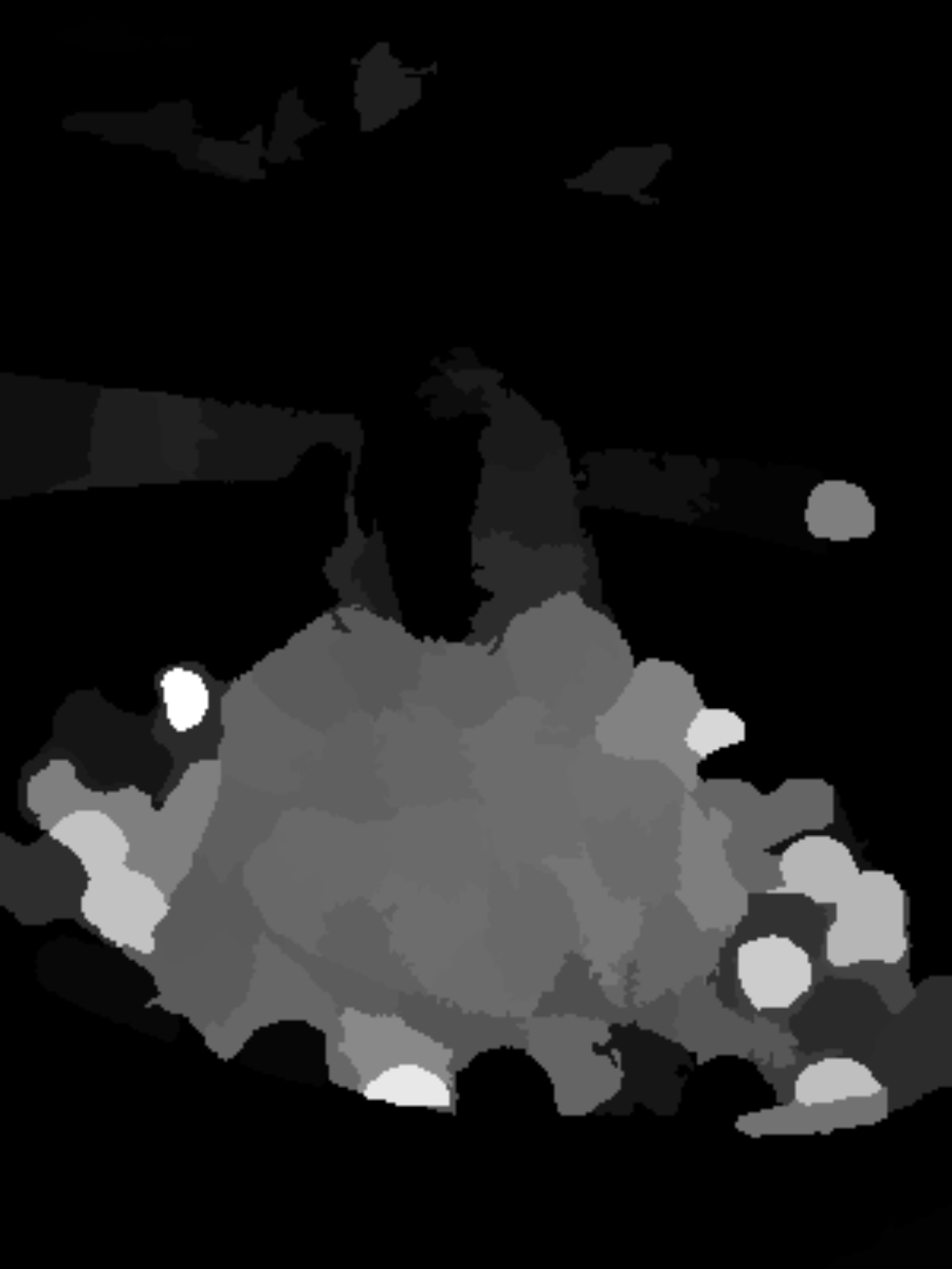}&
	\addExample{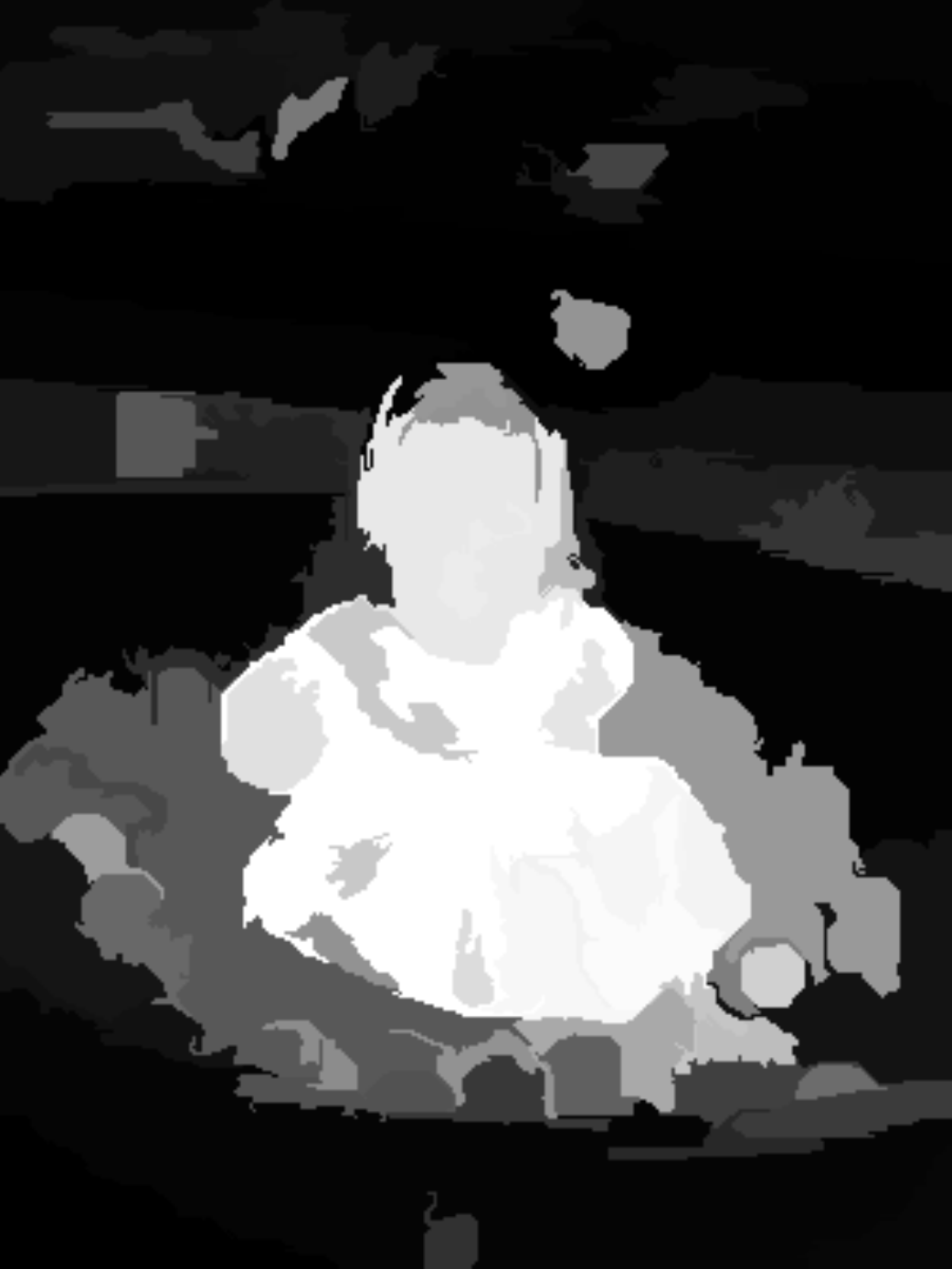}&
	\addExample{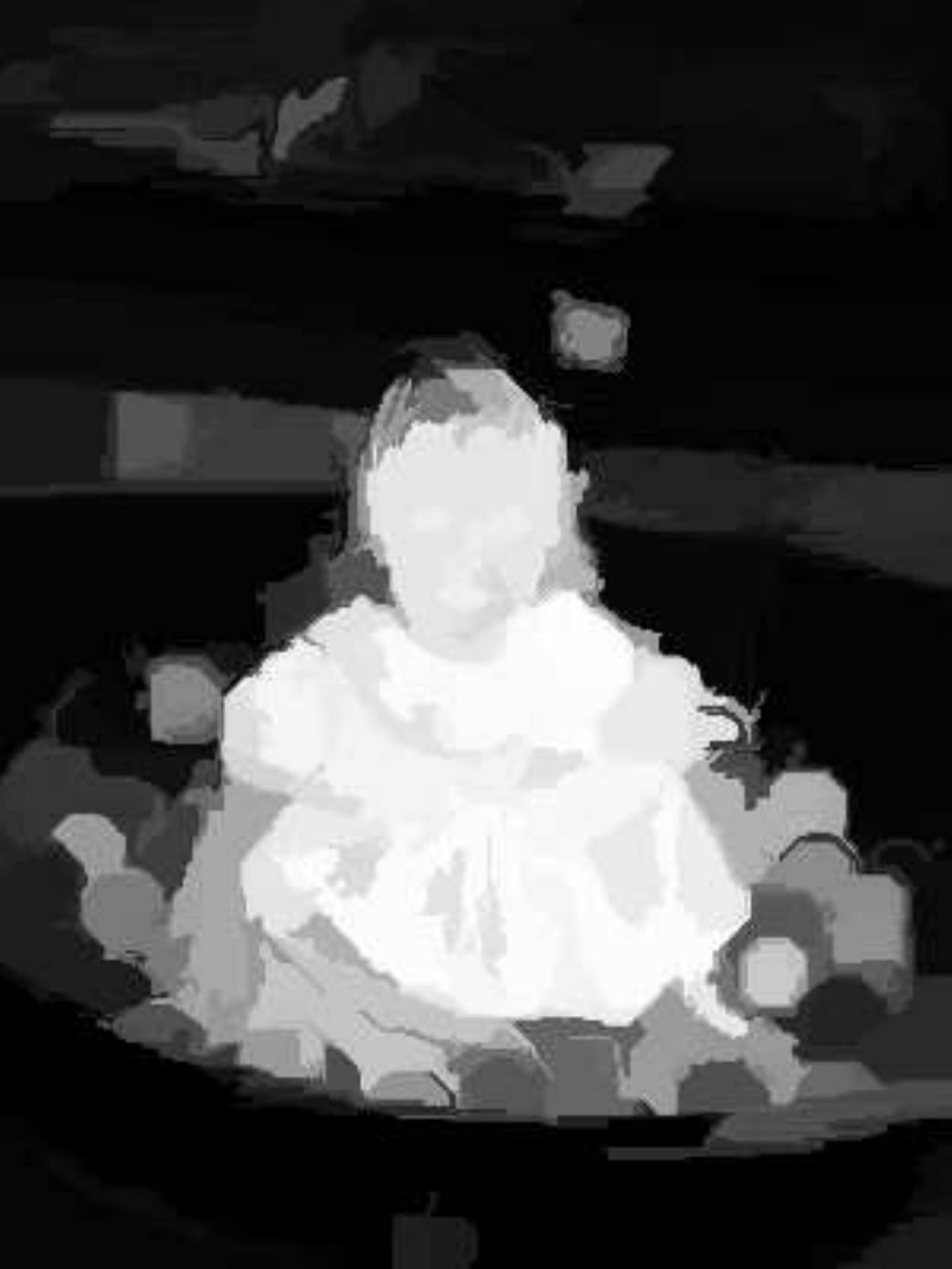}
	\\
	% \vspace{-1.8pt}
	\addExample{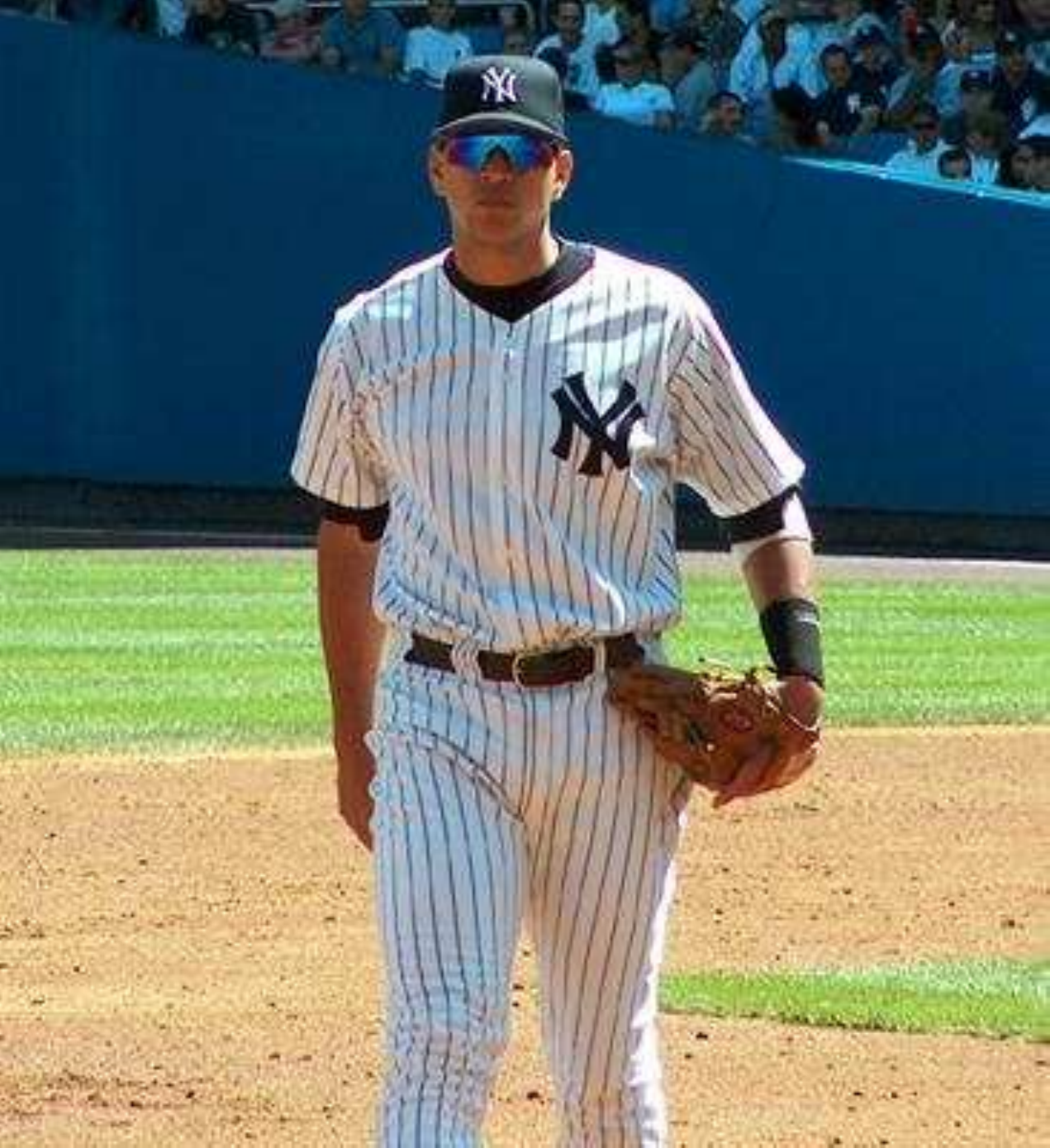}&
	\addExample{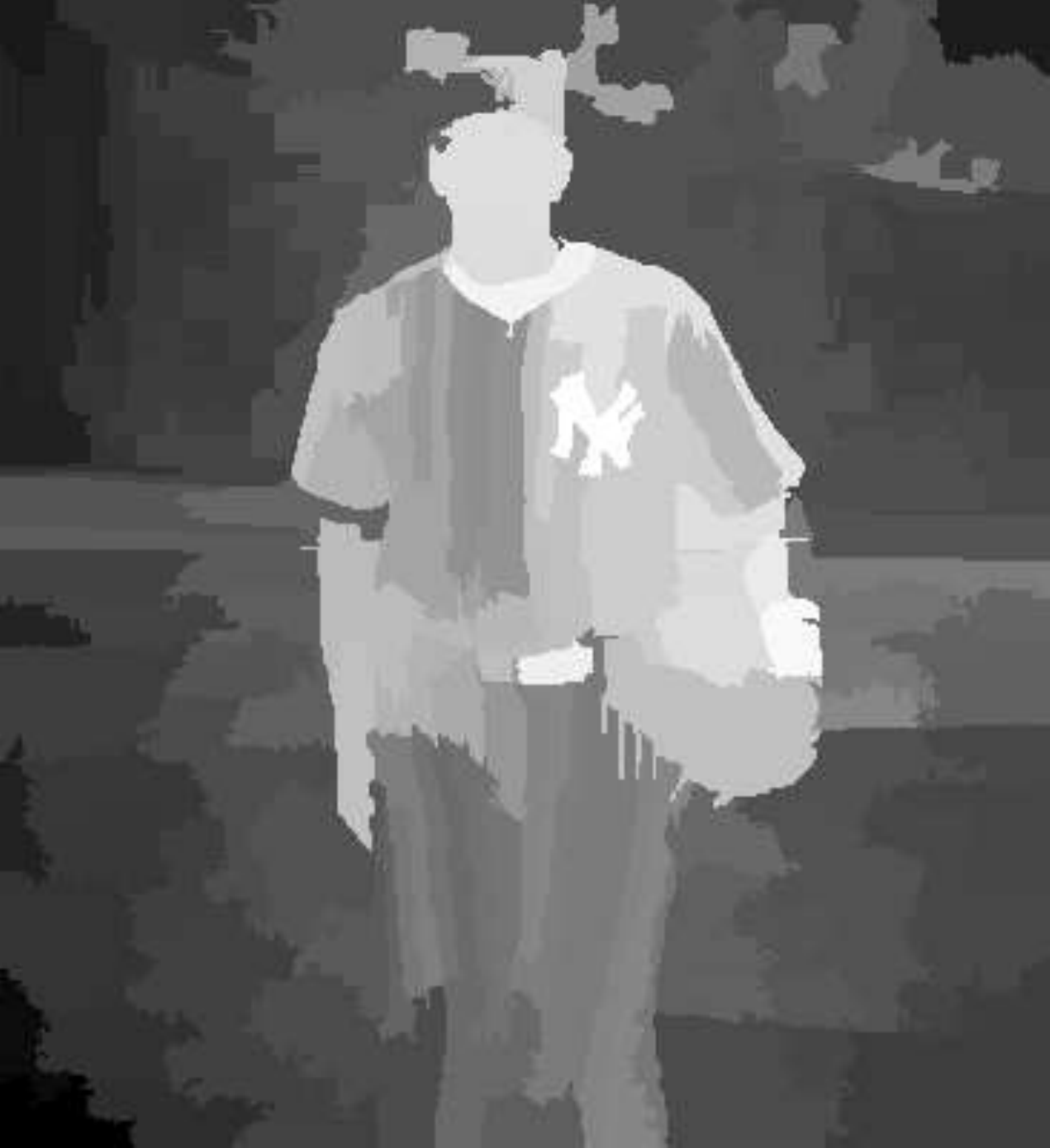}&
	\addExample{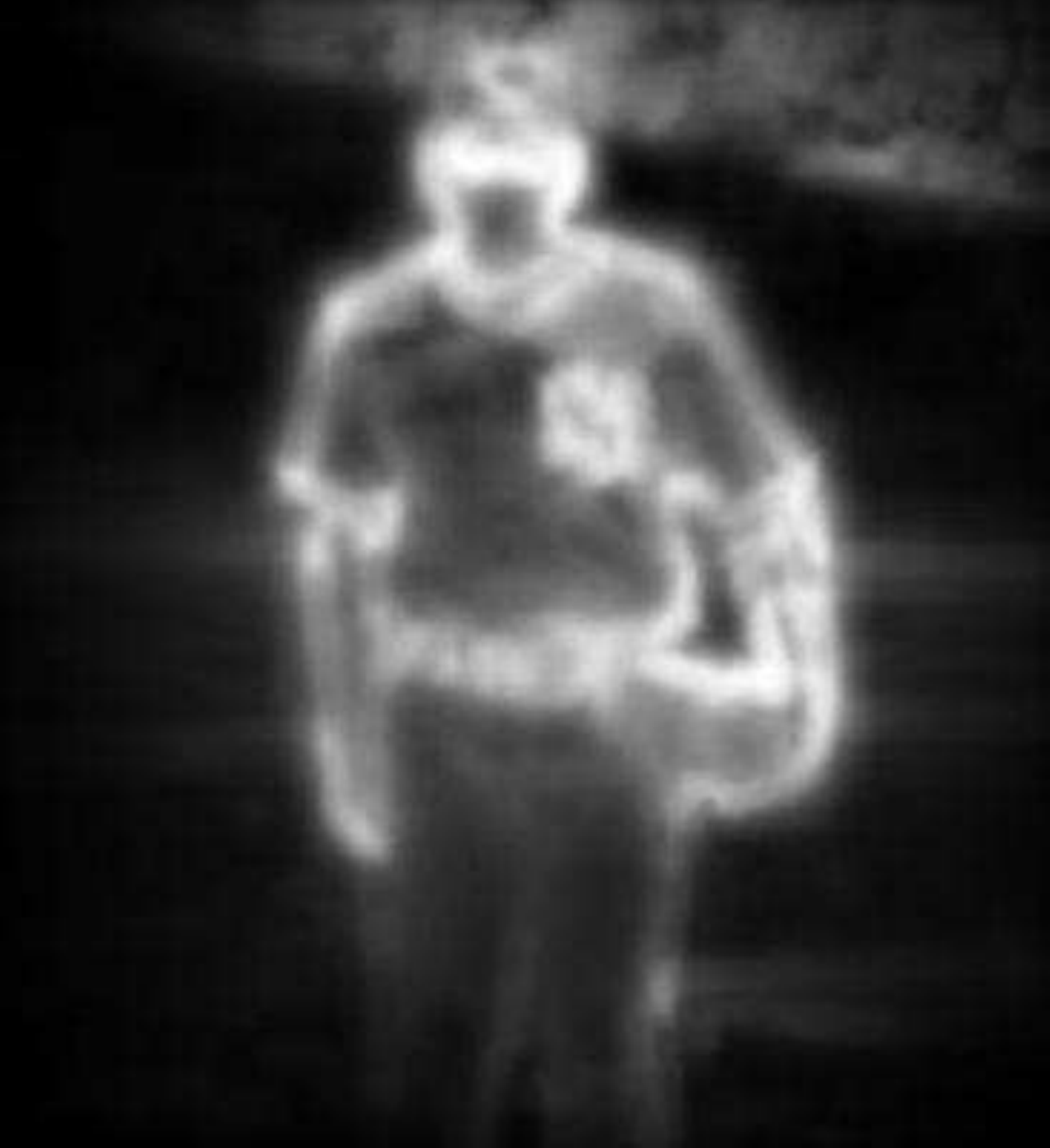}&
	\addExample{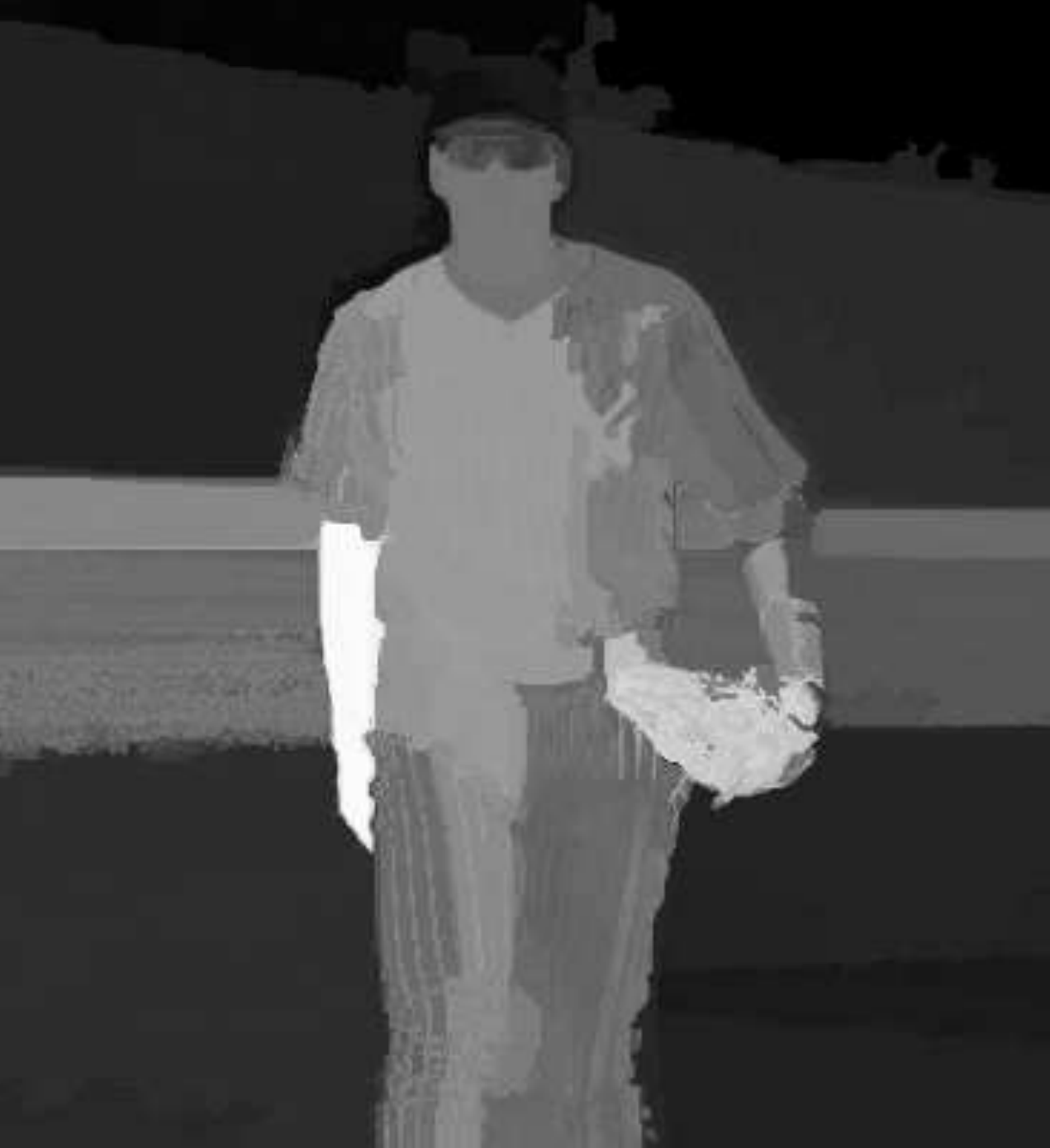}&
	\addExample{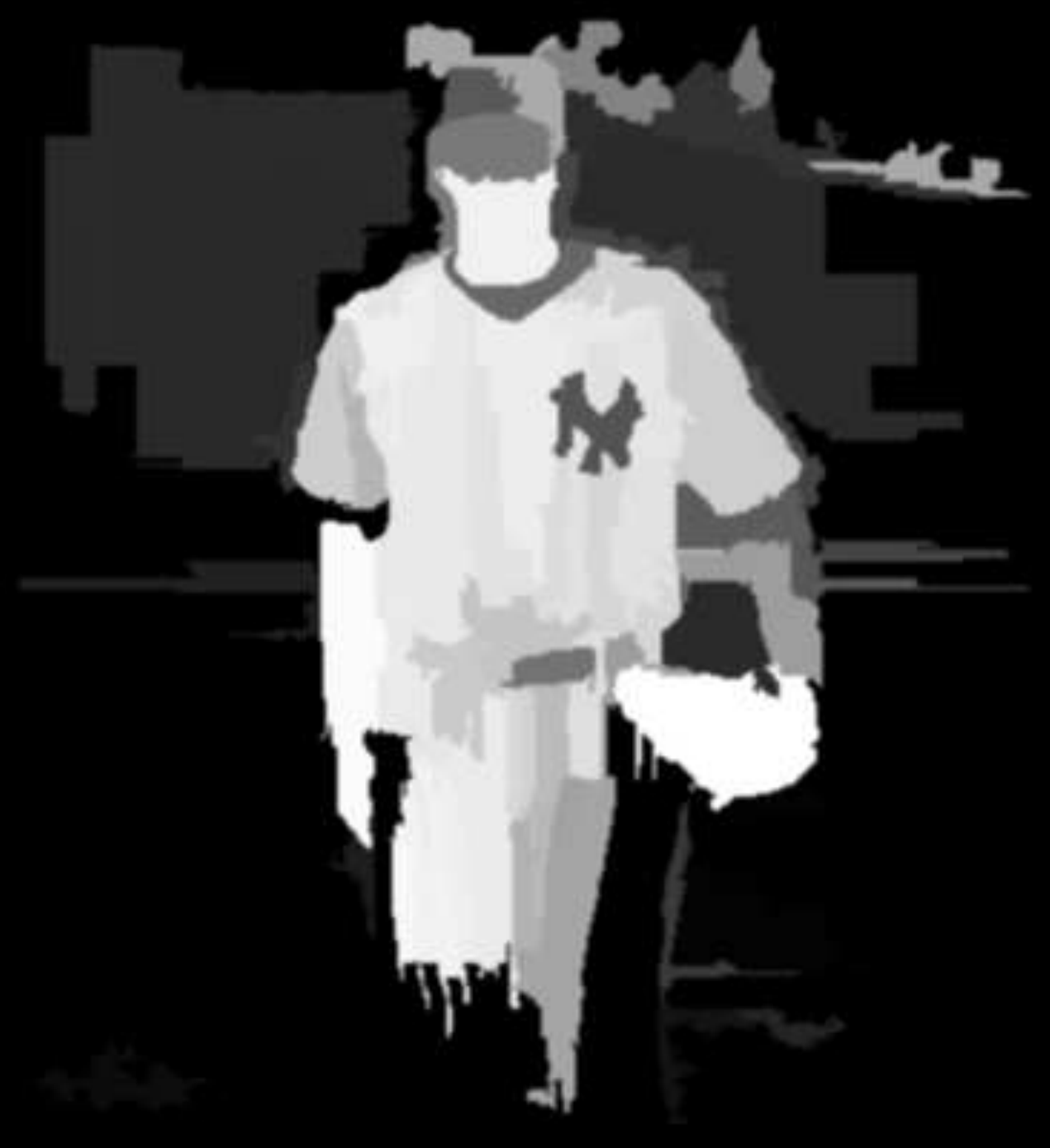}&
	\addExample{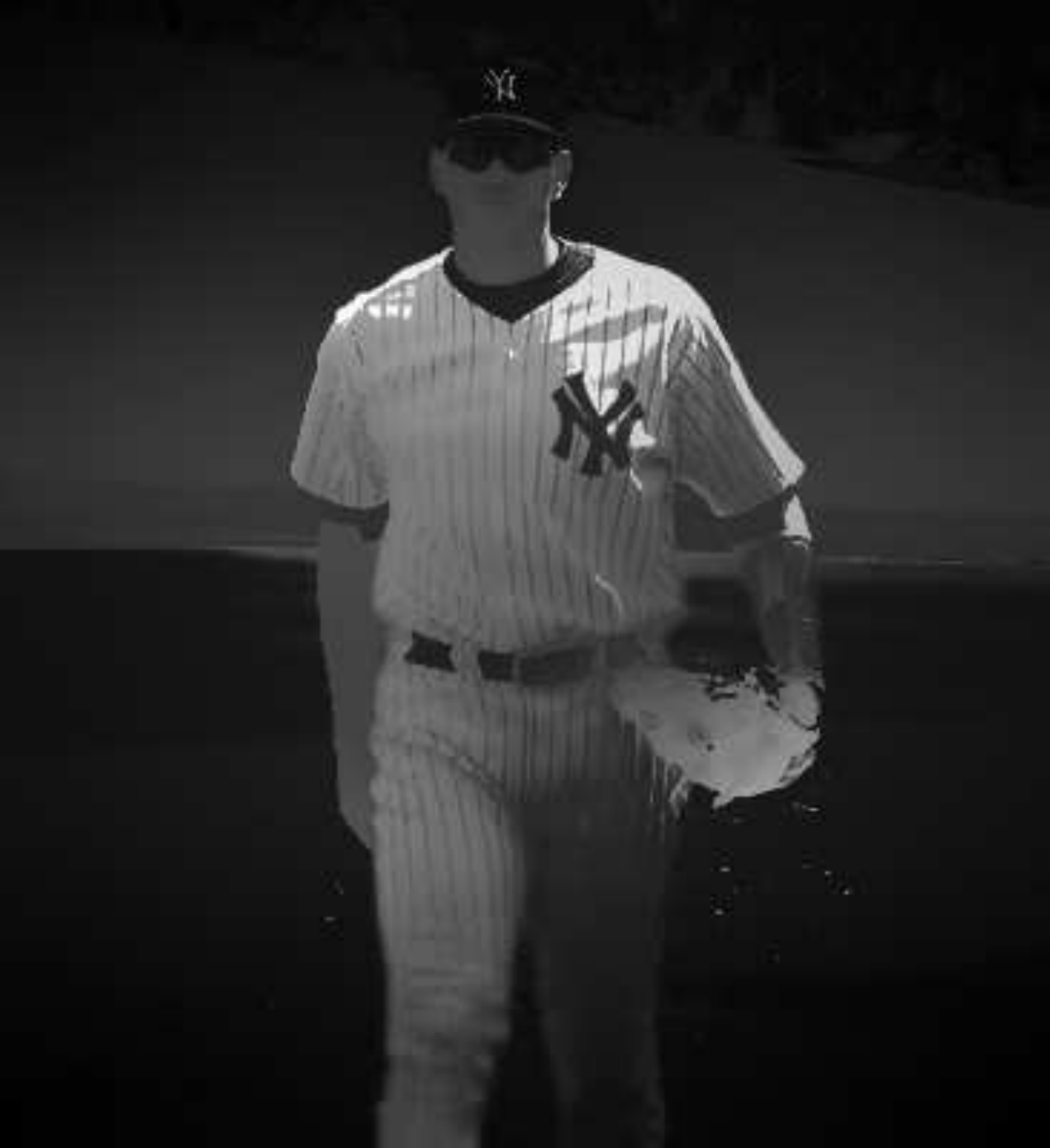}&
	\addExample{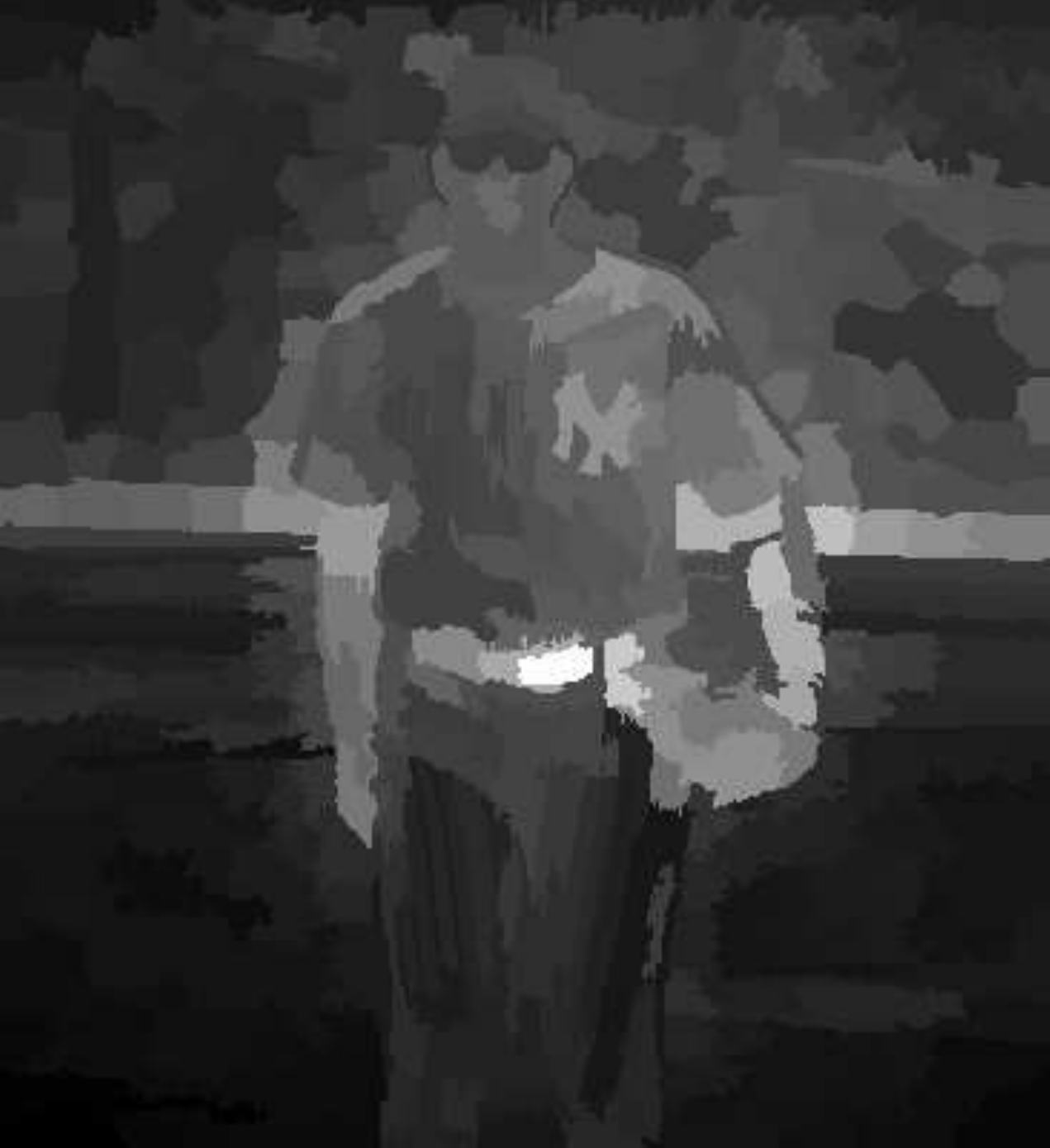}&
	\addExample{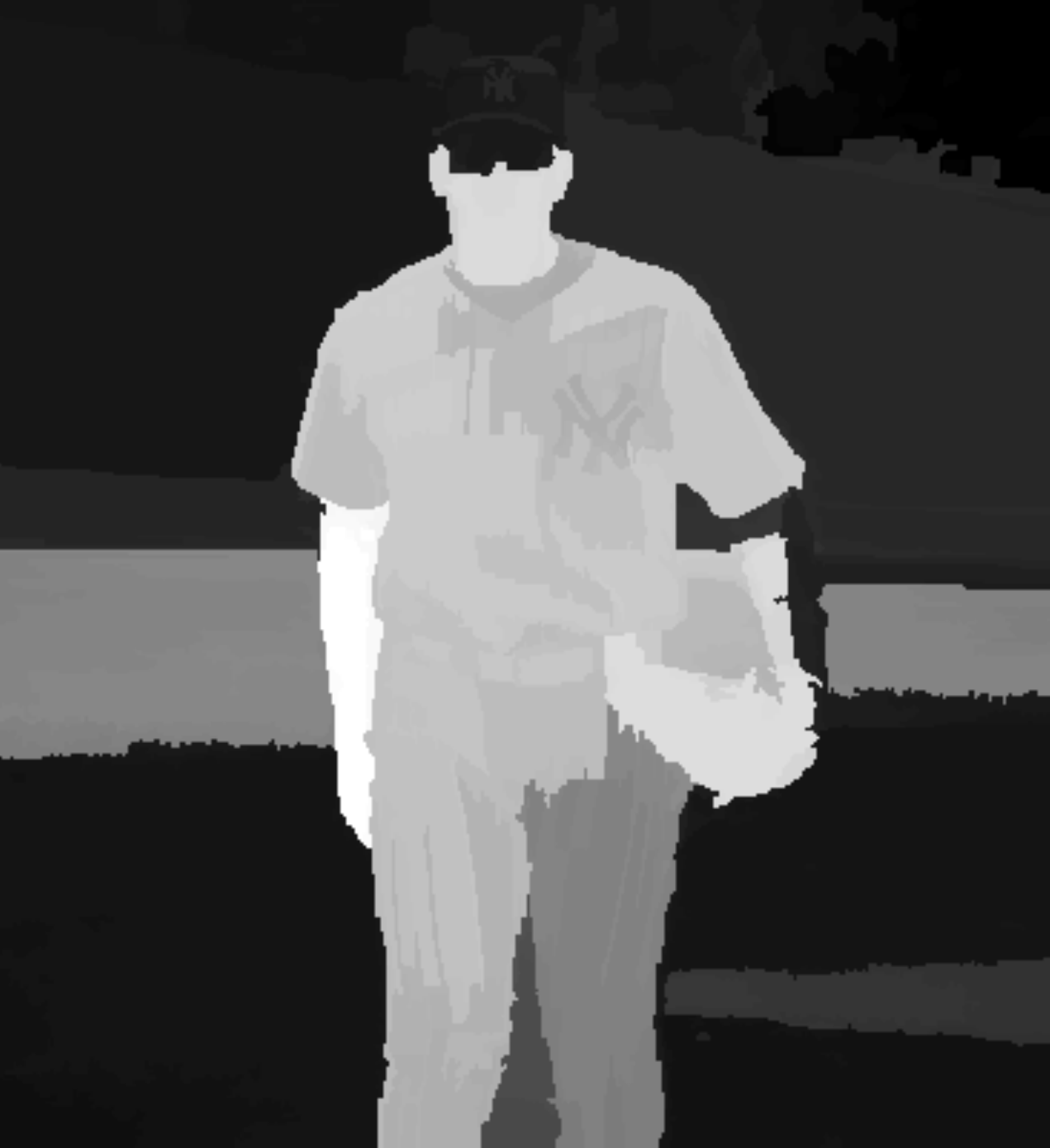}&
	\addExample{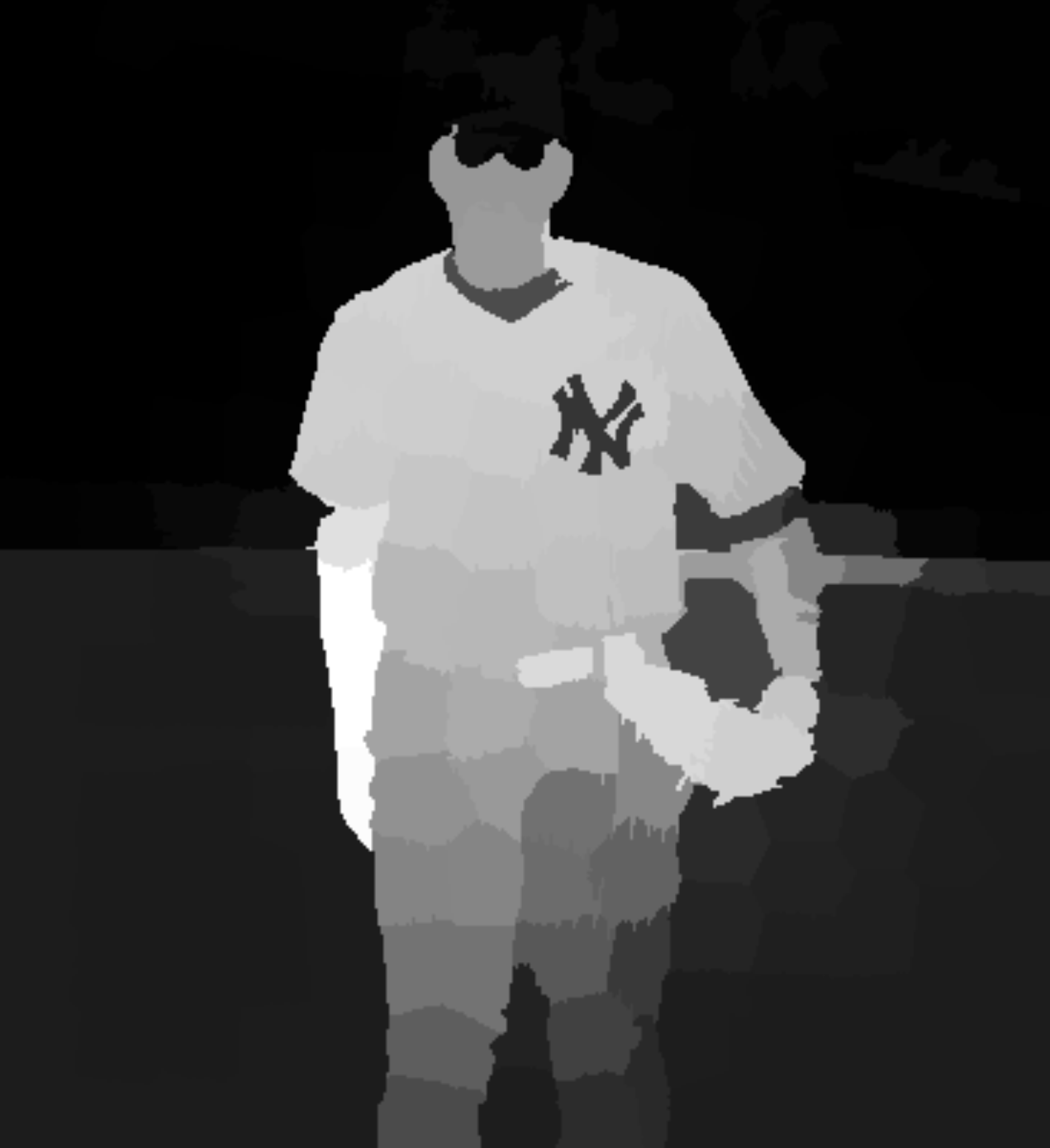}&
	\addExample{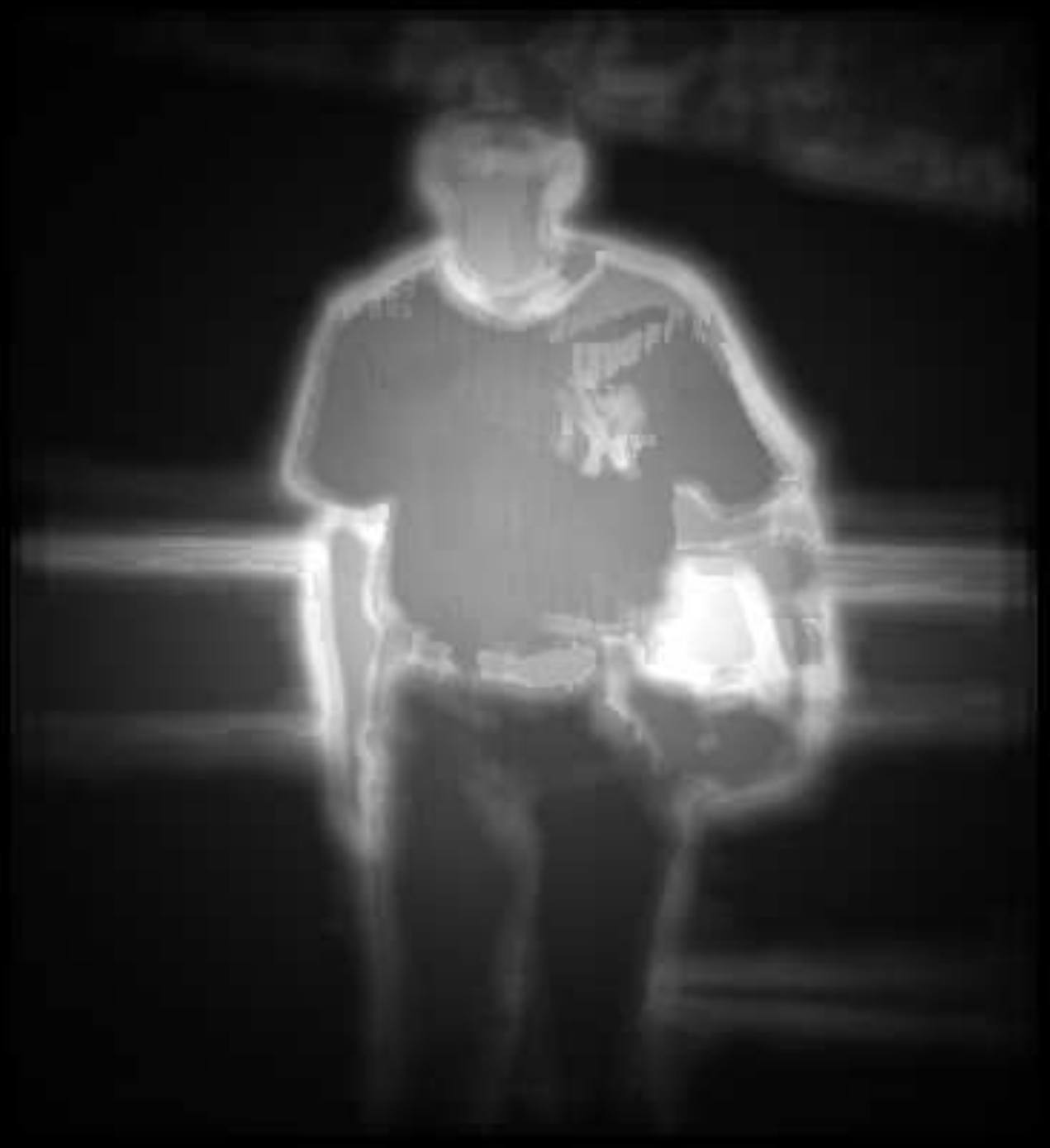}&
	\addExample{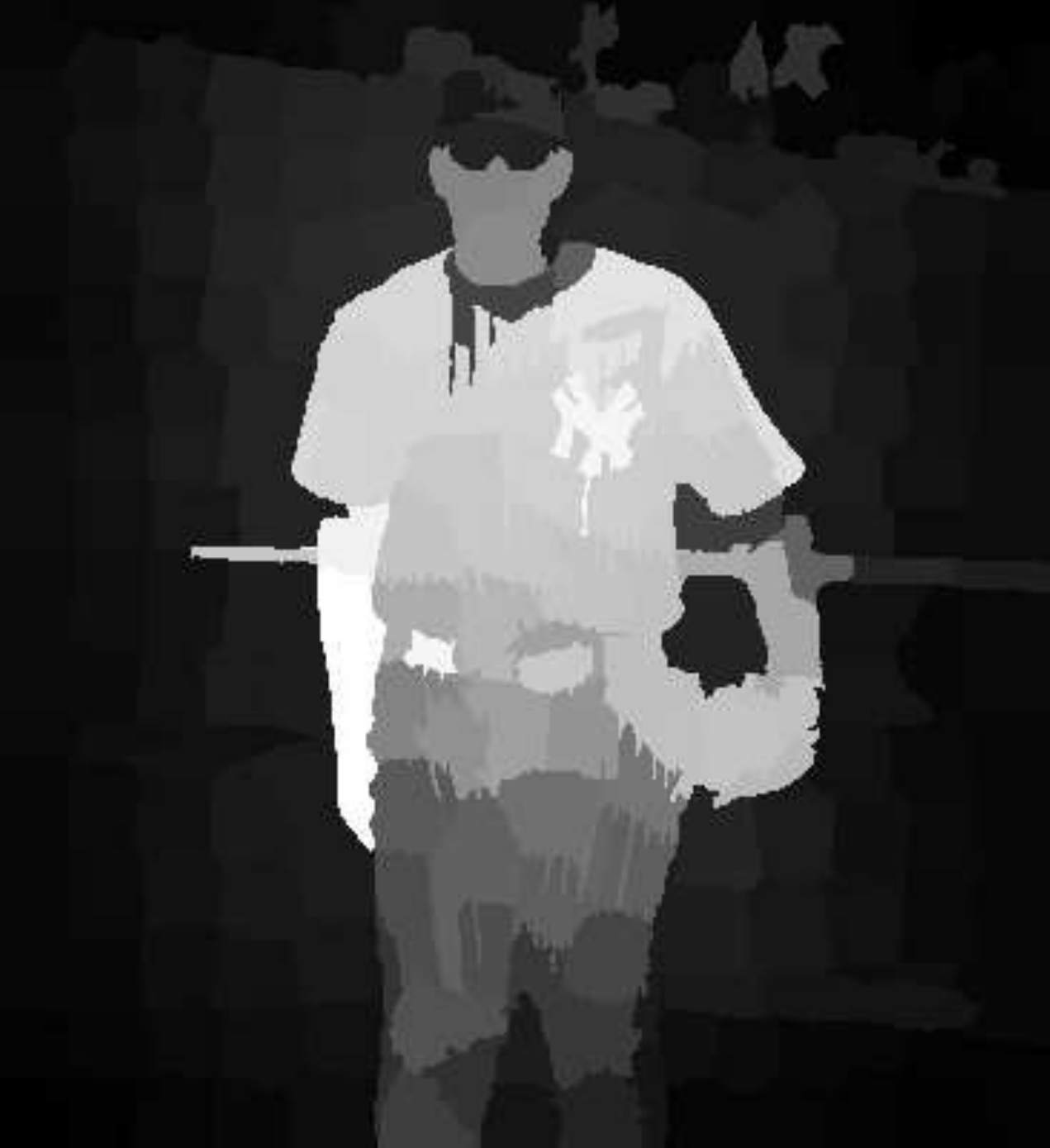}&
	\addExample{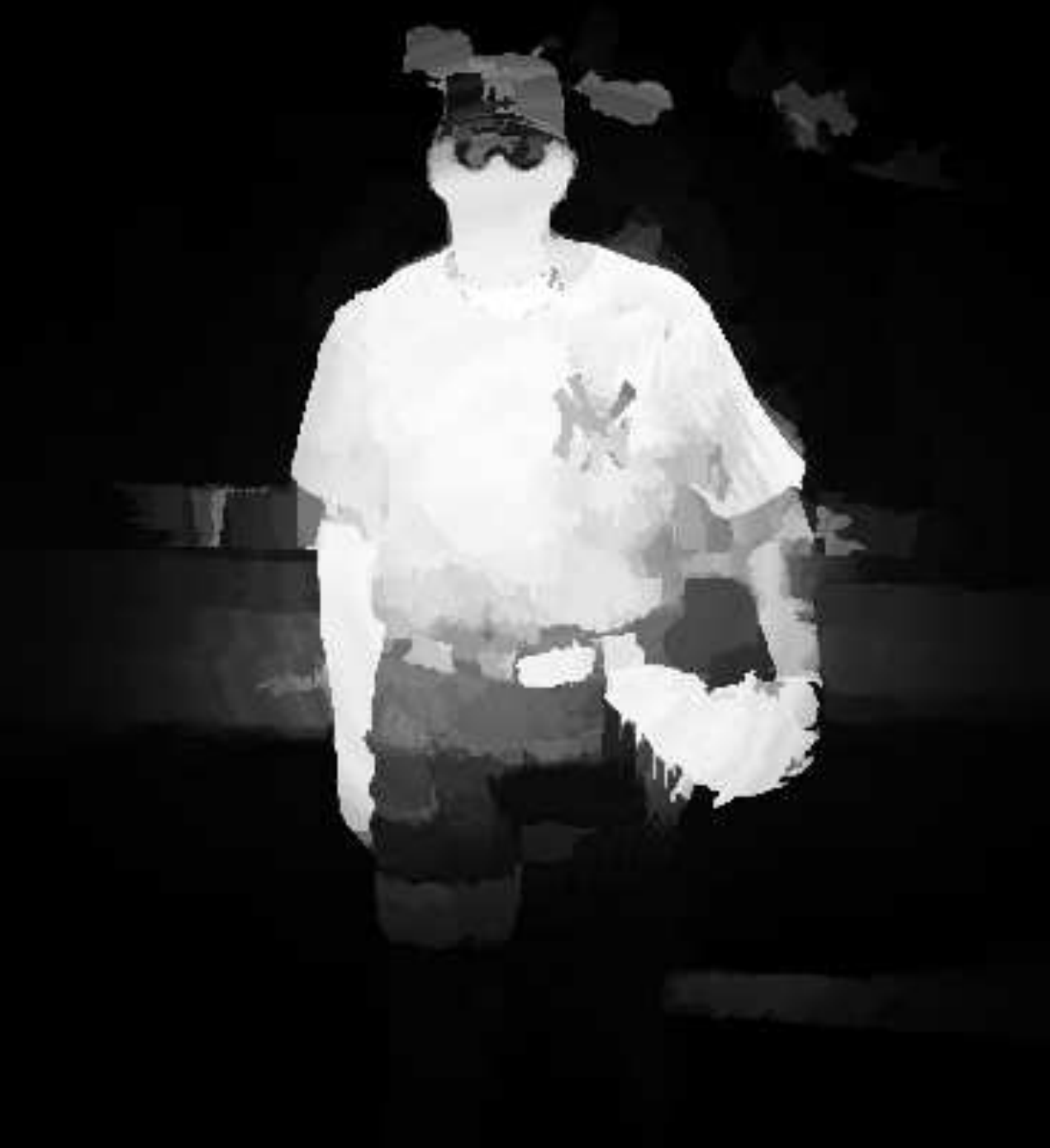}&
	\addExample{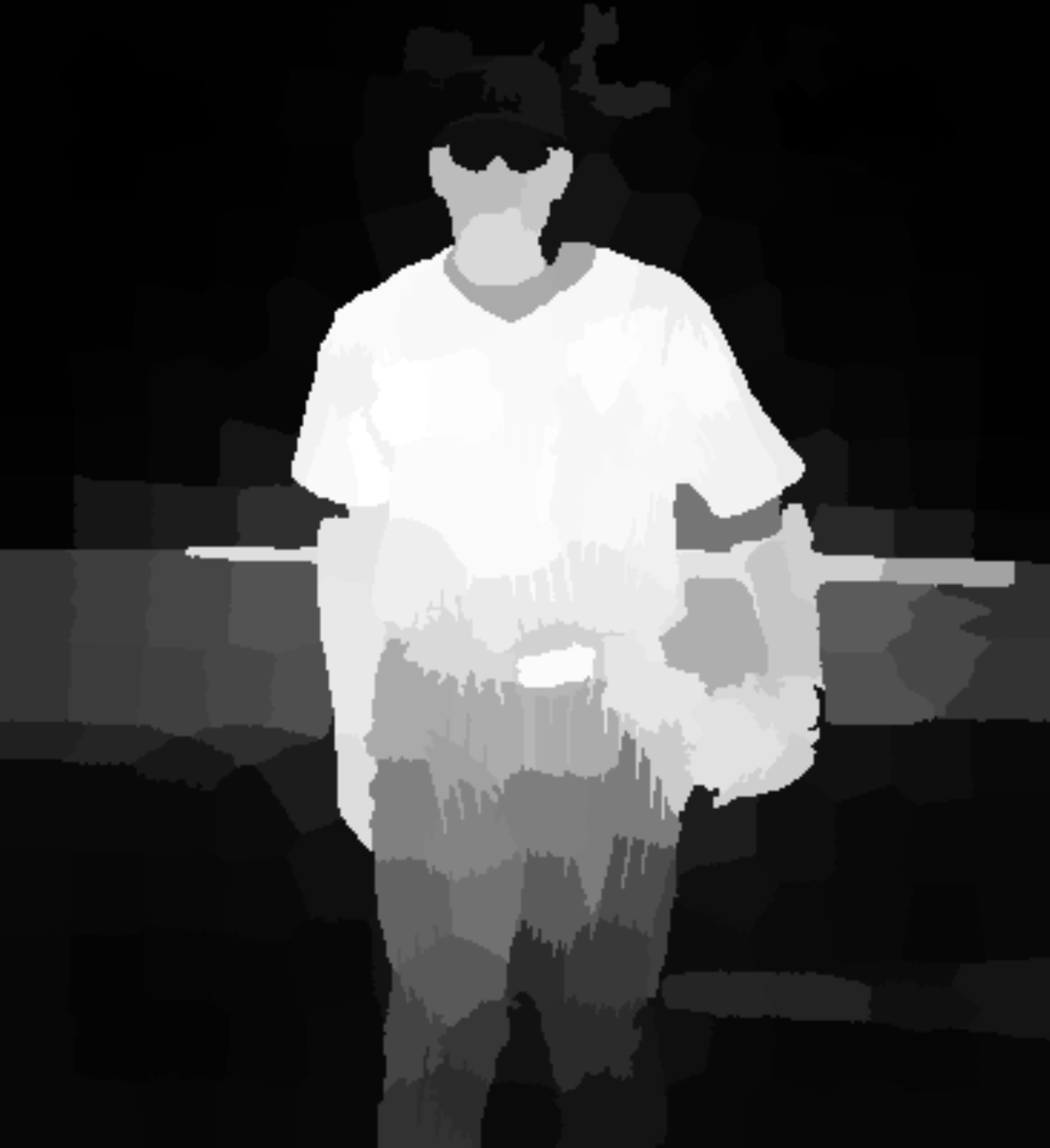}&
	\addExample{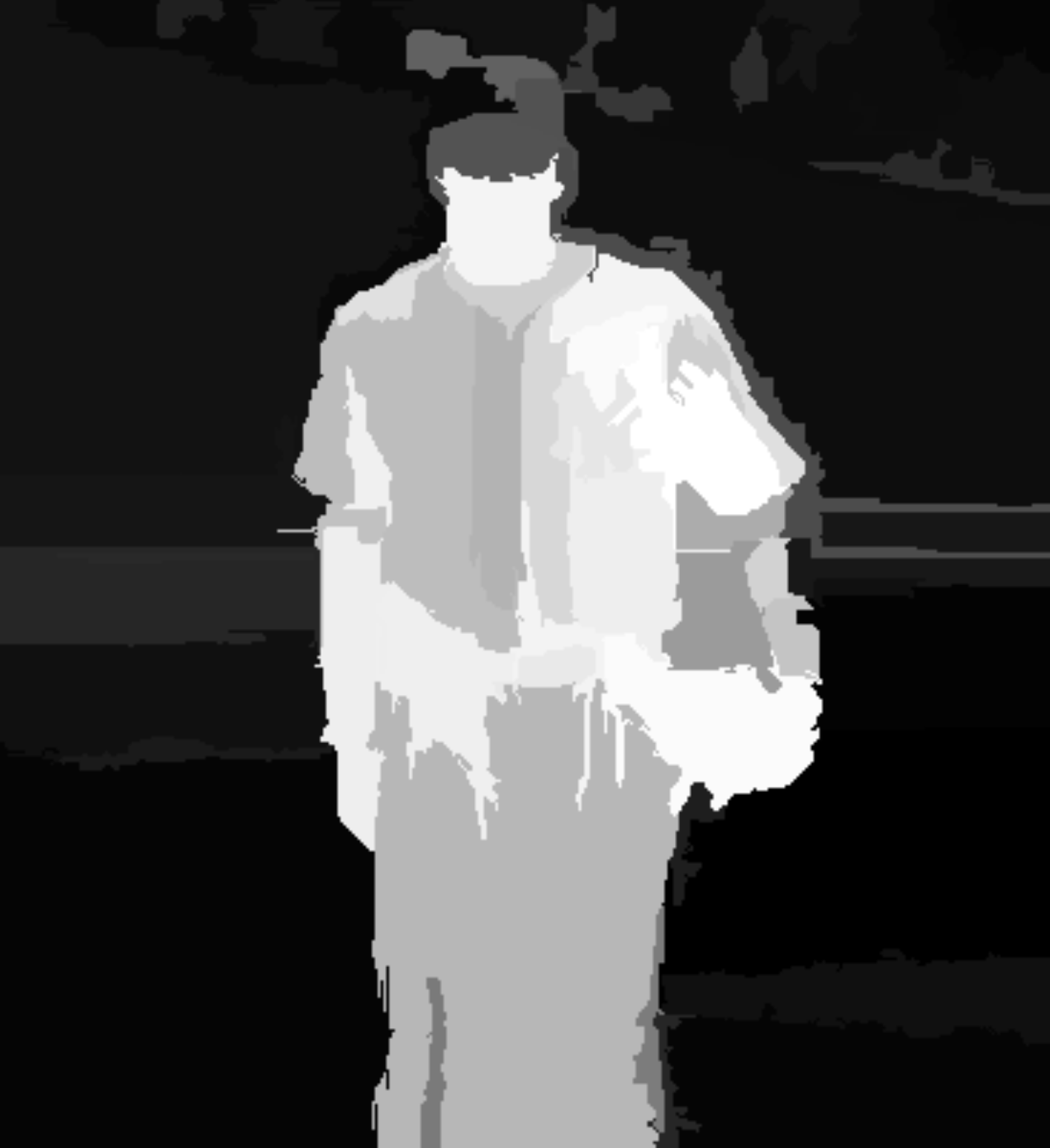}&
	\addExample{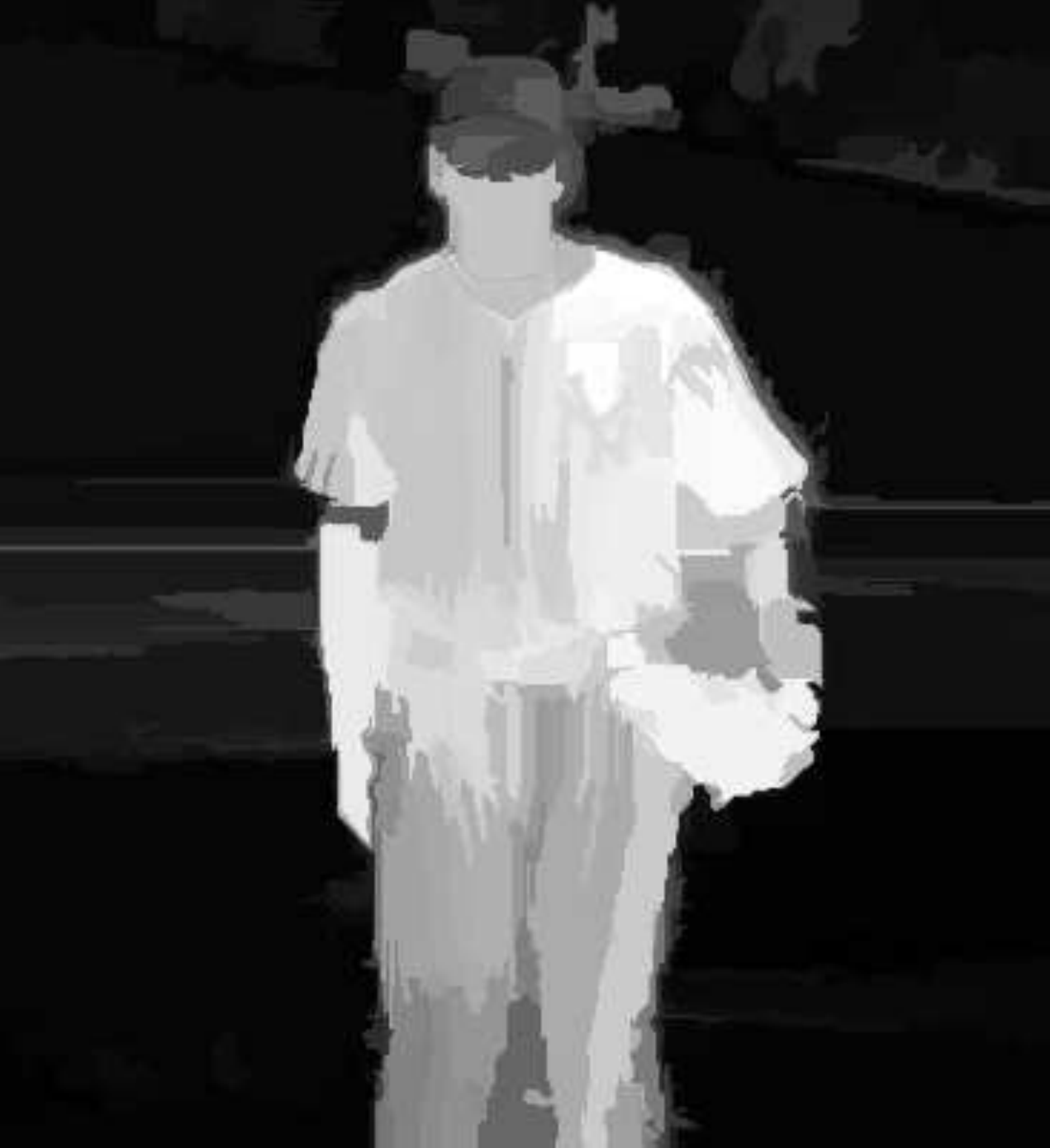}
	\\
    (a) input & (b) SVO & (c) CA & (d) CB  & (e) RC  & (f) SF & (g) LRK & (h) HS & 
    (i) GMR   & (j) PCA & (k) MC & (l) DSR & (m) \scriptsize{RBD} & (n) \scriptsize{DRFIs} & (o) DRFI \\
\end{tabular}
\caption{Visual comparison of the saliency maps. Our method (DRFI) consistently generates better saliency maps.}
% \vspace{-15pt}
\label{fig_smap_visual_comp}
\end{figure*} 

\myPara{Qualitative comparison}
We also provide the qualitative comparisons of different methods in \Figref{fig_smap_visual_comp}.
As can be seen, our approach (shown in \Figref{fig_smap_visual_comp} (n)(o)) can deal well
with the challenging cases
where the background is cluttered.
For example, in the first two rows,
other approaches may be distracted by the textures on the background while our method almost successfully highlights the entire salient object. It is also worth pointing out that our approach performs well when the object touches the image border, \eg, the first and last third rows in \Figref{fig_smap_visual_comp}, even though it violates the pseudo-background assumption. With the multi-\layer~enhancement, more appealing results can be achieved.

%\section{Discussion and Future Work}
%{\color{blue} The confusion matrix of the saliency classification will be given. And we will point out that it is easier to correctly classify the background than the salient object.
%
%For the future, we consider more features and speedup our implementation using the parallel computing and GPU acceleration.}

%\begin{table}[t]
%\setlength{\abovecaptionskip}{-50pt}
%\setlength{\belowcaptionskip}{-10pt}
%\centering
%\caption{\small Confusion matrix on the MSRA-B test set.}
%{\footnotesize
%\begin{tabular}{ccc}
%\hline
%\ & Salient Object & Background \\
%\hline
%Salient Object & {\bf 0.853} & 0.147 \\
%Background & 0.087 & {\bf 0.913} \\
%\hline
%\end{tabular}
%}
%\label{tab:confusion_matrix_msra}
%\end{table}

\subsection{Robustness Analysis}
As suggested by~\Figref{fig:feature_rank} and~\Figref{fig:featImpDataset}, the region backgroundness and property descriptors, especially the geometric properties, play important roles in our approach. In natural images, the pseudo-background assumption may not be held well. Additionally, the distributions of salient objects may be different from our training set. It is natural to doubt that whether our approach can still perform well on these challenging cases. To this end, we select 635 images from DUT-OMRON dataset (we call it DUT-OMRON*), where salient objects touch the image border and are far from the image center. Quantitative comparisons with state-of-the-art approaches are presented in~\tabref{tab:AUC}, ~\Figref{fig:smapCompPR}, and~\Figref{fig:smapCompROC}. Check our project website \href{http://jianghz.com/drfi}{jianghz.com/drfi} and the supplementary material for more details.
% , where the last column demonstrates the average annotation map of all these 635 images. We can observe strong off-the-center bias and border-touching among these images, which are suitable to verify the robustness of our approach.

% , by around $11.37\%$.
Not surprisingly, performances of all approaches decline. But our approach DRFI still significantly outperforms other methods in terms of PR curve, ROC curve and AUC scores. Even with a single~\layer, our approach DRFIs performs slightly better than others (ranked as the second best in terms of AUC scores). In specific, DRFIs and DRFI are better than the top-performing method by around $2.22\%$ and $1.36\%$ (compared with $2.20\%$ and $1.26\%$ on the whole DUT-OMRON data set) according to the AUC scores. 
% However, our approach still performs worse than RBD and DSR in terms of MAE scores.

\subsection{Efficiency}
Since the computation on each \layer~of the multiple segmentations is independent, we utilize the multi-thread technique to accelerate our C++ code. \tabref{tab:runtimes} summarizes the running time of different approaches, tested on the MSRA-B data set with a typical 400$\times$300 image using a PC with an Intel i5 CPU of 2.50GHz and 8GB memory. 8 threads are utilized for acceleration. As we can see, our approach can run as fast as most existing approaches. If equipped as a pre-processing step for an application, \eg, picture collage, our approach will not harm the user experiences.

For training, it takes around 24h with around 1.7 million training samples. As training each decision tree is also independent to each other, parallel computing techniques can also be utilized for acceleration.

\begin{figure}[t]
    \centering
    \scriptsize
    \begin{tabular}{|c|c|c|c|c|c|c|c|}
     \hline
        \textbf{Method}      & SVO & CA & CB & RC & SF & LRK & HS  \\
     \hline
        \textbf{Time(s)}     & 56.5 &   52.3  & 1.40 &   0.138  & 0.210  & 11.5 & 0.365  \\
    % \hline
        \textbf{Code}       & M+C &   M+C &   M+C &   C  &  C &   M+C &   EXE  \\
    \hline
    \hline
    \textbf{Method} & GMR & PCA & MC & DSR & RBD & DRFIs & DRFI* \\
    \hline
    \textbf{Time(s)} &  1.16 &  2.07 & 0.129  &  4.19 &   0.267  &  0.183 & 0.418 \\
    % \hline
    \textbf{Code} &  M+C &   M+C   & M+C   & M+C  &  M   & C  & C \\
    \hline
    \end{tabular}
    \caption{Comparison of running time. M indicates the code is written in MATLAB and EXE is corresponding to the executable. (*8 threads are used for acceleration.)
    }\label{tab:runtimes}
\end{figure} 

\section{Discussions and Future Work}
\label{sec:Conclusion}

\subsection{Unsupervised vs Supervised}
As data-driven approaches, especially supervised learning methods, dominate other fields of computer vision, it is somewhat surprising that the potential of supervised salient object detection is relatively under exploited. The main research efforts of salient object detection still concentrate on developing heuristic rules to combine saliency features. Note we are not saying that heuristic models are useless. We instead favor the supervised approaches for following two advantages. On one hand, supervised learning approaches can \emph{automatically} fuse different kinds of saliency features, which is valuable especially when facing high-dimensional feature vectors. It is nearly infeasible for humans, even domain experts, to design rules to integrate the 93 dimensional feature vector of this paper. For example, the sixth important feature $p_{12}$ (variance of the L* values of a region) seems to be rather obscure for salient object detection. Yet integration rules discovered from training samples indicate it is highly discriminative, more than the traditional regional contrast features.

On the other hand, data-driven approaches always own much better generalization ability than heuristic methods. In a recent survey of salient object detection~\cite{BorjiSI12b}, none of existing unsupervised algorithms can consistently outperforms others on all benchmark data sets since different pre-defined heuristic rules favor different settings of the data set (\eg, the number of objects, center bias, etc). Leveraging the large amount of training samples (nearly two milliion), our learned regional saliency regressor almost performs the best over all six benchmark data sets. Even though it is trained on a single data set, it performs better than others on challenging cases which are significantly different from the training set.

One potential reason that learning-based approaches is not favored for salient object detection might be related to the efficiency. Though a lot of training time is required to train a classifier, testing time is more likely to be the major concern of a system rather than the offline training time. Once trained, the classifier can be used as an off-shelf tool. In this paper, we demonstrate that a learning-based approach can run as fast as some heuristic methods.

\subsection{Limitations of Our Approach}
\begin{figure}[t]
    \centering
    \renewcommand{\arraystretch}{.9}
    \renewcommand{\tabcolsep}{.5mm}
    \begin{tabular}{ccc}
    	\includegraphics[height=0.11\textwidth,keepaspectratio]{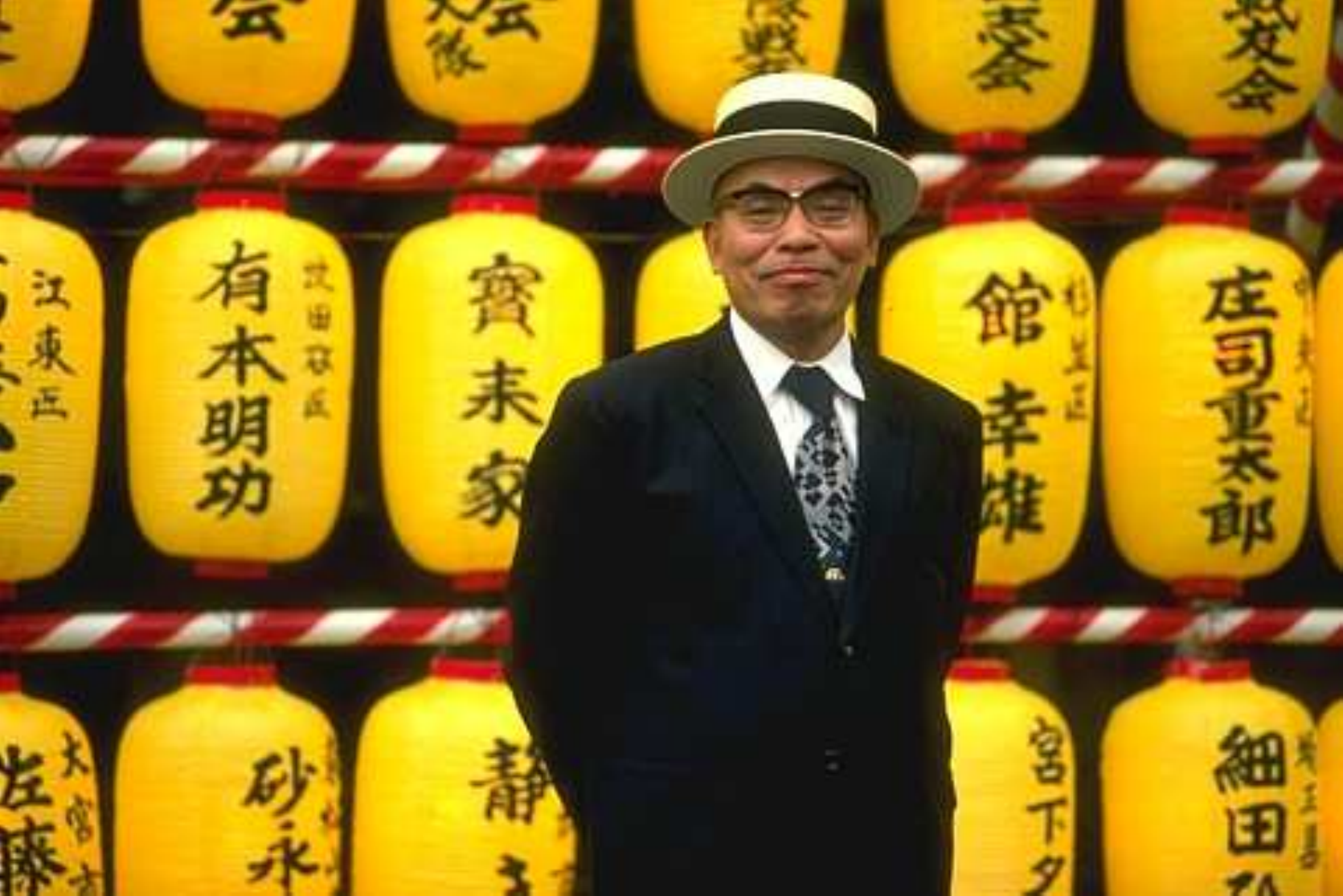}&
    	\includegraphics[height=0.11\textwidth,keepaspectratio]{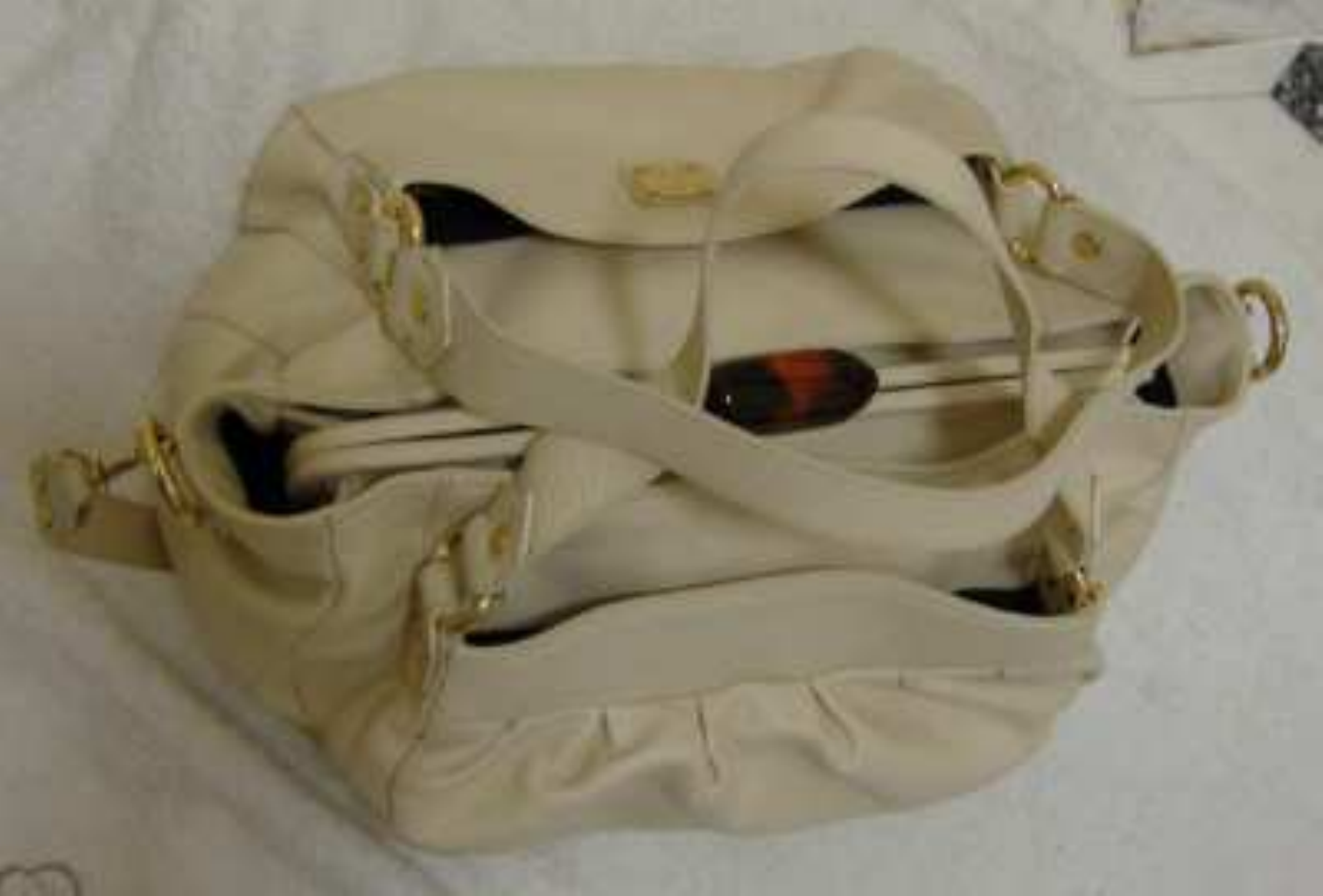}&
    	\includegraphics[height=0.11\textwidth,keepaspectratio]{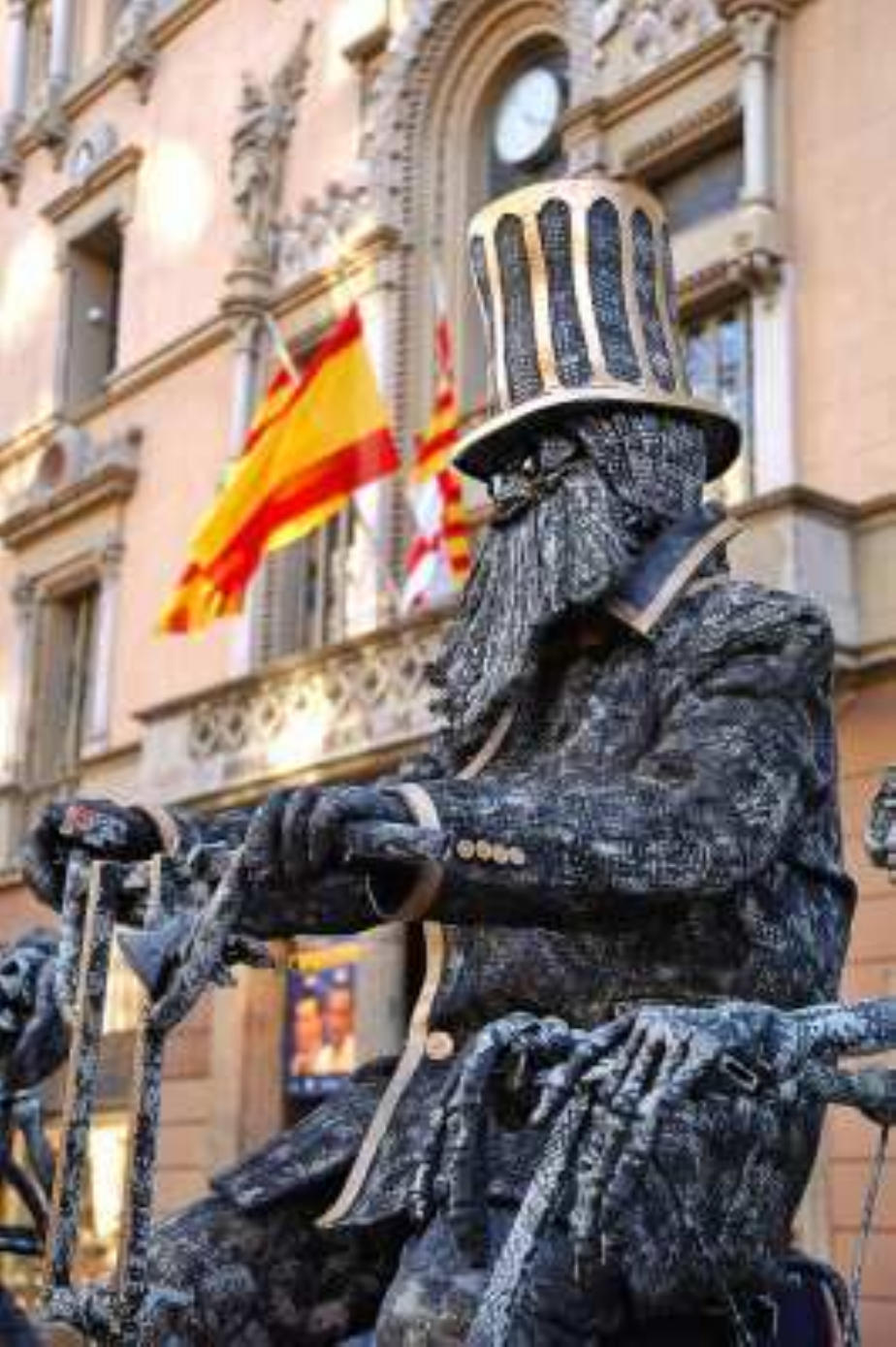}\\
    	\includegraphics[height=0.11\textwidth,keepaspectratio]{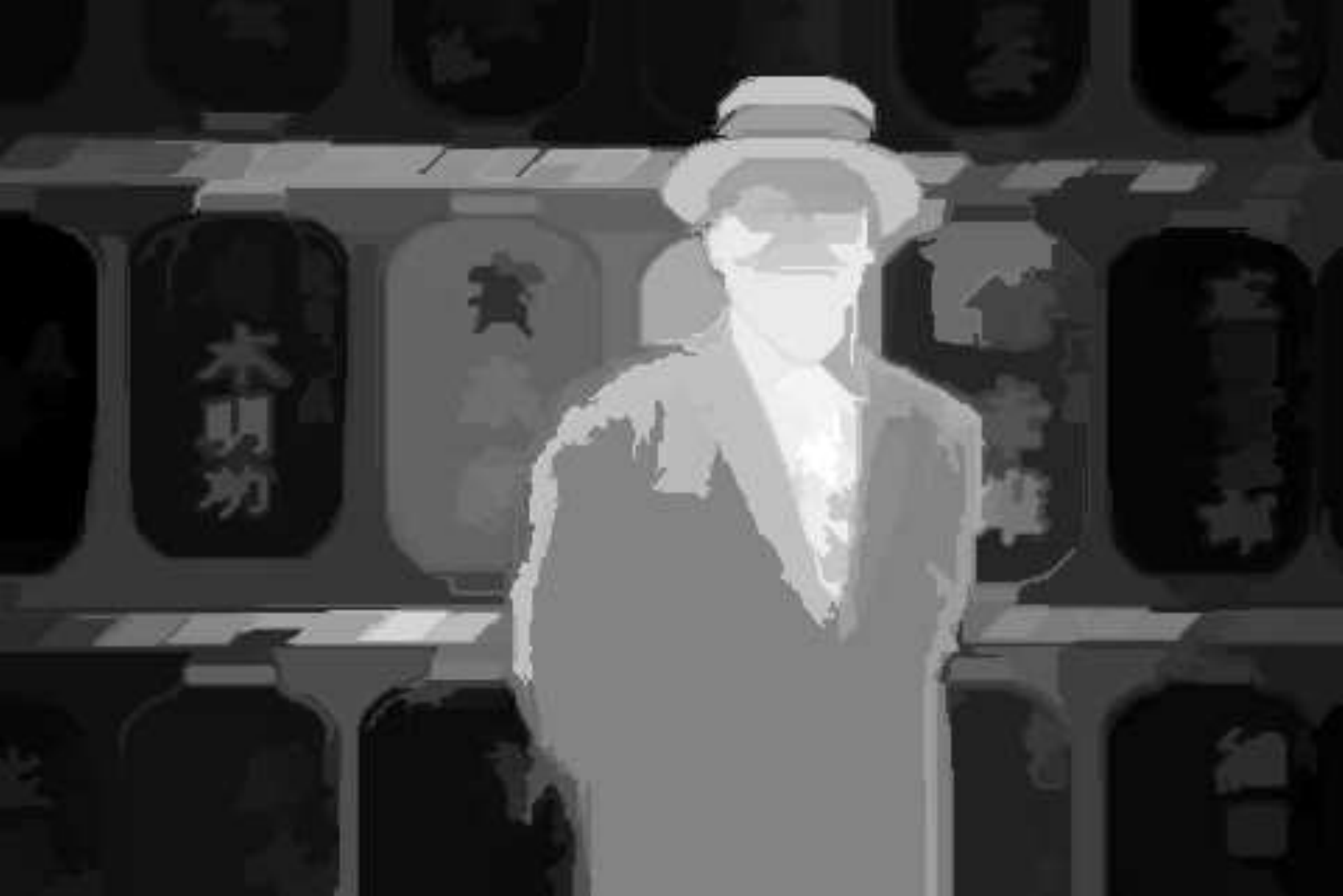}&
    	\includegraphics[height=0.11\textwidth,keepaspectratio]{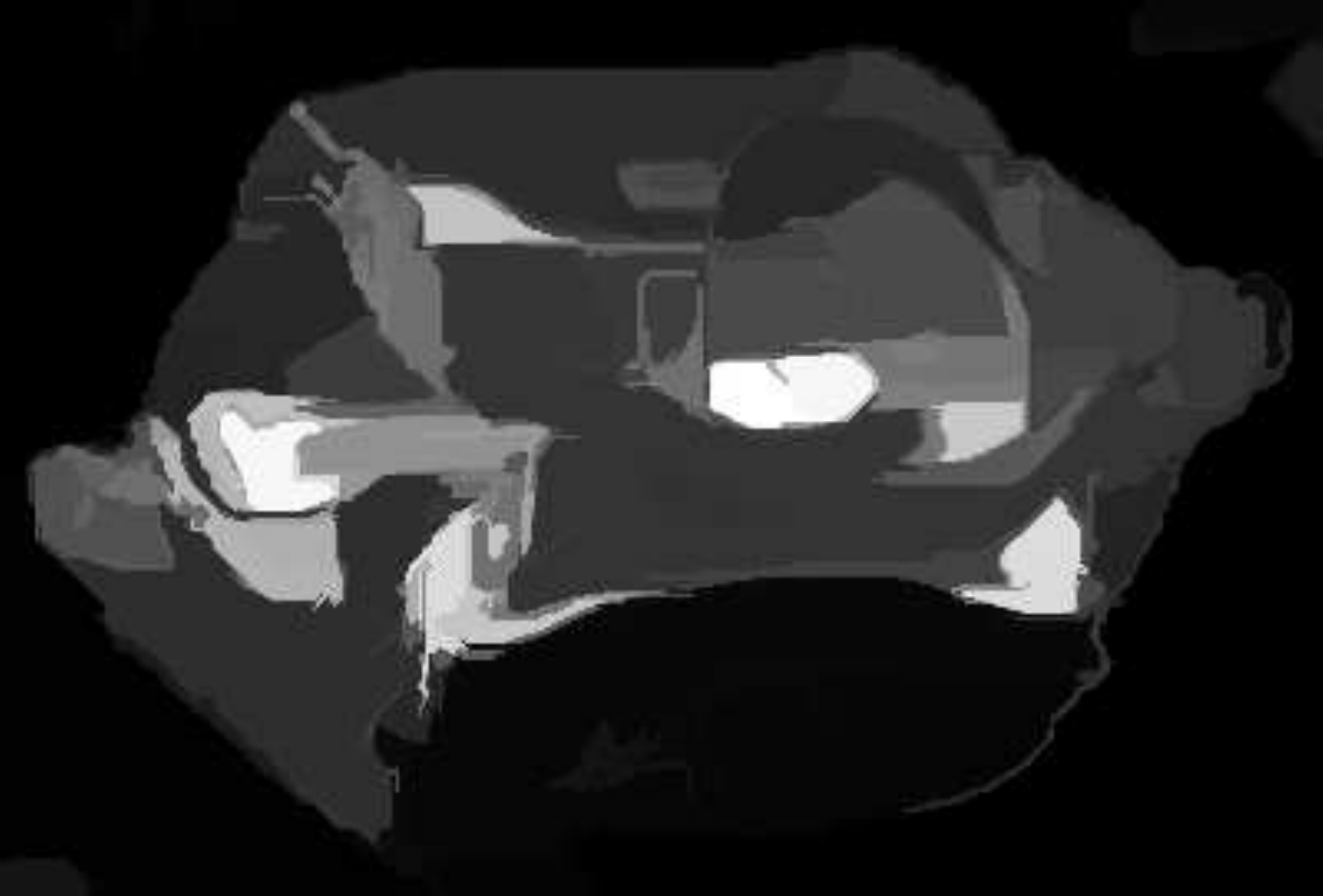}&
    	\includegraphics[height=0.11\textwidth,keepaspectratio]{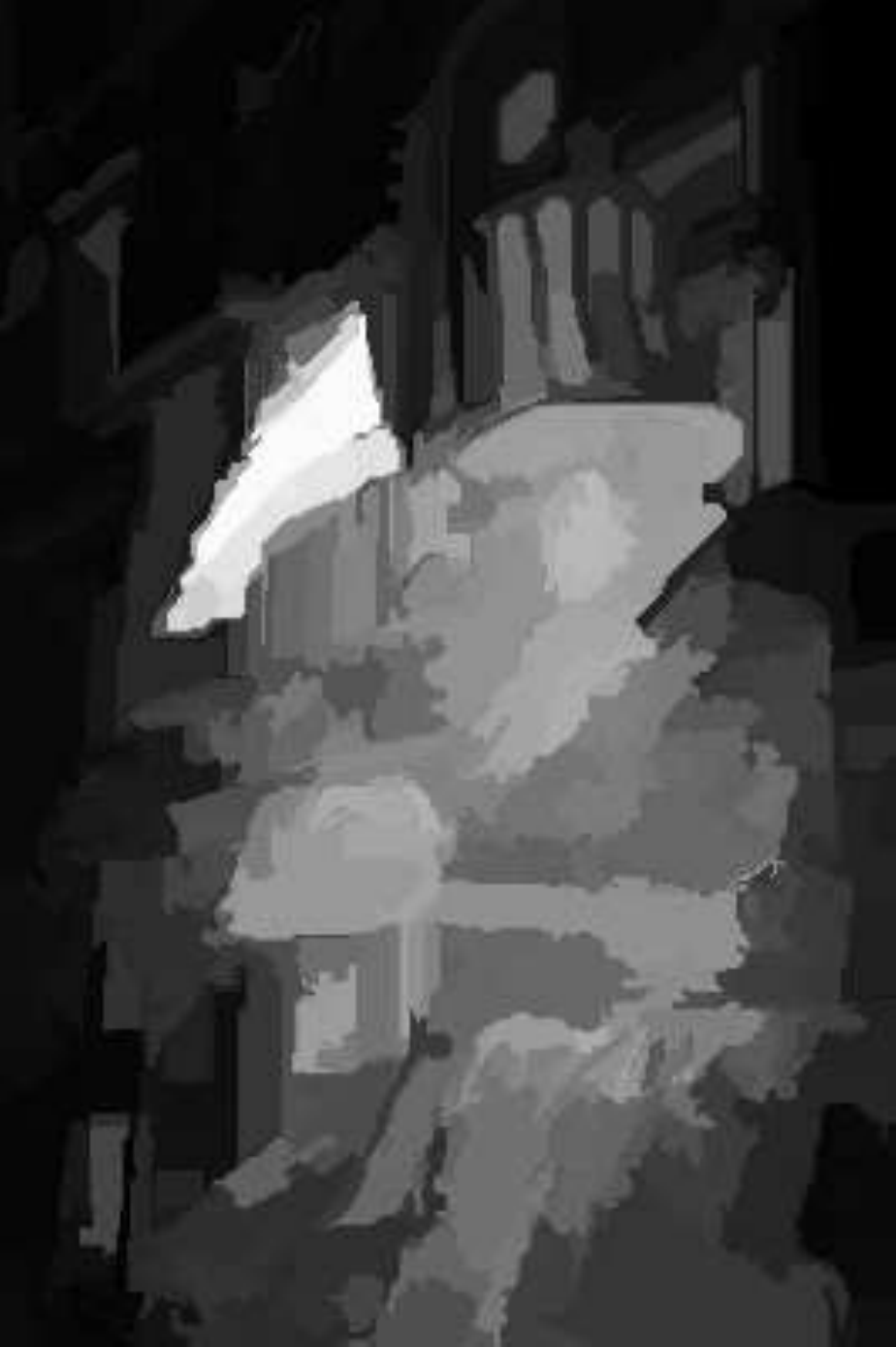}\\
    \end{tabular}
    \caption{Failure cases of our approach.}
    \label{fig:failure}
\end{figure}

Since our approach mainly consider the regional contrast and backgroudness features, it may fail on cluttered scenes. See~\Figref{fig:failure} for illustration. In the first column, high saliency values are assigned to the texture ares in the background as they are distinct in terms of either contrast or background features. The salient object in the second column have similar color with the background and occupies a large portion of the image, making it challenging to generate good detection result. For the third column, it is not fair to say that our approach completely fails. The flag is indeed salient. However, as the statue violates the pseudo-background assumption and occupying large portions as well, it is difficult to generate an appealing saliency map using our approach.

% the texture in the background make the contrast features less informative.

% and third columns,  Meanwhile, the pseudo-background assumption are violated. Thus the output saliency map is less appealing. 

\subsection{Conclusion and Future Work}
In this paper,
we address the salient object detection problem
using a discriminative regional feature integration approach.
The success of our approach
stems from the utilization of supervised learning algorithm: we learn a \rforest~regressor to automatically
integrate a high-dimensional regional feature descriptor to
predict the saliency score and automatically discover the most discriminative features.
Experimental results validate that compared with traditional approaches, which heuristically compute saliency maps from different types of features
% and combine them to get the saliency map, our approach is more superior and robust on different benchmark data sets. Though far from perfect on challenging cases, %where the salient objects may touch the image border or are far from the image center, 
% our approach produces more appealing salient object detection results than state-of-the-art algorithms.

Our approach is closely related to the image labeling method~\cite{DBLP:journals/ijcv/HoiemEH07}. The goal is to assign predefined labels (geometric category in~\cite{DBLP:journals/ijcv/HoiemEH07} and object or background in salient object detection) to the pixels. It needs further study to investigate the connection between image labeling and salient object, and if the two problems are essentially equivalent. Utilizing the data-driven image labeling approaches for salient object detection is also worth exploring in the future.

% In essence, our approach is closely related to the image labeling method~\cite{DBLP:journals/ijcv/HoiemEH07}. The goal is to assign pre-defined labels (geometric category in~\cite{DBLP:journals/ijcv/HoiemEH07} and object or background in salient object detection) to the pixels. A slight difference is that the ``soft'' value of pixels (\ie, the saliency map denoting the probability of being salient) is preferred in addition to the ``hard'' labeling for salient object detection. The relationship between image labeling and salient object detection is an open problem. Utilizing the data-driven image labeling approaches for salient object detection is also worth exploiting in the future.

Additionally, there exist some obvious directions to further improve our approach.
\begin{itemize}
\item Incorporating more saliency features. In this paper, we consider only contrast, backgroundness, and generic property features of a region. By considering more saliency features, better detection results can be expected. For example, background connectivity prior~\cite{zhu2014saliency} can be incorporated to relax the pseudo-background assumption. Additionally, spatial distribution prior~\cite{LiuYSWZTS11, PerazziKPH12}, focusness prior~\cite{JiangLYP13UFO}, diverse density score~\cite{jia2013category} based on generic objectness, and graph-based manifold ranking score~\cite{YangZLRY2013} can also be integrated.

\item Better fusion strategy. We simply investigate the linear fusion of saliency maps with any post optimization step. As a future work, we can utilize the optimization step of other approaches to enhance the performance. For example, we can run saliency detection on hierarchical detections and fuse them as suggested in~\cite{yan2013hierarchical}. The optimization method proposed in~\cite{zhu2014saliency} is also applicable.

\item Integrating more cues. A recent trend on salient object detection is to integrate more cues in addition to traditional RGB data. Our approach is natural to be extedned to consider cues such as depth on RGB-D input, temporal consistency on video sequences, and saliency co-occurrence for co-salient object detection.

% \item Applications. Saliency maps generated by our approach are potentially useful for many applications
\end{itemize}

\section*{Acknowledgements} % {\small \noindent\textbf{Acknowledgements}\
This work was supported in part by the National
Basic Research Program of China under Grant No. 2015CB351703 and 2012CB316400, and the National Natural Science
Foundation of China under Grant No. 91120006.

% Can use something like this to put references on a page
% by themselves when using endfloat and the captionsoff option.
\ifCLASSOPTIONcaptionsoff
  \newpage
\fi

\bibliographystyle{IEEEtran}
\bibliography{REF}

\end{document}